"\pdfoutput=1"
\documentclass[12pt,a4paper]{book}

\usepackage{packages}
\usepackage{tocbibind}
\usepackage{thesis_layout}

\usepackage{bm}
\usepackage{subfigure}
\usepackage{epsfig}
\usepackage{epstopdf}
\usepackage{float}
\usepackage{booktabs}

\renewcommand{\baselinestretch}{1.5}

\begin{document}
\bibliographystyle{IEEEtranS}


\begin{titlepage}
\begin{center}

\mbox{}\vfill {\Large\bf\sc Collision free autonomous navigation and formation building for non-holonomic ground robots}

\vspace{20mm}



{\large {\bf  Chao Wang}}

\vspace{50mm}














\end{center}

\vfill

\end{titlepage}

\pagenumbering{roman}

\begin{center}
 Abstract
\end{center}

The primary objective of a safe navigation algorithm is to guide the object from its current position to the target position while avoiding any collision with the en-route obstacles, and the appropriate obstacle avoidance strategies are the key factors to ensure safe navigation tasks in dynamic environments. The basic requirement for an appropriate obstacle avoidance strategy is to sense or detect obstacles and make proper decisions when the obstacles are nearby. By fulfilling the basic requirement, the more advanced obstacle algorithms should have other additional features. 
\par
In this report, three different obstacle avoidance strategies for safe navigation in dynamic environments have been presented. All of them are applicable in the non-holonomic systems by which motions of many objects can be described. The biologically-inspired navigation algorithm (BINA) is efficient in terms of avoidance time, it is also simple and easy to compute. The equidistant based navigation algorithm (ENA) is able to achieve navigation task with in uncertain dynamic environments, and it is suitable for a variety of situations due to its flexibility. The navigation algorithm algorithm based on an integrated environment representation (NAIER) allows the object to seek a safe path through obstacles in unknown dynamic environment in a human-like fashion and it is very efficient in numerous particular scenarios where other algorithm are found inefficient or even impossible to solve. The performances and features of the proposed navigation algorithms are confirmed by extensive simulation results and experiments with a real non-holonomic mobile robot. Furthermore, the performance of these algorithms are compared with each other in various aspects.
\par
The algorithms have been implemented on two real control systems: intelligent wheelchair and robotic hospital bed. The performance of the proposed algorithms with SAM and Flexbed demonstrate their capabilities to achieve navigation tasks in complicated real time scenarios. The proposed algorithms are easy to be implemented in real time and costly efficient.
\par 
An extra study on networked multi-robots formation building algorithm is presented in this paper. A constructive and easy-to-implement decentralised control is proposed for a formation building of a group of random positioned objects. Furthermore, the problem of formation building with anonymous objects is addressed. This randomised decentralised navigation algorithm achieves the convergence to a desired configuration with probability 1.

\tableofcontents
\listoffigures
\listoftables
\newpage
\pagenumbering{arabic}

\pdfoutput=1
\chapter{Introduction} \label{intro}

	A safe navigation is defined as the process of ascertaining the object's current position and planning its route from this current position to the target position on the premise of the safety of the object. An appropriate obstacle avoidance strategy is the key factor to ensure a safe navigation process. Numerous existing researches are focused on the development of navigation algorithms in static environments (see e.g. \cite{SMM10,MJ04,JGO1996,KY91}), which have limited applications in many real world scenarios where the obstacle are dynamically moving. The problem of safe navigation in dynamic environments poses a serious challenge in the field of mobile robotics. This challenge is further extended when the informations about the dynamic obstacles are scarce, especially some basic prior informations which are assumed to be available in many researches. These prior informations include the complete map of the environments, the position and the orientation of the obstacles in the map, the nature of the obstacles (whether the shape of the obstacles are consistent or deforming over time) and the motion of the obstacles (whether the obstacle is moving with a constant or time-varying velocity) etc. 
\par
	The basic requirement for an appropriate obstacle avoidance strategy is to sense or detect obstacles and make proper decisions when the obstacles are nearby. By fulfilling the basic requirement, the more advanced obstacle algorithms should have one or more additional features such as: obstacle avoidance with minimum effort, optimum obstacle avoidance time, avoidance with limited information, and computational efficiency. Although most of the existing obstacle avoidance algorithms achieve the basic requirement, few of them have the additional features necessary to be considered as more advanced algorithms.
\par
	In this report, three advanced obstacle avoidance strategies for safe navigation in dynamic environments have been presented. The proposed navigation algorithms are based on controller switching.  Mathematical analysis of such switched controller systems including global stability analysis canbe found in~\cite{MAS00,SAV02,SAV99,SEE99}. All of the proposed algorithms are applicable for the non-holonomic unicycle mobile systems with bounded speed and angular velocity. It is well-known that the motion of many wheeled robots, missiles and unmanned aerial vehicles can be described by this model; see e.g. \cite{TS10,MTS11,MTS11ronly} and references therein. Each of the proposed strategies has more than one additional feature besides the basic requirement for an appropriate obstacle avoidance strategy. 

\par
The detailed description of the proposed navigation algorithms are provided. Their features  are shown in extensive computer simulations. The implementation of the proposed navigation algorithms on different platforms, including ActivMedia Pioneer 3-DX mobile robot, a electric-powered wheelchair and a hospital bed control system, are presented to demonstrate their applicability in a wide range of real world scenarios. Furthermore, the performance of three obstacle avoidance strategies are compared with each other across various scenarios, which gives the insight to the merits and advantages of each of the algorithm. 
\par
	The problem of networked multi-robots formation building, another interesting challenge in the field of navigation, is also investigated. A constructive and easy-to-implement decentralised navigation algorithm is proposed for a group of randomly positioned objects. Furthermore, we present a  randomised decentralised navigation algorithm that achieves the convergence to a desired configuration with probability of 1. Its performance is confirmed by the simulations results and the implementation on the ActivMedia Pionner 3-Dx mobile robots.
\par

	The following section presents a literature review on the relevant topics in this report.
\par

	\section {Obstacle Avoidance Strategies for Safe Navigation Algorithm}

	The concept of safe navigation has been reflected in different aspects of our life, for example, pedestrians avoid colliding each other in the street, vehicles follow traffic rules in the driving lanes, animals use simple motion control rules that result in remarkable intelligent behaviors. The implementation of navigation algorithms can also be found various applications such as agricultural applications \cite{MASH10}, GPS navigation systems \cite{PS02}, auto-pilot function \cite{SJ08}, missile guidance \cite{MIR04, MIR06}, robots designed for specific purposes (rescue \cite{JJS97}, multiple robot navigation~\cite{HMA12} and explore\cite{Z92}, etc).   Therefore, it is hardly surprising that safe navigation in dynamic environments has become one of the most interesting and attractive  research topics, and has already generated a large number of successful results.
\par
	The existing navigation approaches fall into two main categories: global and local (reactive) navigation approaches.
\par
	Global navigation algorithms \cite{S00, KR97, KRR98, LA92,VSEP96,BA05} assume that a complete set of priori information about the environment, i.e. "map", is known, so that optimal theoretical  solutions based on different policies can be computed. Within this framework, several techniques (surveyed in e.g., \cite{LaLaSh05,KuVu06}) have been developed for dynamic scenes, including nonholonomic planners \cite{QuWaPl40} and state-time space \cite{Fraich99,ReSh94} approaches. One major problem of the global navigation algorithms is that they are computationally expensive which makes them inapplicable in many real time implementations. Also, their performance can be easily plagued by incomplete or erroneous data, so that in many applications, the paths computed by the global navigation algorithm need to be revised and evaluated online, which is time consuming and cost inefficient. Furthermore, the performances of these algorithms are very likely to be ruined by the unexpected "changes" in the environments, such as an unexpected obstacle appearing in the environment. 
\par
	Local navigation algorithms \cite{LZL07,DIS07,TS10,MTS11,SAVH13}, on the other hand, use sensory systems to observe a small fraction of an unknown environment. The best decisions are made iteratively each time using the information available. These algorithms are more realistic in real time implementation where the sensory systems have finite detection range. These navigation algorithms are more computationally efficient when compared to global navigation algorithms as relatively less information needs to be processed each time. However, many of the researches only focus themselves on static environments, approaches such as dynamic window \cite{SeMaPe05,FBTh97}, the curvature velocity \cite{Simm96}, and the lane curvature \cite{NaSi98} approaches, consider only the simple problem of navigation in static environments. The obstacle avoidance with moving obstacles is much harder because the object only has finite time and limited options to avoid the obstacles when they are detected within a certain range. Some approaches are proposed to deal with obstacle avoidance with moving obstacles, for example, velocity obstacles
\cite{FS98,LaLaSh05}, collision cones \cite{ChGh98}, or inevitable collision states \cite{FTA04,OE06}. These approaches require the deterministic knowledges about the obstacles' velocity and a moderate rate of its change. Finally, many of these papers \cite{S00, KR97, KRR98, LA92} do not take into account non-holonomic constraints on robot's motion which is a severe limitation in practice. 
\par
	The following subsections briefly discuss some of the well-know navigation algorithms. 
\par

	\subsection {BUG Algorithms}

	The family of BUG algorithms are examples of biologically-inspired algorithms. The concept of first two members, Bug1 and Bug2, of the Bug algorithm family is introduced in \cite{LV86}. The maneuver of BUG algorithms are inspired from animals (i.e. rats) unidirectionally following the obstacle (i.e. walls) boundary to the target location in a known environment. In general, Bug2 is more "aggressive" than Bug1 in terms of finding the possible paths to the target, and work in many scenarios where BUG are found impossible to accomplish. The advantage of the Bug algorithms is that it allows the object to navigate in an environment with obstacles whose shapes and coordination are not known, which is the primary assumption for many approaches. However, the problem with these algorithms is that they do not guarantee global convergence to the target, i.e. the object may be "looping " around the same obstacle infinitely. This is partially because the BUG algorithms are often proposed as pseudo codes. The obstacle boundary following methods and the leave conditions are not specifically defined which makes the implementations of the BUG algorithms inconsistent for different types of control systems. Also, these algorithm are only investigated in the static environments where the obstacles are stationary, which is not applicable in many real world scenarios where the obstacles are dynamically moving. The later varieties of the BUG algorithms includes: Alg1 and Alg2\cite{SAVM90, SAV90}, DistBug\cite{KIRE97}, RoveerBug\cite{LSL99}, Rev1 and Rev2\cite{NH00}, TangentBug\cite{KI98},and others \cite{LV90,NHM99}. The performance of the BUG algorithms varies in different environments (as investigated in \cite{NJ07}), for example, TangentBug\cite{KI98} works best in the environments with wider spaces that allow the equipped range sensors to be used efficiently . A detailed performance comparison of most of the BUG algorithm family is carried out in \cite{NJ07}.
\par

	\subsection {Model Predictive Control}

	Model Predictive Control (MPC) is an iterative computation process which computes a set of future control signals which minimises some cost functions at every time instant, only the first control signal is applied to the control system. This process is repeated at next time instant until the control objective is achieved. The concept of (MPC) can be employed to solve the problems of safe robot navigation. In case of robot navigation, the robot computes a set of future control signals for $t \in [t, t+T)$ based on the feedbacks from sensory systems at time $t$ and the control signal at time $t$ is applied. This control scheme is more flexible than global navigation algorithm because the robot has access to the updated information from sensory systems at every time instant, any changes in the environments can be adapted according to the control strategies. The final convergence achieved when the robot arrived at the target location.
\par
	There are a number of varieties of MPC being proposed in different literatures. The robust Model Predictive Control is favourable for the implementation of navigation algorithms on real control systems.  There are three main approaches to robust Model Predictive Control: 1) Min-Max MPC\cite{SPO98} 2) Constraint Tightening MPC \cite{RAH06,KYR07} and 3) Tube MPC \cite{LWC04}. The common features of these robust Model Predictive Control is demonstrated by their performances in the presence of set bounded disturbance which is a desired property in real robotic applications. The general MPC is applicable for vehicles with simple linear models. However, many of the vehicle exhibits more complicated non-holonomic characteristics with bounded speed and angular velocity. The non-linear MPC is a more suitable control approach to these non-linear vehicle types.
\par

	\subsection{Velocity Obstacle Approach}
	The problem of safe navigation in dynamic environments is much more challenging. A number of approaches have been proposed to solve the problem. Velocity Obstacle Approach (\cite{FP93,FP98}) is one of the most well-known approaches which achieves navigation tasks in dynamic environments by computing a set of potential collision velocities, any velocity within this set will cause a collision in future time. The primary objective is to choose a velocity that is outside of the collision velocities set to guarantee the safety of the object, and then find the optimal velocity (depending on different criteria) among these safe velocities.  This approach has the merit of taking into account the velocities of the obstacles, predicting potential collision velocities in the future, allowing various motion planning options and guaranteeing the safety of the object in dynamic environments. This approach has proven its successfulness in theoretical basis, however, since the approach does not consider the non-holonomic constraints of the real robot, it may not be suitable for many of the real time implementations. Moreover, the set of collision velocities are computed based on the assumption that the obstacles maintain their velocities in the future time. For obstacles moving with non-linear velocities, a more advanced Non-Linear Velocity Obstacles \cite{ZS01,LFS02,LF05} has to be estimated which is computationally expensive. Many other navigation approaches are inspired by the concept of Velocity Obstacle Approach, including Reciprocal Velocity Obstacle Approach \cite{VD08}, Probabilistic Velocity Obstacle \cite{FC07}, Generalised Velocity Obstacle \cite{WD09}  etc.  
\par
	In \cite{OE05, OE06}, an approach similar to velocity obstacle is proposed, which computes the collision velocities in Velocity Space. This approach takes into account the non-holonomic constraints of the object during computation. Due to the nature of the computation method, this approach can only apply to obstacles that are moving in straight lines or other paths consisting of segments of short straight lines. Furthermore, this approach assumes the shapes of the obstacles to be either circles or polygons. The Collision Cone Approach \cite{CA98} is one of a few approaches which allows the objects to avoid moving obstacles regardless of the shapes of the obstacle. Once again, the non-holonmic constraints are not considers in \cite{CA98}.  Examples concerning nonholonomic robots include artificial potential approach, combined with sliding-mode control for gradient climbing \cite{LinHuLa06,FeRu07}, and kinematic control based on polar coordinates and Lyapunov-like analysis \cite{ChQuPoFa10}.
\par

	\subsection {Other Navigation Algorithms}

	The Artifical Potential Fileld (APF) \cite{K86} approach treats the robot as a point in a potential field. The robot is influenced by two forces: the repulsive force from the obstacles and the attractive force from the target. The advantage of APF is that the approach is able to generate a safe path to the target with little computational effort. However, the robot can be trapped in local minima when APF is employed which leads to failure of navigation task. This problem has been investigated by many researcher and possible solutions have been proposed (see e.g. \cite{MA10,MA12}). The APF is also improved to adapt robots of unicycle model in \cite{RJM08} .
\par
	Sliding Mode Control (SMC) is a non-linear control scheme which offers a number of benefits in robotic navigation. The most well-known advantage of SMC is its robustness against external disturbance and internal state uncertainties. This is critically important in the cases of noisy sensory measurements or uncertain state estimations \cite{MASH10}. Since the execution of SMC involves simply switching between several states (usually two or three), it is generally simple to be implemented in a wide range of dynamic systems. A number of sliding mode controller has been designed for robot navigation: in \cite{MTS11,MASH13},  a sliding mode controller is proposed for the problem of border patrolling (boundary following) and obstacle avoidance; the environmental extremum seeking problem in an unknown environment is investigated in \cite{MTS1}; a navigation algorithm based on range-only measurements is proposed in \cite{TS10}; navigation problems in maze-like environment \cite{MAH13} and environmental level tracking problem is considered in \cite{AS12}
	
\par
	The motion safety for robotic systems are studies on \cite{FTA04, FT07, PR07}. In these papers, three safety criteria which assess the safety state of an object in a risky environment. These three criteria are summarised as: 1) robotic system's dynamic 2) environment obstacle's future behavior and 3) time horizon. The safety of the robotic systems are not guaranteed if any of these criteria is violated. Based on the criteria, the authors propose a state in which robotic systems will always collide with the obstacles regardless of the future motion of the robotic system is, namely Inevitable Collision State. A survey regarding other recent navigation algorithms can be found in~\cite{MH13}.

	\section {Formation Building Control Strategies}

		Formation building control strategies focus on the coordinated control of a group of mobile robots, which is another topic in the field of autonomous navigation. The formation building control strategies allow the groups of mobile robots to move in a desired pattern, which is found particularly useful in a wide range of applications including robotic sensing networks \cite{CJ04,OP04}, rescue robot teams \cite{KGS06,SH10}, research and explore operation in unknown environments \cite{SA04} etc. The employ of formation building strategies in these applications has advantages of increasing system efficiency and robustness and reducing the system cost and operation effort \cite{CYQ05}.
\par
		Numerous formation building control strategies have been proposed in recent years. The leader-follower approach \cite{PKC91,PKC96,DJP98} is simple to understand: one of the robots in the group is assigned as the leader which has the group formation information to guide the movements of the followers. This centralised formation approach is easy to implement in a group of real mobile robots. However, the formation group is not efficient as the leader is responsible for most of the computation tasks, and the performance of the group relies heavily on the leader. Moreover, the group is not robust as any failure on the leader (e.g. communication error, computation error) will cause the whole group to collapse. Finally, as surveyed in \cite{RW08}, the typically leader-follower approach is lack of inter-vehicle feedback throughout the group.
\par
		The study of decentralised control laws for groups of autonomous mobile robots has emerged as a challenging new research area in recent years (see, e.g., \cite{FJT02, FJ02, IHKS09, YGH06, JA03, GMM10, SAV04, HY08,ST10a,CS11}). Broadly speaking, this problem falls within the domain of decentralised control, but the unique aspect of it is that groups of mobile robots are dynamically decoupled, meaning that the motion of one robot does not directly affect that of the others. In decentralised formation control schemes, each member in the group is responsible for deciding its own motion based on the state of itself and its neighbors. The group's objective can still be achieved even when failure of any members in the group occurs.
\par
		A discrete time model of a system consisting of several autonomous agents is presented in \cite{VCBCS95}, This model can be viewed as a special case of a computer model proposed in \cite{REY86} for the computer animation industry and mimicking animal aggregation. This model proves that control of the motion of multiple autonomous agents can be achieved in decentralised fashion. A modification of the model in \cite{VCBCS95} is presented in \cite{JA03}, and a mathematical analysis of the model is presented. The main results of \cite{JA03} are sufficient conditions for coordination of the system of agents that are given in terms of a family of graphs characterizing all possible neighbouring relationships among agents. In behavior-based approach \cite{BT98,LD07,MJA06}, a set of desired behaviors are prescribed for each member in the group. These desired behavior may includes: collision avoidance to other members in the group, avoidance of exterior obstacles, formation building, and formation keeping. The motion of each member is based on the weighting of these behaviors.  The performance of the behavior-based approach under the influence of member failure is studied in \cite{RJ99}. Other types of formation control strategies can be found in e.g.\cite{GV03,KFT03,YJ03,CYQ05}.
\par
		It should be pointed out that many papers in this area consider simplest first- or second-order linear models for the motion of each robot; see, e.g., \cite{OSR04, THG03} and \cite{THG03ii}. Therefore, the obtained results are heavily based on tools and methods from linear system theory, such as stochastic matrices or graph Laplacians. It is known that there are examples of unrealistic physically embodied behaviour that would be possible under such simplified models. In particular, the robot motion in such linear models does not satisfy the standard hard constraint on either robot speed or angular velocity. Furthermore,  it can be  shown that the models proposed in \cite{SAV04} and many other papers will result in arbitrarily large robot angular velocity and arbitrarily small robot turning radius, which is impossible on actual wheeled robots.

	\section{Main Contuributions of This report}

		The main contributions of this report are described as follows:

 		\begin{enumerate}

			\item Three novel obstacle avoidance strategies are proposed for safe navigation of mobile robots of non-holonomic unicycle model. The detailed descriptions of the proposed navigation algorithms are presented.

			\item The computer simulation results and experiment results with real P-3 mobile robot are presented with detailed descriptions and explanations. The results demonstrate the features of the proposed algorithms and their performance with a real non-holonomic systems.

			\item These navigation algorithms are compared with each other across various aspects including measurements required for computation, computation complexity and performance comparison in different scenarios. Each of the proposed algorithm has unique characteristics over the others which make itself more efficient in one or more of the mentioned aspects.

			\item The algorithms have been implemented on two real control systems: intelligent wheelchair (SAM) and autonomous hospital bed (Flexbed). The objective is to show the applicabilities of the algorithms with real control systems. The performance of the proposed navigation algorithms with the SAM and Flexbed demonstrate their capabilities to achieve navigation tasks in complicated real time scenarios. The applications of the proposed algorithms are not limited to these two control systems, but can be easily implemented into many other control systems.

			\item The problem of formation building for a group of mobile robots is considered. We propose a robust decentralised formation building algorithm for a group of mobile robots to move in the defined geometric configurations. Furthermore, we consider a more complicated formation problem with a group of anonymous robots, these robots are not aware of their position in the final configuration and have to reach a consensus during the formation process. We propose a randomized algorithm for the anonymous robots which achieves the convergence to a desired configuration with probability 1. The performance and applicability of the proposed algorithm is confirmed by the computer simulation result as well as the experiments with a group of real mobile robots.

		\end{enumerate}

	\section {Chapter Outline} 

		This report is orangised as follows: the detailed descriptions of the navigation algorithms Biologically-Inspired Navigation Algorithm (BINA), Equidistant Navigation Algorithm (ENA) and Navigation Algorithm based on Integrated Environment Representation (NAIER) are presented in Chapter~\ref{C2},~\ref{C3} and~\ref{C4}, respectively. We also present the simulation results and the experimental results with the real robots for each of the proposed navigation algorithms in the corresponding chapters. These results demonstrate the features, effectiveness and applicability of the proposed navigation algorithms. These three navigation algorithms are compared with each other across various aspects to give a deeper insight to the merits of each of the proposed algorithm in Chapter~\ref{C5}. The implementation of the proposed algorithms on two real control systems, wheelchair and hospital bed, are shown in Chapter~\ref{C6} and Chapter~\ref{C7}. The performance and applicability of the algorithms are confirmed by real life experimental results. We proposed a decentralised control strategy for formation building of a group of mobile robots and a randomised algorithm for formation building with anonymous robots in Chapter~\ref{C8}. Finally, Chapter~\ref{C9} states the conclusion of this report. 
		
\pdfoutput=1

\chapter{Biologically-Inspired Obstacle Avoidance Strategy for Safe Navigation} \label{C2}

	Researchers in the area of robot navigation are finding much inspiration from biology, where the problem of controlled animal motion is a central one. Animals, such as insects, birds, or mammals, are believed to use simple, local motion control rules that result in remarkable and complex intelligent behaviours. The algorithm proposed in this paper also belongs to the class of biologically inspired or  biomimetic navigation algorithms. In biology, a similar obstacle avoidance strategy is called negotiating obstacles with constant curvatures (see e.g. \cite{Lee98}). An example of such a movement is a squirrel running around a tree. We confirm the performance of our real-time navigation strategy with
extensive computer simulations and experiments with a Pioneer P3-DX mobile wheeled robot.

	\section{Problem Description} \label{PD1}
\begin{figure}[h]
		\centering
		\includegraphics[width=4.0in]{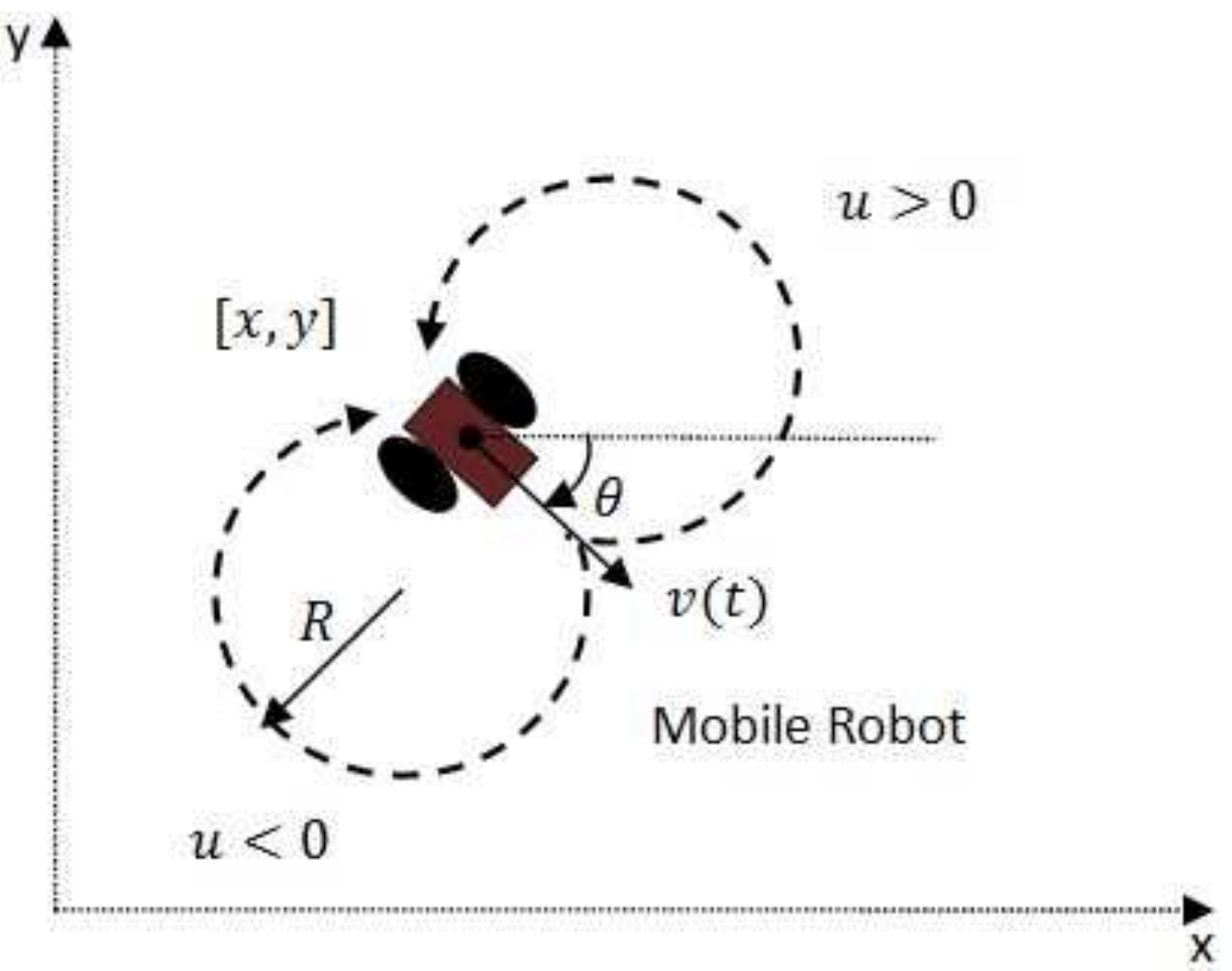}
		\caption{Coordinate and orientation of the mobile robot and its minimal turning radius $R$}
		\label{para}
		\end{figure}

		We consider a wheeled mobile robot, modeled as unicycle, travels in a two dimensional plane. The position of the wheeled mobile robot is represented by the absolute Cartesian coordinates as $(x,y)$ of the reference point, which is the center of mass of the mobile robot. The orientation of the mobile robot, measured in counterclockwise direction from the reference axis, is give by $\theta$, see Fig.~\ref{para}. The motion of the mobile robot exhibits non-holonomic characteristics with hard constraints on its speed and angular velocity. The mathematical model of the mobile robot is as follows:
		\begin{equation}
		\label{bi_1}
		\begin{array}{l}
		\dot{x} = v(t) \cos \theta,
		\\
		\dot{y} = v(t) \sin \theta,
		\\
		\dot{\theta} = u 
		\end{array},~~~ 
		\begin{array}{l}
		x(0) = x_0,
		\\
		y(0) = y_0,	
		\\
		\theta(0) = \theta_0
		\end{array} 
		\end{equation}
		where 
		\begin{equation}
		\label{bi_max}
		 u\in [-U_{max},U_{max}],~~~v(t)\in [0,V_{max}].
		\end{equation} 

		here $v(t)$ and $u(t)$ are the speed and angular velocity of the mobile robot, respectively. $V_{max}$ and $U_{max}$ are the non-holonomic constraints.  This is the standard non-holonomic model with hard constraints on the angular and linear velocities. It is well-known that the motion of many wheeled robots, missiles and unmanned aerial vehicles can be described by this model; see e.g. \cite{TS10,MTS11,MTS11ronly,low07} and references therein.  The non-holonomic constraints establish restrictions on the forward and rotational movements of the mobile robot, see Fig.~\ref{para}; in particular, for given $v$, they limit the turning radius of the mobile robot from below by 
		\begin{equation}
		\label{bi2_mini}
		R = v/U_{max}.
		\end{equation}
		\par

		There are also a steady point-wise target $\bf T$ and several moving obstacles. These obstacles are moving in random directions with various speeds. Each of the obstacles at time $t$ occupies a certain domain $D_i (t)$ for $i = 1, 2,\ldots,n $, each of the domain $D_i (t)$ is assumed to be a closed bounded planar set.
\par	

		The objective for safe navigation of the mobile robot in the cluttered dynamic environment is to avoid the enroute obstacles with a safety distance $d_{safe}>0$ and drive the mobile robot to the target location at time $t_f>0$, i.e, $c(t_f) = \bf{T}$ and  $d_i(t)\geq d_{safe}~\forall i=1,2,\ldots,k, ~~\forall t\in [0,t_f]$.
\par
\begin{figure}[h]
	\centering
	\subfigure[]{\scalebox{0.50}{\includegraphics{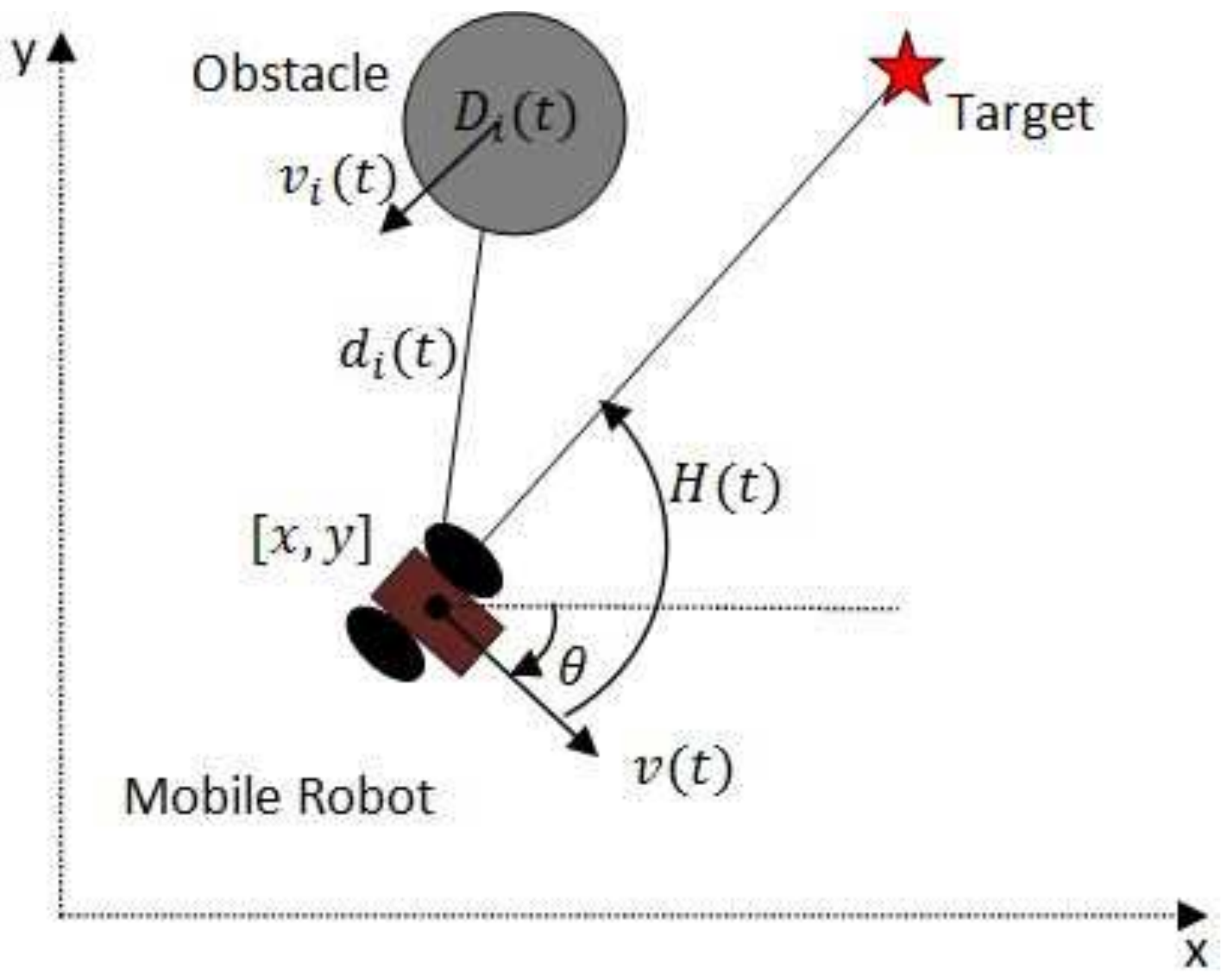}}
	\label{para1}}
	\subfigure[]{\scalebox{0.50}{\includegraphics{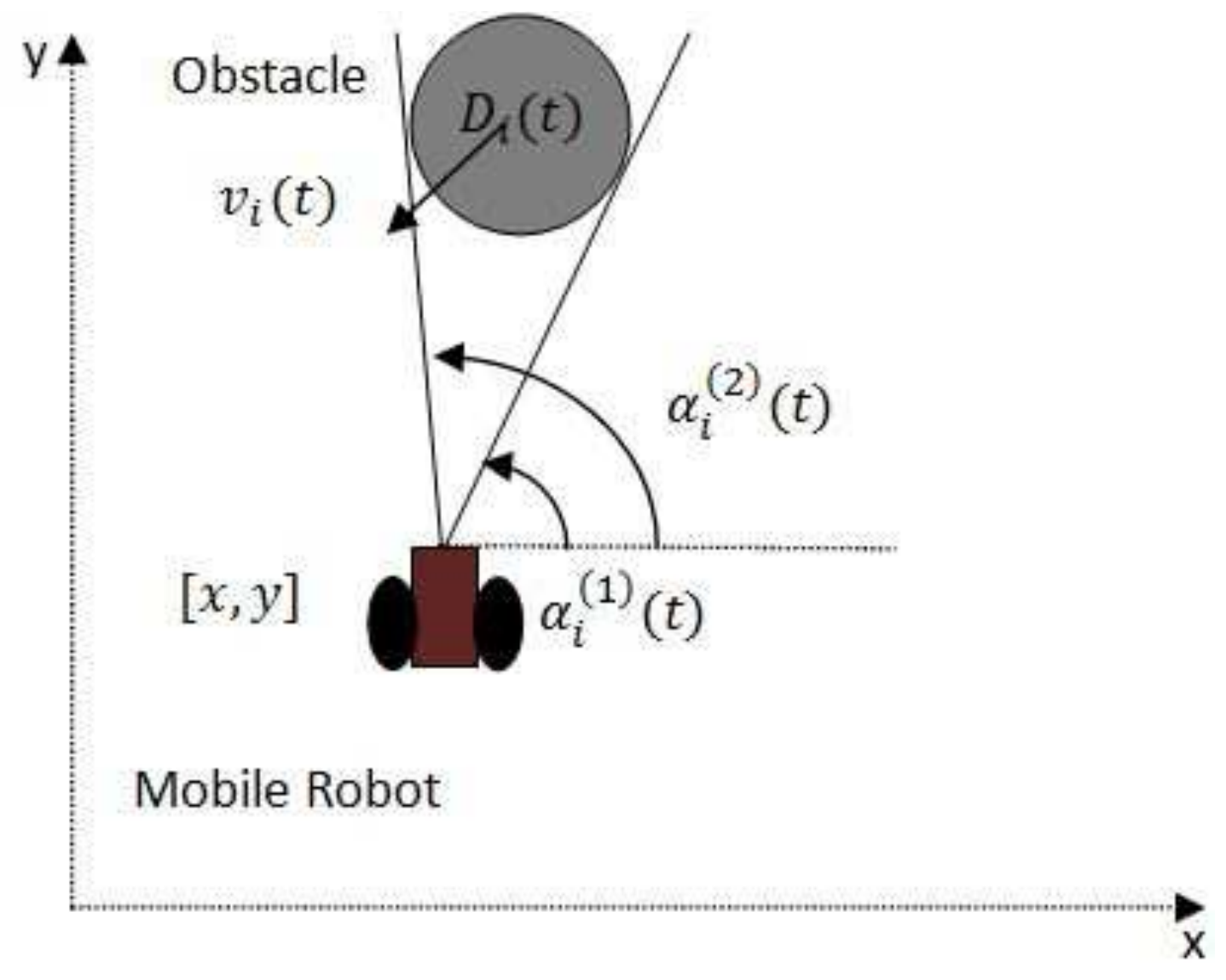}}
	\label{para2}}
	\caption{(a)Minimum distance $d_i(t)$ and angular difference $H(t)$; (b) the Vision cone $\alpha_i^{(1)}$ and $\alpha_i^{(2)}$}
	\end{figure}

		To achieve the navigation task, the following information is required:

			\begin{enumerate}
  			\item The minimum distance $d_i(t)$ between the mobile robot and the border of the obstacle $i$, see Fig.~\ref{para1} 
  				\begin{equation*}
					d_i(t) :=\min_{r\in D_i(t)}\|r-c(t)\|.
			            \end{equation*}
  	          where $ c(t):=[x(t),y(t)]$ is vector of  the the mobile robot's Cartesian coordinate. Note that $\|\cdot\|$ denotes the standard Euclidean vector norm. And $\dot{d_i}(t)$ is also accessible.
			\par
 
 			 \item  The velocity $v_i(t)$ of the moving obstacle.
 			 We assume that any  obstacle $i$   bounded and moving in the plane without rotations with the time-varying velocity $v_i(t)$. and the velocities $v_i(t)$ satisfy the following constraint:
     				\begin{equation}
    				 \label{cons1}
    					   \|v_i(t)\| \leq V<V_{max}
    				 \end{equation}
 			 for  all times $t$ and for all obstacles $i$.  It is obvious that if the constraint (\ref{cons1}) does not hold, the safe navigation of the mobile robot is impossible.

  			\item The mobile robot measures $\alpha_i^{(1)}$ and $\alpha_i^{(2)}$ with respect to the reference axis, see Fig.~\ref{para2}, these are the boundaries of the vision cone from the mobile robot to obstacle $i$. 
   			 \item The angular difference $H(t)$ between its current orientation and the direction of the target, see Fig.~\ref{para1}

			\end{enumerate}

	\section {Biologically-Inspired Reactive Navigation Algorithm} \label{A1}

		In this section, we present the description of the biologically-inspired or biomimetic navigation algorithm. 	
		\par

\begin{figure}[h]
		\centering
		\includegraphics[width=3.5in]{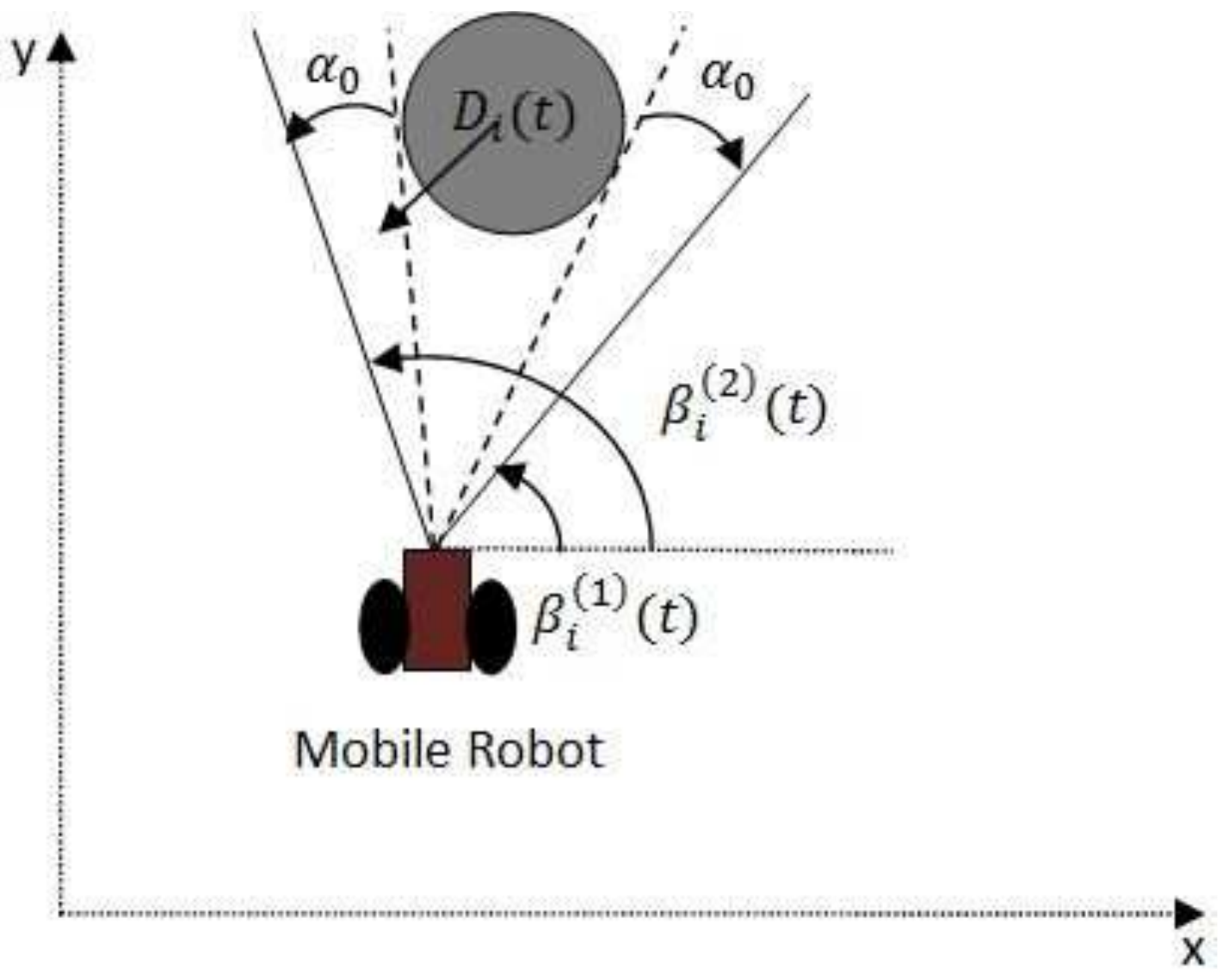}
		\caption{The enlarged vision cone}
		\label{enlarged}
		\end{figure}

		We first enlarge the vision cone $\alpha_i^{(1)}$ and $\alpha_i^{(2)}$ measured from the mobile robot to the obstacle $i$ by a given angle $\alpha_0$. $0 < \alpha_0 < \pi$, we denote the boundaries of the enlarged vision cone for obstacle $i$ as:
		\begin{equation}
		\label{enlarged_vc}
		\beta_i^{(1)}(t):=\alpha_i^{(1)}(t)-\alpha_0,~~~\beta_i^{(2)}(t):=\alpha_i^{(2)}(t)+\alpha_0;
		\end{equation}
		
		see Fig.~\ref{enlarged}, we also introduce the following vectors $l_i^{(1)}$ and $l_i^{(2)}$ ($l_i^{(1)},l_i^{(2)}\in\mathbb{R}^2$) as:
		\begin{equation}
		\label{vector}
		 l_i^{(j)}(t):=(V_{max}-V)[\cos(\beta_i^{(j)}(t)),\sin(\beta_i^{(j)}t))]~~~~\forall j=1,2
		\end{equation}
		\par
		where $V_{max}$ and $V$ are the constants from (\ref{bi_max}) and (\ref{cons1}). A geometric interpretation of (\ref{vector}) is straightforward. The equations (\ref{vector}) define the two lines which are the boundaries of an enlarged vision cone of the obstacle. These lines are also known as occlusion lines.
\par
		\par

		Let $r_1$ and $r_2$ be non-zerp two-dimensional vector. We use $\gamma (r_1, r_2)$ to represent the relative angular difference measured  from $r_1$ to $r_2$ in counterclockwise direction.  We need the following $f$ function to assign different values depending $\gamma (r_1, r_2)$ :
		\begin{eqnarray}
		\label{F}
		 f(r_1,r_2):= \left\{ 
                    		\begin{array}{ll}
			0&  \gamma(r_1,r_2) =0\\
			1&  0<\gamma(r_1,r_2) \leq \pi\\
			-1&-\pi<\gamma(r_1,r_2) <0,
			\end{array} \right.
		\end{eqnarray}
		where $r_1, r_2 \in\mathbb{R}^2$, $\gamma (r_1,r_2)\in(-\pi,\pi]$.
		\par

		Furthermore, let $C > 0$ be the switching distance.
		\par 
		When the distance between the mobile robot and the obstacle $i$ is reduced below $C$ at $t_0$, the robot measures the vision cone ($\alpha_i^{(1)}, \alpha_i^{(2)}$) and calculates the vector $l_i^{(1)}$ and $l_i^{(2)}$ by (\ref{vector}).
		\par
		The velocity of the obstacle $v_i(t)$ is added to both $l_i^{(1)}$ and $l_i^{(2)}$, and we want to find the one which makes the smallest angle between itself and the mobile robot's current velocity vector $v(t)$,  i.e, the minimum condition in (\ref{bi_mini}) is satisfied, for all time $t$  during the process of avoiding obstacle $i$ and $t \ge t_0$.
			\begin{equation}
			\label{bi_mini}
			\min_{j=1,2} |\gamma(v_i(t)+l_i^{(j)}(t),v(t))|
			\end{equation}
			\par
\begin{figure}[h]
			\centering
			\subfigure[]{\scalebox{0.50}{\includegraphics{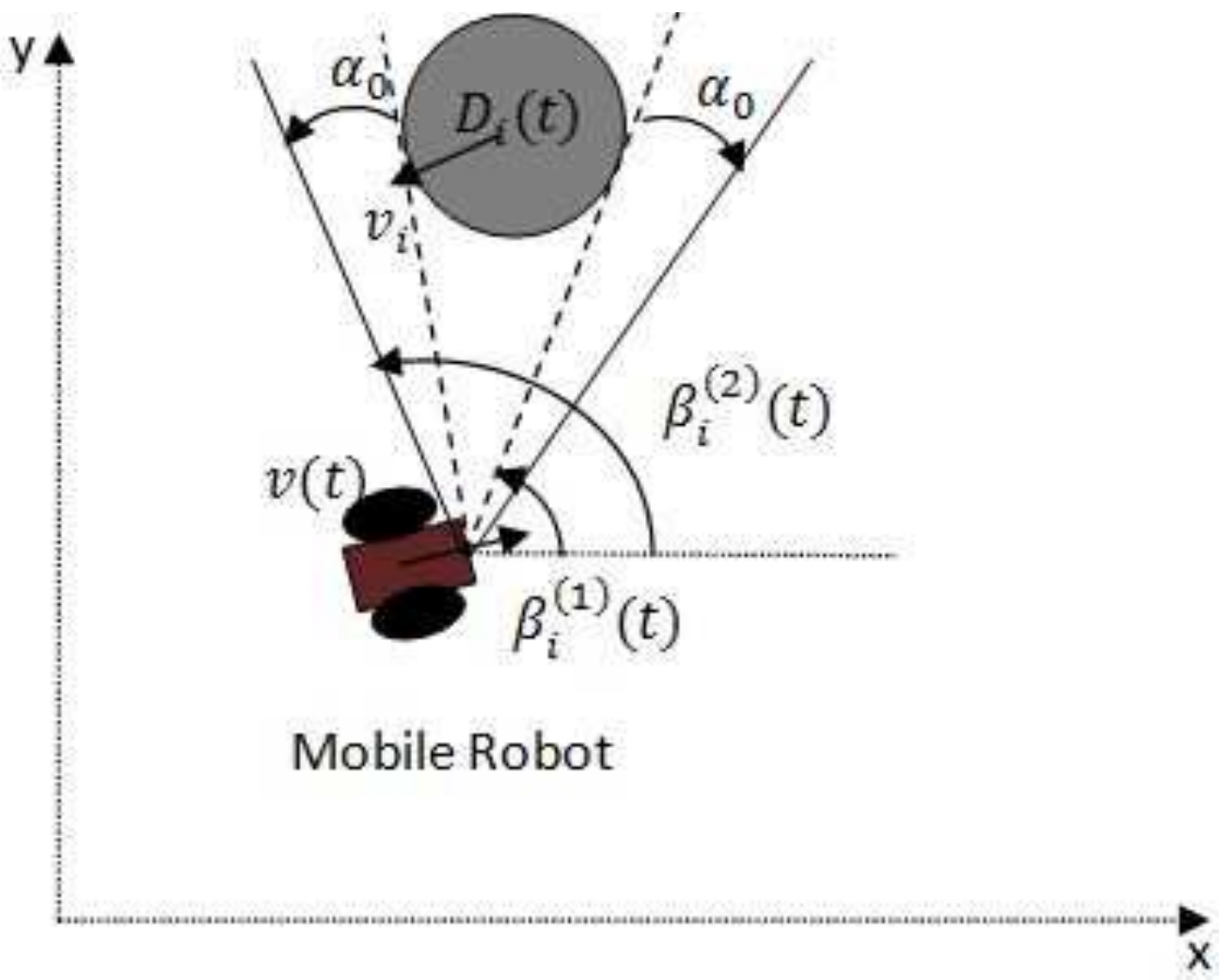}}
			\label{illu1a}}
			\subfigure[]{\scalebox{0.50}{\includegraphics{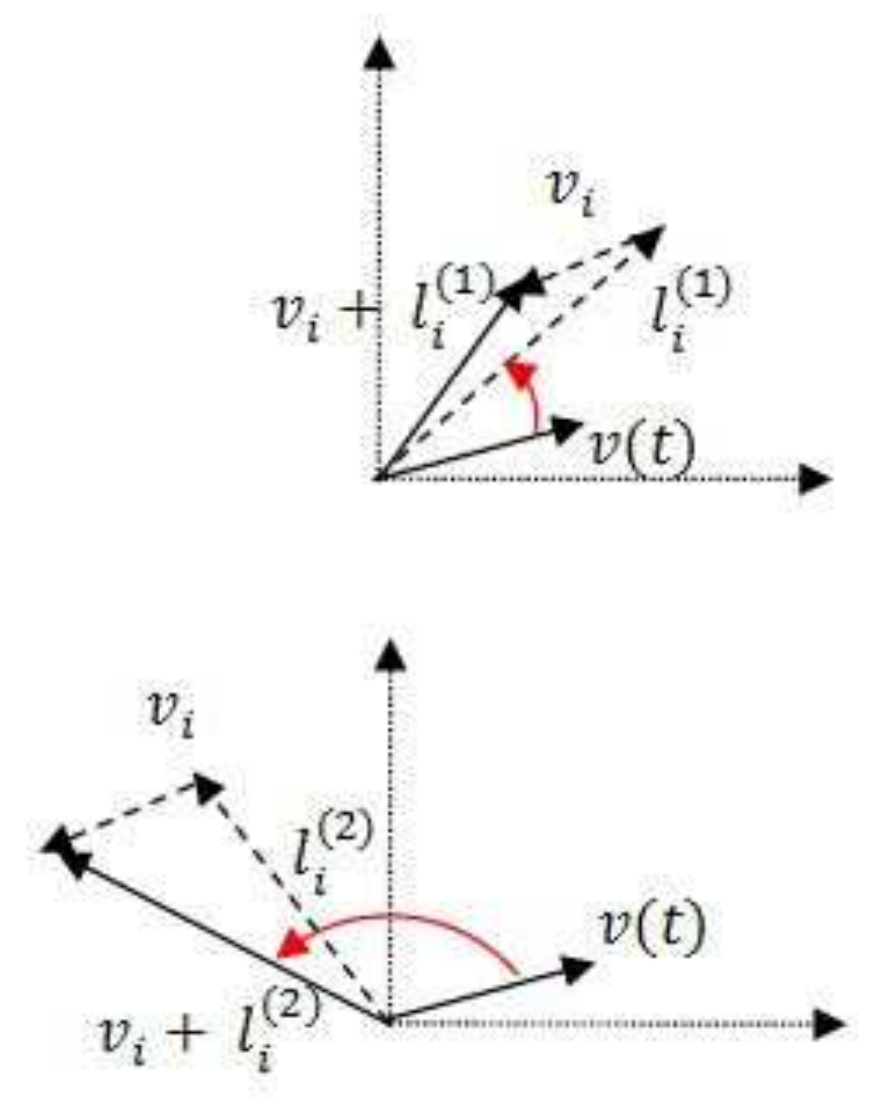}}
			\label{illu1b}}		
			\caption{(a)Illustrative example; (b)Two possible angle differences}
			\end{figure}

		An illustrative example of (\ref{bi_mini}) is given in Fig.~\ref{illu1a}. The angular difference between $v(t)$ and $v_i(t)+l_i^{(j)}(t),v(t))$ for $j = 1,2 $ is shown in curved arrows. It can be observed from Fig.~\ref{illu1b} that the vector $v_i(t)+l_i^{(1)}(t)$ is closer to the robot's current moving direction $v(t)$.

			\par

		Let $h$ be the index for which (\ref{bi_mini}) is achieved, we proposed the following obstacle avoidance law:

			\begin{eqnarray}
			\label{bi_cont}
			 u(t)=U_{max}f(v_i(t)+l_i^{(h)}(t),v(t));
			\nonumber
			\\
			V(t)=\|v_i(t)+l_i^{(h)}(t)\|.
			\end{eqnarray}
			\par
		
		Notice that the control inputs defined in (\ref{bi_cont}) satisfy the constraints (\ref{bi_max}). (\ref{F}) implies that the non-holonomic constraint on the angular velocity is satisfied, i.e. $|u(t)|\leq U_{max}$. Furthermore, (\ref{bi_cont}), (\ref{cons1}) and (\ref{vector}) imply that 
			\begin{eqnarray*}
			V(t)=\|v_i(t)+l_i^{(h)}(t)\|\leq \|v_i(t)\|+\|l_i^{(h)}(t)\|\\
			\leq V+ (V_{max}-V)=V_{max}.
			\end{eqnarray*}
Therefore the second constraint (\ref{bi_max}) on the linear velocity  holds as well.
		\par

		The intuition behind the obstacle avoidance strategy (\ref{bi_cont}) can be explained as follows: The law (\ref{bi_cont}) belongs to the class of sliding mode control laws; see e.g. \cite{UT92}.  The closed-loop system (\ref{bi_1}), (\ref{bi_cont}) is a system of ordinary differential equations with a discontinuous right-hand side. The control law (\ref{bi_cont})  steers the velocity vector $v_R(t)$ of the robot to a switching surface on which the robot's velocity becomes equal to the vector 
$v_i(t)+l_i^{(h)}(t)$. Now consider a coordinates system moving with the velocity $v_i(t)$. In these coordinates, the obstacle $i$ becomes stationary, and the robot is moving with the velocity $l_i^{(h)}(t)$. Furthermore, the motion with the velocity $l_i^{(h)}(t)$ is an obstacle avoidance strategy which keeps a constant avoiding  angle $\alpha_0$  between the instantaneous moving direction of the robot and one of the two boundary rays of the vision cone of the obstacle $i$ from the robot. In other words, the control law consists of steering toward the closer of the two edges of the enlarged obstacle vision cone defined by (\ref{vector}). The idea of this obstacle avoidance strategy originates from biology where it was called negotiating obstacles with constant curvatures (see e.g. \cite{Lee98}). An example of such a movement is a squirrel running around a tree. A similar strategy for avoiding stationary obstacles was proposed and studied in \cite{TeSav10rb}.
		\par


		The complete navigation algorithm combines the obstacle avoidance law (\ref{bi_cont}) with the target pursuit law (\ref{bi_pur}).

		\begin{equation}
		\label{bi_pur}
		\begin{array}{l}
		 u(t)=0;  
		\\   v(t) = V_{max}.
		\end{array}
		\end{equation}
		\par 

		Introduce the following constants $\alpha_i>0$:
		\begin{equation}
		\label{ai}
		 a_i:=\frac{R_i}{\cos\alpha_0}-R_i~~~\forall i=1,2,\ldots,k.
		\end{equation}
		\par

		The execution of the navigation algorithm requires proper switching between the obstacle avoidance law (\ref{bi_cont}) and the target pursuit law (\ref{bi_pur}) obeys the following rules:

		{\bf R1:} (\ref{bi_pur})$\rightarrow$(\ref{bi_cont}) when the distance between the mobile robot and the obstacle $i$ reduced below trigger distance $C$ at time $t_0$.i.e, $d_i(t_0)=C$ and $\dot{d_i}(t_0)<0$
\par
		{\bf R1:} (\ref{bi_cont})$\rightarrow$(\ref{bi_pur}) when the distance is greater than certain values $1.1\alpha_i$ and the mobile robot is oriented towards the target at time $t_*$. i.e, $\dot{d_i}(t_*)\leq1.1\alpha_i$  and $\theta(t_*)=H(t_*)$ 

\par
	Our main theoretical result requires a number of assumptions.
	\begin {assumption}
		\label{as21}
		We assume that every obstacles $i$ can be covered by a circular disk of radius $R_i>0$. moving 
with the constant velocity $v_i\neq 0$. Furthermore, $V_i<V_{max}~~\forall i=1,2,\ldots,k$
where $V_i:=\|v_i\|$, and the vectors $v_i$ and $v_j$ are not collinear for all $i\neq j$. 
\end {assumption}

		\begin{assumption}
		\label{as23}
		The maximum angular velocity of the mobile robot $U_{max}$ satisfies:
		\begin{equation}
		\label{cons3}
		 U_{max} >F_i
		\end{equation}
		where $F_i$ is defined as
		\begin{equation}
		\label{Fi}
		 F_i:=\frac{(V_i+V_{max})R_i}{(R_i+d_{safe})^2\sqrt{1-\frac{R_i^2}{(R_i+d_{safe})^2}}}~~~~\forall i=1,2,\ldots,k.
		\end{equation}
		\end{assumption}

		\begin{assumption}
		\label{as22}
		 The switching distance $C$ for which the obstacle avoidance law to be activated satisfies the following constraint:

		\begin{equation}
		\label{cons2}
		 C \geq \frac{\pi V_{max}}{U_{max}-F_i}+1.1a_i~~~~\forall i=1,2,\ldots,k.
		\end{equation}
		\end{assumption}

		\begin{assumption}
		\label{as24}
		The avoiding angle $\alpha_0$ in (\ref{enlarged_vc}) satisfies the following condition:
		\begin{equation}
		\label{cons5}
		\alpha_0\geq \arccos\left(\frac{R_i}{R_i+d_{safe}}\right).
		\end{equation}
		\end{assumption}
		 It is obviously follows from Assumption \ref{as24} and (\ref{ai}) that $a_i\geq d_{safe}$ for all $i$.
		\par

		\begin{assumption}
		\label {as25}
		We assume the distance between different obstacle $i$ and obstacle $j$ satisfies:
		\begin{equation}
		\label{cons4}
		Dist_{ij}(t)\geq 2C+\frac{\pi V_{max}}{U_{max}-F_i};
		\end{equation}
		and the distance between obstacle $i$ and the target $\bf{T}$ satisfies:
		\begin{equation*}
		Dist_{i}\geq 1.1a_i
		\end{equation*}
		\end{assumption}

\begin{theorem}
\label{T21} Let $d_{safe}>0$,  $0<\alpha_0<\frac{\pi}{2}$ and $C>0$ be given and suppose  that Assumptions~{\rm \ref{as21} -- \ref{as25}}
hold.
Then  the control rule {\bf R1, R2} provides a
target reaching strategy with collision avoidance.
\end{theorem}

		The mathematical proof of Theorem~\ref{T21} is proposed in~\cite{SAW13}

	\section{Computer Simulation Results}

	In this section, we present the computer simulation results for the proposed navigation algorithm. With these simulation results, we demonstrate the capability of the proposed navigation algorithm to guide the mobile robot to the target position in cluttered dynamic environments.
\par

	In the following simulation results, we give figures for each simulation and provide simulation data to explain the features of the proposed navigation algorithm when necessary. In these figures, The mobile robot, the obstacles and the target are depicted by gray dick, black circles and red star, respectively, and the instantaneous moving direction of the mobile robot and the obstacles are shown in black arrows.
\par


		\begin{figure}[h]
		\centering
		\subfigure[]{\scalebox{0.4}{\includegraphics{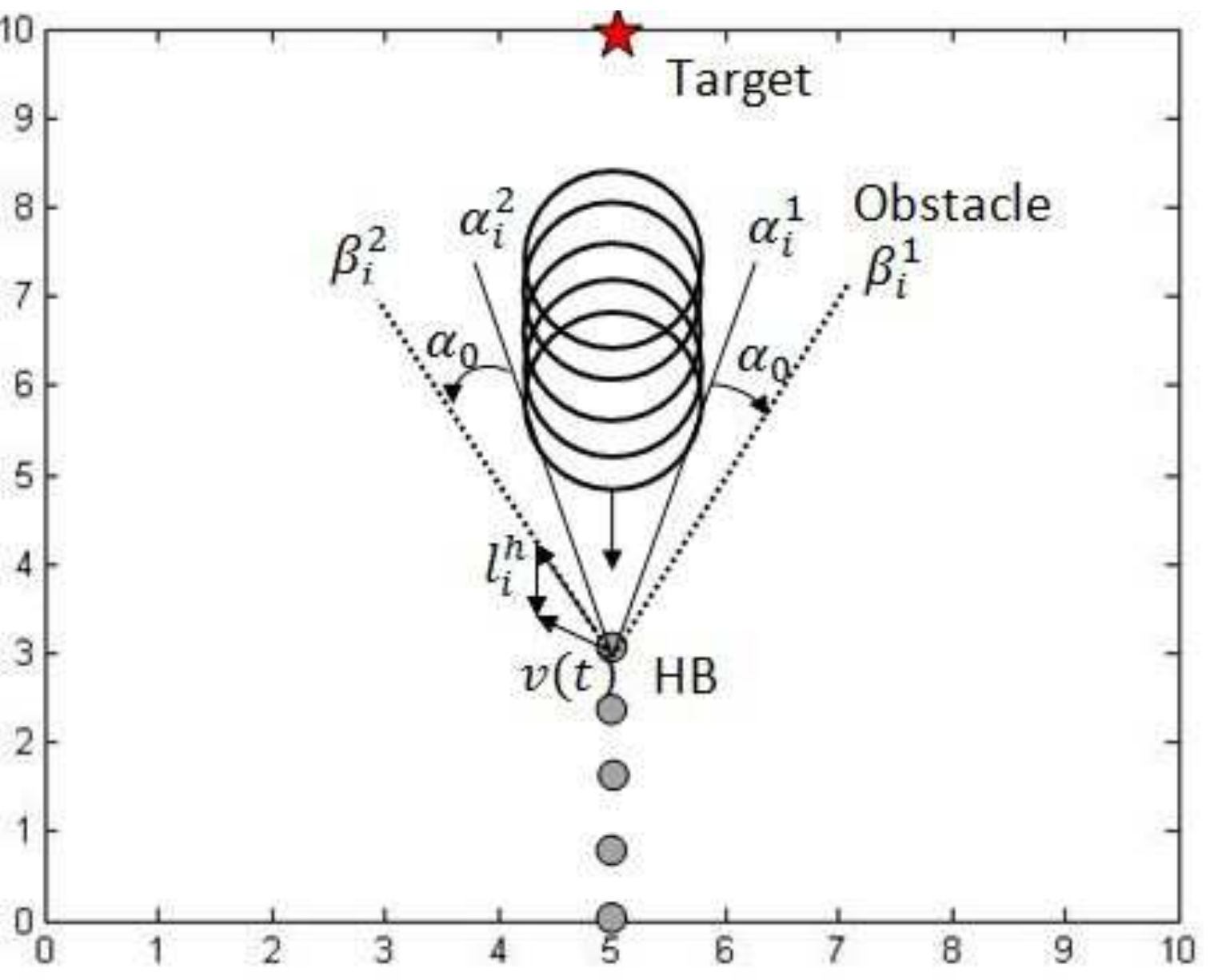}}
		\label{c1.sim11}}
		\subfigure[]{\scalebox{0.4}{\includegraphics{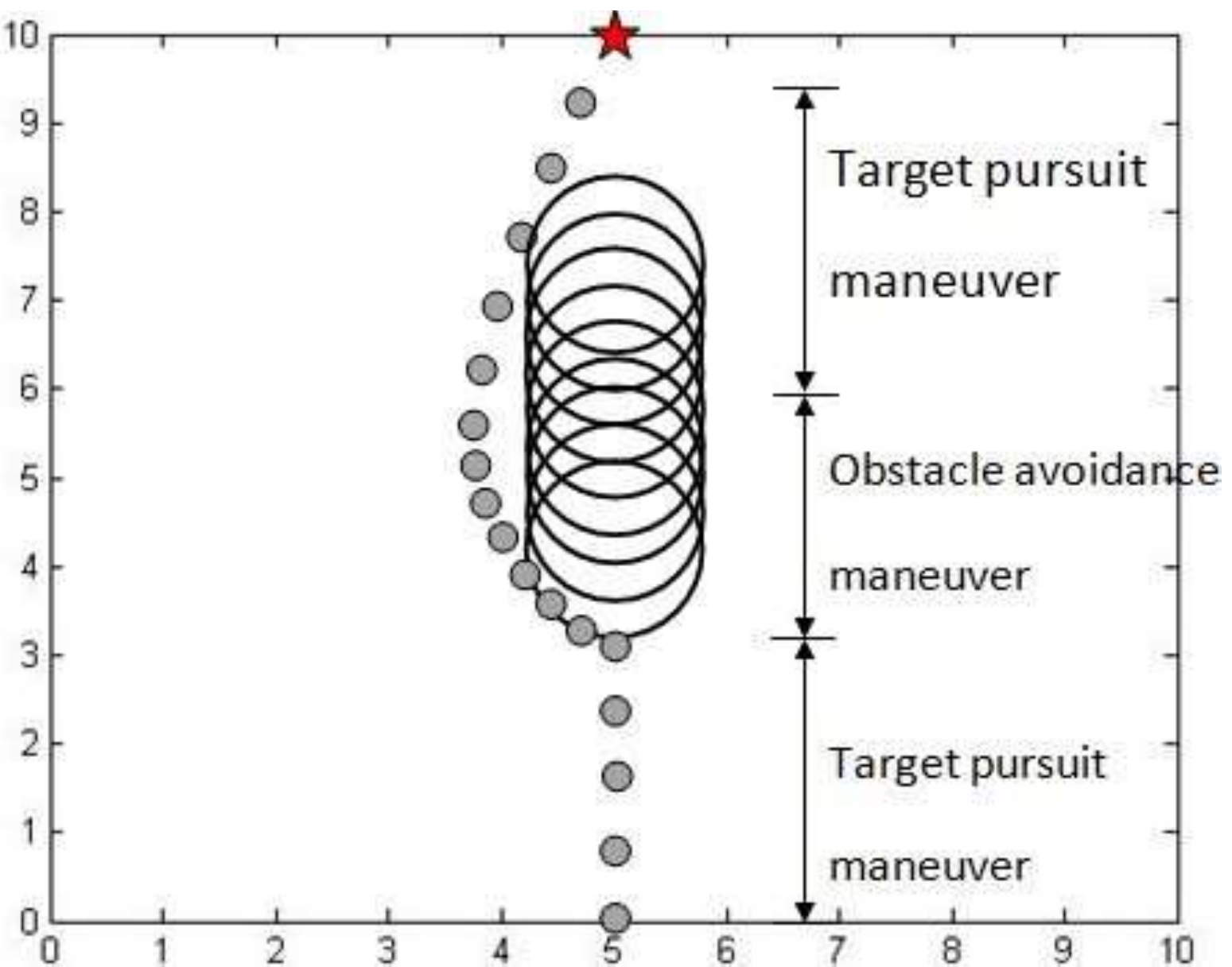}}
		\label{c1.sim12}}
		\subfigure[]{\scalebox{0.38}{\includegraphics{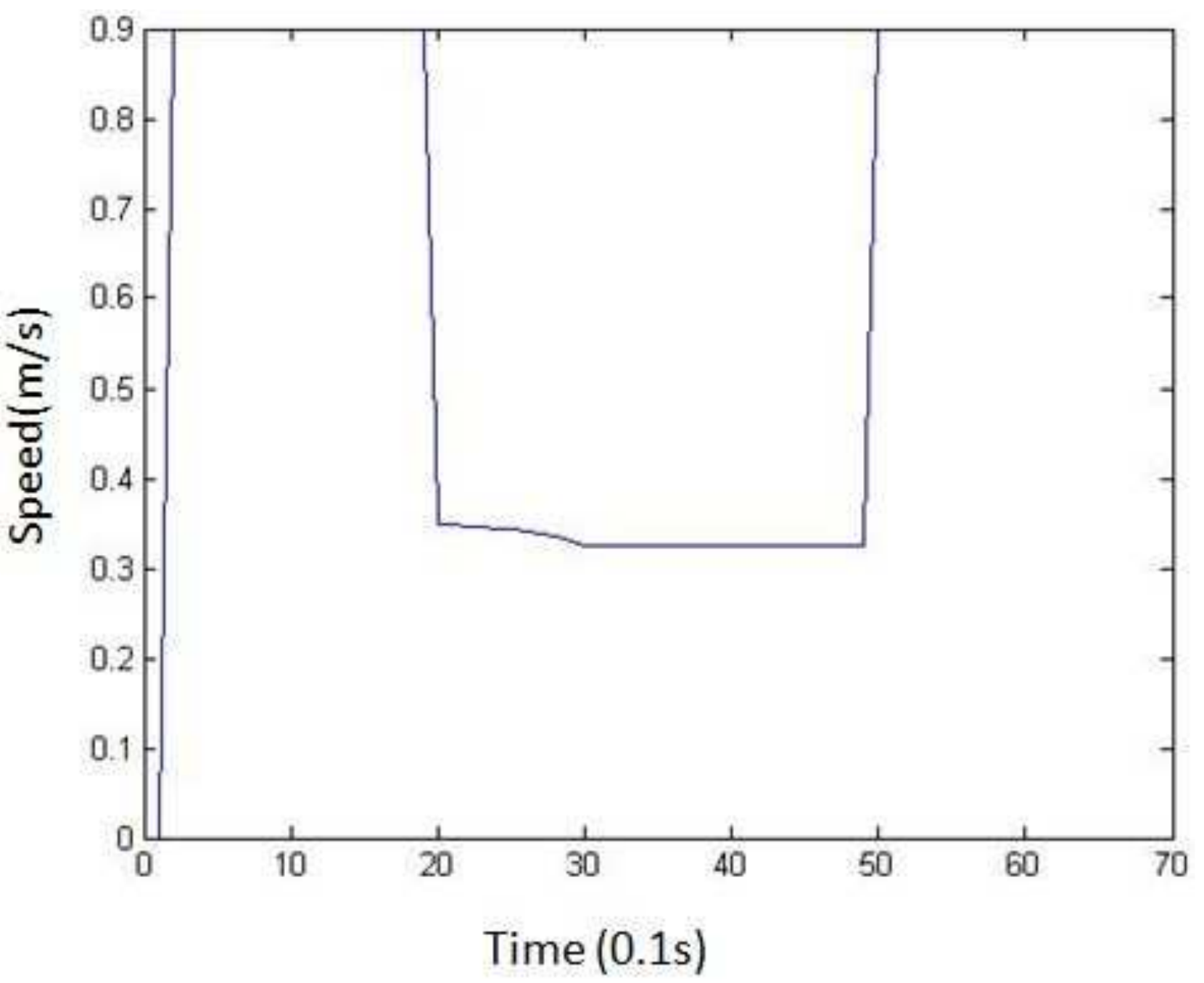}}
		\label{c1.sim13}}
		\subfigure[]{\scalebox{0.38}{\includegraphics{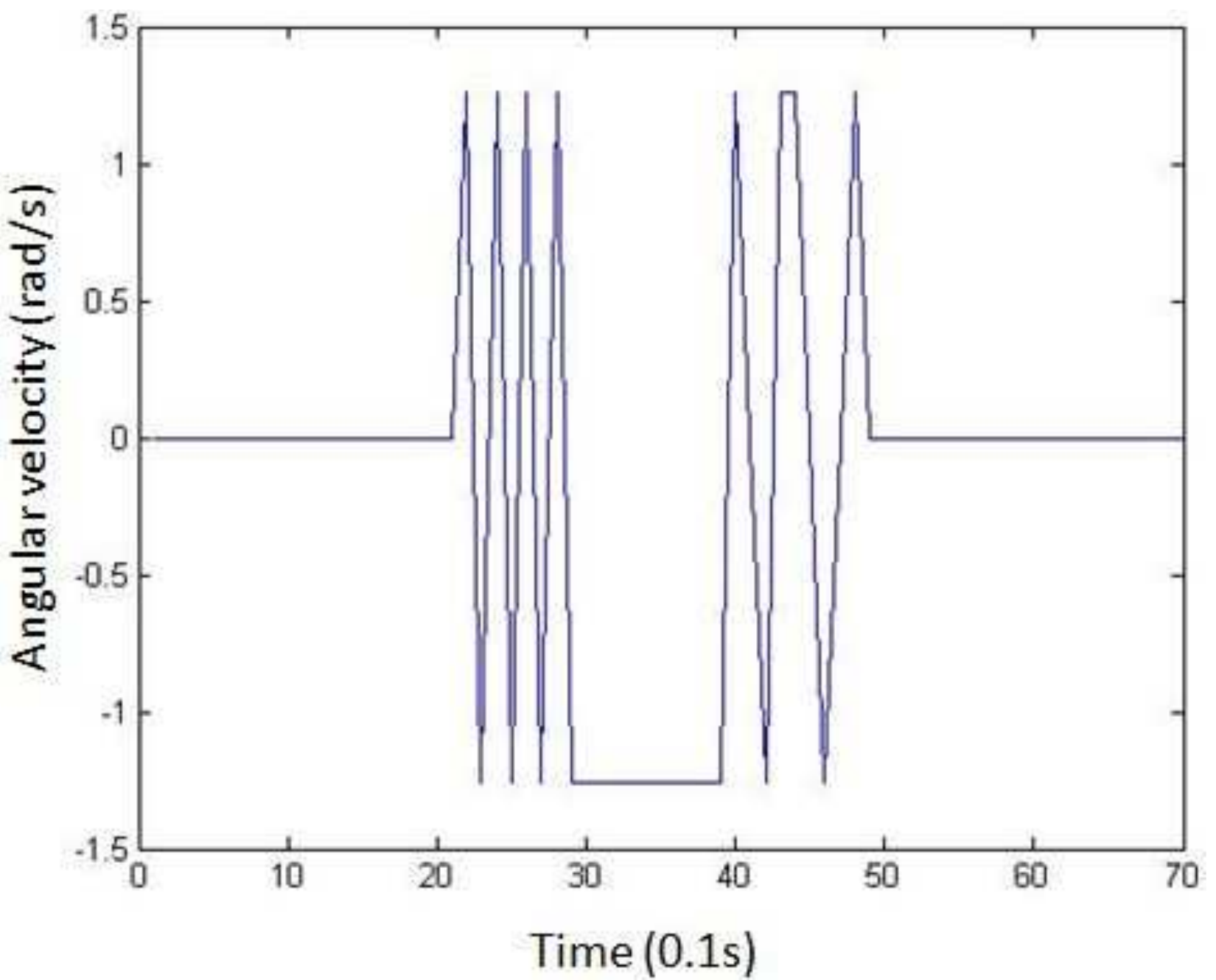}}
		\label{c1.sim14}}
		\caption{Illustration of the proposed navigation algorithm with one obstacle scenario}
		\label{c1.sim1}
		\end{figure}
		
	In Fig.~\ref{c1.sim1}, we demonstrate the execution of the proposed navigation algorithm with a simple scenario. The mobile robot measures the vision cone ($\alpha_i^{(1)}$, $\alpha_i^{(2)}$) from itself to the obstacle, calculates the necessary parameters ($\beta_i^{(1)}$, $\beta_i^{(2)}$, $l_i^{(h)}$) by (\ref{enlarged_vc}), (\ref{vector}), (\ref{bi_mini}) and the control signal by (\ref{bi_cont}), see Fig.~\ref{c1.sim11}. The mobile robot safely arrives the target location while avoiding the moving obstacle with the proper switching between (\ref{bi_cont}) and (\ref{bi_pur}), the indication of different maneuvers taken by the mobile robot during its path to the target location are shown in Fig.~\ref{c1.sim12}. The speed and angular velocity of the mobile robot is shown in Fig.~\ref{c1.sim13} and Fig.~\ref{c1.sim14}, respectively.
\par
		\begin{figure}[!h]
		\centering
		\subfigure[]{\scalebox{0.33}{\includegraphics{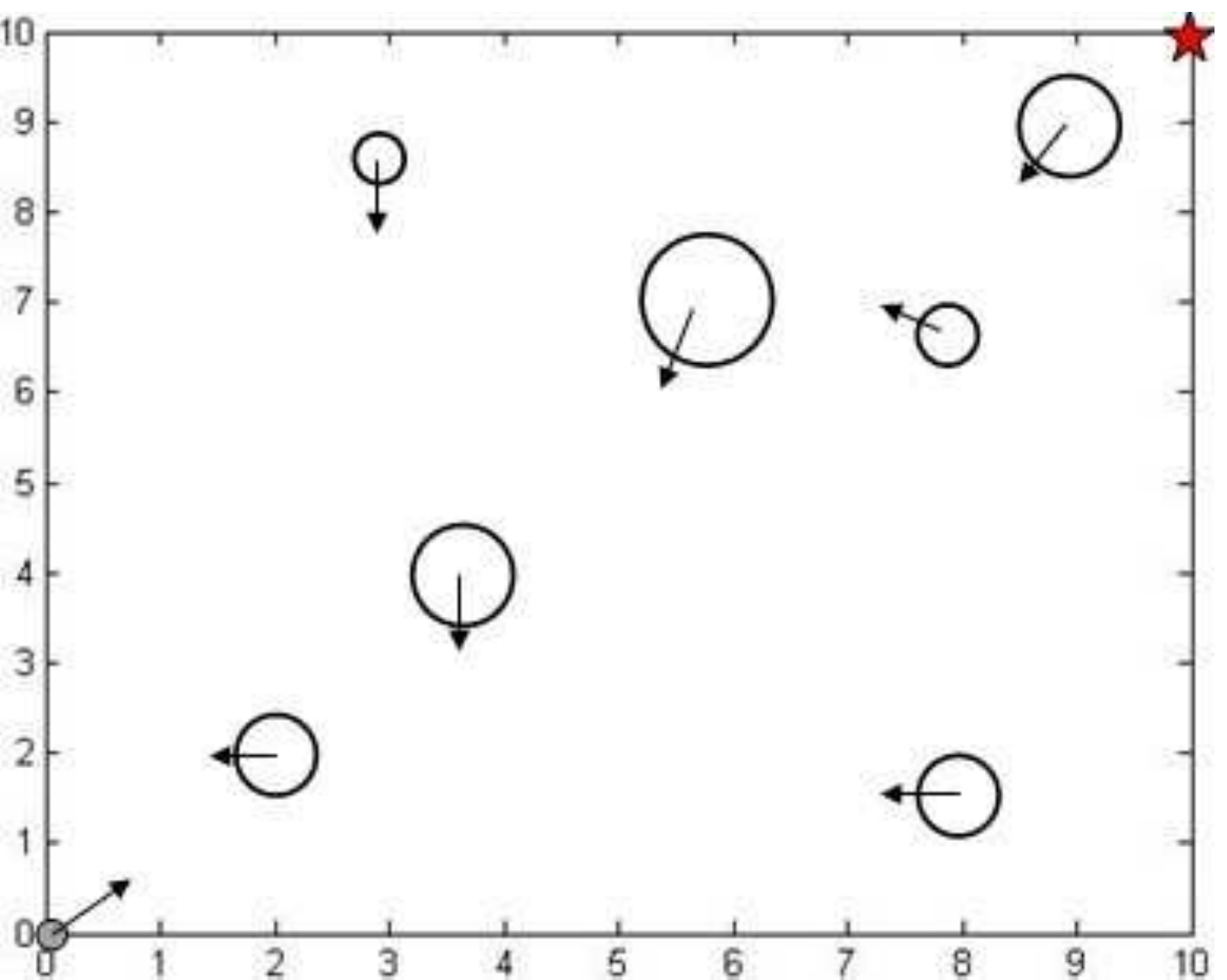}}
		\label{c1.sim21}}
		\subfigure[]{\scalebox{0.33}{\includegraphics{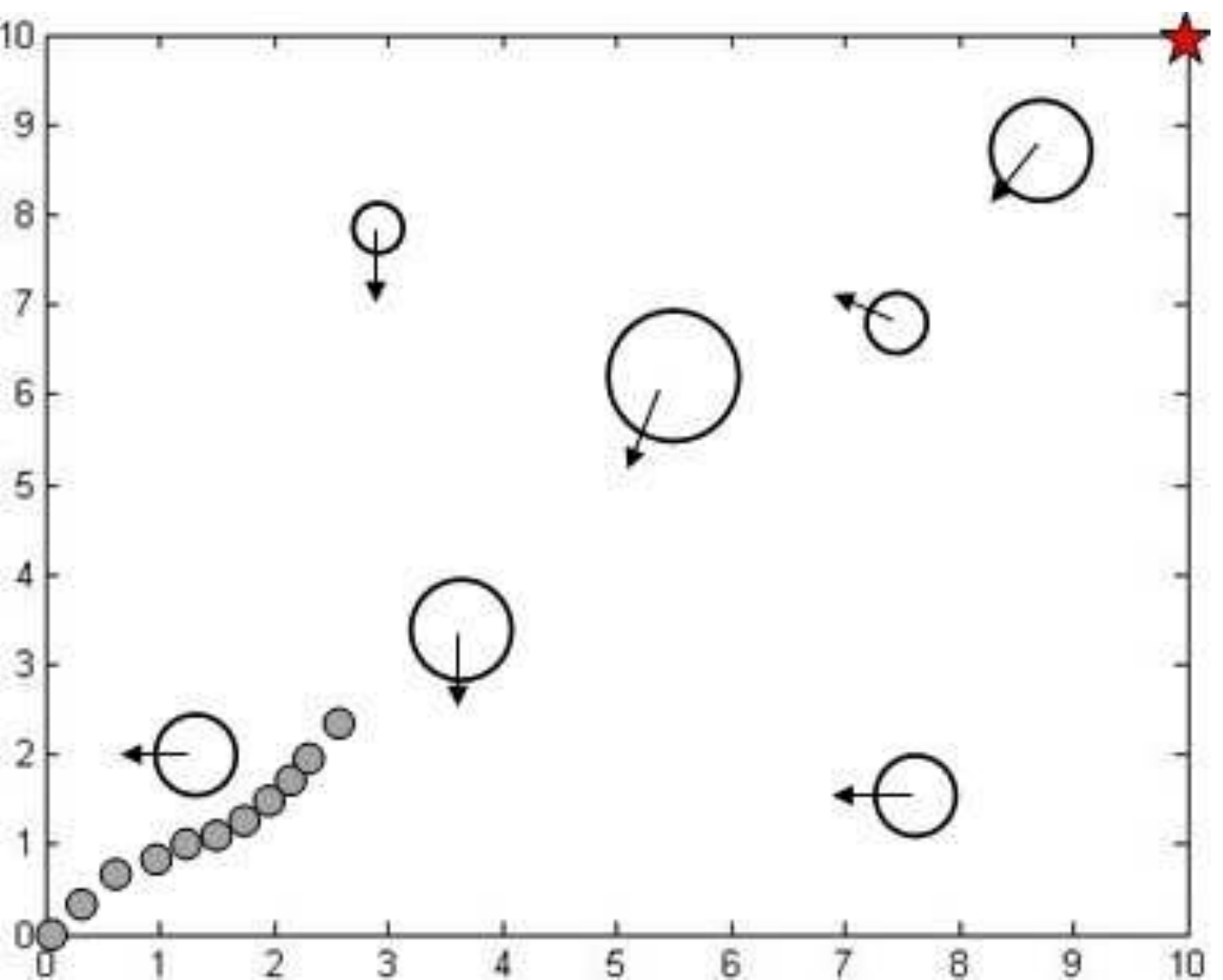}}
		\label{c1.sim22}}
		\subfigure[]{\scalebox{0.33}{\includegraphics{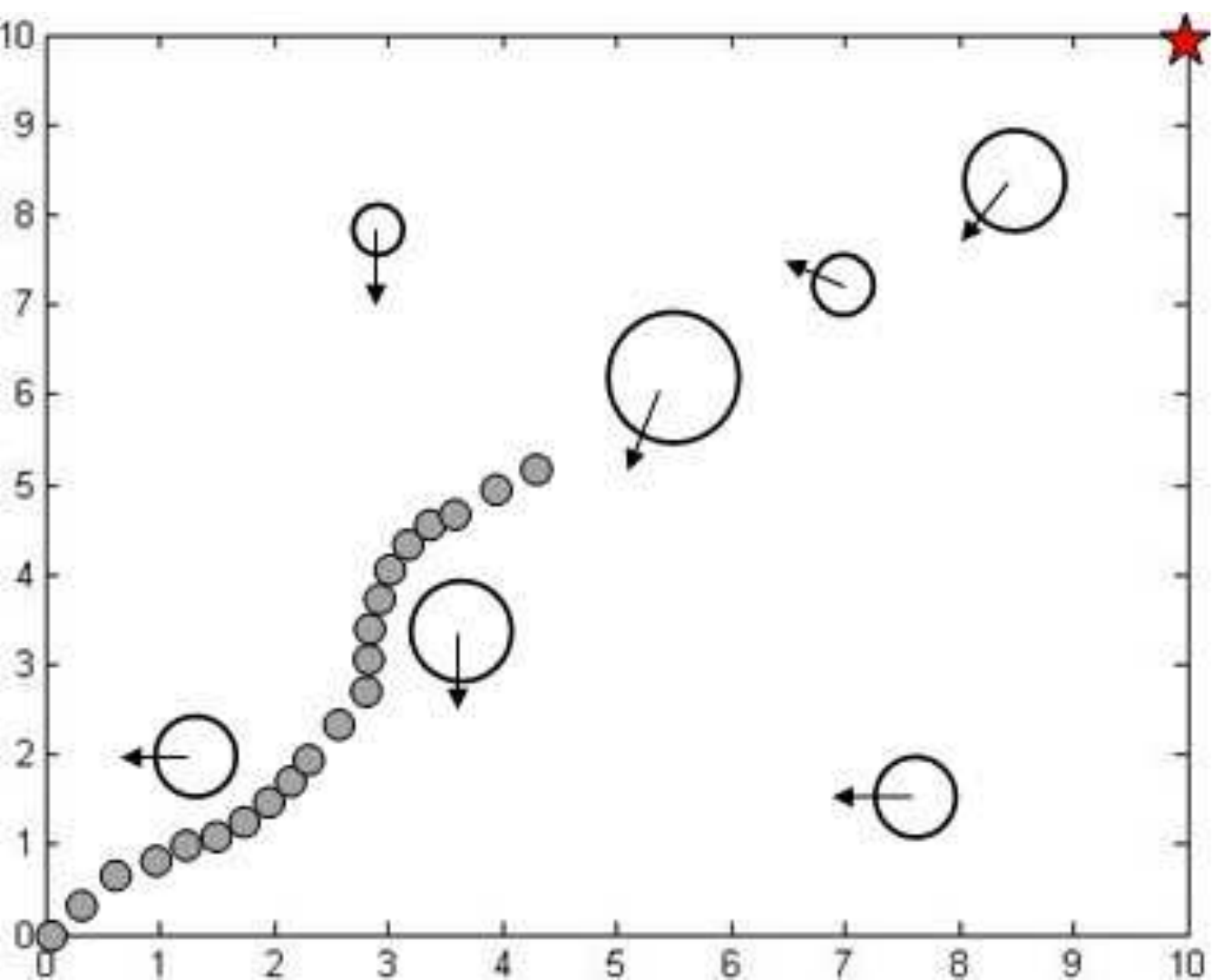}}
		\label{c1.sim23}}
		\subfigure[]{\scalebox{0.33}{\includegraphics{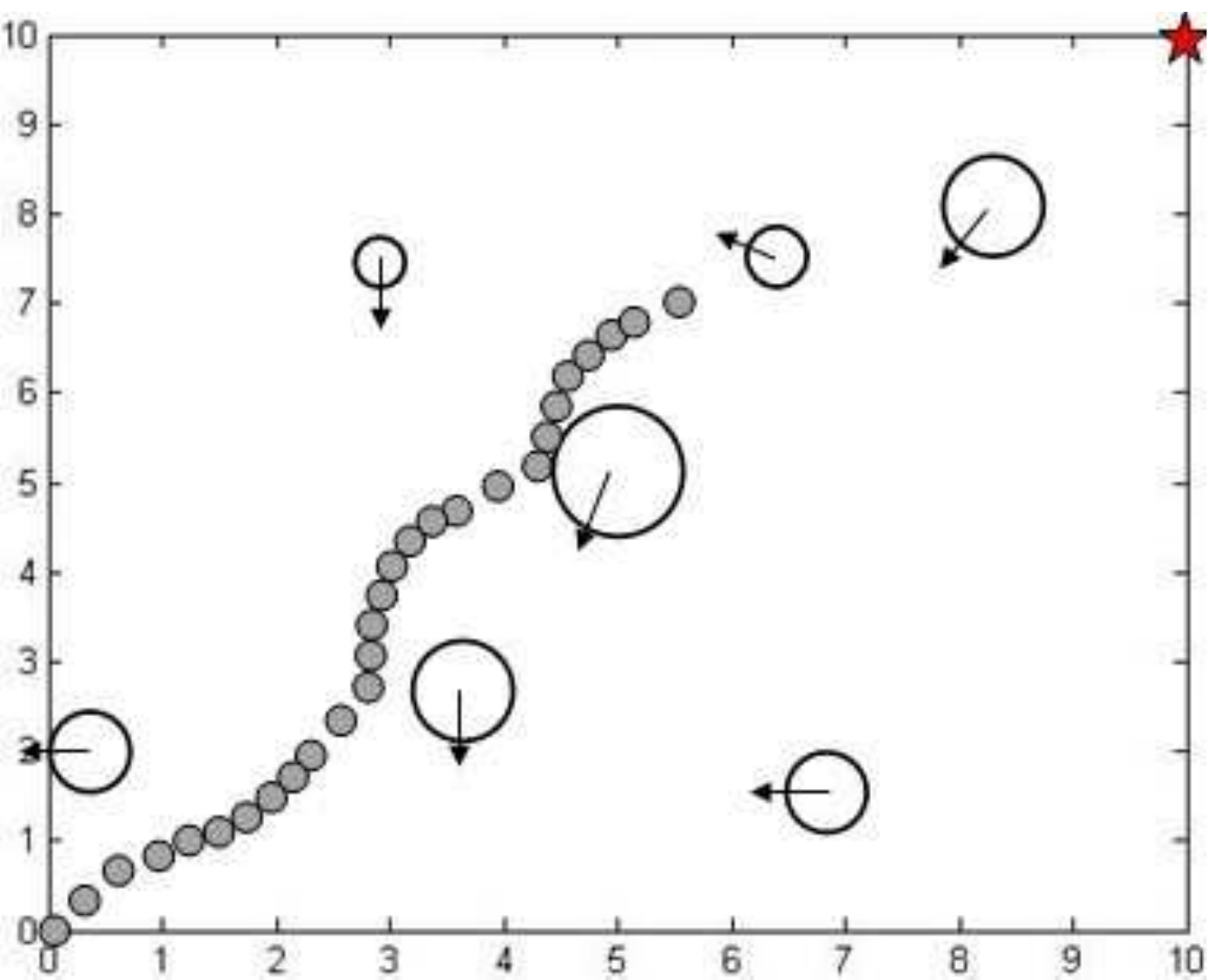}}
		\label{c1.sim24}}
		\subfigure[]{\scalebox{0.33}{\includegraphics{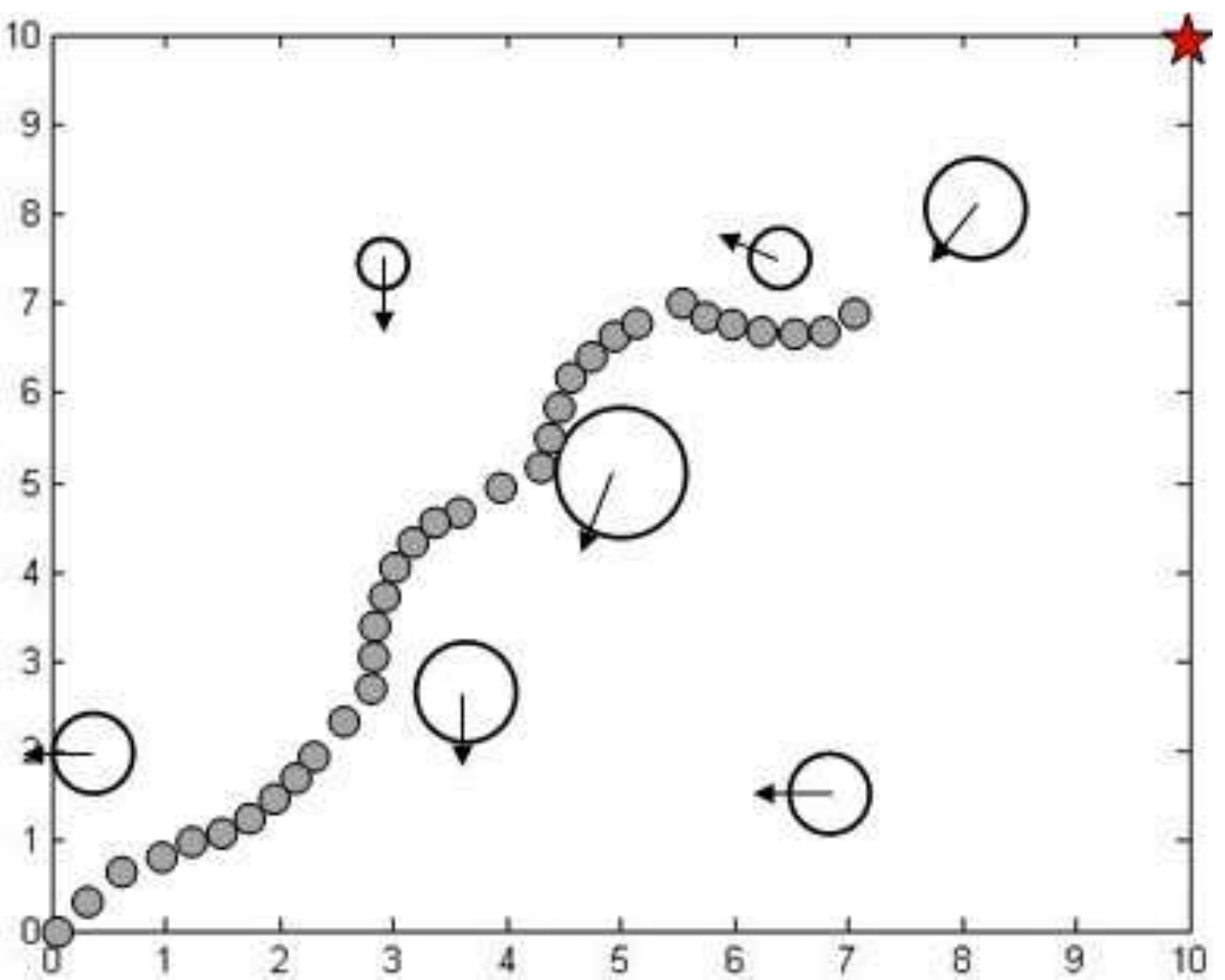}}		
		\label{c1.sim25}}
		\subfigure[]{\scalebox{0.33}{\includegraphics{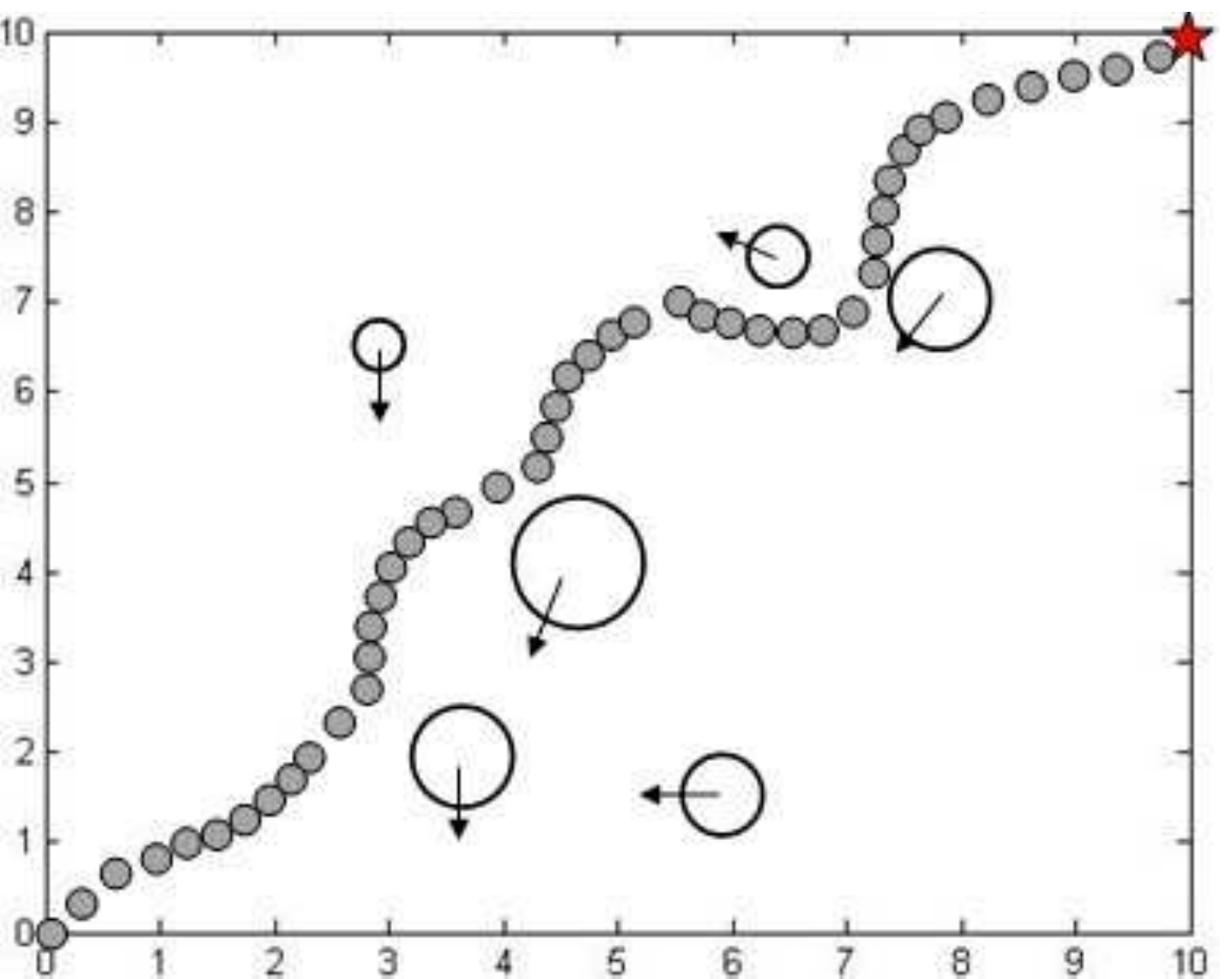}}
		\label{c1.sim26}}
		\subfigure[]{\scalebox{0.33}{\includegraphics{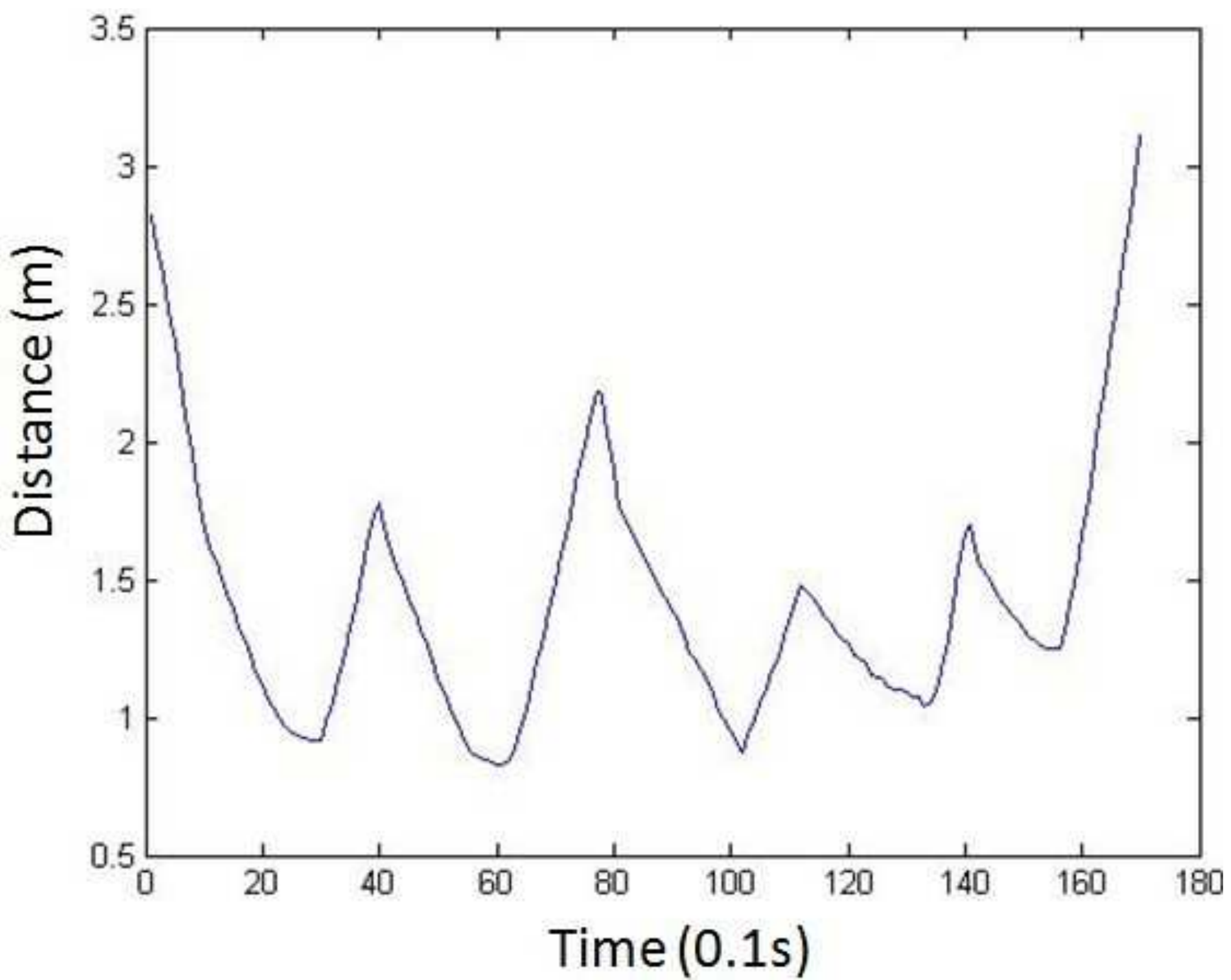}}
		\label{c1.sim2.dis}}
		\caption{Mobile robot navigating in a cluttered environment with multiple moving obstacle}
		\label{c1.sim2}
		\end{figure}

	The next simulation shows the performance of the proposed navigation algorithm in a cluttered dynamic environment. This scenario can be treated as an extension of the simple scenario shown in Fig.~\ref{c1.sim1}. In Fig.~\ref{c1.sim2}, the obstacles of different radii are moving at various velocities, the mobile robot treats and bypasses each of these obstacles with a similar fashion shown in Fig.~\ref{c1.sim1}. The proposed navigation algorithm ensures the safety of the mobile robot in this cluttered dynamic environment, and delivers it to the target position. The crucial moments for the mobile robot to avoid each of the obstacles with the proposed obstacle avoidance law (\ref{bi_cont}) are shown in Fig.~\ref{c1.sim22}, Fig.~\ref{c1.sim23}, Fig.~\ref{c1.sim24}, Fig.~\ref{c1.sim25}, and the overall path taken by the mobile robot is shown in Fig.~\ref{c1.sim26}. The safety of the mobile robot is confirmed by the distance between itself and the closest obstacle which is shown in Fig.~\ref{c1.sim2.dis}. The minimum value ($d_{safe}$) of the distance in this simulation is $0.832m$.

\begin{figure}[!h]
		\centering
		\subfigure[]{\scalebox{0.4}{\includegraphics{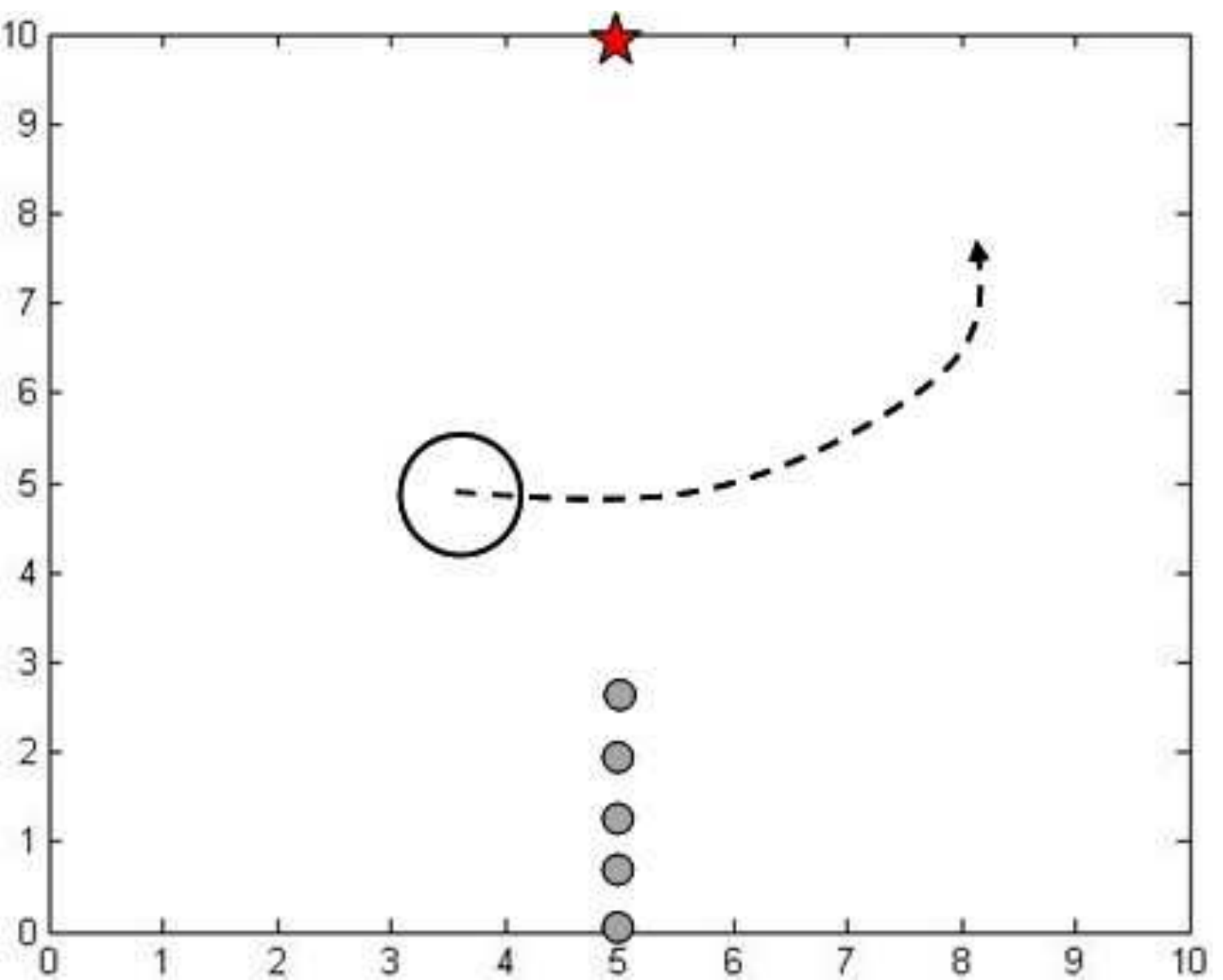}}
		\label{c1.sim31}}
		\subfigure[]{\scalebox{0.4}{\includegraphics{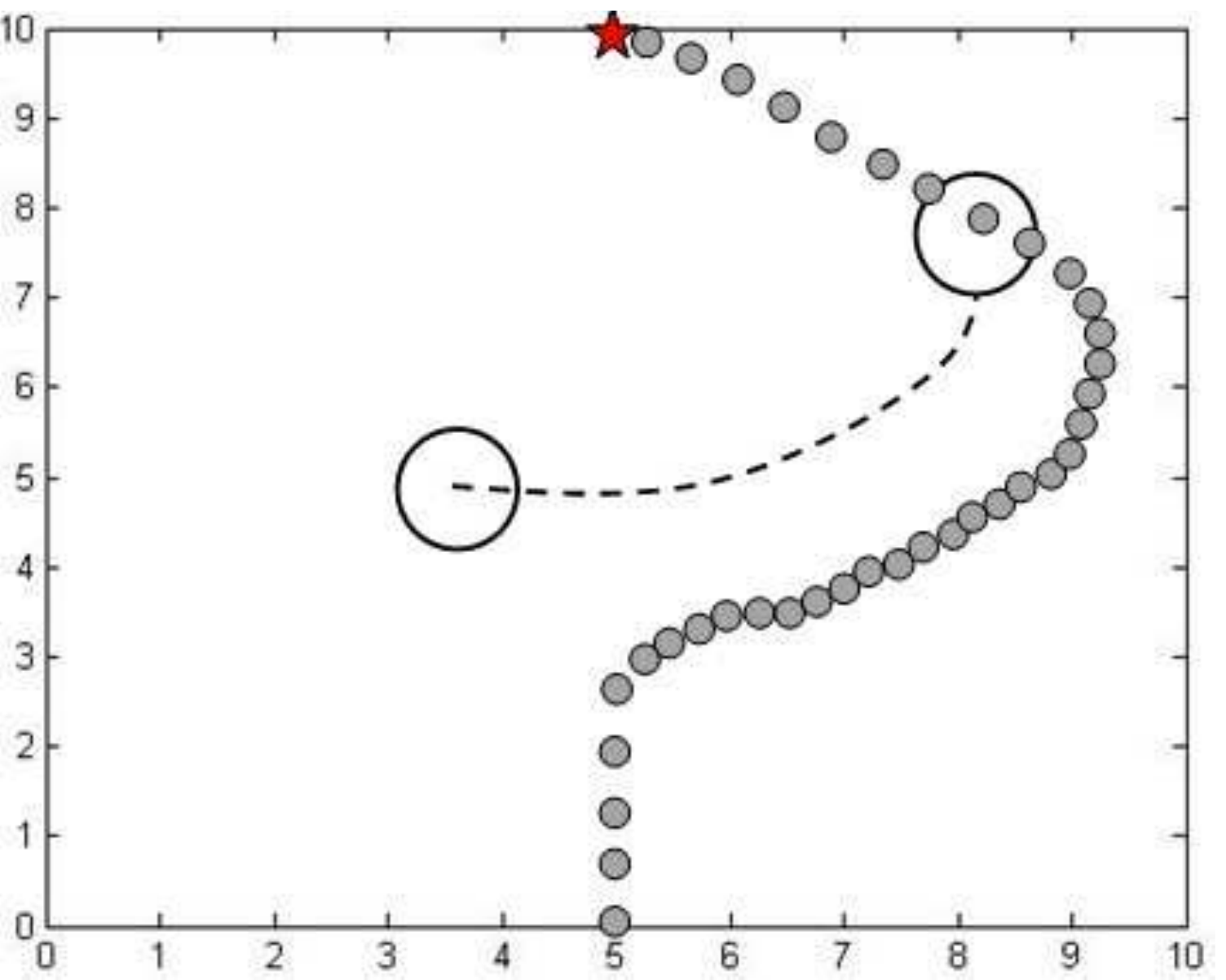}}
		\label{c1.sim32}}
		\caption{Mobile robot avoids an obstacle moving along an arc}
		\label{c1.sim3}
		\end{figure}

\begin{figure}[!h]
		\centering
		\subfigure[]{\scalebox{0.4}{\includegraphics{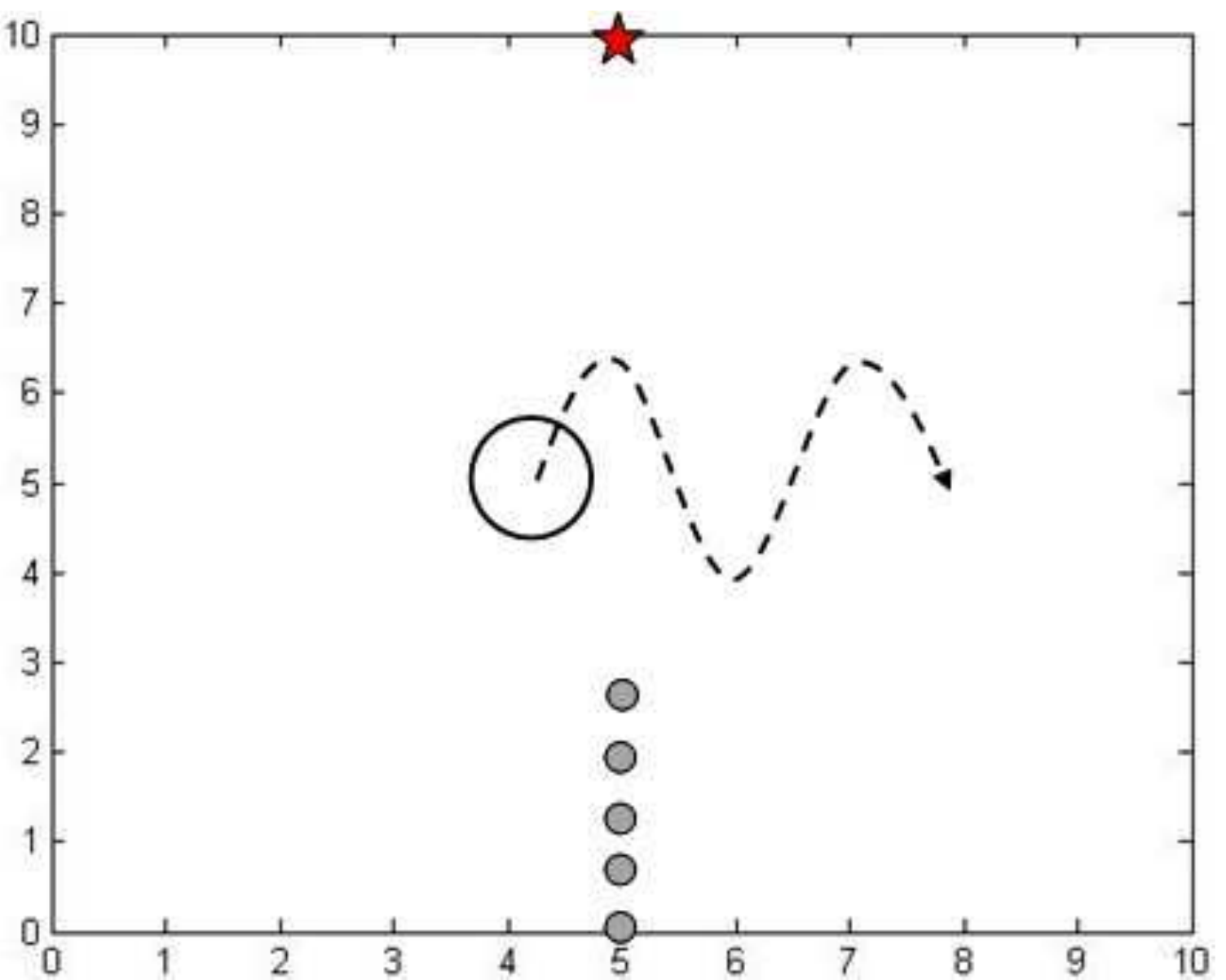}}
		\label{c1.sim41}}
		\subfigure[]{\scalebox{0.4}{\includegraphics{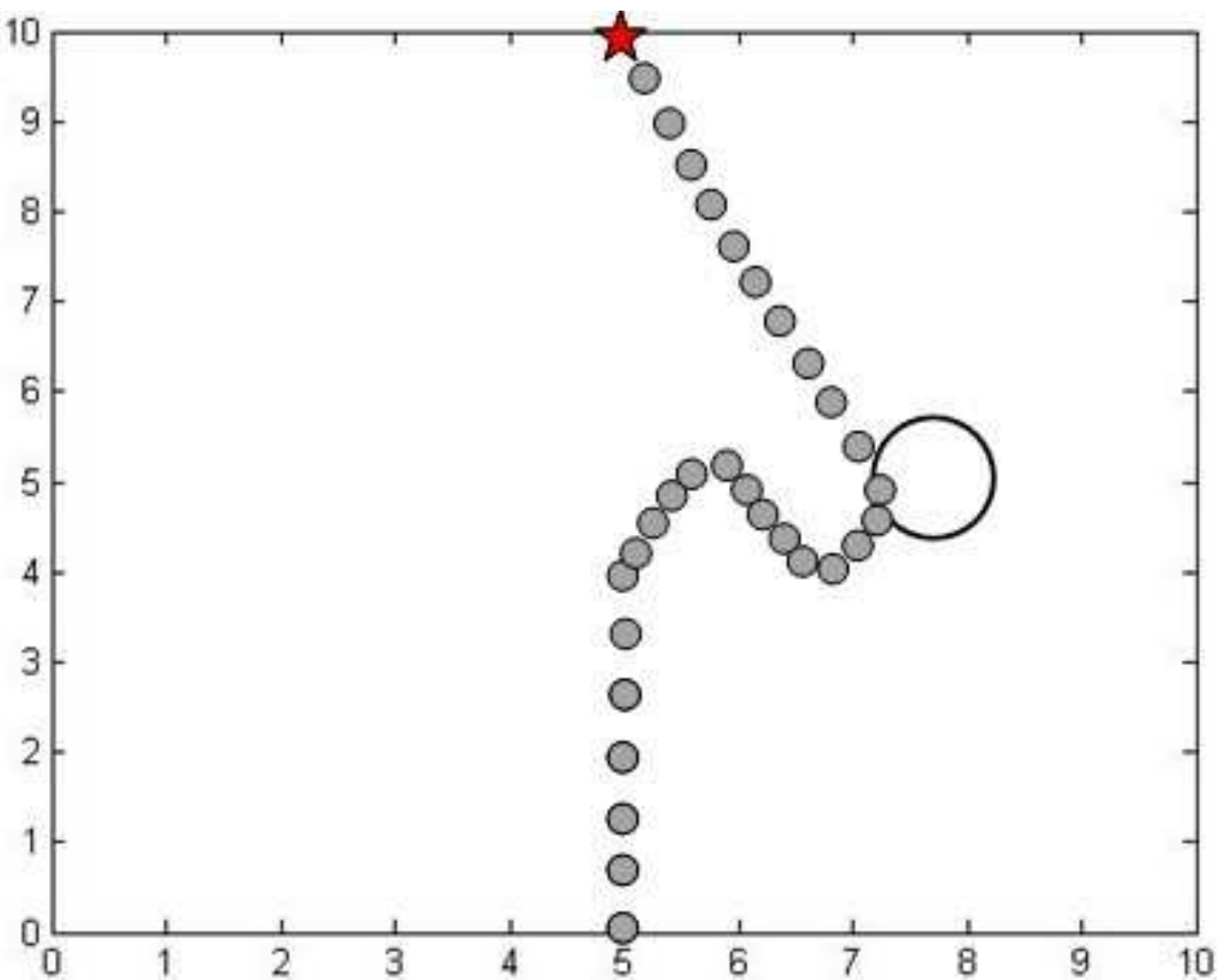}}
		\label{c1.sim42}}
		\caption{Mobile robot avoids an obstacle in a sinusoidal path}
		\label{c1.sim4}
		\end{figure}

	In many real life scenarios, the movements of the obstacles are usually more complicated than simple straight motion. Fig.~\ref{c1.sim3} and Fig.~\ref{c1.sim4} show the avoidance of the obstacles with non-linear velocities by the proposed navigation algorithm. The path taken by the obstacle are depicted in dashed lines, see Fig.~\ref{c1.sim31} and Fig.~\ref{c1.sim41}. The proposed navigation algorithm takes account of the non-linear velocity $v_i(t)$ in (\ref{bi_mini}) and (\ref{bi_cont}), it always steers the mobile robot so that the orientation of mobile robot is $\alpha_0$ degree away from the $l_i^{(h)}$.The overall path taken by the mobile robot is shown in Fig.\ref{c1.sim32} and Fig.\ref{c1.sim42}.
	\par
	
		\begin{figure}[!h]
		\centering
		\subfigure[]{\scalebox{0.40}{\includegraphics{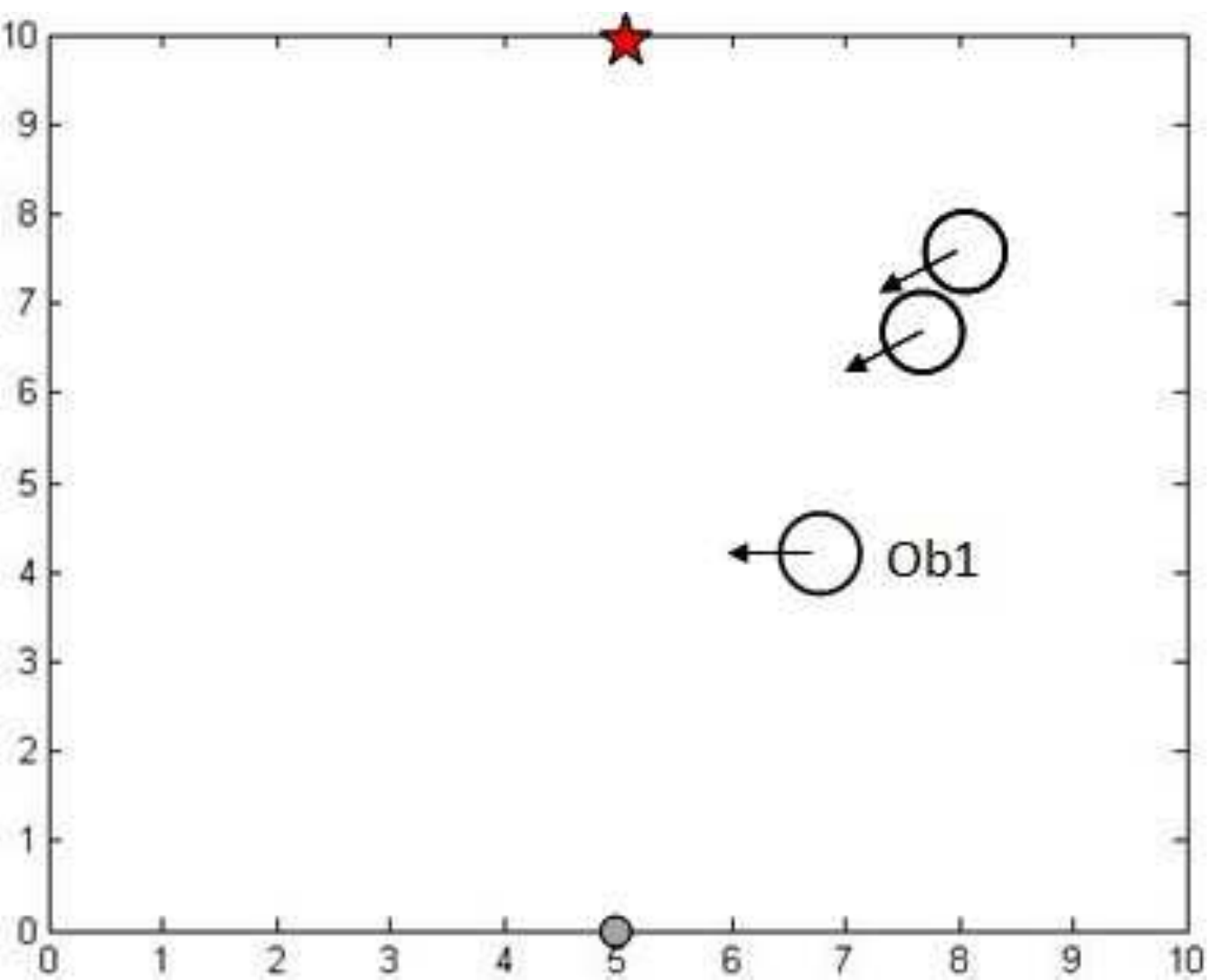}}
		\label{c1.sim51}}		
		\subfigure[]{\scalebox{0.40}{\includegraphics{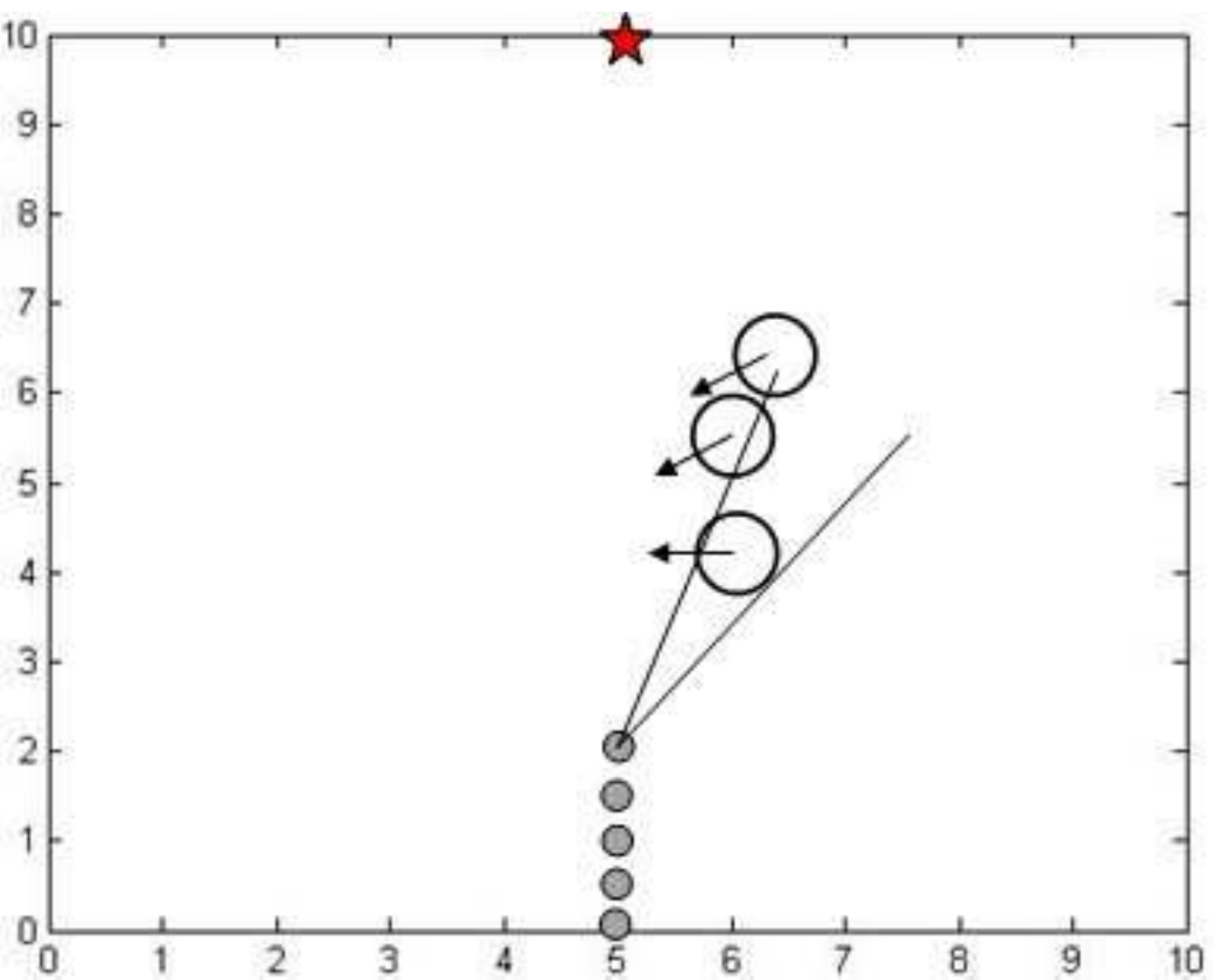}}
		\label{c1.sim52}}
		\subfigure[]{\scalebox{0.40}{\includegraphics{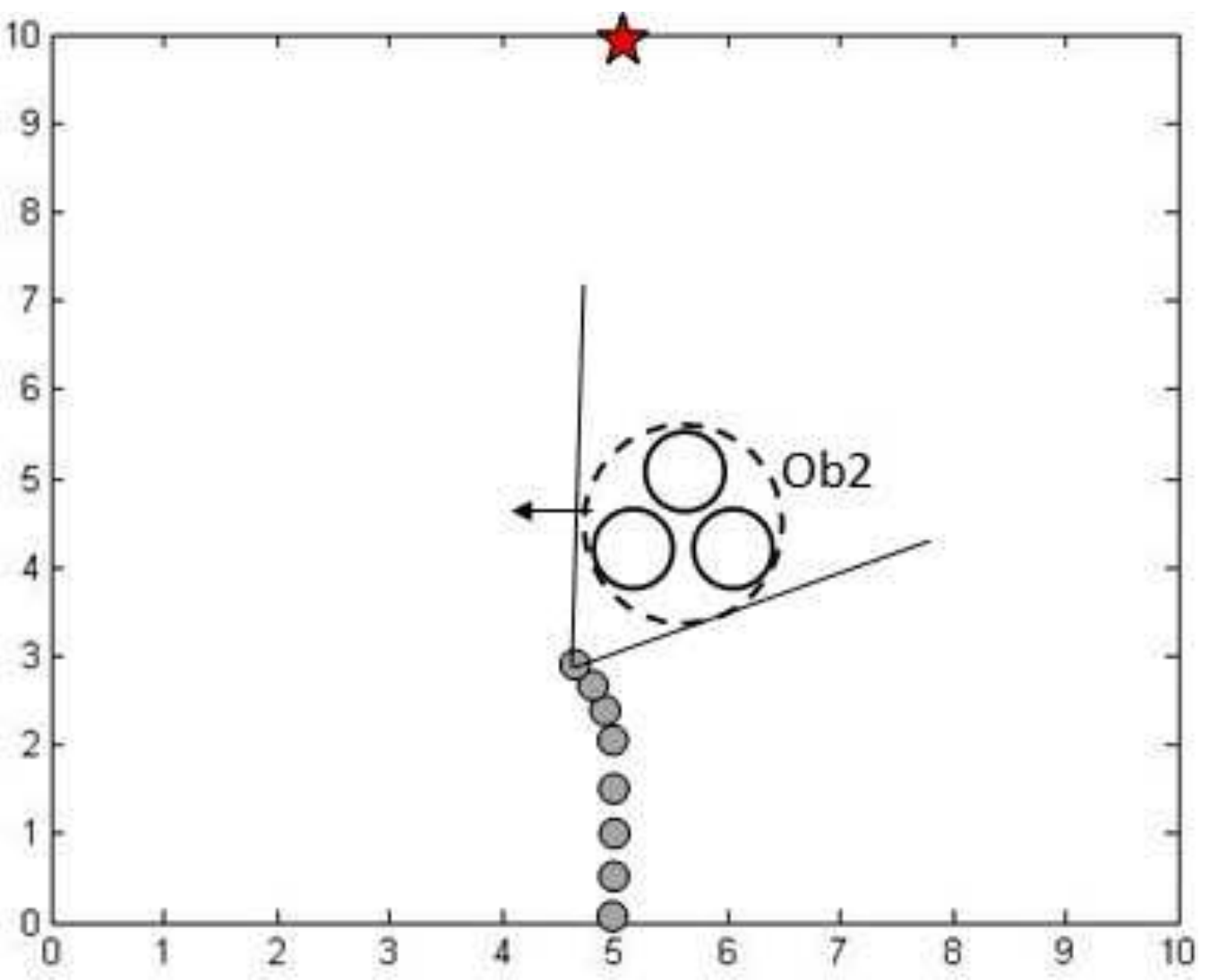}}
		\label{c1.sim53}}
		\subfigure[]{\scalebox{0.40}{\includegraphics{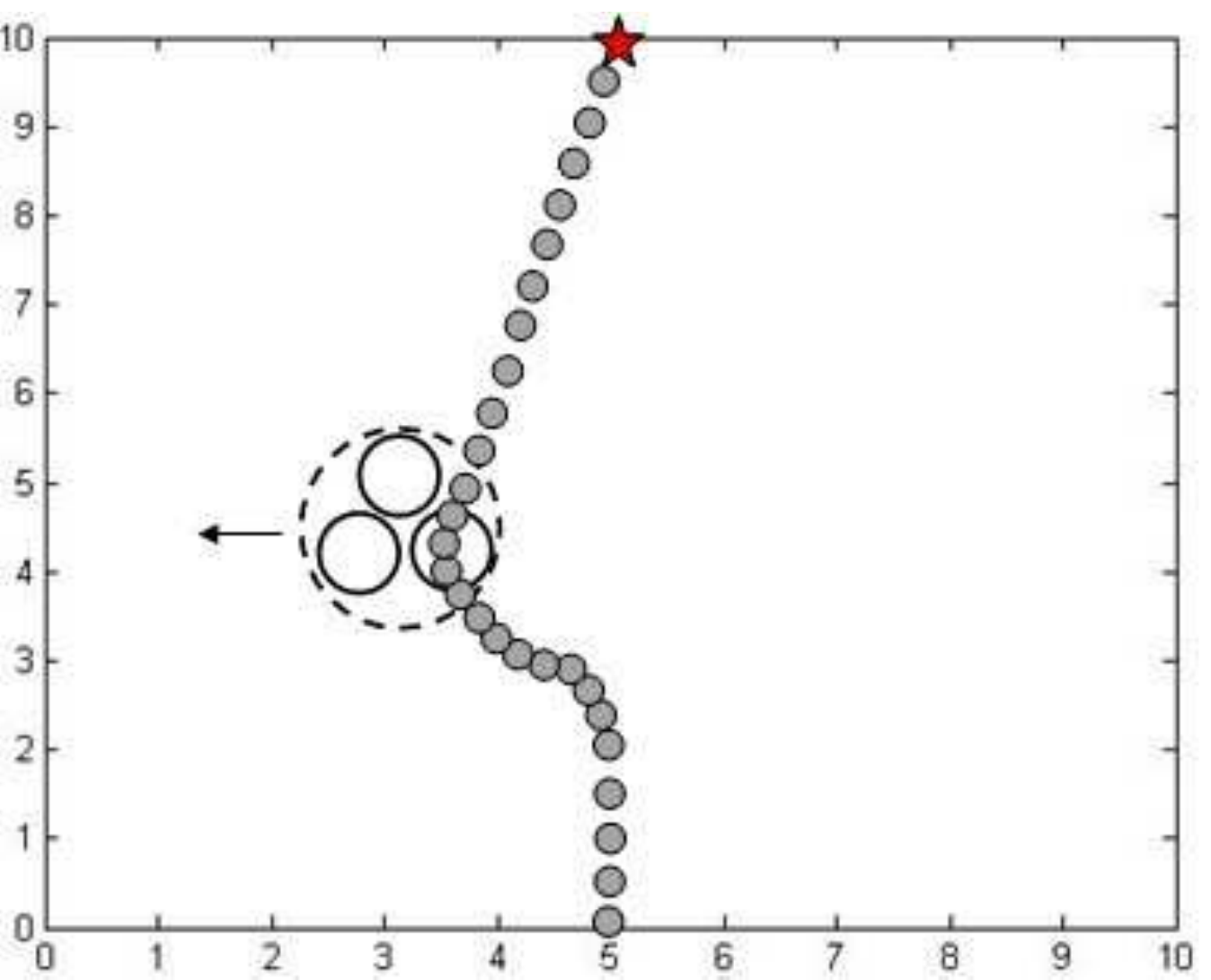}}
		\label{c1.sim54}}
		\caption{Mobile robot avoids an obstacle with dynamically changing radius}
		\label{c1.sim5}
		\end{figure}
	We investigate the performance of the proposed navigation algorithm when the radius of the obstacle are dynamically changing. This is a common behavior especially in workspaces or other populated places, for example, when the mobile robot sensors one obstacle within the switching distance $C$ and starts to avoid it, at the meantime, two other obstacles approaches the first obstacle and forms a group of three obstacles with larger radius. Fig.~\ref{c1.sim5} shows a simulation of such scenario where the radius of the obstacle are dynamically changing. The change of radius of the obstacle "seen" by the mobile robot are shown in Fig.~\ref{c1.sim52} and Fig.~\ref{c1.sim53}.  The mobile robot arrives the target location safely in Fig.\ref{c1.sim54} and its path is depicted.
\par

		\begin{figure}[H]
		\centering
		\subfigure[]{\scalebox{0.33}{\includegraphics{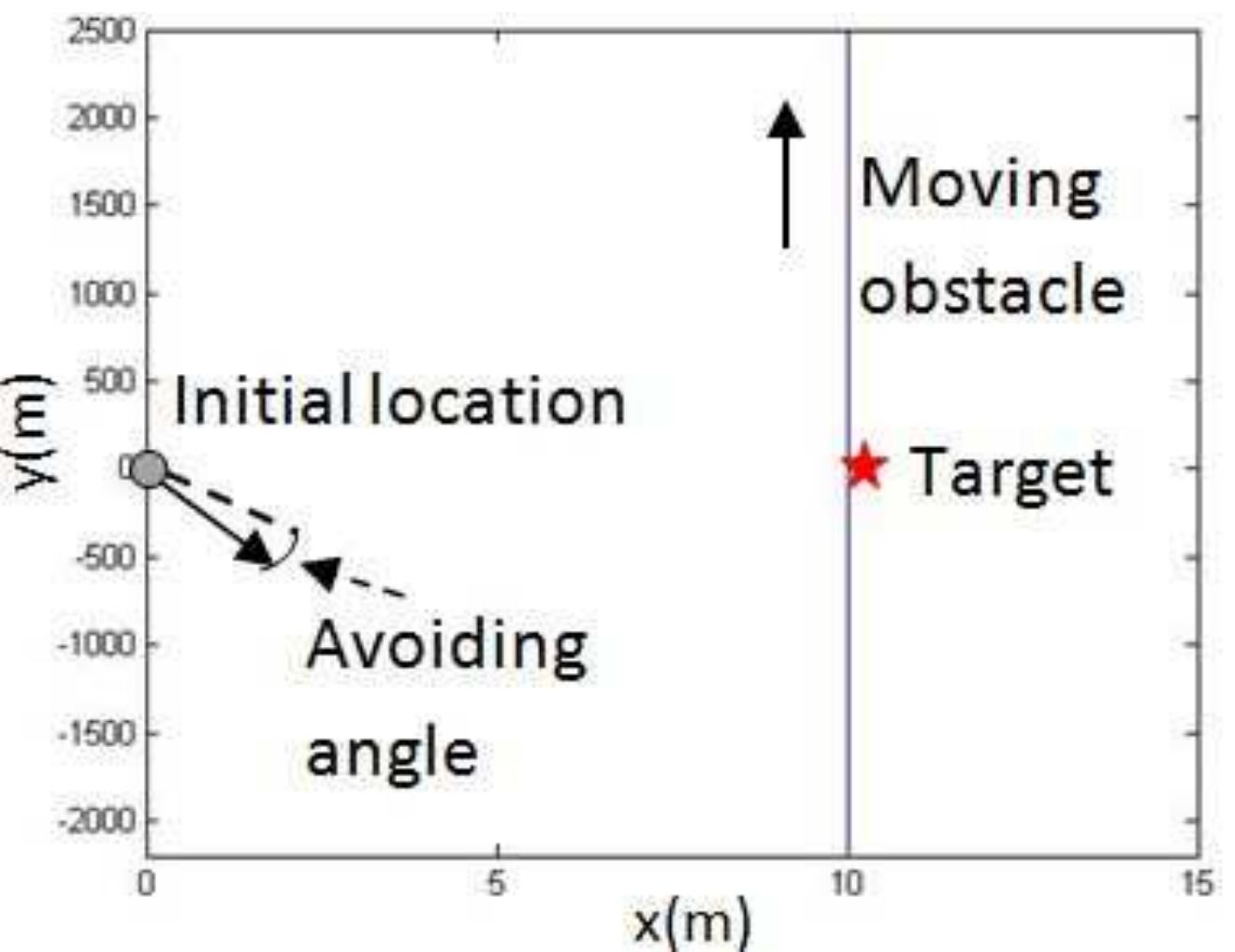}}
		\label{exp3_11.fig}}
		\subfigure[]{\scalebox{0.33}{\includegraphics{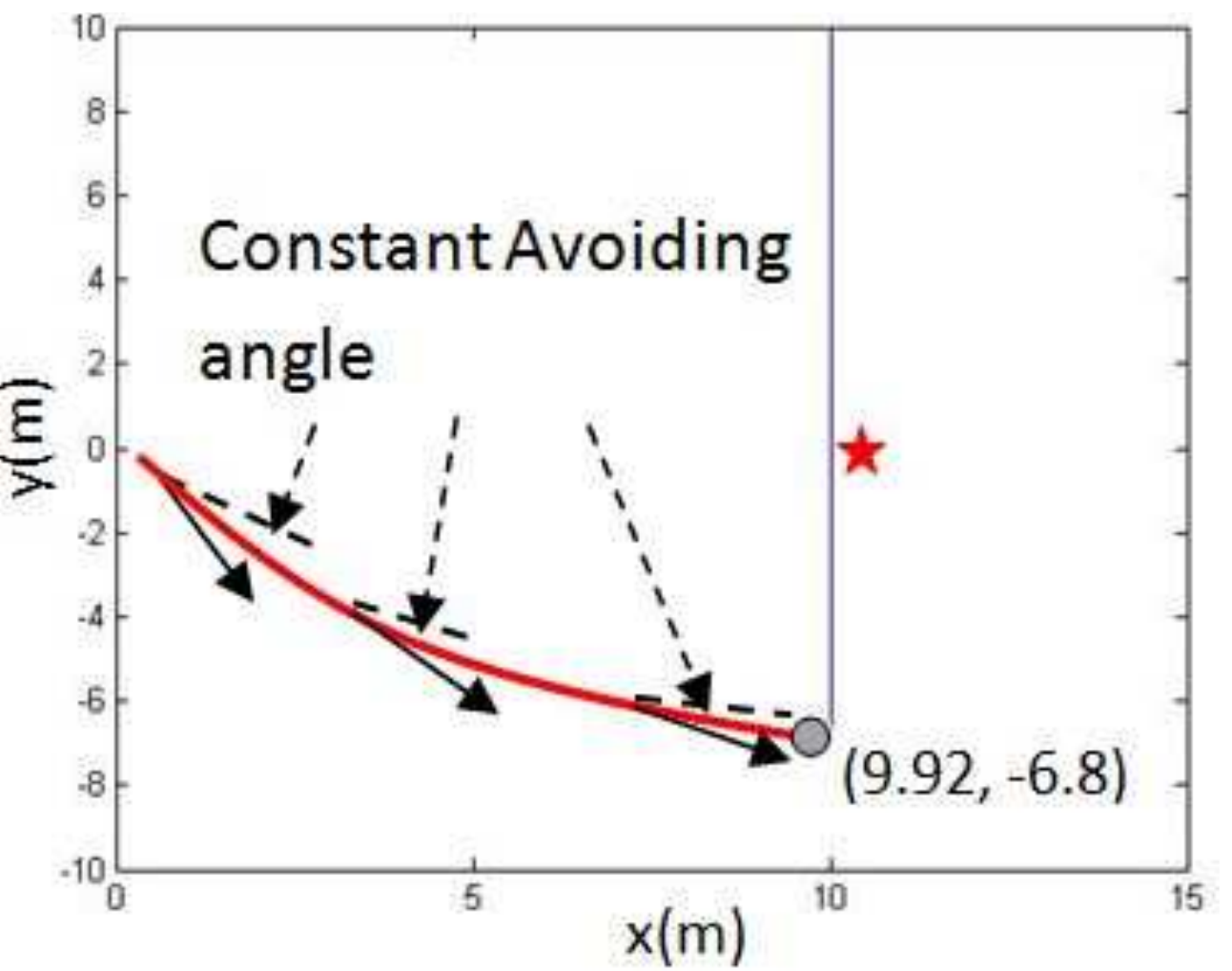}}
		\label{exp3_22.fig}}
		\subfigure[]{\scalebox{0.33}{\includegraphics{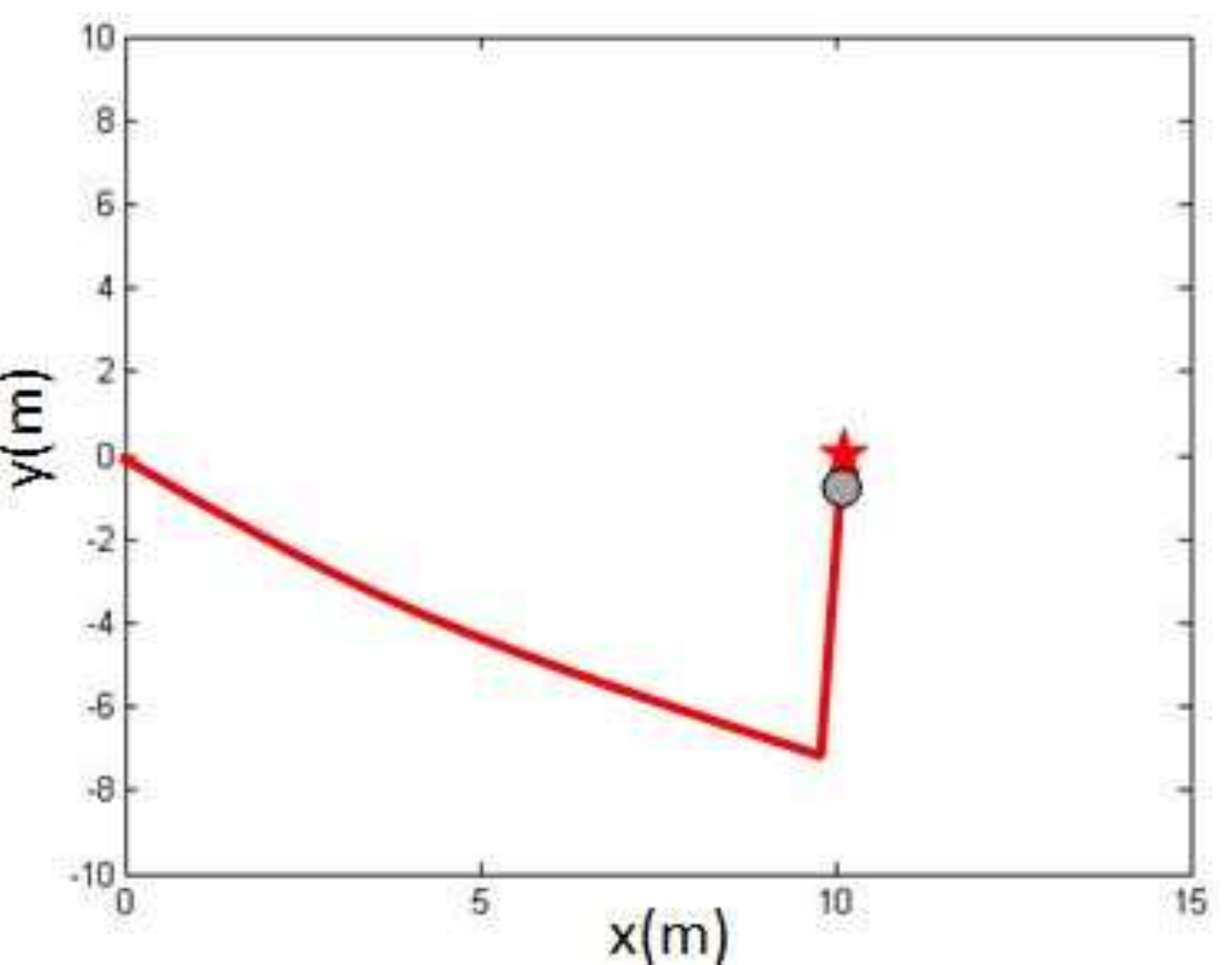}}
		\label{exp3_33.fig}}
		\subfigure[]{\scalebox{0.33}{\includegraphics{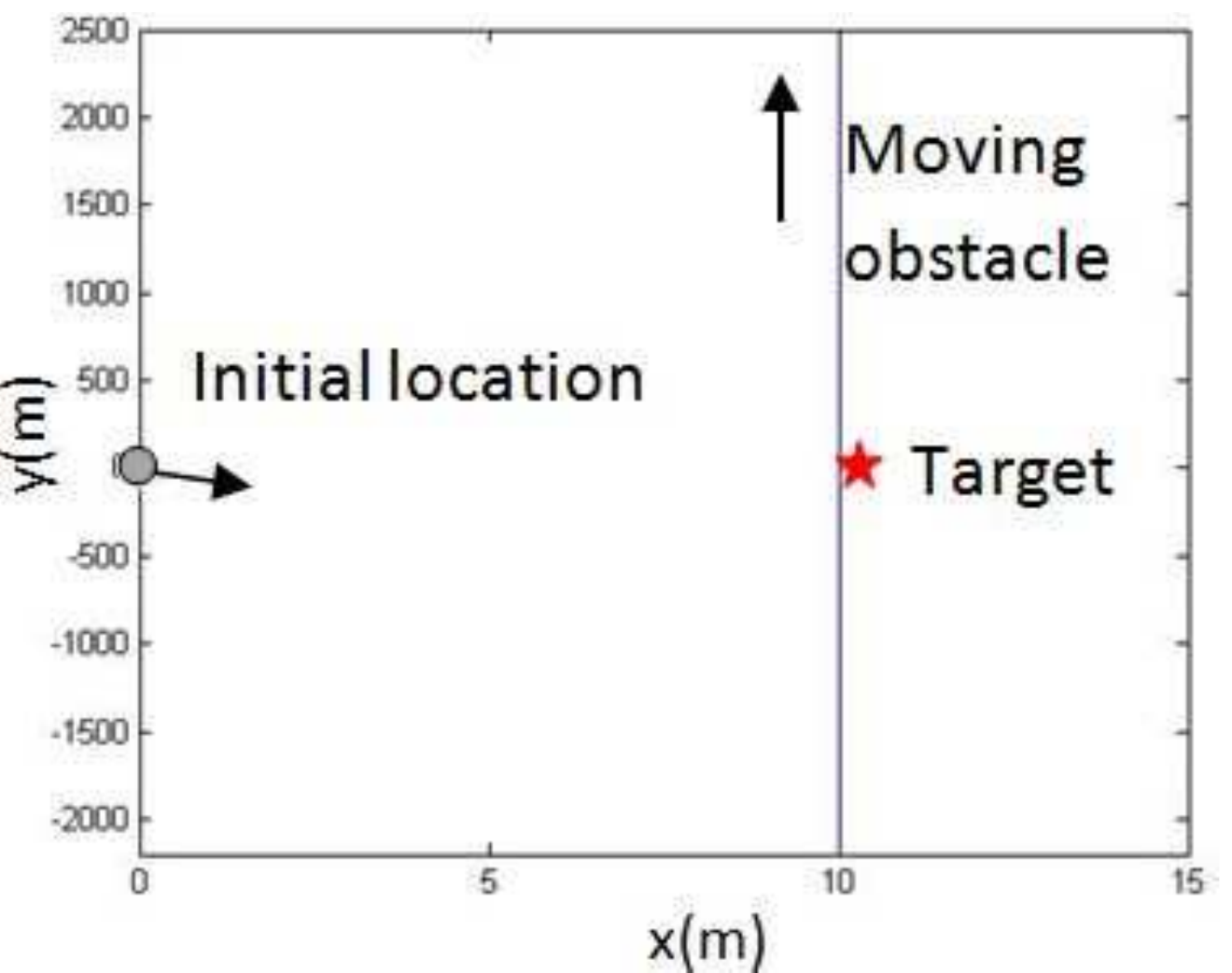}}
		\label{exp3_44.fig}}
		\subfigure[]{\scalebox{0.33}{\includegraphics{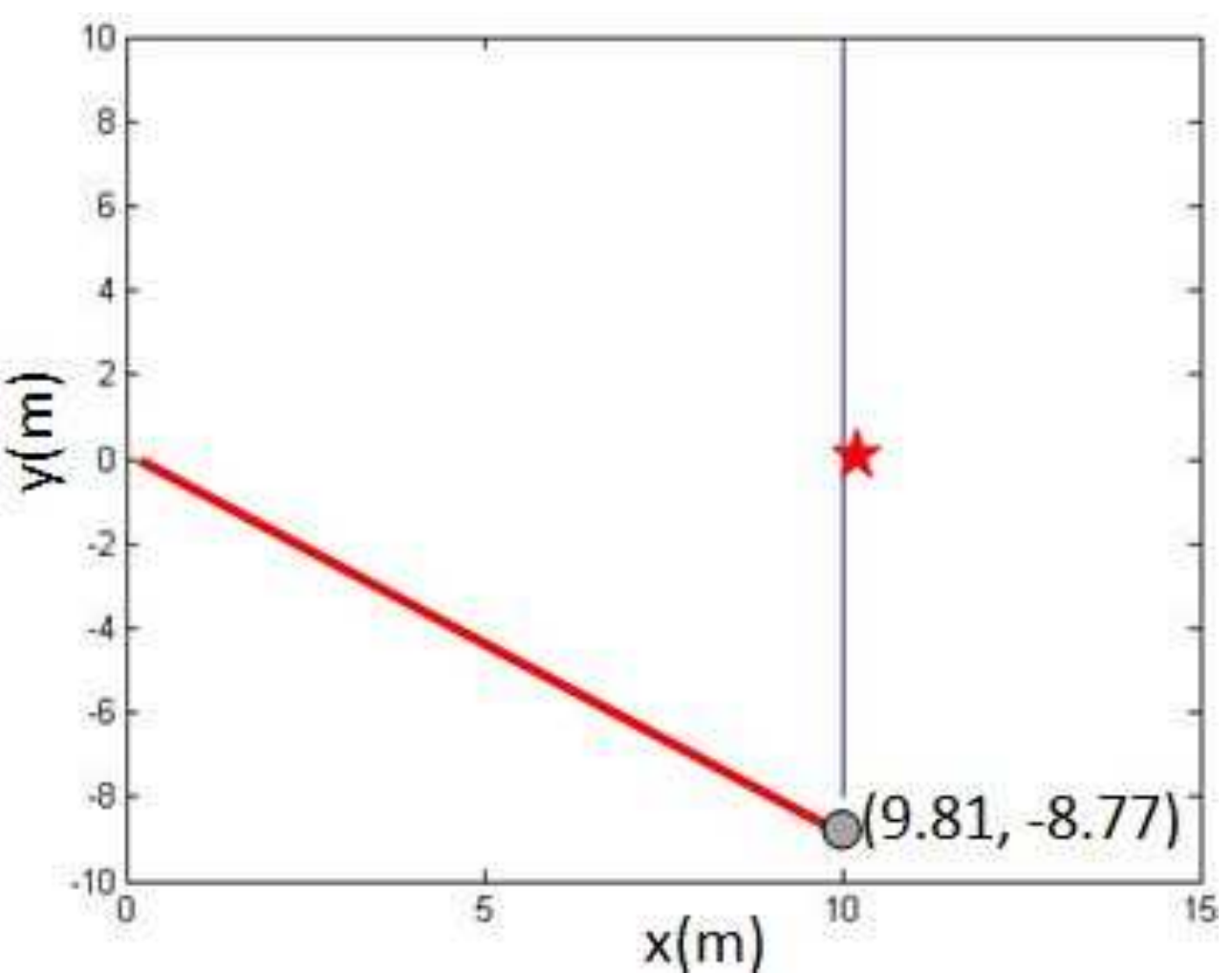}}
		\label{exp3_55.fig}}
		\subfigure[]{\scalebox{0.33}{\includegraphics{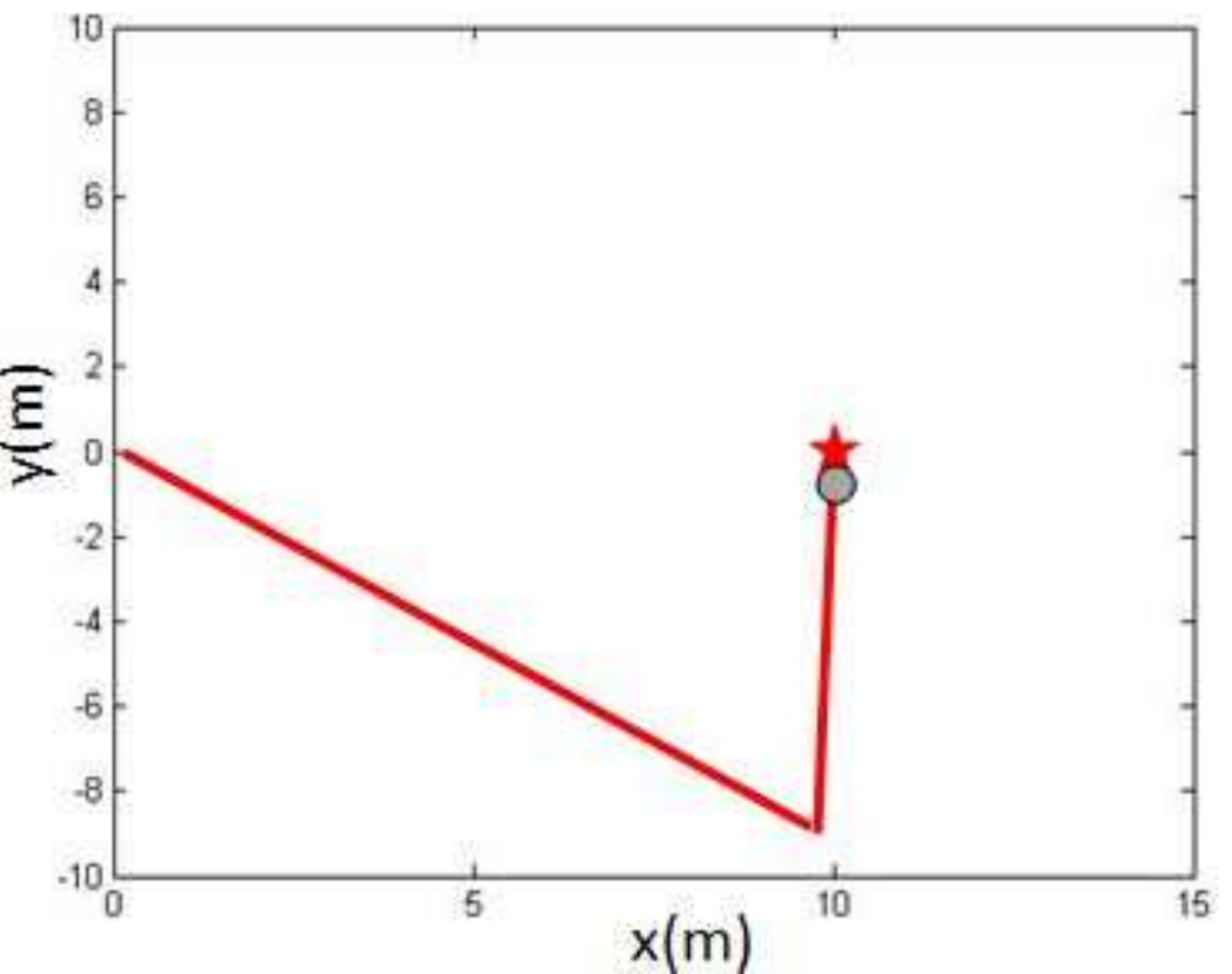}}
		\label{exp3_66.fig}}
		\caption{Proposed navigation algorithm (upper row) versus VOA (lower row)}
		\label{c1_sim6}
		\end{figure}
		\par

	The performance of our biologically-inspired algorithm was compared with that of other well-known navigation algorithms. Fig.~\ref{c1_sim6} shows the comparison between our algorithm and the Velocity Obstacle Approach (VOA) \cite{FS98,LaLaSh05}. In Fig.~\ref{exp3_11.fig} and ~\ref{exp3_22.fig}, a long obstacle is moving perpendicularly to the path between the mobile robot and the target. Fig.~\ref{exp3_33.fig} and ~\ref{exp3_44.fig} show the moment when the mobile robot is about to pass the lower end of the obstacle. As it can be observed, the  biomimetic algorithm avoids the obstacle by keeping a constant avoiding angle between the mobile robot's moving direction and the lower end of the obstacle. On the other hand, VOA drives the mobile robot towards the lower end of the obstacle in a straight line and then move towards the target position. The complete paths for mobile robot to reach the target are shown in Fig.~\ref{exp3_55.fig} and Fig.~\ref{exp3_66.fig}. The overall maneuver time for the biomimetic algorithm is 17.3s and for VOA is 18.7s. Hence,  the biomimetic algorithm  outperforms VOA in this scenario in terms of maneuver time.
\par

	\section {ActivMedia Pioneer 3-DX Wheeled Mobile Robot} \label{p3}

	The navigation algorithms proposed in this report are implemented on ActivMedia Pioneer 3-DX wheeled mobile robot to demonstrate their performance and features. The description of Pioneer 3-DX mobile robot is presented in this section, Pioneer 3-DX (abbreviated as P3) is one of the most commonly used testbeds in a robotic laboratory (see e.g.\cite{PLE04,FN08}) due to its versatility, reliability and durability, manufactured by ActivMedia Robotics. 
\par

	The base of P3 contains onboard computer, swappable batteries (8-10 hours run-time with 3 batteries) and other circuitries. The body of the base is made of tough ($1.6mm$) aluminum which protects these important internal components from physical damages. It is a two-wheeled differential driven robot. The tire of P3 is made of foam-filled rubber which provides sufficient traction on various types of floor. The encoders are attached to the motors of both wheels. There are 8 front and 8 rear ultrasonic sensors equipped around the sides of P3. Furthermore, P3 has a wide range of accessories and extensions, such as Laser sensor and navigation package, manipulator arms and gripper, color stereo camera etc.

		\begin{figure}[h]
		\centering
		\includegraphics[width=4.0in]{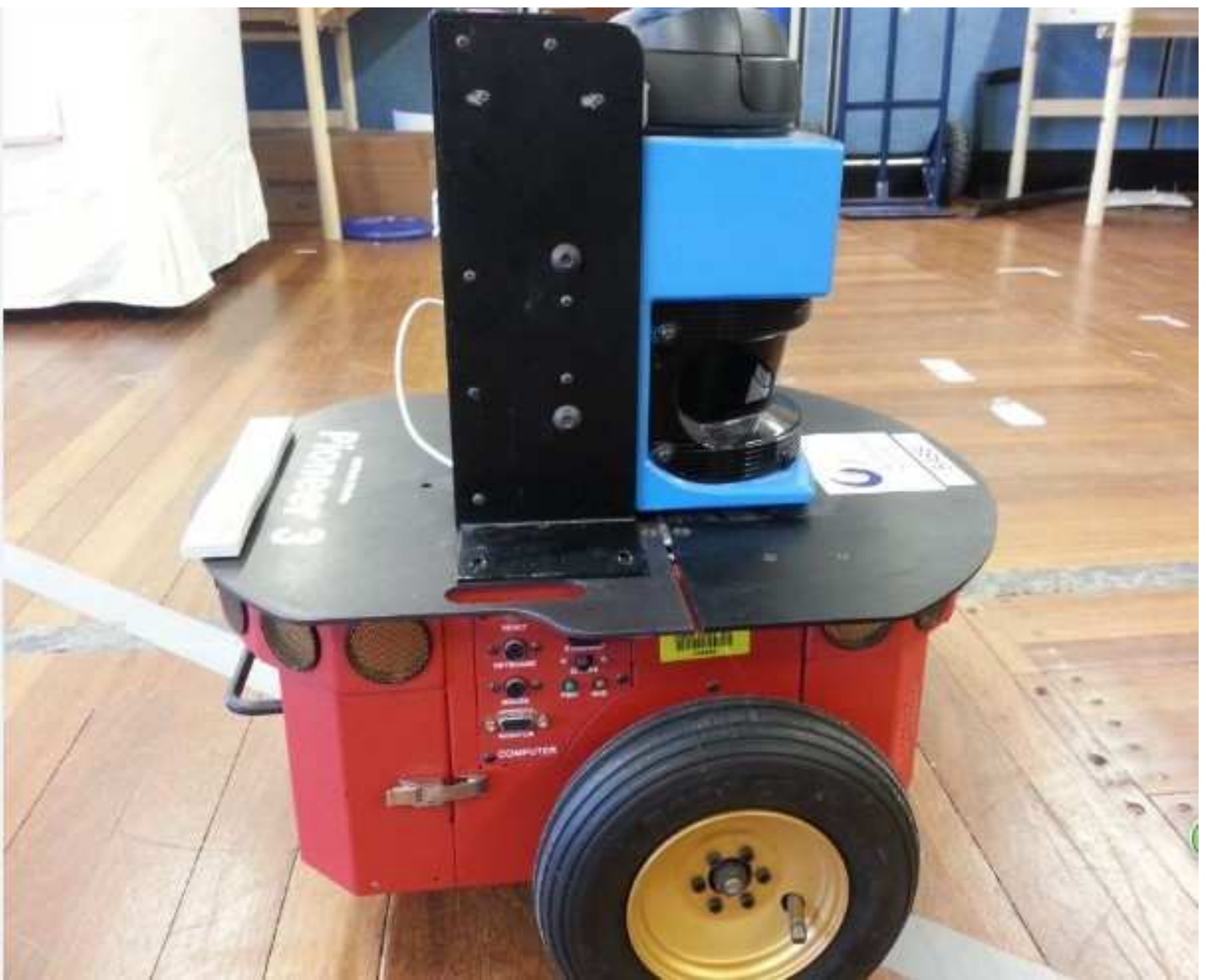}
		\caption{Pioneer 3-DX wheeled mobile robot with SICK laser range finder}
		\label{pionner}
		\end{figure}

	The following components are closely related to the implementation of the proposed navigation algorithm, thus their specification and functions should be emphasised:

	\begin{itemize}

		\item An onboard computer or an internal notebook with wireless capability with the proposed navigation algorithm is the centre piece of the P3 control system. All data are sent to the computer, where the corresponding control signals are computed and sent to the motor of P3.
		\item The differential driven motor system allows P3 to drive forward and backward at maximum speed of $1.2 m/s$, and rotation speed of $300 degree/s$.
		\item The 500-tick encoders are attached to the motors so that the relative position and the orientation of P3 regarding to the reference point can be estimated.
		\item the sensory system is consisted of a SICK laser range finder which provides only $190 degree$ field of view, and totally 16 ultrasonic sensors which provide $360 degree$ field of view. In general, the SICK range finder is more robust to noise ($1 mm$ distance resolution and $0.25 degree$ angular resolution with $\pm0.25 mm$ measurement error) and has a further detection range (maximum $80 m$). On the other hand, the sets of ultrasonic sensor is able to provide  $360 degree$ field of view. Therefore, a better utilisation of the laser range finder and the ultrasonic senors will optimise the performance of sensory system. 
		\end{itemize}
	The more detailed description and specification of P3 system can be found in \cite{mr,ds}.

	\section{Experiments wih Real Mobile Robot}

	The proposed navigation algorithm is implemented and experiments are carried out on an ActivMedia Pioneer 3-DX (P3) robot (see detailed description and specification in Section \ref{p3}) to demonstrate its applicability and performance on a real non-holonomic system. 
\par
		\begin{figure}[!h]
		\centering
		\subfigure[]{\scalebox{0.33}{\includegraphics{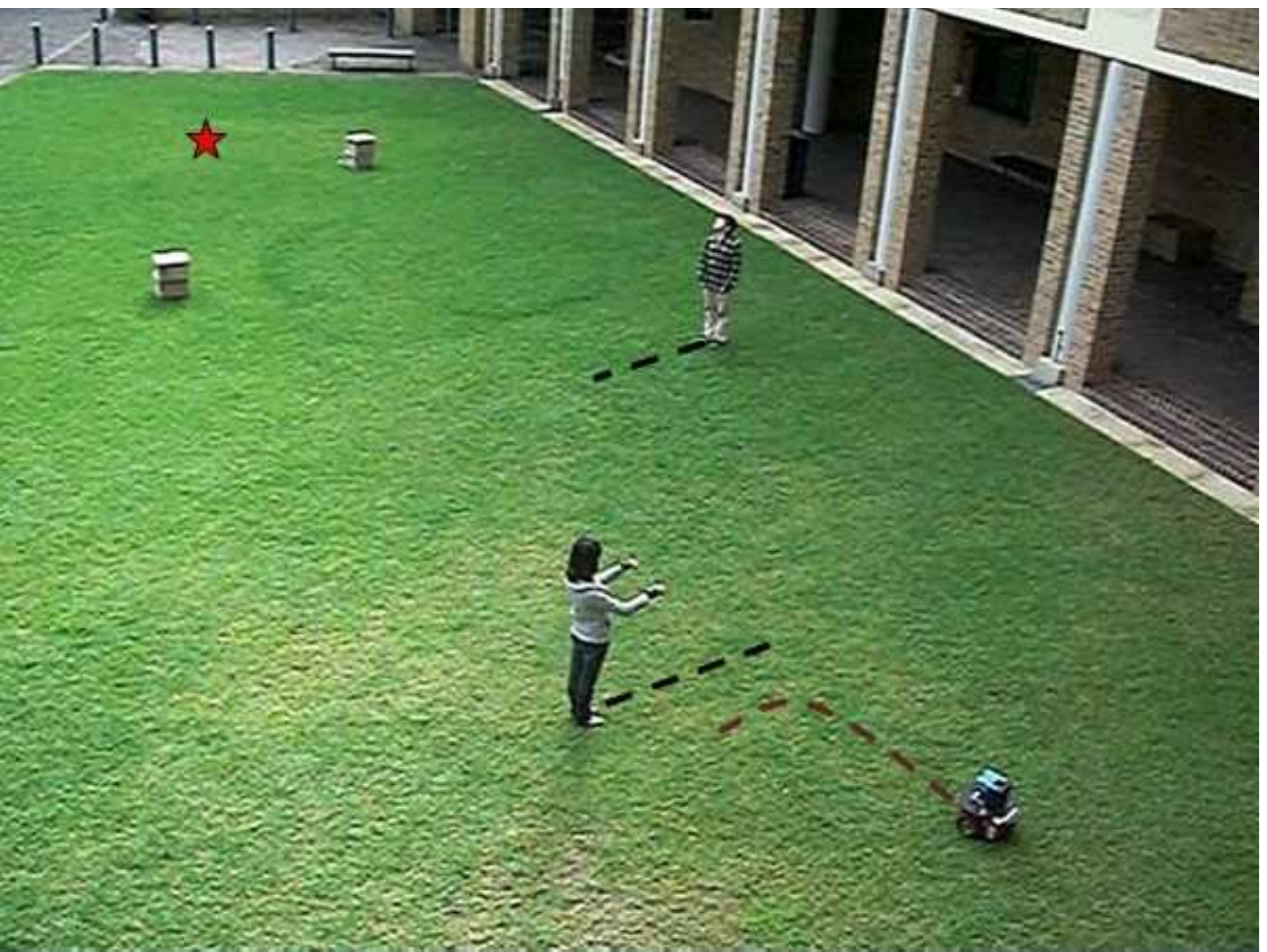}}
		\label{c1.exp31}}		
		\subfigure[]{\scalebox{0.33}{\includegraphics{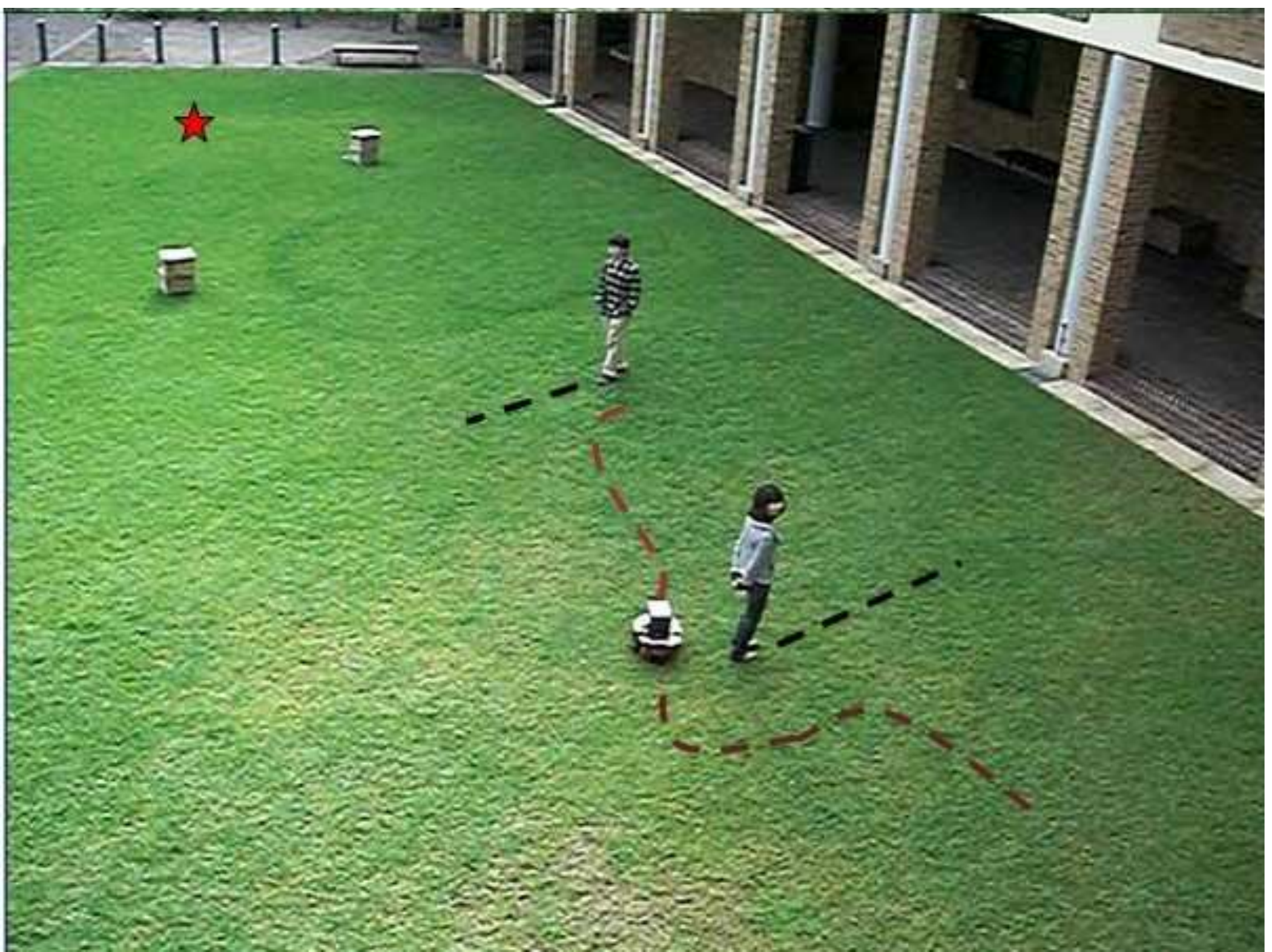}}
		\label{c1.exp32}}
		\subfigure[]{\scalebox{0.33}{\includegraphics{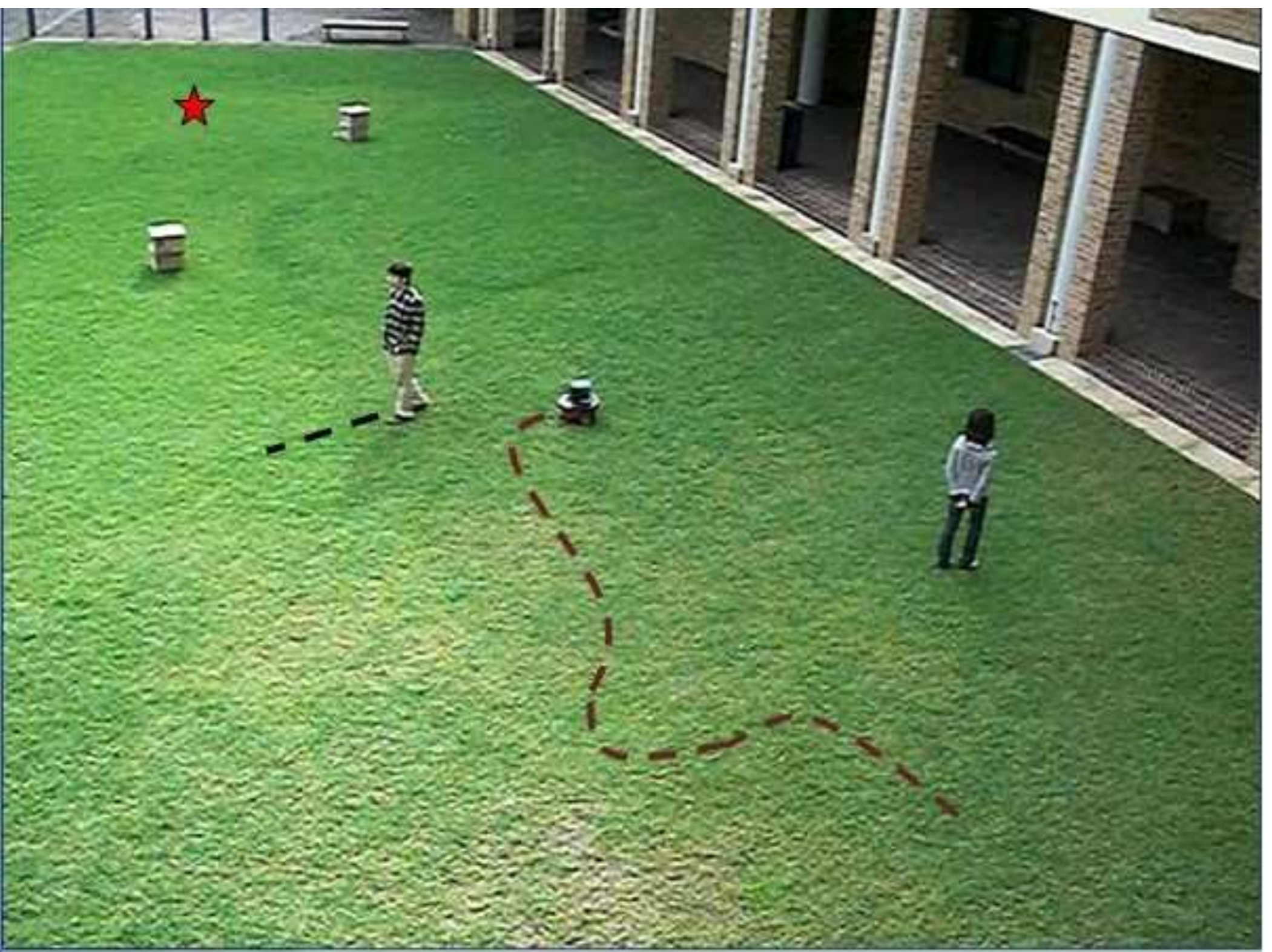}}
		\label{c1.exp33}}
		\subfigure[]{\scalebox{0.33}{\includegraphics{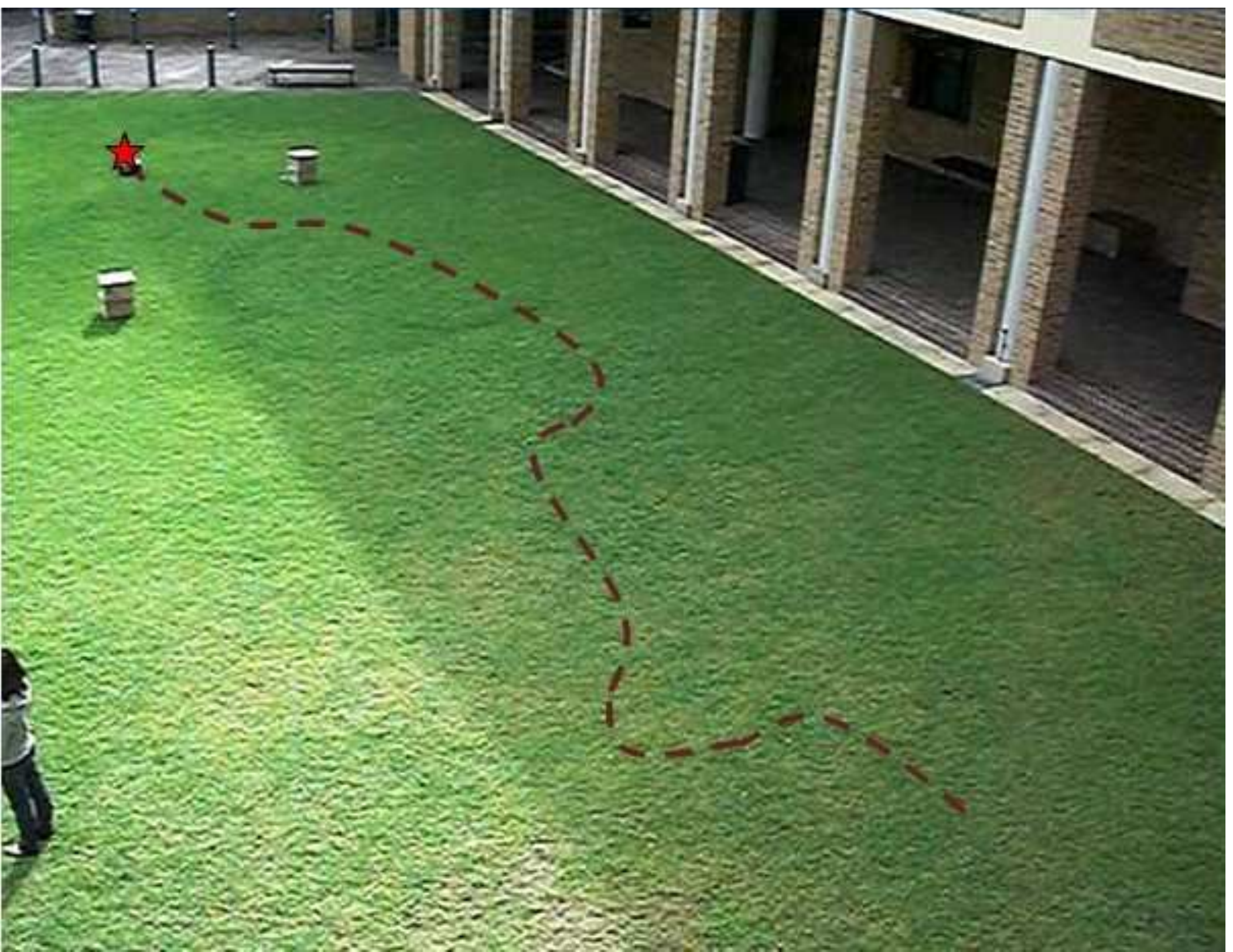}}
		\label{c1.exp34}}
		\caption{Robot navigating in an open area}
		\label{c1.exp3}
		\end{figure}
	In these experiments, the distance ($d_i(t)$) and the vision cone ($\alpha_i^{(1)}, \alpha_i^{(2)}$) from P3 to the closest obstacle are measured by SICK laser range finder (for the front area) and ultrasonic sensors (for the rear area). The position ($x(t),y(t)$) and orientation ($\theta(t)$) of P3 are fed back from encoders attached to the motors of both wheels. The target position ($T_x, T_y$) is prescribed to P3 and the heading ($H(t)$) can be calculated by simple geometry.  These information is sent to a notebook where the appropriate control signals are computed. 
\par

	The first experiment is carried out in a general open area. The movements of the robot is governed by our biologically-inspired navigation algorithm. The proper switching between the obstacle avoidance law (\ref{bi_cont}) and the target pursuit law (\ref{bi_pur}) allows the robot to efficiently avoid the obstacles (see Fig.~\ref{c1.exp31}, Fig.~\ref{c1.exp32} and Fig.~\ref{c1.exp33}) and safely arrive at the target location (see Fig.~\ref{c1.exp34}).
\par
\begin{figure}[!h]
		\begin{minipage}{.5\textwidth}	
			\centering	
			\subfigure[]{\scalebox{0.4}{\includegraphics{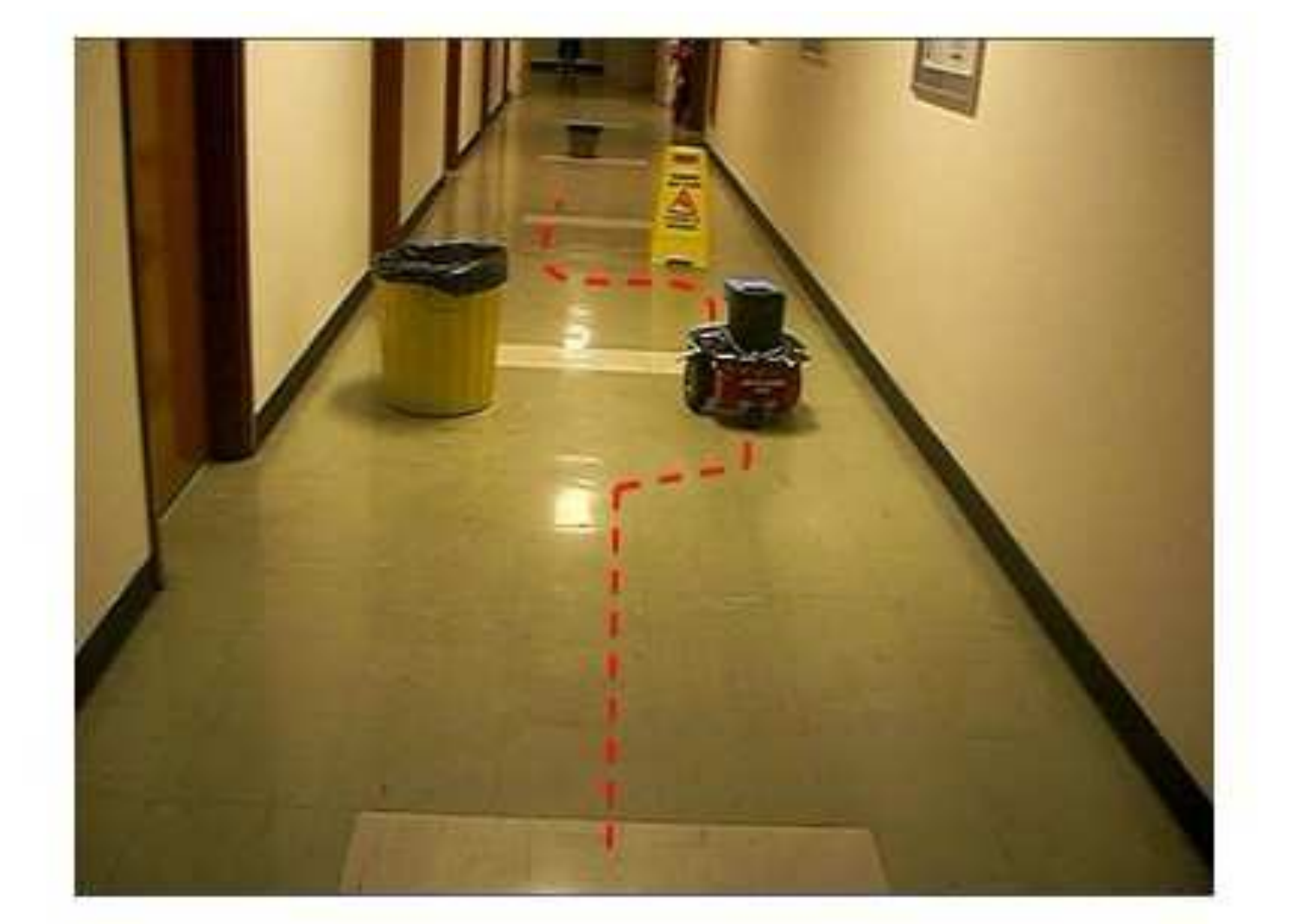}}
			\label{c1.exp11}}
			\subfigure[]{\scalebox{0.4}{\includegraphics{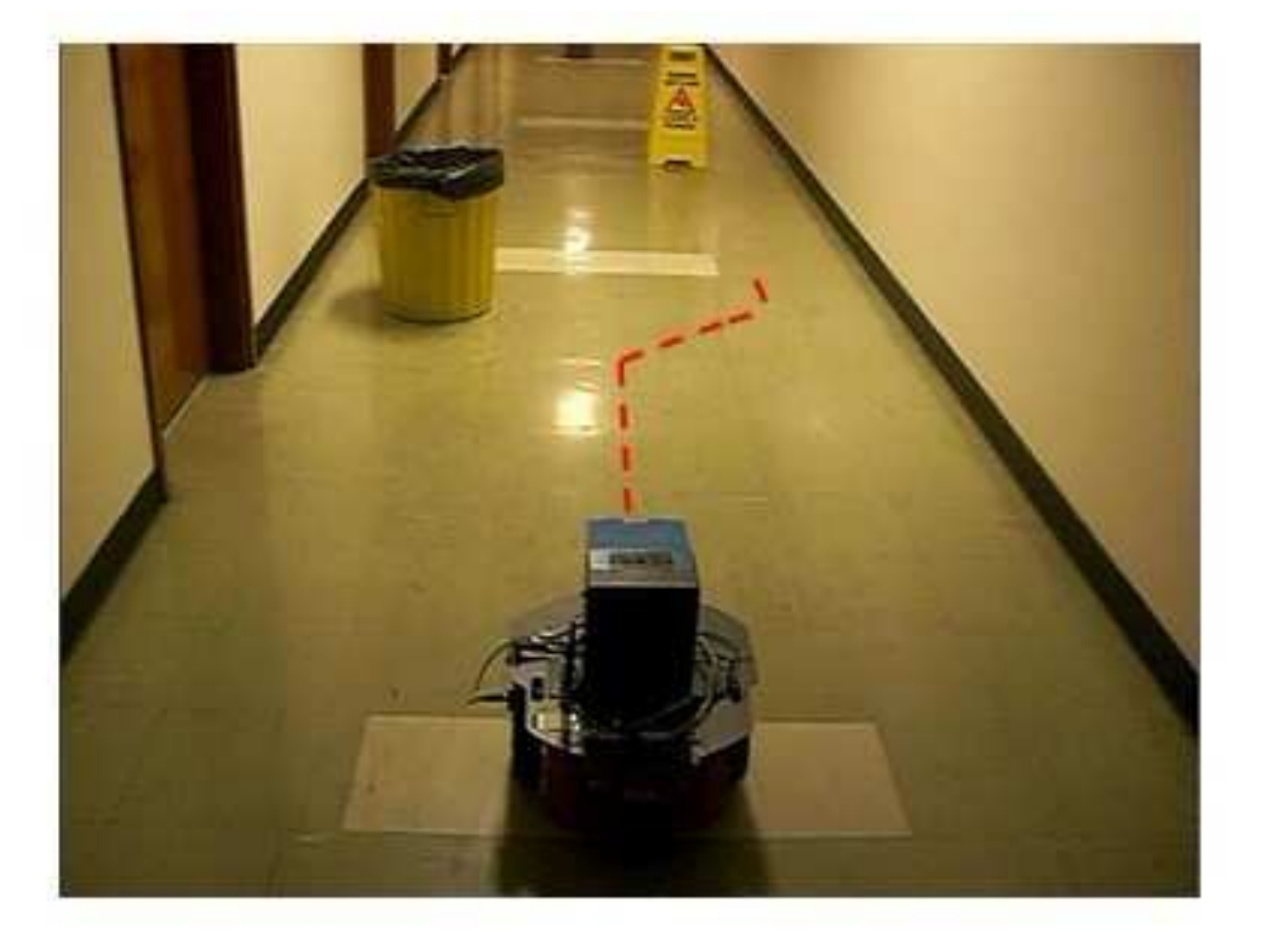}}
			\label{c1.exp12}}
			\subfigure[]{\scalebox{0.4}{\includegraphics{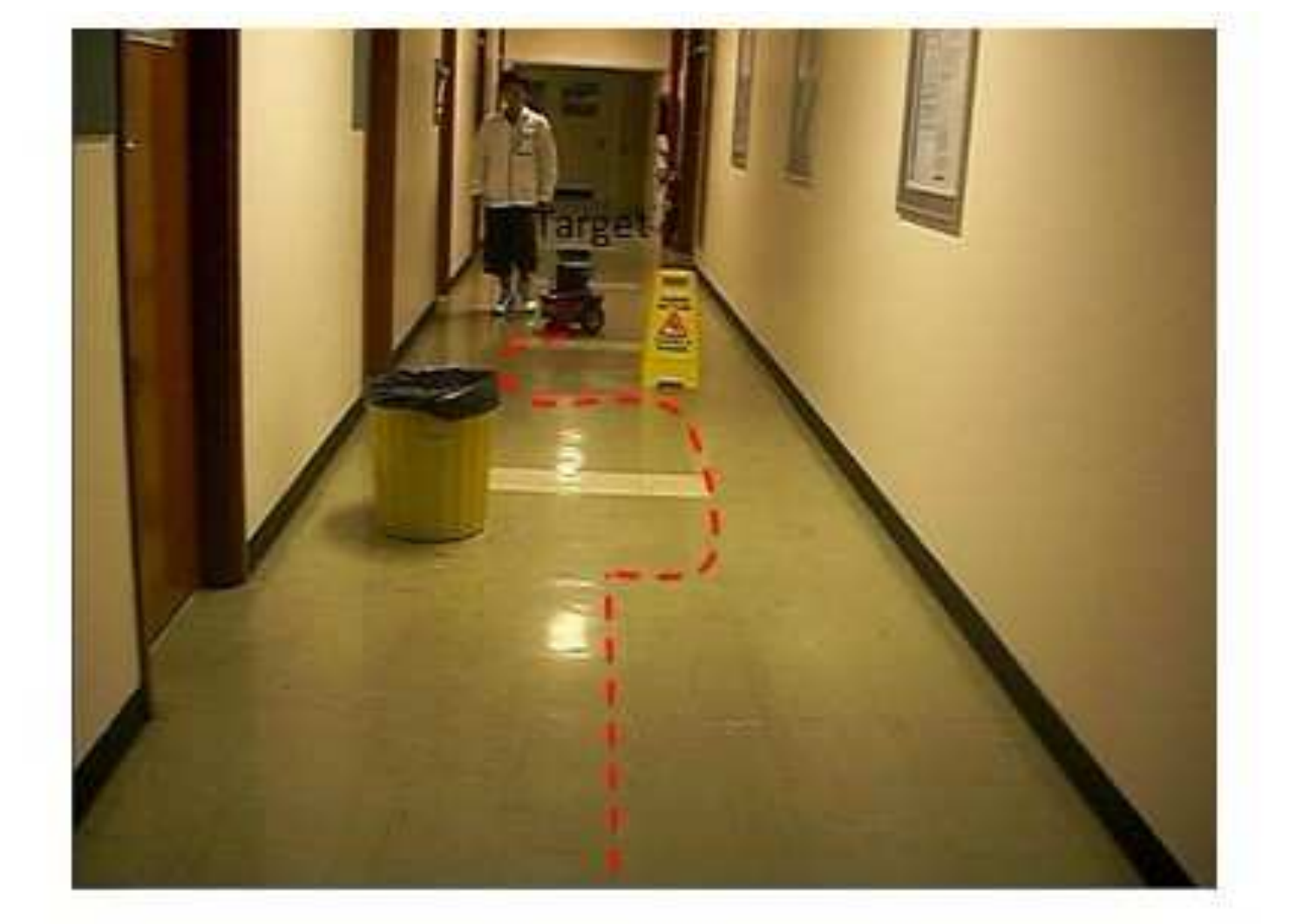}}
			\label{c1.exp13}}
		\end{minipage}
		\begin{minipage}{0.5\textwidth}
			\subfigure[]{\scalebox{0.8}{\includegraphics{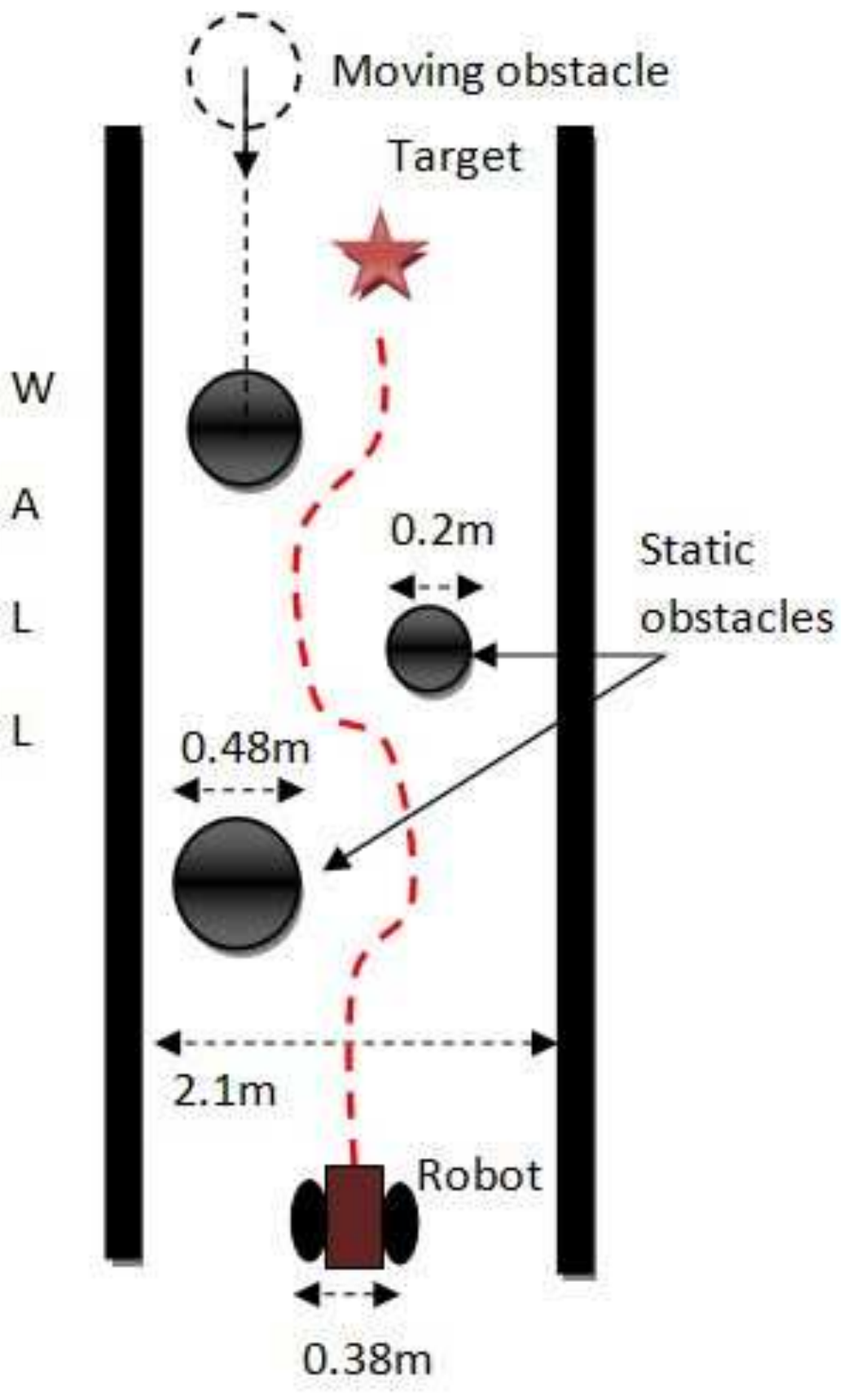}}
			\label{c1.exp14}}
		\end{minipage}
		\caption{Robot navigating in a corridor with stationary and moving obstacles case 1}
		\label{c1.exp1}
		\end{figure}
\begin{figure}[!h]
		\begin{minipage}{.5\textwidth}	
			\centering	
			\subfigure[]{\scalebox{0.34}{\includegraphics{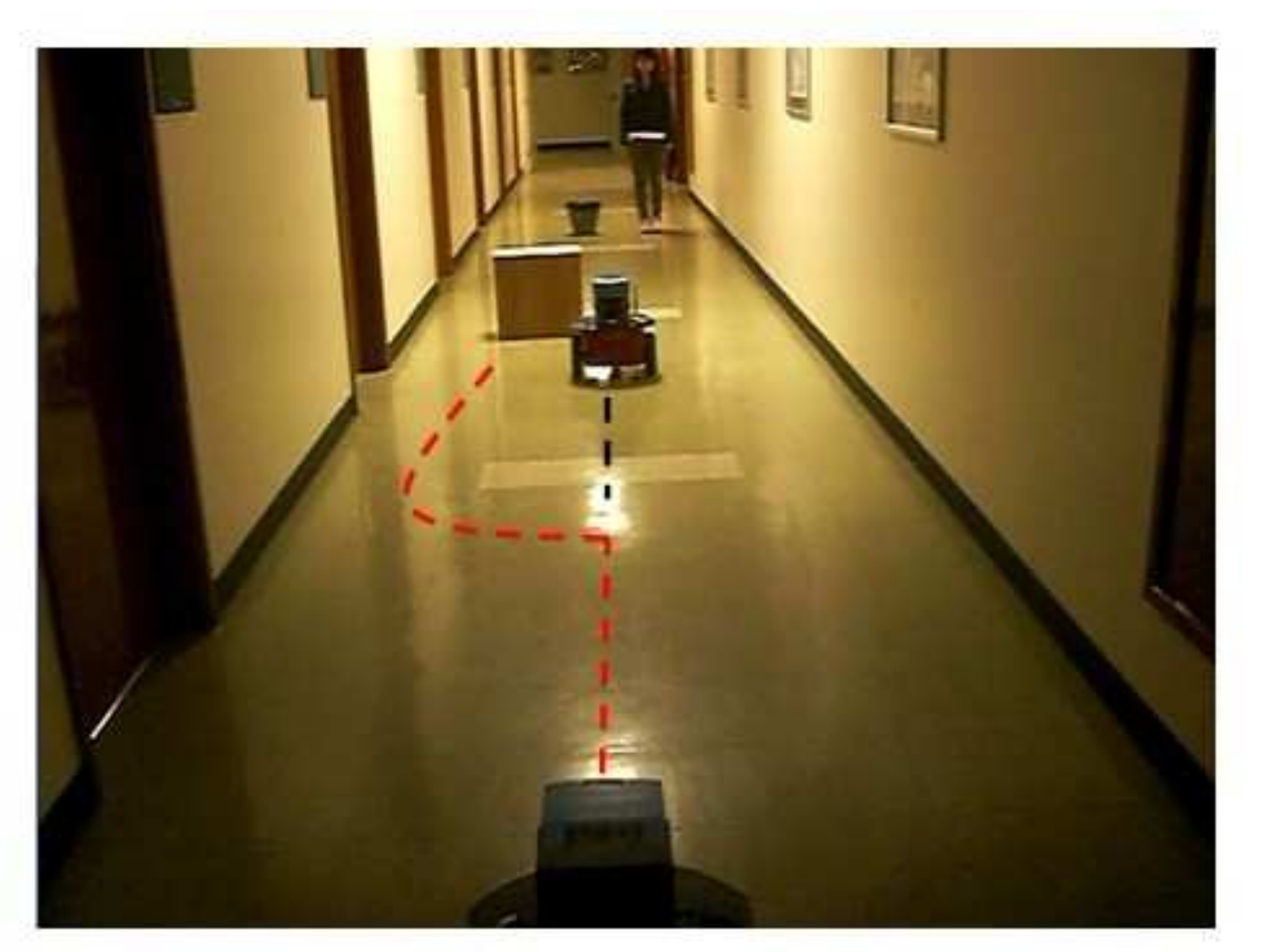}}
			\label{c1.exp21}}
			\subfigure[]{\scalebox{0.34}{\includegraphics{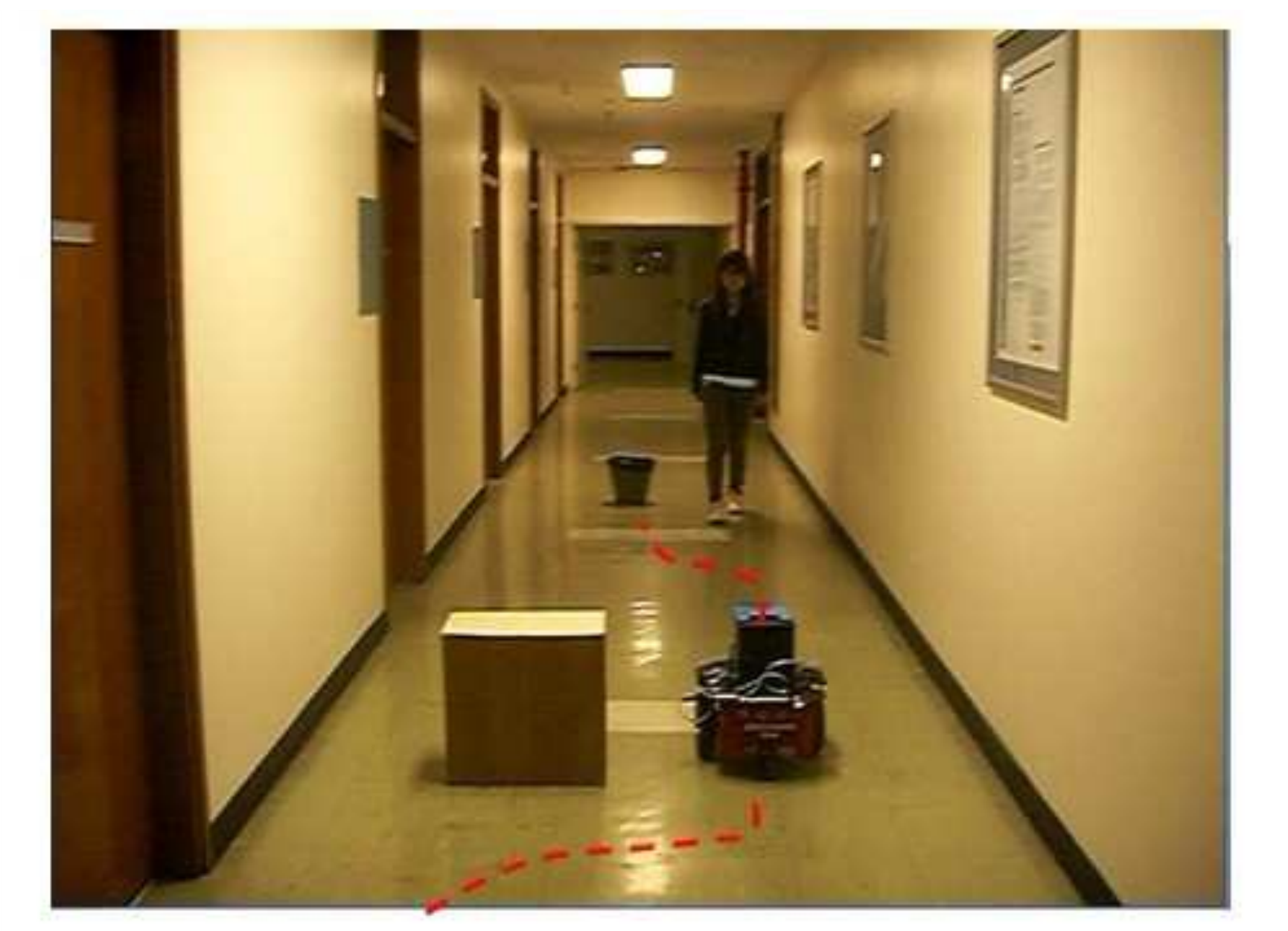}}
			\label{c1.exp22}}
			\subfigure[]{\scalebox{0.34}{\includegraphics{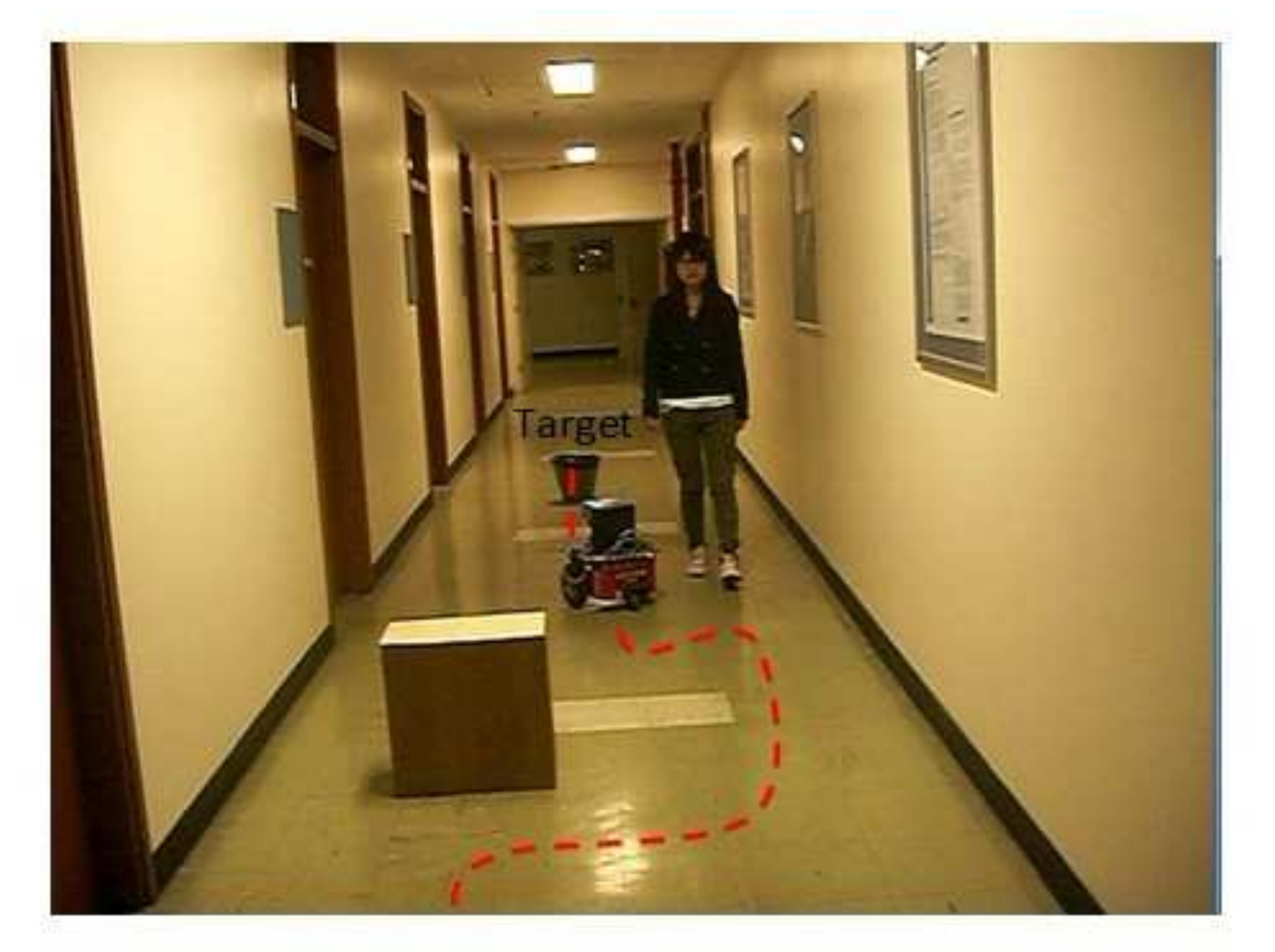}}
			\label{c1.exp23}}
		\end{minipage}
		\begin{minipage}{0.5\textwidth}
			\subfigure[]{\scalebox{0.8}{\includegraphics{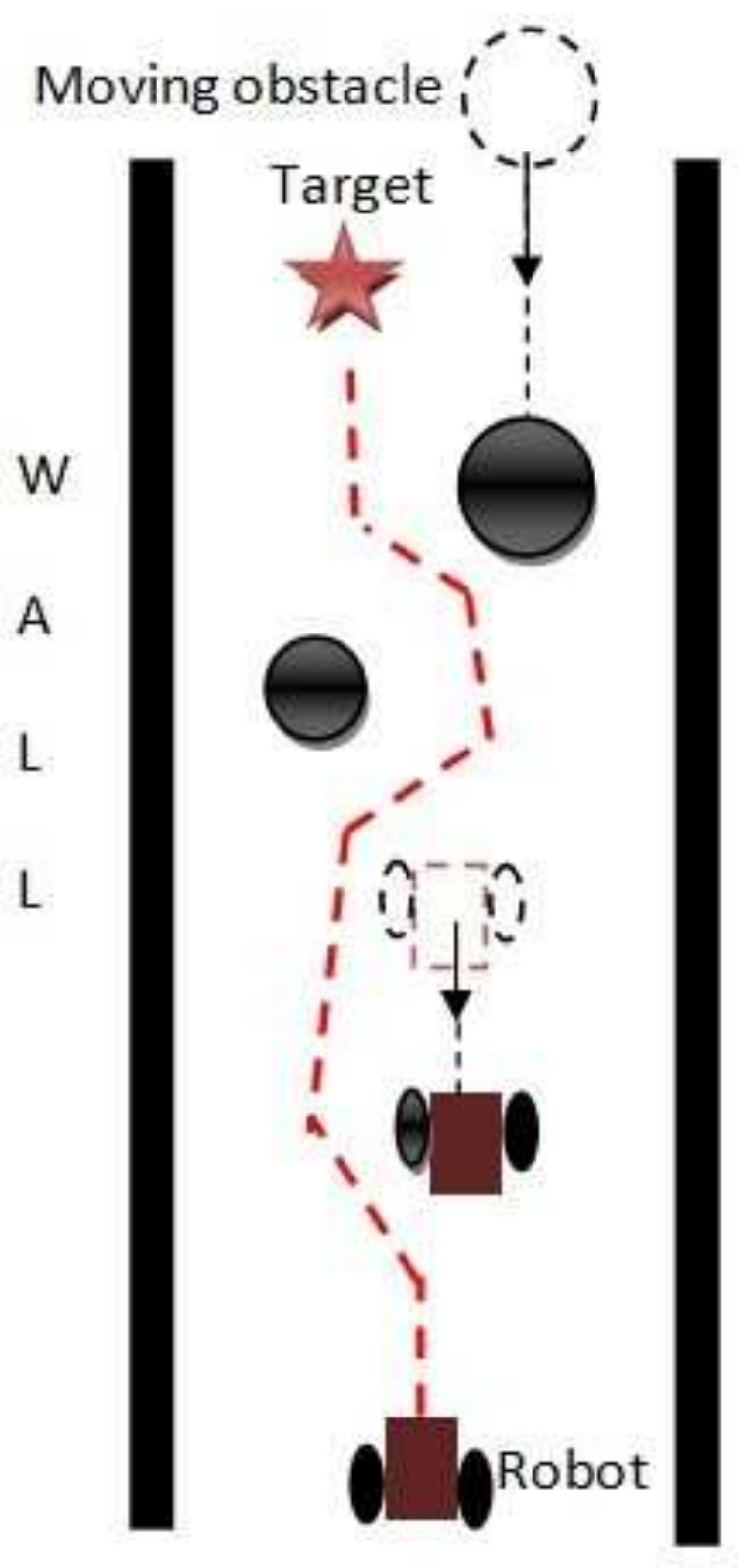}}
			\label{c1.exp24}}
		\end{minipage}
		\caption{Robot navigating in a corridor with stationary and moving obstacles case 2}
		\label{c1.exp2}
		\end{figure}

	In the next experiments, we examine the performance of the proposed navigation algorithm in a corridor with stationary and moving obstacles.  The robot needs to avoid these obstacles while not colliding into the walls. Note that in the experiments, the vision cone to the obstacles are finite. In particular, the vision cone of a wall consists of straight lines connecting the robot and points of the wall at the distance less than value $D$.  In Fig.~\ref{c1.exp1}, the robot avoids two stationary obstacle in Fig.~\ref{c1.exp11}, Fig.~\ref{c1.exp12} and a moving obstacle in Fig.~\ref{c1.exp13}.  During the avoidance maneuver, the robot is able to keep a certain distance to the obstacles and the walls. The complete path taken by the robot is shown in Fig.~\ref{c1.exp14}. A similar experiment is presented in Fig.~\ref{c1.exp2}, in this experiment, one more moving obstacle is brought into the scene and the stationary obstacles are moved close to the center of the corridor which restricts the maneuver area for the robot.

\begin{figure}[!h]
		\centering
		\subfigure[]{\scalebox{0.33}{\includegraphics{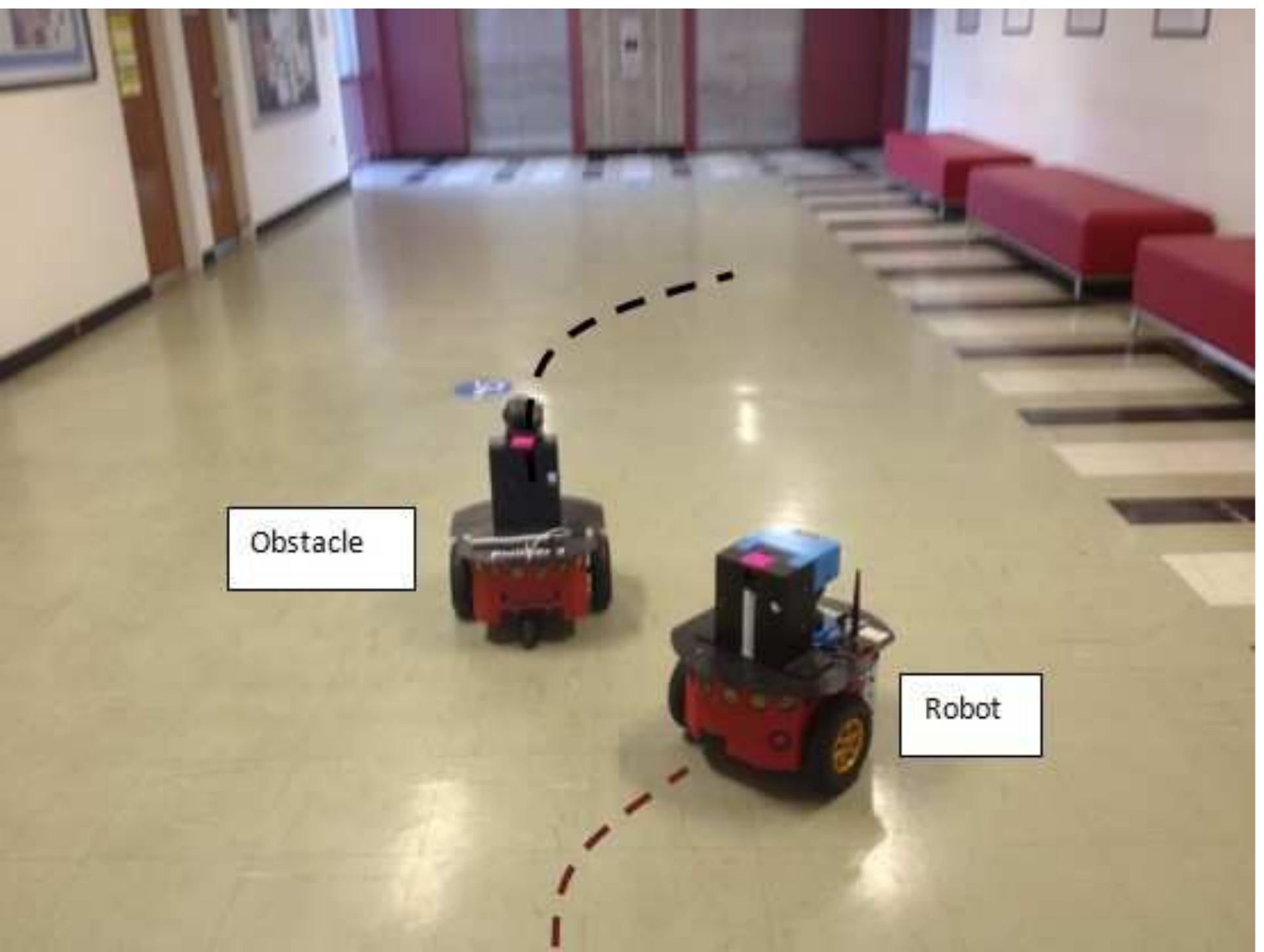}}
		\label{c1.exp41}}		
		\subfigure[]{\scalebox{0.33}{\includegraphics{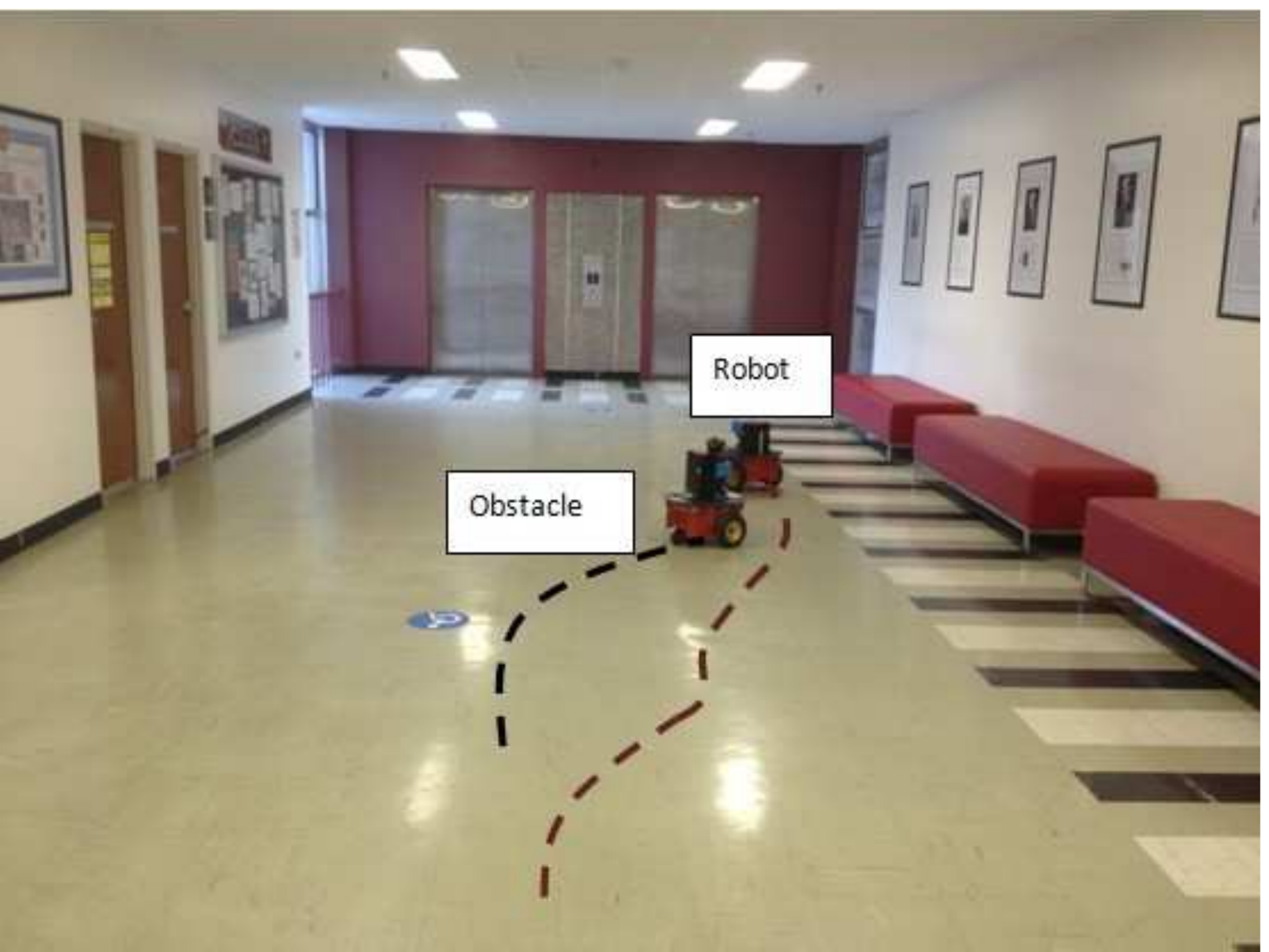}}
		\label{c1.exp42}}
		\caption{Robot avoiding obstacle with non-linear velocity case 1}
		\label{c1.exp4}
		\end{figure}

\begin{figure}[!h]
		\centering
		\subfigure[]{\scalebox{0.33}{\includegraphics{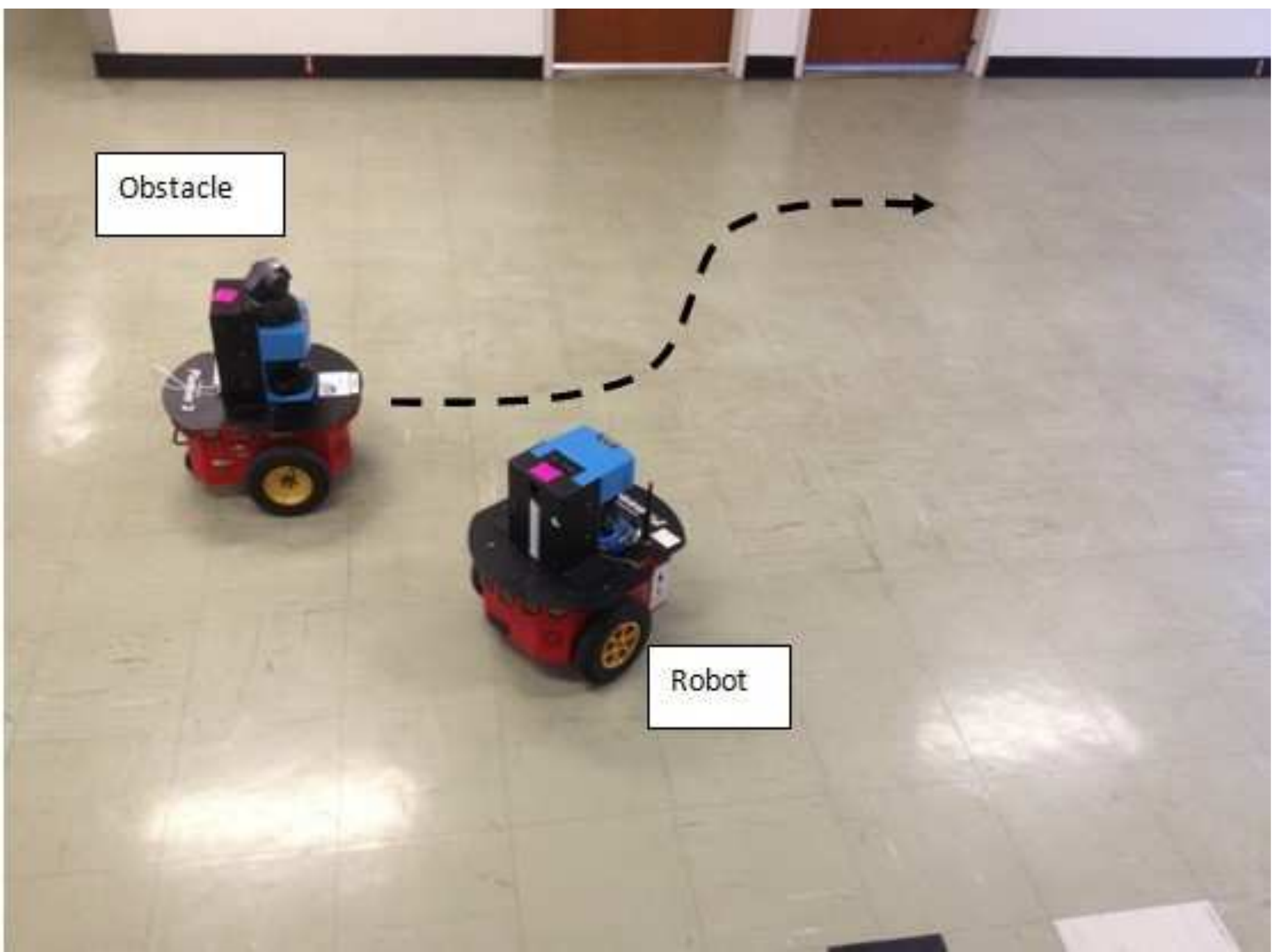}}
		\label{c1.exp51}}		
		\subfigure[]{\scalebox{0.33}{\includegraphics{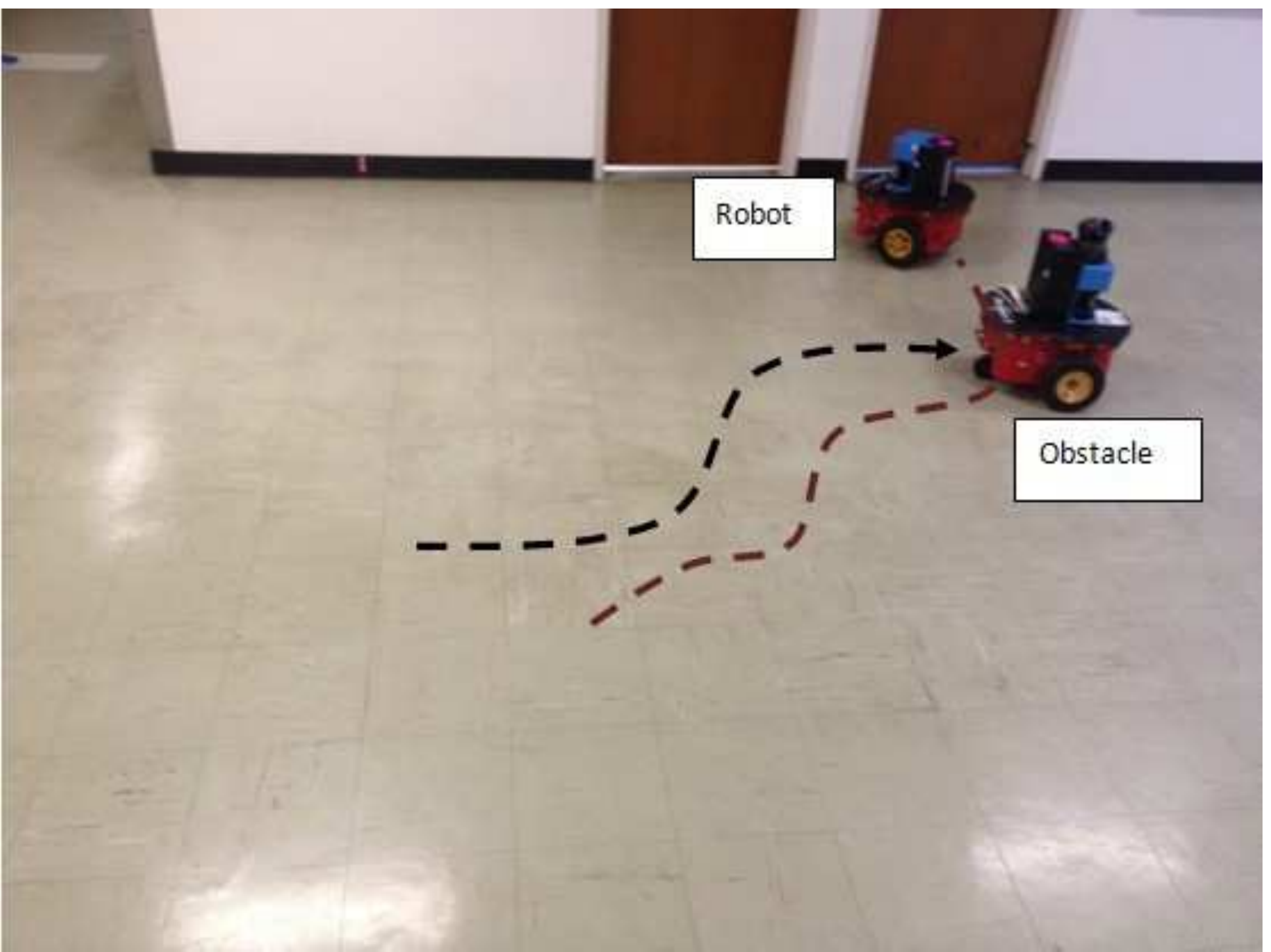}}
		\label{c1.exp52}}
		\caption{Robot avoiding obstacle with non-linear velocity case 2}
		\label{c1.exp5}
		\end{figure}

	The last experiment shows the ability of the proposed navigation algorithm to avoid obstacles with non-linear velocities. The avoidance of obstacles with non-linear velocities is much more difficult since the future trajectory of the obstacles is more unpredictable and the non-linear velocities are much difficult to estimate (possible solution can be found in \cite{LHZ11,SYX05,ILJ06}). In this experiment, we use another P3 robot as the obstacle, the maneuver of this obstacle is limited by non-holonomic constraints and its angular velocity is severely limited for the obstacle avoidance to be possible. Fig.~\ref{c1.exp4} and Fig.~\ref{c1.exp5} show two examples of avoidance of obstacle with non-linear velocity. 
\par

	\section {Summary}

	In this section, we present a biologically-inspired navigation algorithm for non-holonomic mobile robots. The proper switching between the obstacle avoidance law and the target pursuit law guarantees the safety of the mobile robot in cluttered dynamic environment, and  safely drives the mobile robot to the target position. The performance of our real-time navigation strategy is confirmed by extensive computer simulations and experiments with a Pioneer 3-DX mobile robot.
\chapter {Range-only Based Obstacle Avoidance Strategy for Safe Navigation in Uncertain Dynamic Environments} \label{C3}

	The navigation of mobile robot in uncertain dynamic environments is a very challenging topic in robotics due to limited information available. The term uncertain implies the mobile robot may observe only  a small part of the obstacle, may not distinguish between its points, and so may be unable to estimate many of its parameters, especially the size, center, edge and full velocity etc. In this section, we present a reactive navigation algorithm for non-holonomic mobile robot. Unlike the previous researches on navigation in uncertain environments, such as BUG algorithm\cite{LV86}, the obstacles are assumed stationary in the environment, our proposed navigation algorithm deals with dynamic obstacles in uncertain environments. The proposed navigation algorithm does not need to build a complete map for the environment or to recognise individual obstacle in the scene, it does not require data, either exact or estimated, about the obstacle' size, position, orientation and velocities, which is required by many other existing algorithms such as Velocity Obstacle Approach  \cite{FP93,FP98}.  Additionally, the obstacles are not assumed to be rigid or even solid. They are continuums of arbitrary and time-varying that may rotate, twist, wring, skew. wriggle or be deformed in any other way. The only information available is the minimum distance from the mobile robot to the nearest obstacle.

	\section {Problem Description} \label{PDC3}

	We consider a mobile robot which travels in a plane. Its motion is controlled by its speed $v$ and angular velocity $u$ both limited by given constants $V$ and $\bar{u}$. The position of mobile robot is represented by the absolute Cartesian coordinate $(x,y)$ of the reference point, which is the center of axle connecting the two driving wheels. Its orientation is given by the angle $\theta$ between the mobile robot centerline and the abscissa axis. The mathematical model of the mobile robot is described by the following equations:
	\begin{equation}
	\label{1}
	\begin{array}{l}
	\dot{x} = v \cos \theta,
	\\
	\dot{y} = v \sin \theta,
	\\
	\dot{\theta} = u
	\end{array},~~~ 
	\begin{array}{l}
	 x(0) = x_0,
	\\
	y(0) = y_0,
	\\
	\theta(0) = \theta_0 
	\end{array} 
	\end{equation}
	where 
	\begin{equation}
	\label{max}
	 u\in [-U_{max},U_{max}],~~~v\in [0,V].
	\end{equation} 
	These equations describes the ability of the mobile robot to move forward with a certain speed in the forward direction along planar curves whose curvature radius exceeds a given threshold. In this case, this minimal turning radius equals
	\begin{equation}
	 R=v/\bar{u}
	\end{equation}
	\par
	There are multiple unknown disjoint static and moving obstacles in the same plane as the mobile robot. At time $t$, each of them occupies a certain domain $D_i$, where $i = 1,2,\dots,n$ is the serial number of the obstacle. The information regarding to these obstacle is scarce: the positions of them are not known in advance and they may undergo arbitrary motions. Furthermore, there is no specifications on the shapes of the obstacles and the obstacles are not necessarily a rigid body: they may perpetually change their shapes and undergo any displacements and deformations, including stretches, bends, twists, etc. We do not limit ourselves to a particular type of the obstacle kinematics and do not specify whether the obstacles are rigid, elastic, or plastic solid or fluid.  
	\par
\begin{figure}[h]
	\centering
	\subfigure[]{\scalebox{0.49}{\includegraphics{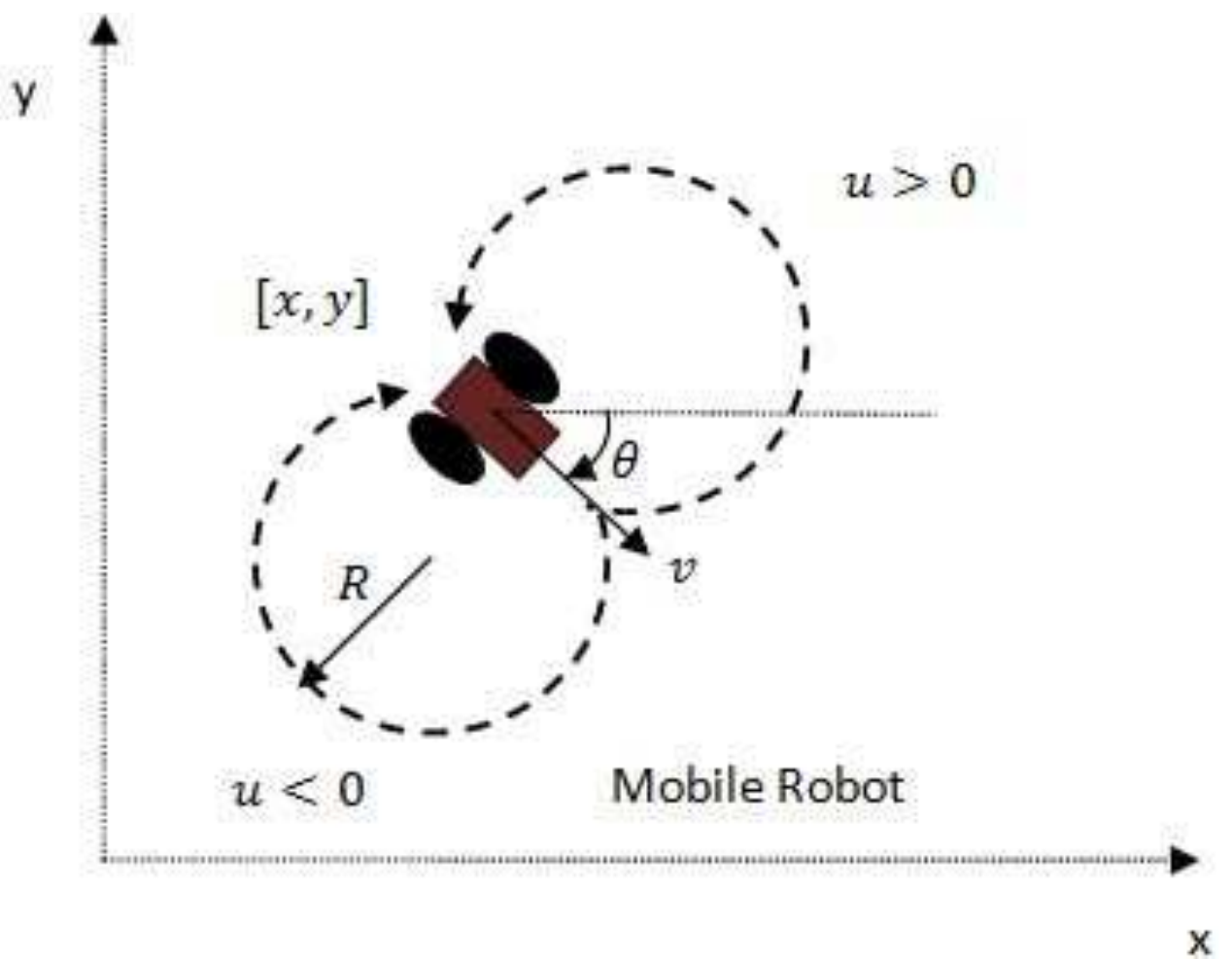}}
	\label{non}}
	\hfill
	\subfigure[]{\scalebox{0.48}{\includegraphics{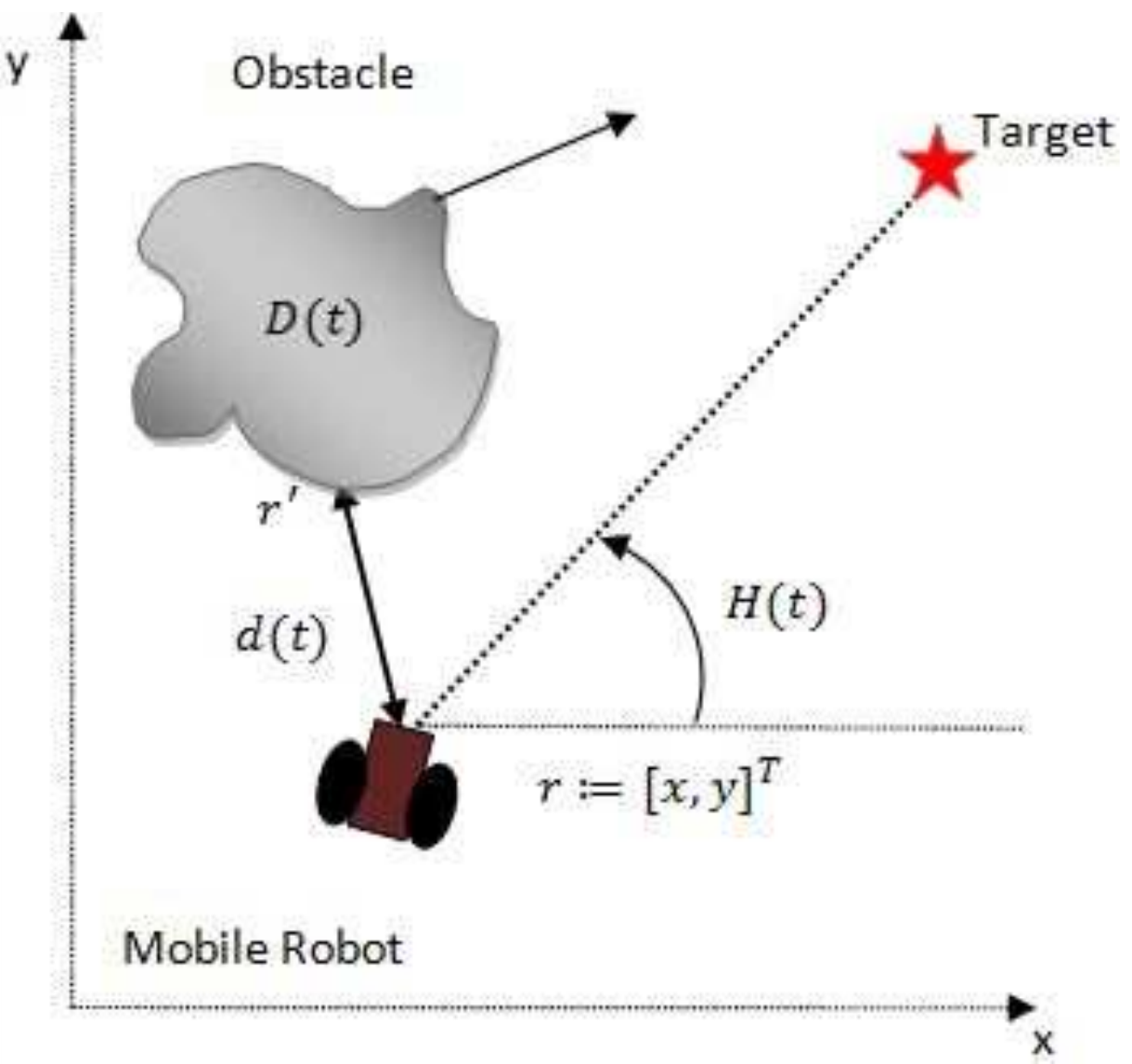}}
	\label{d}}
	\caption{(a)Coordinates and orientation of the mobile robot and its minimal turning radius $R$; (b)Moving domain $D(t)$ and heading to target $H(t)$}
	\end{figure}

	The only information that is available to the mobile robot is the current distance $d(t) := \text{\bf dist}_{D(t)}[r(t)]$ to the closest obstacle $D(t)$ and the rate $\dot d(t)$ at which this measurement evolves over time. Here $r:=[x, y]^{\top}$ is the vector of the coordinates of the mobile robot and
	\begin{equation}
	\label{ch3:dist} \text{\bf dist}_{D}(r) :=
	\min_{r^\prime \in D}
	\|r-r^\prime\|
	\end{equation}
	(see Fig.\ref{d}), where $\|\cdot\|$ denotes the standard Euclidean vector norm and the minimum is achieved if $D$ is closed.
	\par
	
	Finally, there is a steady point-wise target $\bf{T}$ in the plane and the mobile robot is able to measure the heading $H(t)$ towards the target, see Fig.\ref{d}.
The objective is to guide the mobile robot through the obstacle-free part of the  plane and reach the target $\bf{T}$ at a certain time $t_f>0$:
	\begin{equation*}
				r(t_f) = \text{\bf T};
		\qquad
            r(t) \not\in \bigcup_{i=1}^n D_i(t)\quad \forall t \in [0,t_f].
			\end{equation*}
Moreover, the distance between the mobile robot and any obstacle should constantly exceed a given safety margin $d_{\text{safe}}>0$:
	\begin{equation}
	\label{dist}
		\text{\bf dist}_{D_i(t)}\left[ r(t) \right]\geq d_{\text{safe}}\quad \forall t \in [0,t_f], i=1,\ldots,n .
	\end{equation}

\section{Reactive Navigation Algorithm} \label{A22}

	In this section, we present a description of the proposed navigation algorithm, which combines a sliding mode obstacle avoidance strategy and a target reaching strategy. The obstacle avoidance strategy is activated when an obstacle is nearby and the target reaching strategy is put to use when there is no threat of collision.
\par
\begin{figure}[h]
\centering
\includegraphics[width=3.5in]{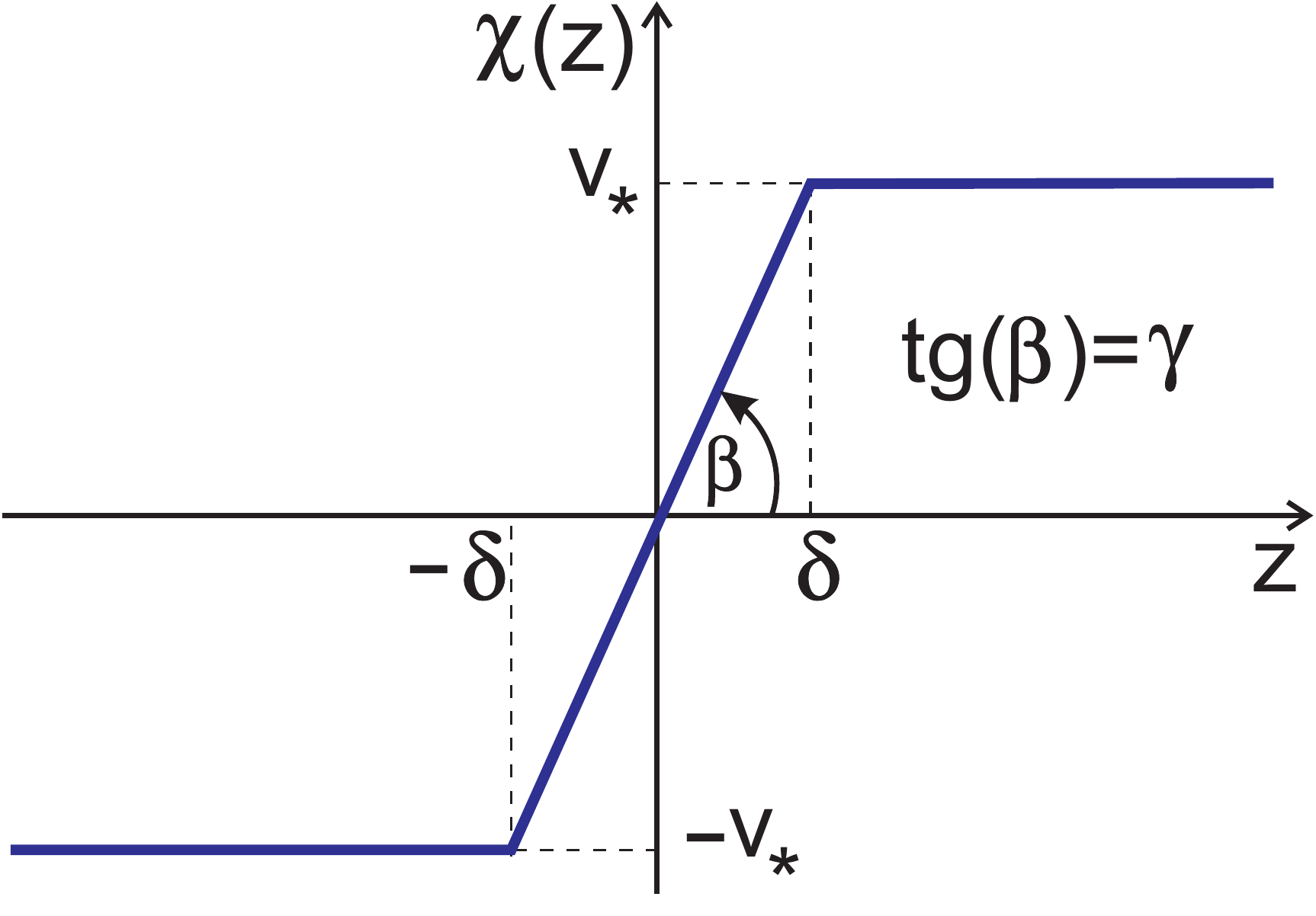}
\caption{Linear function with saturation}
\label{satur}
\end{figure}

We first examine the obstacle avoidance strategy:
\begin{gather}
\label{avoid}
	u(t) = U_{max} \cdot \sgn \big\{ \dot{d}(t) + \chi[d(t)-d_0]\big\} , \qquad
 	\text{where}
 	\\
 	\label{chi}
	\chi(z):=
	\begin{cases} \gamma z & \text{if} \; |z|\leq \delta
	\\
	v_{\ast} \sgn (z)& \text{if} \; |z| > \delta
	\end{cases} \qquad (v_{\ast} := \gamma\delta )
\end{gather}
is the linear function with saturation, see Fig.\ref{satur}, whereas $\gamma>0, \delta >0$, and $d_0 > d_{\text{safe}}$ are the controller parameters, with the last of them $d_0$ being the desired distance to the obstacle when bypassing it.
\par
\par
The proposed obstacle avoidance strategy belongs to the class of sliding mode control algorithms; see e.g. \cite{UT92}.
The intuition behind this strategy is that in the sliding mode, the equation $\dot{d}+\chi(d-d_0)=0$ of the sliding surface is satisfied,
according to which the mobile robot is steered towards the desired distance $d_0$ to the obstacle. For this to take effect, the sliding mode maneuver should be at least realistic. Since the derivative $\dot{d}$ does not exceed $|\dot{d}| \leq v_r$ the relative speed $v_r$ of the mobile robot with respect to the obstacle, this means that in (\ref{chi}), the saturation level $v_\ast$ should not exceed this speed. This can be achieved by proper tuning of the controller parameters $\gamma$ and $\delta$ based on available estimates of the speeds of the obstacles.
If initially the mobile robot is not on the sliding surface, the control law \eqref{avoid} quickly drives it to this surface after a short initial turn
with the maximal steering actuation, see \cite{AM12} for details. So sliding motion is the main part of the obstacle avoidance maneuver.
\par
\begin{figure}[h]
\centering
\subfigure[]{\scalebox{0.26}{\includegraphics{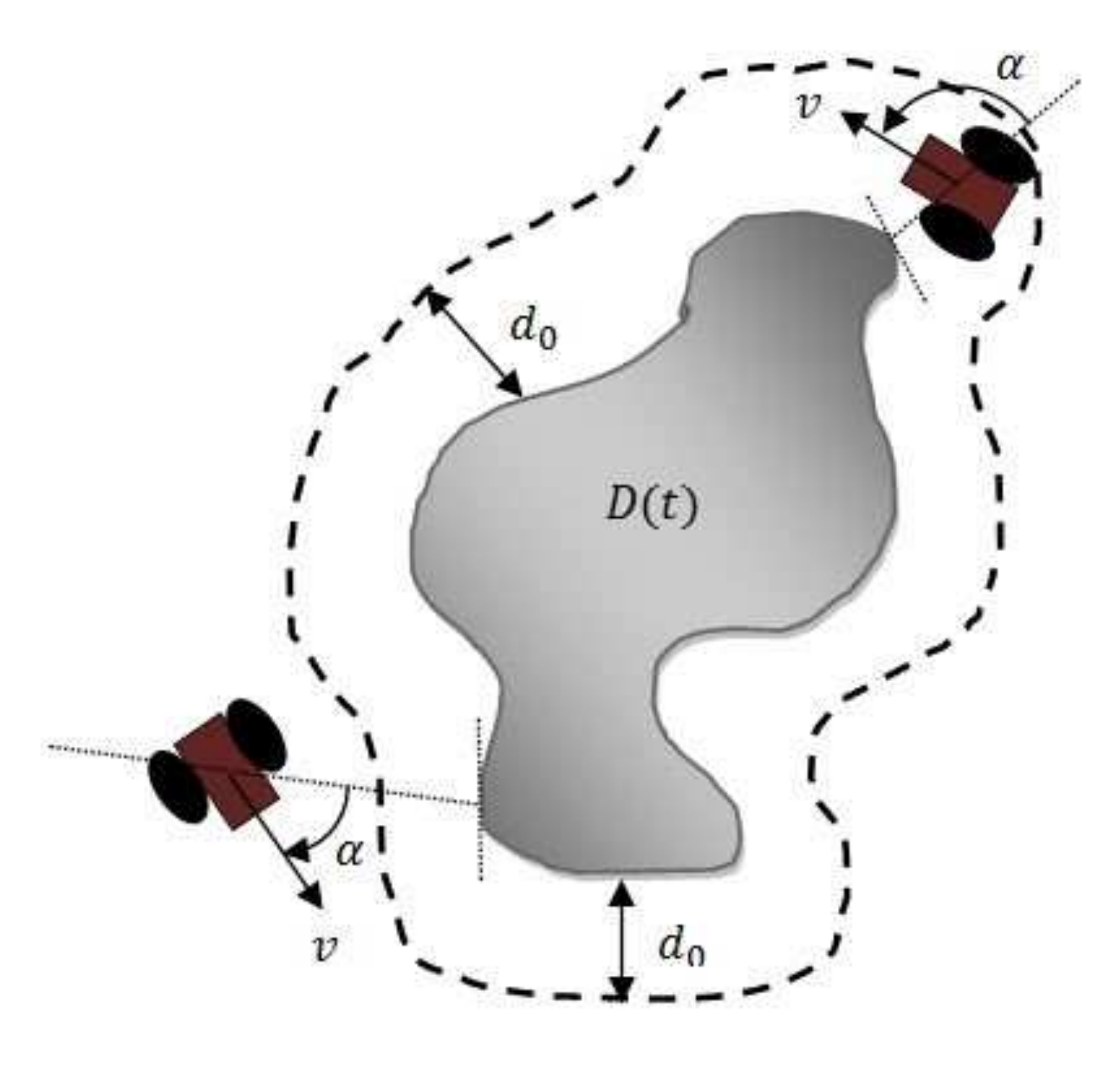}}
\label{sli}}
\hfill
\subfigure[]{\scalebox{0.26}{\includegraphics{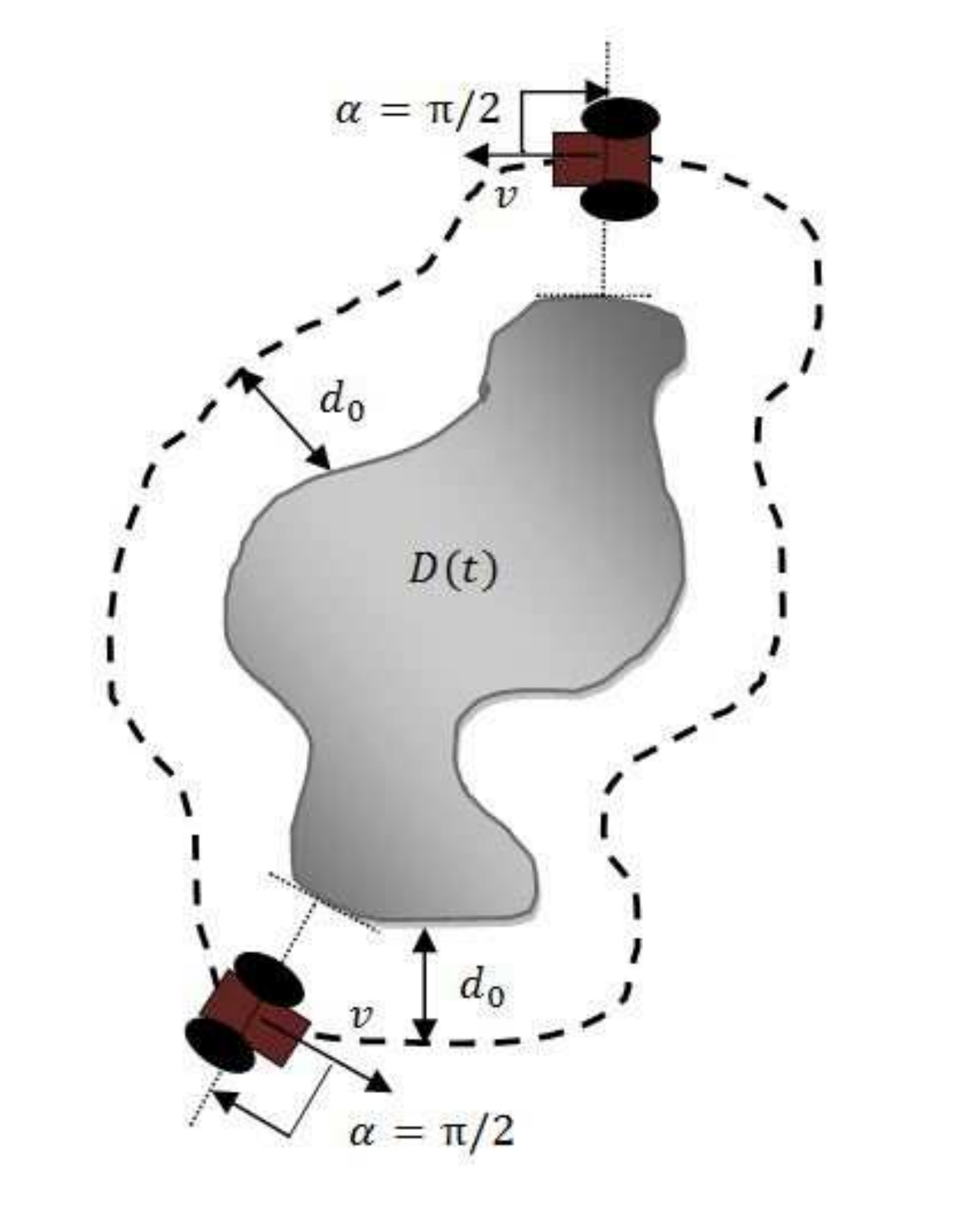}}
\label{lin}}
\caption{(a)Sliding mode maneuver; (b)linear mode maneuver}
\end{figure}

 
In more details, this motion looks as follows. The equation $\dot{d}+\chi(d-d_0)=0$ means that 
the angle $\alpha$ between the relative velocity of the mobile robot $\vec{v}_r$ and the line of sight at the nearest point of the obstacle equals $\alpha = \arccos\frac{\chi(d-d_0)}{v_r}$. It follows that the angle $\alpha$ is obtuse for $d<d_0$ and acute for $d>d_0$, and so the mobile robot is driven towards the desired distance $d_0$ to the obstacle. In doing so, the angle $\alpha$ is kept constant $\alpha = \arccos[\frac{v_\ast}{v_r} \sgn \chi(d-d_0)]$ in the saturation zone $|d-d_0| > \delta$. As $d$ leaves this zone and approaches $d_0$, the angle goes to $\frac{\pi}{2}$. 
In the limit where $d=d_0$, the mobile robot is oriented parallel to the border of the obstacle $\alpha = \pi/2$, which means traveling along the $d_0$-equidistant curve, see Fig.\ref{lin}.
\par
The above control law \eqref{avoid} is activated in a close vicinity of an en-route obstacle. Whenever the mobile robot is far away from them, it is driven towards the target in a straight line and so 
\begin{equation}
\label{pur}
u(t) = 0;
\end{equation}
Switching between two control laws \eqref{avoid} and \eqref{pur} is based on two parameters: $C>0$ --- the threshold distance at which the obstacle avoidance maneuver is activated, and $\epsilon >0$ --- the mismatch between the current and desired $d_0$ distances to the obstacle with which termination of the obstacle avoidance maneuver is permitted. Specifically, the switching rules are as follows. Switching from (\ref{pur}) to (\ref{avoid}) occurs whenever the distance to the nearest obstacle reduces to $C$; switching from (\ref{avoid}) to (\ref{pur}) occurs when $d \leq d_0+\epsilon$ and the mobile robot is oriented towards the target. This switching strategy allows the mobile robot to keep a safe distance to the obstacles and reach the target position in a dynamic environment \cite{AM12}. 
\par
\begin{figure}[h]
\centering
\includegraphics[width=1.7in]{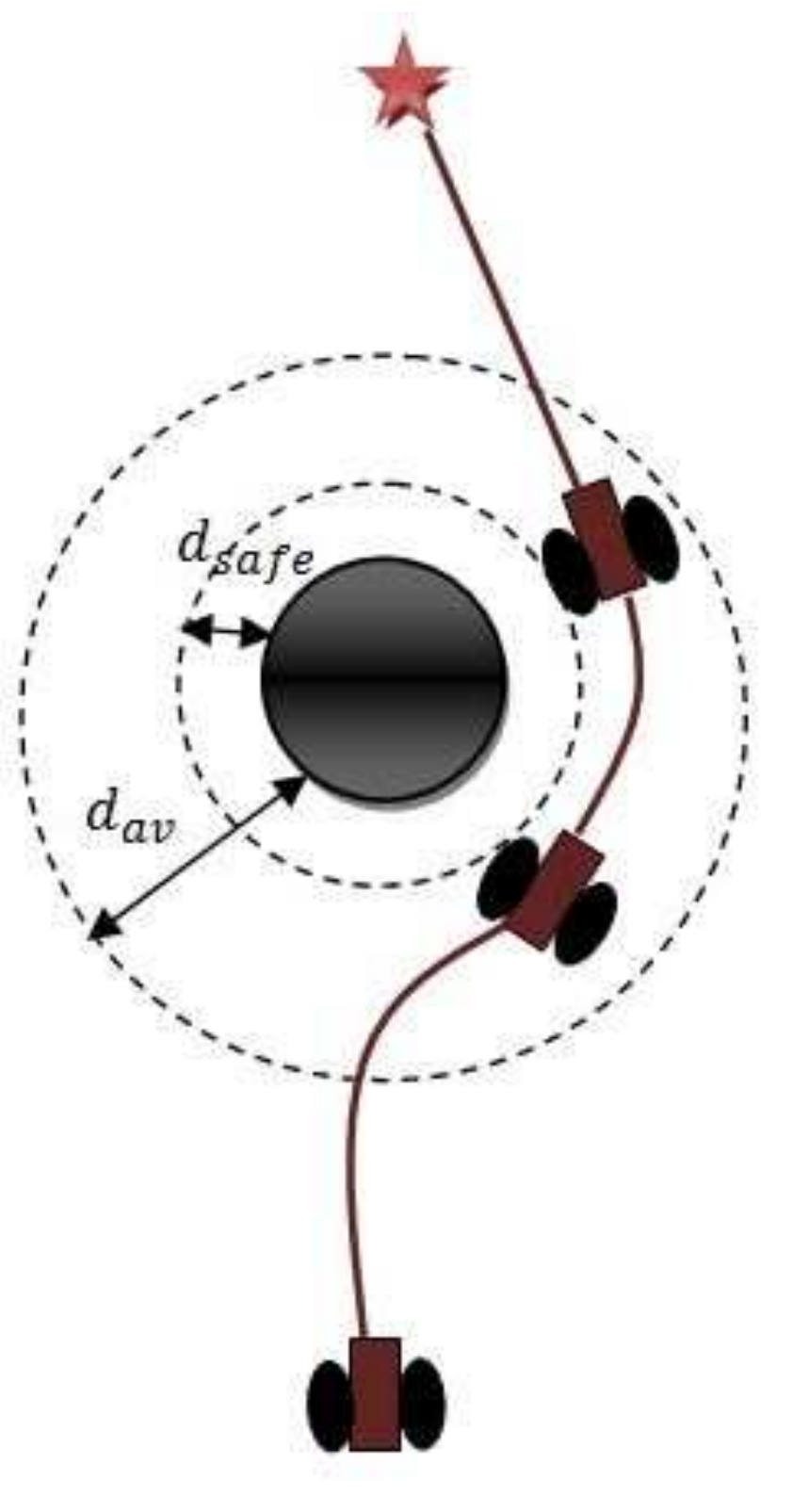}
\caption{Execution of the proposed navigation algorithm in a simple scenario}
\label{ex}
\end{figure}

An example of its execution in a simple scenario is depicted in Fig. \ref{ex}.

\par

Mathematically rigorous analysis of the presented algorithm under some technical assumptions was given in~\cite{AM12}.

\section{Computer Simulation Results}

In this section, the performance and features of the proposed navigation algorithm in uncertain dynamic environments are demonstrated by computer simulations performed in Matlab. In the following simulation figures presented, we depict the mobile robot gray disk and obstacles by black circle/block, the paths taken by the mobile robot to avoid the obstacles are shown by a series of gray disk and tis current heading is presented by an arrow.
\par

\begin{figure}[!h]
\centering
\subfigure[]{\scalebox{0.4}{\includegraphics{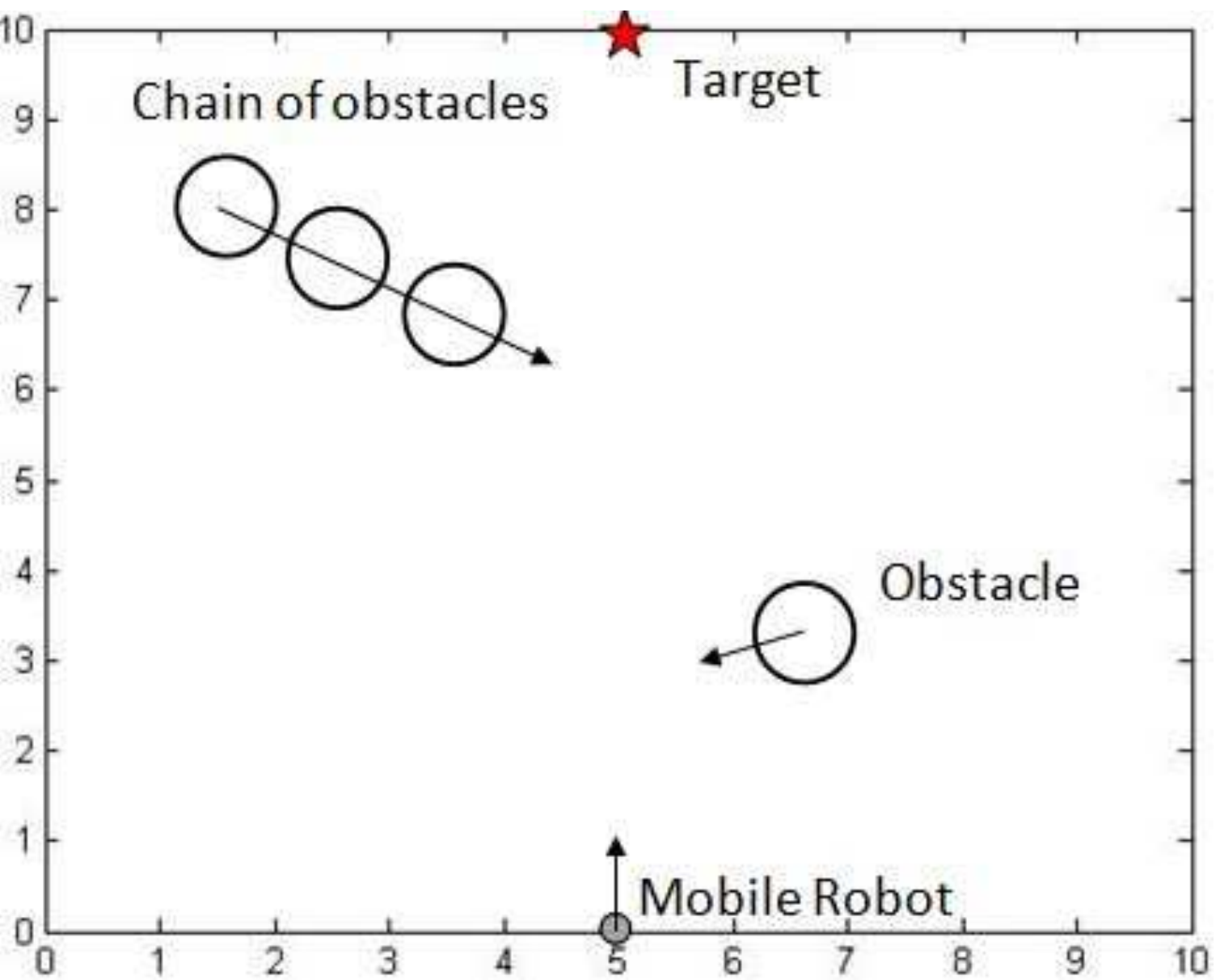}}
\label{c3.sim11}}
\subfigure[]{\scalebox{0.4}{\includegraphics{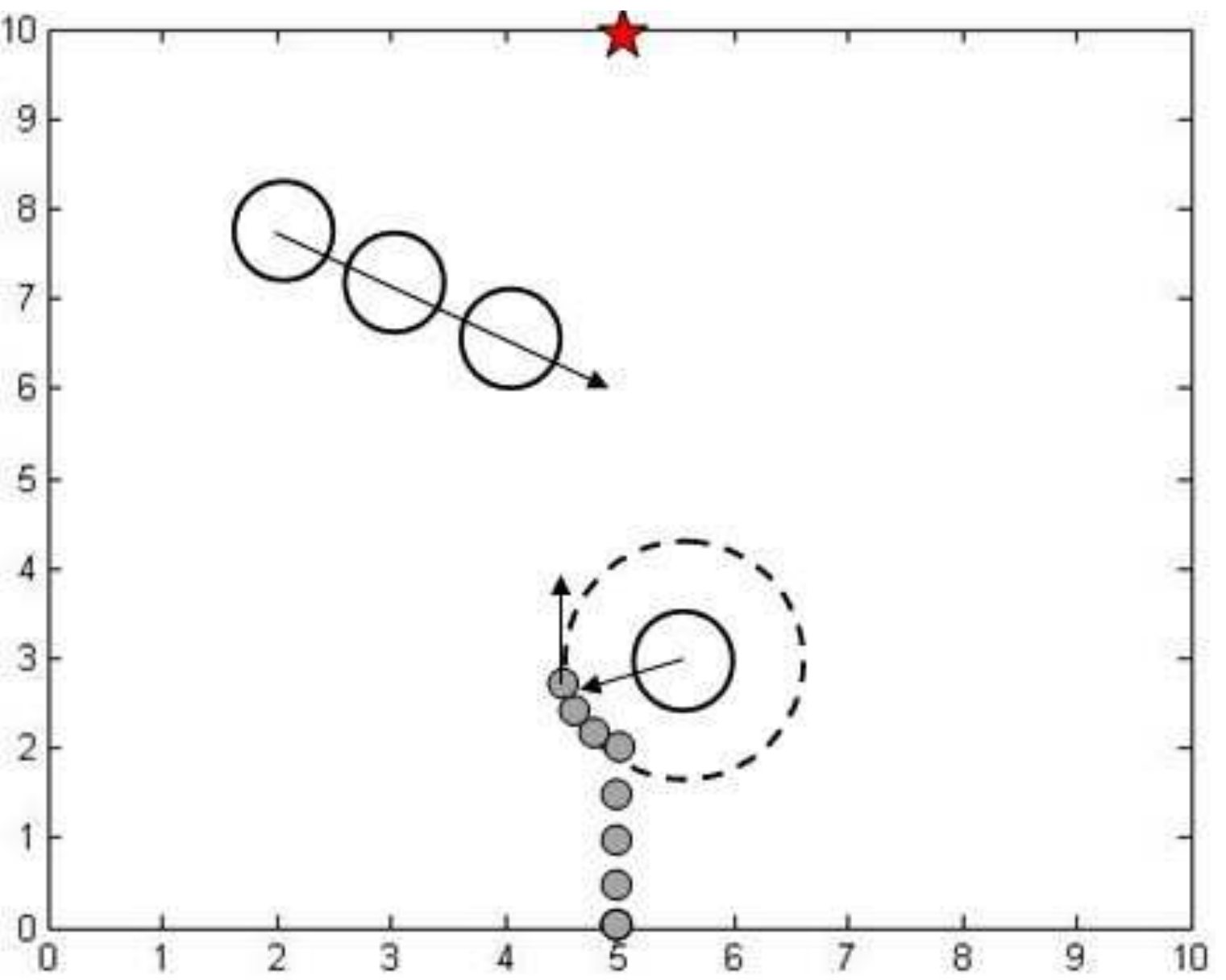}}
\label{c3.sim12}}
\subfigure[]{\scalebox{0.4}{\includegraphics{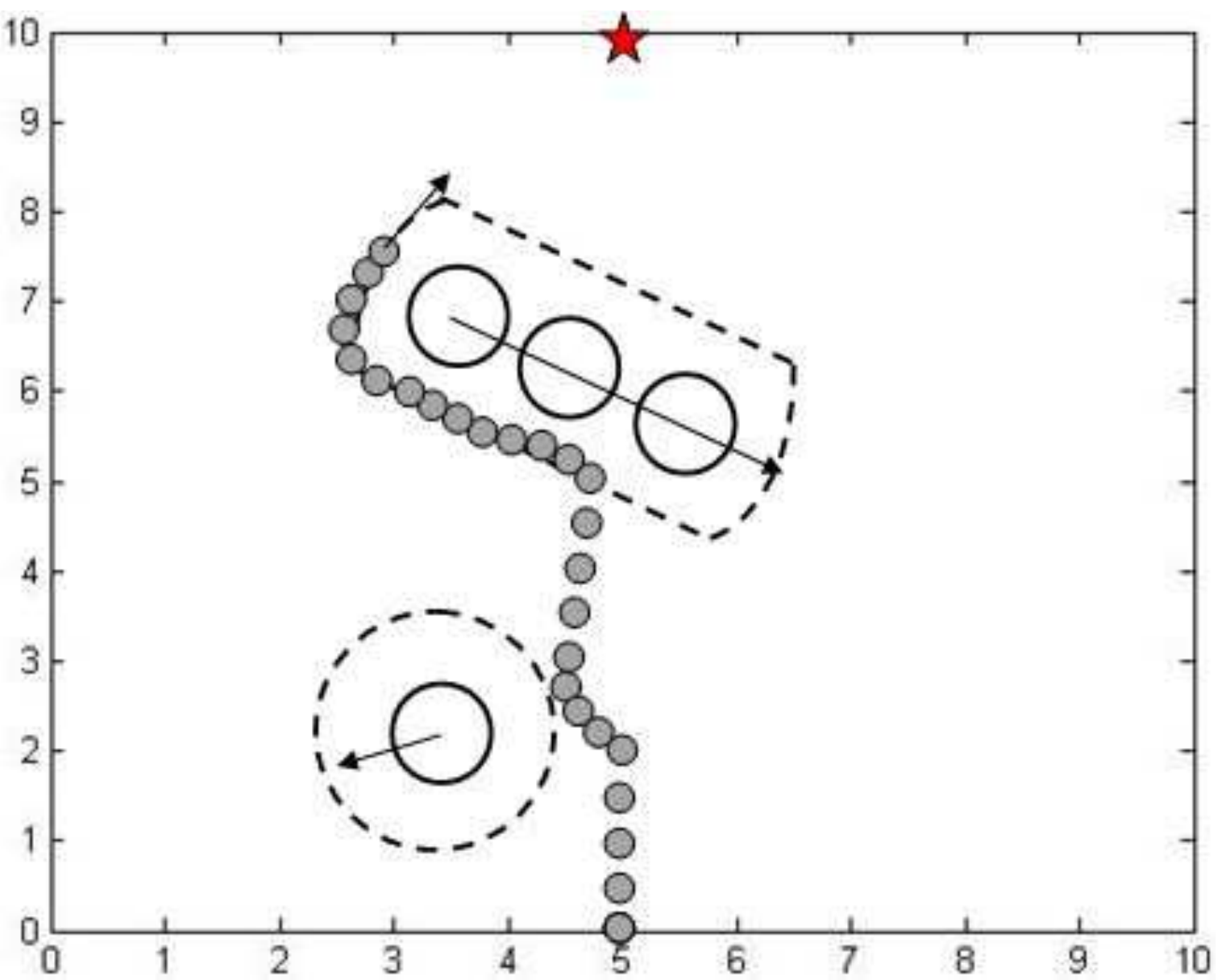}}
\label{c3.sim13}}
\subfigure[]{\scalebox{0.4}{\includegraphics{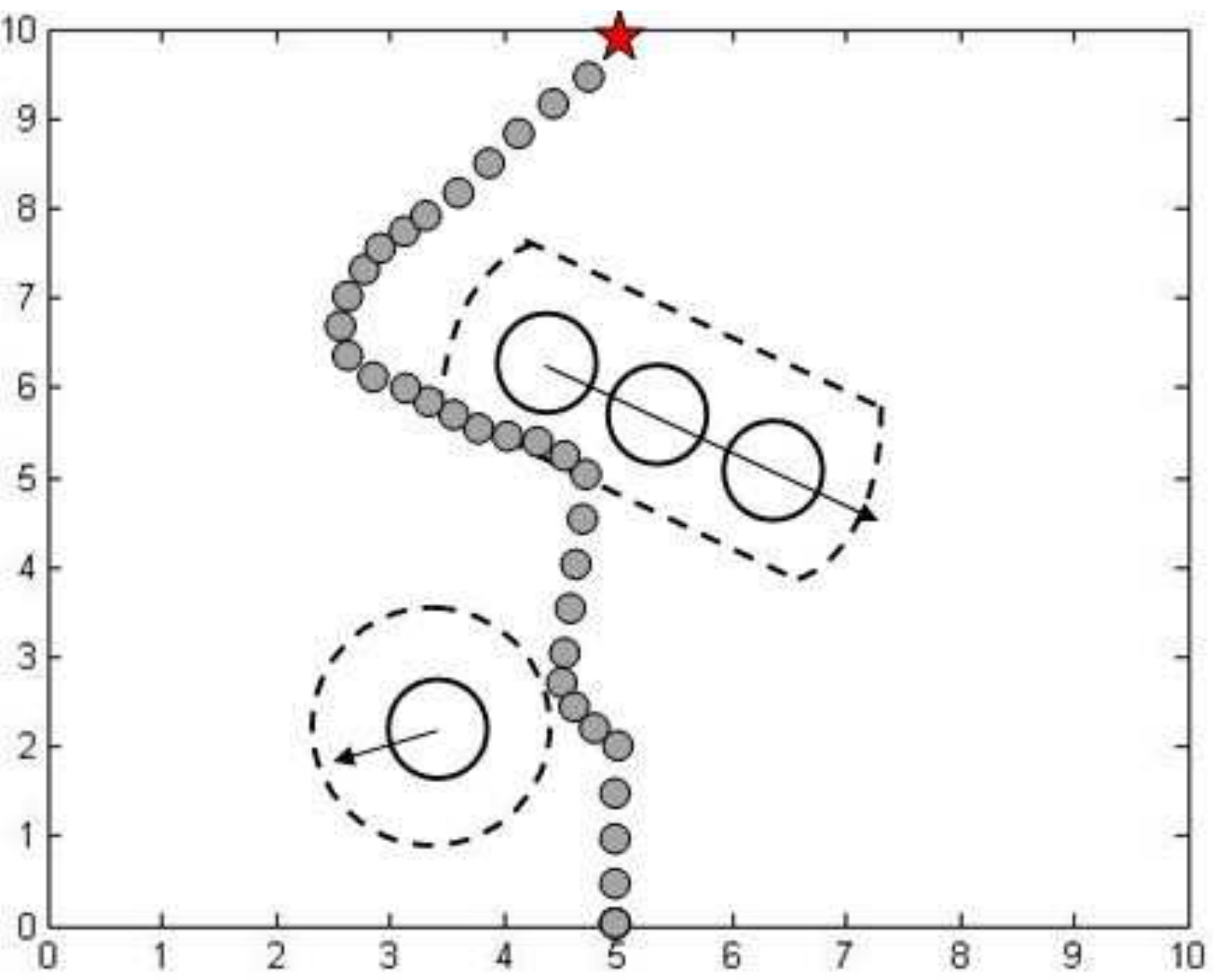}}
\label{c3.sim14}}
\subfigure[]{\scalebox{0.5}{\includegraphics{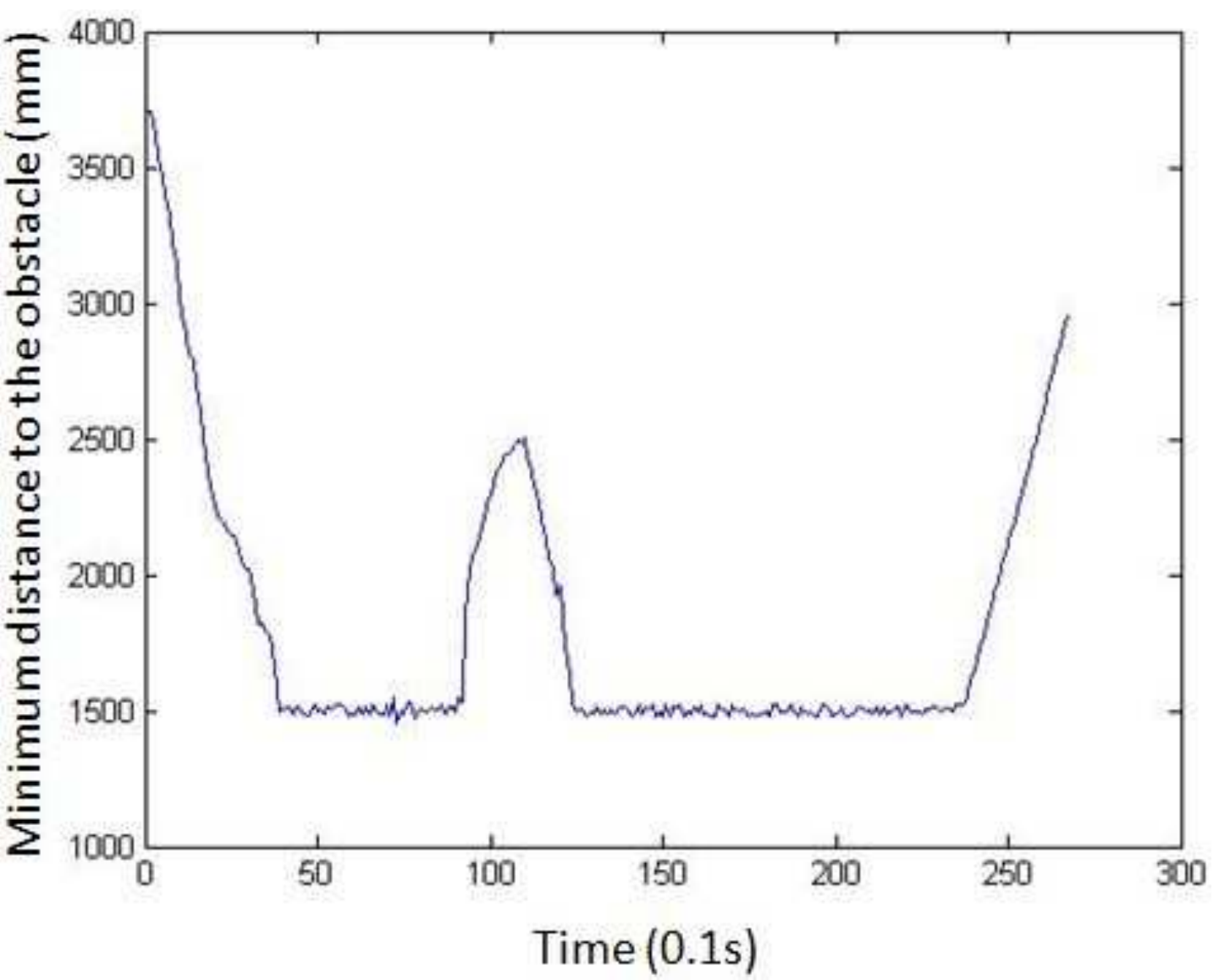}}
\label{c3.sim15}}
\caption{Mobile robot navigating in uncertain dynamic environments }
\label{c3.sim1}
\end{figure}

We first demonstrate the performance of the proposed navigation algorithm in uncertain dynamic environment, see Fig.~\ref{c3.sim1}. Notice that the robot does not require any information about the properties of the obstacles in advance. In Fig.~\ref{c3.sim11}, it can be observed that the two different types of obstacles are shown:  one single obstacle and a chain of obstacles (this group is treated as one obstacle). The initial position and moving direction of these obstacles are also shown. In Fig.~\ref{c3.sim12} and Fig.~\ref{c3.sim13}, the robot tracks the $d_0$-equidistant curves (dashed lines around the obstacles) of the obstacles regardless the shape of the obstacle when the obstacle avoidance mode is activated. The complete path taken by the robot to the target location while avoiding en-route obstacles are shown in Fig.~\ref{c3.sim14}. During the whole simulation, the only data measurable by the robot is the minimum distance between itself and the closest obstacle, this data is shown in Fig.~\ref{c3.sim15}. It can be seen that the robot maintains a constant distance ($1.5m$) to the closest obstacle during its obstacle avoidance maneuver.

\par

\begin{figure}[!h]
\centering
\subfigure[]{\scalebox{0.4}{\includegraphics{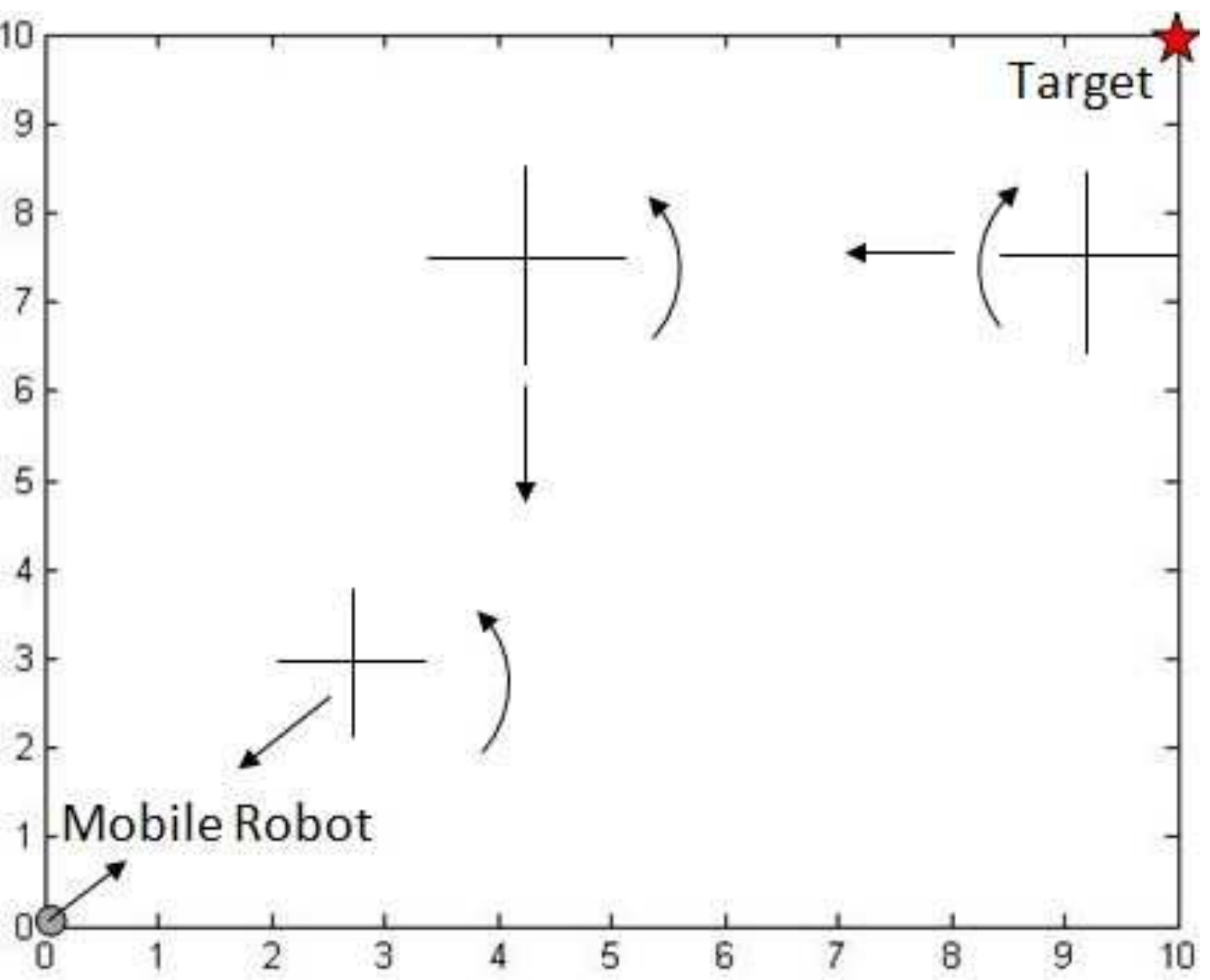}}
\label{c3.sim21}}
\subfigure[]{\scalebox{0.4}{\includegraphics{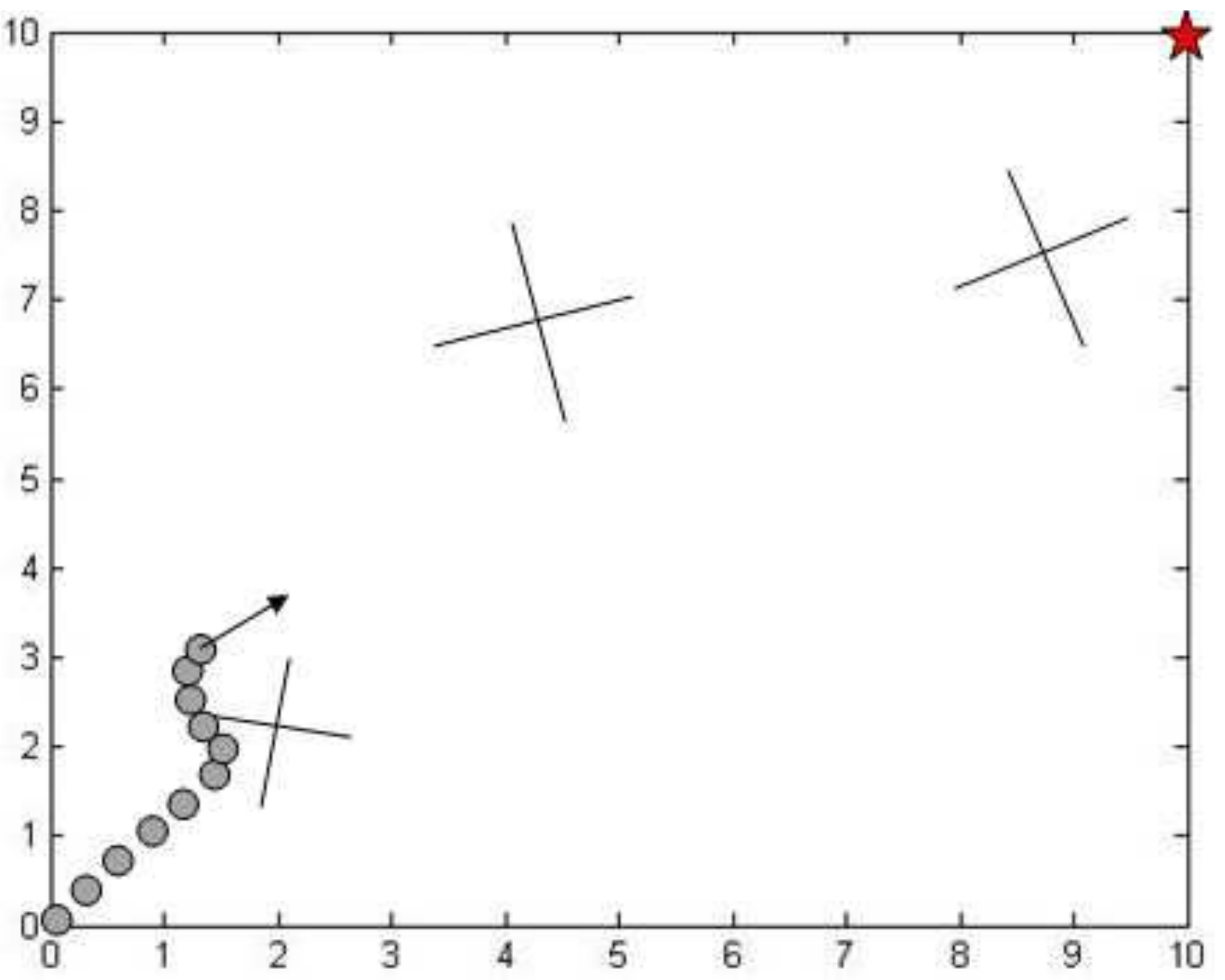}}
\label{c3.sim22}}
\subfigure[]{\scalebox{0.4}{\includegraphics{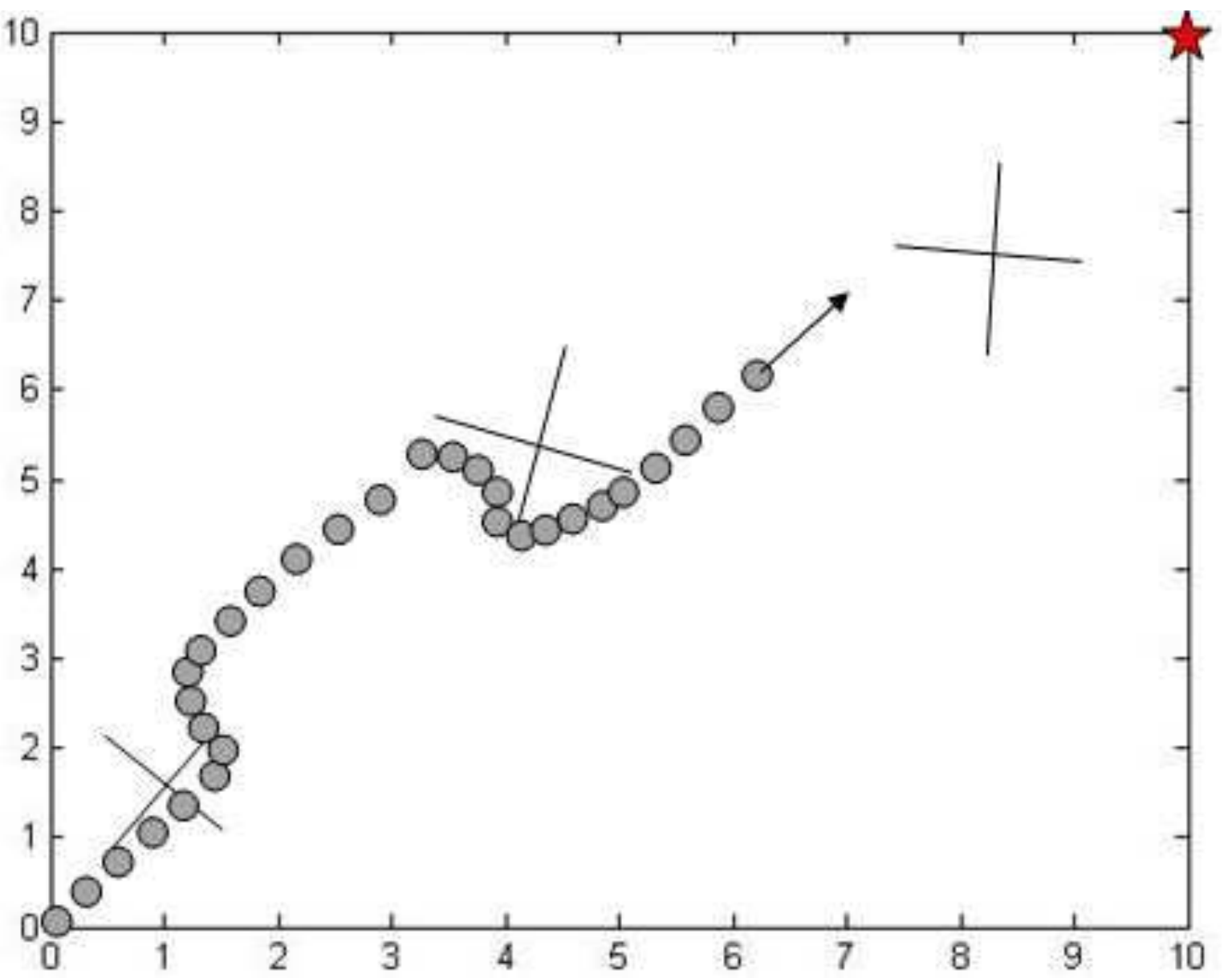}}
\label{c3.sim23}}
\subfigure[]{\scalebox{0.4}{\includegraphics{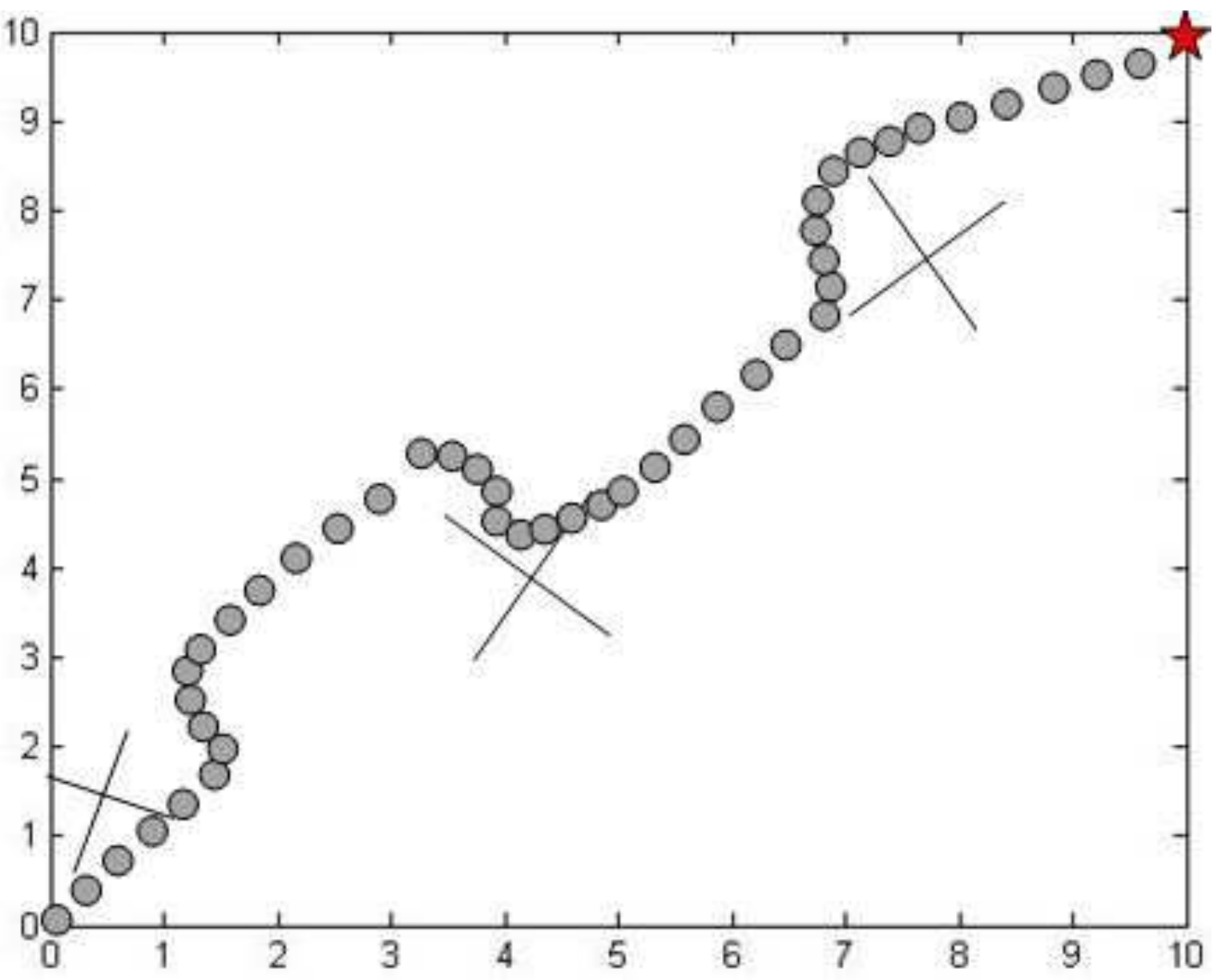}}
\label{c3.sim24}}
\caption{Mobile robot avoids moving and rotating obstacles}
\label{c3.sim2}
\end{figure}

We enhance the mobility of the obstacle in the next simulation. In Fig.~\ref{c3.sim2}, the obstacles (crosses) are not only moving in straight line but also rotating slowly around their center points, the direction of their movement and rotation are depicted in Fig.~\ref{c3.sim21}. In order to avoid these obstacles, the proposed navigation algorithm is able to adapt the rotational movement of the obstacle and keep itself away from any part of the obstacle, especially the "wings" of the crosses. The robot successfully avoids the obstacles and arrives the target position in Fig.~\ref{c3.sim22}, Fig.~\ref{c3.sim23} and Fig.~\ref{c3.sim24}. This simulation shows that the proposed navigation algorithm can cope with obstacles undergoing complex motions.
\par

\par

\begin{figure}[!h]
\centering
\subfigure[]{\scalebox{0.4}{\includegraphics{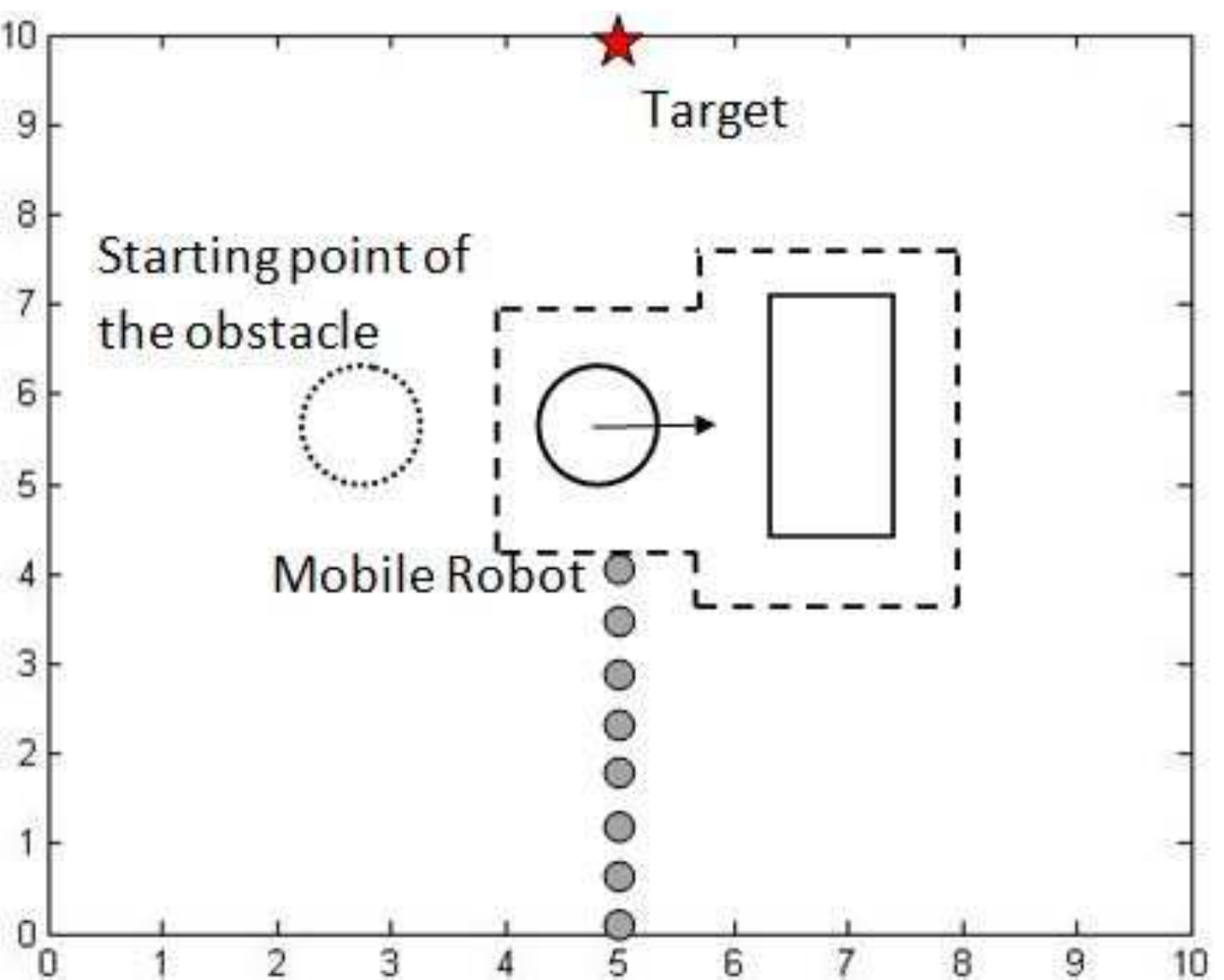}}
\label{c3.sim31}}
\subfigure[]{\scalebox{0.4}{\includegraphics{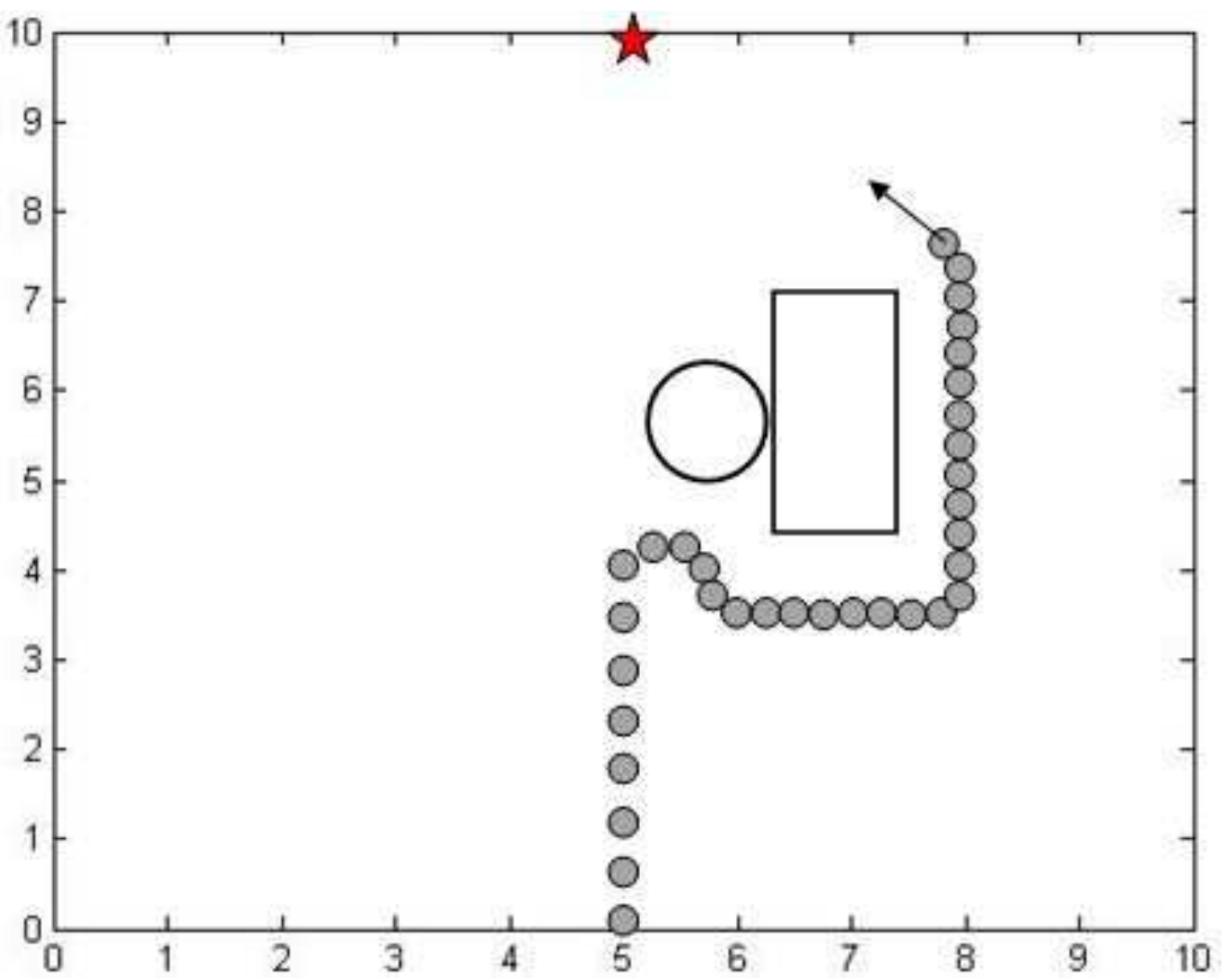}}
\label{c3.sim32}}
\subfigure[]{\scalebox{0.4}{\includegraphics{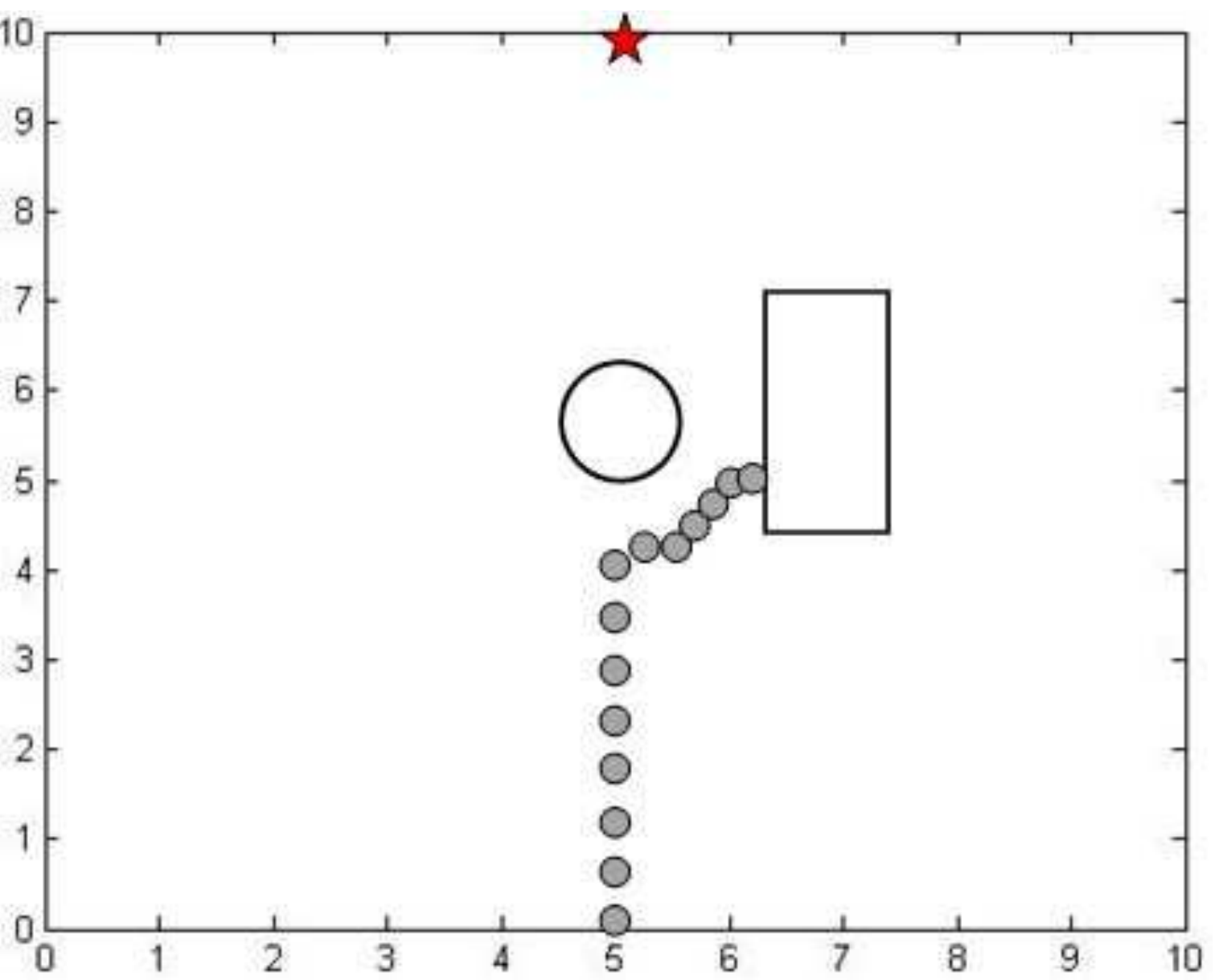}}
\label{c3.sim33}}
\caption{The implementation of interpolation technique with the proposed navigation algorithm}
\label{c3.sim3}
\end{figure}

The performance of our proposed navigation algorithm can be improved by the on-the-fly interpolation technique, which treats several closely positioned obstacles one single deformed object. This technique gives the proposed algorithm extra flexibility in the scenes with multiple stationary and moving obstacles,  such as corridor with multiple pedestrians and train stations with many passengers. The utilisation of the interpolation technique together with the proposed navigation is presented in Fig.~\ref{c3.sim3}, where the mobile robot is facing one moving obstacle which travels towards a stationary block obstacle. In Fig.~\ref{c3.sim31}, the mobile robot treats these two obstacles as one single obstacle and avoids it in a path shown in Fig.~\ref{c3.sim32}. If the interpolation technique is not put to use in this case, the robot will avoid the moving obstacle once the distance between them is reduced below threshold distance $C$, and ignore the existence of the static obstacle, which will result a collision as shown in Fig.~\ref{c3.sim33}.

\par

The next simulation result demonstrate the ability of the proposed navigation algorithm to avoid dynamic deforming obstacles. The dynamic deforming obstacles with time-varying shapes covers many complex real life scenarios. For example, scenarios with reconfigurable rigid obstacles, rigid obstacles with large moving parts, forbidden zones between moving obstacles, like inter-vehicle those in a dense platoon, flexible underwater obstacles, like fishing nets, schools of big fish, bunches of cables, etc., and virtual obstacles, like areas corrupted with hazardous chemicals, vapor, radiation, high turbulence etc. or on-line estimated areas of operation of a hostile agent.
\par
\begin{figure}[!h]
\centering
\subfigure[]{\scalebox{0.4}{\includegraphics{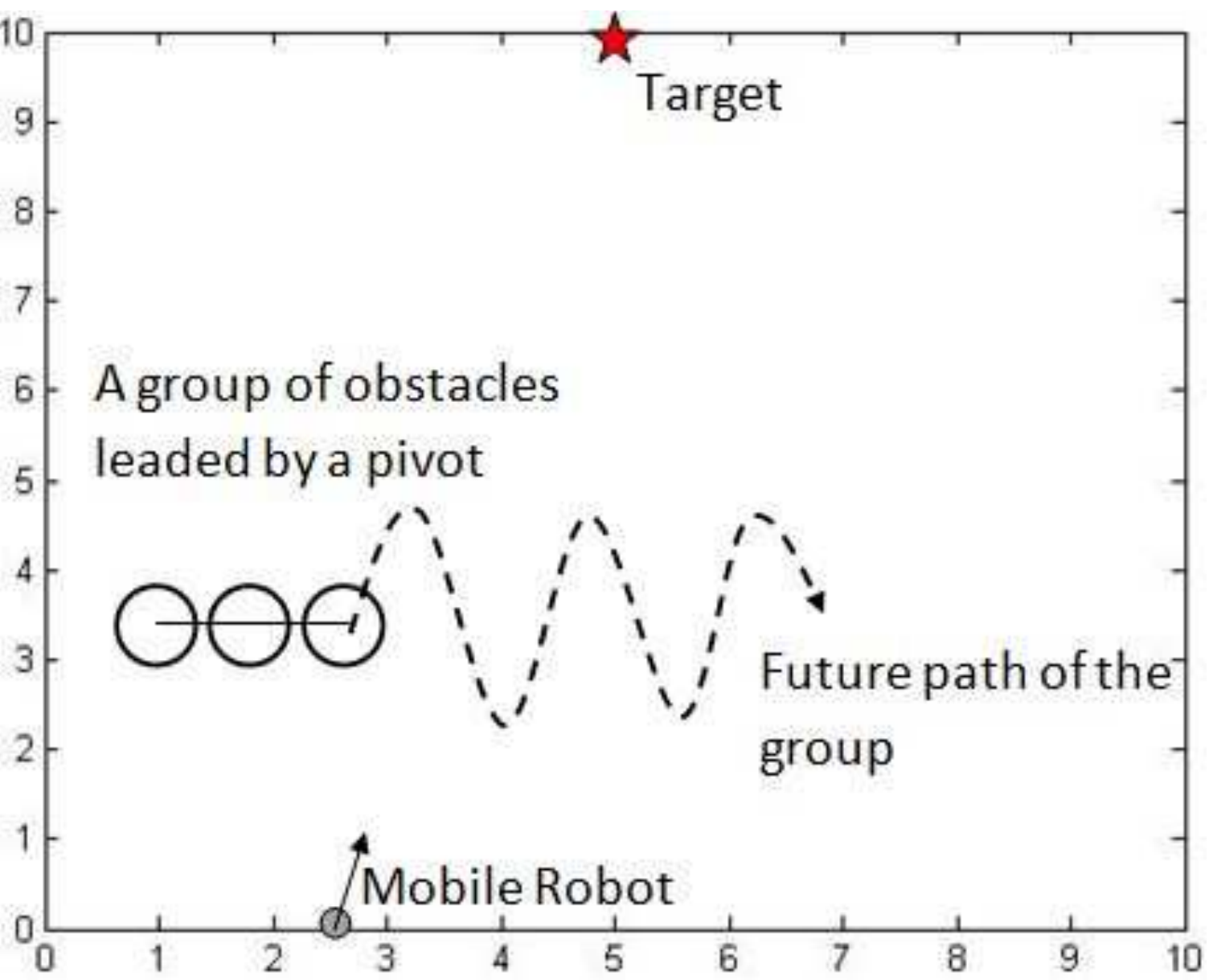}}
\label{c3.sim41}}
\subfigure[]{\scalebox{0.4}{\includegraphics{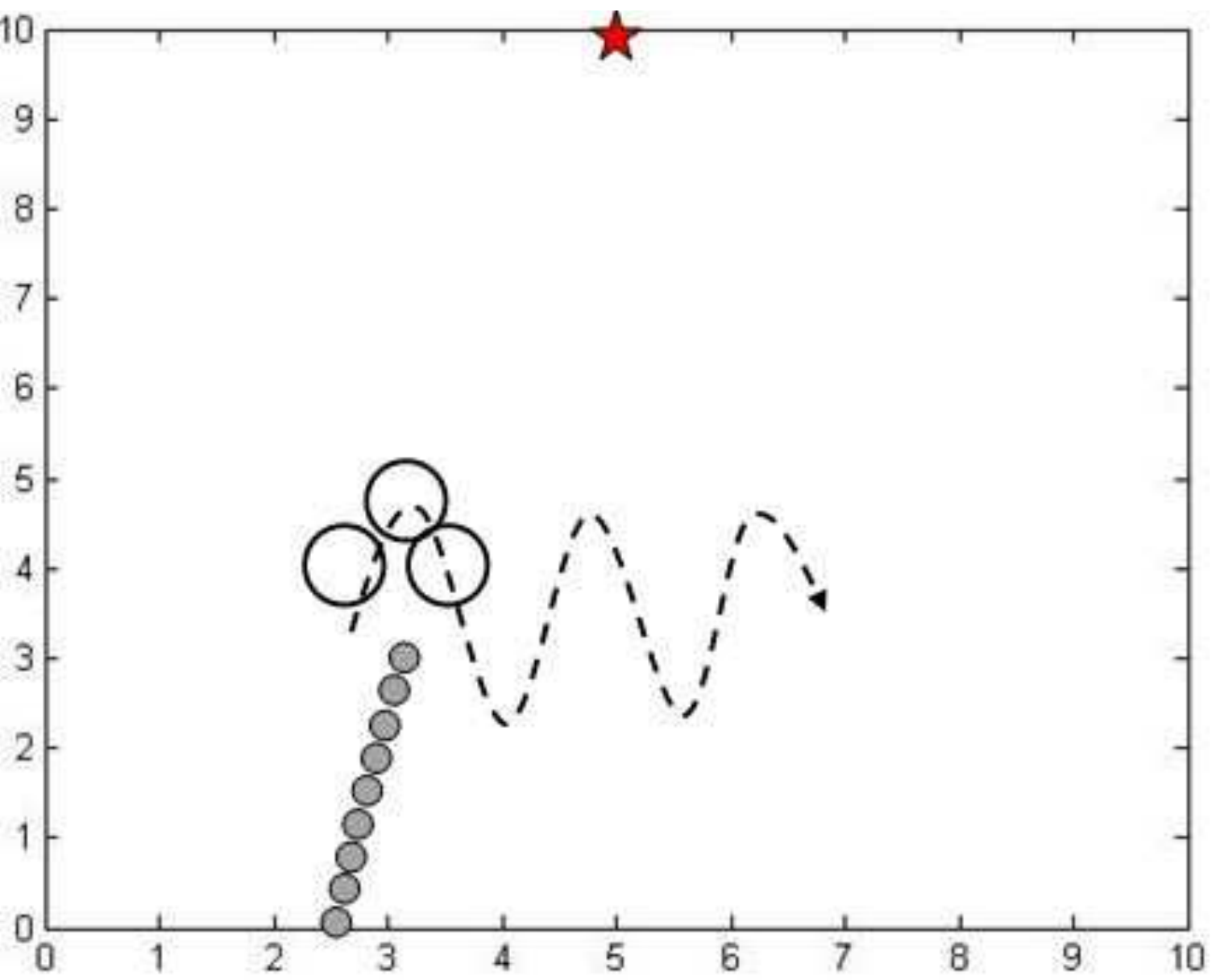}}
\label{c3.sim42}}
\subfigure[]{\scalebox{0.4}{\includegraphics{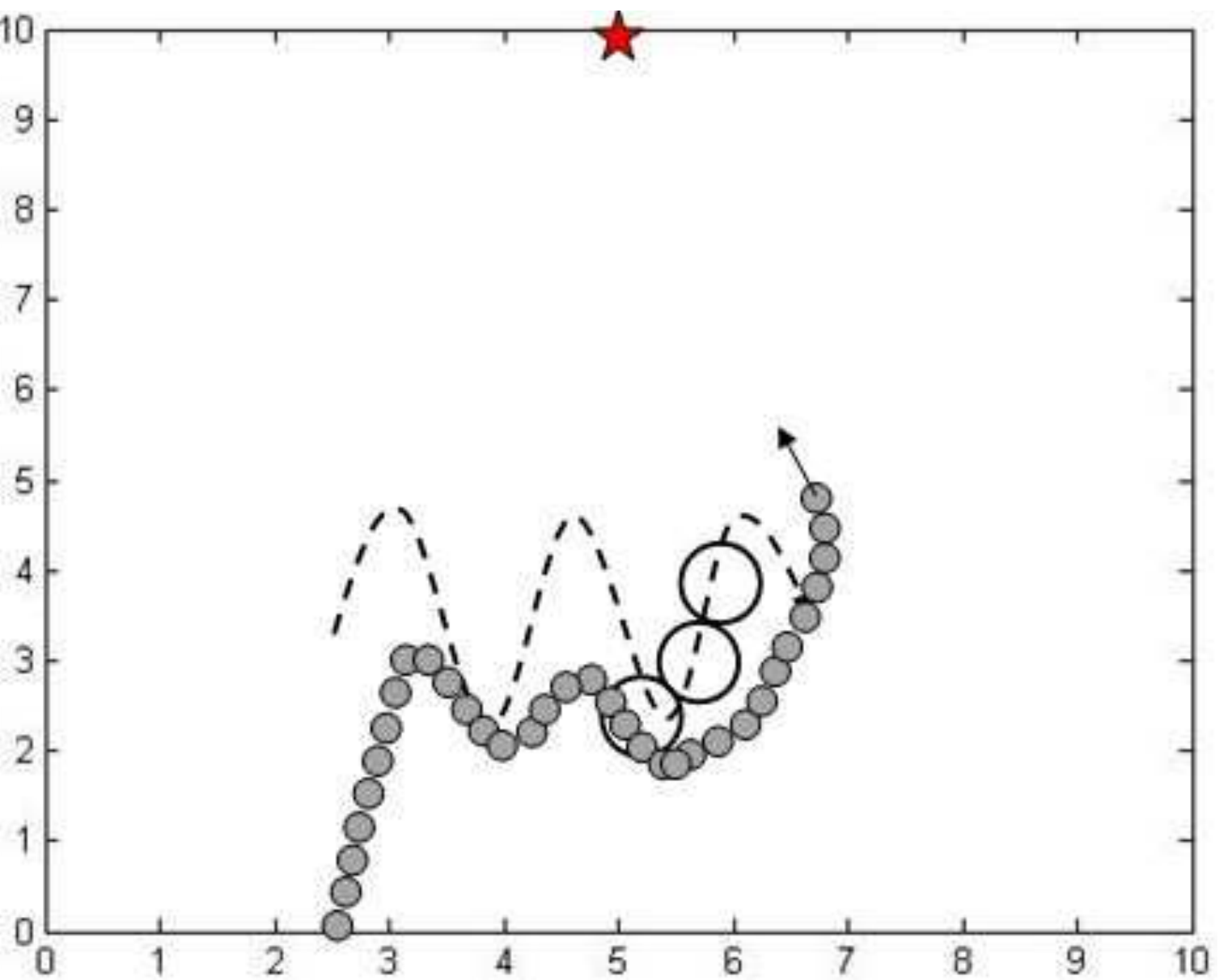}}
\label{c3.sim43}}
\caption{Mobile robot avoids dynamic deforming obstacle}
\label{c3.sim4}
\end{figure}

In Fig.~\ref{c3.sim4}, a group of three obstacles are interpolated as one obstacle, and its movement is led by the first obstacle (the leader), the obstacle moves in a sinusoidal fashion as shown in Fig.~\ref{c3.sim41}. Fig.\ref{c3.sim42} and Fig.\ref{c3.sim43} show the moments for which the mobile robot avoids this dynamic deforming obstacle, notice that the shape of the obstacle is different from each other in each of the figures.

\par

\begin {figure}
\begin{center}
\subfigure[]{\scalebox{0.42}{\includegraphics{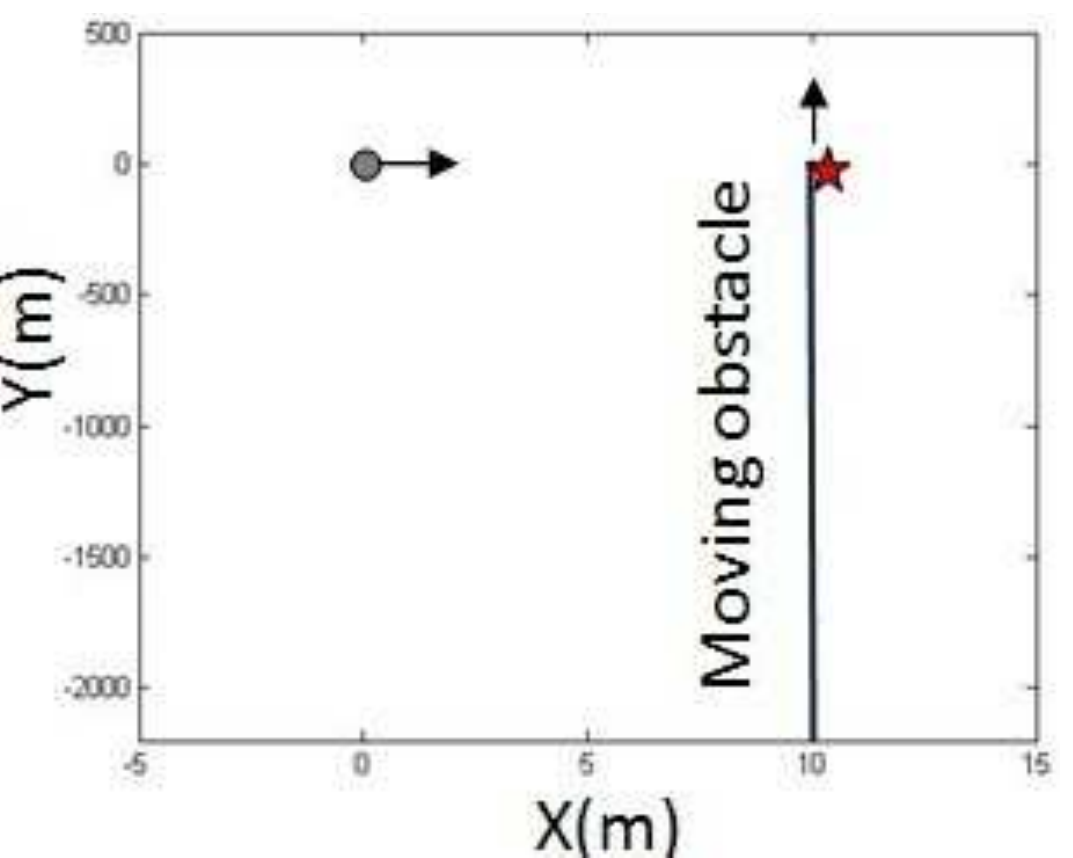}}
\label{c3.sim51}}
\subfigure[]{\scalebox{0.42}{\includegraphics{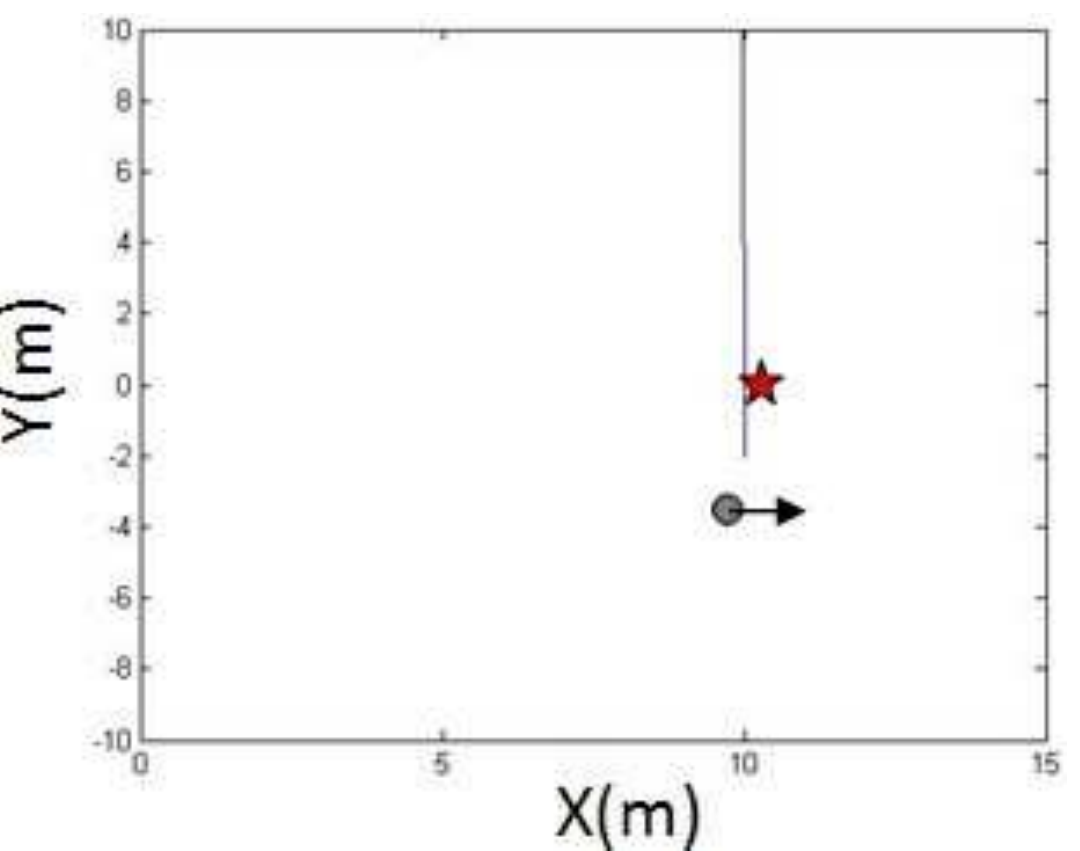}}
\label{c3.sim53}}
\subfigure[]{\scalebox{0.42}{\includegraphics{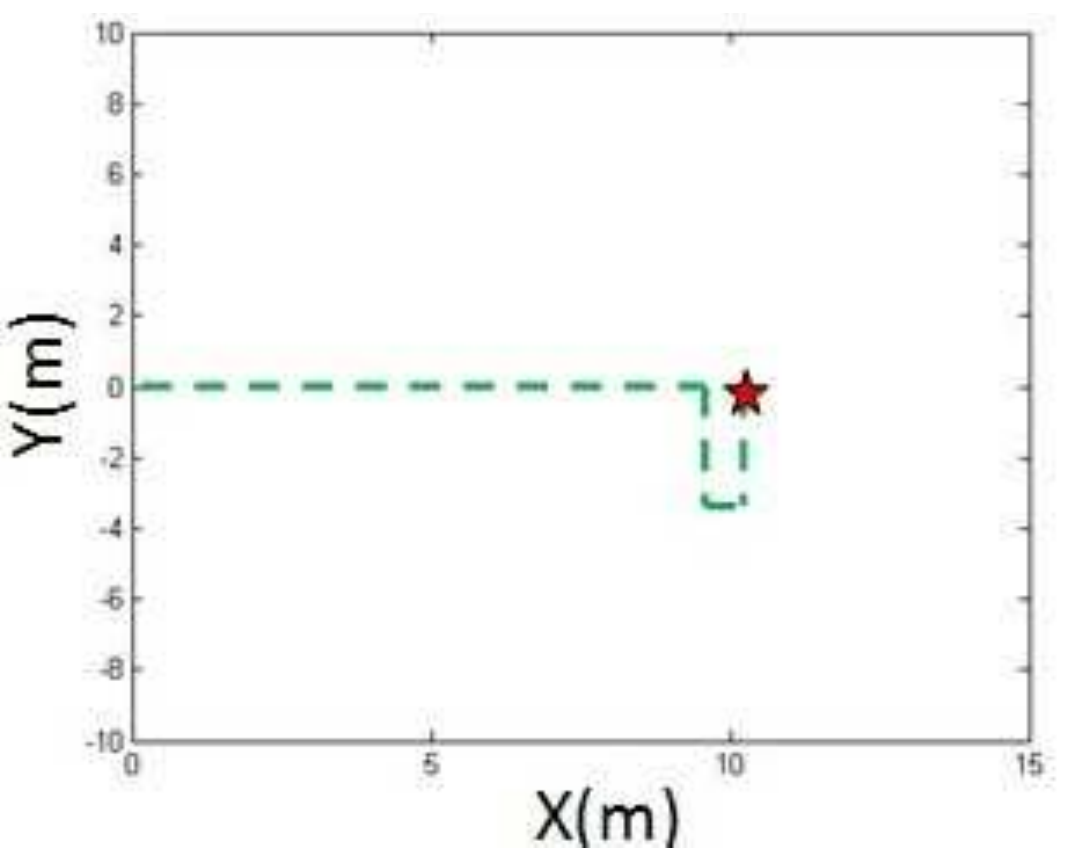}}
\label{c3.sim55}}
\subfigure[]{\scalebox{0.42}{\includegraphics{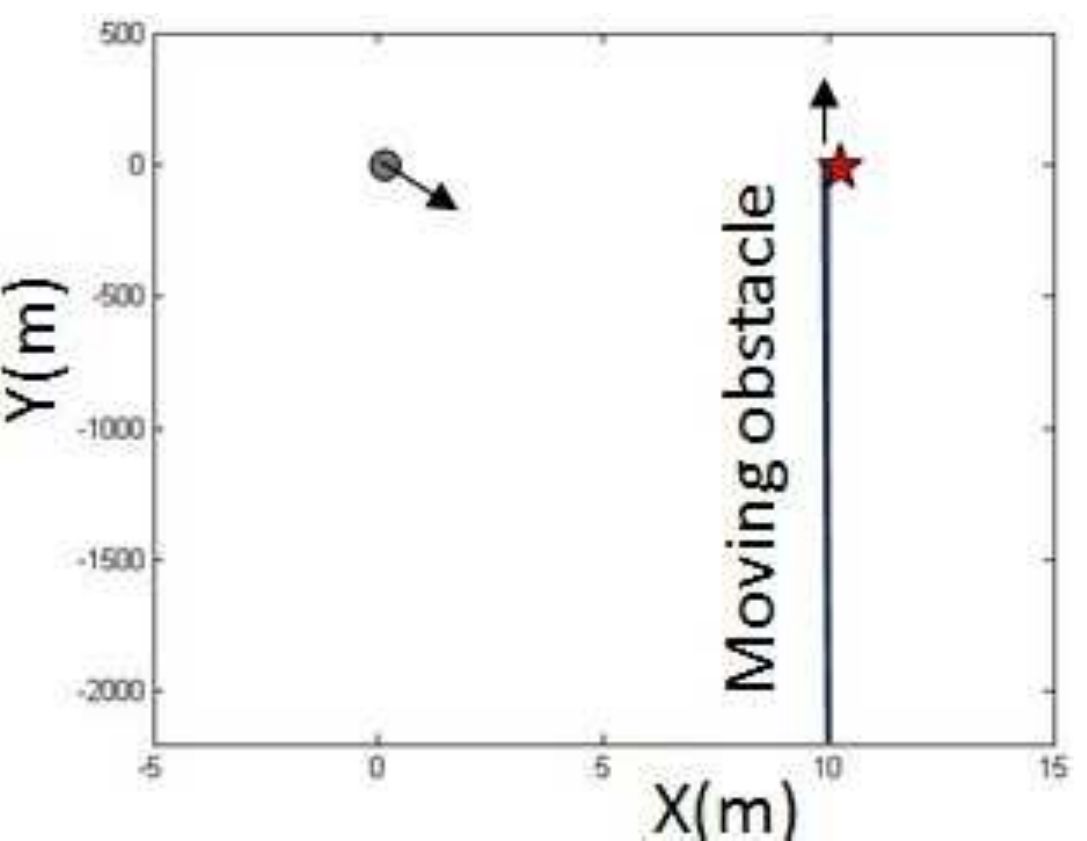}}
\label{c3.sim52}}
\subfigure[]{\scalebox{0.42}{\includegraphics{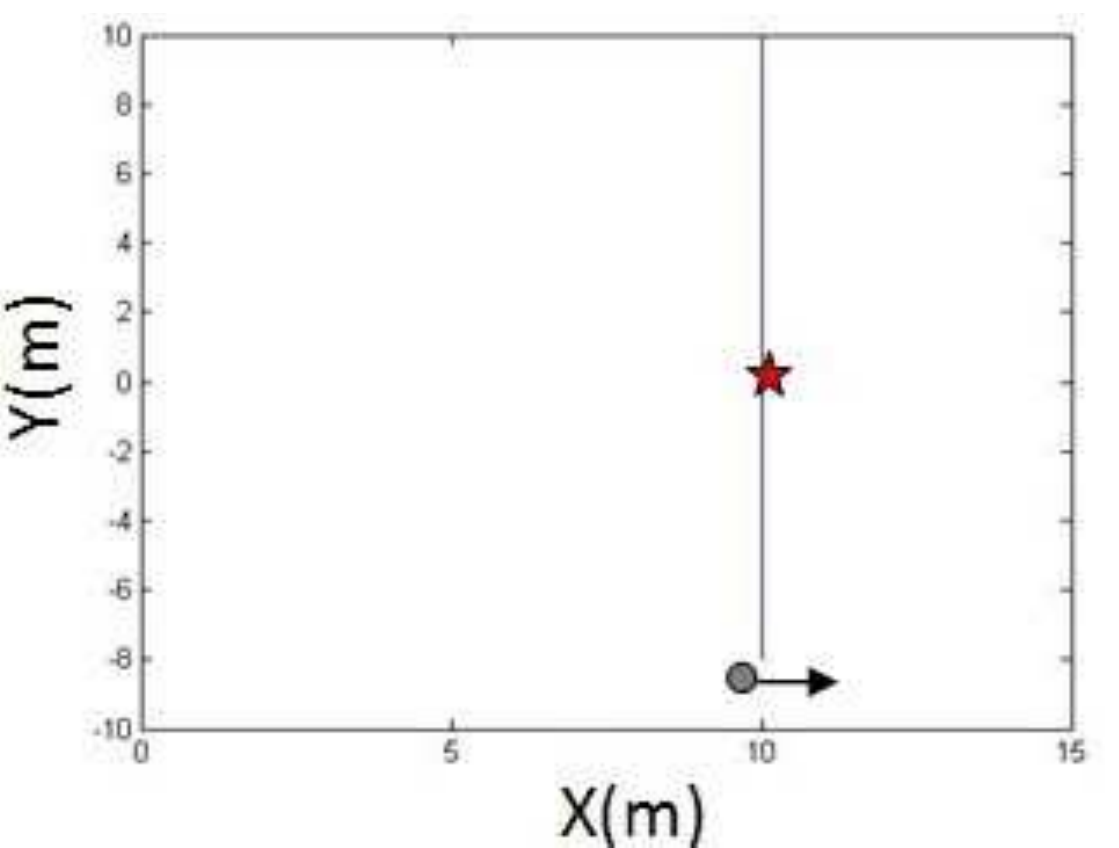}}
\label{c3.sim54}}
\subfigure[]{\scalebox{0.42}{\includegraphics{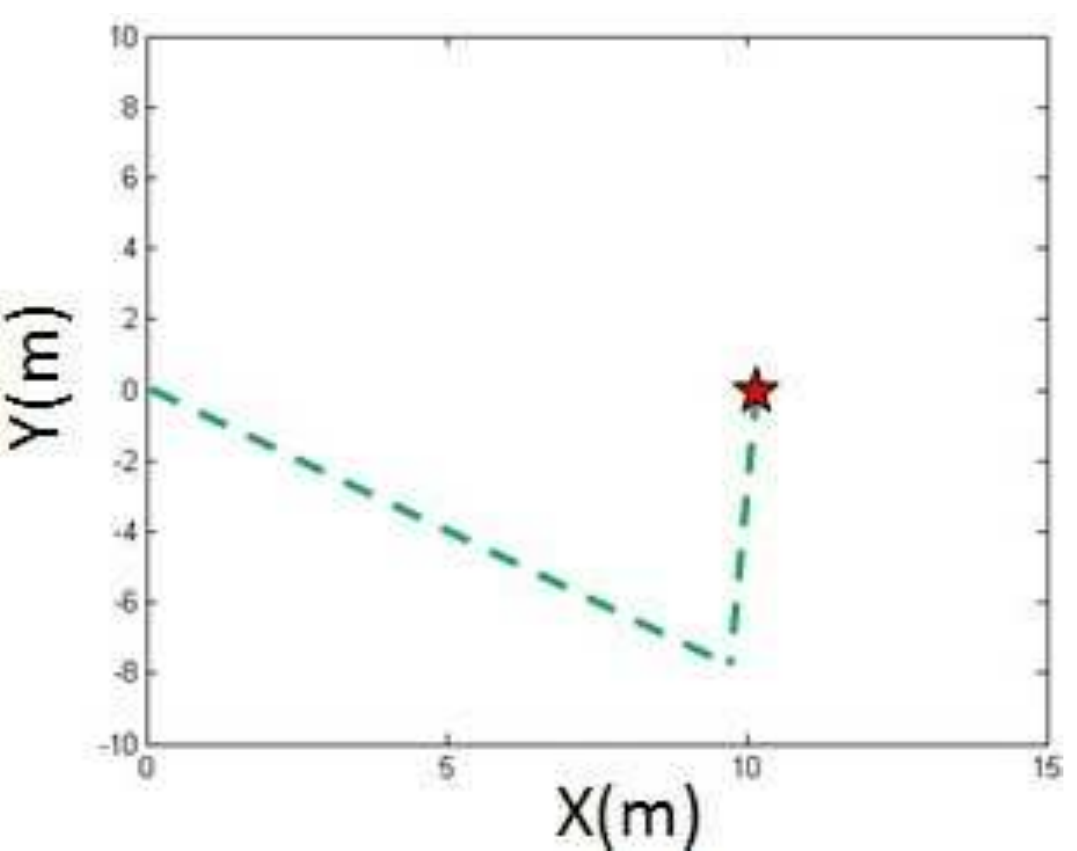}}
\label{c3.sim56}}
\caption{Proposed navigation algorithm (upper row) versus VOA (lower row)}
\label{c3_sim5}
\end{center}
\end{figure}

The last simulation result presents the comparison between the performance of the proposed navigation algorithm with that of Velocity Obstacle Approach (VOA) \cite{FS98,LaLaSh05} in the scenario where a long obstacle perpendicularly intervenes between the robot and the target, which is hidden just behind the obstacle, see Fig.~\ref{c3.sim51} and Fig.~\ref{c3.sim52}. Fig.~\ref{c3.sim53} and Fig.~\ref{c3.sim54} display the situations at the moments when the robot transverses the path of the obstacle. The entire paths of the robot during target reaching with obstacle avoidance are depicted in Fig.~\ref{c3.sim55} and Fig.~\ref{c3.sim56}.
(We considered VOA with two choices of velocity per maneuver, with the second choice being made at the favorable moment of bypassing the obstacle.) The proposed navigation algorithm first drives the robot directly to the target in a straight line until the trigger threshold is trespassed and then undertakes a relatively small de-tour by following the obstacle boundary. VOA basically drives the robot to the target along two straight lines with bypass of the obstacle in a close range at the rear.
Thus the both guidance laws do drive the robot to the target. However PGL does this faster: the ratio of the maneuver times is as follows
$\frac{T_{\text{Propsed}}}{T_{\text{VOA}}}= \frac{14.51}{18.61} \approx 0.78$. Thus in this experiment, the proposed navigation algorithm outperforms VOA.

\section{Experiments with Real Mobile Robot}

The performance and applicability of the proposed navigation algorithm is examined on ActivMedia Pioneer 3-DX (P3) robot.
\par
In these experiments, the only information available to the mobile robot P3, which is the distance between itself and the closest obstacle, is measured by the SICK laser range (for the front area) finder and ultrasonic sensors (for the rear area). Any other information about the obstacles is not known to P3 prior and during the experiments. 
\par

\begin{figure}[!h]
\centering
\subfigure[]{\scalebox{0.4}{\includegraphics{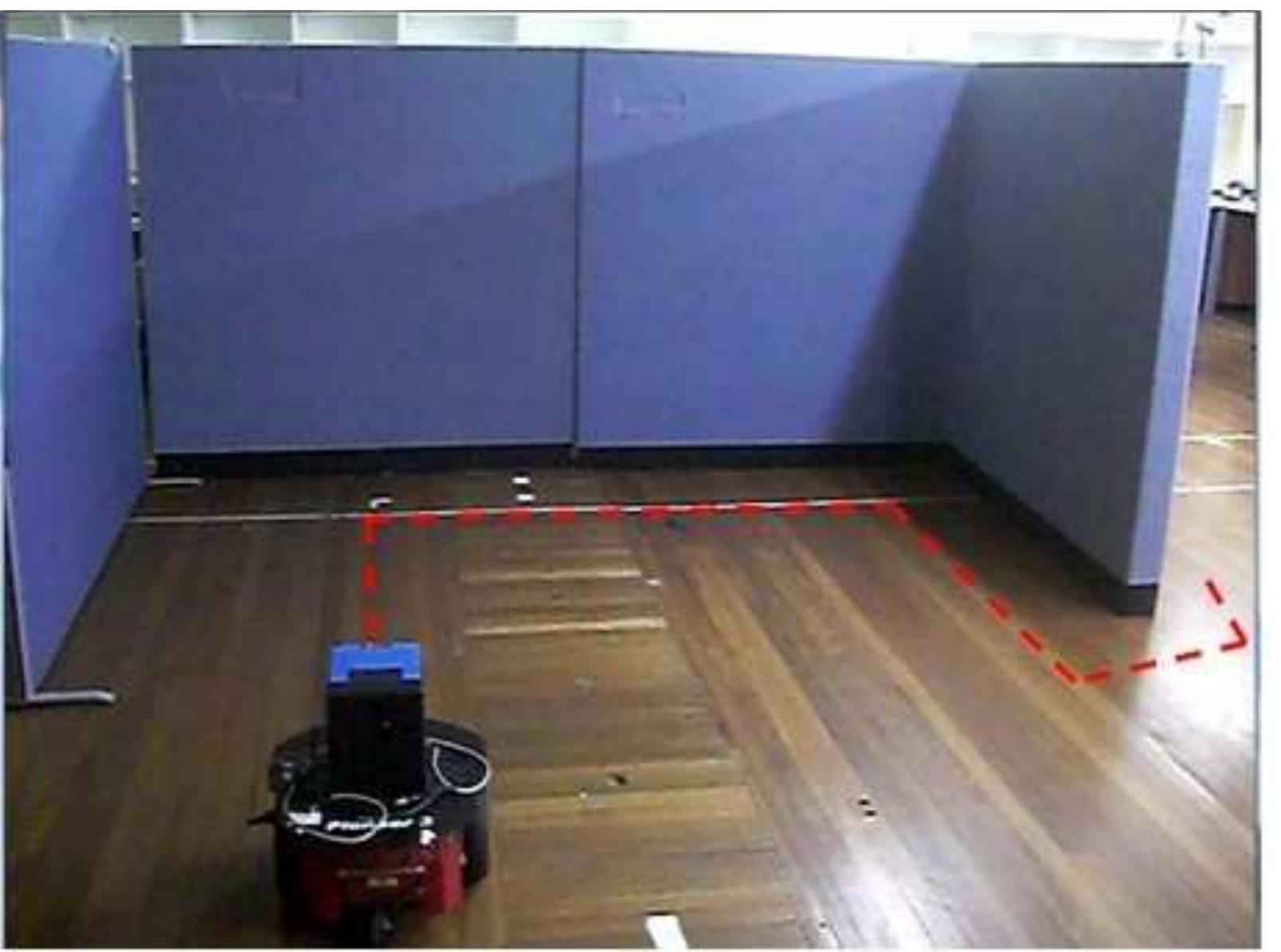}}
\label{c3.exp41}}
\subfigure[]{\scalebox{0.4}{\includegraphics{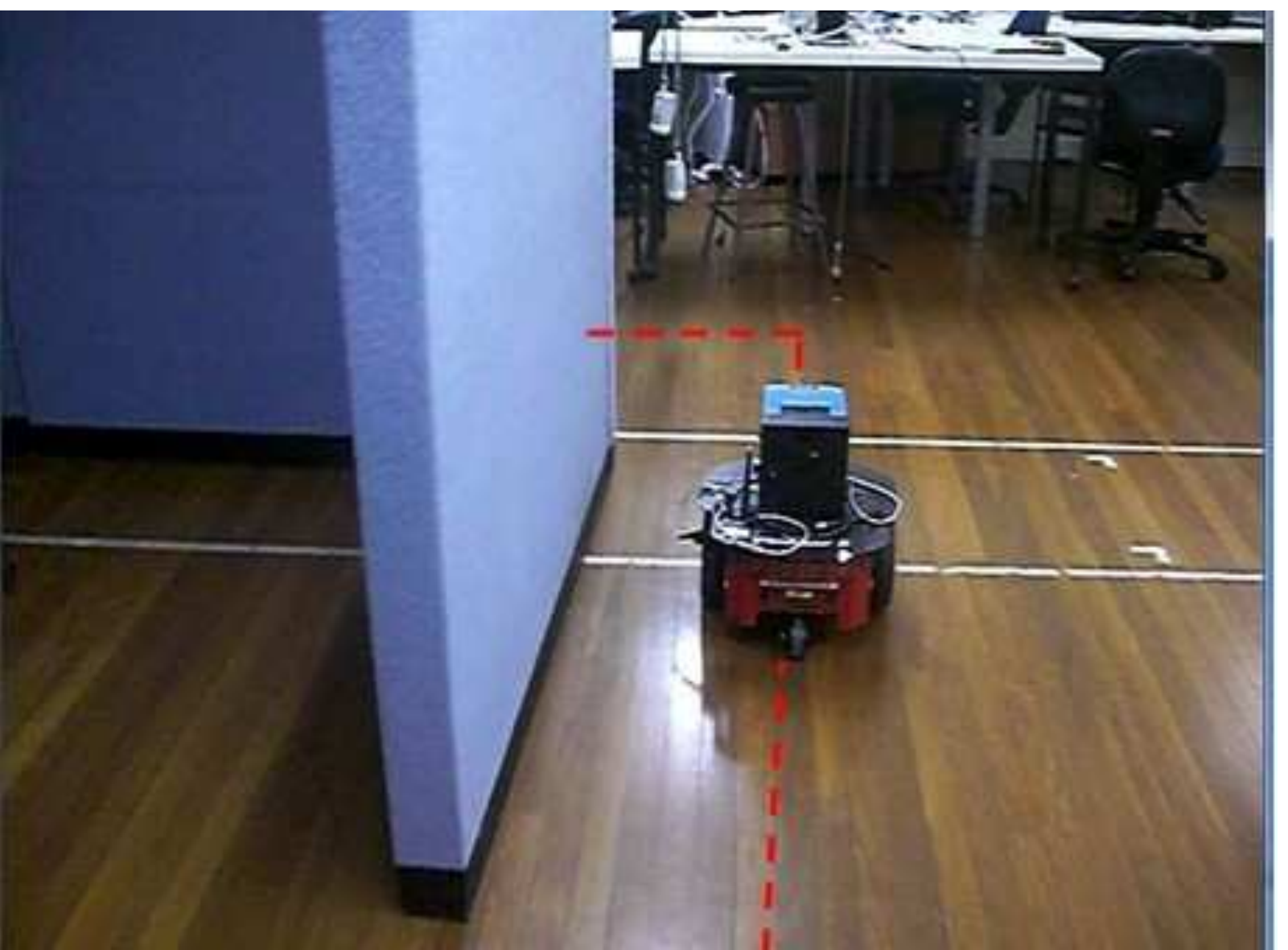}}
\label{c3.exp42}}
\subfigure[]{\scalebox{0.4}{\includegraphics{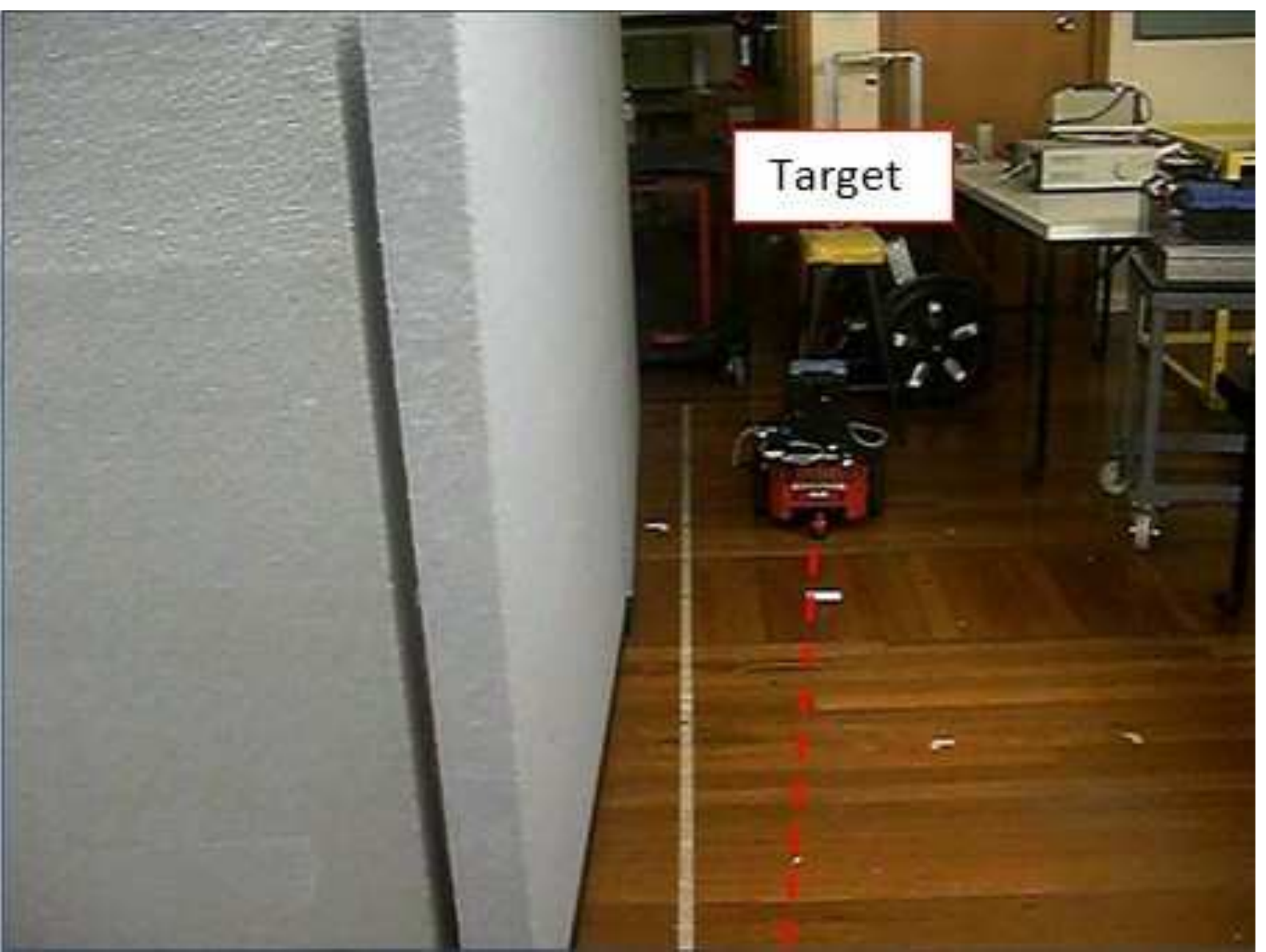}}
\label{c3.exp43}}
\subfigure[]{\scalebox{0.6}{\includegraphics{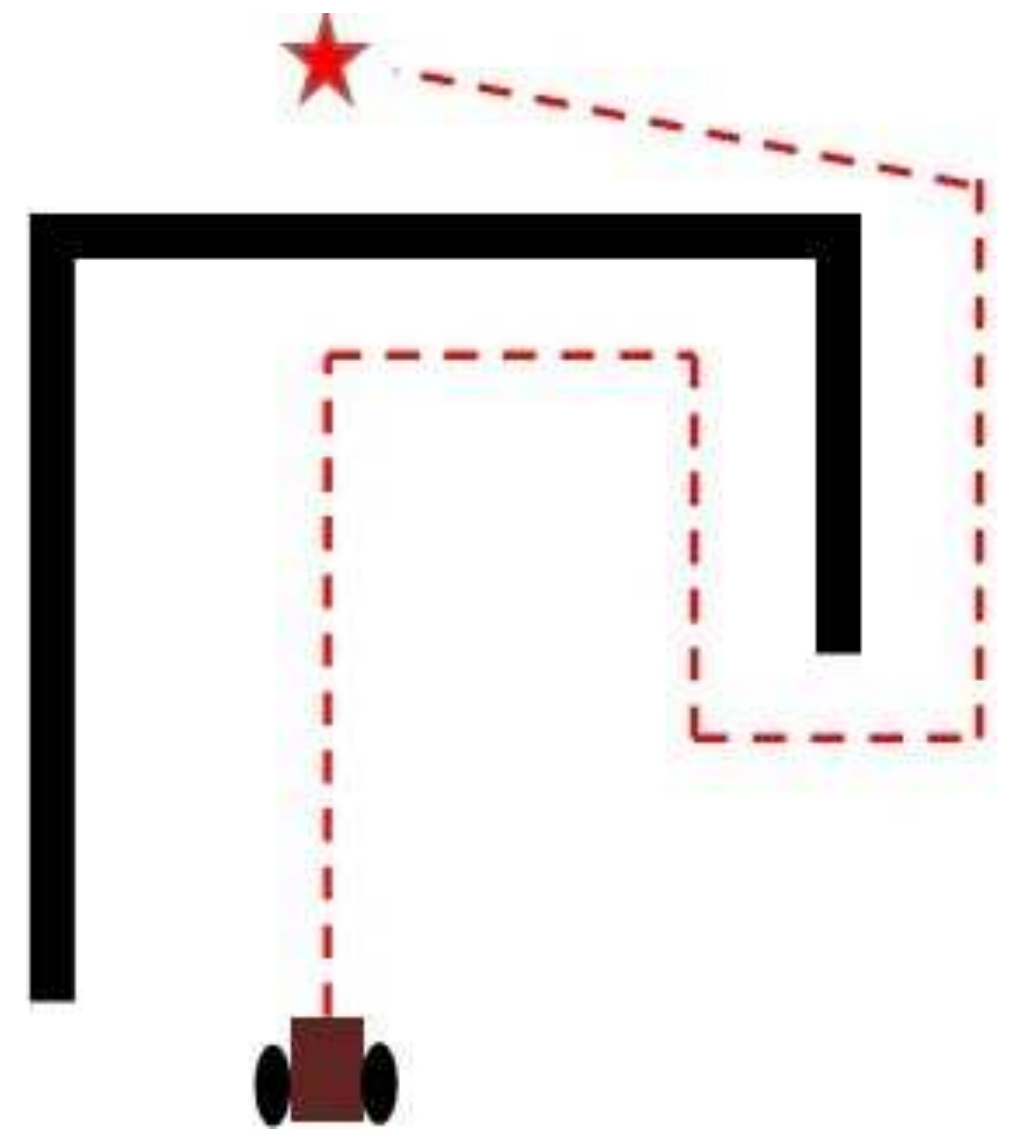}}
\label{c3.exp44}}
\caption{Robot avoids U-shape obstacle}
\label{c3.exp4}
\end{figure}
\par

The first experiment shows the avoidance of a U-shape obstacle using the proposed navigation algorithm. the robot move straight towards the target location, which is behind the obstacle, until it senses the obstacle within the threshold distance. The robot starts to avoid the obstacle by keeping constant distance to it, see Fig.~\ref{c3.exp41} and Fig.~\ref{c3.exp42}. In Fig.~\ref{c3.exp43}, the robot is oriented towards the target and safely move towards it. The complete path is depicted in Fig.~\ref{c3.exp44}.

\begin{figure}[!h]
\centering
\subfigure[]{\scalebox{0.33}{\includegraphics{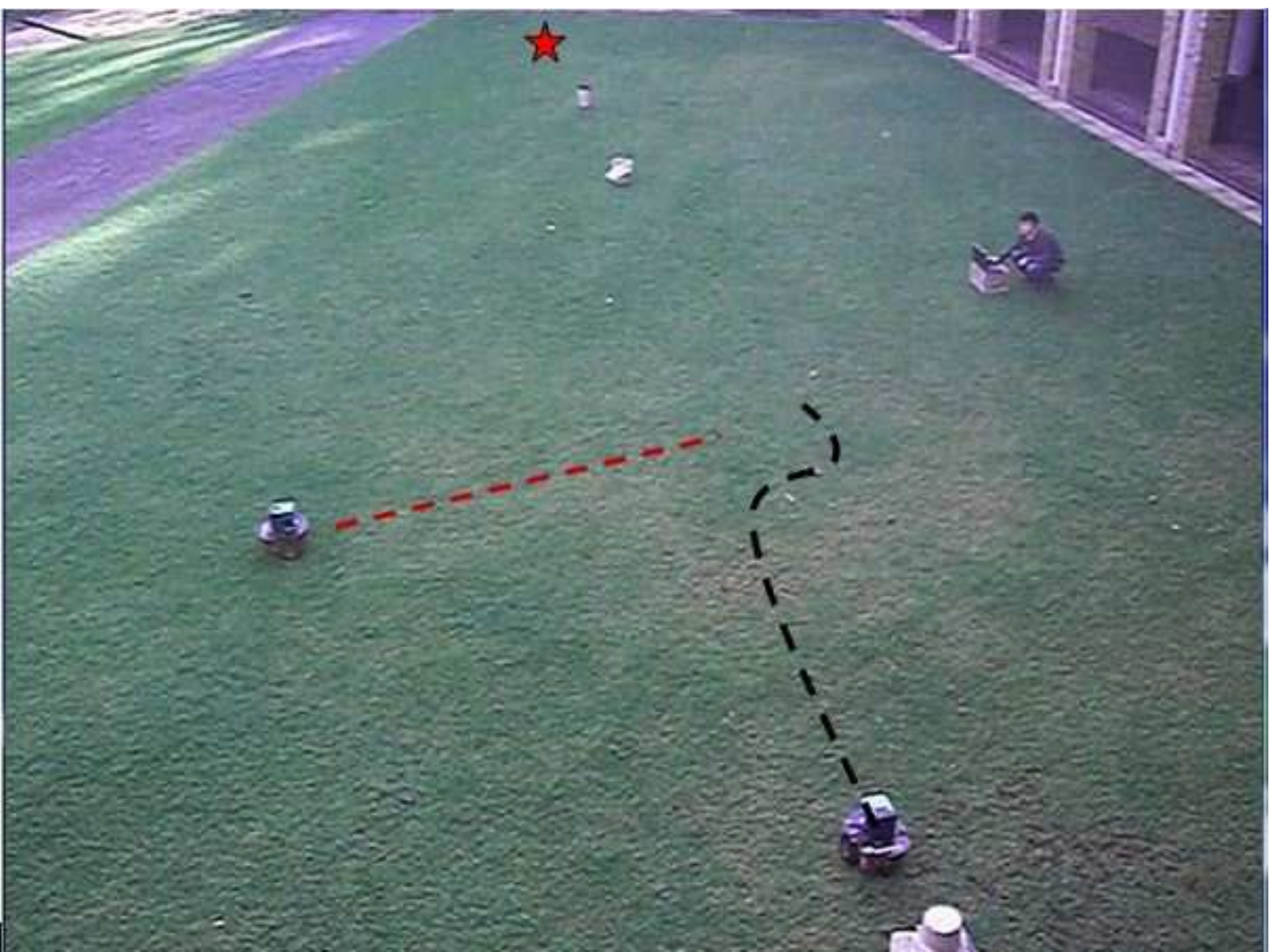}}
\label{c3.exp11}}
\subfigure[]{\scalebox{0.33}{\includegraphics{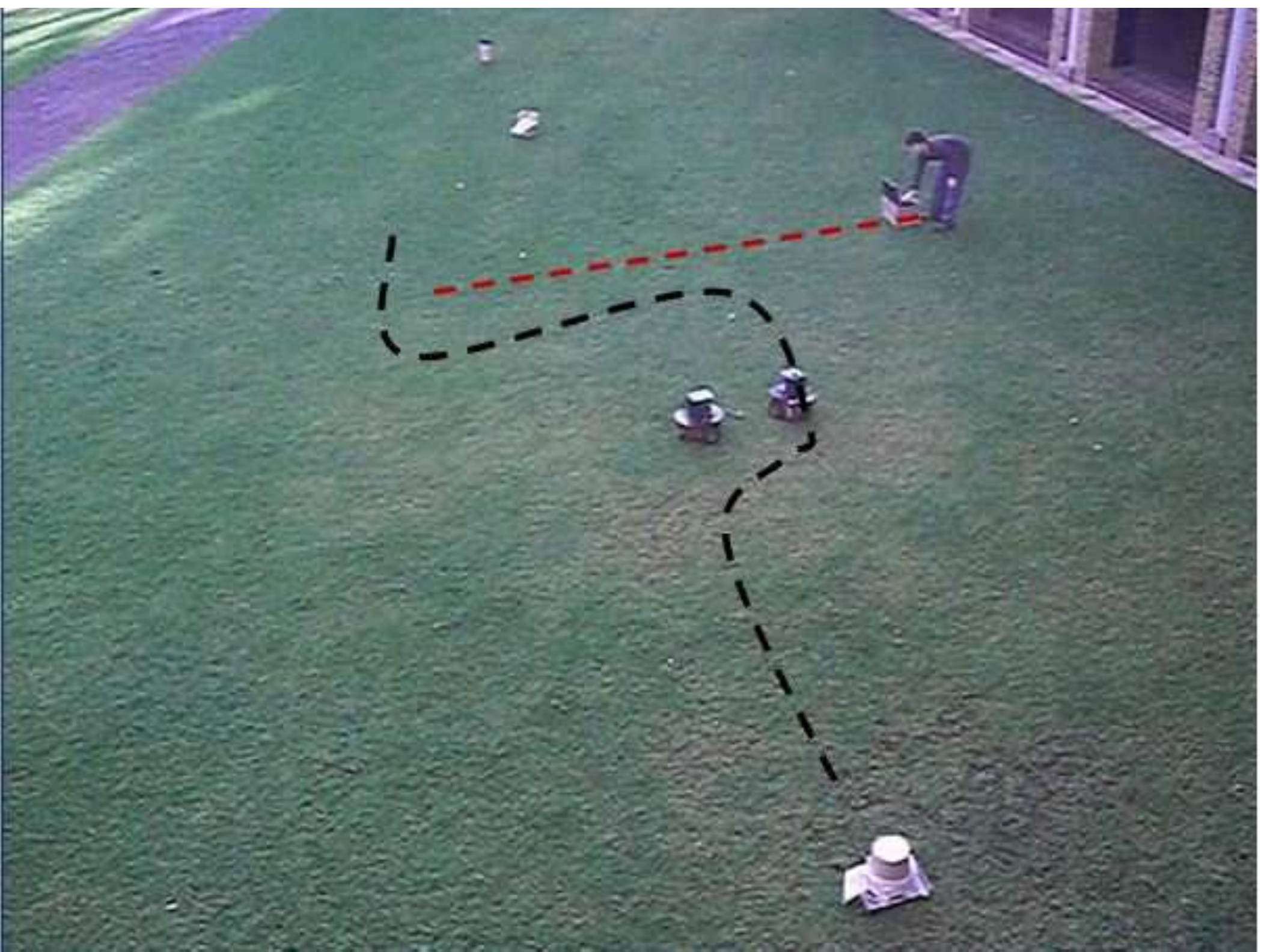}}
\label{c3.exp12}}
\subfigure[]{\scalebox{0.33}{\includegraphics{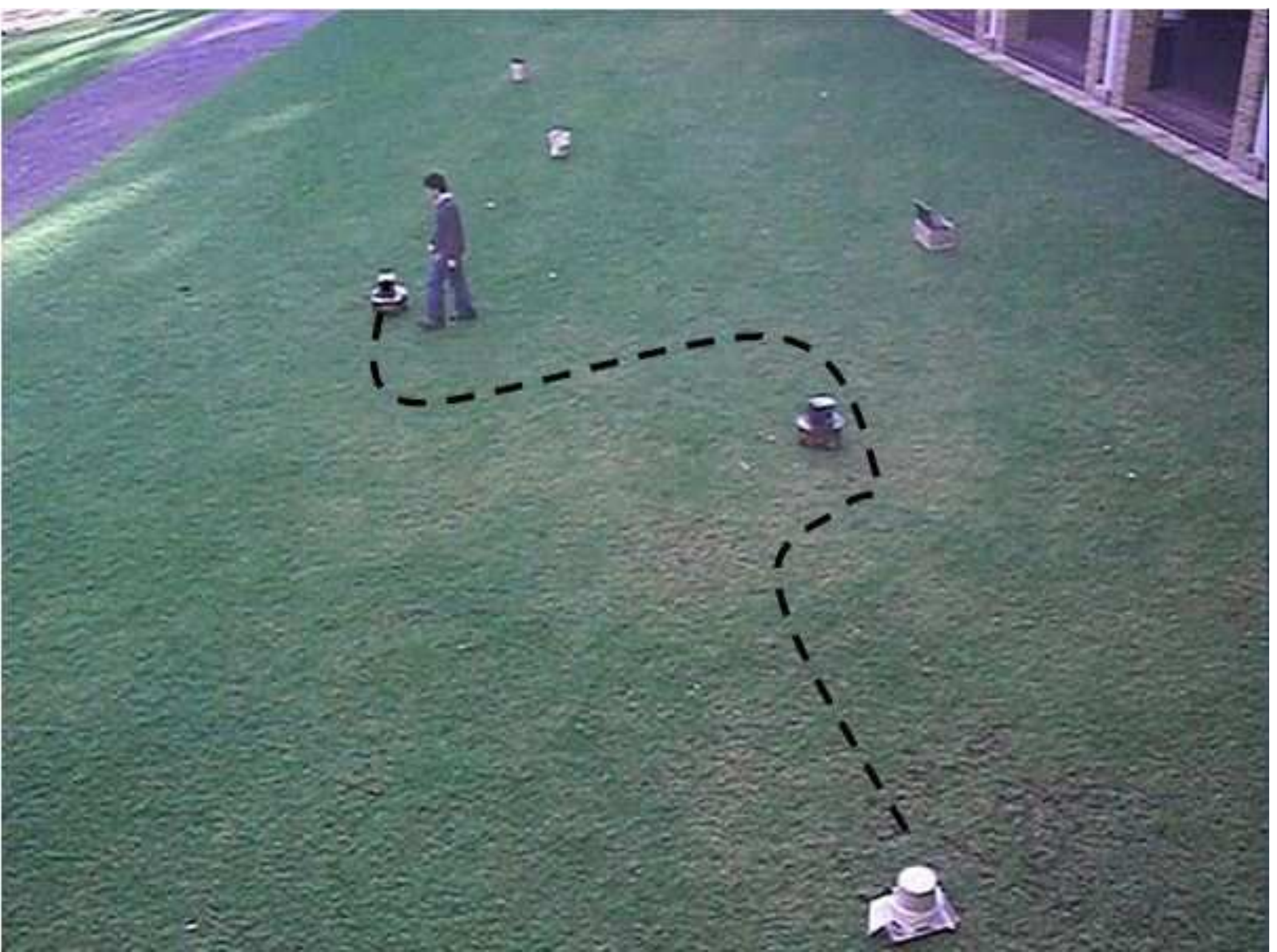}}
\label{c3.exp13}}
\subfigure[]{\scalebox{0.33}{\includegraphics{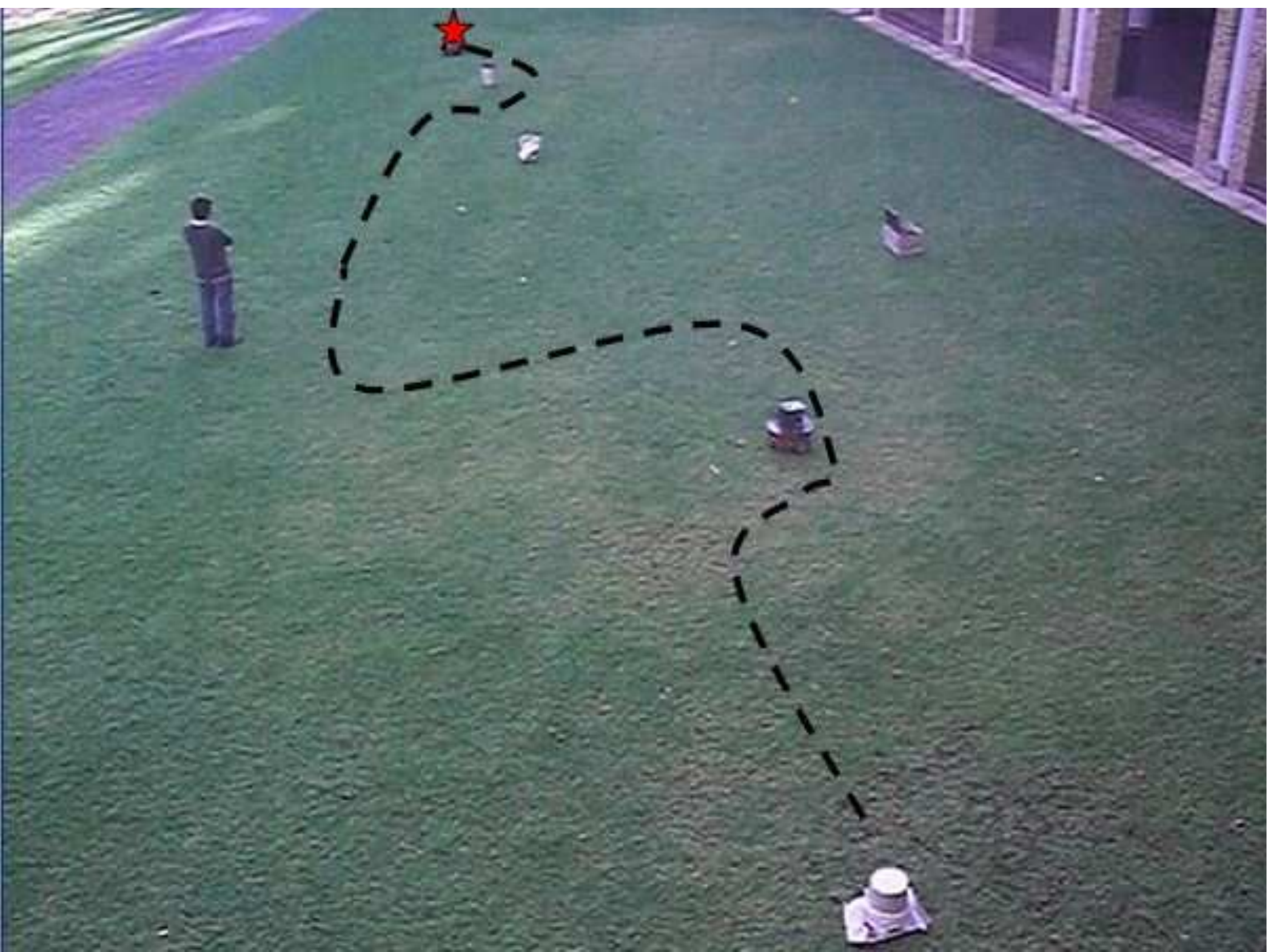}}
\label{c3.exp14}}
\caption{Robot navigating in an environment with stationary and dynamic obstacles}
\label{c3.exp1}
\end{figure}
\par

The next experiment demonstrate the basic ability of the proposed navigation algorithm to guide the mobile robot in an environment with stationary and dynamic obstacles. At the beginning of the experiment, the mobile robot moves towards the target because there is no threat of collision. When there is an obstacle detected within the threshold distance $C$, the mobile robot avoid colliding into the obstacle by keeping a constant distance between itself and the obstacle, i.e. tracking the $d_0$-equidistant curve, the mobile robot continues to pursuit the target when $d \le d_0+ \epsilon $ and the robot is oriented towards the target. see Fig.~\ref{c3.exp11}, Fig.~\ref{c3.exp12} and Fig.~\ref{c3.exp13}. The complete path is shown in Fig.~\ref{c3.exp14}.
\par

\begin{figure}[!h]
\centering
\subfigure[]{\scalebox{0.45}{\includegraphics{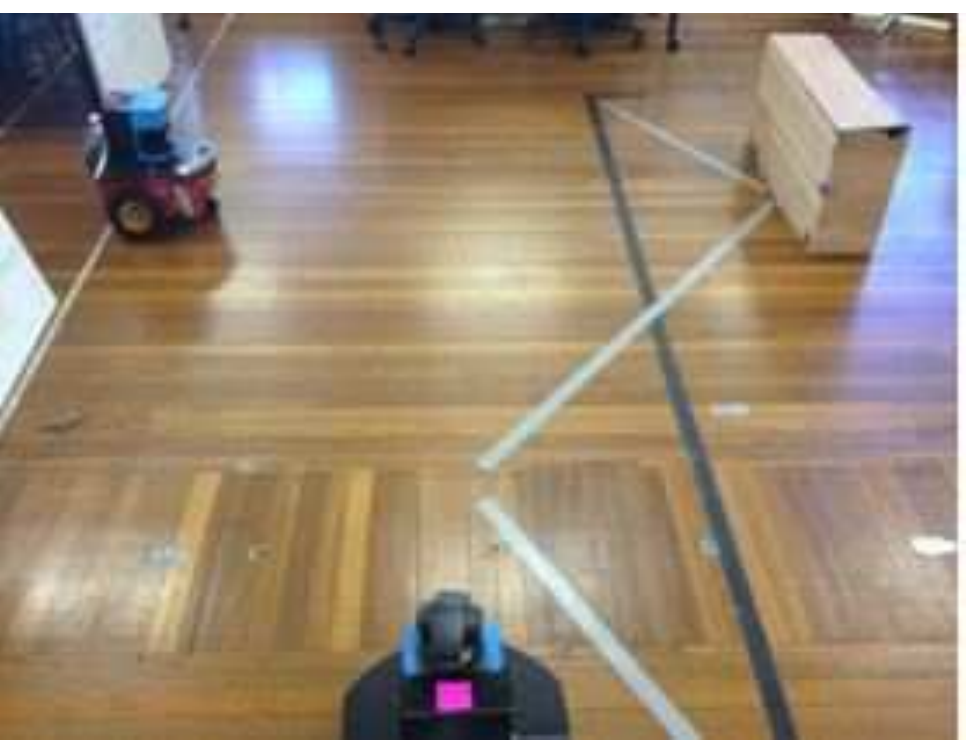}}
\label{c3.exp24}}
\subfigure[]{\scalebox{0.45}{\includegraphics{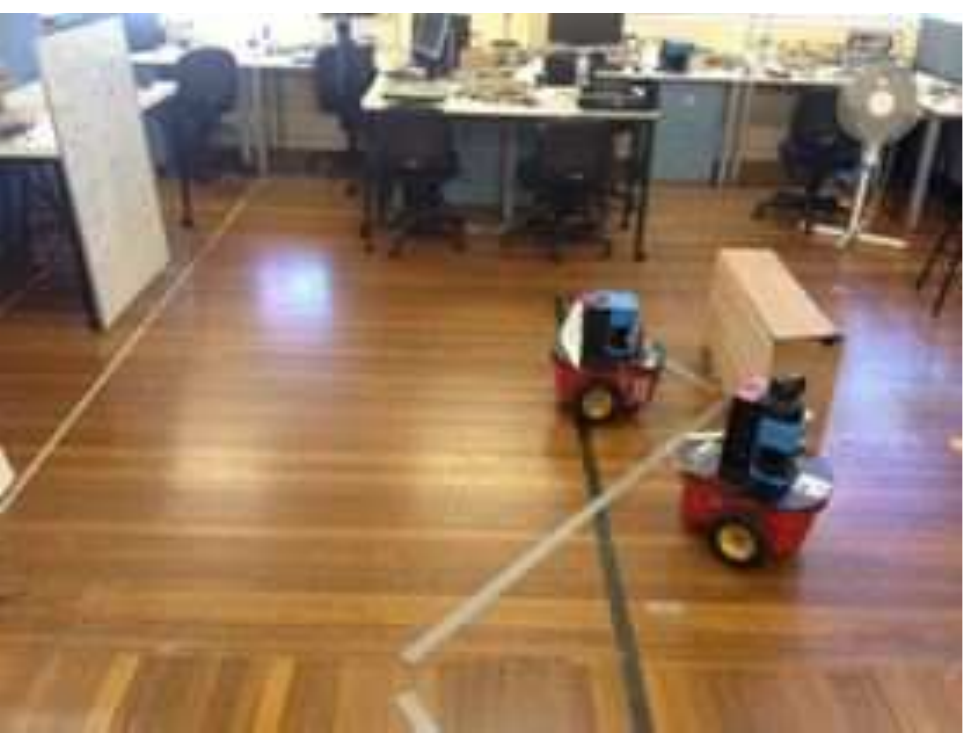}}
\label{c3.exp25}}
\subfigure[]{\scalebox{0.45}{\includegraphics{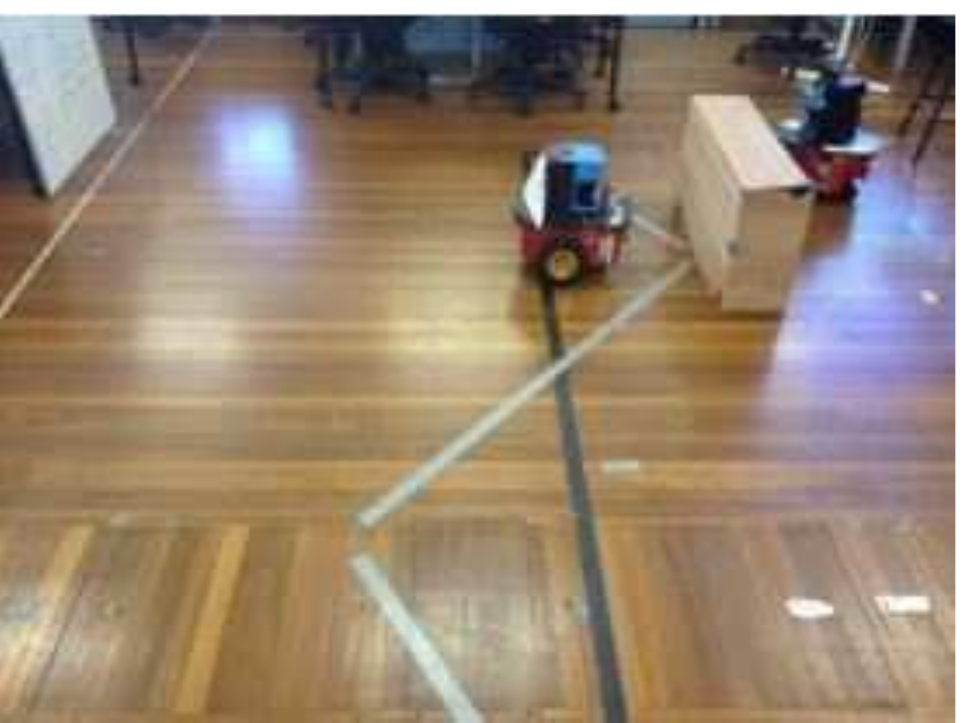}}
\label{c3.exp26}}
\subfigure[]{\scalebox{0.45}{\includegraphics{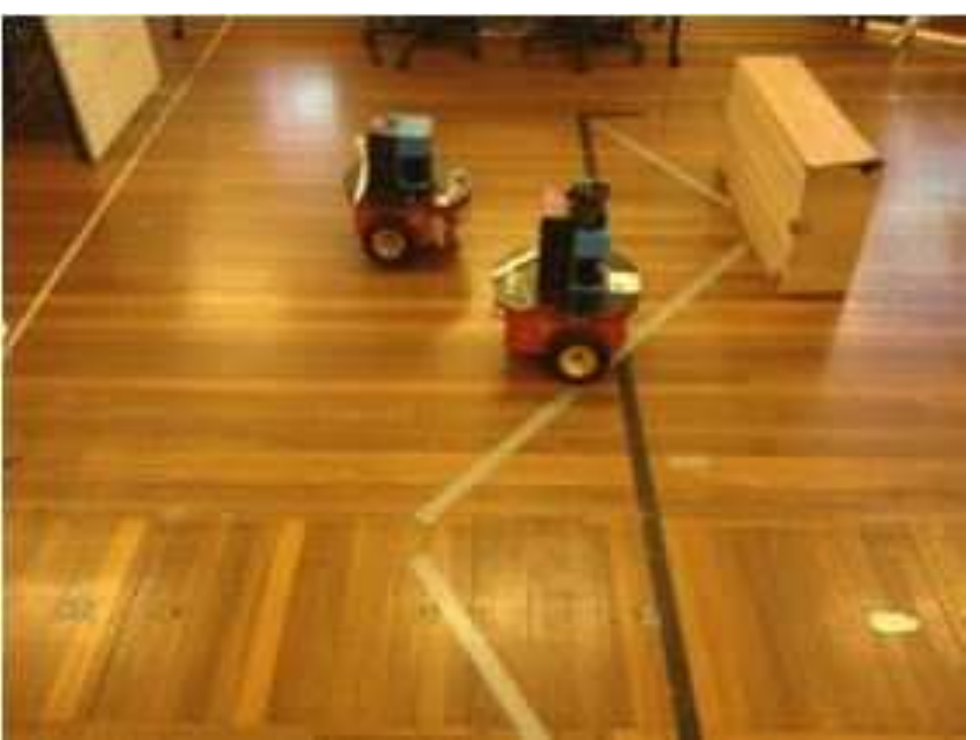}}
\label{c3.exp21}}
\subfigure[]{\scalebox{0.45}{\includegraphics{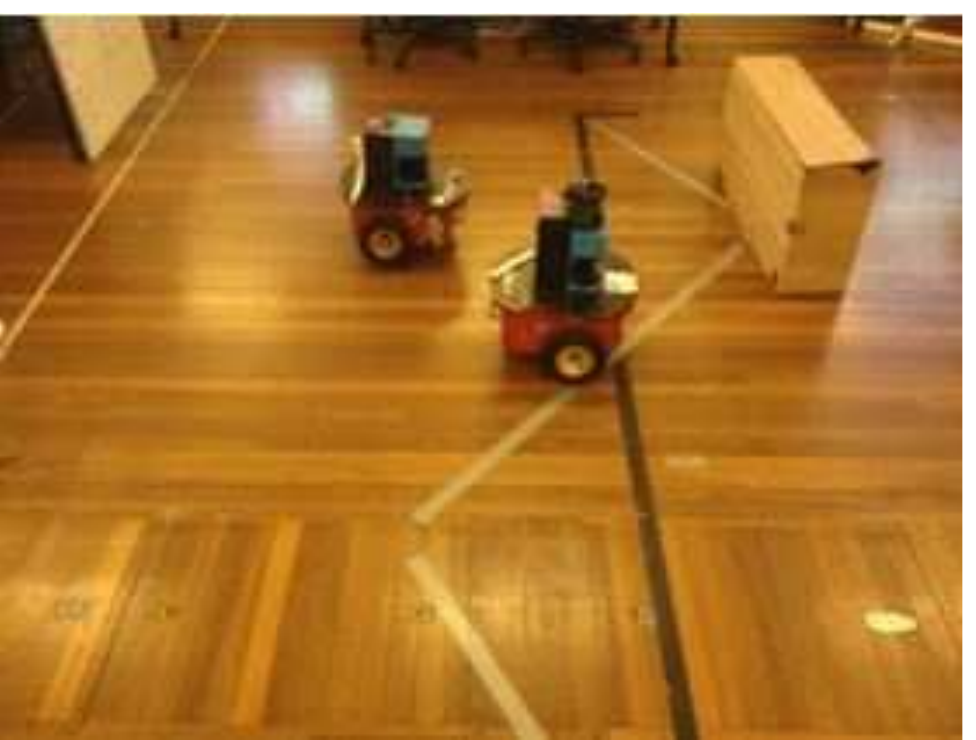}}
\label{c3.exp22}}
\subfigure[]{\scalebox{0.45}{\includegraphics{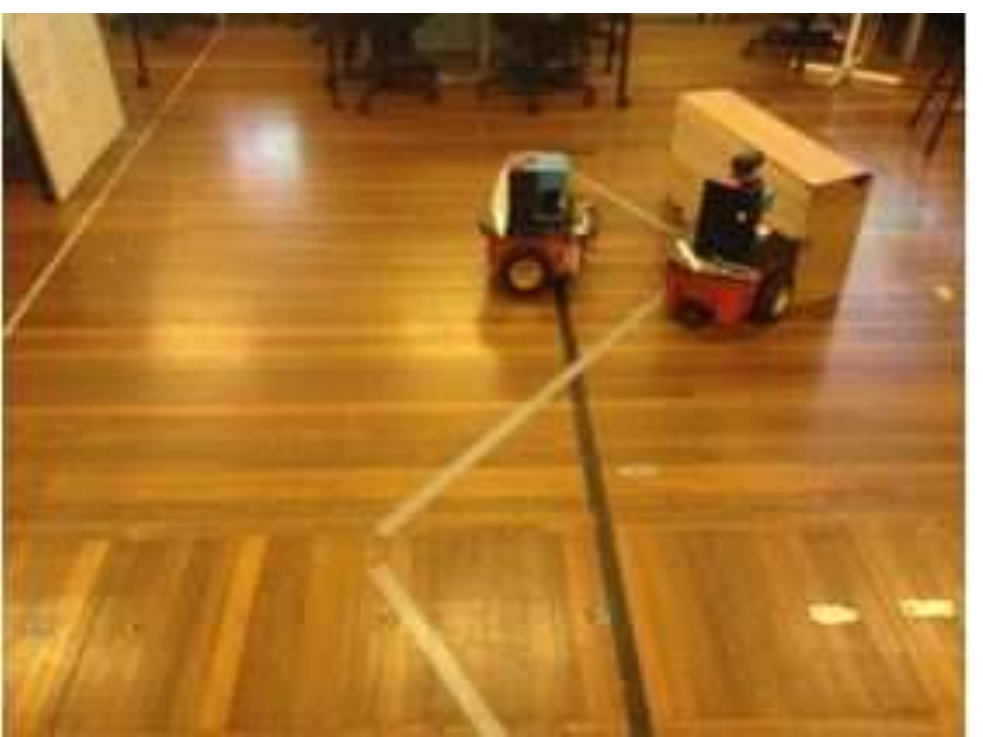}}
\label{c3.exp23}}
\subfigure[]{\scalebox{0.55}{\includegraphics{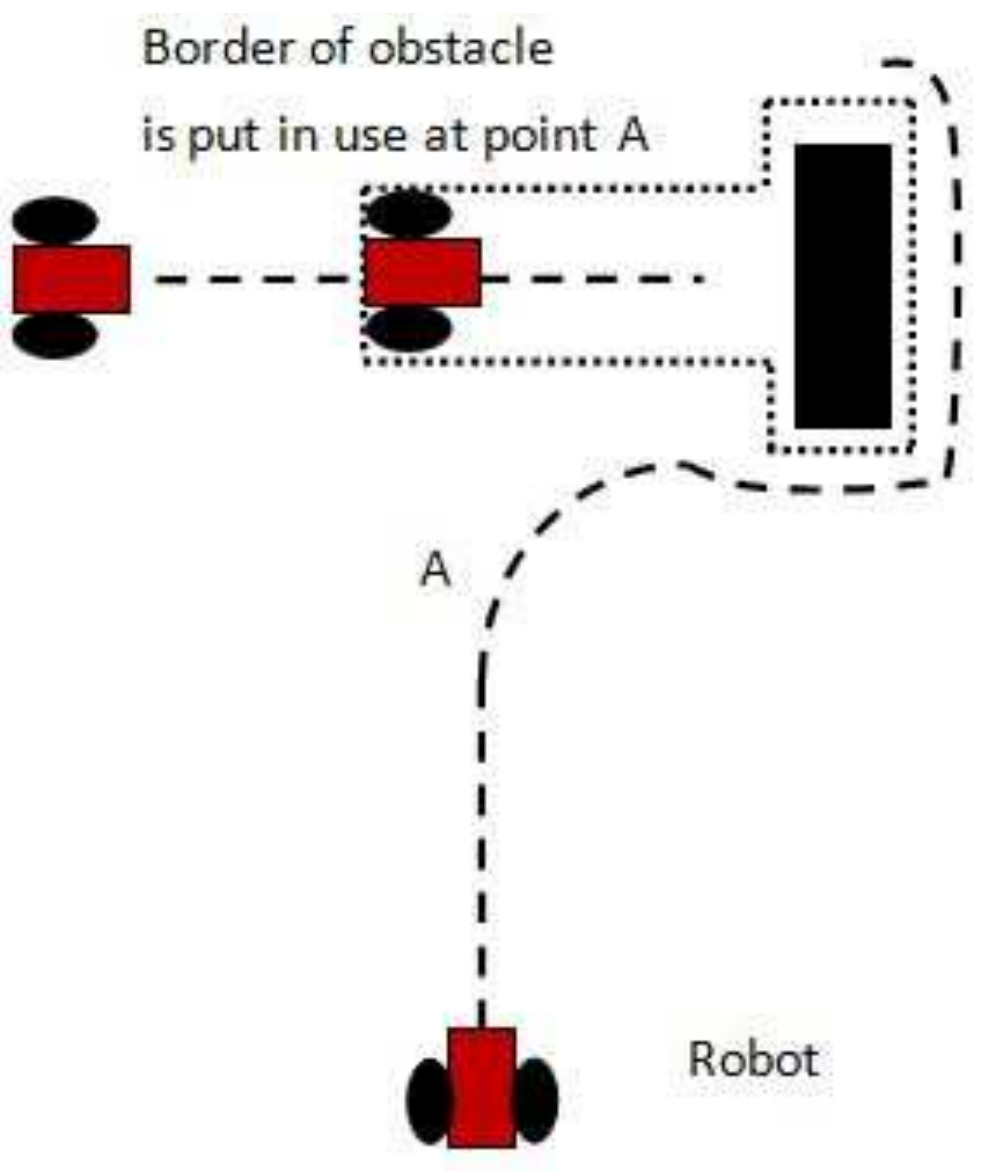}}
\label{c3.exp28}}
\subfigure[]{\scalebox{0.5}{\includegraphics{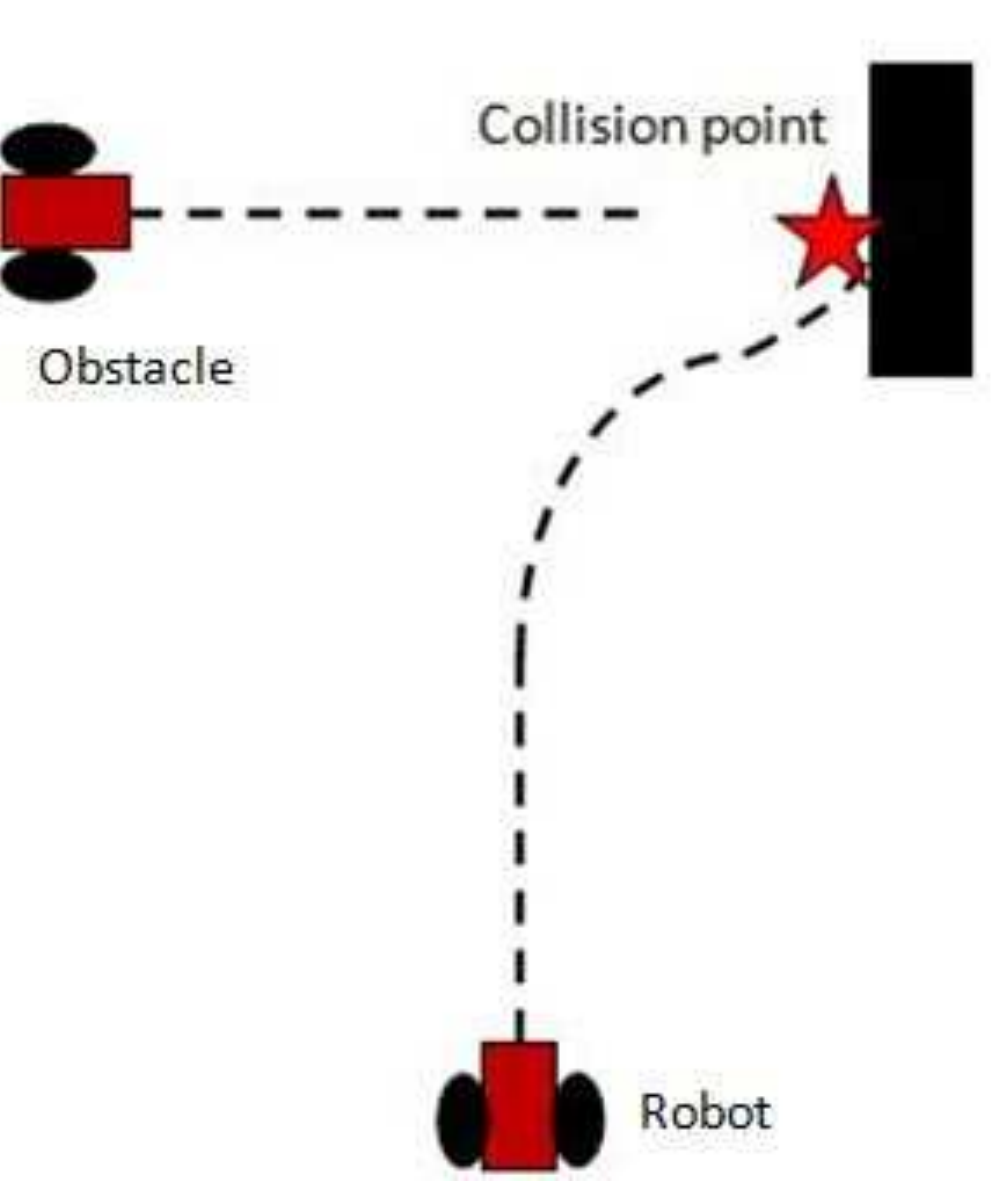}}
\label{c3.exp27}}
\caption{The implementation of the proposed navigation algorithm with the interpolation technique}
\label{c3.exp2}
\end{figure}
\par

In Fig.~\ref{c3.exp2}, the implementation of the proposed navigation algorithm with the interpolation technique (as discussed in Fig.~\ref{c3.sim3}) is shown. In this scenario, the robot successfully avoid the moving obstacle and the stationary obstacle by considering them as one obstacle using interpolation technique, see Fig.~\ref{c3.exp24}, Fig.~\ref{c3.exp25} and Fig.~\ref{c3.exp26} . Note that the interpolation technique is put in use when the obstacles are closely positioned, for example, as shown in Fig.~\ref{c3.exp28}, the interpolation is put in use at point A rather than at the beginning of the experiment, which otherwise would create a very large obstacle with spacious maneuverable gap in between them.  However, the robot collides into the box while avoiding the moving obstacle if the interpolation technique is not put in use, as seen in Fig.~\ref{c3.exp21}, Fig.~\ref{c3.exp22} and Fig.~\ref{c3.exp23}. The comparison of the overall paths taken  by the robot for these two cases are present in Fig.~\ref{c3.exp28} and Fig.~\ref{c3.exp27}.

\begin{figure}[!h]
\centering
\subfigure[]{\scalebox{0.33}{\includegraphics{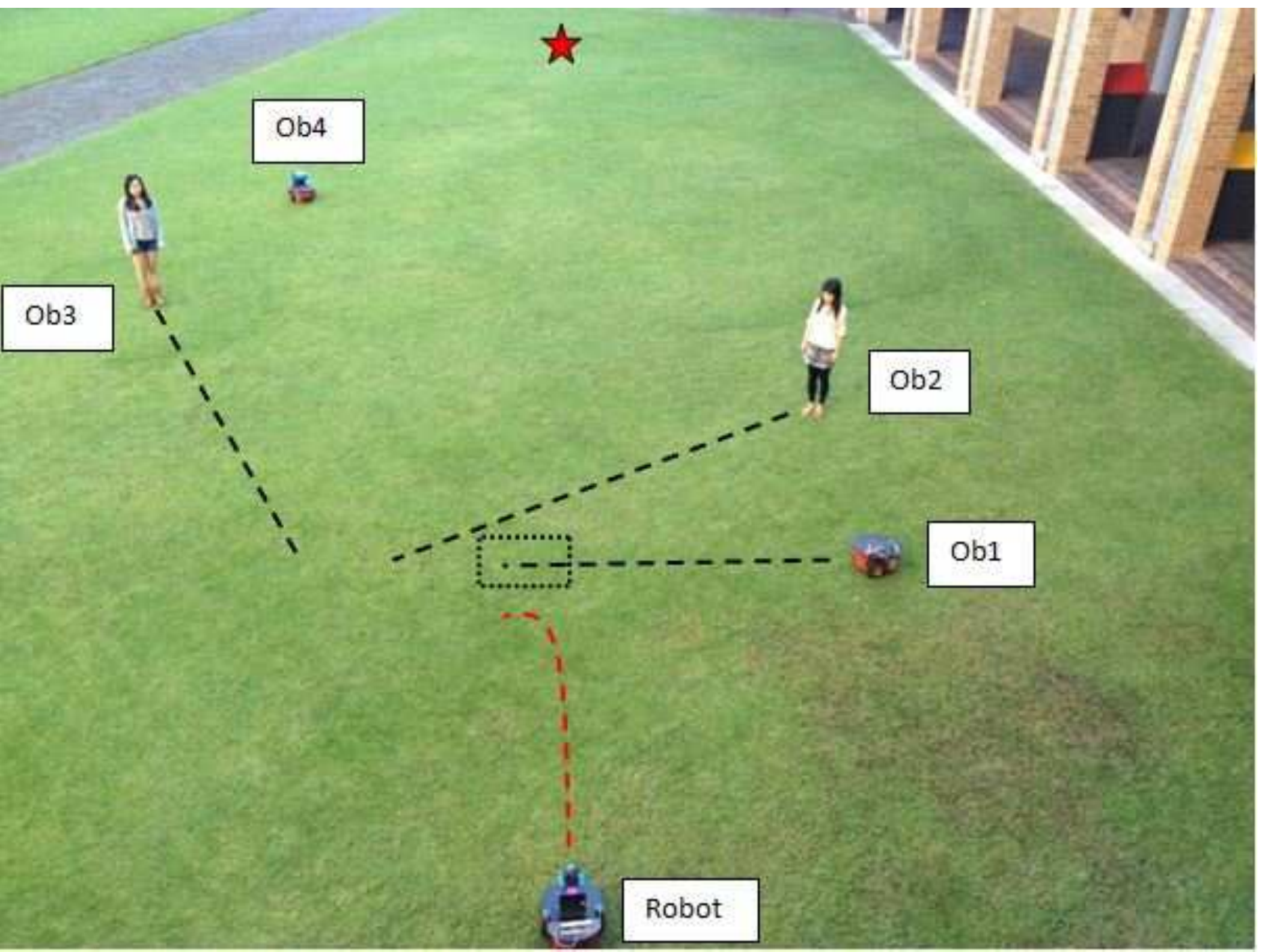}}
\label{c3.exp31}}
\subfigure[]{\scalebox{0.33}{\includegraphics{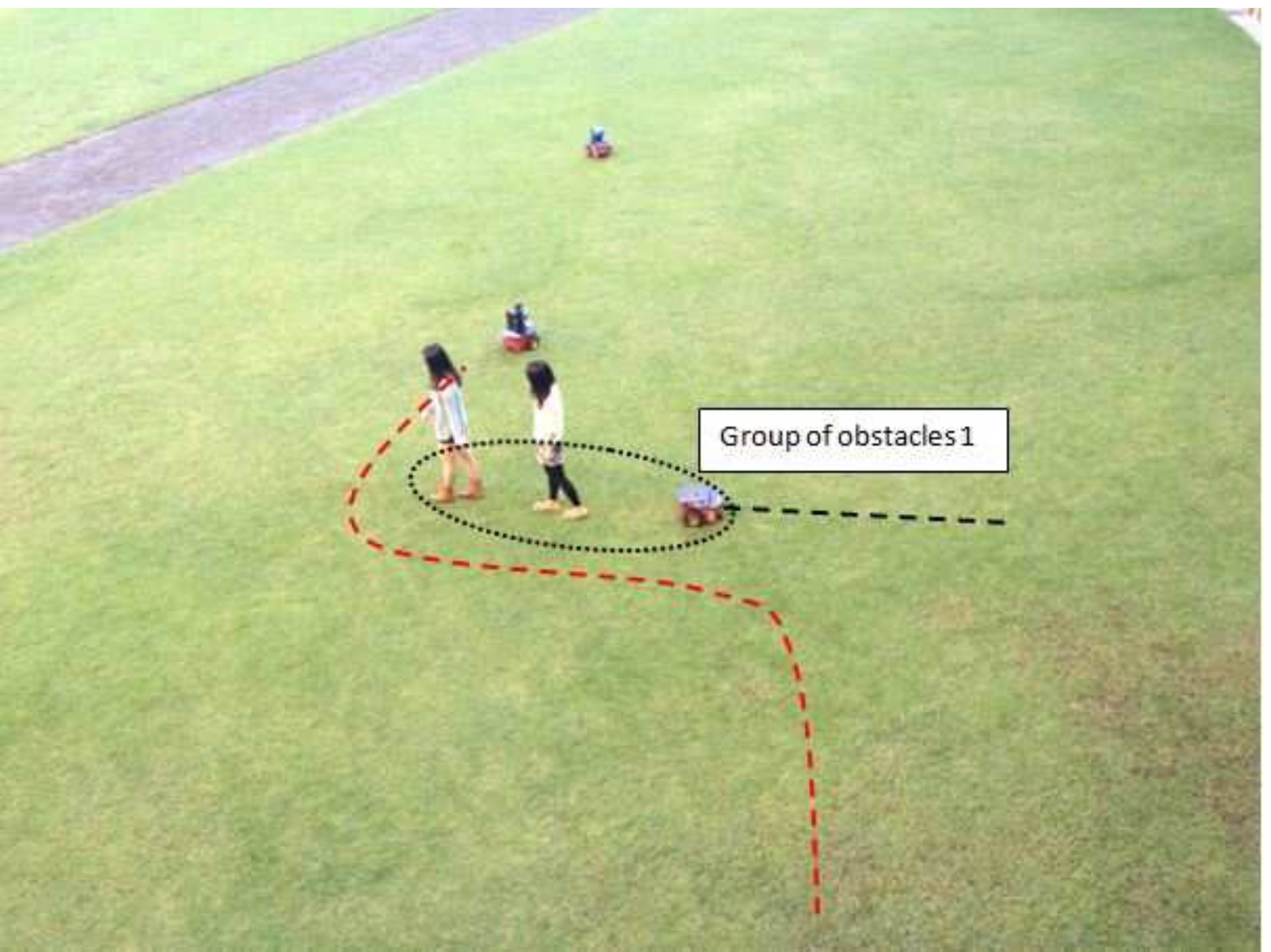}}
\label{c3.exp32}}
\subfigure[]{\scalebox{0.33}{\includegraphics{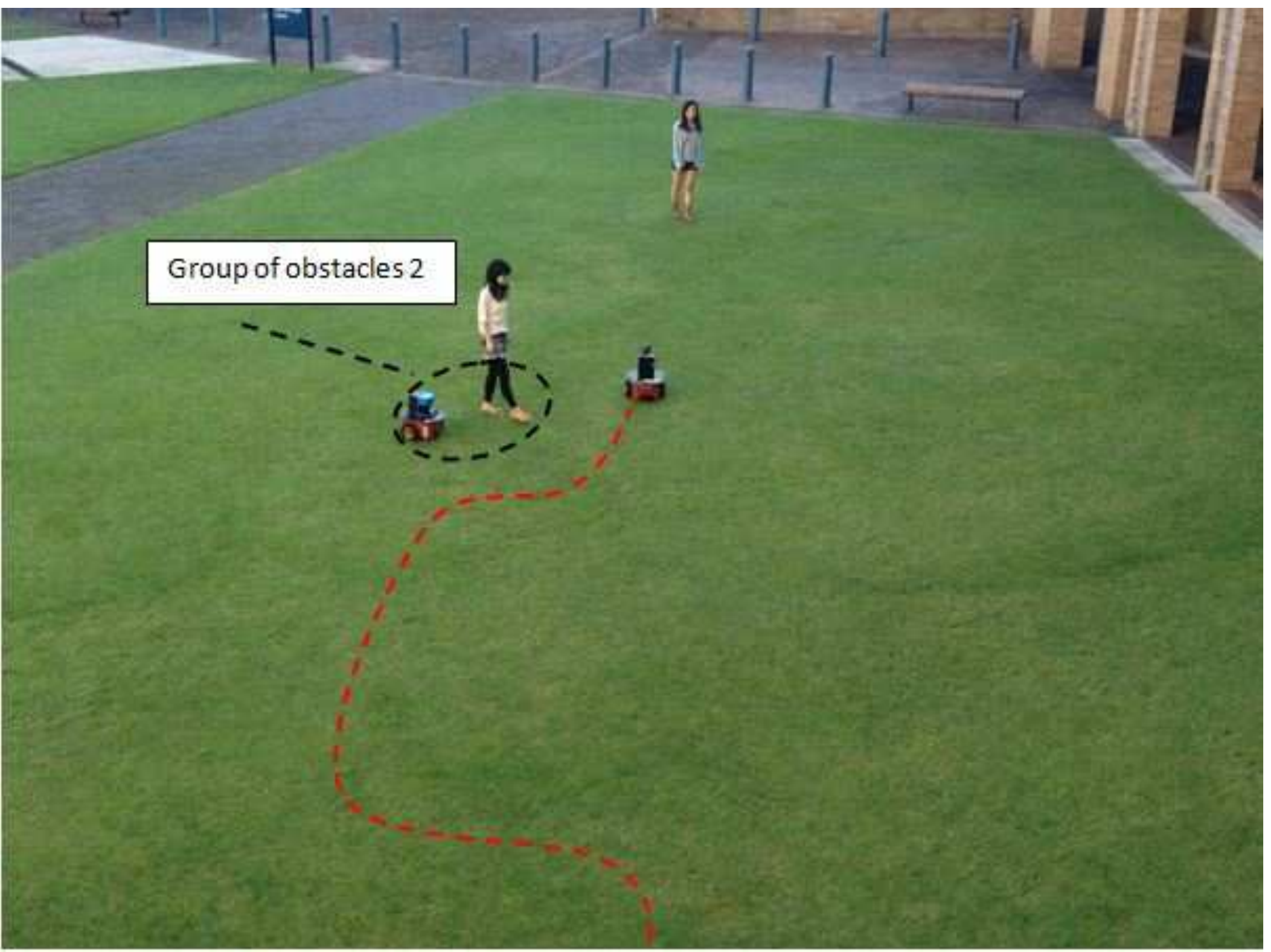}}
\label{c3.exp33}}
\subfigure[]{\scalebox{0.33}{\includegraphics{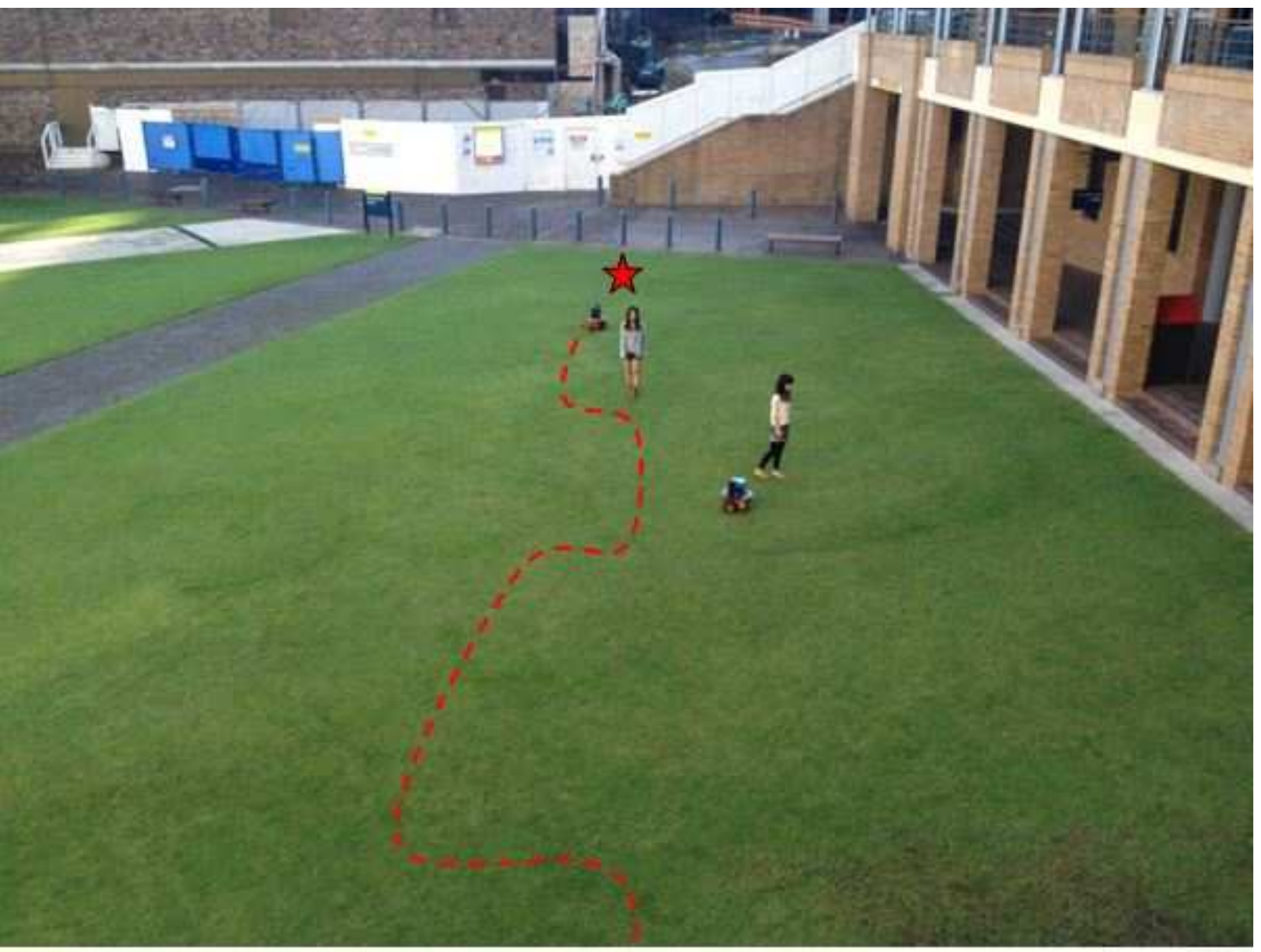}}
\label{c3.exp34}}
\caption{Robot avoids tight clusters of obstacles with dynamical changing radii}
\label{c3.exp3}
\end{figure}
\par

The proposed navigation algorithm with interpolation technique can be used to deal with more complicated real life scenario such as obstacles with dynamical changing radii and tight clusters of obstacles. Fig.~\ref{c3.exp3} shows an example of such scenario. In Fig.~\ref{c3.exp31}, the robot starts to bypass obstacle 1 (ob1), the obstacles ob2 and ob3 approach ob1, see Fig.~\ref{c3.exp31}, thus forming a tight cluster (group of obstacles) which move in a common direction of ob1. The imaginary border of this group of obstacles is indicated by the dashed line. The robot is able to keep a constant distance to the imaginary border of the group and avoid the group as a whole, see Fig.~\ref{c3.exp32}. After this, the robot encounters the second group of obstacle 2 and one single obstacle with different radii, the robot is able to avoid both of them and safely arrive the target location, see Fig.~\ref{c3.exp33} and Fig.~\ref{c3.exp34}. 
\par 

\begin{figure}[!h]
\centering
\subfigure[]{\scalebox{0.33}{\includegraphics{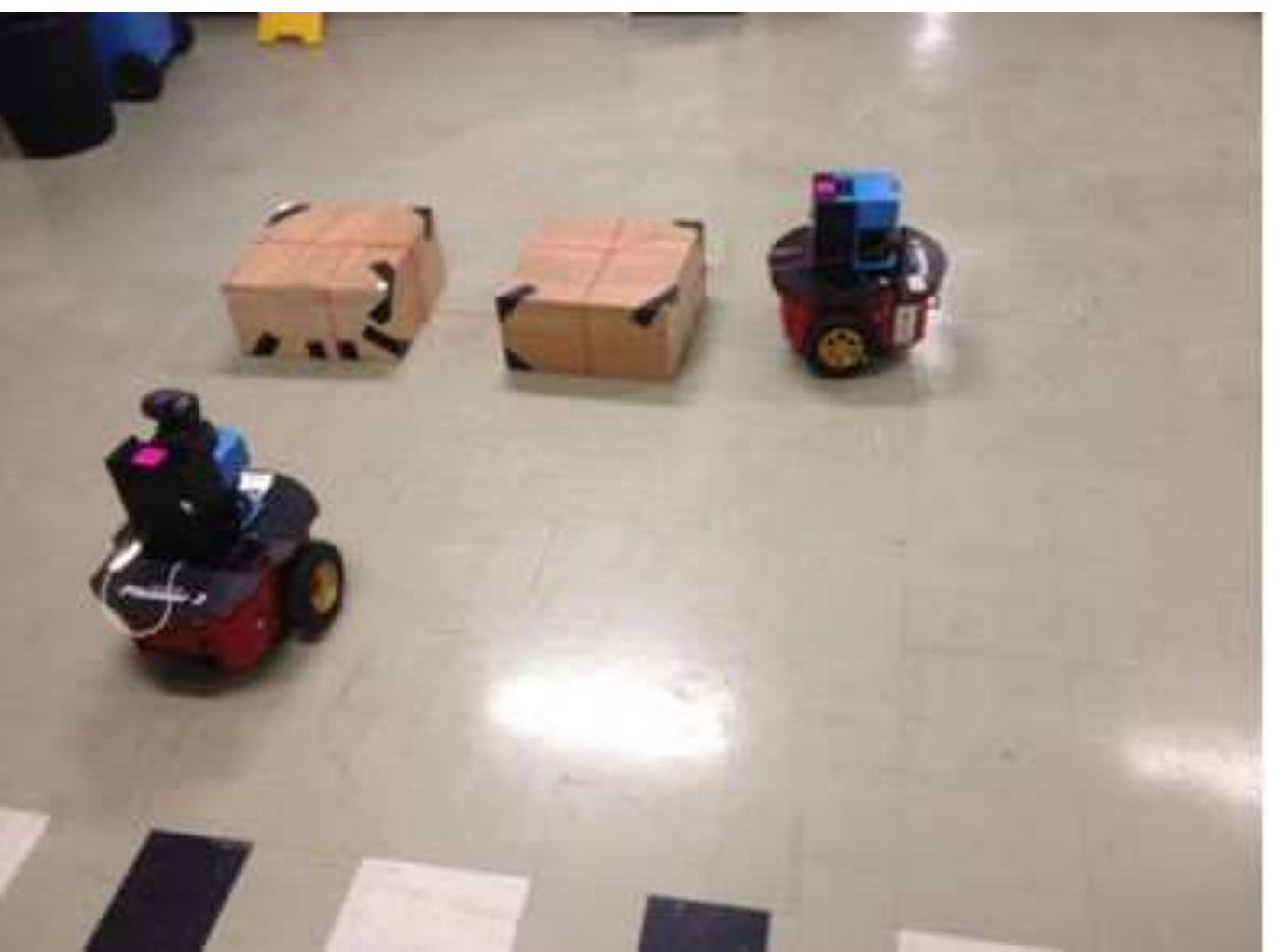}}
\label{c3.exp51}}
\subfigure[]{\scalebox{0.33}{\includegraphics{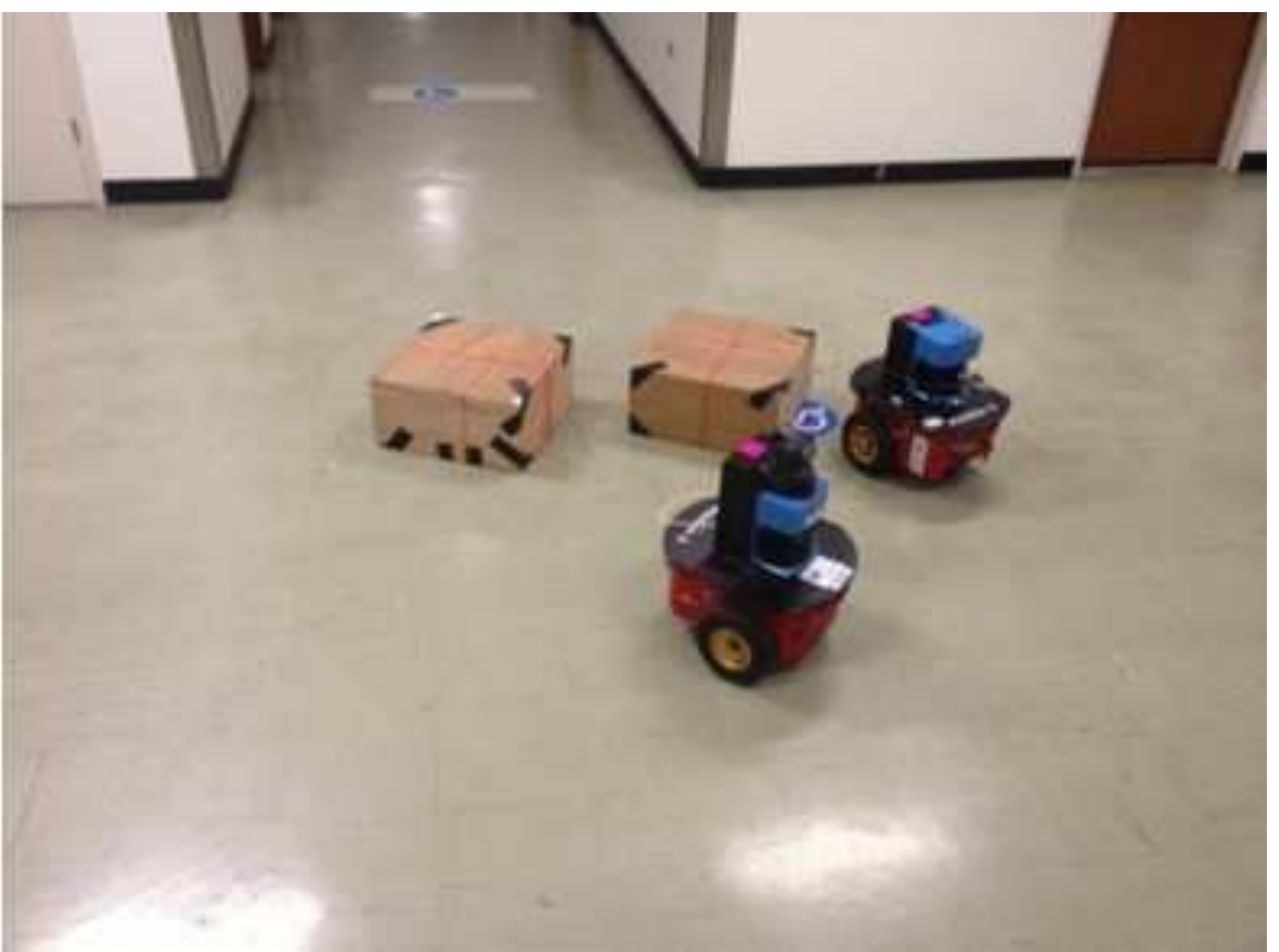}}
\label{c3.exp52}}
\subfigure[]{\scalebox{0.33}{\includegraphics{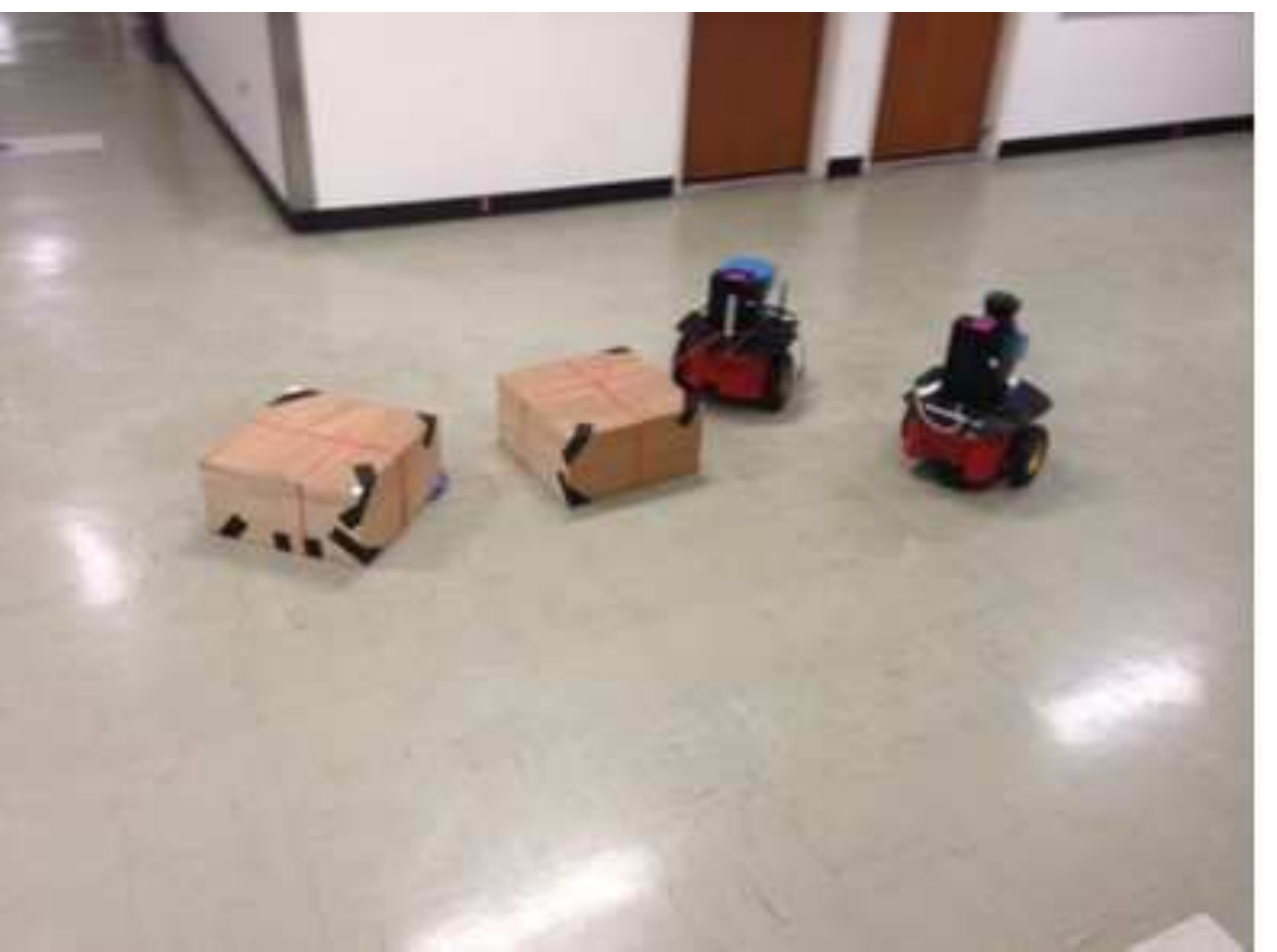}}
\label{c3.exp53}}
\subfigure[]{\scalebox{0.33}{\includegraphics{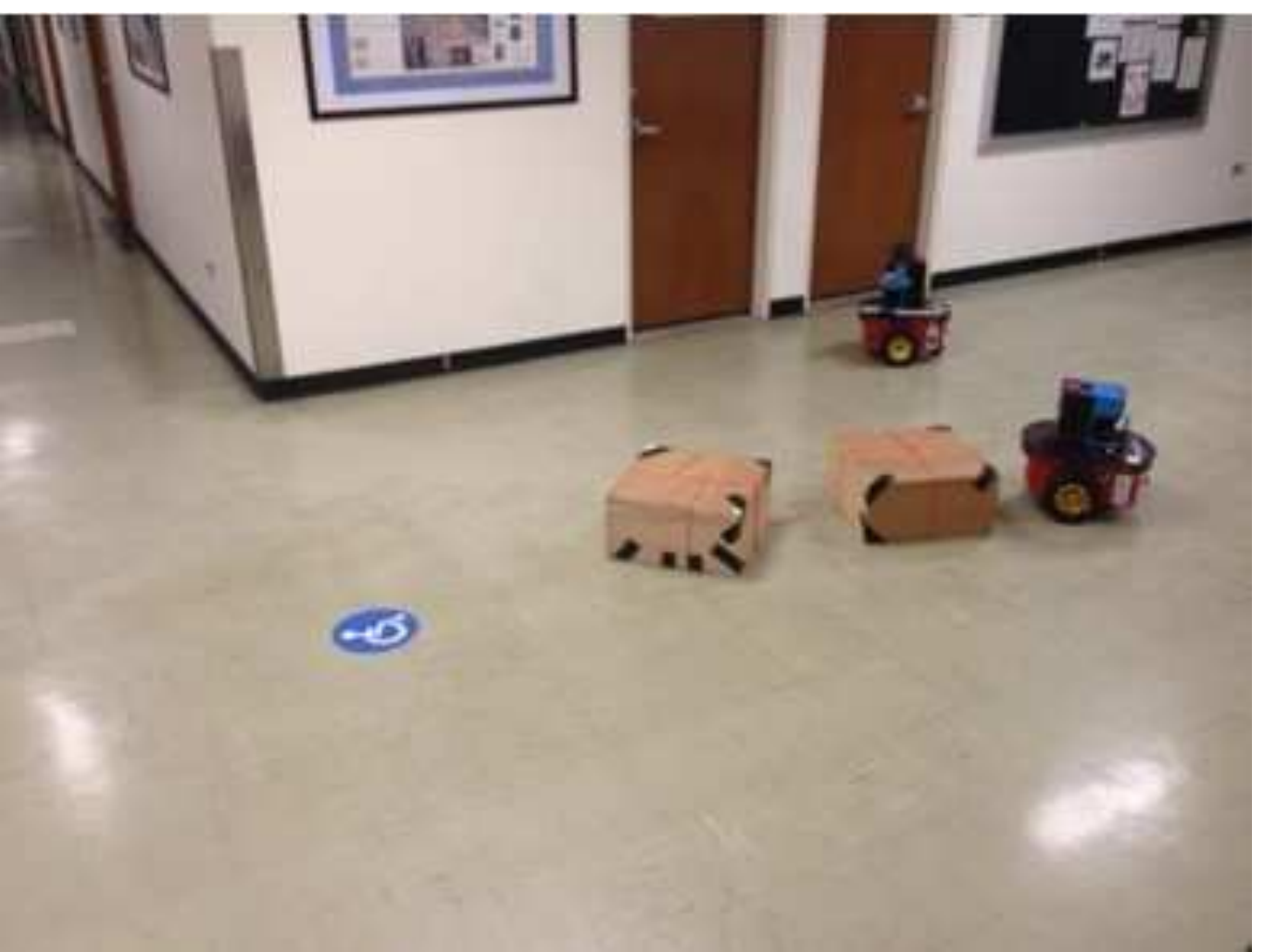}}
\label{c3.exp54}}
\subfigure[]{\scalebox{0.33}{\includegraphics{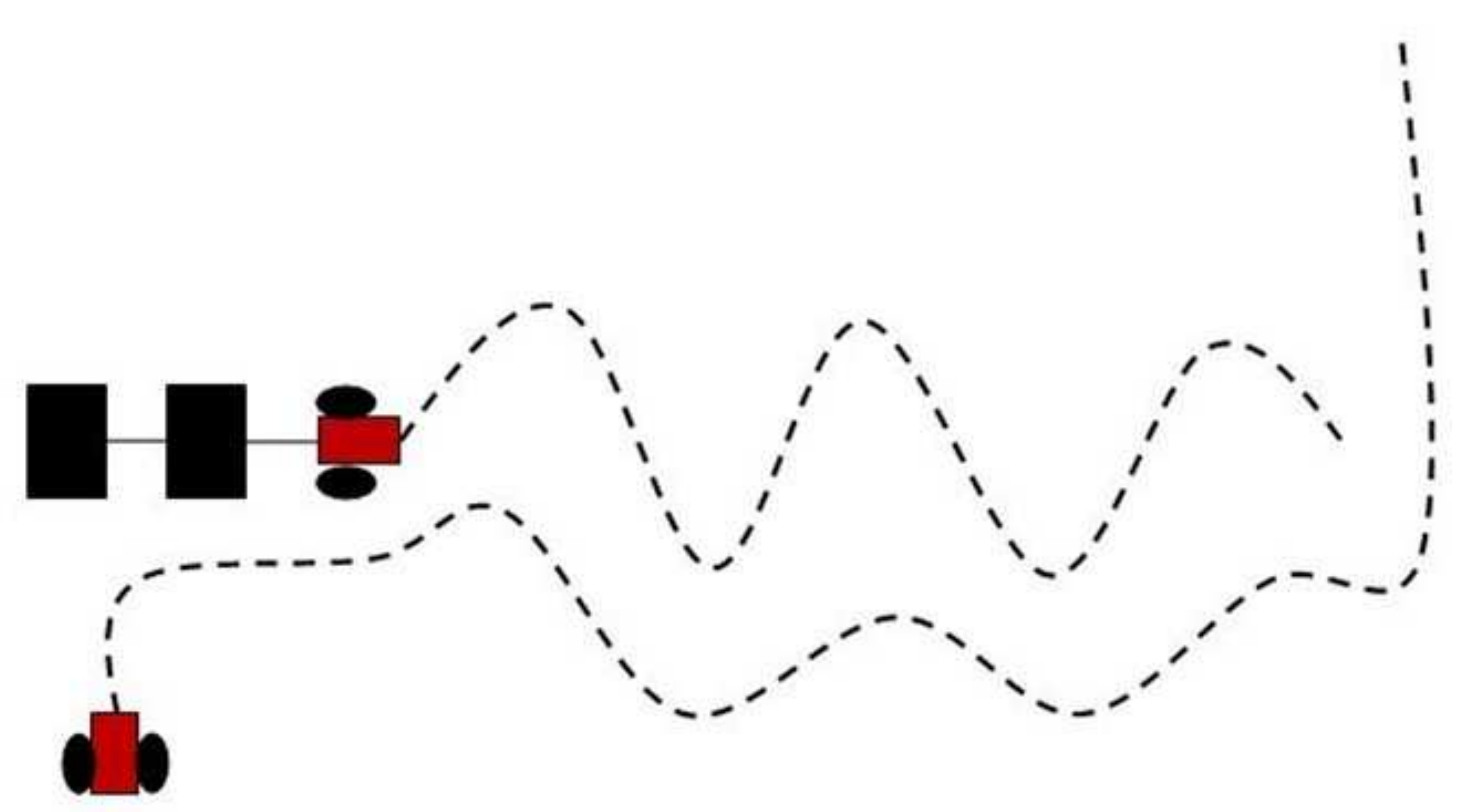}}
\label{c3.exp55}}
\caption{Robot avoids deforming obstacle}
\label{c3.exp5}
\end{figure}
\par

In the last experiment, the robot faces an obstacle with deforming shape. This obstacle consists of a P3 robot towing two boxes forming a chain of interconnected obstacles. The inter-element free spaces in the chain are interpolated so that the robot has to avoid the resultant snake-like obstacle. The towing robot moves in a sinusoidal fashion so that the obstacle "snake" squirms. The challenge is that the robot needs to keep a constant distance to the border of the obstacle which is time-varying, see Fig.~\ref{c3.exp51}, Fig.~\ref{c3.exp52}, Fig.~\ref{c3.exp53} and Fig.~\ref{c3.exp54}. The complete path for the robot to avoid this obstacle with deforming shape is depicted in Fig.~\ref{c3.exp55} 

\section{Summary}

A sliding mode based navigation algorithm for non-holonomic mobile robots has been proposed. The proposed navigation algorithm is safe and implement cost-efficient, which guarantees the safety of the mobile robots in uncertain dynamic environments. The efficiency of the proposed algorithm are illustrated by extensive computer simulations. The applicability of the proposed algorithm in many challenging real life scenario are demonstrated by experiments with P3 mobile robot.
\newtheorem{remark}{Remark}[section]
\chapter {Navigation Algorithm Based on an Integrated Environment Representation} \label{C4}

	A typical obstacle avoidance strategy, when performing navigation tasks in complex environments, avoids "the closest" or "the most dangerous" obstacle and continues to move in a desired direction until another dangerous obstacle is detected. The strategy of this kind raises several issues, e.g, the concept of "dangerous" is hard to be defined and may vary in different scenarios, difficulties in dealing with obstacles which are hidden or outside the sensing range, and the ambiguous case where two or more dangerous obstacles presented at the same time. In this section, we propose instead a novel reactive navigation approach based
on an integrated environment representation which overcomes the above problems. This approach is partially inspired by the paradigm shift from binary interaction models to an integrated treatment of multiple interactions which are typical for social interactions in human crowds or animal swarms  suggested  in \cite{MHT11} for analysis of pedestrian behaviour in crowds. Our approach is based on an integrated
representation of the information about the environment in which  the combined effect of close multiple stationary and moving obstacles is implicitly included in the representation of a sensing  field of the robot. An advantage of our approach is that we do not have to separate obstacles and approximate their shapes by discs or polygons. Moreover, our algorithm does not require any information on the obstacle's velocities or any other derivatives of measurements. Our approach results in a very efficient and intelligent robotic behaviour. Instead of being repelled by a crowd of obstacles, as it happens for many other navigation algorithms,the robot seeks a path through this crowd.

	\section{Problem Description} \label{PDC4}

		Consider a non-holonomic mobile robot with hard constraints on its angular velocity. It travels in a two-dimensional plane. The vector of the robot's Cartesian coordinate in this plane is presented by $[x(t),y(t)]$, and its heading is given by $\theta(t)$. The movement of this mobile robot is controlled by both its speed $V>0$ and its angular velocity $u(t)$. The mathematical model of the robot is as follows:
		\begin{equation}
		\label{ch4:1}
		\begin{array}{l}	
		\dot{x(t)} = V \cos \theta(t),
		\\
		\dot{y(t)} = V \sin \theta(t),
		\\
		\dot{\theta(t)} = u(t) 
		\end{array}
		\end{equation}
		where 
		\begin{equation}
		\label{ch4:max}
		 u\in [-U_{max},U_{max}].
		\end{equation}  
Here $U_{max}$ is the non-holonomic constraint on the angular velocity which is given. The angle $\theta(t)\in (-\pi,\pi]$ is measured in the counterclockwise direction from the $x-$axis.
\par
		Let $\delta>0$ be the sampling period. The robot updates its angular velocity $u(t)$ at discrete times $0,\delta,2\delta,3\delta, \ldots$.Also, let 
		\begin{eqnarray}
		\label{cR}
		 c_R(t):=[x(t),y(t)]
		\end{eqnarray}
		 be the vector of the robot's coordinates, and 
		\begin{equation}
		\label{vR}
		v_R:(t)=\dot{c}_R(t)
		\end{equation}
		 be the robot's velocity vector.
\par

		We assume that there exists an unknown dynamic environment, which consists of any number of stationary or moving obstacles which can be deformable. In other words, the environment is described by a time-varying planar subset ${\cal E}(t)$ which is not known to the robot.The current distance $d(t)$ from the robot to  the environment  ${\cal E}(t)$ is defined as 
	\[
		 d(t) :=\min_{r\in {\cal E}(t)}\|r-c_R(t)\|.
	\]
		Here  $\|\cdot\|$
		denotes the standard Euclidean vector norm. 
\par
		Let $\theta_0)\in (-\pi,\pi)$ be a given desired direction and $d_{safe}>0$ be a given distance.The objective of our proposed navigation algorithm is to navigate the robot in the direction $\theta_0$ while satisfying the safety constraints $d(t)\geq d_{safe}$ for all $t$, i.e. $d(t) \ge d_{safe}$ for all $t \ge 0$
\par

\begin{figure}[h]
		\centering
		\includegraphics[width=5.5in]{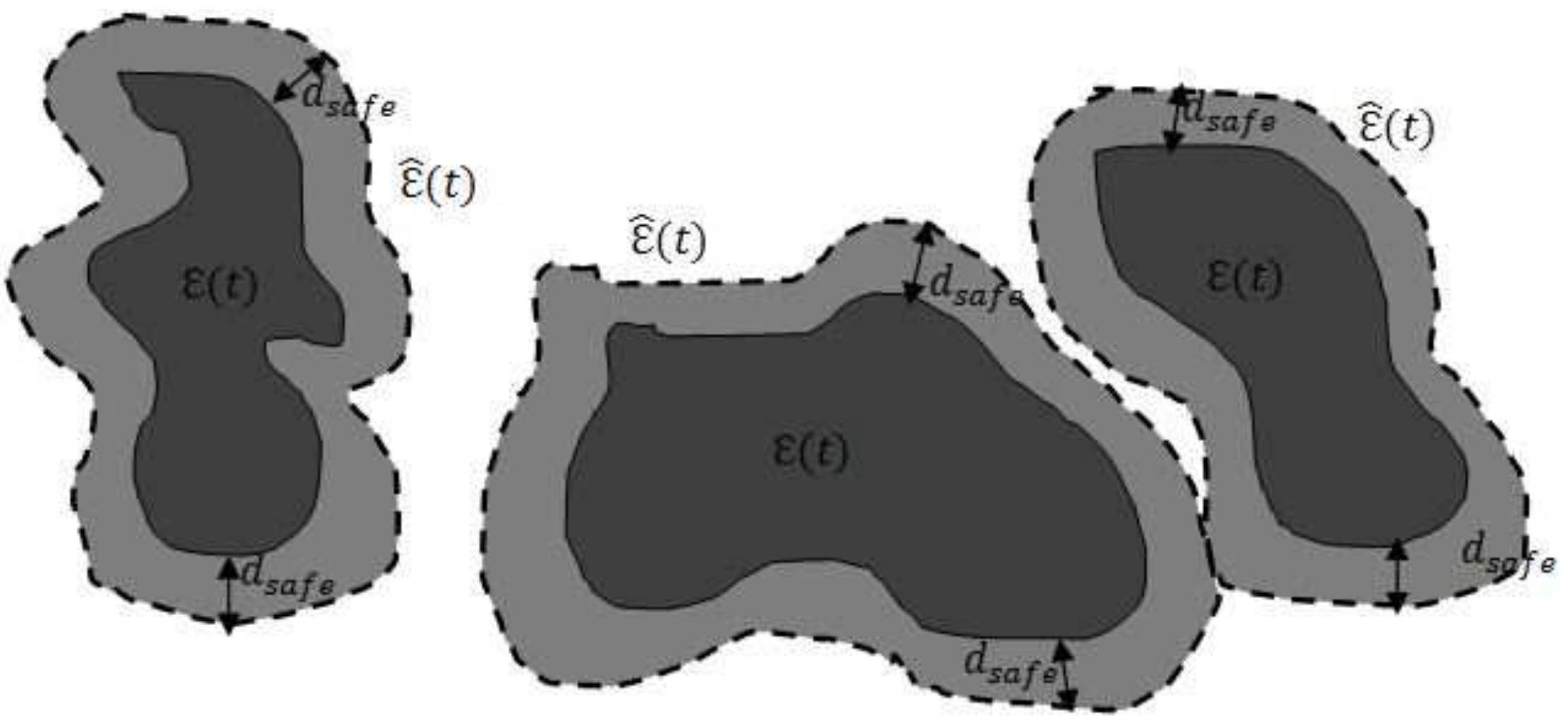}
		\caption{The environment (interior darker region) ${\cal E}(t)$, The enlarged environment (the gray larger region covers ${\cal E}(t)$) $\hat{{\cal E}}(t)$ and the $d_{safe}-$neighbourhood}
		\label{Fp1}
		\end{figure}

		To further ensure the safety of the robot, we define the $d_{safe}$-neighborhood of ${\cal E} (t)$ as the enlarged environment $\hat{\cal E} (t)$, see Fig.~\ref{Fp1}. If the robot can avoid colliding into the enlarged environment $\hat{\cal E} (t)$, the robot is considered as safe with respect to the environment ${\cal E} (t)$.

		\begin{assumption}
		\label{As-2}
		The set $\hat{{\cal E}}(t)$ is closed with piecewise analytic boundary.
		\end{assumption}

\begin{figure}[h]
		\centering
		\subfigure[]{\scalebox{0.55}{\includegraphics{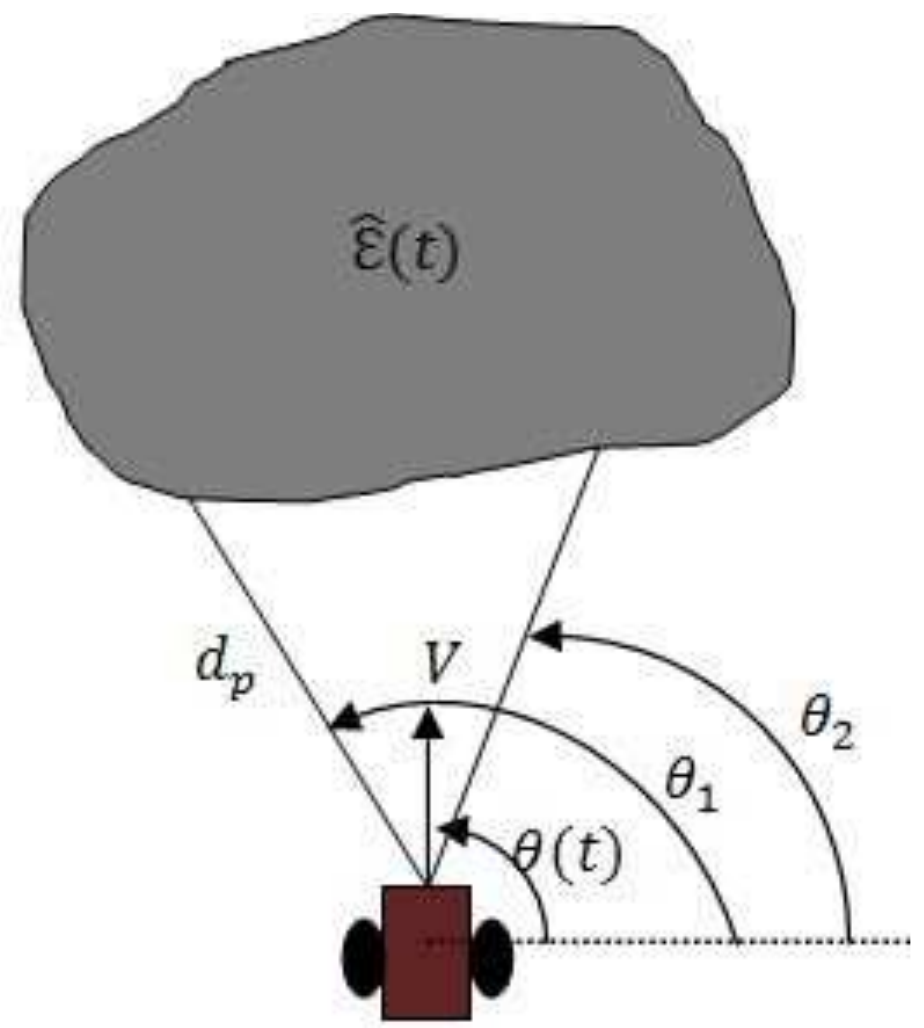}}
		\label{Fp2a}}
		\hfill
		\subfigure[]{\scalebox{0.55}{\includegraphics{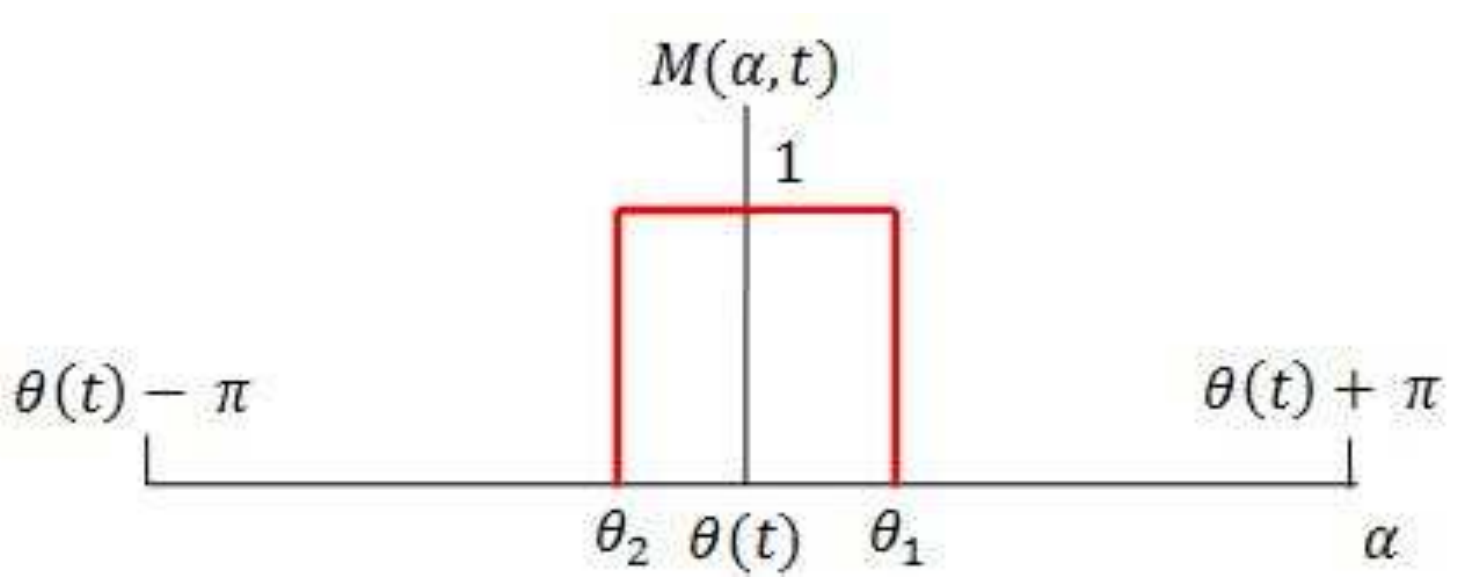}}
		\label{Fp2b}}
		\caption{Illustration of the  binary function $M(\alpha,t)$}
		\label{Fp2}
		\end{figure}

		Let $D_s > 0$ be the sensing range of the robot. Introduce the binary function $M(\alpha,t)\in \{0,1\}$ defined for all $t\geq 0$ and all $\alpha\in (\theta(t)-\pi,\theta(t)+\pi)$ as follows: $M(\alpha,t)=1$ in the direction defined by the angle $\alpha$  and the robot's position at time $t$, there is a point $P$ of an enlarged environment $\hat{{\cal E}}(t)$ at the distance $d_P(t)$ such that
		\begin{equation}
		\label{Pdist}
		 d_P(t)\leq D_s\cos\alpha;
		\end{equation}
		and $M(\alpha,t)=0$ otherwise; see Fig. \ref{Fp2a}.

		\par

\begin{figure}[h]
		\centering
		\includegraphics[width=2.5in]{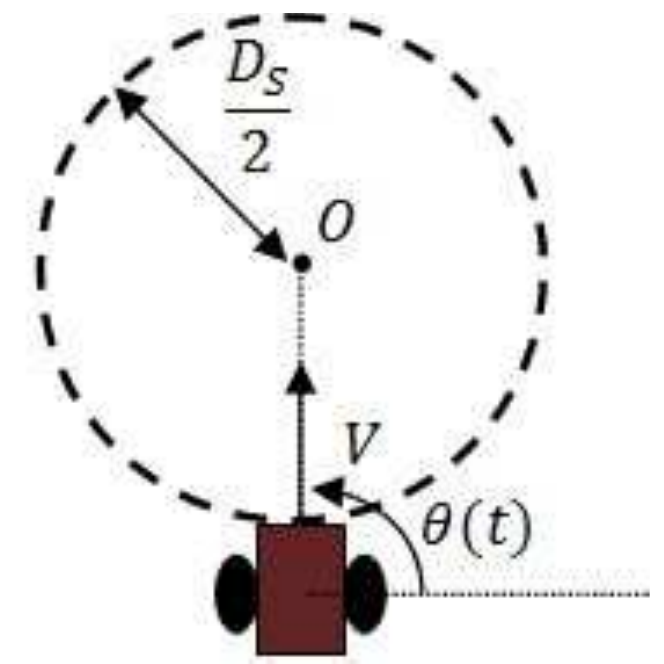}
		\caption{The disc $D(t)$ of diameter $D_s$}
		\label{Fp3}
		\end{figure}

		The equation in (\ref{Pdist}) defines a disc $D$ with diameter $D_s$ centered at point O that is $D_s/2$ ahead of the robot's position at time $t$ in the direction of its current heading $\theta (t)$, see Fig.~\ref{Fp3}. It is straight forward that the binary function $M(\alpha,t)$ indicate whether the boundary of the area $D$ has been invaded by some points of the unknown dynamic environments in the direction defined by the angle $\alpha$.
\par

		In fact, the binary function $M(\alpha,t)$ about the environment ${\cal E} (t)$ is the only available measurement to the unknown dynamic environment for our algorithm. This measurement is relatively easy to be obtained in practice comparing to that required by other existing approaches. Our algorithm may not require any additional information about the obstacles, these obstacles are not necessarily be rigid and their shapes can be time-varying or deformable. Moreover, our algorithm does not required the exact information or estimation about the velocities of the obstacles and their derivatives. 
\par
		Finally, the robot has the knowledge of its own heading $\theta (t)$ and the the desired direction $\theta_0$ .

		Now we wish to illustrate a geometric sense of our choice of  the disc $D(t)$ and the related inequality (\ref{Pdist}) as the core of the 		measurement available to the robot.

		Suppose that ${\cal T}$ is a point-wise obstacle moving with an arbitrary  time-varying velocity $v_T(t)$ satisfying the only constraint 
		\begin{equation}
		\label{VT}
		\|v_T(t)\|\leq V.
		\end{equation}
		 It is obvious that if (\ref{VT}) does not hold, it is impossible for the robot to surely avoid collisions with ${\cal T}$ for any navigation strategy.

		Now we are in a position to state the following proposition.

		\begin{proposition}
		\label{P1}
		Let $t_0>0$ be a given time and let $[x_T(0),y_T(0)]$ be the initial condition of the obstacle ${\cal T}$. Suppose that the robot moves 
ahead with the control input $u(t)=0$ for $t\in [0,t_0]$. Then the obstacle ${\cal T}$ does not collide with the robot over time interval $t\in [0,t_0]$ for any obstacle's satisfying (\ref{VT}) if and only if the initial condition  $[x_T(0),y_T(0)]$ does not belong to the disc $D(0)$ shown in Fig. \ref{Fp3} with the diameter $D_s=2Vt_0$.
\end{proposition}

The proof of Proposition \ref{P1} is presented in~\cite{SAWC13}.

	\section{Navigation Algorithm} \label{A3}

		In this section, we present a description of the proposed navigation algorithm. We start with the necessary assumptions:

		\begin{assumption}
		\label{As-3}
		The initial condition of the robot satisfies $\theta(t)=\theta_0$ and 
		$M(\alpha,0)=0$ for all $\alpha\in (\theta_0-\pi,\theta_0+\pi)$.
		\end{assumption}

		Furthermore, introduce the function $m(t)$ as follows:
		\begin{eqnarray}
		\label{mm}
		 m(t):&=&0~~~~if~~~~M(\alpha,0)=0~~~~ \forall \alpha\in (\theta_0-\pi,\theta_0+\pi);\nonumber\\
		 m(t):&=&1~~~~otherwise.
		\end{eqnarray}	

		The value $m(t)=1$ indicates that there are finite number of open intervals $(\theta_0-\pi,\theta_0+\pi)$ in the set $\alpha\in (\theta_0-\pi,\theta_0+\pi)$. These intervals are sets of direction that are considered safe to the robot in the sense that they are far enough from the robot's current position. 
\par
		Let $j(t)$ be the index of the interval that is closest to the robot's current heading $\theta (t)$, which has two possible results: if $M(t,\theta(t))=0$ meaning that the robot's current heading $\theta (t)$ is safe, thus $\theta(t)\in (A_i^{-},A_i^{+})$ for some $i$, then $j(t):=i$. If $M(t,\theta(t))=1$, meaning that the enlarged environment has crossed $D$ in the direction of robot's current heading, thus $\theta(t) \notin (A_i^{-},A_i^{+})$, then 
		\begin{equation}
		\label{j}
		 j(t):=arcmin \{|A_i^{-}|,|A_i^{+}|\}
		\end{equation}
		over all $i$. In other words, $j(t)$ is the index $i$ for which the minimum over numbers $\{|A_i^{-}|,|A_i^{+}|\}$ is achieved.
Finally, we define the direction $C(t)$ as the middle of the interval closest to $\theta (t)$, i.e.
		\begin{equation}
		\label{CC}
		 C(t):=\frac{A_{j(t)}^{-}+A_{j(t)}^{+}}{2}.
   		\end{equation}
		Our navigation algorithm law for controlling the robot's angular velocity $u(t)$ as discrete times $0,\delta,2\delta,3\delta, \ldots$ is as follows:

		\begin{eqnarray}
		\label{cont}
		{\bf if}~~~~m(k\delta)&=&0~~~{\bf then}~~~u(t):=U_{max} sign (\theta_0-\theta(t))~~~ \forall t\in [k\delta,(k+1)\delta);\nonumber\\
		{\bf if}~~~~m(k\delta)&=&1~~~{\bf then}~~~u(t):=U_{max} sign (C(t)-\theta(t))~~~\forall t\in [k\delta,(k+1)\delta).
		\end{eqnarray}
		Here $C(t)$ is defined by (\ref{CC}) and $sign(\cdot)$ is the standard sign function
		\begin{eqnarray*}  
		 sign (x) := \left\{
                        \begin{array}{lll}
		-1&  if & x < 0\\
		0 &  if & x = 0\\
		1 & if  & x > 0
		\end{array} \right.
		\end{eqnarray*}
		\par

		The idea behind the navigation law (\ref{cont}) is to make the robot's heading as close as possible to the closest safe interval $C (t)$ when it senses unknown environments inside $D(t)$, on the other hand, when the robot does not sense any environment inside $D(t)$, the navigation law (\ref{cont}) makes the robot's heading as close as possible to the desired direction $\theta_0$.
\par

		We state the assumptions for the proofs of the proposed navigation algorithms as follows"

		We now suppose that the environment ${\cal E}(t)$ consists of  several disjoint stationary or moving obstacles
$D_1(t),\ldots,D_n(t)$
 in the plane. These can be deformable.  Moreover, consider the enlarged obstacles 
 $\hat{D}_1(t),\ldots,\hat{D}_n(t)$ that are defined as the
$d_{safe}-$neighbourhood of $D_1(t),\ldots,D_n(t)$. 

\begin{assumption}
\label{As41}
The enlarged obstacles $\hat{D}_1(t),\ldots,\hat{D}_n(t)$ are convex closed sets with piecewise analytical boundary. 
\end{assumption}

Furthermore, we define the maximum displacement of the set
 $\hat{D}_i(t)$ over time interval $[t_1,t_2)$ as
 \[
 \rho_{i}(t_1,t_2):=\max_{r\in \hat{D}_i(t_2)}
 \min_{h\in \hat{D}_i(t_1)}\|r-h\|.
 \]
 Moreover, introduce the following upper bound on the environment displacement speed:
 \begin{equation}
 \label{VD}
 V_E:=\max_{i=1,\ldots,n}\sup_{t_2>t_1\geq 0}
\frac{ \rho_i(t_1,t_2)}{(t_2-t_1)}.
 \end{equation}

 \begin{assumption}
 \label{As2}
 The following inequalities hold:
 \begin{equation}
 \label{Ase1}
 \frac{V\sin(U_{max}\delta)}{U_{max}}>V_E\delta,
 \end{equation}
 \begin{equation}
 \label{Ase2}
 D_s>2(V+V_E)\delta,
 \end{equation}
 \begin{equation}
 \label{Ase3}
 (D_s-2V\delta)\sin(U_{max}\delta)>2V_E\delta,
 \end{equation}
 \begin{equation}
 \label{Ase4}
 (D_s-2V\delta)(1-\cos(U_{max}\delta))>2V_E\delta.
 \end{equation}
 \end{assumption}
  
\begin{remark}
It should be noticed that Assumption \ref{As2}
is not very restrictive. Indeed, for small $\delta$, 
$sin(U_{max}\delta)$ is close to $U_{max}\delta)$, therefore
the requirement (\ref{Ase1}) becomes close to the inequality
$V>V_E$. It is obvious that if this inequality does not hold, it is impossible for the robot to surely avoid collisions for any navigation strategy. Furthermore, the requirements (\ref{Ase2}), (\ref{Ase3}) and (\ref{Ase4}) can be satisfied
by taking a large enough sensing parameter $D_s$.
\end{remark}

We define the distance between enlarged obstacles
$\hat{D}_i(t)$ and $\hat{D}_j(t), i\neq j$
 as 
\[
d_{ij}(t) :=\min_{r\in \hat{D}_i(t),h\in \hat{D}_j(t)}\|r-h\|.
\]
\begin{assumption}
 \label{As43}
 For all $i\neq j, t\geq 0$ the following inequality 
$d_{ij}(t)>D_s+2V\delta$
 holds.
 \end{assumption}
 \begin{assumption}
 \label{As4}
We also assume that for any time $t$, there is no a point
of the enlarged environment $\hat{{\cal E}}(t)$ in the 
direction that is opposite to the robot's current velocity vector and at the distance less than $2V\delta$ from the robot's current position.
 holds.
 \end{assumption}
 
 \begin{theorem}
 \label{T41}
 Suppose that Assumptions \ref{As-3}--\ref{As4} hold. Then the robot navigation strategy (\ref{cont}) is  safe collision free.
 \end{theorem}

		The mathematical proof of the theorem~\ref{T41} is presented in~\cite{SAWC13}.

	\section{Computer Simulation Results}

		In this section, we present computer simulation results for a wheeled mobile robot navigating in unknown environments. The movement of the mobile robot is guided by the proposed navigation algorithm. The simulations are performed using MobotSim and Matlab. The desired direction is same as the initial direction of the robot, i.e. $\theta_0 = \theta (0)$.
\par
\begin{figure}[h]
		\centering
		\includegraphics[width=3.5in]{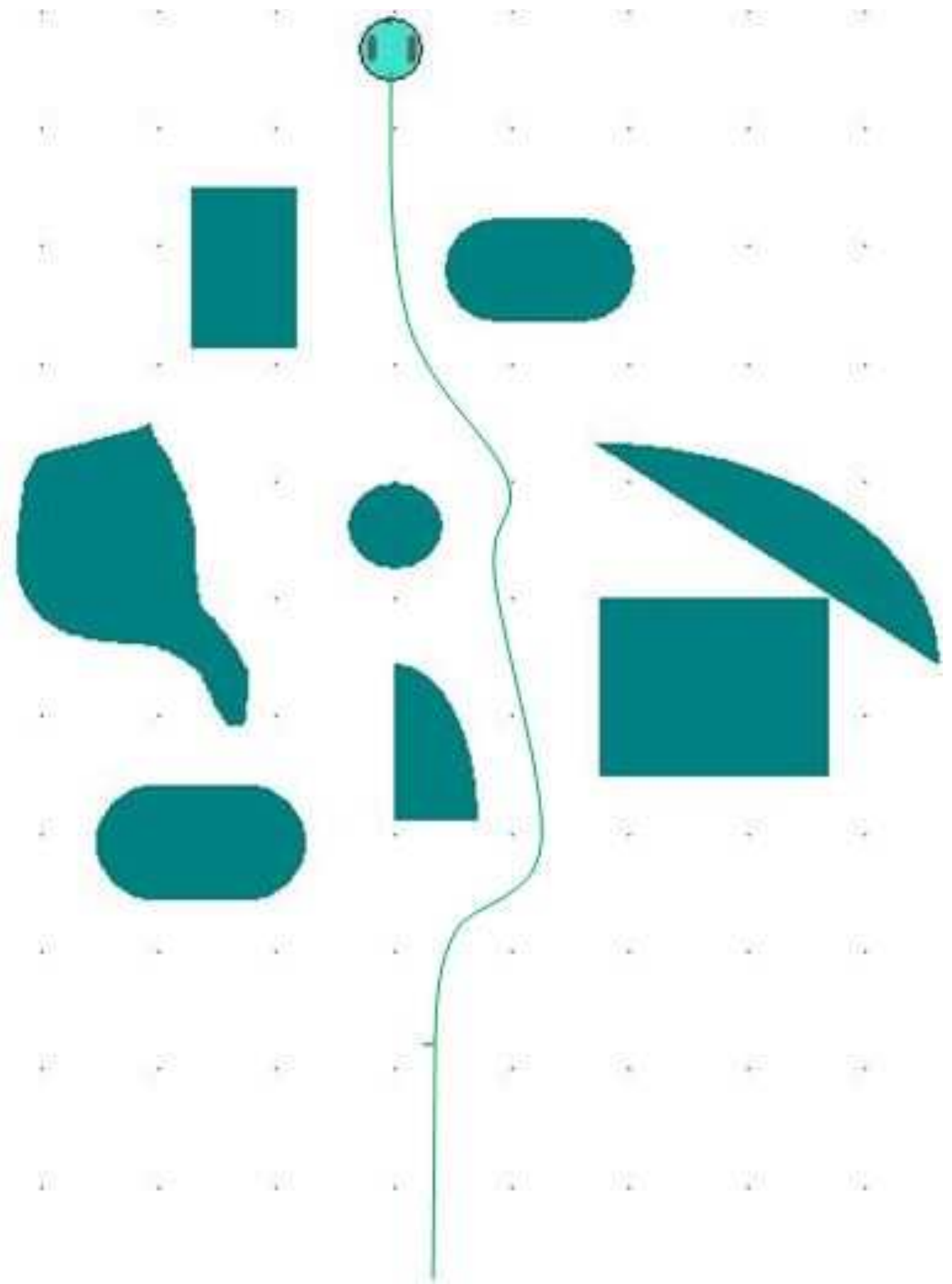}
		\caption{Robot navigating in unknown static environment with irregular obstacles}
		\label{c4.sim1}
		\end{figure}

		We start with a simple scenario of navigating a mobile robot in an unknown static environment. The mobile robot does not have the information about the position of the obstacle in advance, more importantly, unlike many of the existing navigation algorithms which restrict the shapes of the obstacles for computational simplicity, the shapes of the obstacles are random and irregular in this simulation as shown in Fig.~\ref{c4.sim1}. The proposed navigation algorithm guides the mobile robot through the vacancy between the static obstacles as shown in Fig.~\ref{c4.sim1} 

		In the following simulations performed in Matlab, we depict the mobile robot as a gray disk, we also indicate the current heading of the robot as black arrow and the intended direction (middle of vacancy) as a red arrow. Furthermore, the sensing range of the robot is represented by a dashed circle around it.
\par
\begin{figure}[!h]
		\centering
		\subfigure[]{\scalebox{0.55}{\includegraphics{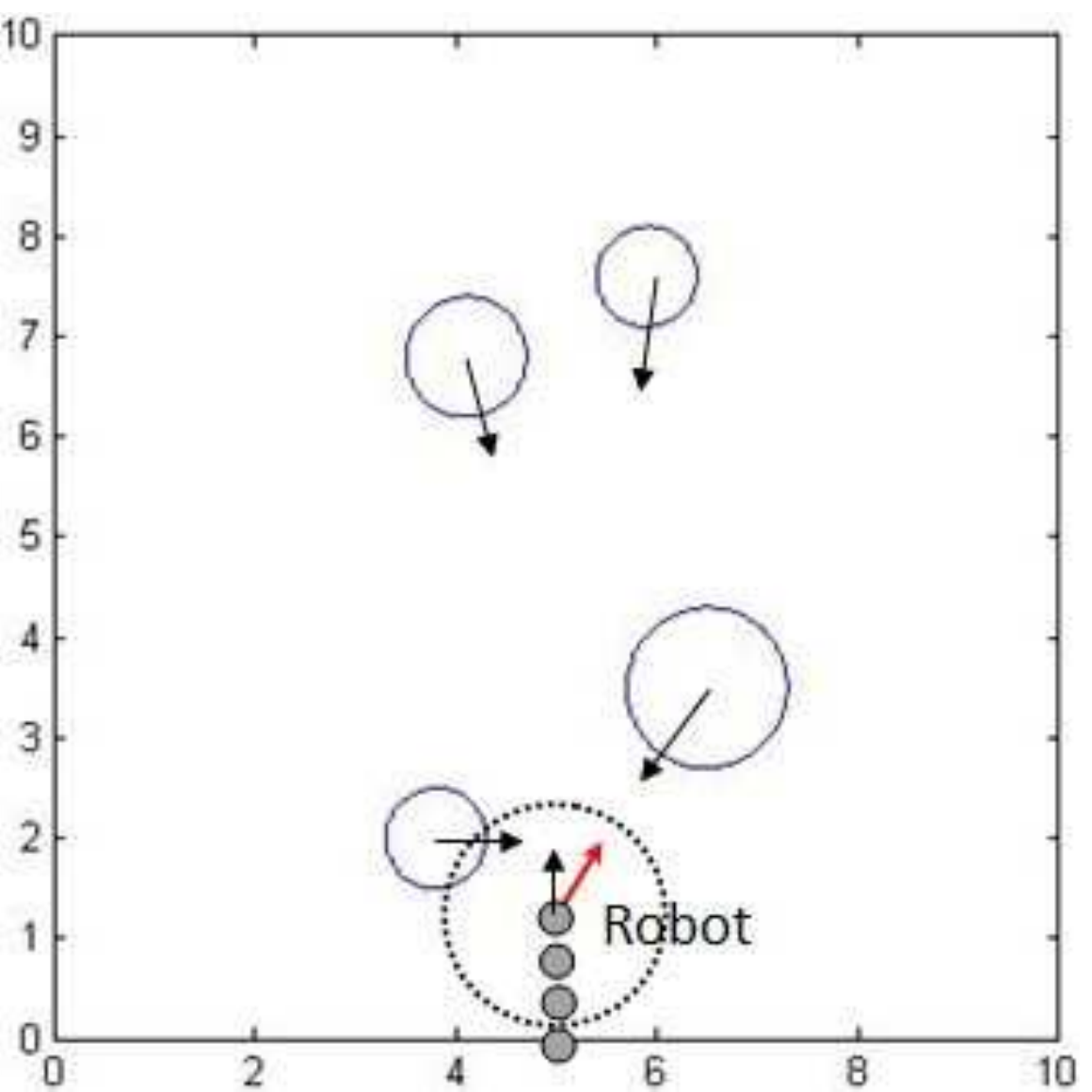}}
		\label{c4.sim21}}
		\subfigure[]{\scalebox{0.55}{\includegraphics{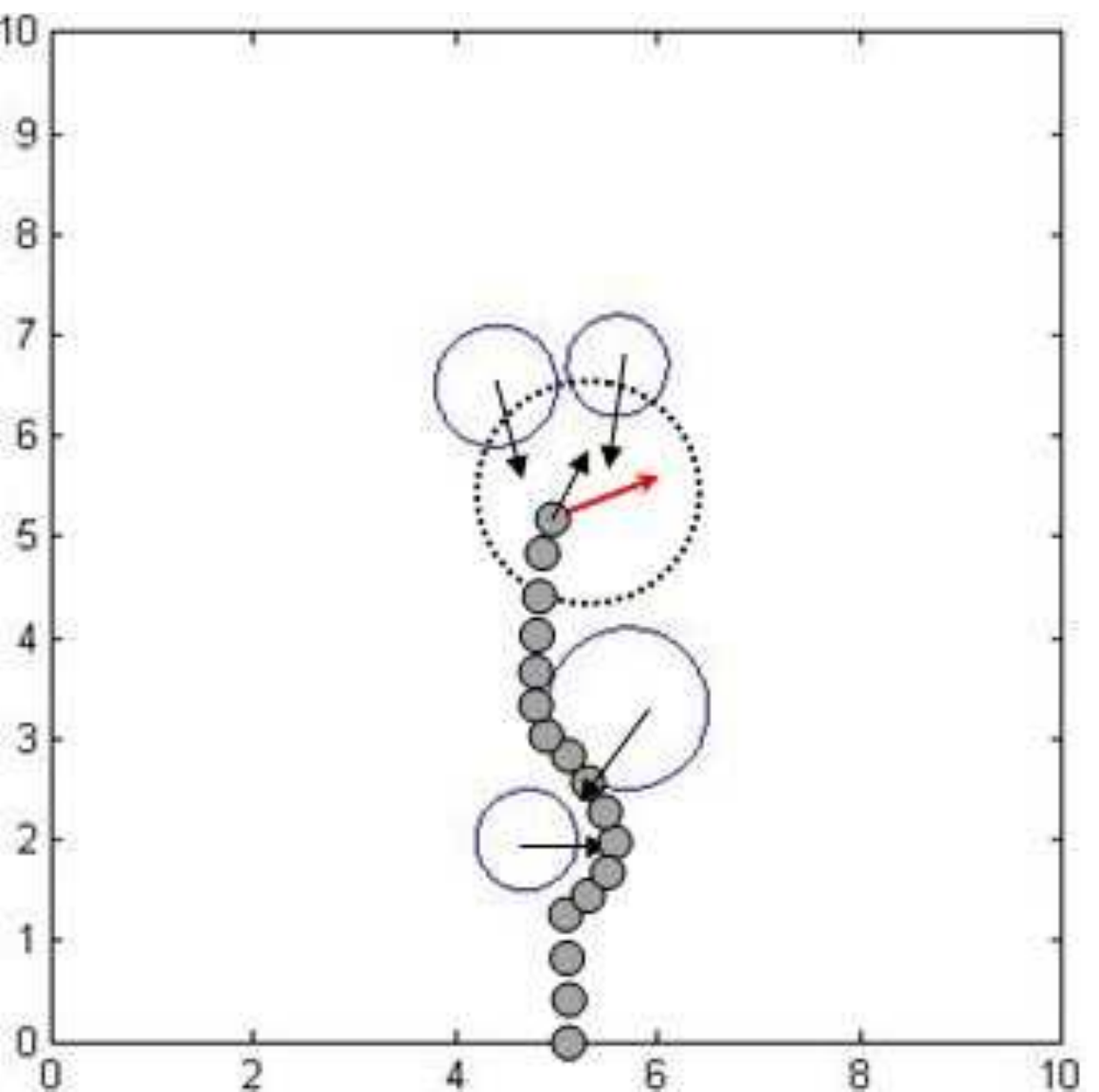}}
		\label{c4.sim22}}
		\subfigure[]{\scalebox{0.55}{\includegraphics{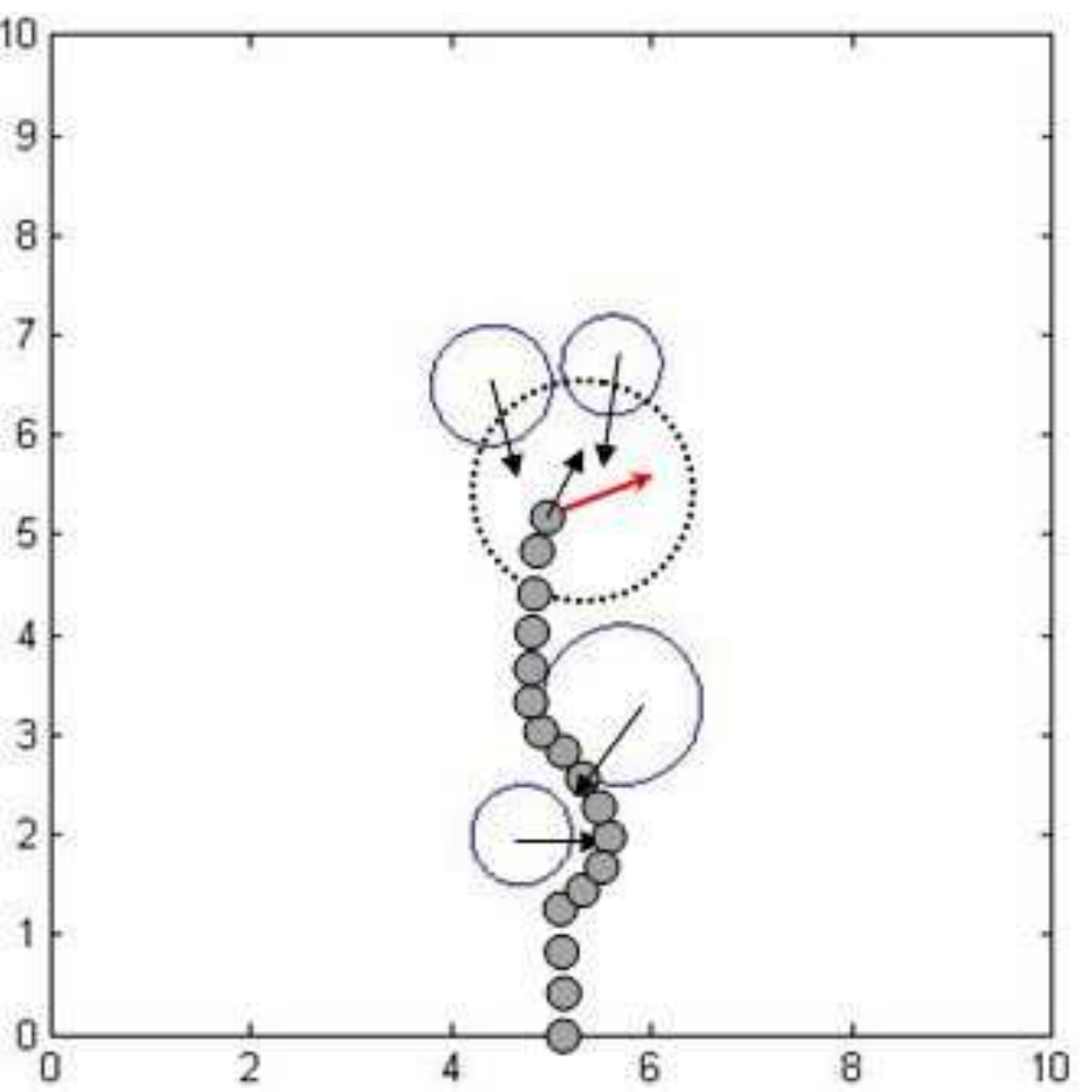}}
		\label{c4.sim23}}
		\subfigure[]{\scalebox{0.55}{\includegraphics{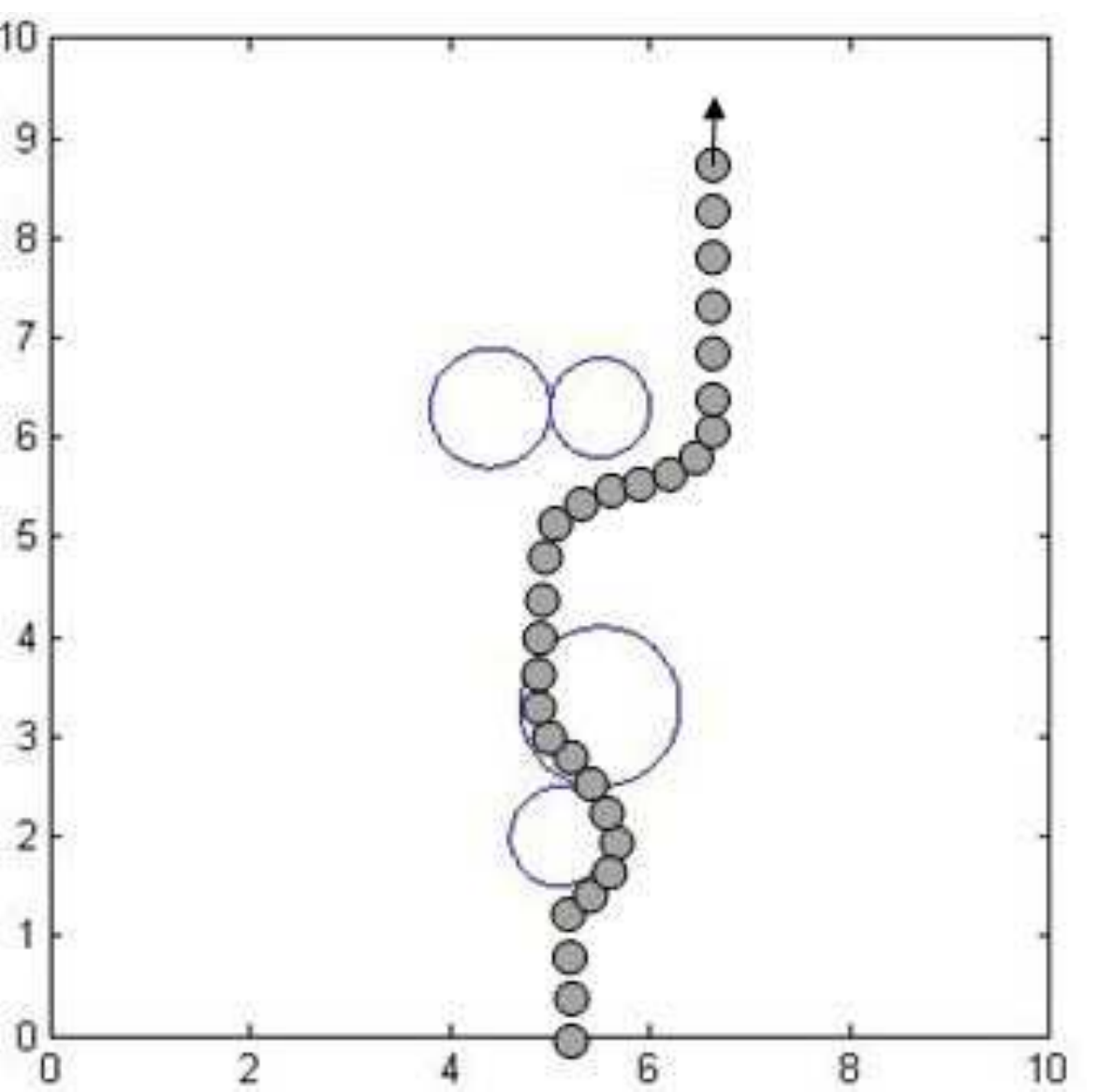}}
		\label{c4.sim24}}
		\caption{Robot navigating in a unknown dynamic environments with multiple moving obstacles }
		\label{c4.sim2}
		\end{figure}

		In Fig.~\ref{c4.sim2}, the robot is navigating in a dynamic environment with four obstacles. Once the robot sense one (Fig.~\ref{c4.sim21}) or more (Fig.~\ref{c4.sim22} and Fig.~\ref{c4.sim23}), it will steer its heading towards a safe direction to avoid the obstacles. Furthermore, in the ambiguous cases where two obstacles are considered as "dangerous" by the robot, the robot is able to decide whether to go through the middle of gap between the obstacles (when the space is sufficient in Fig.~\ref{c4.sim22}) or go around the obstacles (when the space is too small in Fig.~\ref{c4.sim23}). The overall path is shown in Fig.~\ref{c4.sim24}. This simulation demonstrate the capability of the proposed navigating algorithm to distinguish ambiguous situation when multiple obstacles are considered "dangerous" and make the correct decisions on-the-fly, which is found difficult by many of the existing approaches. 
\par

		\par

		The performance of the proposed navigation algorithm is compared with that of the Velocity Obstacle Approach (VOA) \cite{FP93,FP98}, which is one of the most well-known navigation algorithms. It should be mentioned that in order to compare the performance of these two algorithms, we slightly change the objective of our proposed navigation algorithm from "desired direction reaching" to "target position reaching" by making the desired direction time-varying, which equals to the direction from the robot current location to the target position.
\par
		In the following simulation, the velocity obstacles estimated by VOA are depicted by gray circles. Some of the potentially safe area are occupied by the these velocity obstacles because of the estimation, whereas the proposed navigation algorithm efficiently utilises these paths and bypass the obstacles with a shorter path. In other words, with the proposed navigation algorithm, the robot efficiently seeks a free path through the group of obstacles, whereas with the VOA, the robot is avoiding the crowd of obstacles which results in a much longer path. The performance of the proposed navigation algorithm is shown in figures in upper row and that of VOA is shown in lower row

		 \begin{figure}[h]
		\centering
		\subfigure[]{\scalebox{0.3}{\includegraphics{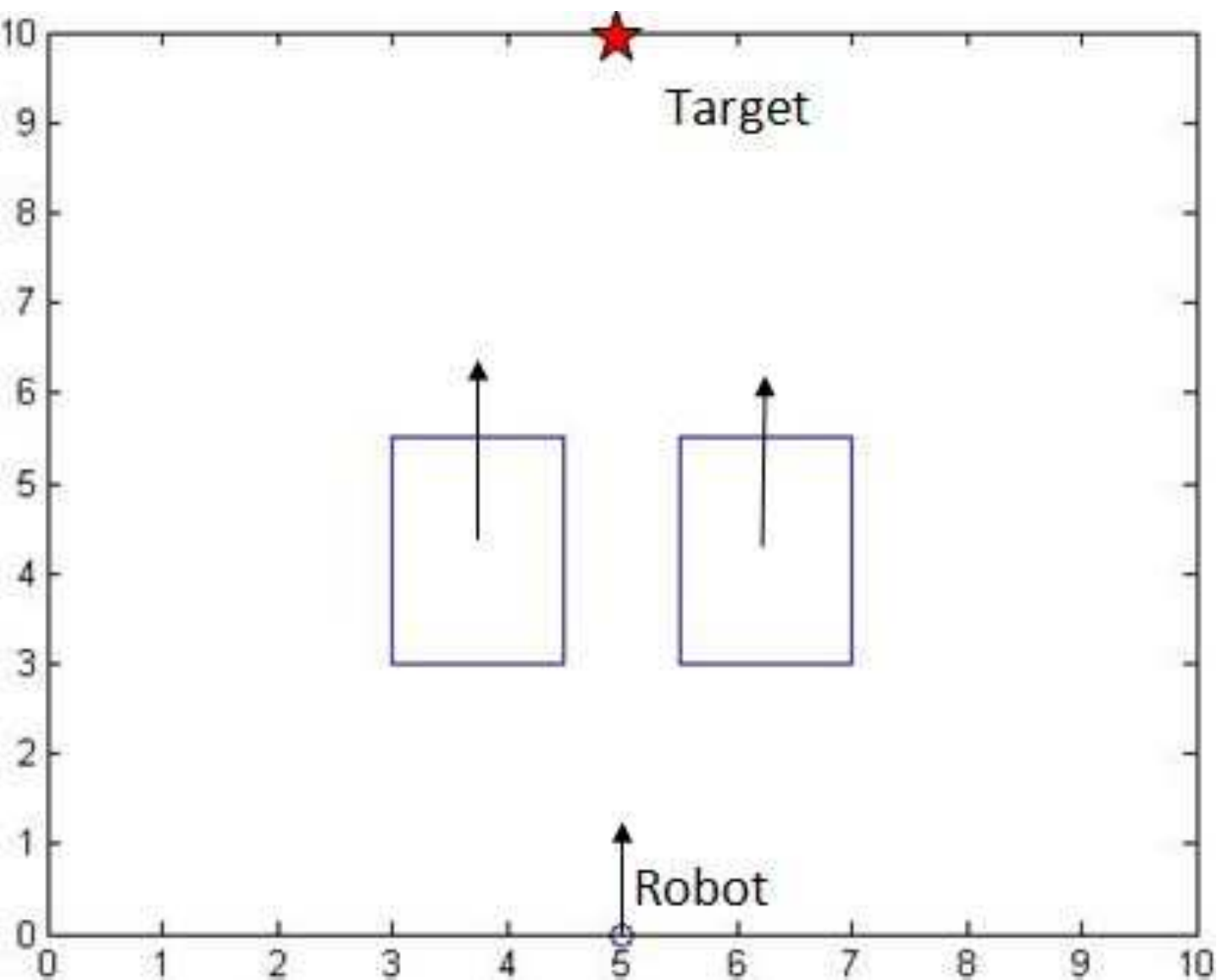}}
		\label{sim31}}
		\subfigure[]{\scalebox{0.3}{\includegraphics{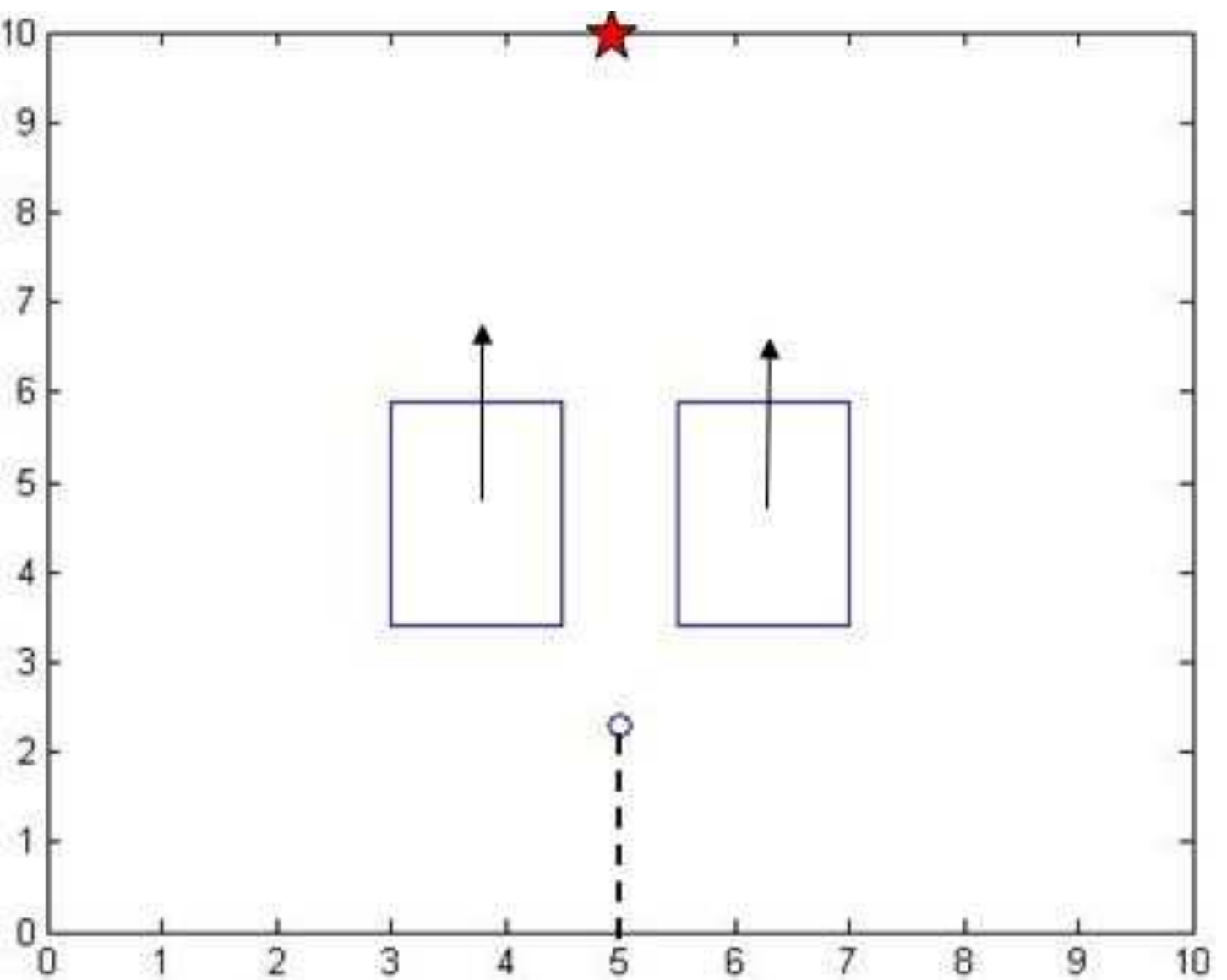}}
		\label{sim34}}
		\subfigure[]{\scalebox{0.3}{\includegraphics{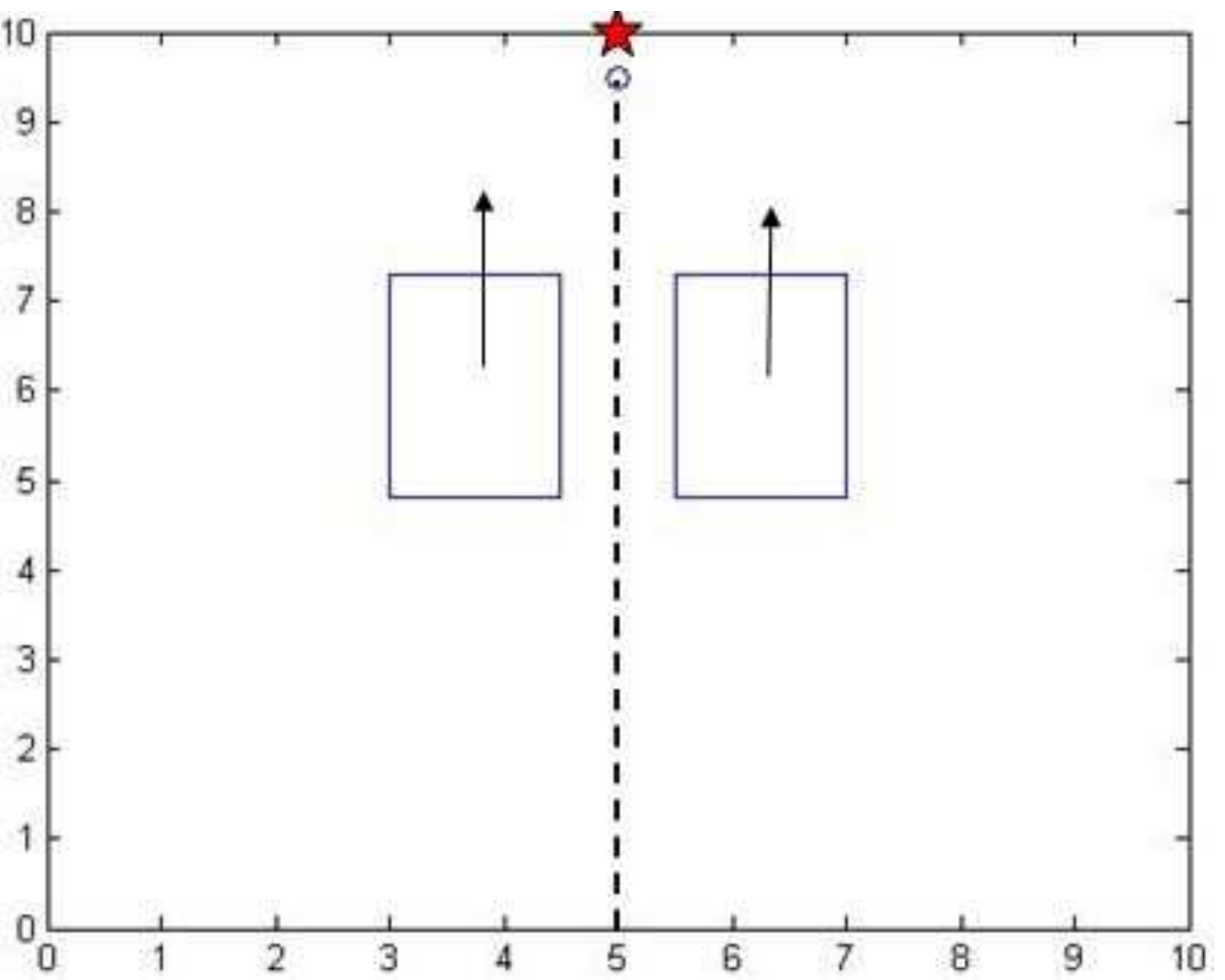}}
		\label{sim32}}
		\subfigure[]{\scalebox{0.3}{\includegraphics{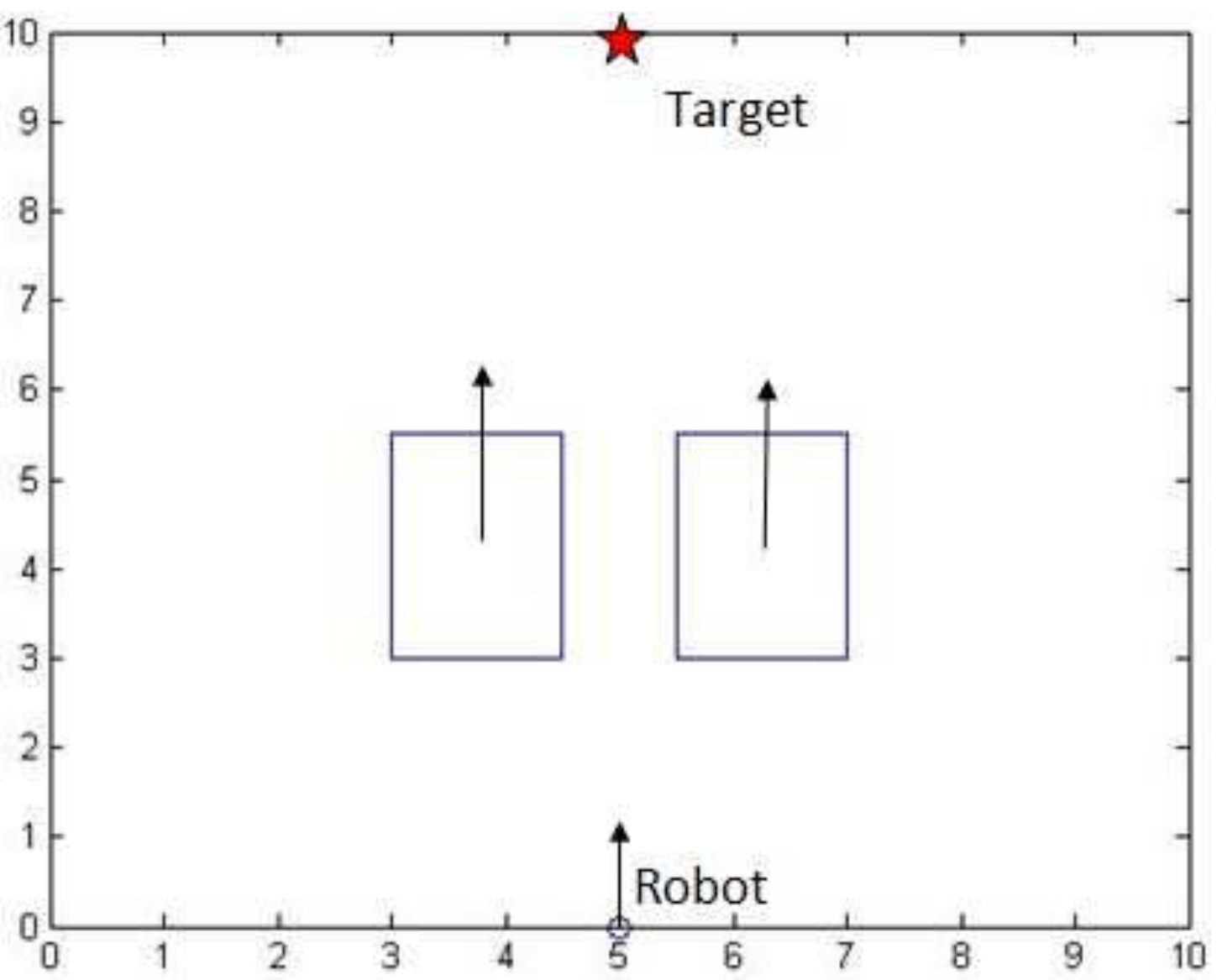}}
		\label{sim35}}
		\subfigure[]{\scalebox{0.3}{\includegraphics{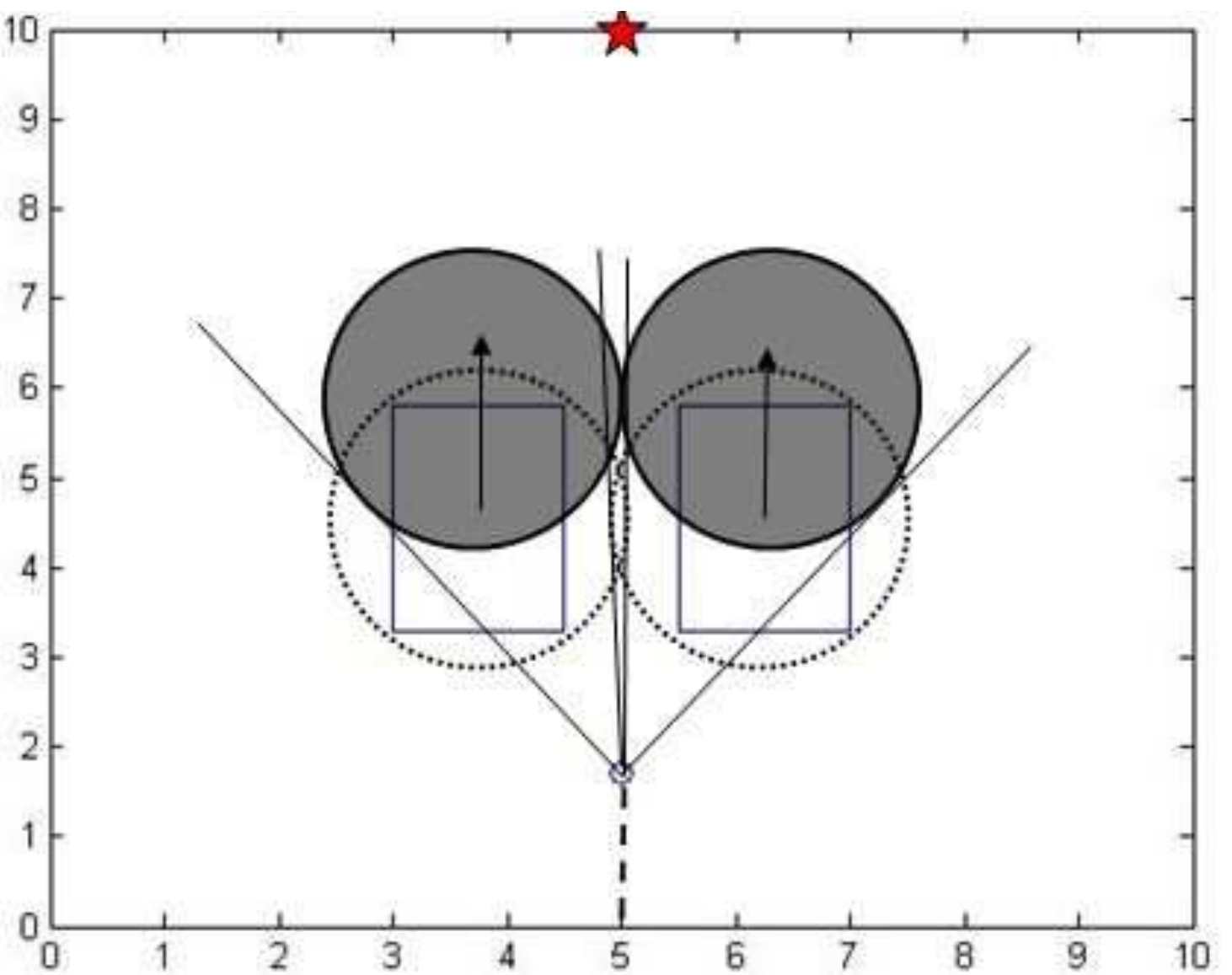}}
		\label{sim33}}
		\subfigure[]{\scalebox{0.3}{\includegraphics{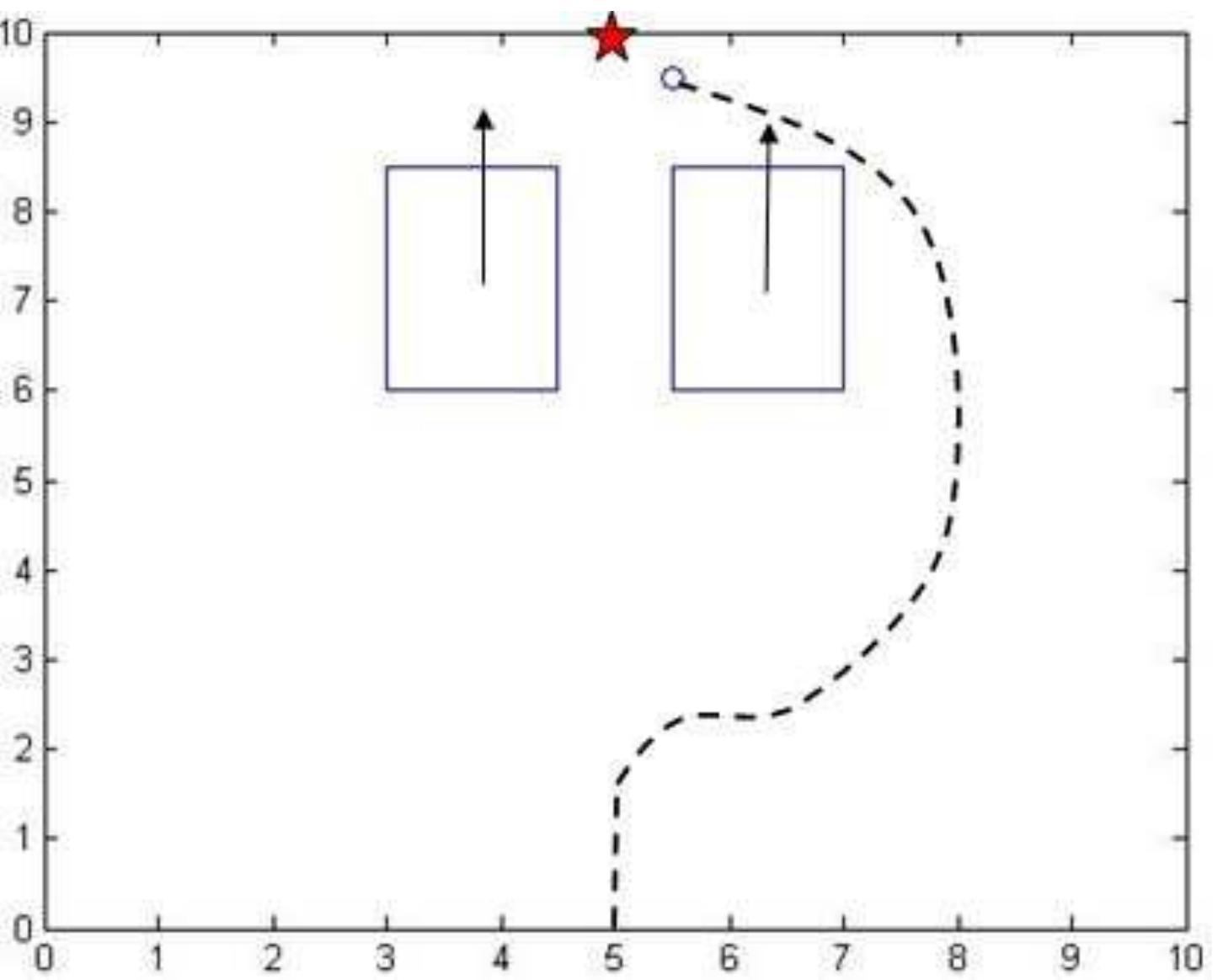}}
		\label{sim36}}
		\caption{Performance comparison: two square obstacles moving side by side}
		\label{sim3}
		\end{figure} 
		\par

		 \begin{figure}[h]
		\centering
		\subfigure[]{\scalebox{0.3}{\includegraphics{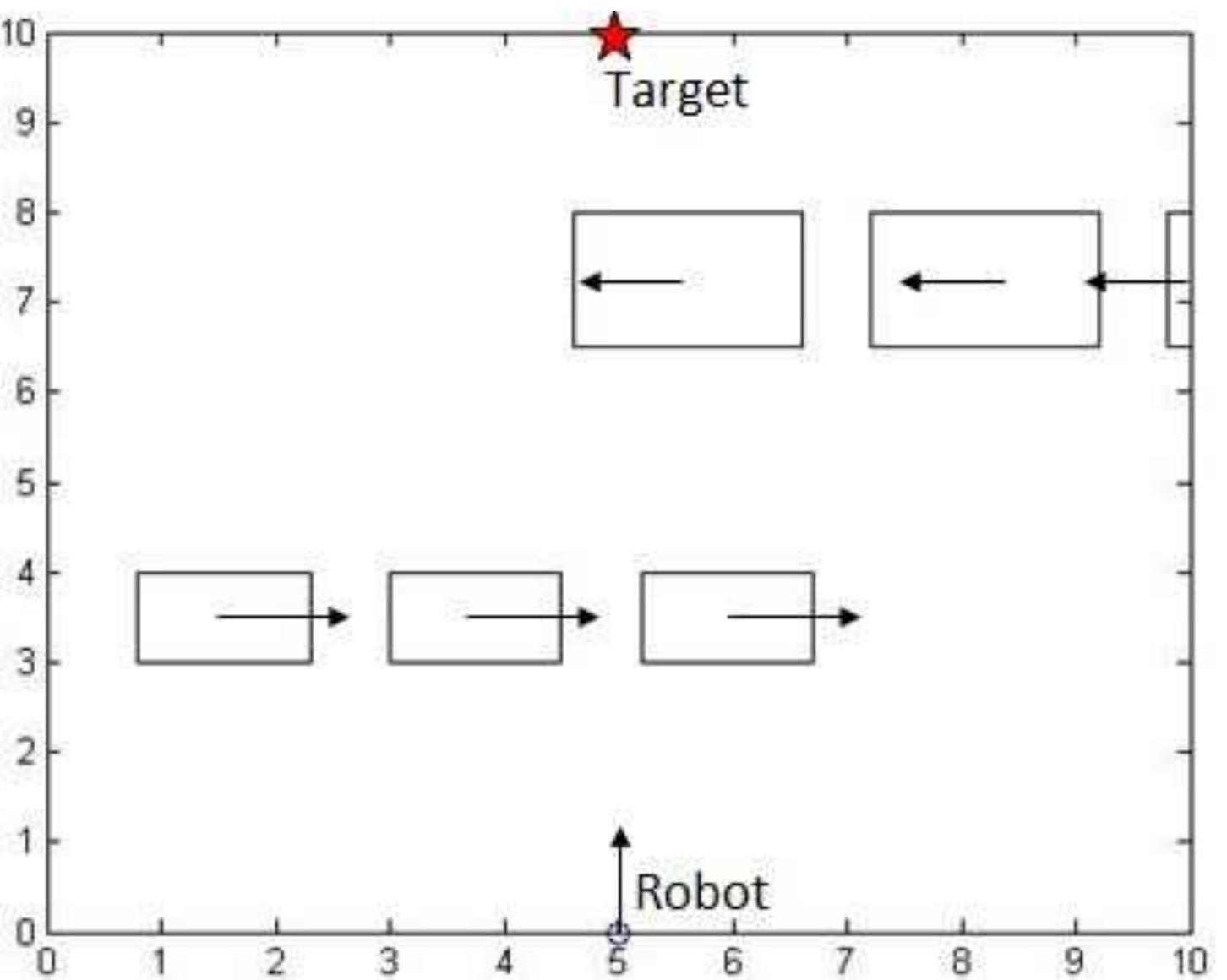}}
		\label{sim41}}
		\subfigure[]{\scalebox{0.3}{\includegraphics{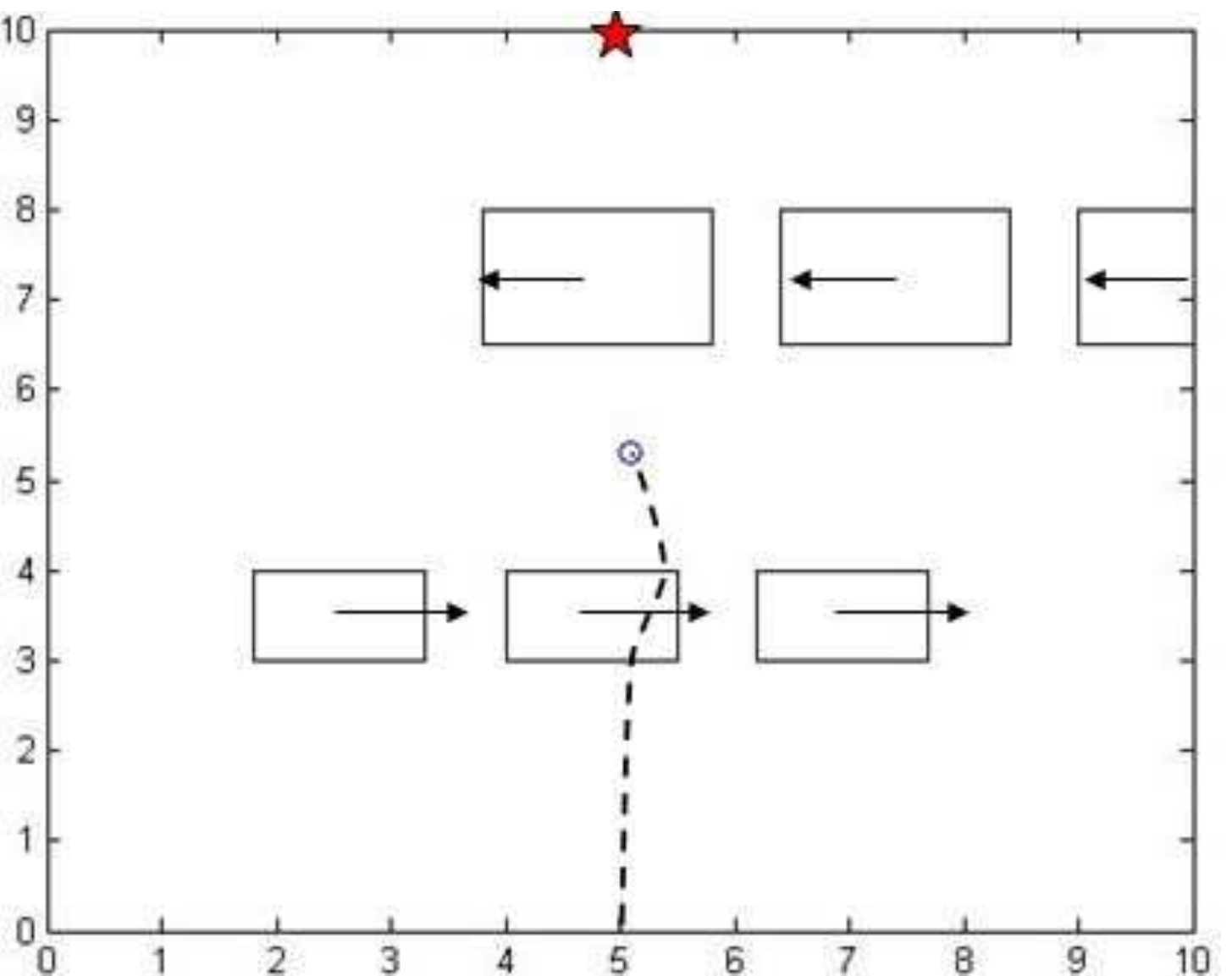}}
		\label{sim44}}
		\subfigure[]{\scalebox{0.3}{\includegraphics{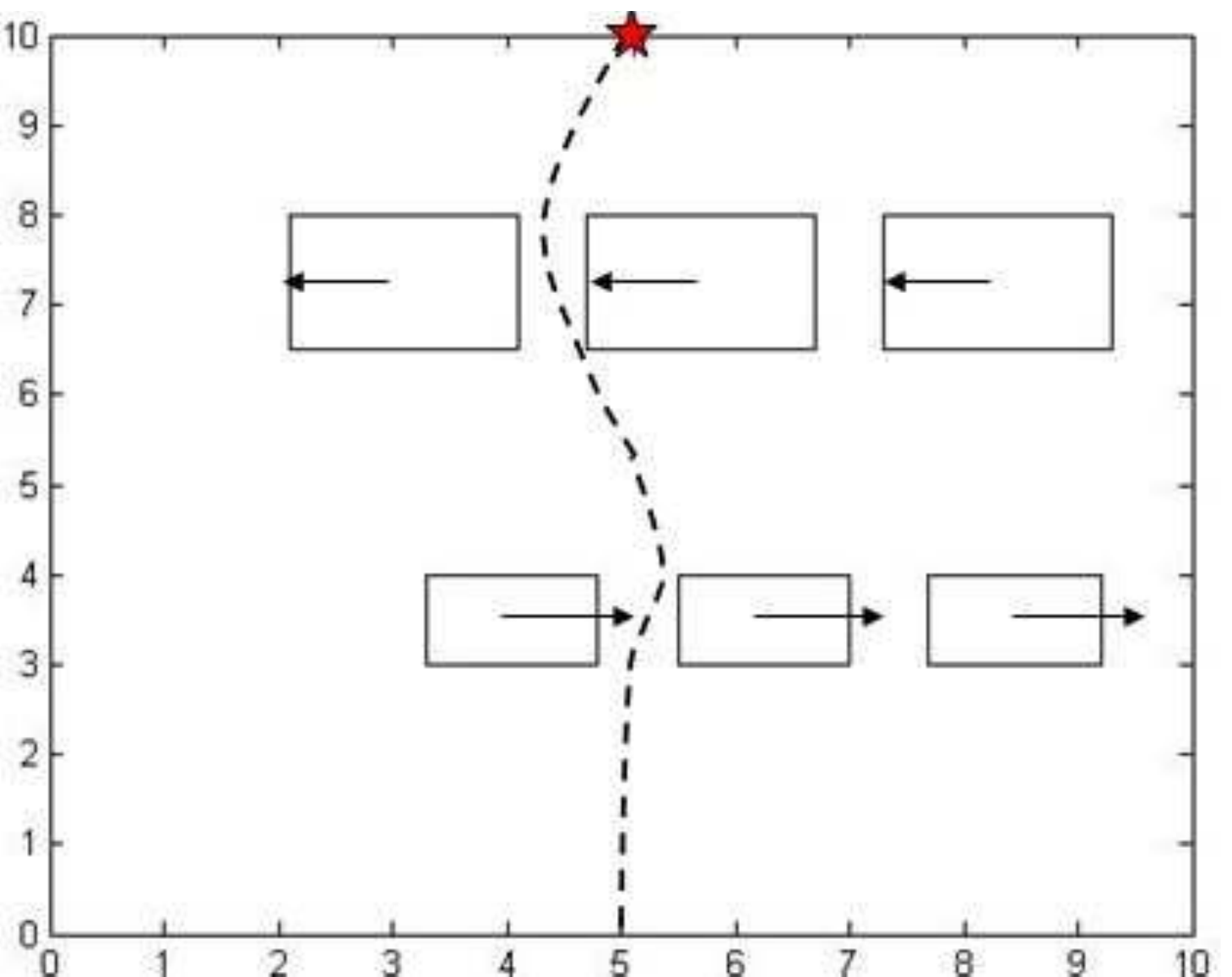}}
		\label{sim42}}
		\subfigure[]{\scalebox{0.3}{\includegraphics{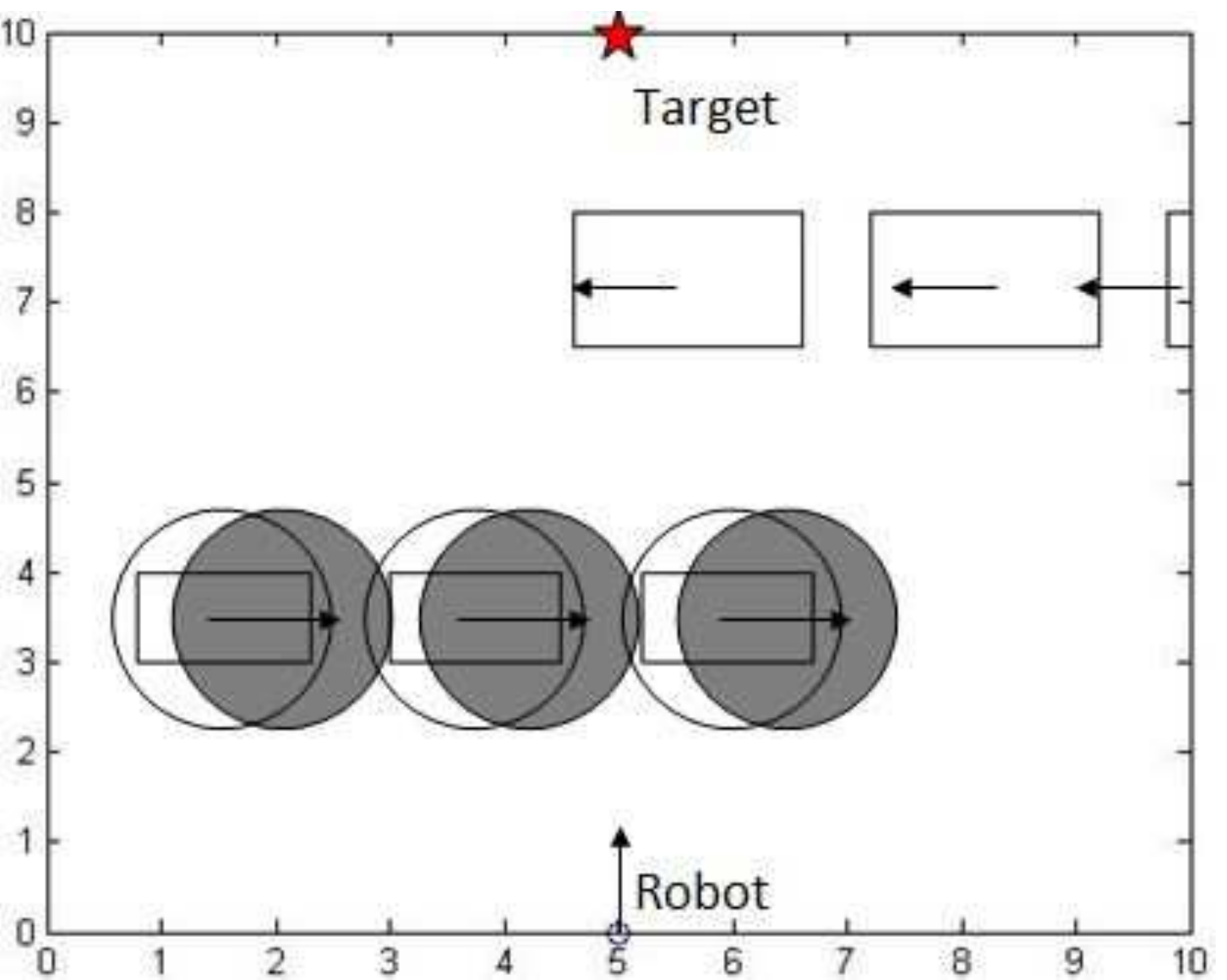}}
		\label{sim45}}
		\subfigure[]{\scalebox{0.3}{\includegraphics{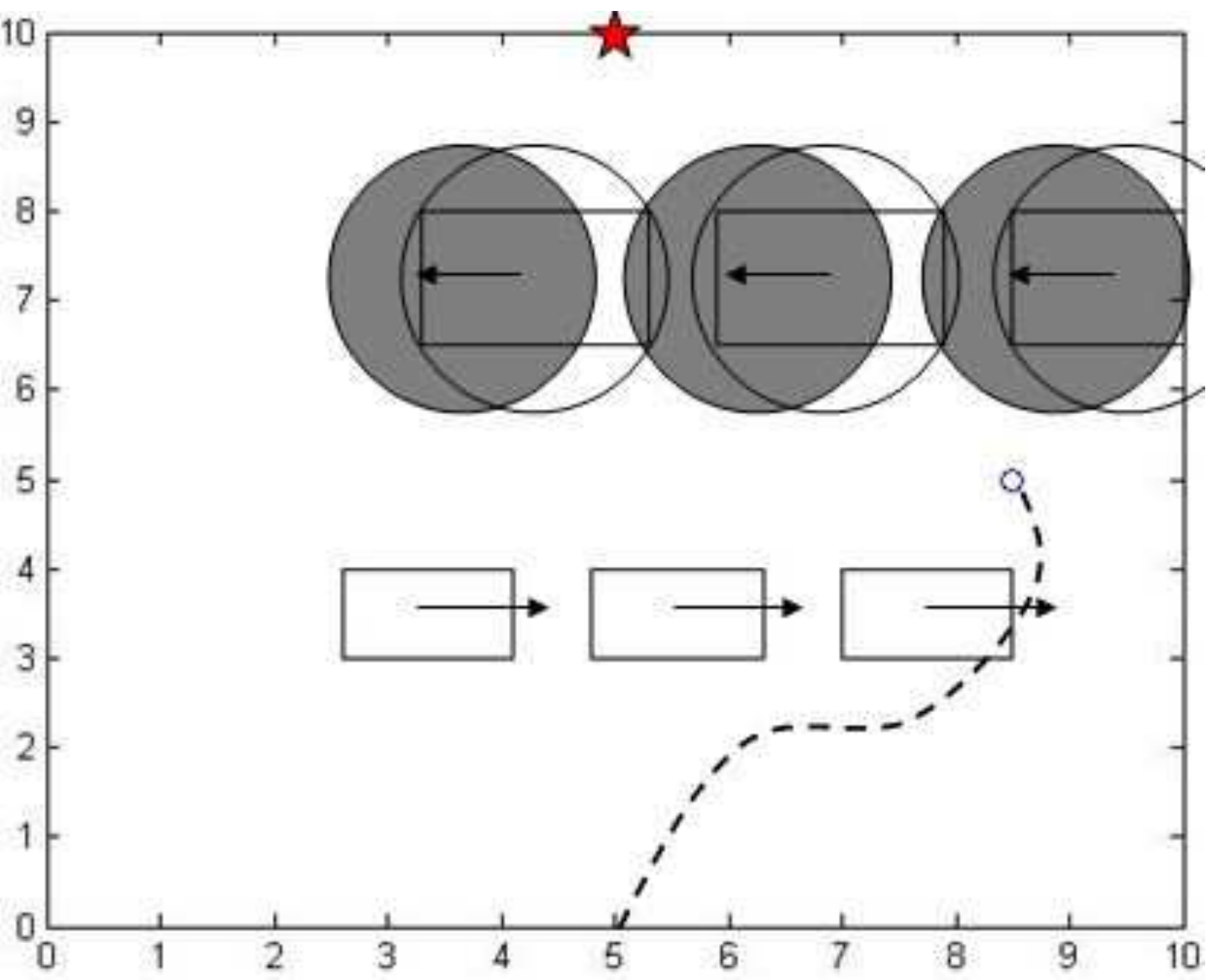}}		
		\label{sim43}}
		\subfigure[]{\scalebox{0.3}{\includegraphics{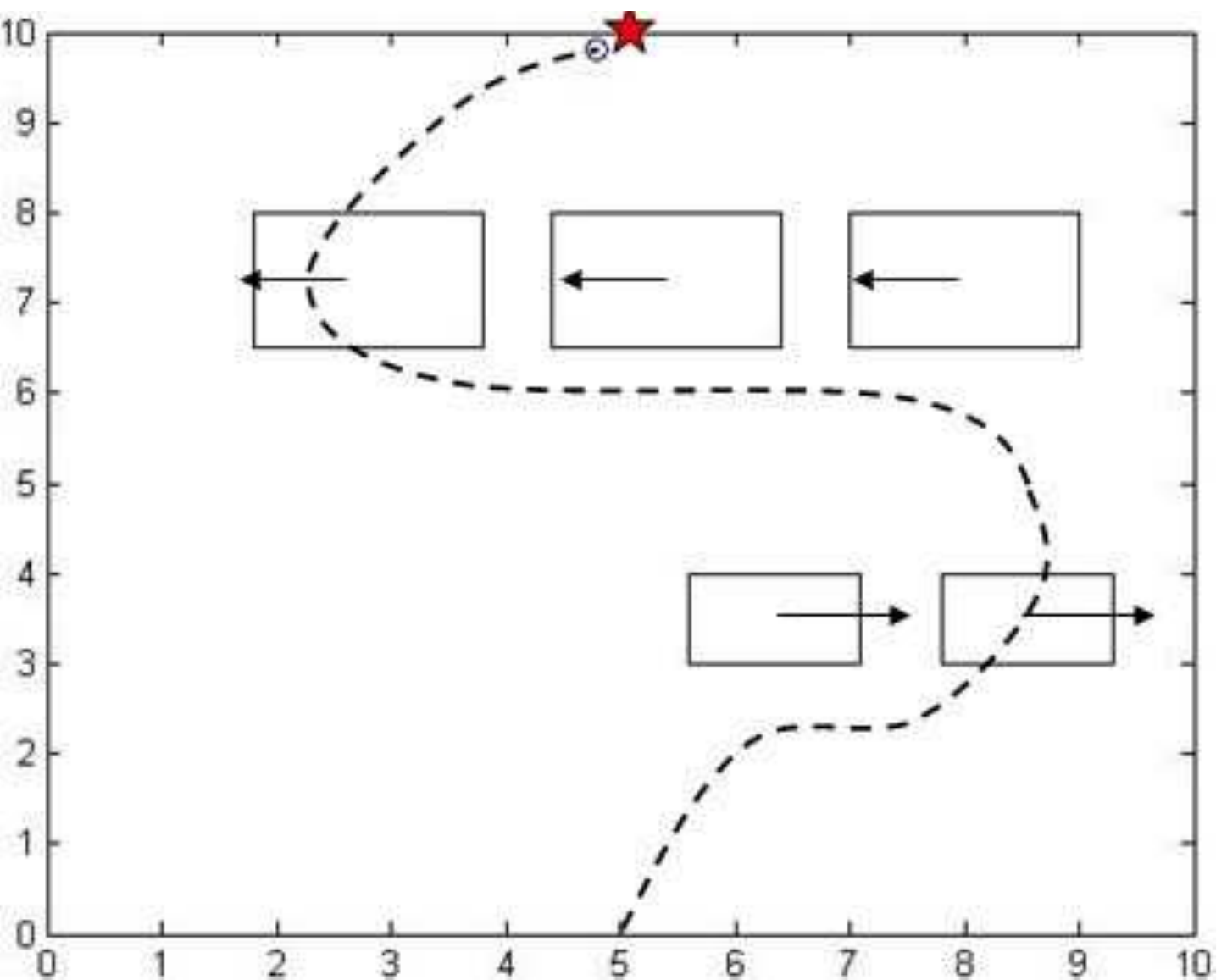}}
		\label{sim46}}
		\caption{Performance comparison 2: chains of obstacles}
		\label{sim4}
		\end{figure} 
		\par

		 \begin{figure}[h]
		\centering
		\subfigure[]{\scalebox{0.42}{\includegraphics{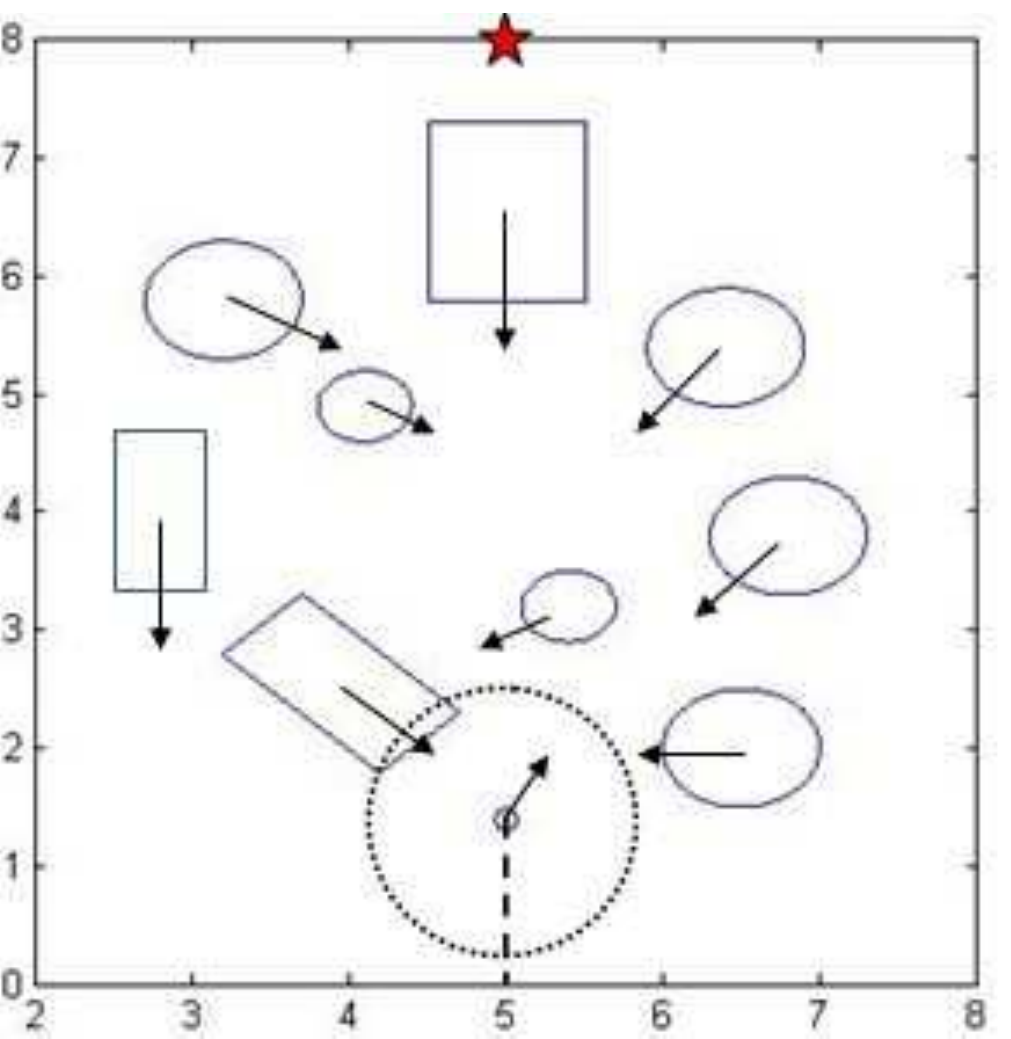}}
		\label{sim51}}
		\subfigure[]{\scalebox{0.42}{\includegraphics{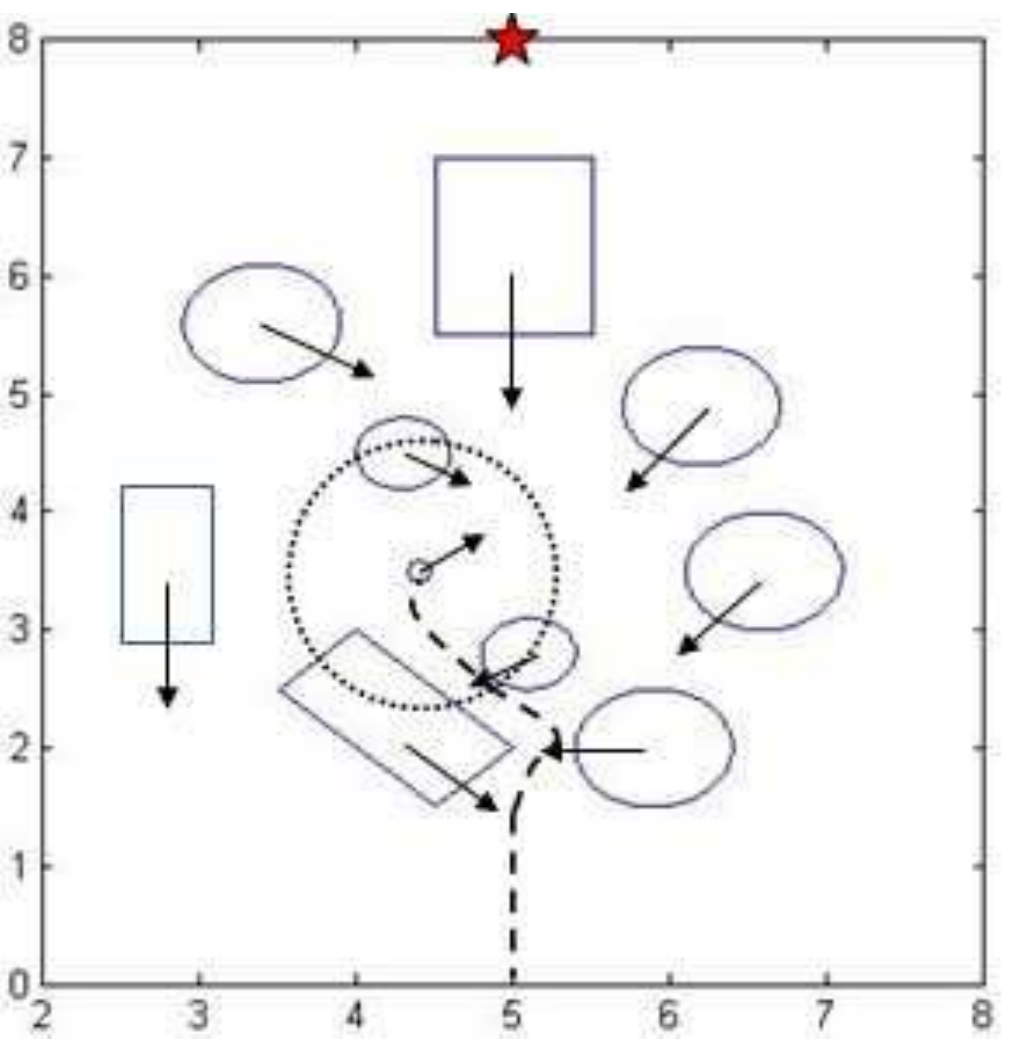}}
		\label{sim54}}
		\subfigure[]{\scalebox{0.42}{\includegraphics{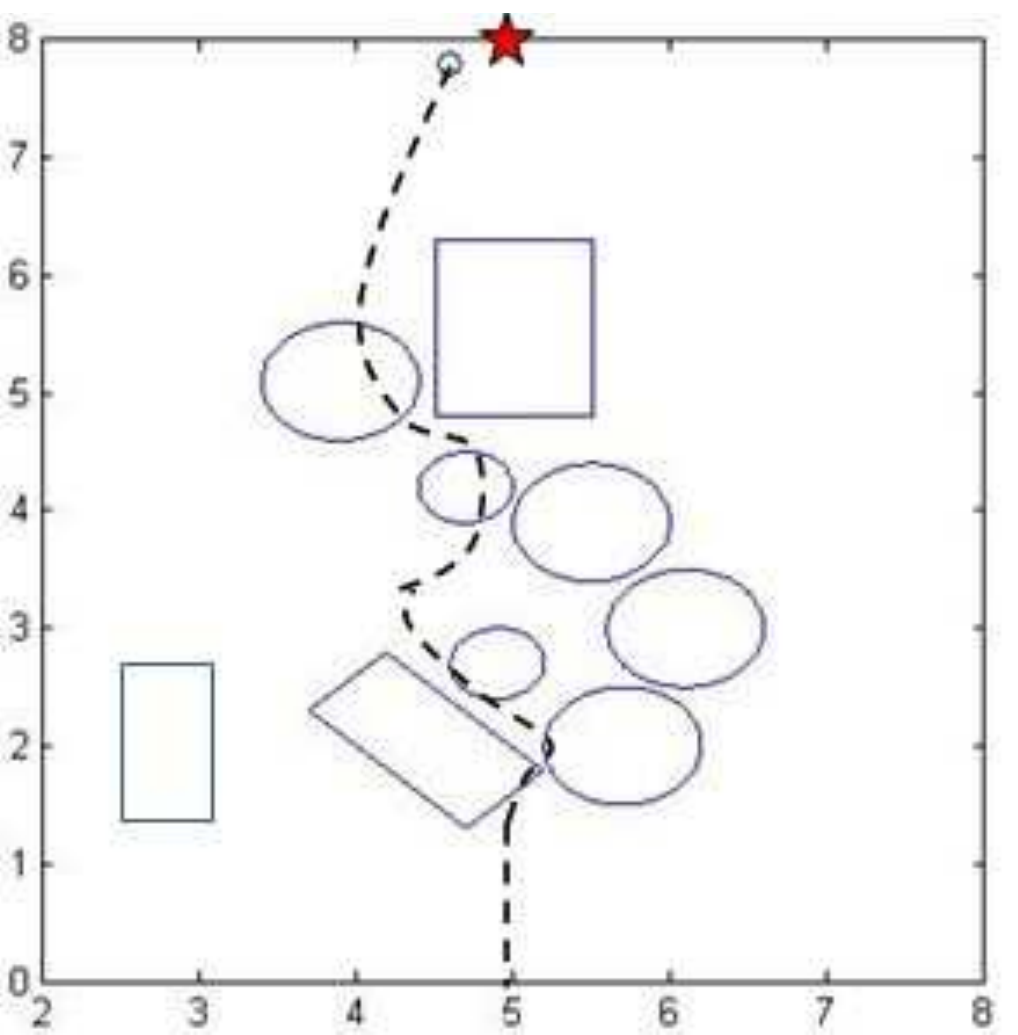}}
		\label{sim52}}
		\subfigure[]{\scalebox{0.42}{\includegraphics{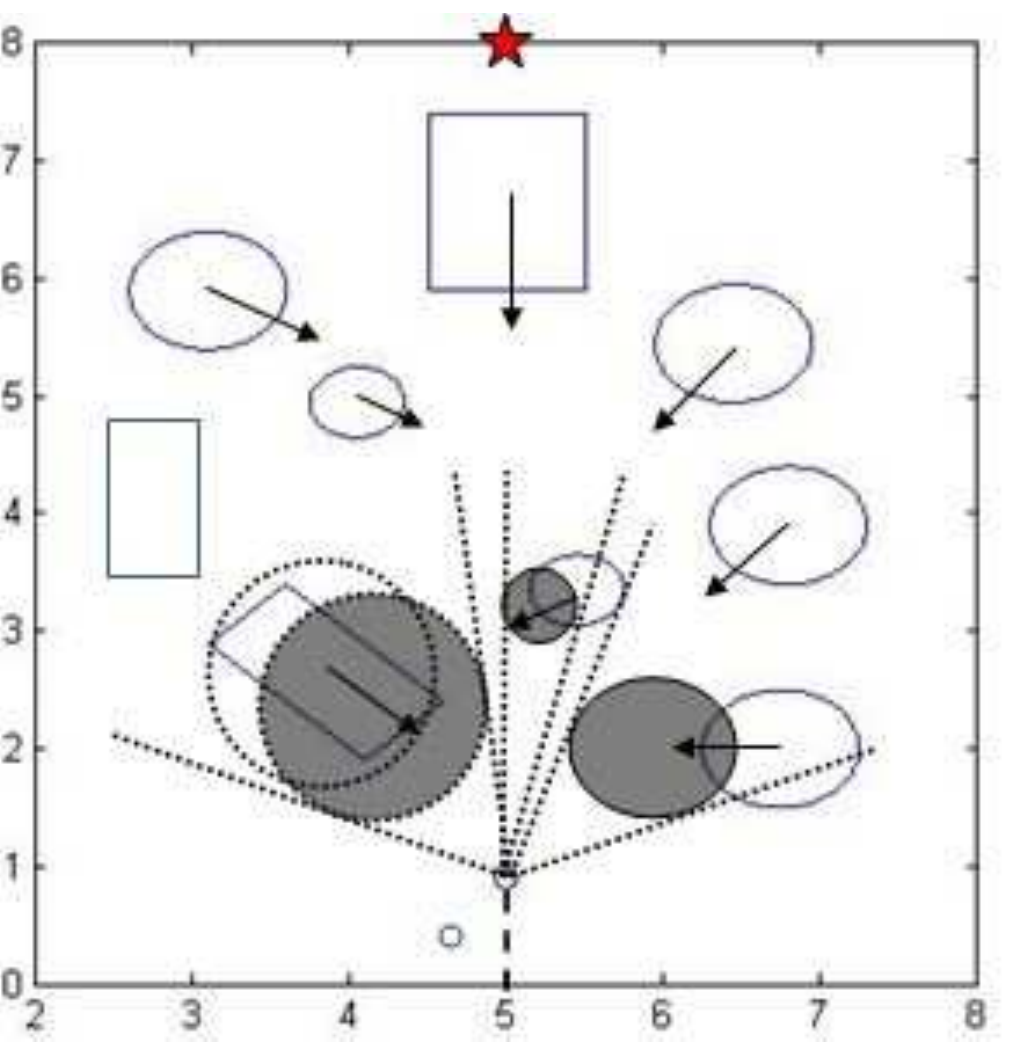}}
		\label{sim55}}
		\subfigure[]{\scalebox{0.42}{\includegraphics{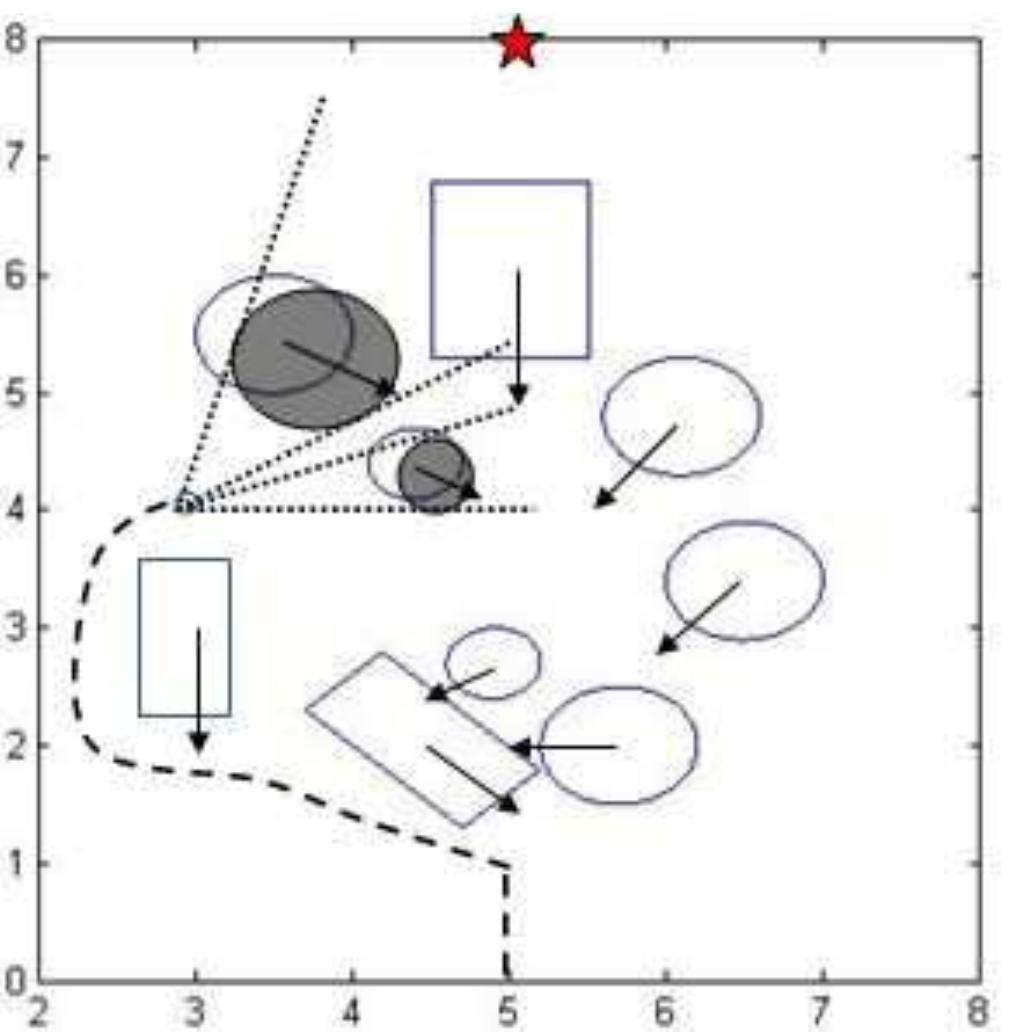}}
		\label{sim53}}
		\subfigure[]{\scalebox{0.42}{\includegraphics{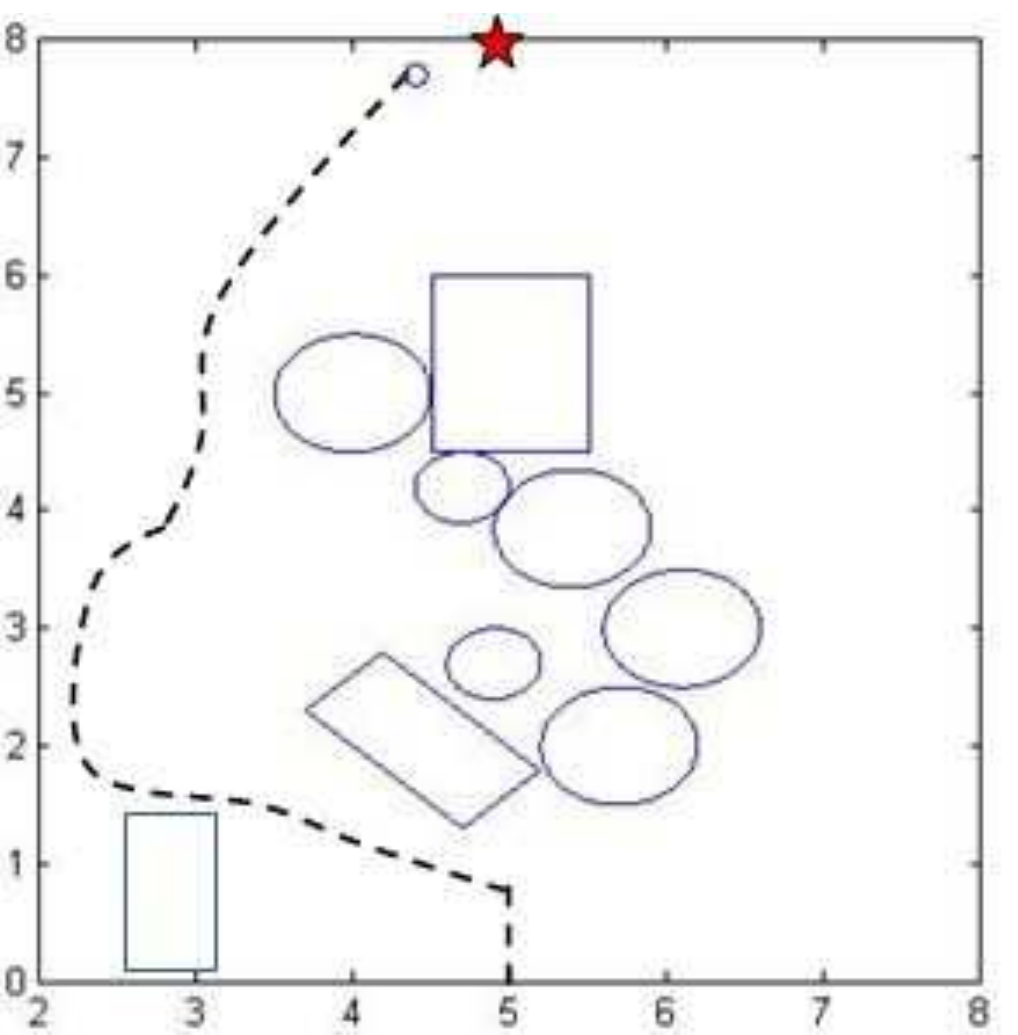}}
		\label{sim56}}
		\caption{Performance comparison 3: robot navigating in a crowded environment}
		\label{sim5}
		\end{figure} 
		\par

	\section {Experiments with Real Mobile Robot}

		The experiments with the proposed navigation algorithm are carried out with an ActiMedia Pioneer 3-DX wheeled mobile robot to demonstrate its applicability in real life scenarios.
\par
\begin{figure}[!h]
		\centering
		\includegraphics[width=5in]{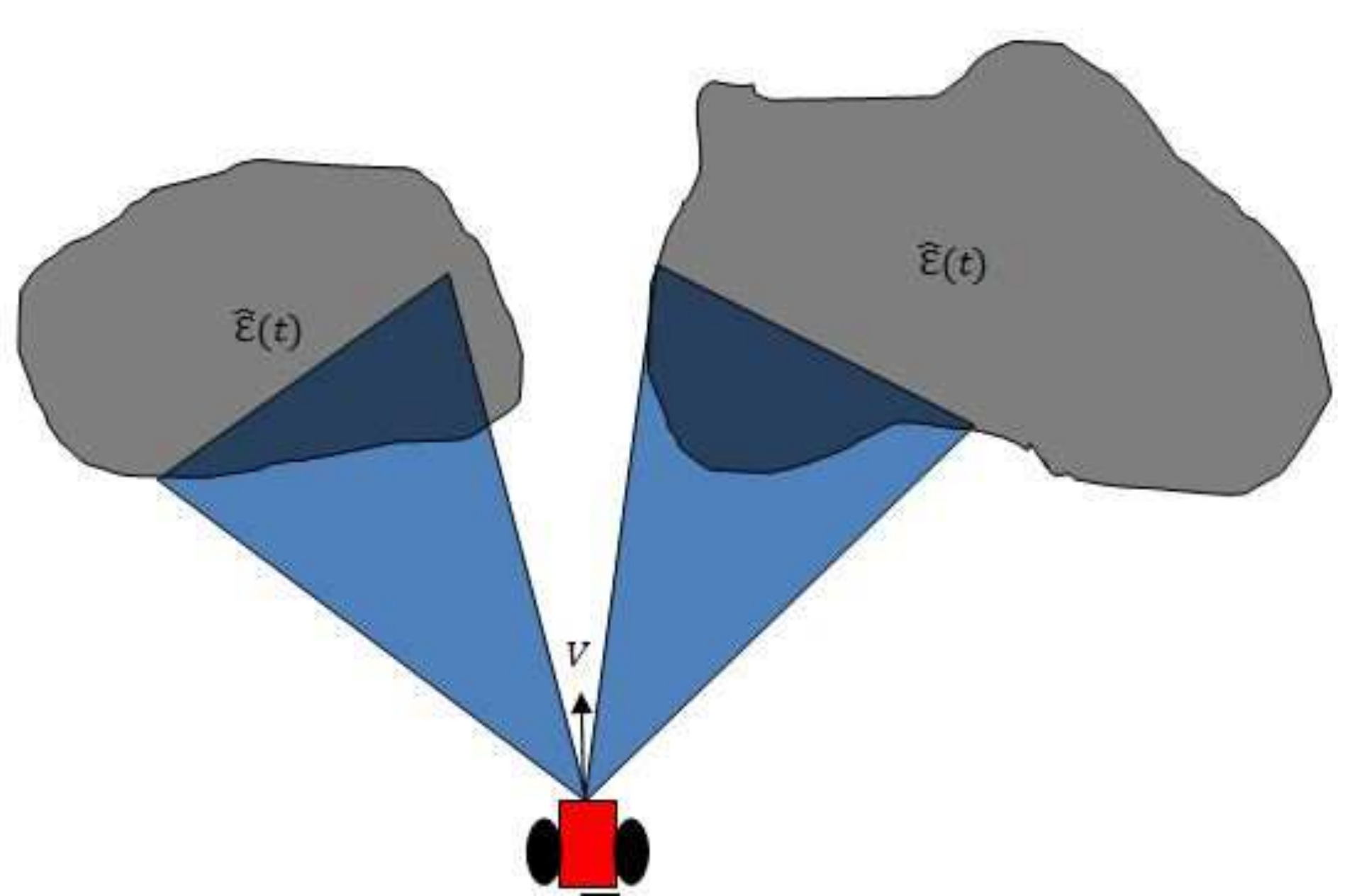}
		\caption{Illustration of obstacle detection}
		\label{c4.illu}
		\end{figure}

		The only information available to the proposed navigation algorithm is measured by SICK range finder and the ultrasonic sensors: the "dangerous" angle ranges at which the enlarged environments $\hat {\cal E} (t)$ cross the distance expressed at (\ref{Pdist}), see Fig.~\ref{c4.illu}. Therefore, the safe open intervals $(A_i^-, A_i^+)$ are found by excluding the "dangerous" directions. The safe interval which is closest to the robot's current heading $\theta (t)$ is found by (\ref{j}) and the desired avoiding direction (middle of vacancy) is calculated by (\ref{CC}). The movements of the robot is controlled by a notebook connected via WIFI connection.
\par

		In the following figures, the planned path of the robot is depicted as a red dashed line and that of the obstacles are shown as black dashed lines
\par
\begin{figure}[h]
		\centering
		\subfigure[]{\scalebox{0.35}{\includegraphics{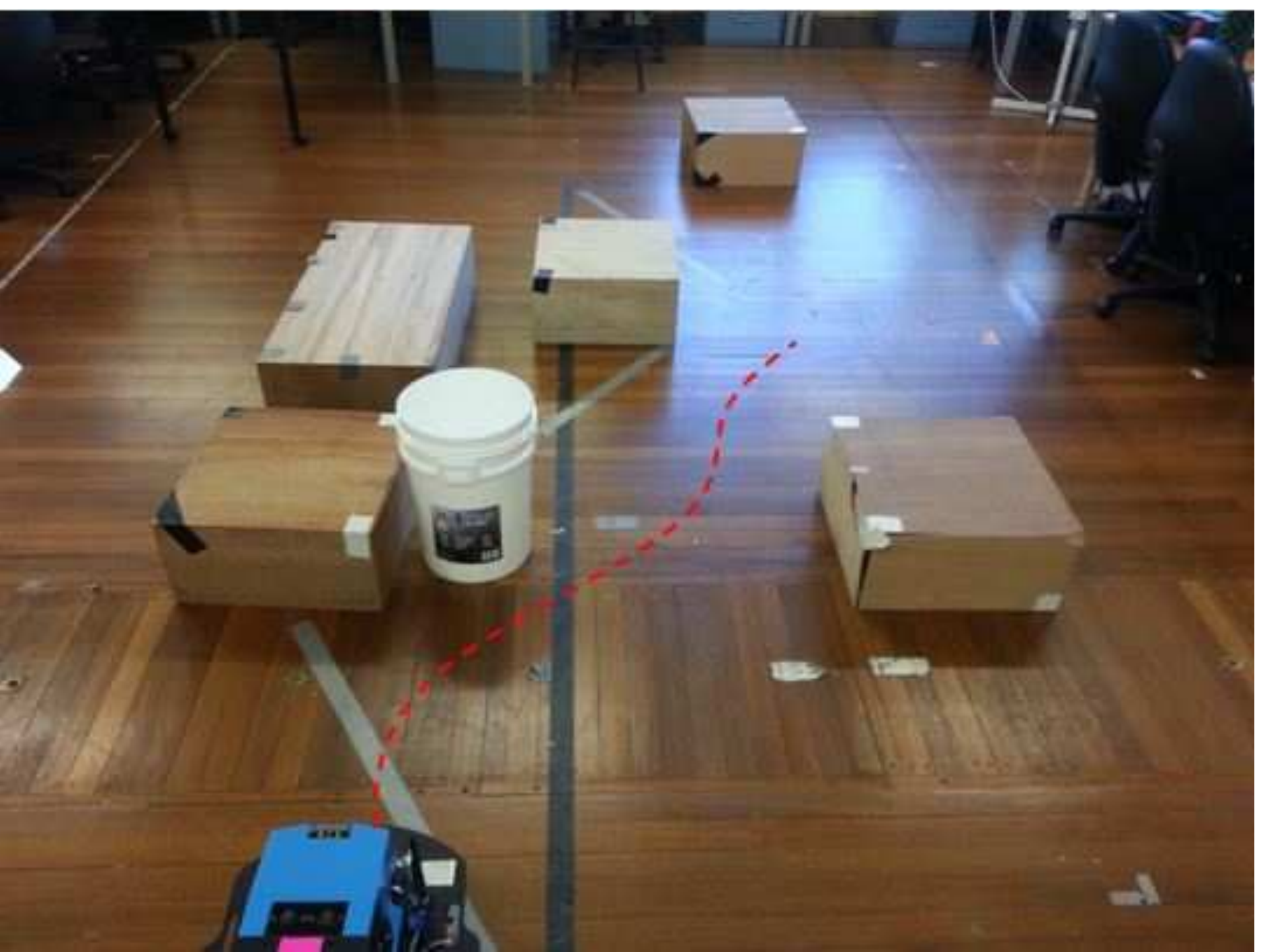}}
		\label{c4.exp11}}
		\subfigure[]{\scalebox{0.35}{\includegraphics{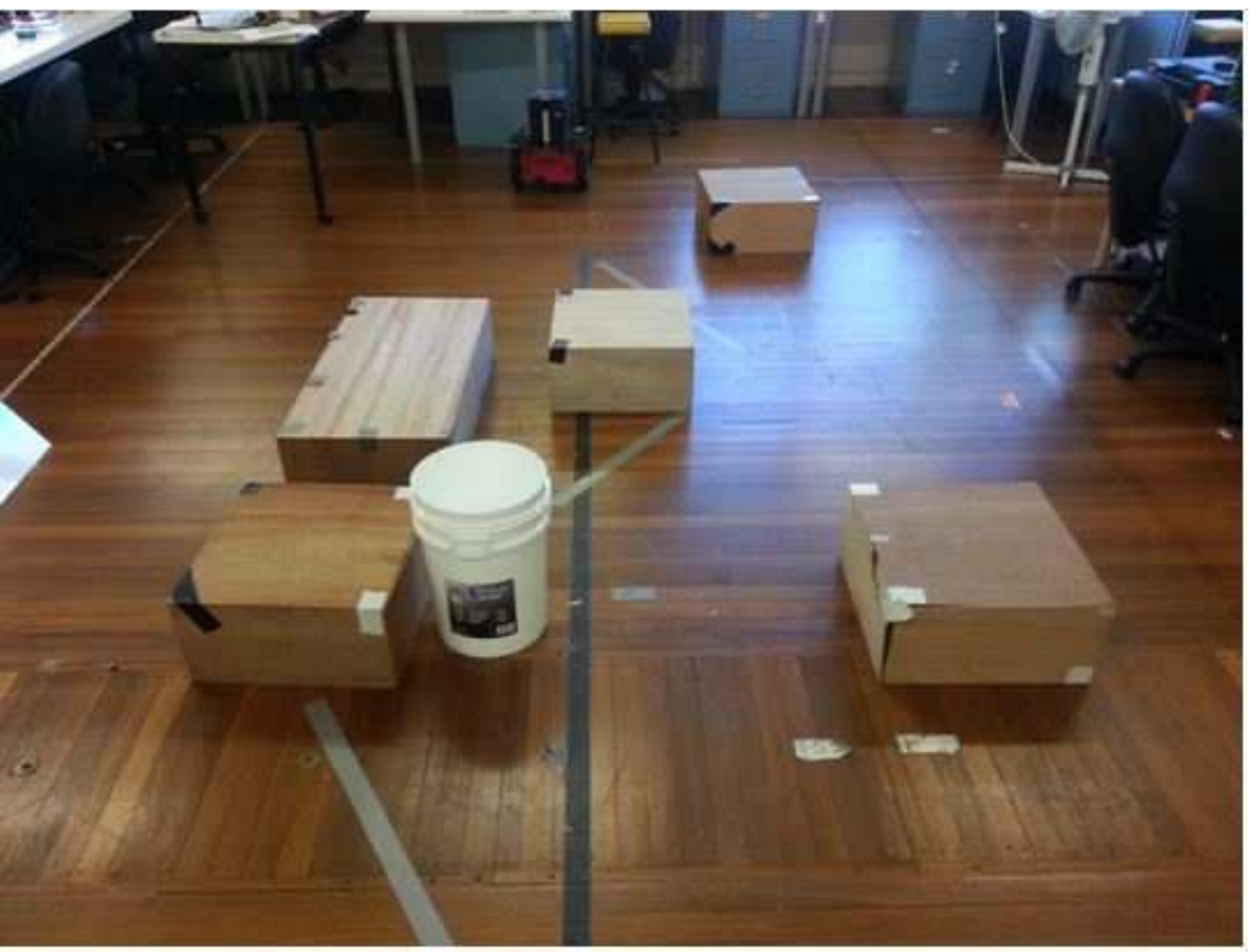}}
		\label{c4.exp12}}
		\subfigure[]{\scalebox{0.35}{\includegraphics{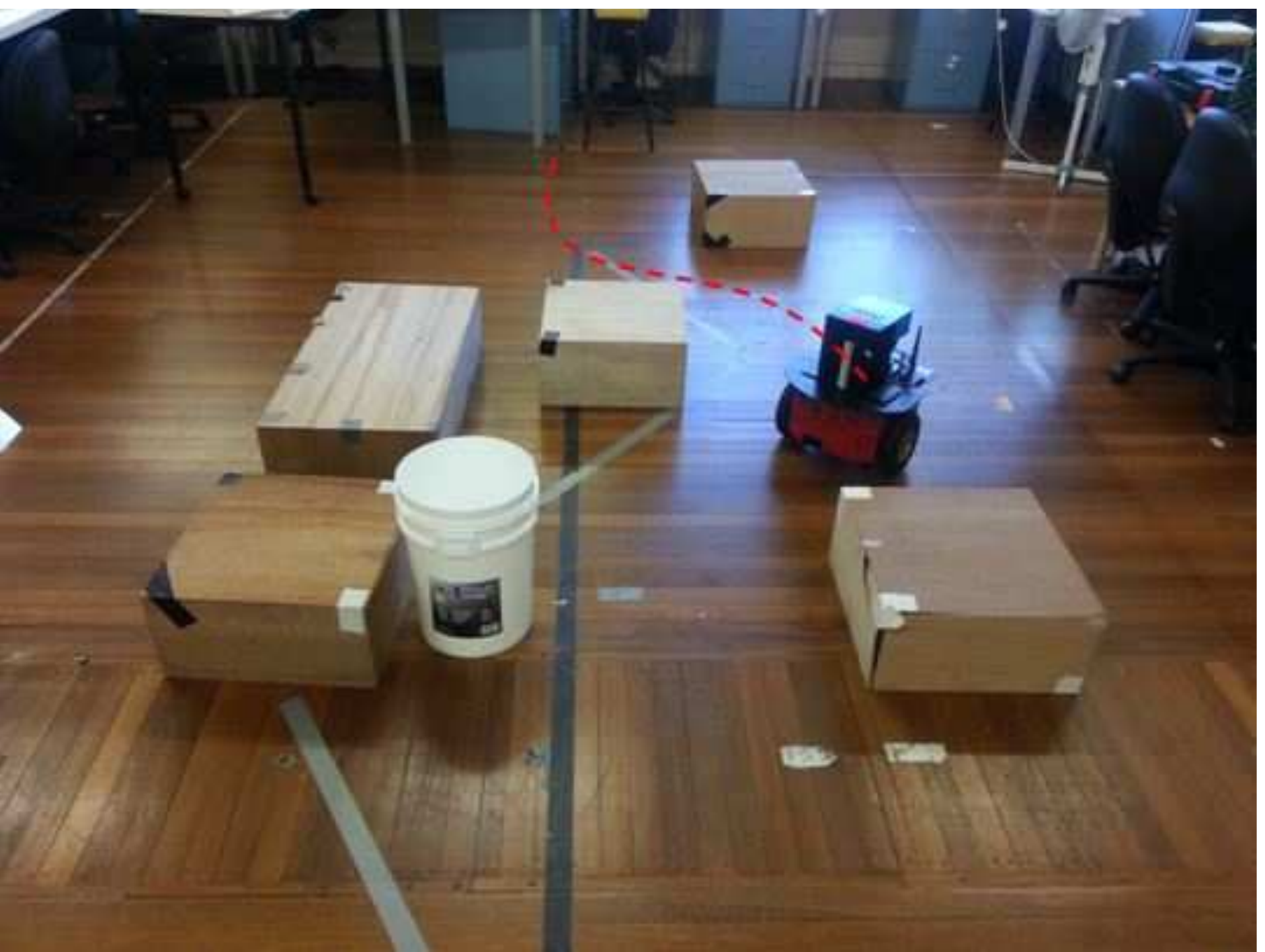}}
		\label{c4.exp13}}
		\subfigure[]{\scalebox{0.5}{\includegraphics{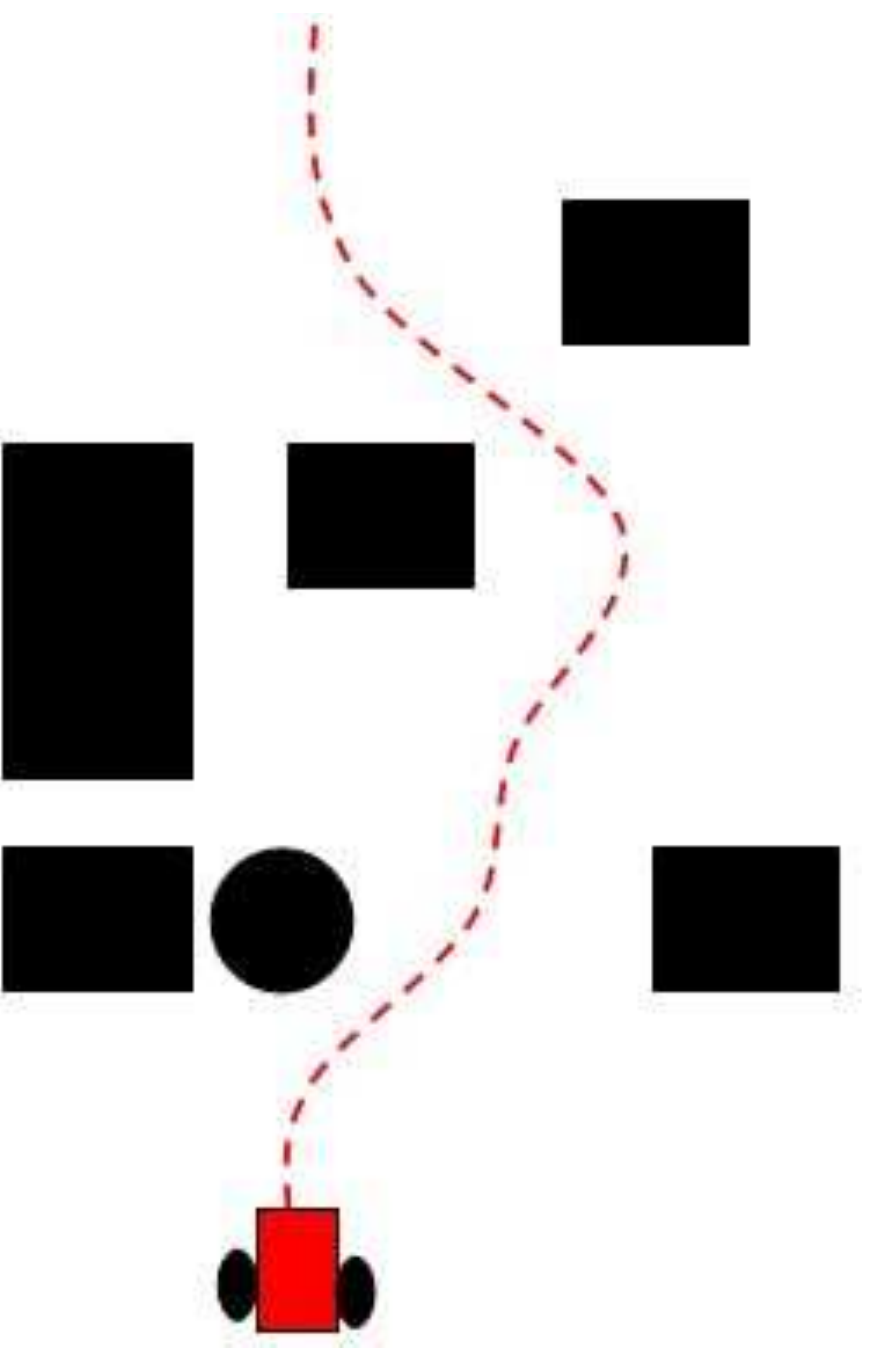}}
		\label{c4.exp14}}
		\caption{Robot avoids static obstacles}
		\label{c4_exp1}
		\end{figure}

		We start with a relatively simple scenario, where the robot is navigating in a static environments with a number of obstacles. The the obstacles are placed so that they block most of the passage, leaving limited safe interval for the robot, also the shapes of the obstacles are different from each other. The robot does not have access to any of this information in prior to the experiment. Fig.~\ref{c4.exp11}, Fig.~\ref{c4.exp12} and Fig.~\ref{c4.exp13} show the moments when the robot bypasses the obstacles through the safe intervals (middle of vacancy). and the overall path is depicted in Fig.~\ref{c4.exp14}.
\par
		\par
		
\begin{figure}[!h]
		\begin{minipage}{.5\textwidth}	
			\centering	
			\subfigure[]{\scalebox{0.3}{\includegraphics{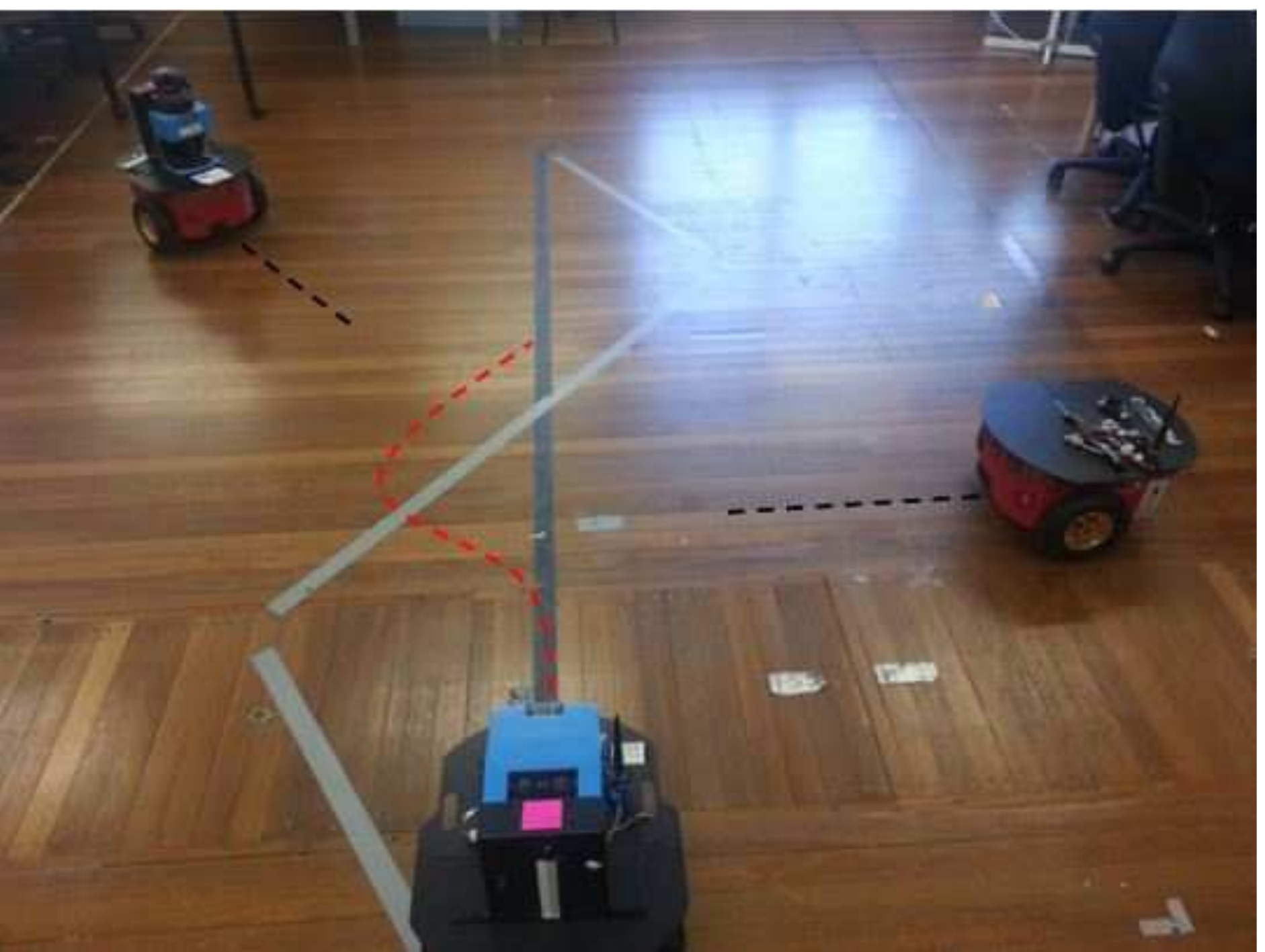}}
			\label{c4.exp21}}
			\subfigure[]{\scalebox{0.3}{\includegraphics{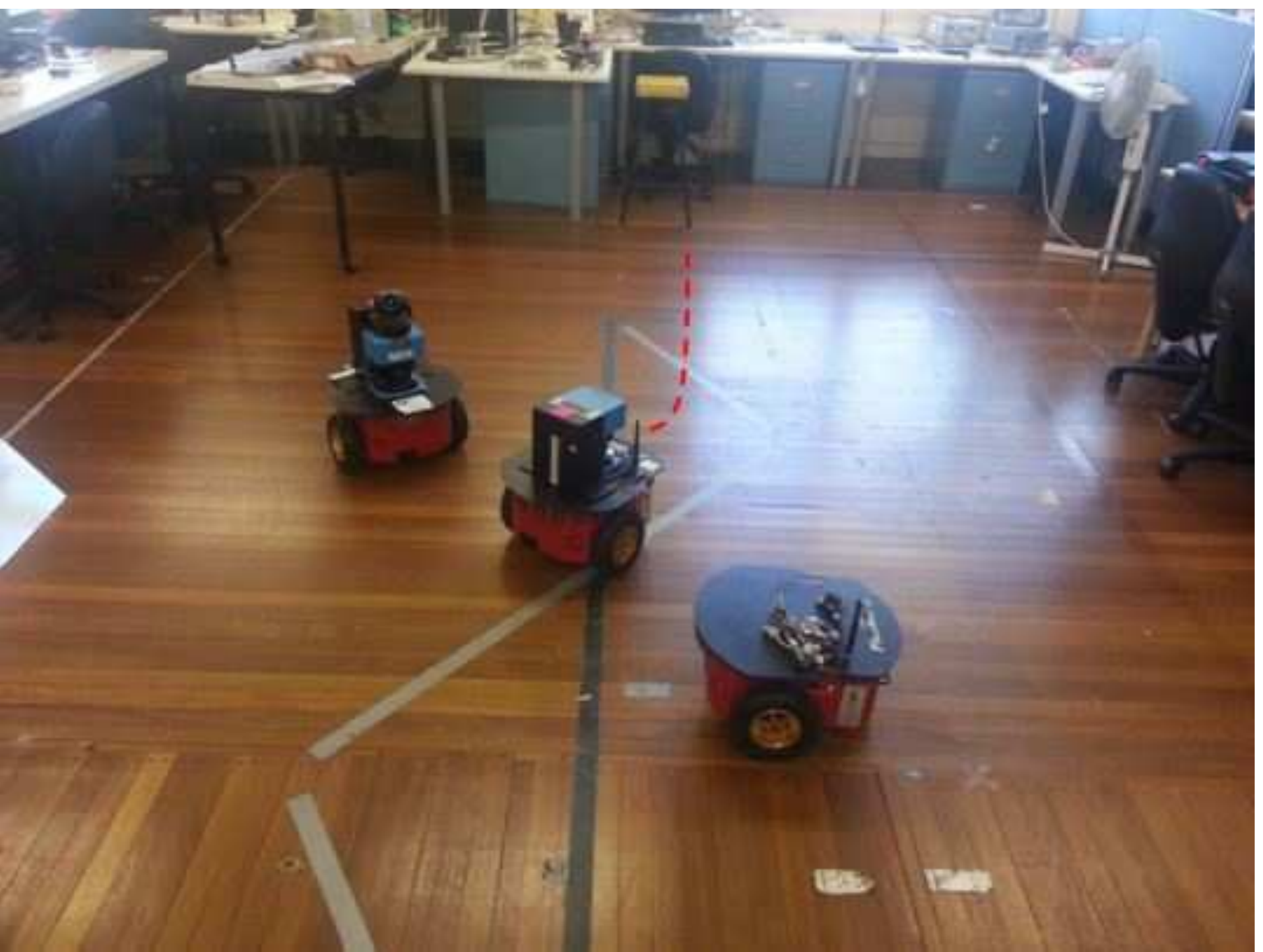}}
			\label{c4.exp22}}
			\subfigure[]{\scalebox{0.3}{\includegraphics{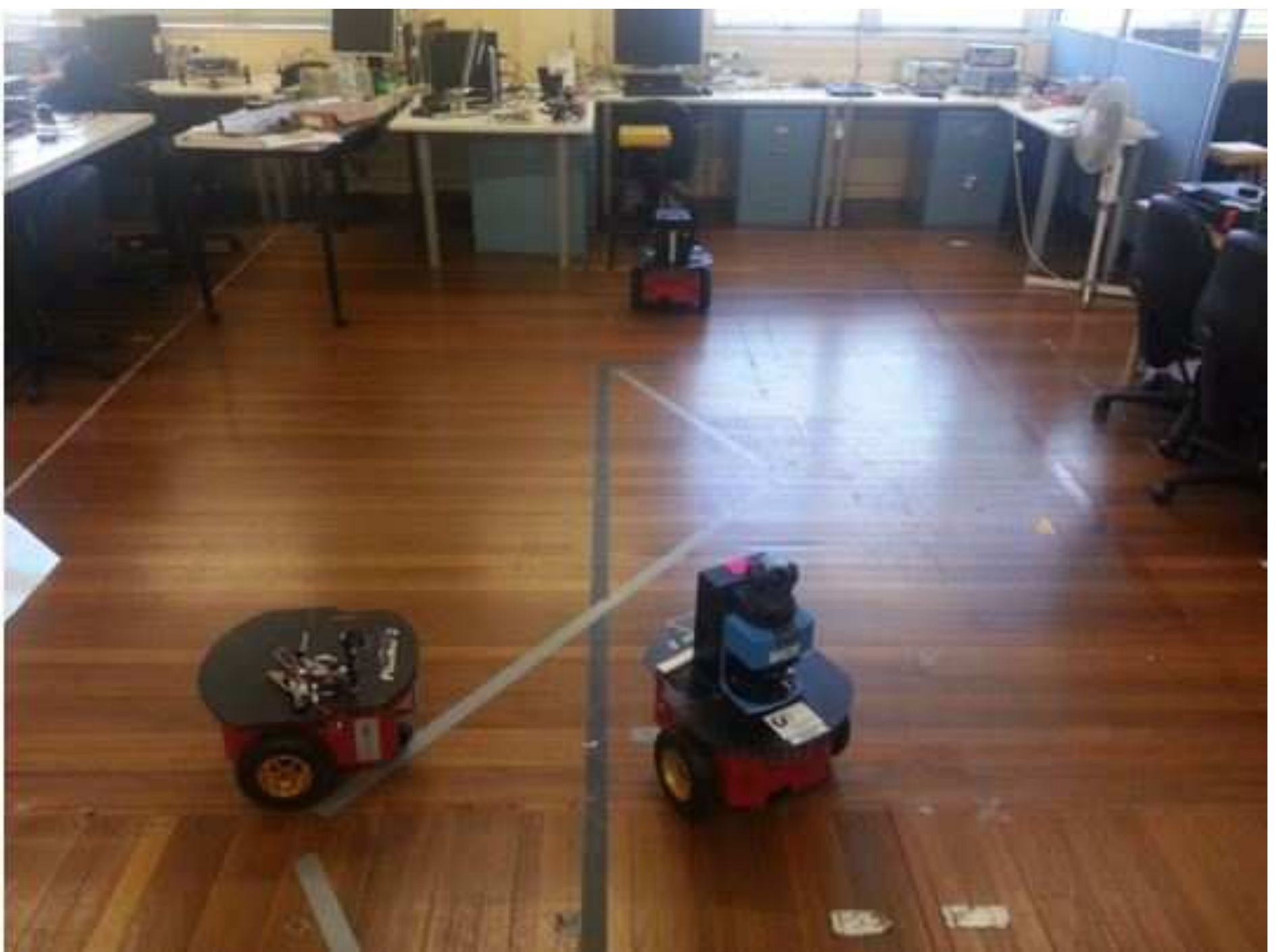}}
			\label{c4.exp23}}
		\end{minipage}
		\begin{minipage}{0.5\textwidth}
			\subfigure[]{\scalebox{0.6}{\includegraphics{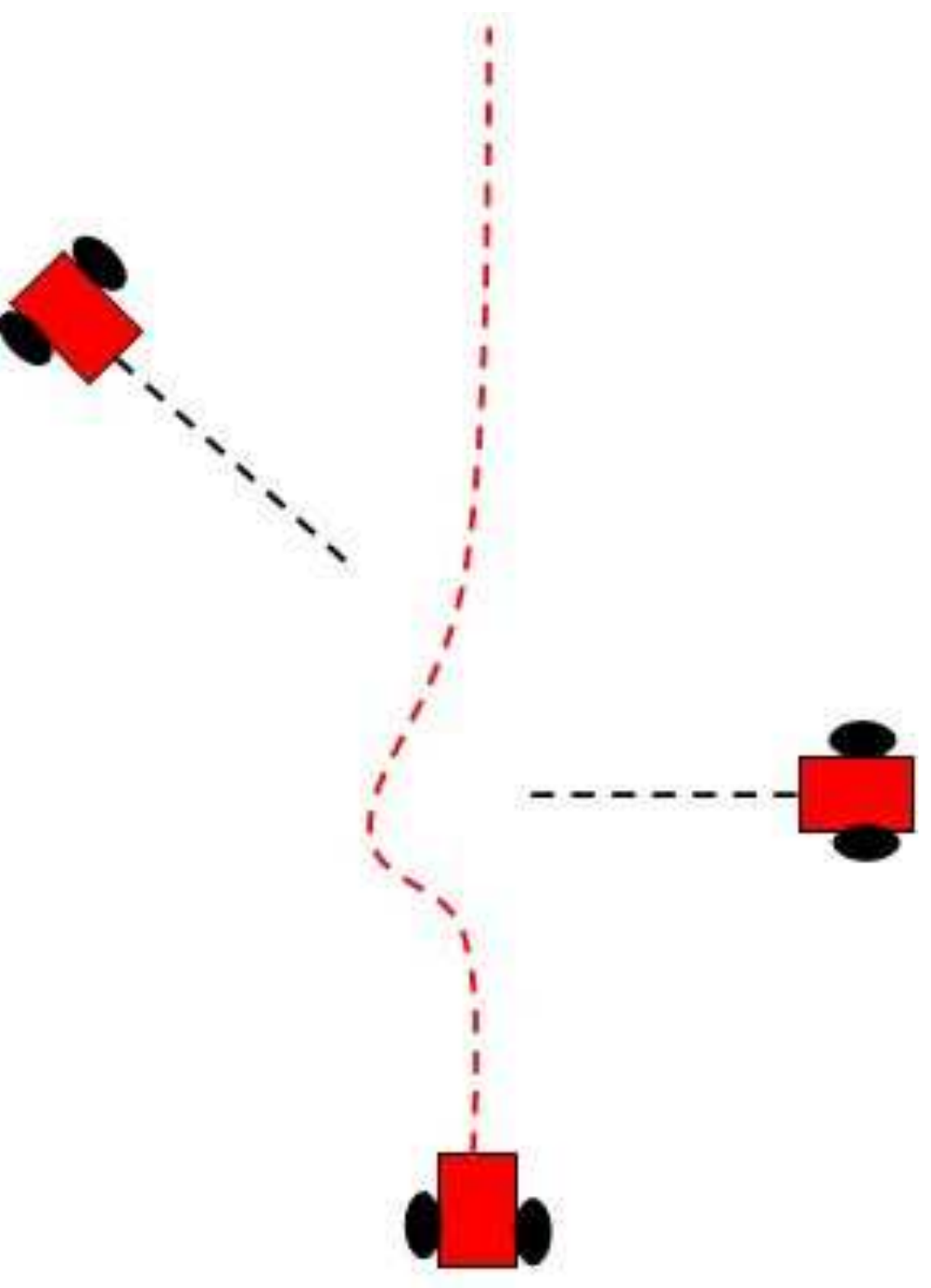}}
			\label{c4.exp24}}
		\end{minipage}
		\caption{Robot avoids obstacles in ambiguous scenario}
		\label{c4.exp2}
		\end{figure}

		The ambiguous situation which is frequently encountered in real life where two or more obstacles are considered as "dangerous" is again investigated in Fig.~\ref{c4.exp2}. In this experiment, two obstacle are moving towards the robot so that the robot finds itself trapped between the obstacles. Unlike many of the existing navigation approaches only consider the "closest" or "most dangerous" obstacle, which likely to result in a longer avoiding path (go around the obstacles). The proposed navigation algorithm guides the robot through the vacancy between the obstacles when the robot senses that the space between the obstacle is large enough for it to bypass. The overall path taken by the robot is shown in Fig.~\ref{c4.exp24} which is very efficient.
\par

\begin{figure}[!h]
		\begin{minipage}{.5\textwidth}	
			\centering	
			\subfigure[]{\scalebox{0.3}{\includegraphics{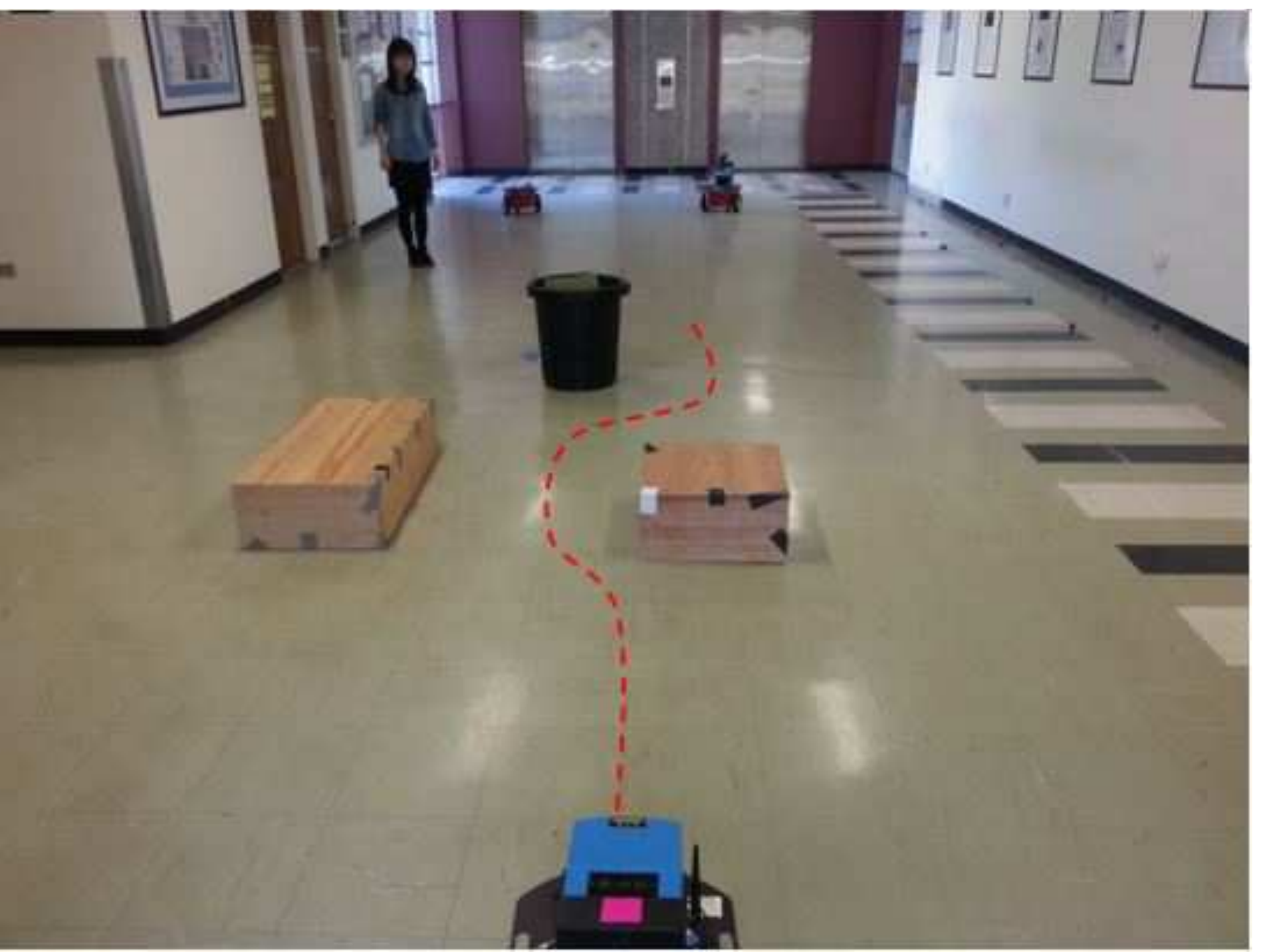}}
			\label{c4.exp31}}
			\subfigure[]{\scalebox{0.3}{\includegraphics{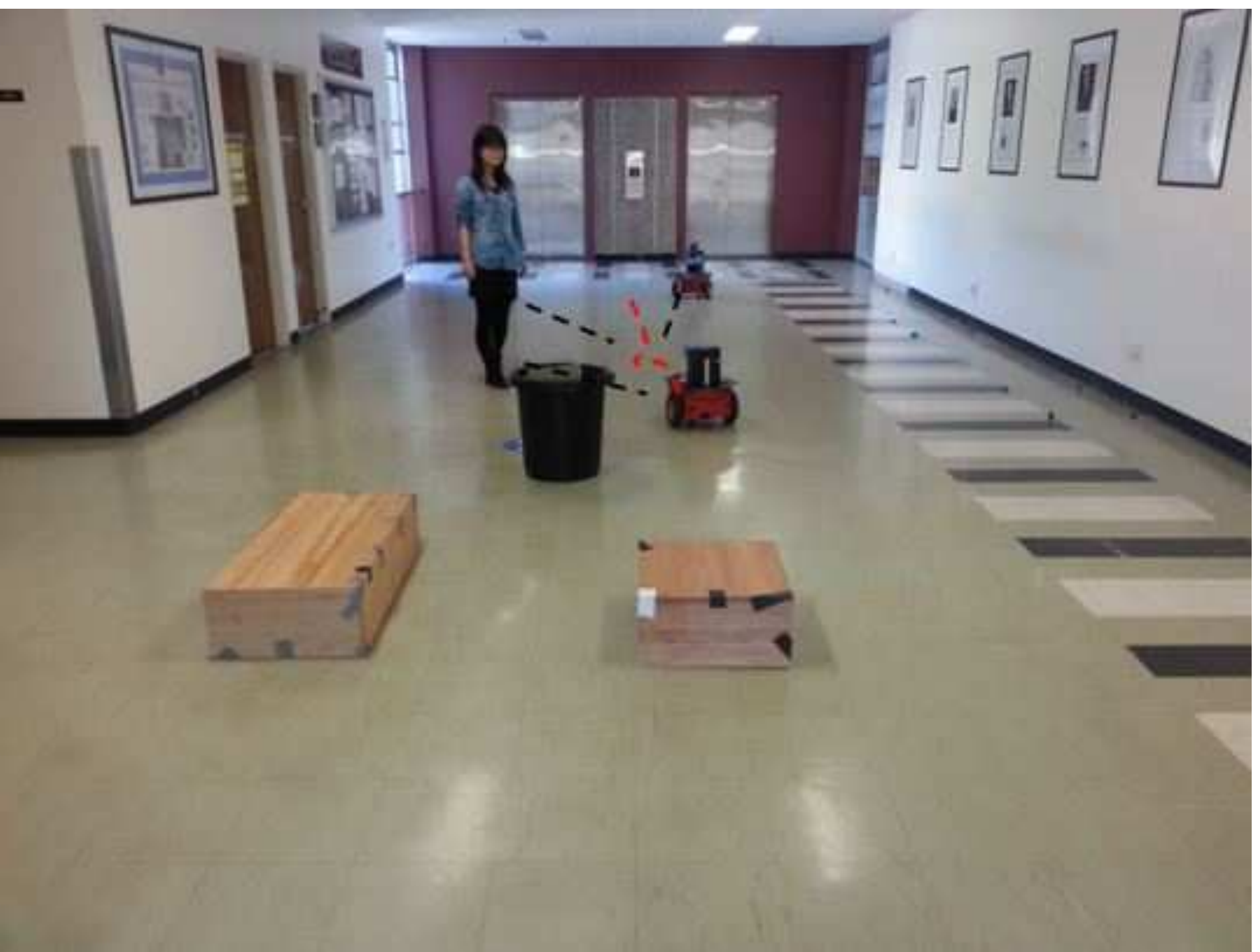}}
			\label{c4.exp32}}
			\subfigure[]{\scalebox{0.3}{\includegraphics{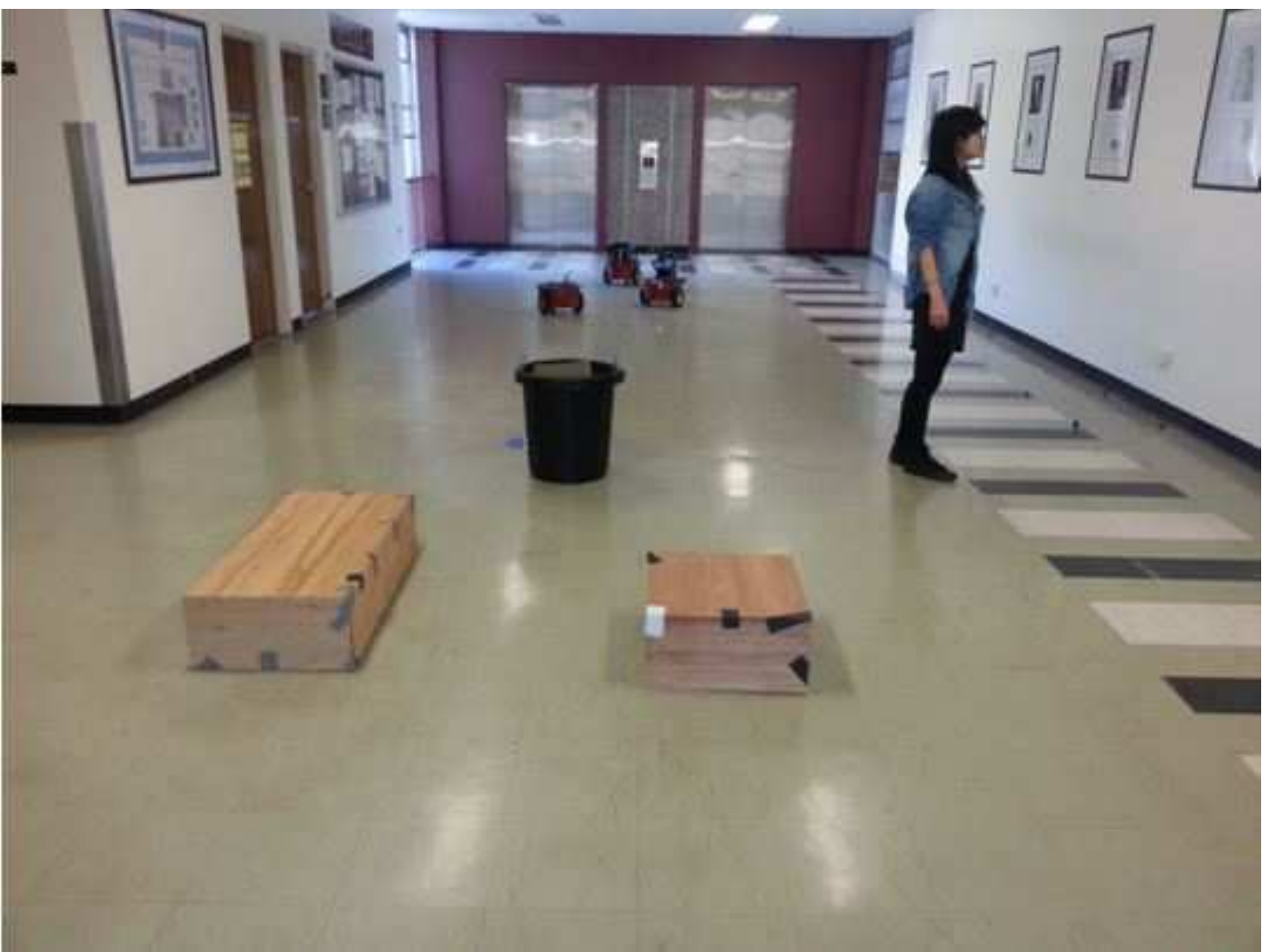}}
			\label{c4.exp33}}
		\end{minipage}
		\begin{minipage}{0.5\textwidth}
			\subfigure[]{\scalebox{0.6}{\includegraphics{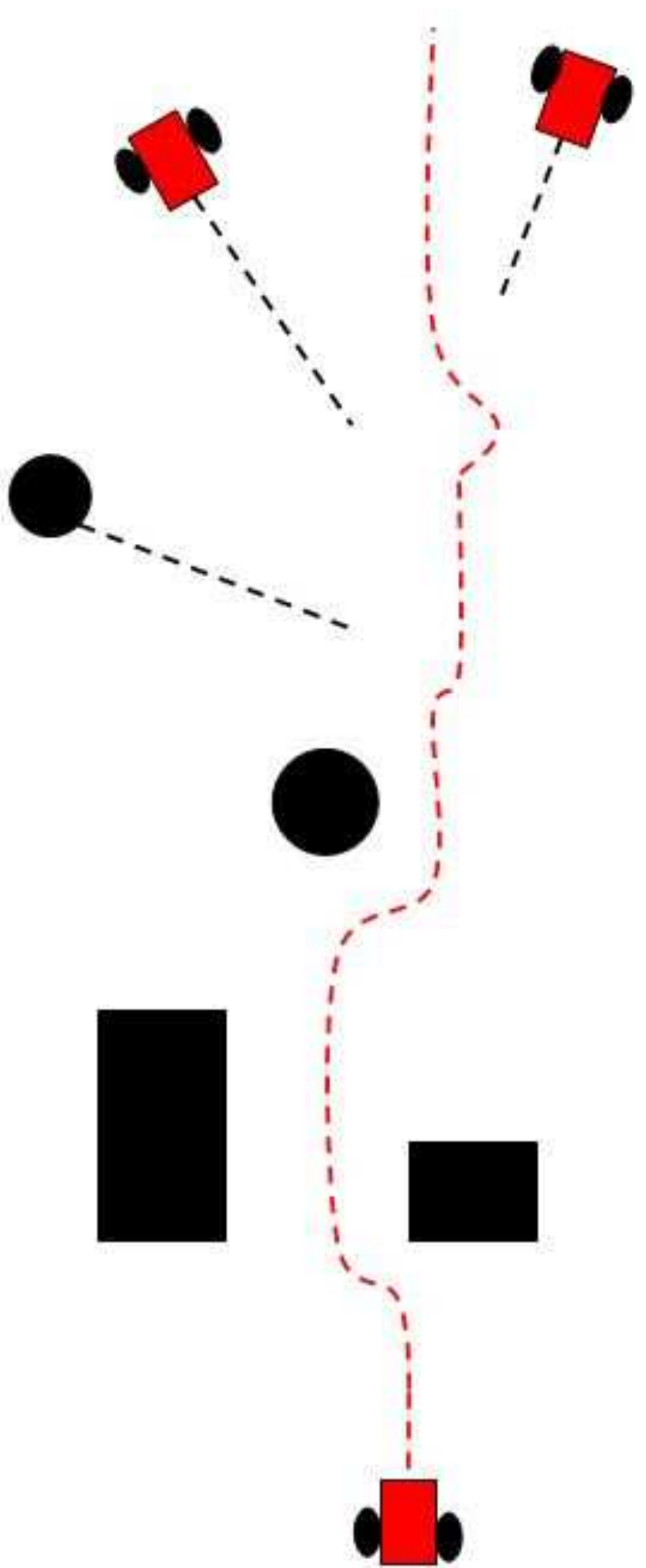}}
			\label{c4.exp34}}
		\end{minipage}
		\caption{Robot navigating in complicated  unknown environment with stationary and dynamic obstacles case 1}
		\label{c4.exp3}
		\end{figure}

\begin{figure}[!h]
		\begin{minipage}{.5\textwidth}	
			\centering	
			\subfigure[]{\scalebox{0.3}{\includegraphics{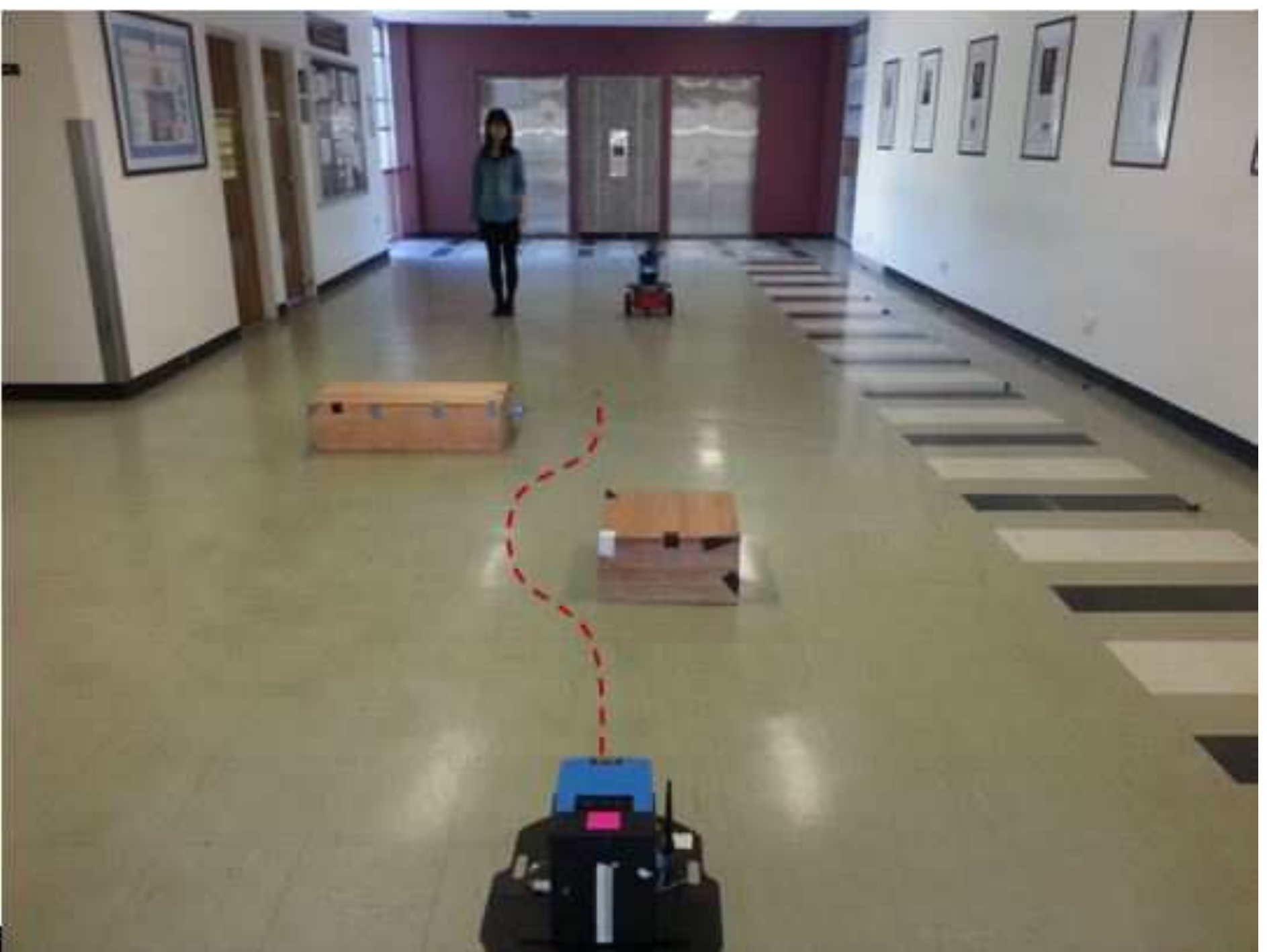}}
			\label{c4.exp41}}
			\subfigure[]{\scalebox{0.3}{\includegraphics{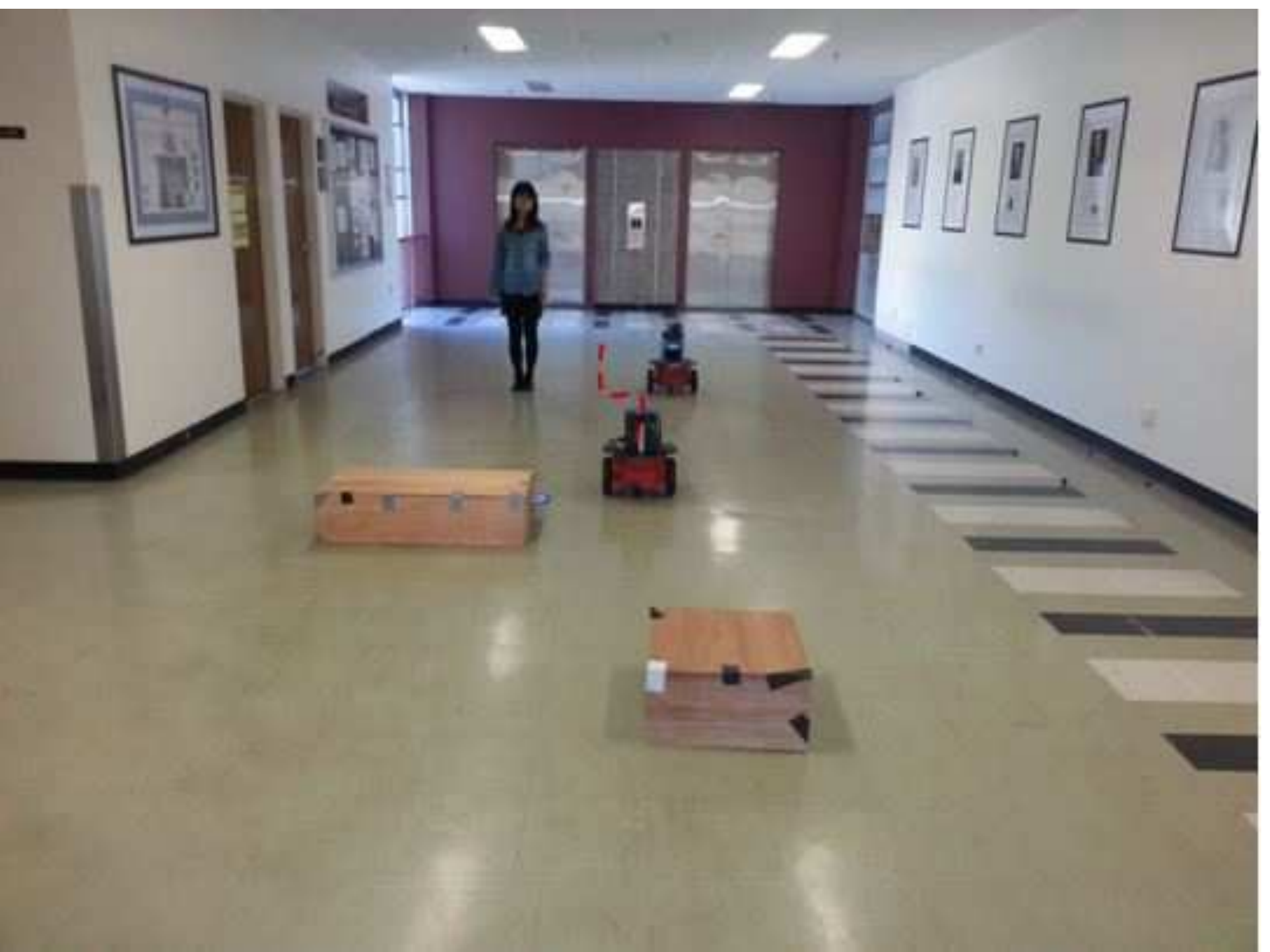}}
			\label{c4.exp42}}
			\subfigure[]{\scalebox{0.3}{\includegraphics{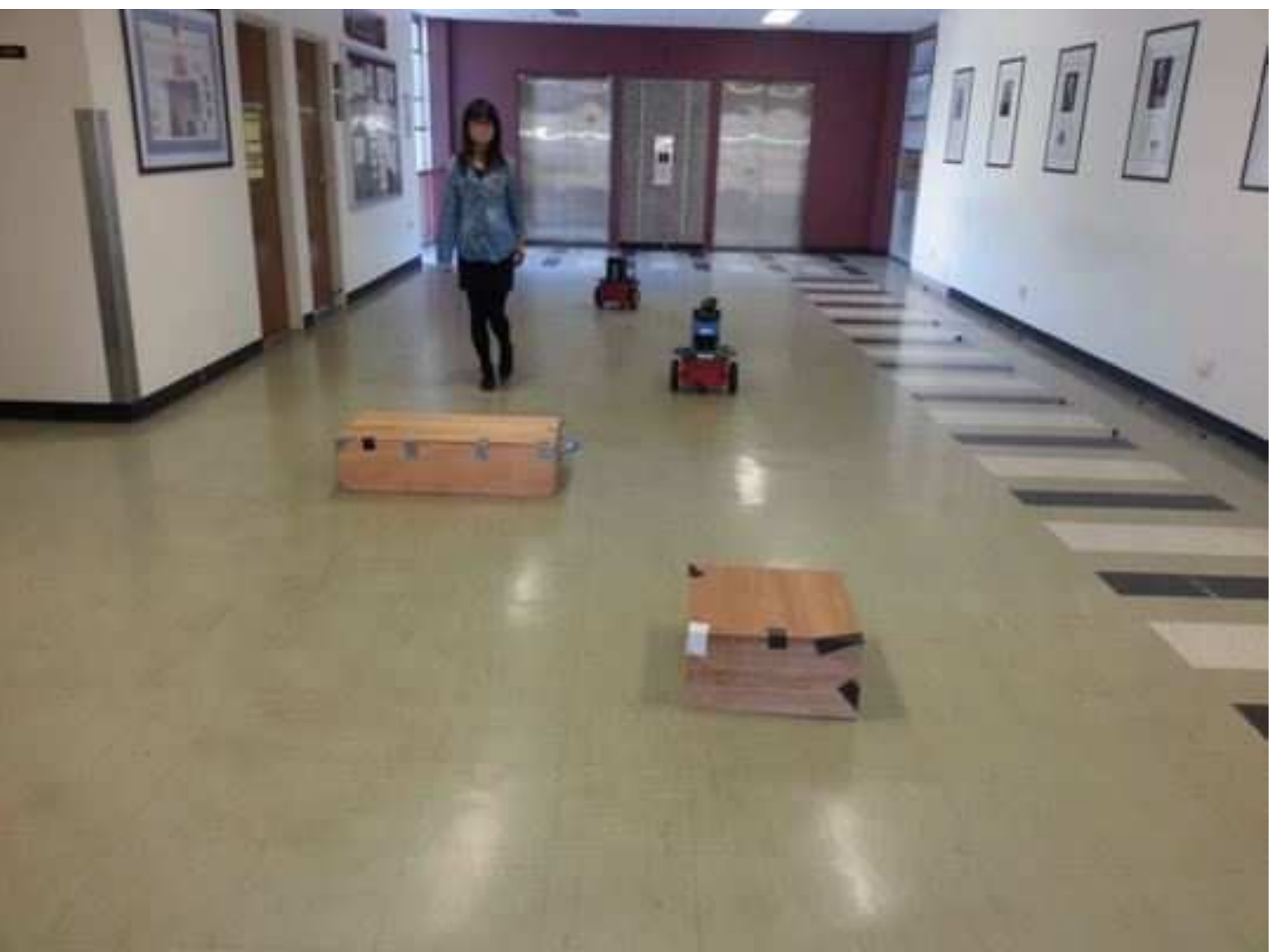}}
			\label{c4.exp43}}
		\end{minipage}
		\begin{minipage}{0.5\textwidth}
			\subfigure[]{\scalebox{0.6}{\includegraphics{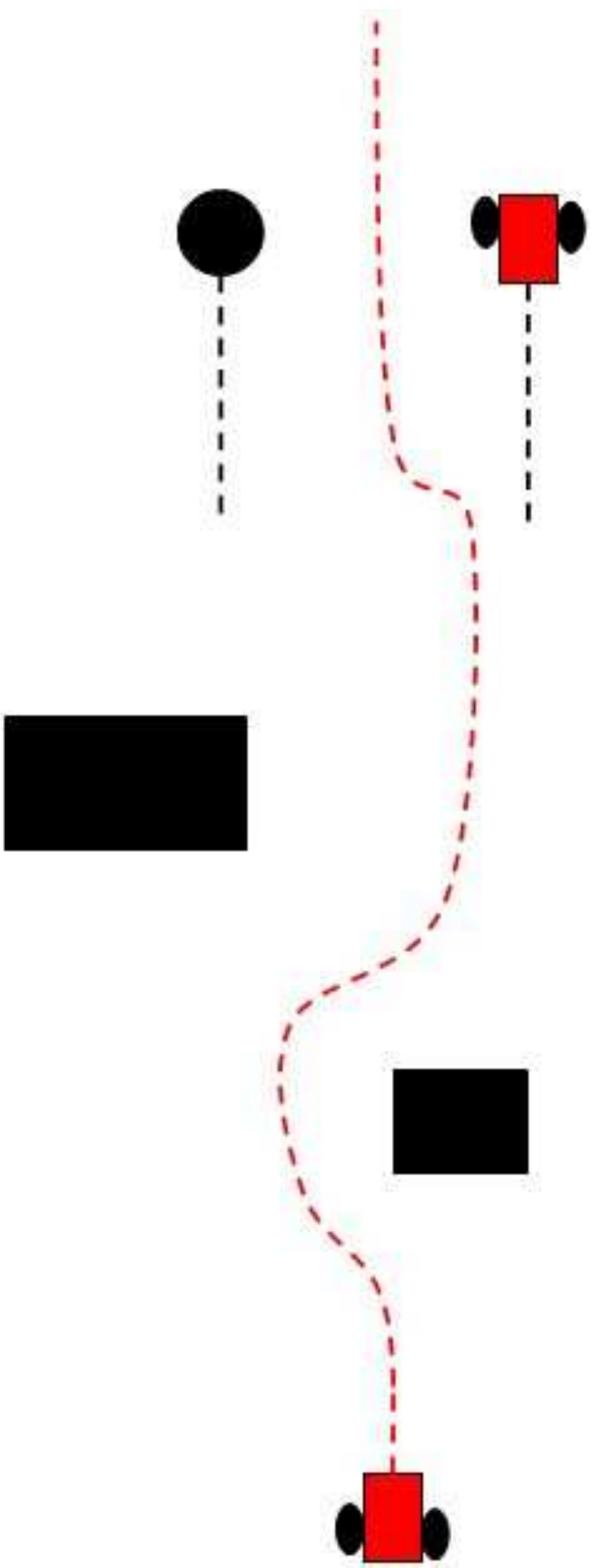}}
			\label{c4.exp44}}
		\end{minipage}
		\caption{Robot navigating in complicated  unknown environment with stationary and dynamic obstacles case 2}
		\label{c4.exp4}
		\end{figure}

\begin{figure}[!h]
		\begin{minipage}{.5\textwidth}	
			\centering	
			\subfigure[]{\scalebox{0.3}{\includegraphics{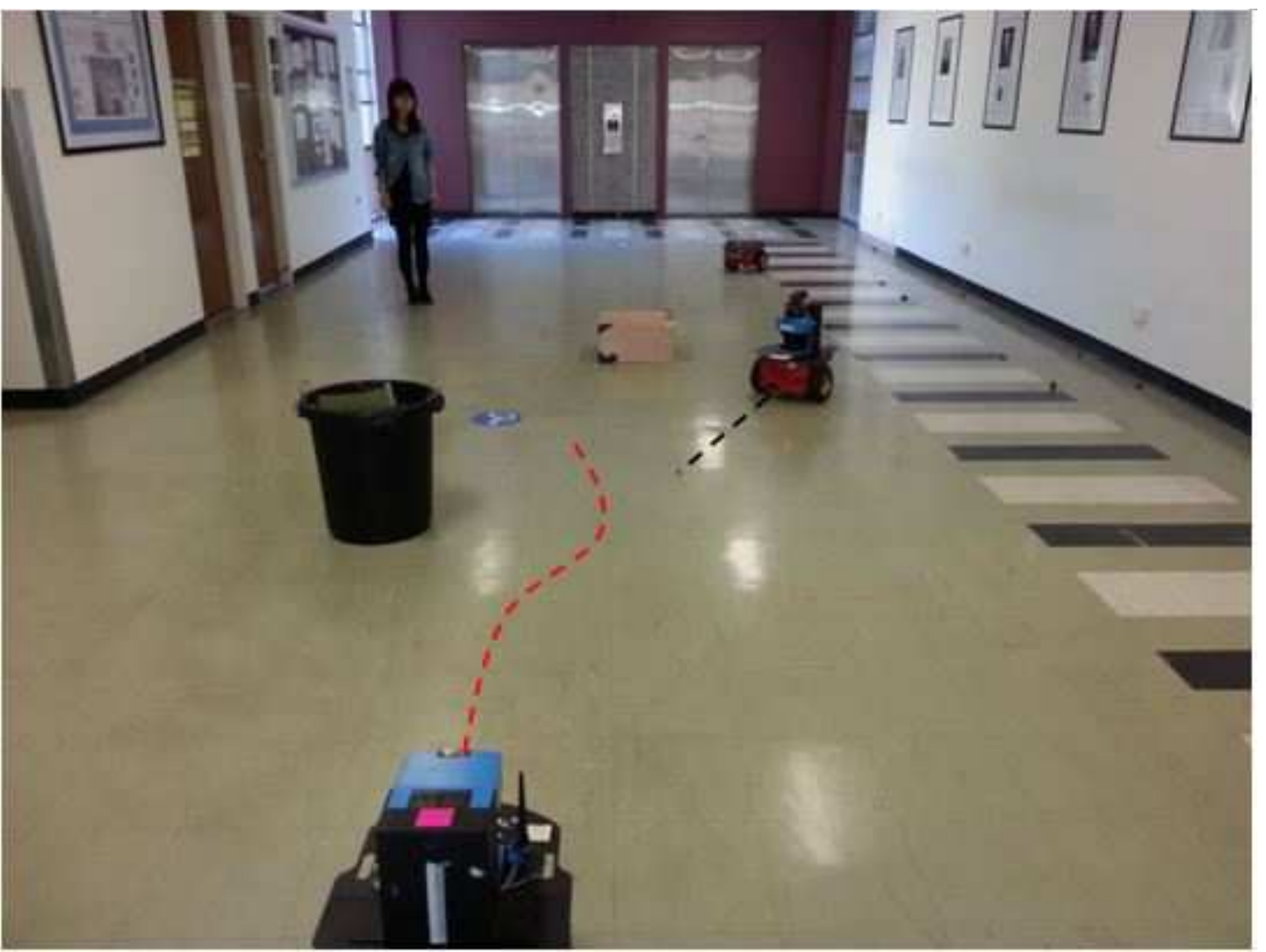}}
			\label{c4.exp51}}
			\subfigure[]{\scalebox{0.3}{\includegraphics{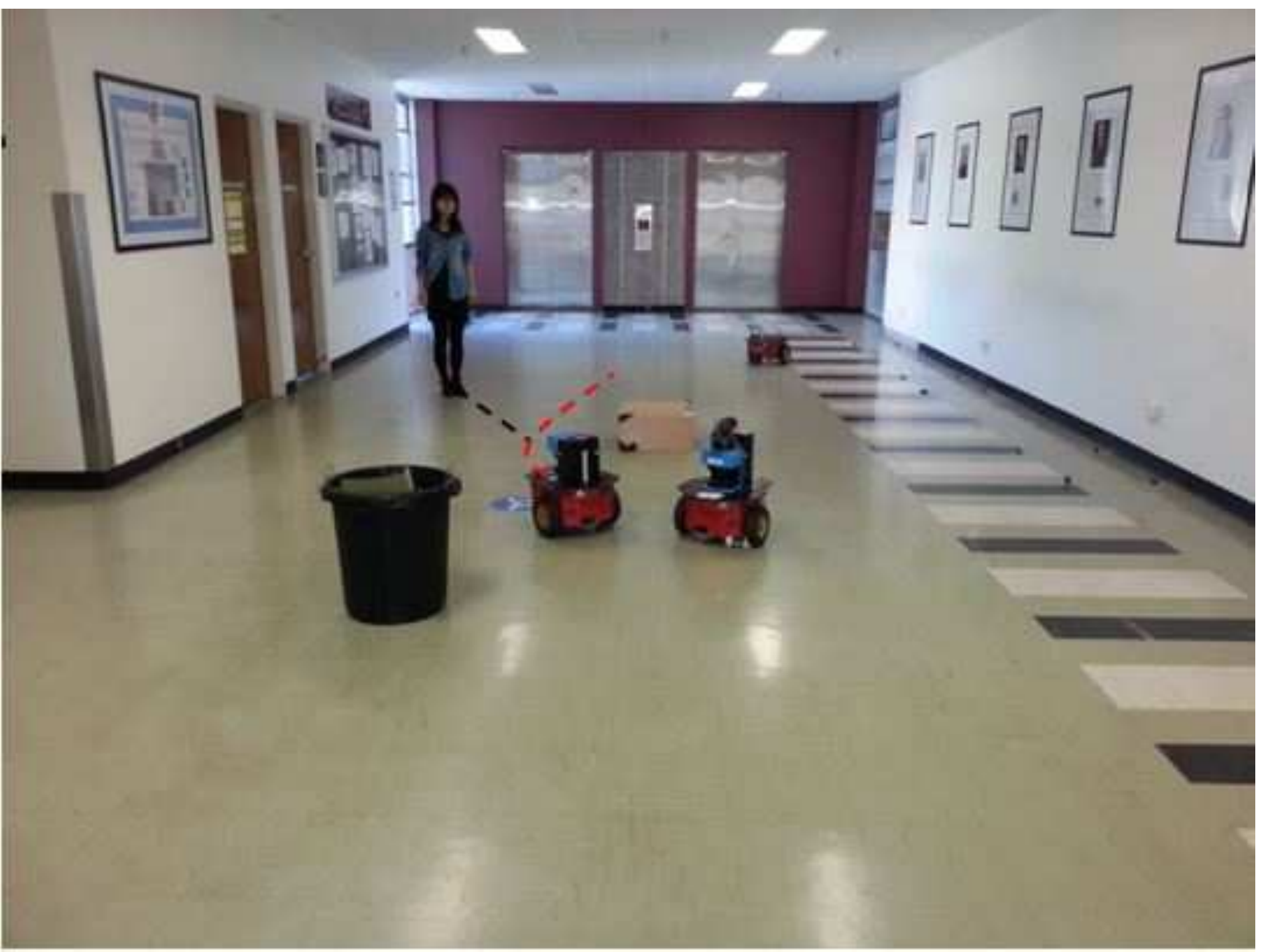}}
			\label{c4.exp52}}
			\subfigure[]{\scalebox{0.3}{\includegraphics{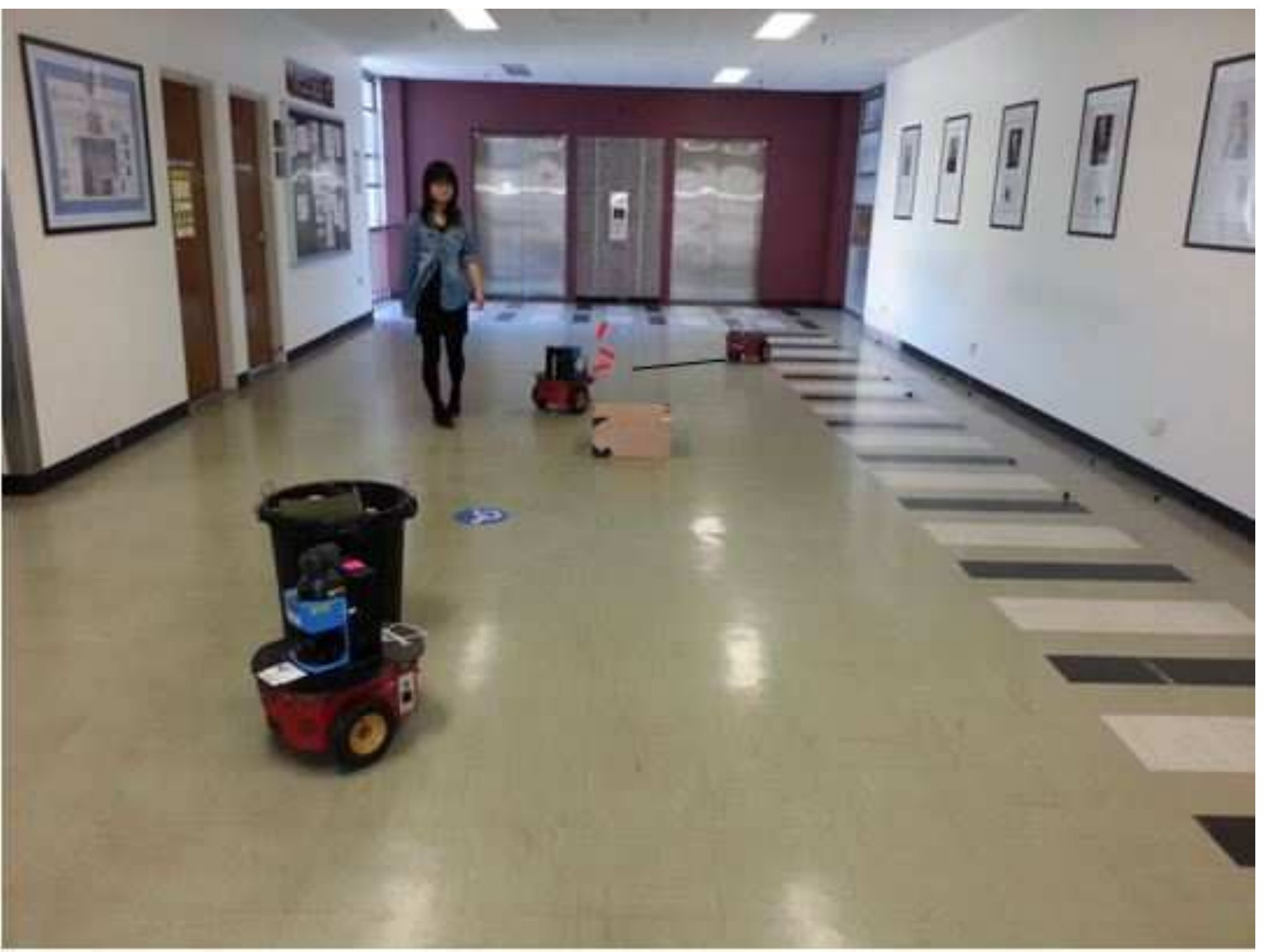}}
			\label{c4.exp53}}
		\end{minipage}
		\begin{minipage}{0.5\textwidth}
			\subfigure[]{\scalebox{0.6}{\includegraphics{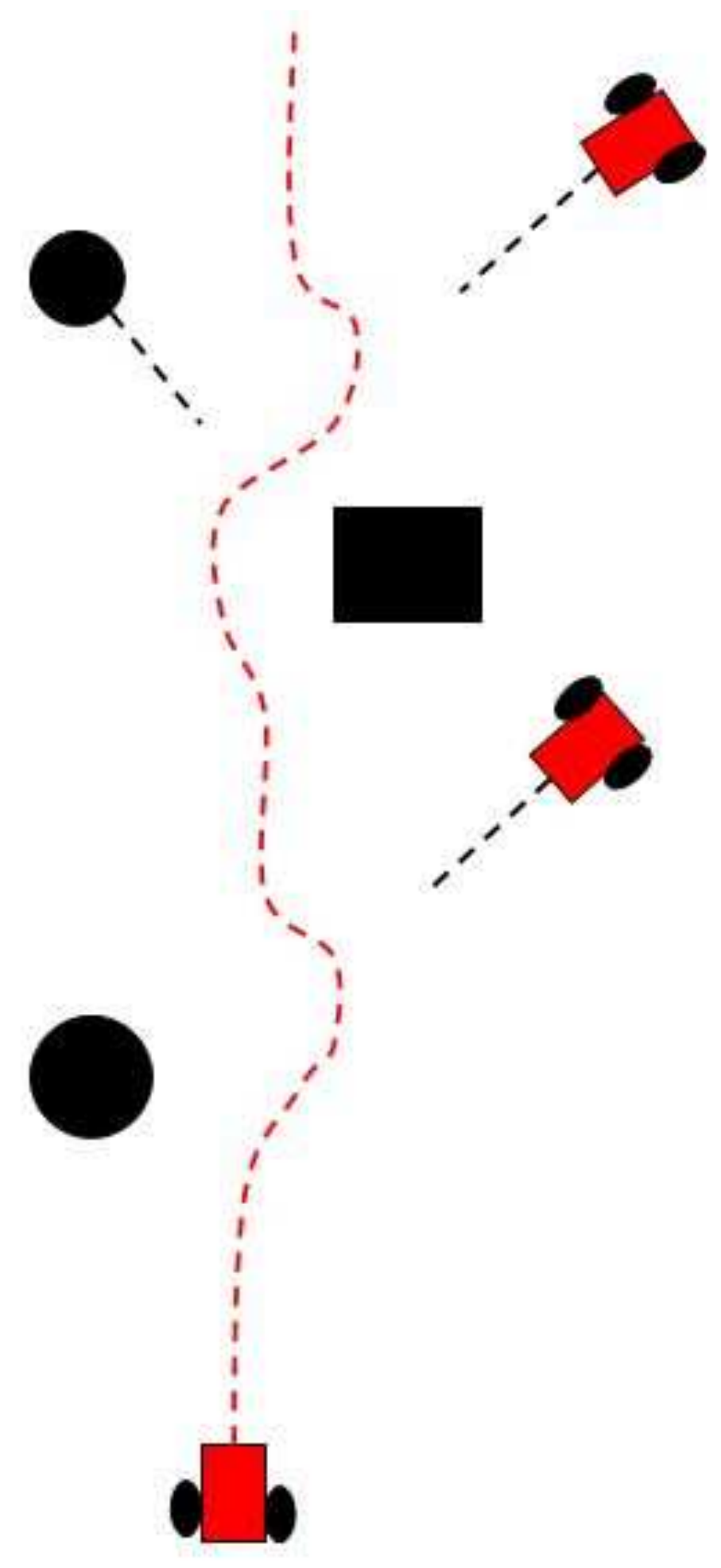}}
			\label{c1.exp54}}
		\end{minipage}
		\caption{Robot navigating in complicated  unknown environment with stationary and dynamic obstacles case 3}
		\label{c4.exp5}
		\end{figure}

		The performance of the proposed navigation algorithm in a complicated unknown dynamic environment with both stationary and dynamic obstacles is examined in Fig.~\ref{c4.exp3}.  The avoidance of dynamic obstacles is more difficult than that of stationary obstacles, especially when the information available to the robot is very limited. In this experiment, the robot use only available information which is the binary function $M(\alpha, t)$ to avoid the obstacles, see Fig.~\ref{c4.exp31}, Fig.~\ref{c4.exp32} and Fig.~\ref{c4.exp33} and achieve the desired heading $\theta_0$ as seen in Fig.~\ref{c4.exp34}. More complicated scenarios of navigating a robot in unknown environments are shown in Fig.~\ref{c4.exp4} and Fig.~\ref{c4.exp5}

	\section{Summary}

		In this chapter,  we propose a navigation algorithm for non-holonomic mobile robot in unknown dynamic environment using integrate environment representation. The proposed navigation is efficient in many scenarios which other algorithms are found difficult or impossible to solve. In particular Unlike many of the existing navigation algorithm which focus on avoiding the "closest" or "most dangerous" obstacle, the proposed navigation seek a safe path through the crowd of obstacles, which is proven to be efficient in many scenarios. The extensive computer simulations show the efficiency of the proposed navigation algorithm and the applicability in real life scenario are demonstrated by experiments with P3 mobile robot.
\chapter {Comparison Between the Navigation Algorithms} \label{C5}

	In this chapter, we compare the navigation algorithms, proposed in Chapter~\ref{C2},~\ref{C3},~\ref{C4} respectively, in various aspects. The purpose of this comparison is to identify the merits of each of the proposed navigation algorithm. It also provides more insights to the proposed navigation algorithms, for example, one algorithm may be more implementation cost efficient than the others for a certain project or one may be more efficient when avoiding obstacles in particular scenarios.
\par

	\section {Essential Measurements}

\begin{figure}[!h]
		\centering
		\subfigure[]{\scalebox{0.50}{\includegraphics{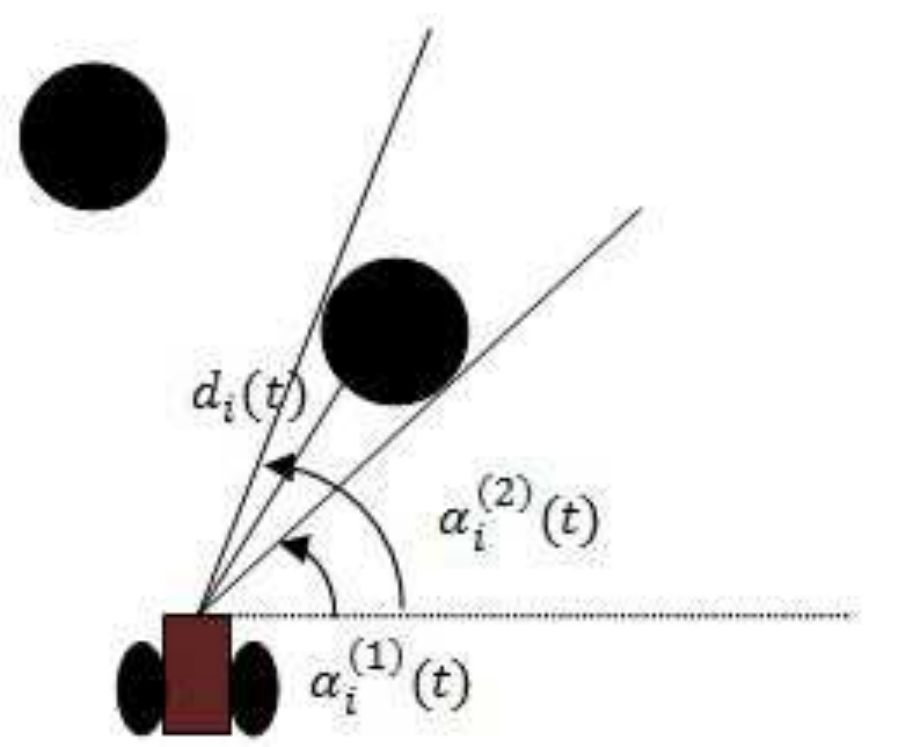}}
		\label{i1}}
		\subfigure[]{\scalebox{0.50}{\includegraphics{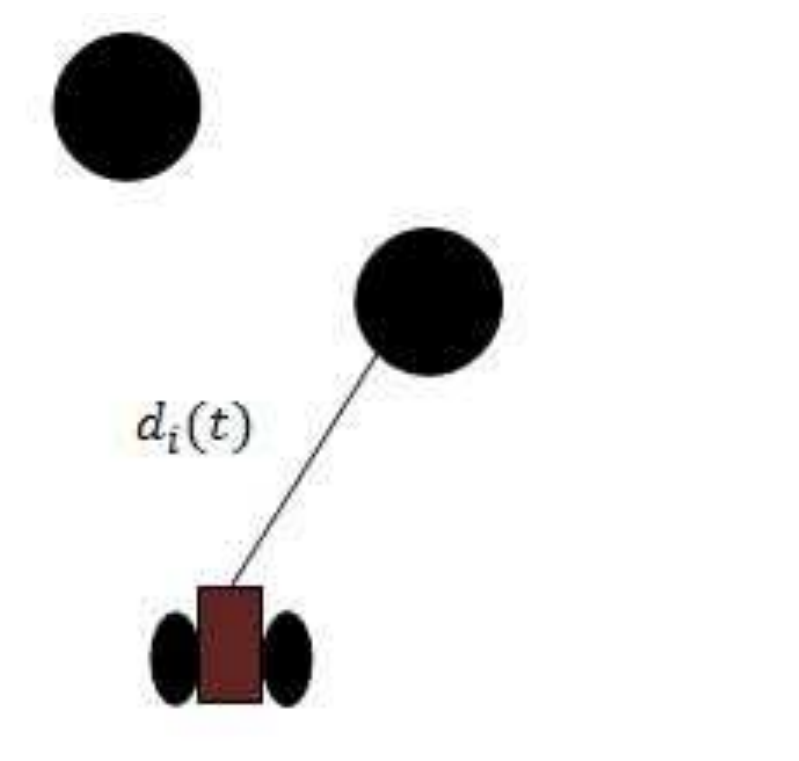}}
		\label{i2}}
		\subfigure[]{\scalebox{0.50}{\includegraphics{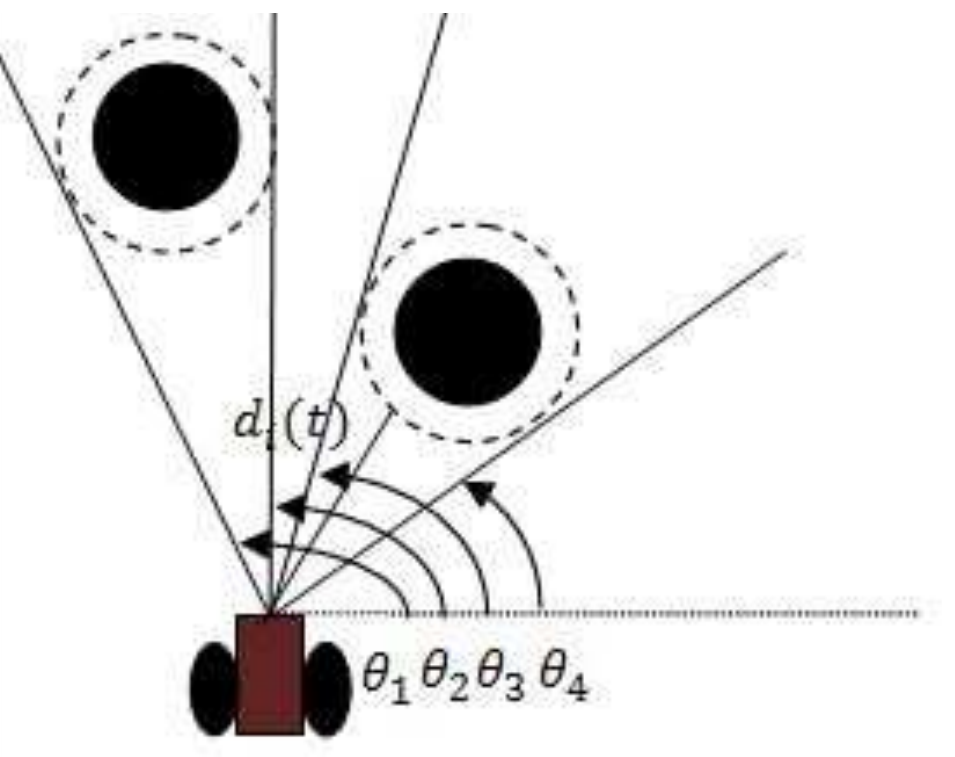}}
		\label{i3}}
		\caption{(a)Essential measurements for BINA (b)Essential measurements for ENA (c)Essential measurements for NAIER}
		\label{illu_meas}
		\end{figure}

		The proper execution of the proposed navigation algorithm relies on the acquisition of several measurements from the environments. Some of these measurements are common for all three algorithms, i.e., the position $(x,y)$ and the orientation $\theta$ of the robot, the angular difference $H(t)$ between the current heading of the robot and the direction of the target.  The required measurements by the algorithms to avoid obstacles are different from one to another, which has been described in section~\ref{PD1}, ~\ref{PDC3} and ~\ref{PDC4}, respectively.  Fig.~\ref{illu_meas} shows the measurements required by the algorithm to avoid obstacles.
\par
		
		It can be seen from Fig.~\ref{illu_meas} that, to successfully avoid the obstacles, ENA only requires the minimum distance $d_i(t)$ between the robot and the obstacle, whereas BINA and NAIER require extra angular measurements regarding to the obstacles, i,e. the vision cone ($\alpha_i^{(1)}, \alpha_i^{(2)}$) for BINA and the angular sectors covered by the enlarged environments for NAIER.
\par
		This implies that when designing hardwares for the algorithms in a control system, we can select a simple sensory device for ENA which only requires the distance measurement. More expensive and accurate sensory devices are required for BINA and NAIER. Therefore, ENA is potentially implementation cost efficient than BINA and NAIER.
\par

	\section {Computation Complexity}

		Computation complexity directly affects the performance of the navigation algorithms in real life scenarios. The amount of computation should be reasonable to obtain the proper control signals within a certain time interval, this is crucial for many time critical navigation tasks such as navigation in a highly populated dynamic environment where the algorithm needs to react to the changes in the environment quickly. The following Fig.~\ref{comp_chart} shows the flowcharts for the proposed navigation algorithms to compute the control signals. We only present the flowcharts for obstacle avoidance strategy for the proposed algorithms since the target reaching strategy is quite similar for all the algorithms. The flowchart starts with the measurements acquired from the environments and ends with the proper control signals which are computed using acquired measurements. Note that we do not include the descriptions of the defined variables and functions.  The detailed description of the variables and functions are explained in Section~\ref{A1}, ~\ref{A22} and ~\ref{A3}, respectively.  
\par

		\begin{figure}[!h]
		\centering
		\subfigure[]{\scalebox{0.50}{\includegraphics{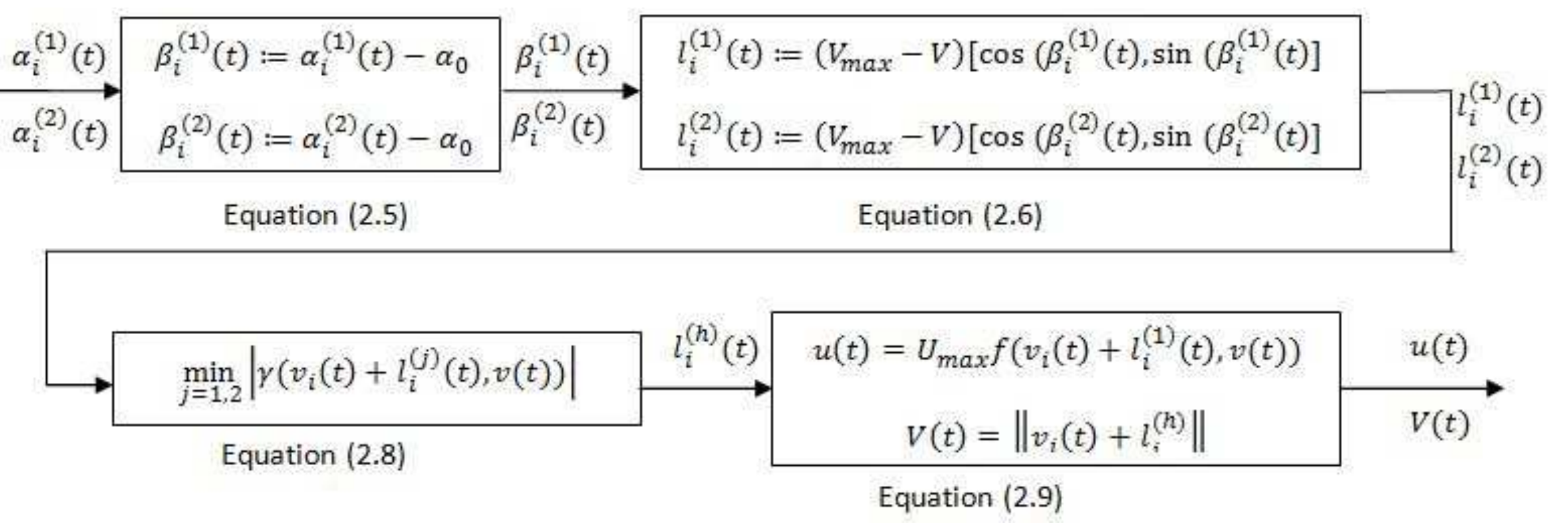}}
		\label{c1}}
		\subfigure[]{\scalebox{0.50}{\includegraphics{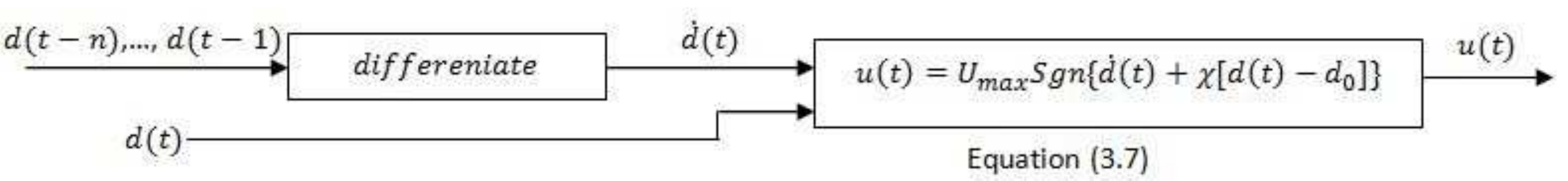}}
		\label{c2}}
		\subfigure[]{\scalebox{0.50}{\includegraphics{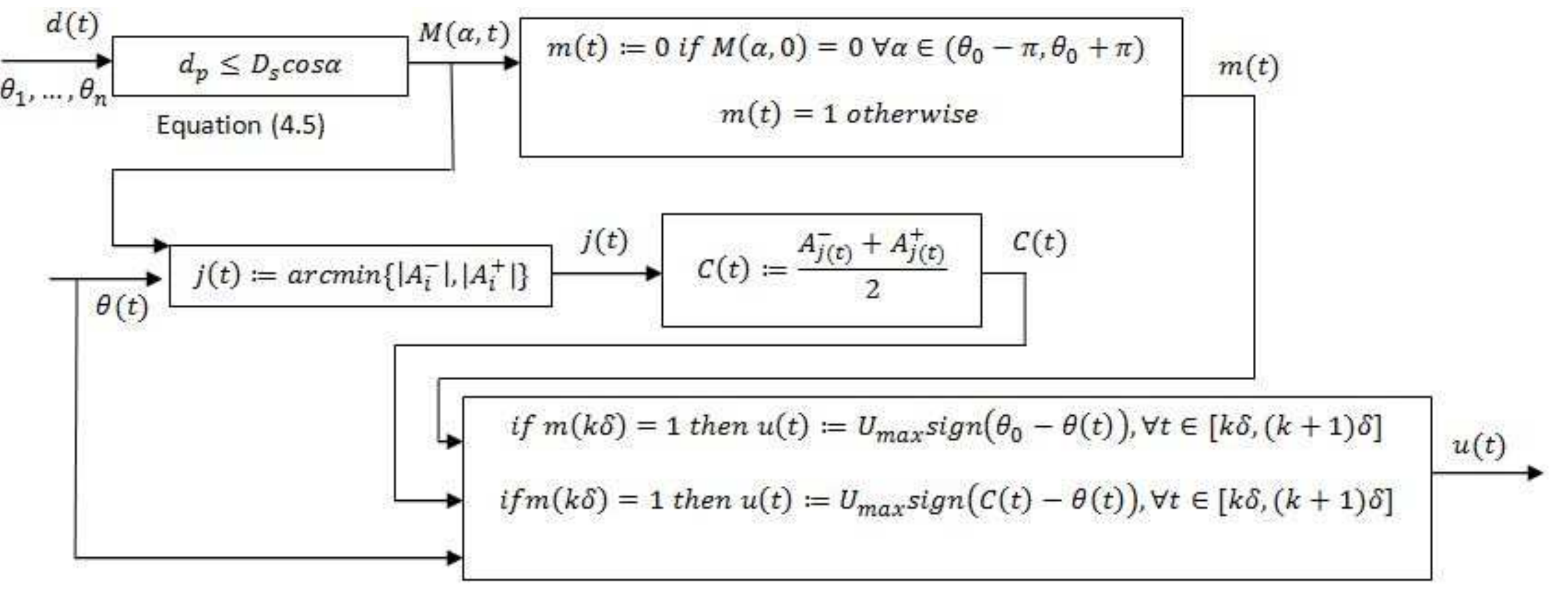}}
		\label{c3}}
		\caption{(a)Flowchart for BINA (b)Flowchart for ENA (c)Flowchart for NAIER}
		\label{comp_chart}
		\end{figure}

		It can be seen from Fig.~\ref{comp_chart}, Among all three algorithm, ENA has the minimum amount of computation steps. The overall performances of the proposed algorithms are quite similar because they do not rely on extensive computations within a short period of time. The computation of all three navigation algorithms require only the arithmetic calculations which is quite easy to compute based on current average computation power. Therefore, all of these proposed navigation algorithms are considered as computationally efficient in certain scenarios.

	\section {Performance Comparison}

		In this section, we compare the performances of the three proposed navigation algorithms in different scenarios. Although all the proposed navigation algorithm are able to accomplish navigation tasks under various challenging scenario as shown in corresponding computer simulation sections and real experiment sections, each of the proposed navigation algorithms has its own unique characteristics which make them more efficient in some particular scenarios. 
\par
 In order to compare these navigation algorithms with the same objective, we slightly change the objective of NAIER from "desired direction reaching" to "target position reaching" by making the desired direction time-varying, which equals to the direction from the robot current location to the target position.

		\subsection {Stationary Obstacles}

		The examples of navigation in the environments with stationary obstacles are very easily found in our everyday life, e.g. the autonomous vacuum cleaner robots clear the dust while avoid colliding with the furnitures and walls inside houses, the transportation robots deliver items from a certain place to another in a warehouse in pre-defined routes which avoid the en-route obstacles. The obstacles in these scenarios can be estimated by circles or regular polygons for simplicity and are easy to avoid. The global navigation algorithms are commonly used to avoid stationary obstacles which require a complete map of the environments and these algorithms are computationally expensive. Our proposed navigation algorithms does not require the complete knowledge of the environments and are able to make proper decision on-the-fly during a navigation task. Furthermore, we do not restrict the shapes of the obstacles to be only circles or regular polygons. The shapes of the obstacles can be irregular (convex) which can be seen in many real life scenarios. For example, obstacles encountered by exploration or rescue robots in hazardous environments are usually rocks and swamp of irregular shapes. 
\par
\begin{figure}[!h]
		\centering
		\subfigure[]{\scalebox{0.45}{\includegraphics{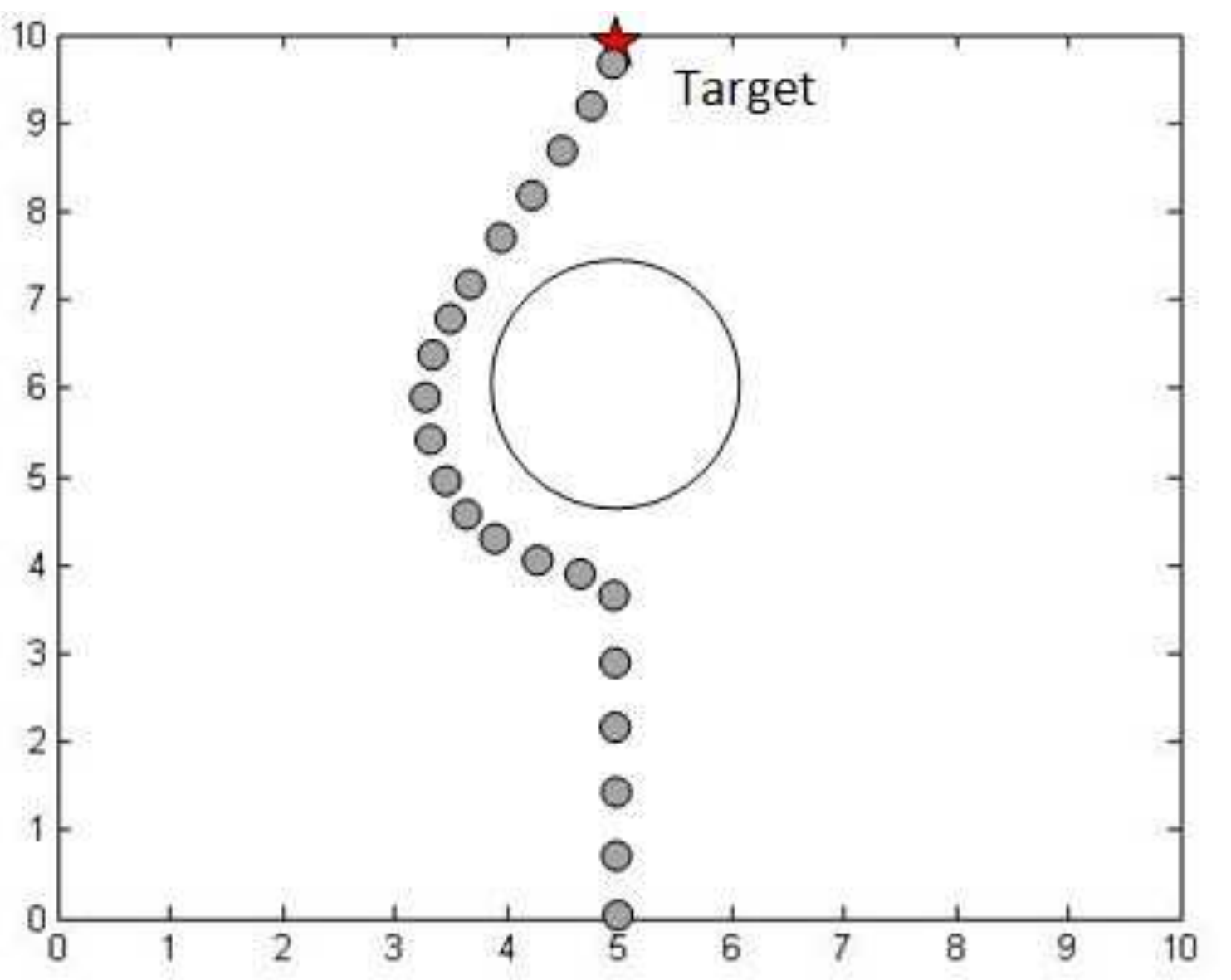}}
		\label{c5.sim11}}
		\subfigure[]{\scalebox{0.45}{\includegraphics{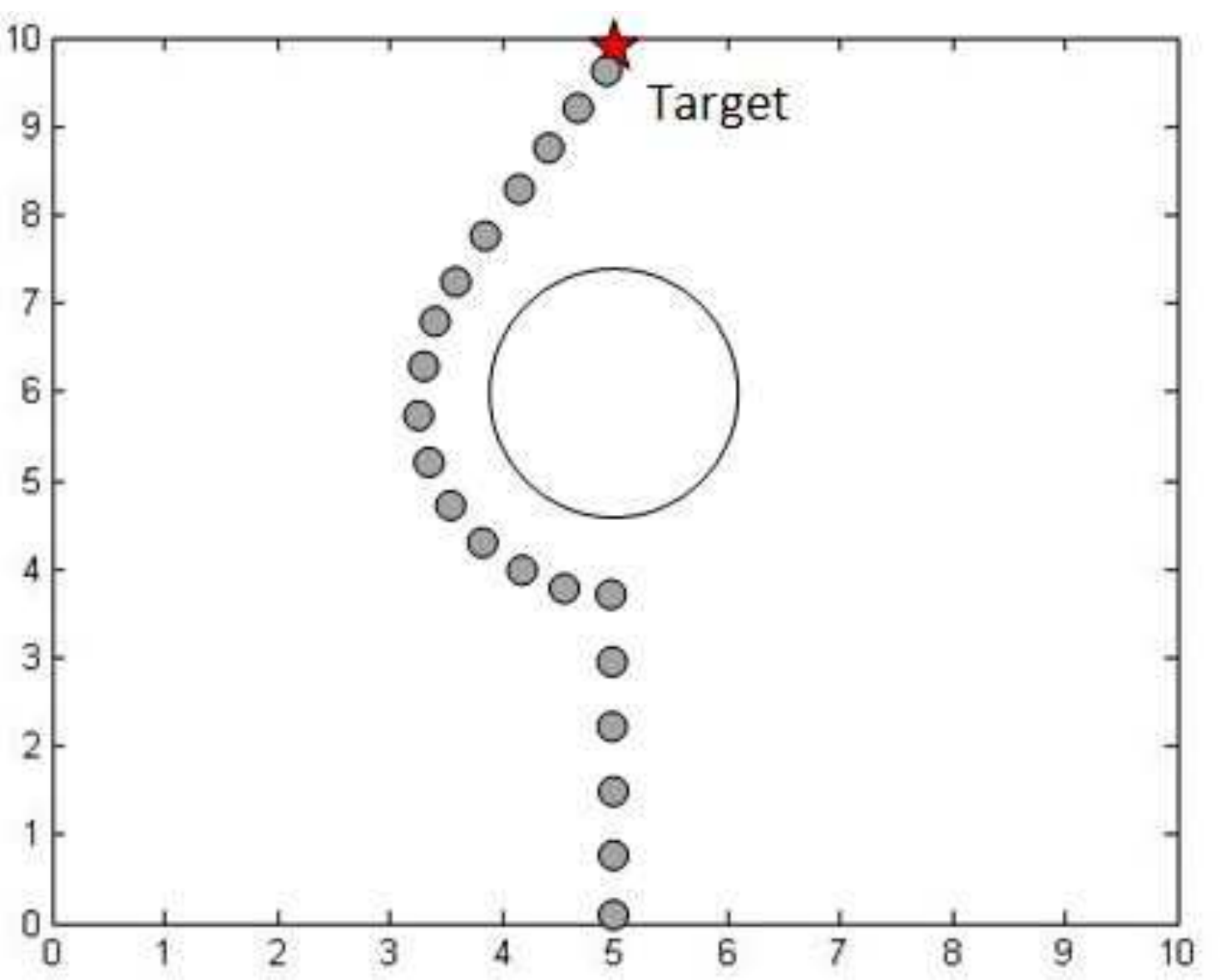}}
		\label{c5.sim12}}
		\subfigure[]{\scalebox{0.45}{\includegraphics{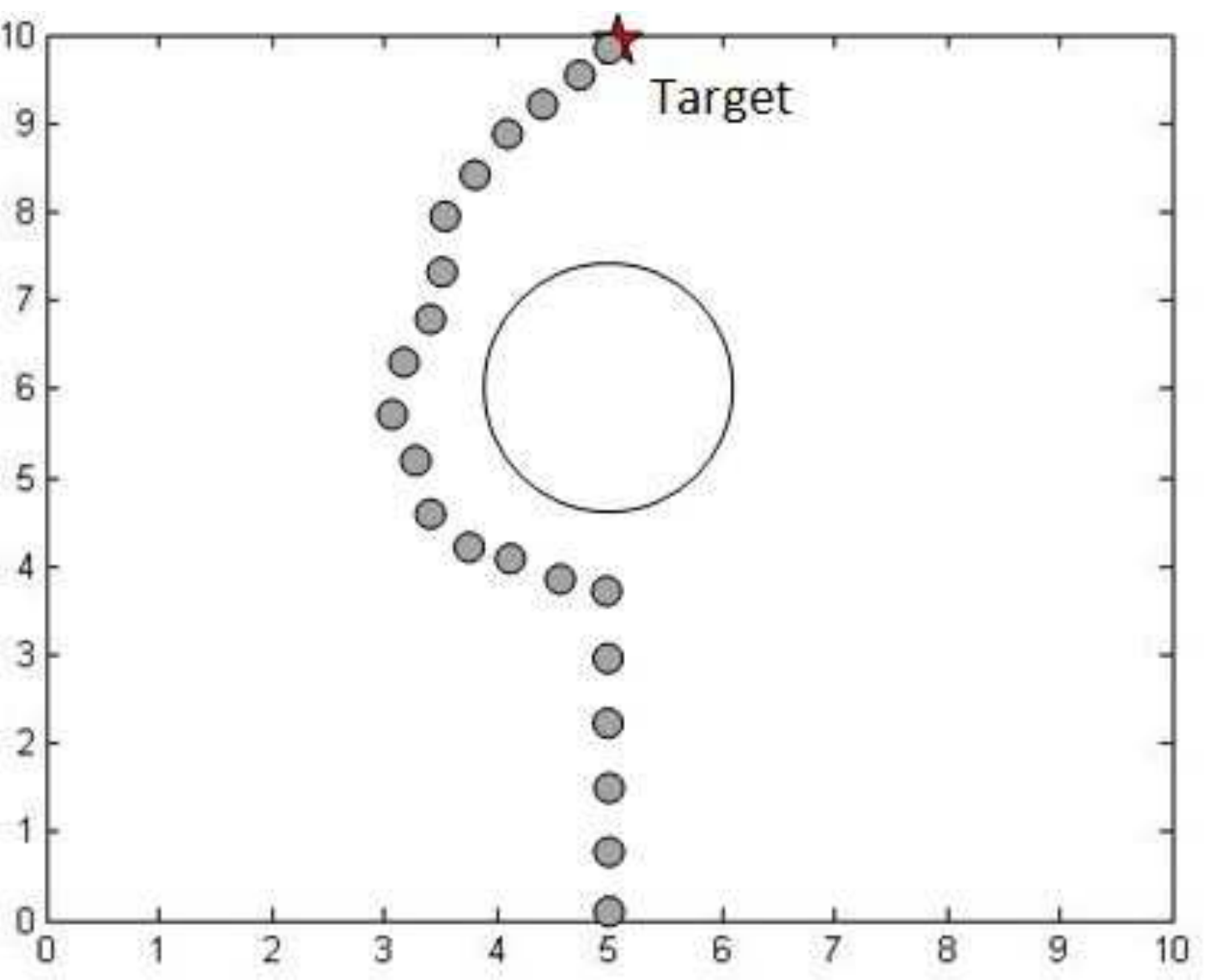}}
		\label{c5.sim13}}
		\caption{Performance comparison: single stationary obstacle for (a) BINA (b)ENA (C)NAIER}
		\label{c5.sim1}
		\end{figure}

		Fig.~\ref{c5.sim1} shows the performance of the proposed algorithms with one single stationary obstacle of circular shape. Many of the real world obstacles can be estimated as circles which are considered as one of the simplest test subjects for obstacle avoidance algorithms. The performance of BINA and ENA is slightly better than that of NAIER since the NAIER has to switch between obstacle avoidance maneuver and target reaching maneuver when the obstacle is nearby (partially due to the change of objective). The performance of BINA and ENA is almost identical which can be explained by the features of both algorithms: BINA guides the robot so that the robot's trajectory converging to the circle of radius $\frac{R_i}{\cos\alpha_0}$ (the deviation is given in \cite{SAW13}), whereas ENA drives the robot to a $d_0$-equidistant curve around the obstacle. Therefore, both algorithms drive the robot to a circle around the obstacle in this particular scenario and it is a quite efficient path to avoid a circular obstacle while retaining a safety margin.
\par
\begin{figure}[!h]
		\centering
		\subfigure[]{\scalebox{0.45}{\includegraphics{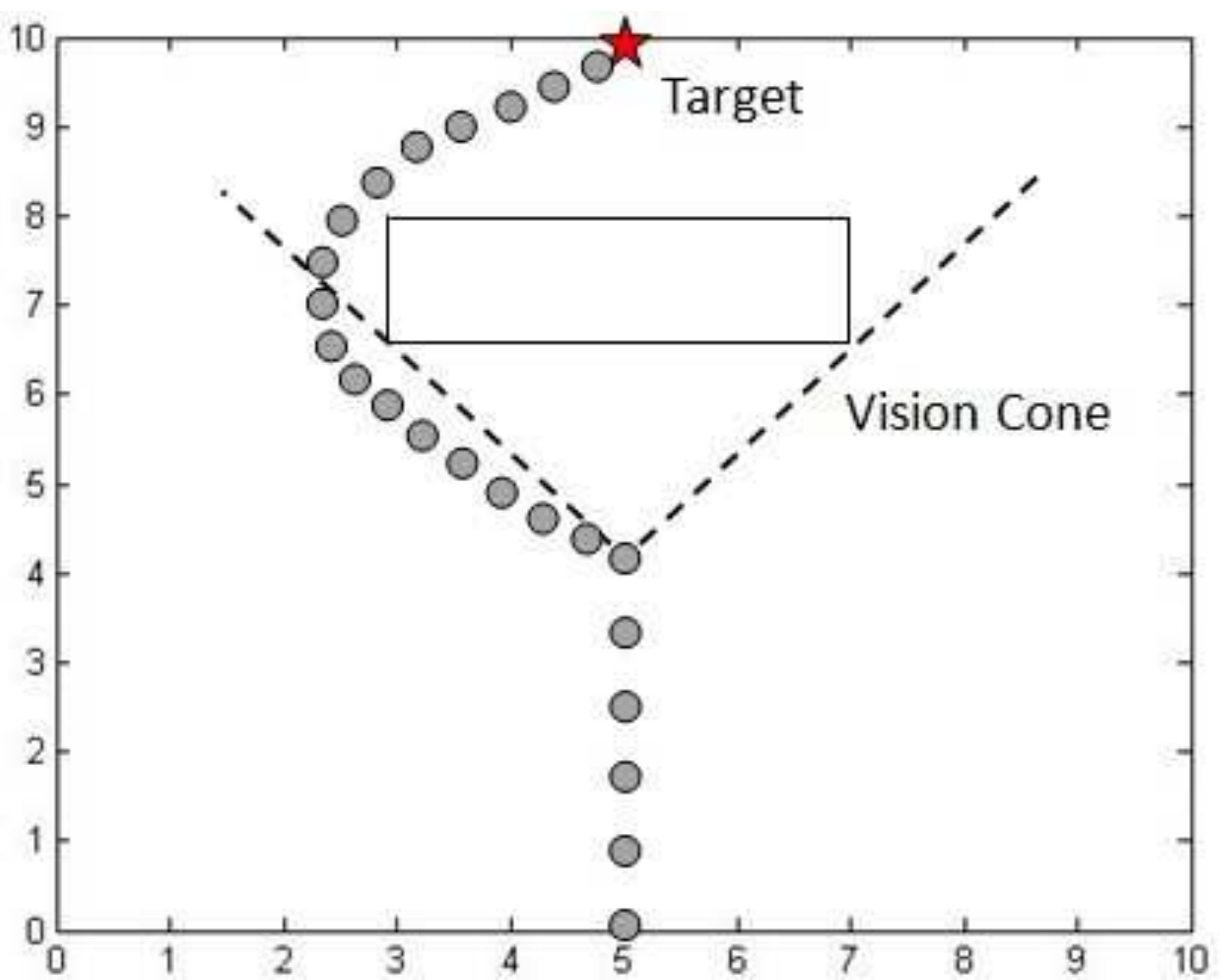}}
		\label{c5.sim21}}
		\subfigure[]{\scalebox{0.45}{\includegraphics{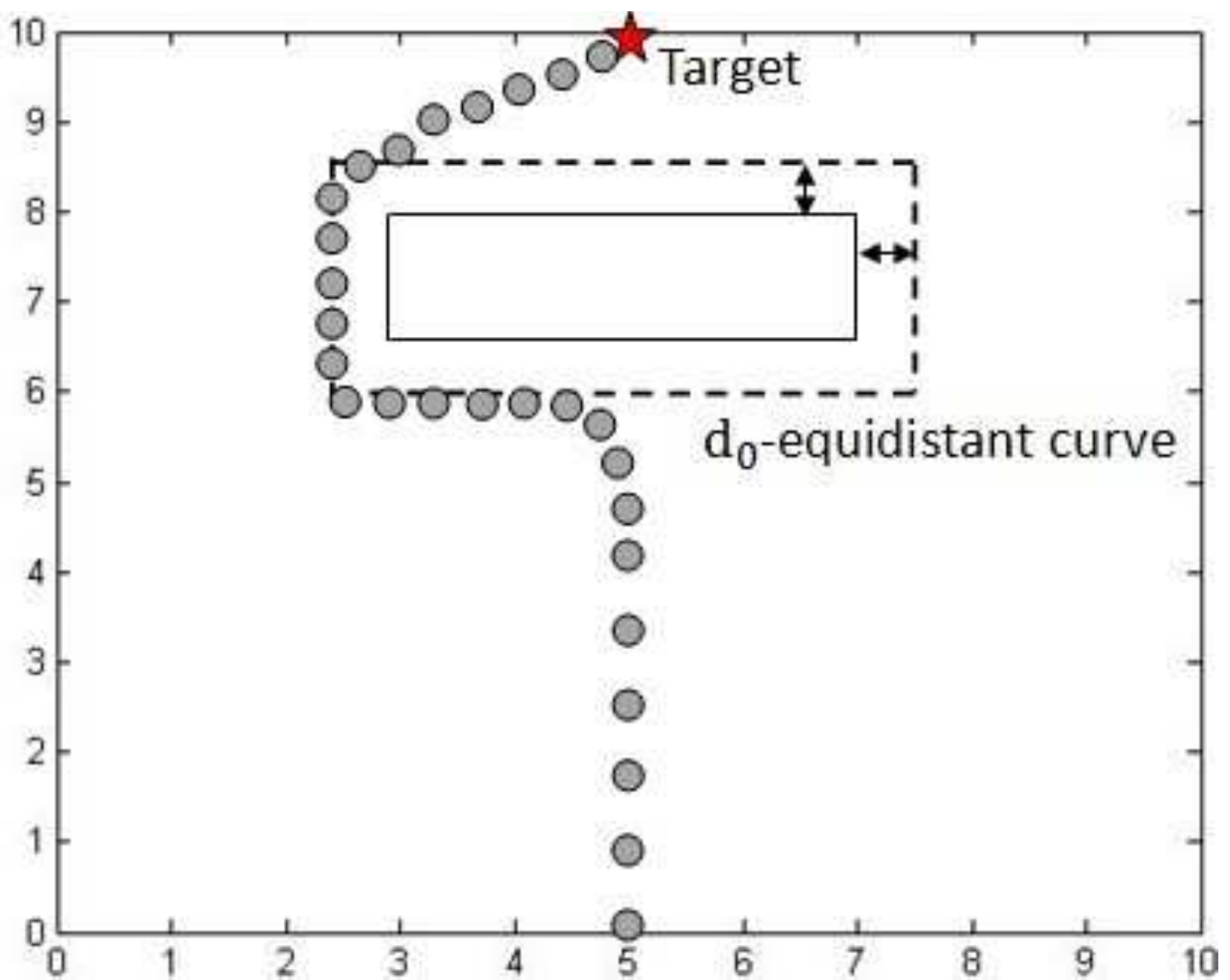}}
		\label{c5.sim22}}
		\subfigure[]{\scalebox{0.45}{\includegraphics{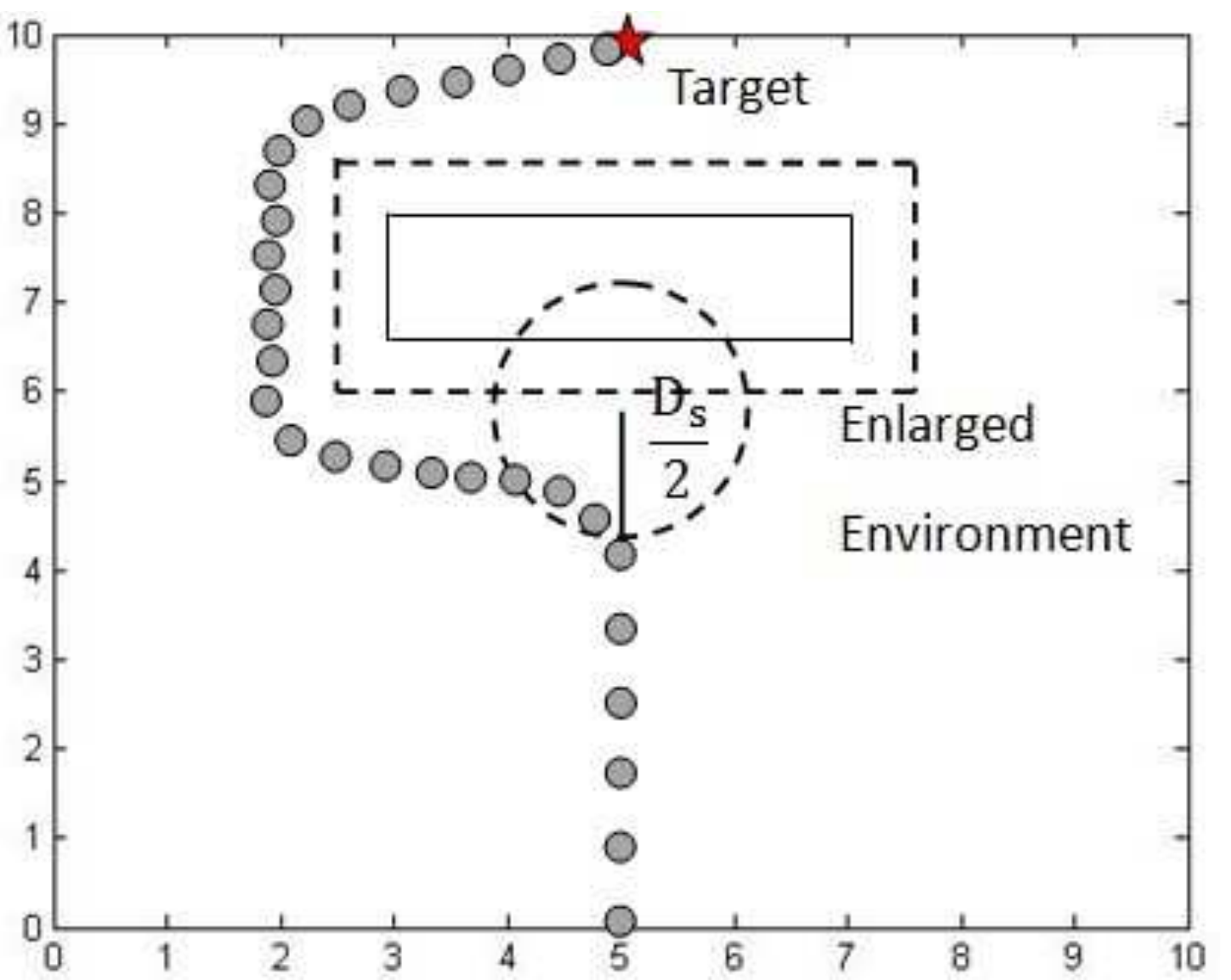}}
		\label{c5.sim23}}
		\caption{Performance comparison: single rectangular obstacle for (a) BINA (b)ENA (C)NAIER}
		\label{c5.sim2}
		\end{figure}

		In the next figure, we replace the circular obstacle with a rectangular obstacle. Obstacles with rectangular shapes are very common in real life scenarios. The shapes of many obstacles, such as tables, benches, can be estimated by a rectangle. In Fig.~\ref{c5.sim2}, we depict the crucial parameters during obstacle avoidance maneuvers for each of the algorithms, i.e., vision cone for BINA, $d_0$-equidistant curve for ENA and the detection range for NAIER. In this simulation, BINA guides the robot to the target location while avoiding the obstacle with minimum time ($23.5 sec$). ENA and NAIER achieve the same navigation task with $27.6 sec$ and $29.8 sec$, respectively.
\par

\begin{figure}[!h]
		\centering
		\subfigure[]{\scalebox{0.4}{\includegraphics{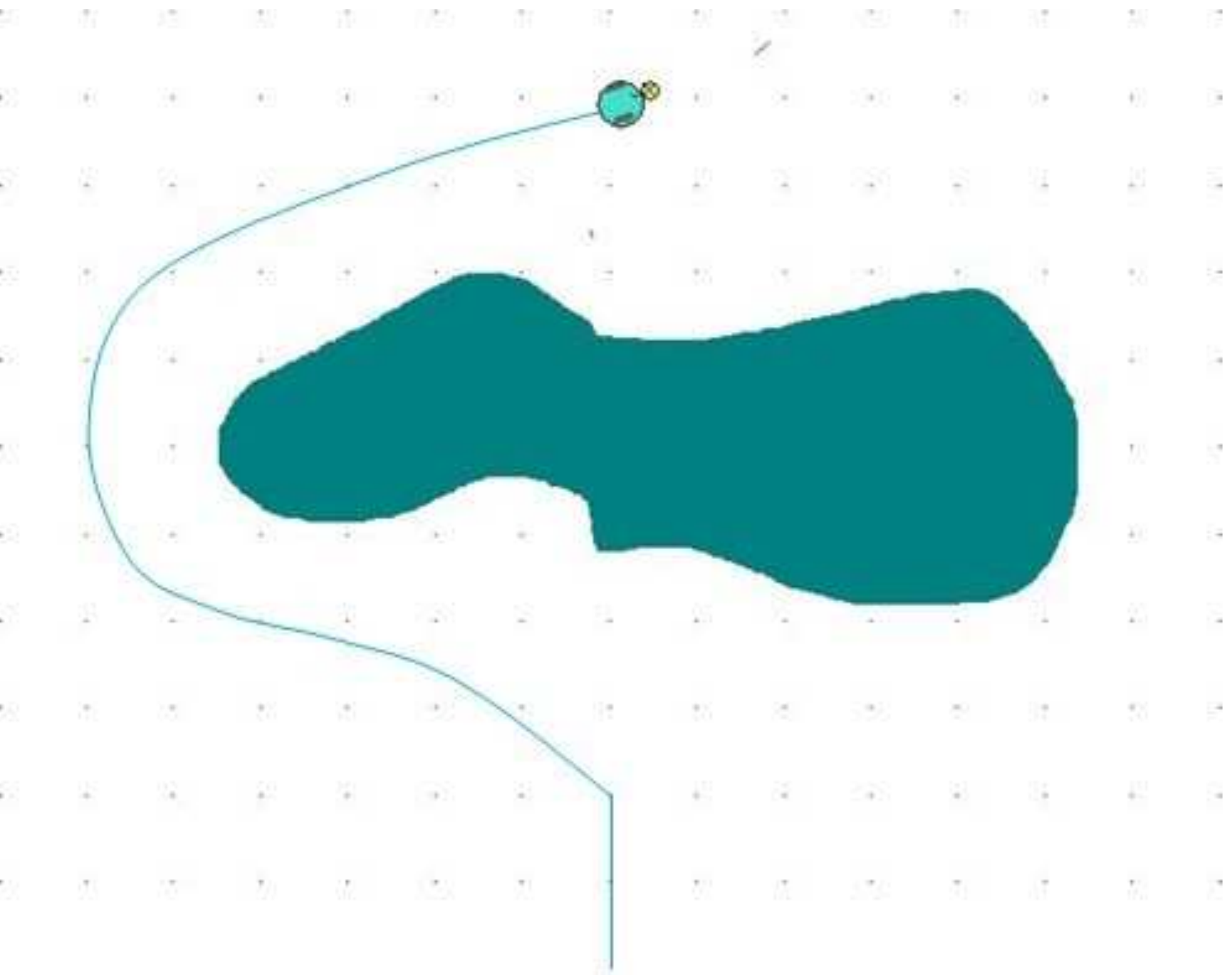}}
		\label{c5.sim31}}
		\subfigure[]{\scalebox{0.4}{\includegraphics{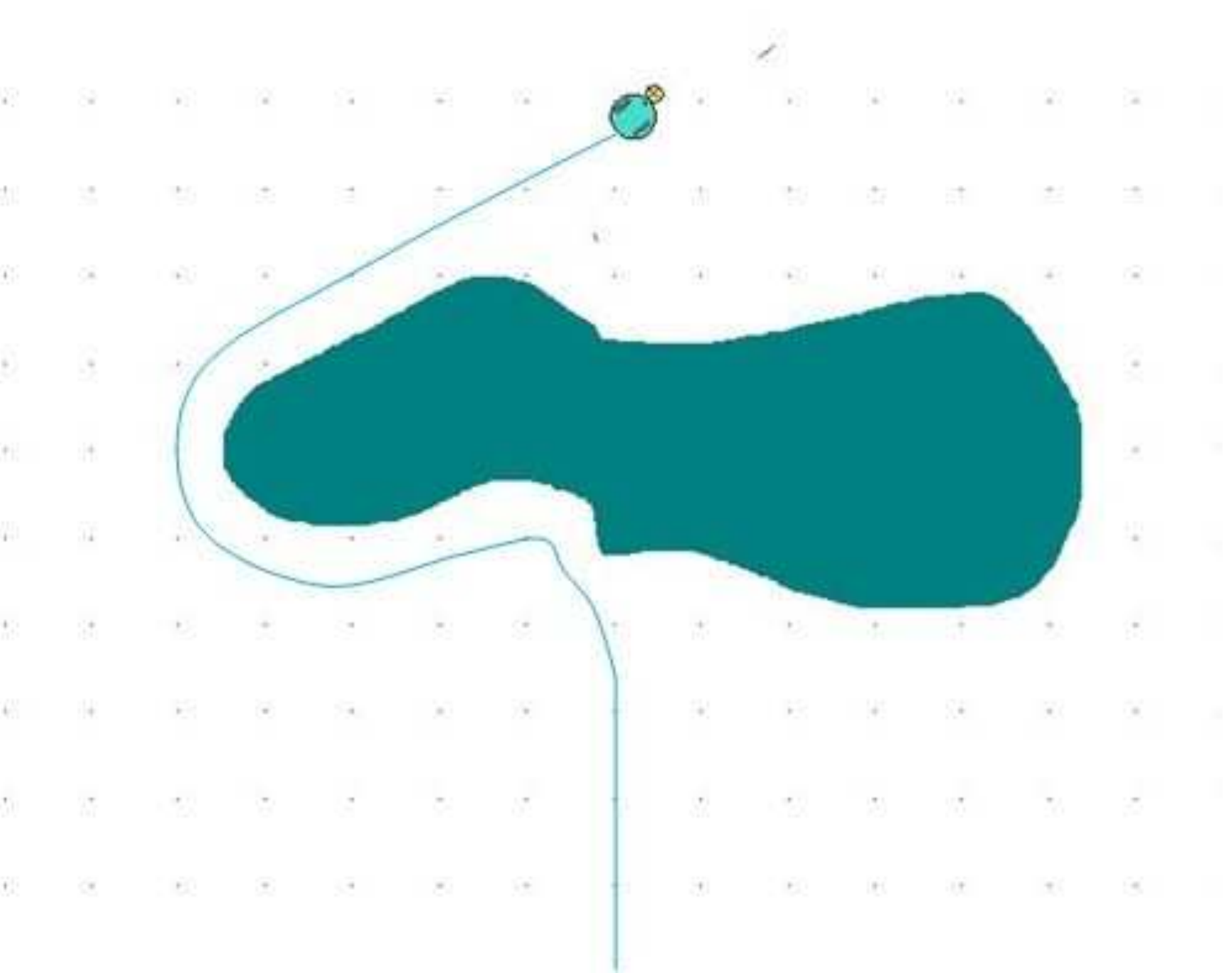}}
		\label{c5.sim32}}
		\subfigure[]{\scalebox{0.4}{\includegraphics{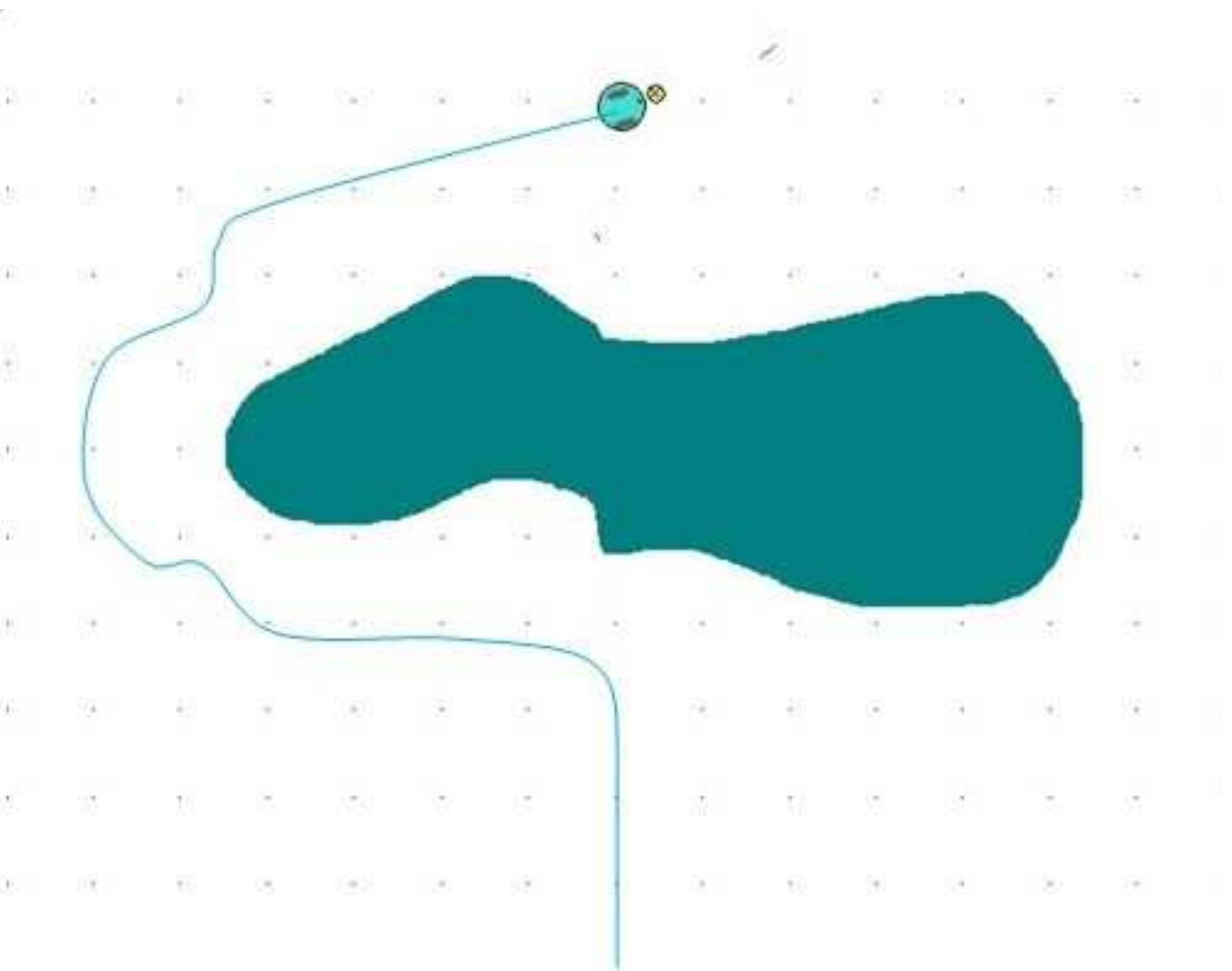}}
		\label{c5.sim33}}
		\caption{Performance comparison: single obstacle with irregular shape for (a) BINA (b)ENA (C)NAIER}
		\label{c5.sim3}
		\end{figure}

		Fig.~\ref{c5.sim3} presents a more challenging scenario with obstacle of irregular shape. Although shapes of most of obstacles in real life can be estimated by circle or rectangle, some of the potentially safe path may be occupied when the estimation is applied. Therefore, it is advantageous for an algorithm to avoid obstacles with irregular shapes. In this case, ENA is most efficient in term of navigation time ($18.1 sec$) since it is able to guide the robot so that it tracks the border of the obstacle ($d_0$-equidistant curve) at close distance. The navigation time for BINA and NAIER are $24.6 sec$ and $23.5 sec$, which are quite close to that of ENA.
\par

		In the last simulation, we show an extension scenario of Fig.~\ref{c5.sim3}. The proposed navigation algorithms need to guide the robot though environments crowded with obstacles of irregular shapes. In this case, NAIER has the best performance with overall navigation time of $21.5 sec$. The advantage of NAIER is that it is able to seek a free path thought obstacles. This is very efficient when the environment is populated with multiple closely positioned obstacles. ENA achieves navigation task with $25.3 sec$. The robot follows the $d_0$-equidistant curve of every en-route obstacles under the guidance of ENA. It takes $35.5 sec$ for BINA to complete the navigation task bacause the use of vision cone forces the robot to take a longer detour to avoid the crowd of obstacles.
\par
		\begin{figure}[!h]
		\centering
		\subfigure[]{\scalebox{0.5}{\includegraphics{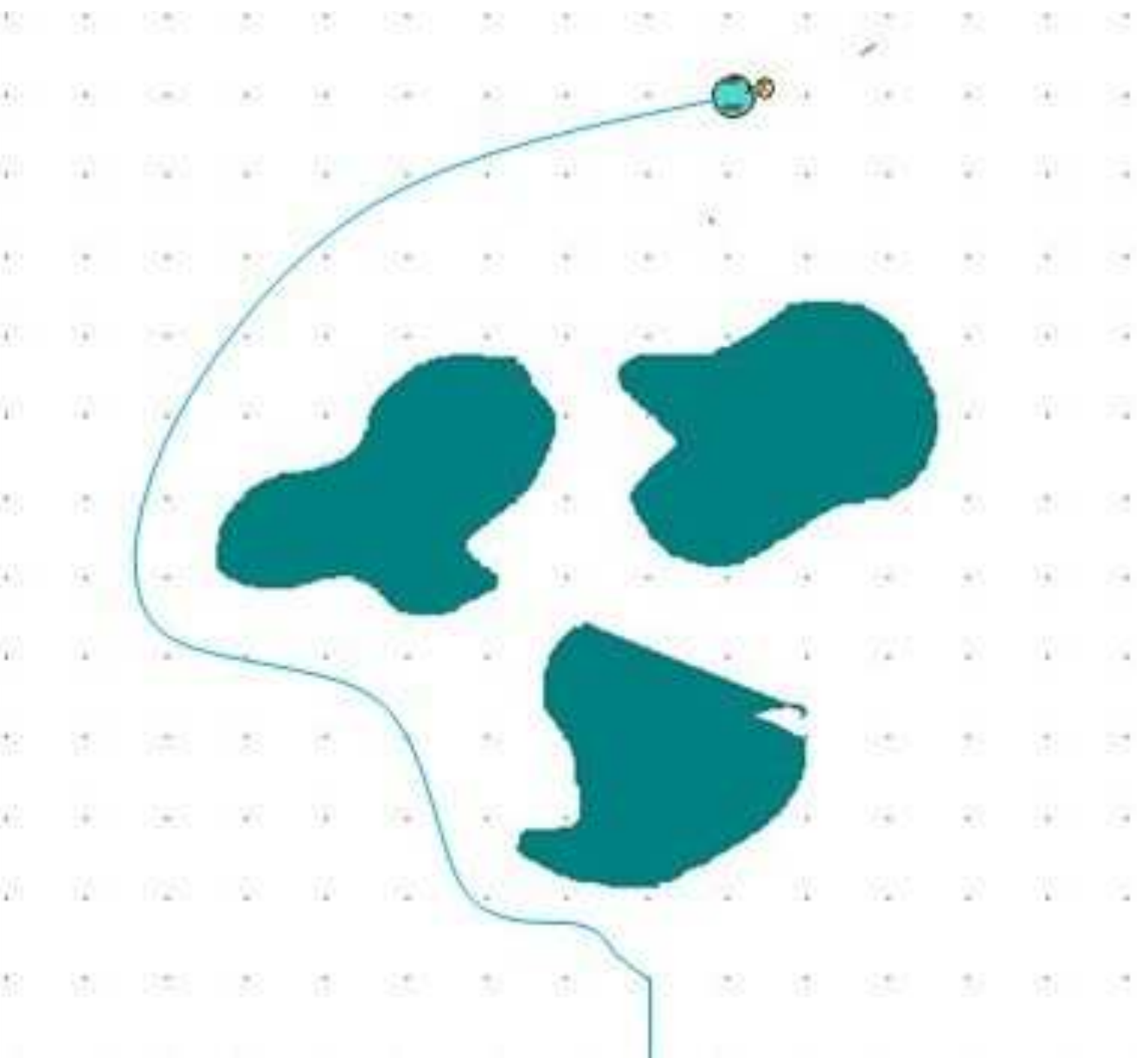}}
		\label{c5.sim41}}
		\subfigure[]{\scalebox{0.5}{\includegraphics{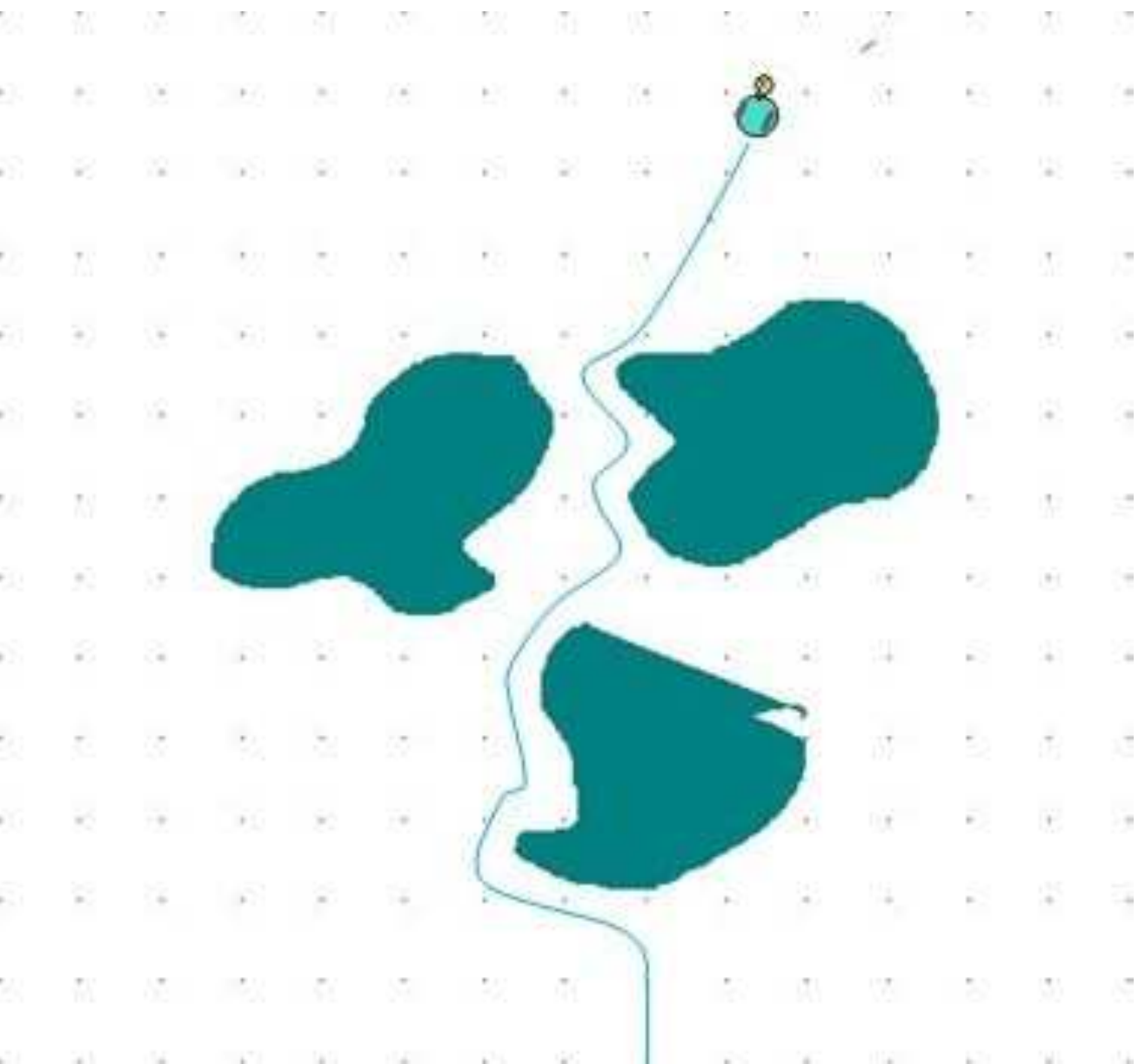}}
		\label{c5.sim42}}
		\subfigure[]{\scalebox{0.5}{\includegraphics{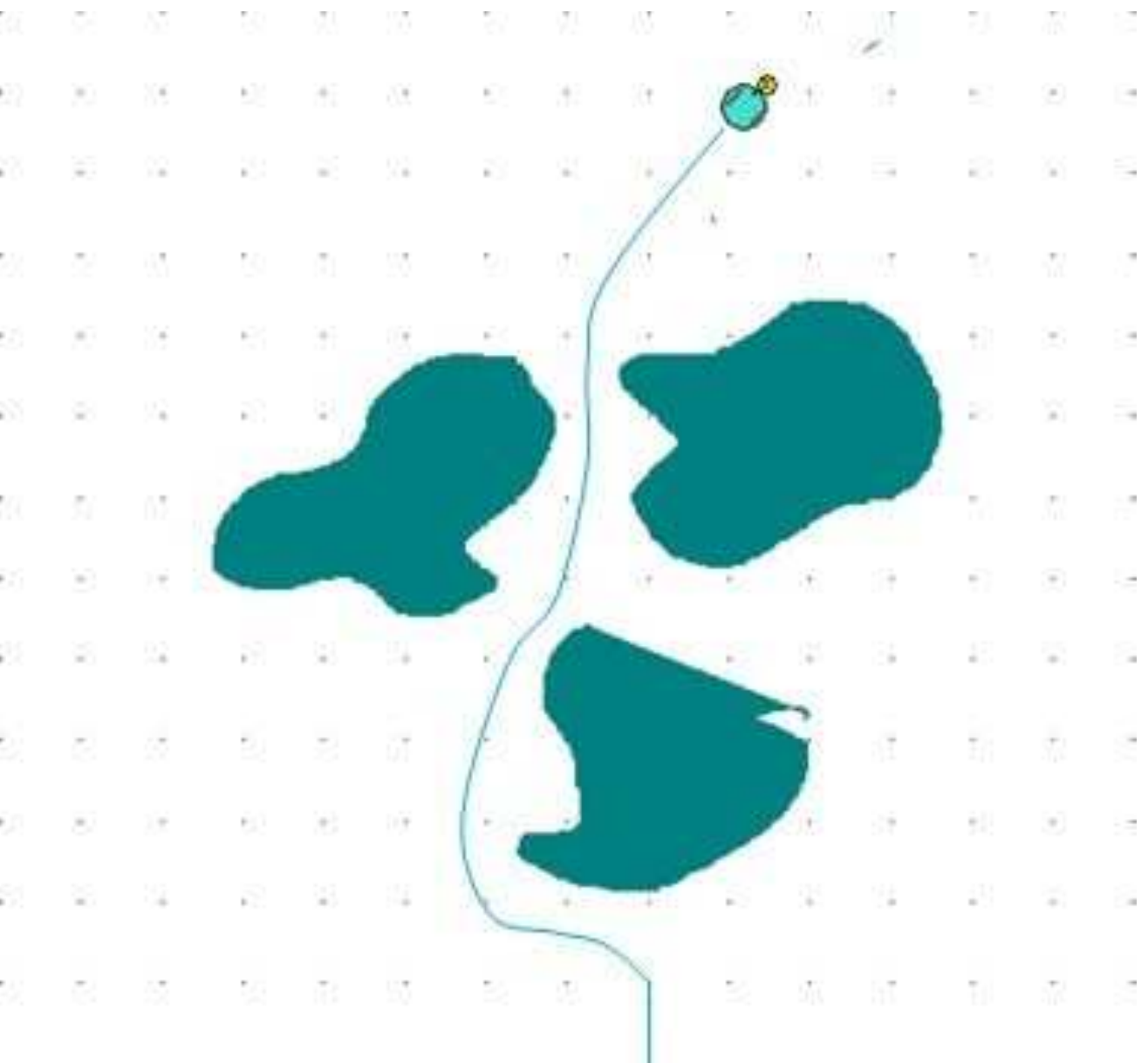}}
		\label{c5.sim43}}
		\caption{Performance comparison: a crowd of obstacles with irregular shapes for (a) BINA (b)ENA (C)NAIER}
		\label{c5.sim4}
		\end{figure}

		\subsection{Dynamic Obstacles}

			In many of real life scenarios, the obstacles are often moving with various speed rather than staying stationary. For example, the museum tour robots and the service robots in nursing houses are usually  operating in the environments with multiple moving objects.  It is much more difficult for an algorithm to navigate a vehicle in a dynamic environment with moving obstacles. There are several reasons for this: firstly, the navigation algorithms need to consider more real time parameters when dealing with moving obstacles, such as obstacles' velocities and accelerations, obstacles' future trajectories. Secondly, the obstacles are not always moving with constant speeds, some obstacles may move with some complicated and unpredictable non-linear velocities. Finally, the shapes of the obstacles may be time-varying in the dynamic environments.
\par
\begin{figure}[!h]
		\centering
		\subfigure[]{\scalebox{0.4}{\includegraphics{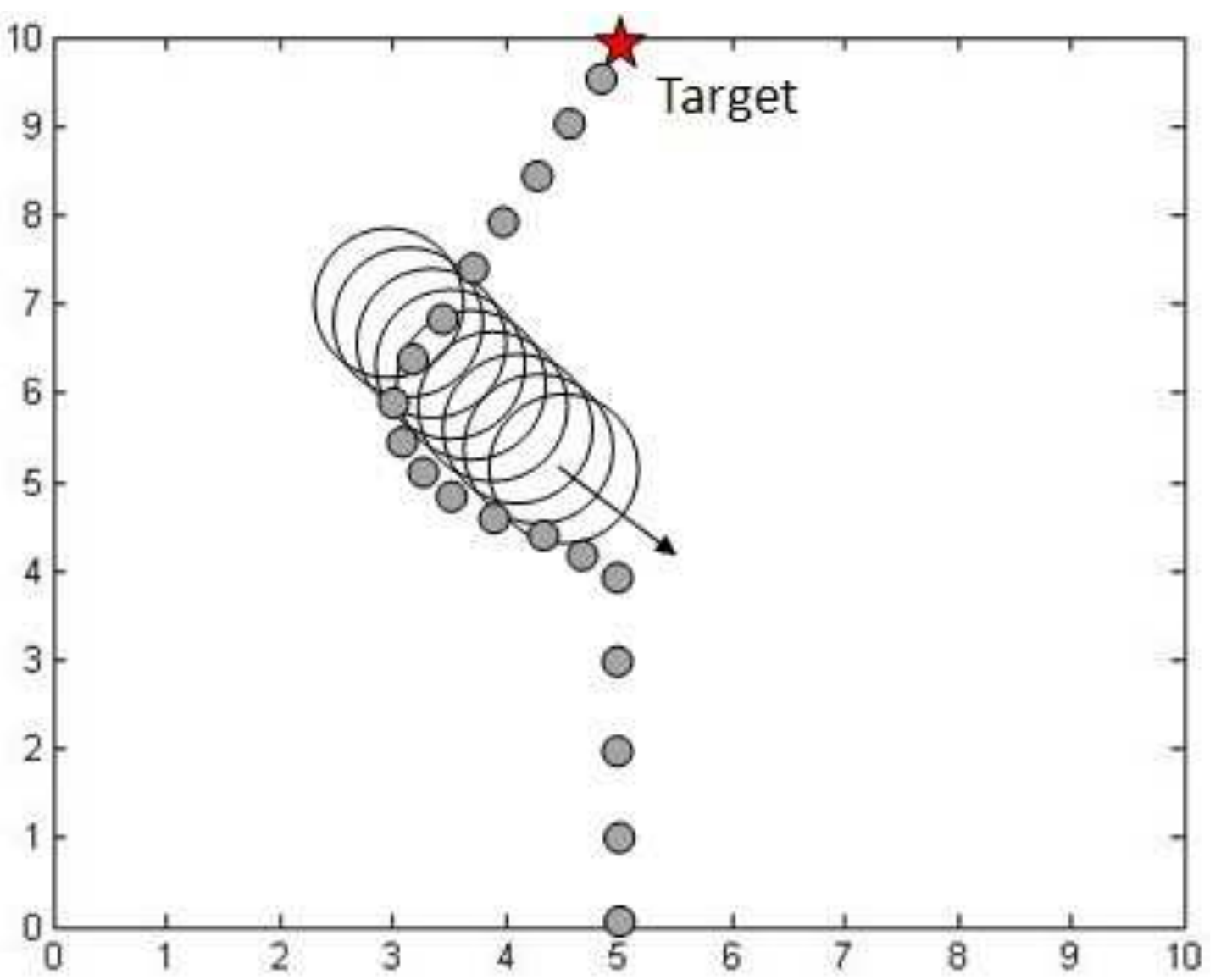}}
		\label{c5.sim51}}
		\subfigure[]{\scalebox{0.4}{\includegraphics{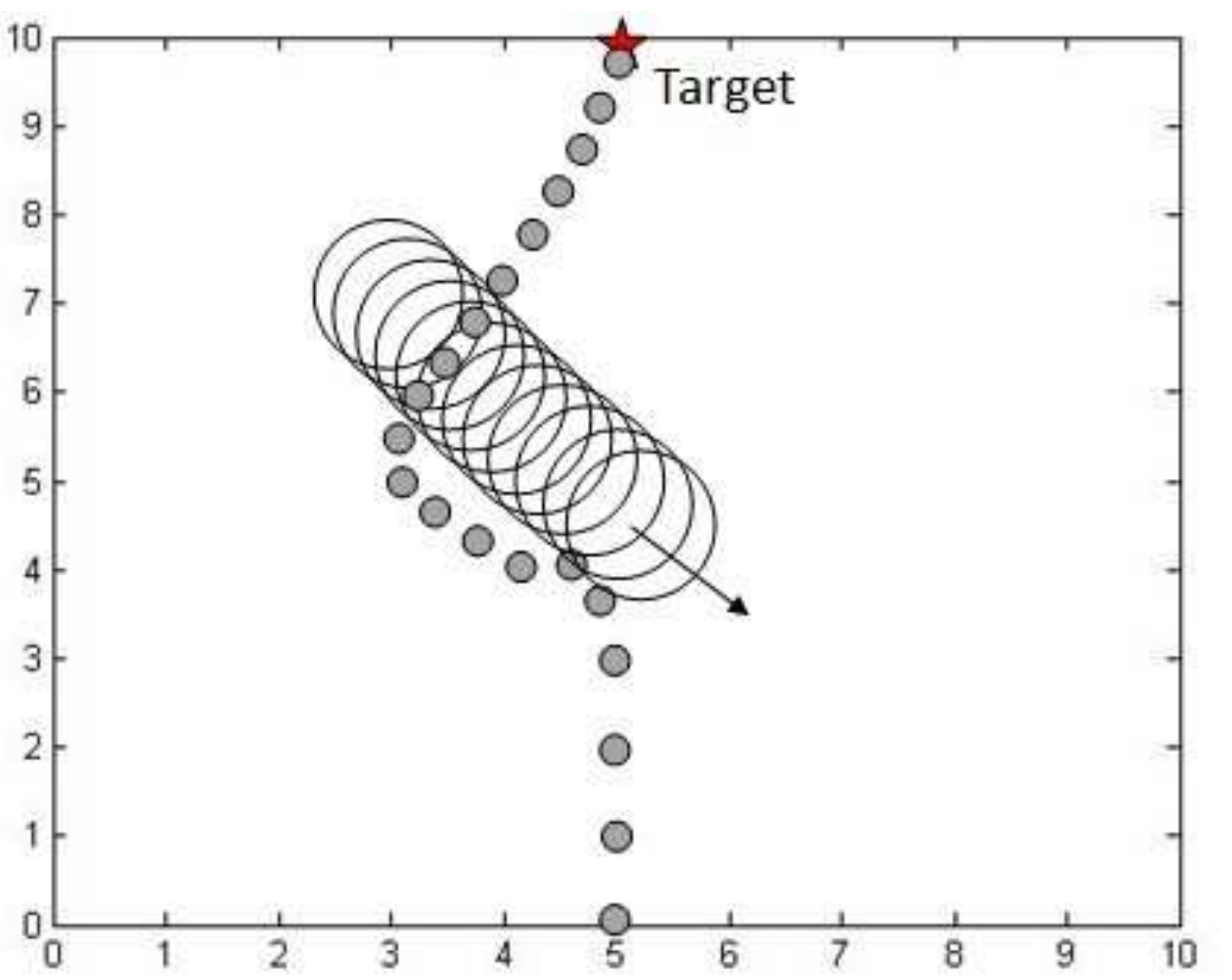}}
		\label{c5.sim52}}
		\subfigure[]{\scalebox{0.4}{\includegraphics{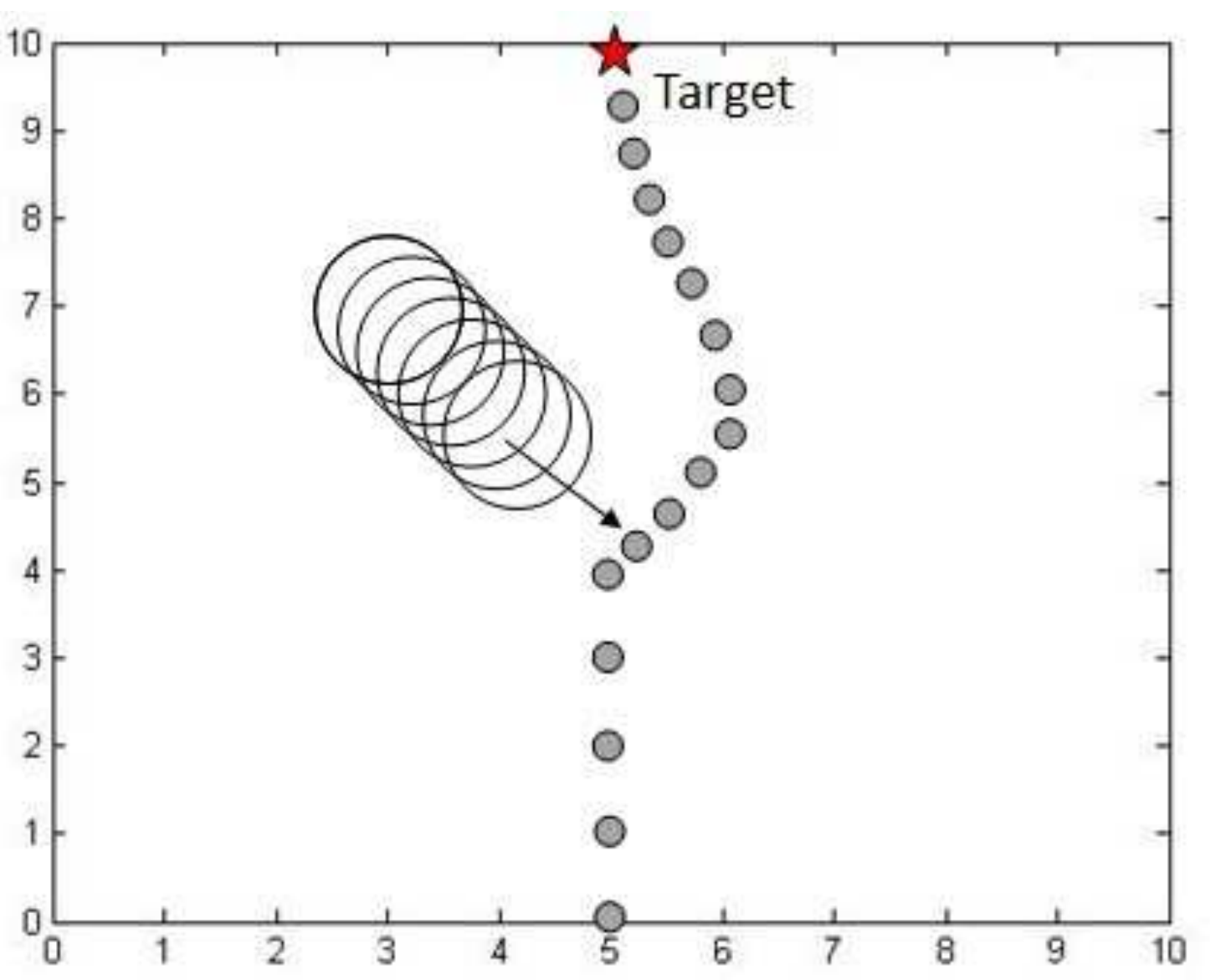}}
		\label{c5.sim53}}
		\caption{Performance comparison: dynamic environment with single moving obstacle for (a) BINA (b)ENA (C)NAIER}
		\label{c5.sim5}
		\end{figure}
\begin{figure}[!h]
		\centering
		\subfigure[]{\scalebox{0.4}{\includegraphics{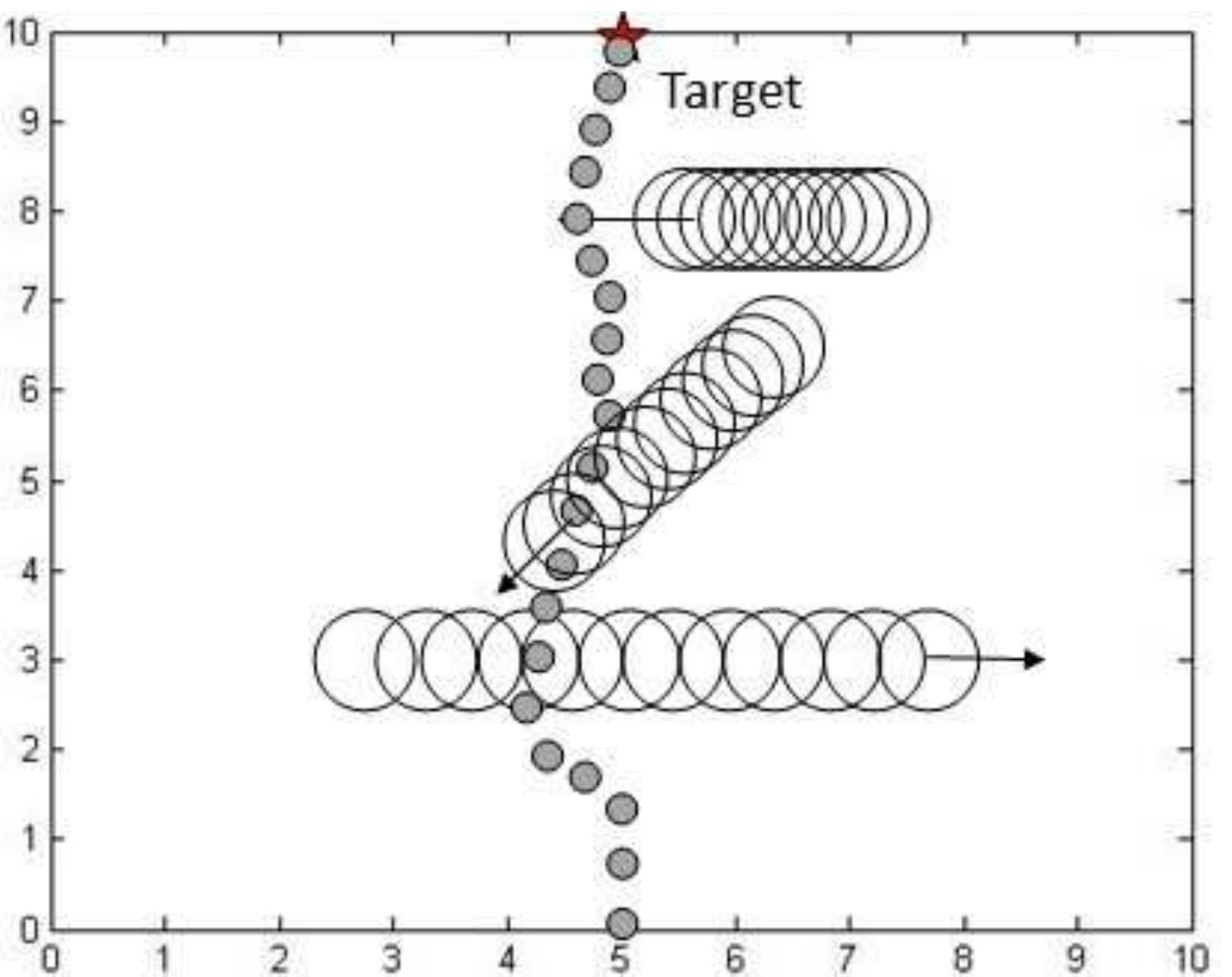}}
		\label{c5.sim61}}
		\subfigure[]{\scalebox{0.4}{\includegraphics{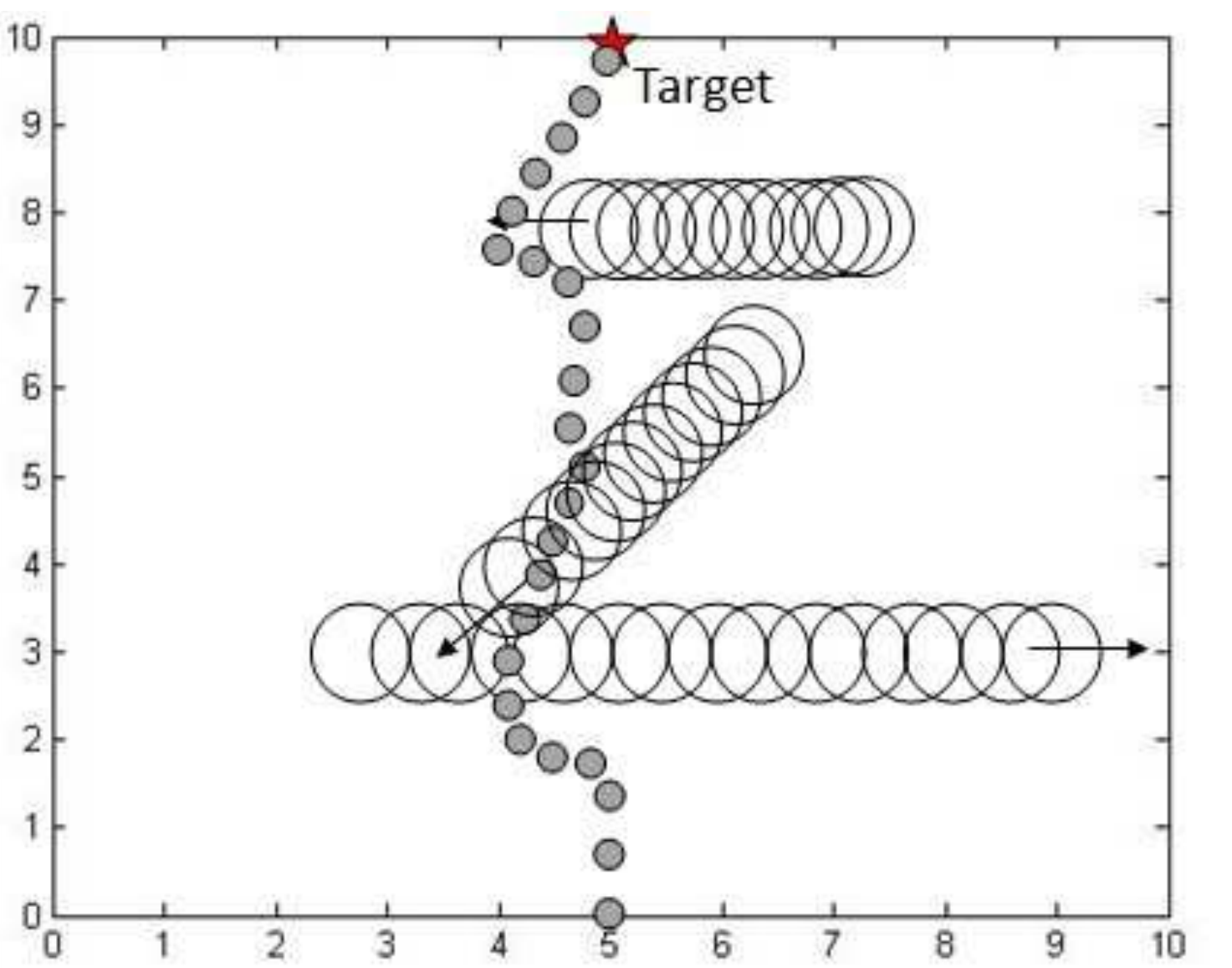}}
		\label{c5.sim62}}
		\subfigure[]{\scalebox{0.4}{\includegraphics{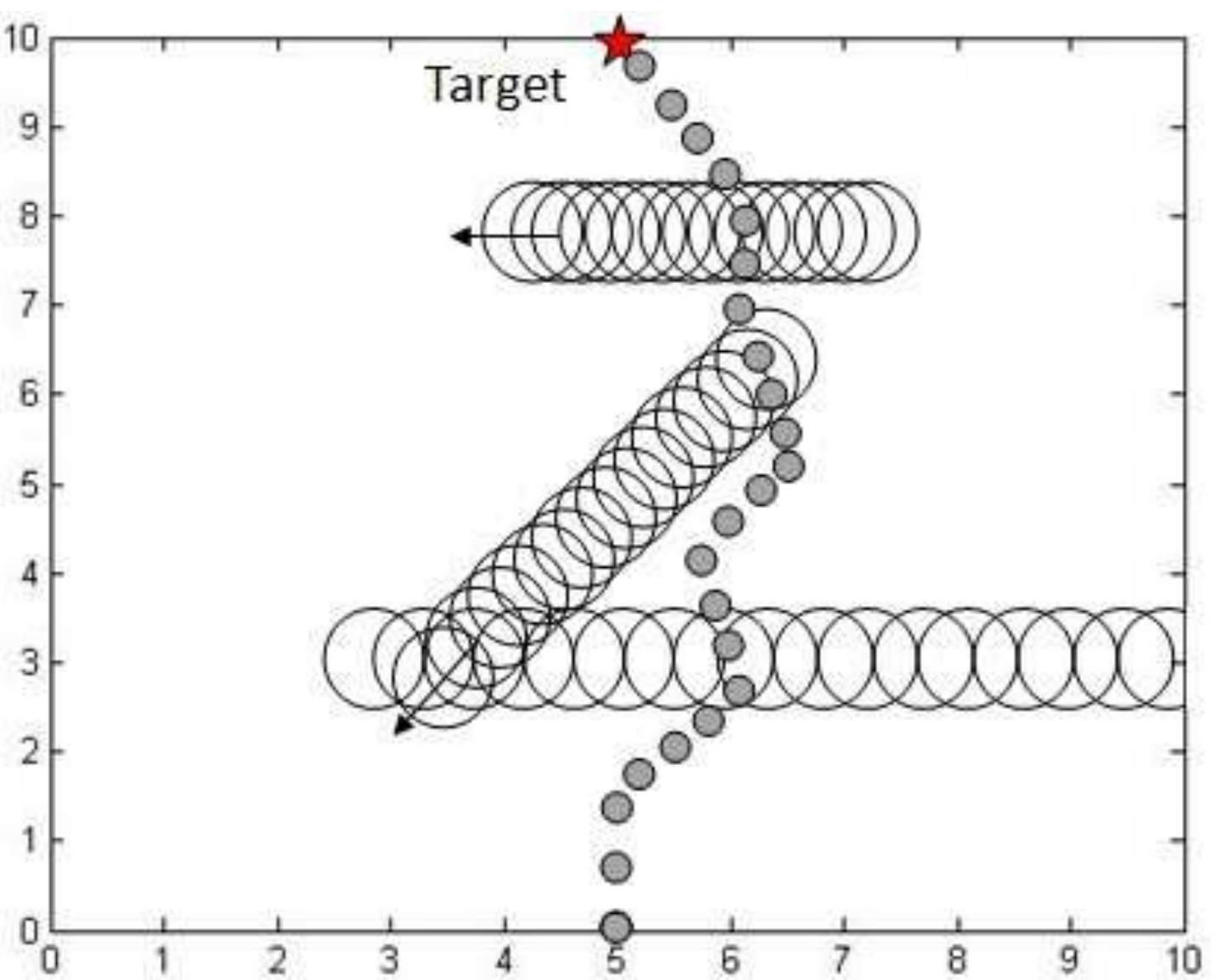}}
		\label{c5.sim63}}
		\caption{Performance comparison: dynamic environment with three moving obstacles for (a) BINA (b)ENA (C)NAIER}
		\label{c5.sim6}
		\end{figure}
\begin{figure}[!h]
		\centering
		\subfigure[]{\scalebox{0.4}{\includegraphics{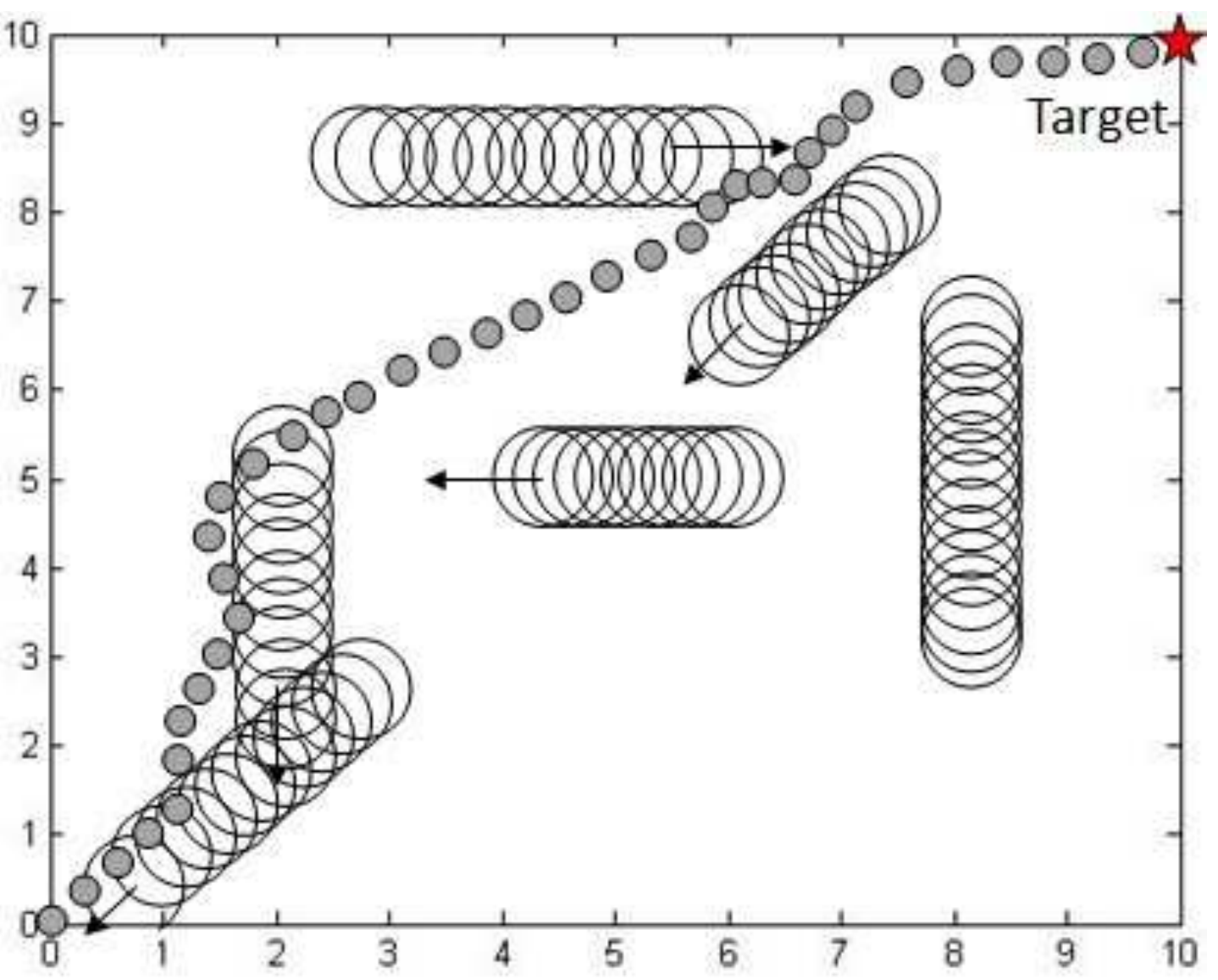}}
		\label{c5.sim71}}
		\subfigure[]{\scalebox{0.4}{\includegraphics{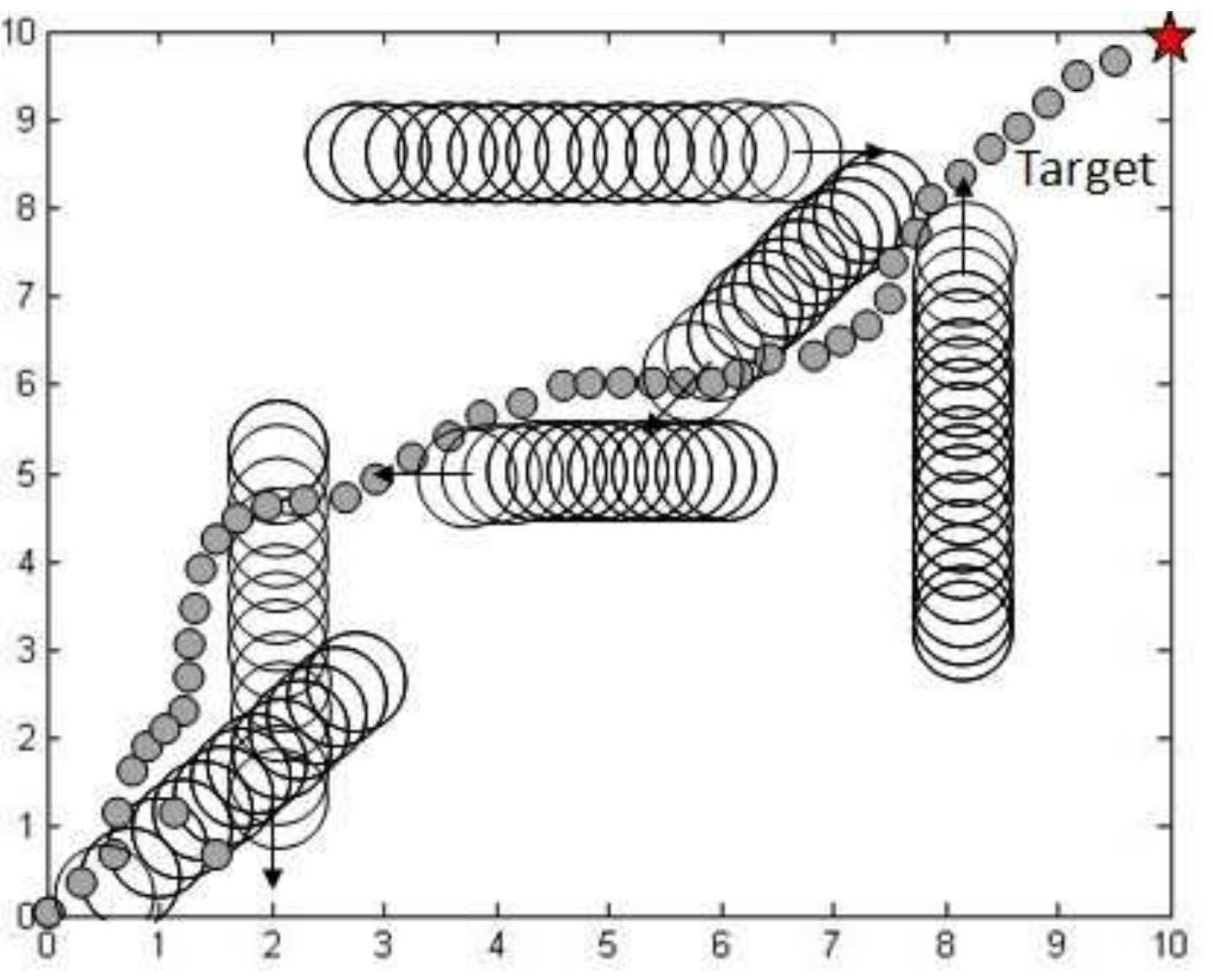}}
		\label{c5.sim72}}
		\subfigure[]{\scalebox{0.4}{\includegraphics{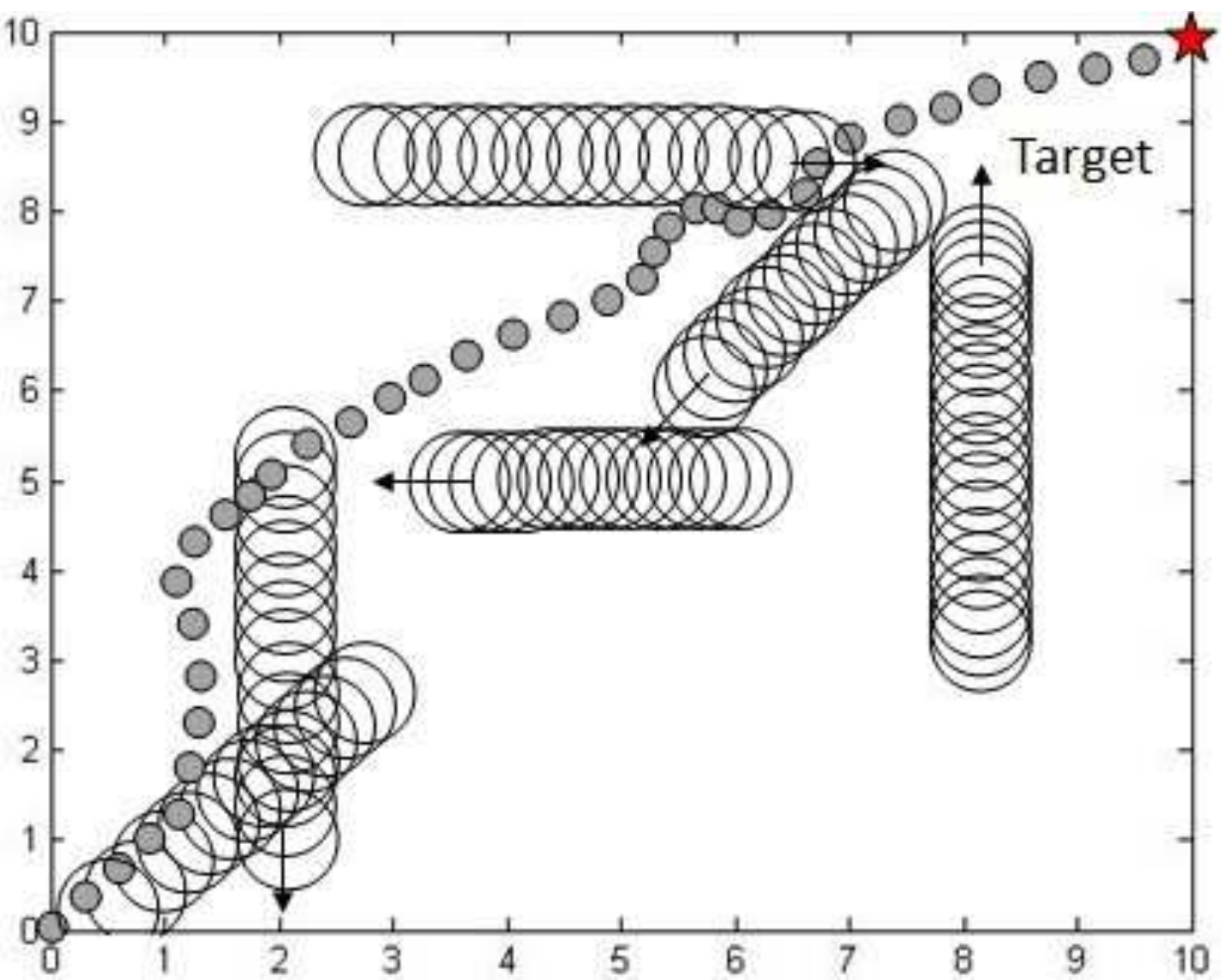}}
		\label{c5.sim73}}
		\caption{Performance comparison: dynamic environment with six moving obstacles for (a) BINA (b)ENA (C)NAIER}
		\label{c5.sim7}
		\end{figure}

			We first examine the performance of the proposed navigation algorithms in the dynamic environments with obstacles moving at constant speeds. In Fig.~\ref{c5.sim5}, Fig.~\ref{c5.sim6} and Fig.~\ref{c5.sim7}, we gradually increase the number of obstacles within the environments. The obstacles are moving with constant speeds at random directions. These simulations show that the proposed navigation algorithms are capable of avoiding moving obstacles and their performance is consistent regardless the number of obstacles.
\par
			Statistically, BINA has the best performance over ENA and NAIER in terms of overall navigation time. Table~\ref{Nav_time} shows the record of navigation time for all three algorithms over $25$ simulation runs. The number of the obstacles is different and also the positions, speeds and moving direction of the obstacles are randomly assigned for all the simulation runs. The results in Table~\ref{Nav_time} shows that BINA has the shortest navigation time for most of the experiments ($23$ out $25$, $92\%$ rates).
\par

		\begin{table}[!h]
		\caption{Navigation time (in $sec$) for BINA, ENA and NAIER over 25 experiment runs (obstacles moving with constant speeds)}
		\centering
		\begin{tabular}{ccccc}
		\hline
				      & Result for BINA  & Result for  ENA & Result for NAIER & Best result\\ 
		\hline
		Experiment 1 & 26.1 & 28.3  &  29.5 &BINA  \\
		Experiment 2 & 23.3 & 24.4  &  27.3 &BINA \\
		Experiment 3 & 30.2 & 32.6  &  31.1  &BINA\\
		Experiment 4 & 35.8 & 38.9  &  40.1  &BINA\\
		Experiment 5 & 22.5 & 26.3  & 26.0   &BINA \\
		Experiment 6 & 27.9 & 29.9  &  28.3  &BINA  \\
		Experiment 7 & 38.8 & 40.1  &  39.9  &BINA  \\
		Experiment 8 & 24.2 &  28.7 &  26.2  &BINA \\
		Experiment 9 & 28.2&   27.1 &  28.9 &ENA \\
		Experiment 10 & 30.1 &  35.0  & 36.2 &BINA\\
		Experiment 11 & 40.1 & 46.3	 & 43.5 &BINA\\
		Experiment 12	& 27.5	& 28.7 & 32.6	&BINA\\                                                                            
		Experiment 13	& 34.1	& 40.2	&38.5	&BINA\\
		Experiment 14	&33.9	& 38.1	& 32.1	&BINA\\
		Experiment 15	&39.1	& 45.8 & 47.2	&BINA\\
		Experiment 16	&28.8	& 30.2 & 32.1	&BINA\\
		Experiment 17	& 26.9 & 30.2	& 27.8	&BINA\\
		Experiment 18	&28.4 & 30.5& 32.9	&BINA\\
		Experiment 19	&29.9	& 28.1& 28.3 &BINA\\
		Experiment 20	&38.2	& 42.1&35.1	& NAIER\\
		Experiment 21 & 32.6 & 35.3 & 34.1&BINA\\
		Experiment 22&  34.1 & 36.3 & 38.8&BINA\\
		Experiment 23& 28.5 & 32.4 & 29.1&BINA\\
		Experiment 24& 29.7 & 31.3 & 32.2&BINA\\
		Experiment 25& 23.9 & 27.3 & 29.3&BINA\\
		\hline
		\label {Nav_time}	
		\end{tabular}	
		\end{table}

		We can not always expect the velocities of the obstacles in environments to be constant in every of the real life scenarios. Obstacles such as many animals (birds, fish, cats etc) are likely to be moving in a more complicated non-linear velocities. The movements of these obstacles are even more difficult to predict or estimate.   
\par
		In the next simulation, the performances of the proposed navigation algorithms against obstacles with non-linear velocities are shown. We assume that the maximum speed and the maximum angular velocities of the obstacles are strictly less than those of the robot, i.e., $\|V_{ob}\|<\|V_{robot}\|$, $\|\omega_{ob}\|<\|\omega_{robot}\|$.  All three proposed navigation algorithms are able to avoid obstacles with non-linear velocities with their own strategies: BINA guides the robot so that it keeps a constant avoiding angle between the instantaneous moving direction of the obstacle. In this case the moving direction of the obstacle is time-varying. ENA drives the robot to the $d_0$-equidistant curve around the obstacle. NAIER always direct the robots to a collision free path by assessing the angular segments within the sensing range. In this particular case, BINA has the shortest navigation time of all three navigation algorithms. However, Table~\ref{Nav_non} shows the record of navigation time for all three algorithms over $20$ simulation runs. The obstacle is assigned with different non-linear velocity and initial position in each simulation run. There is no solid conclusion to which algorithm is more efficient for avoiding obstacle with non-linear velocity. 
\par
		\begin{figure}[!h]
		\centering
		\subfigure[]{\scalebox{0.4}{\includegraphics{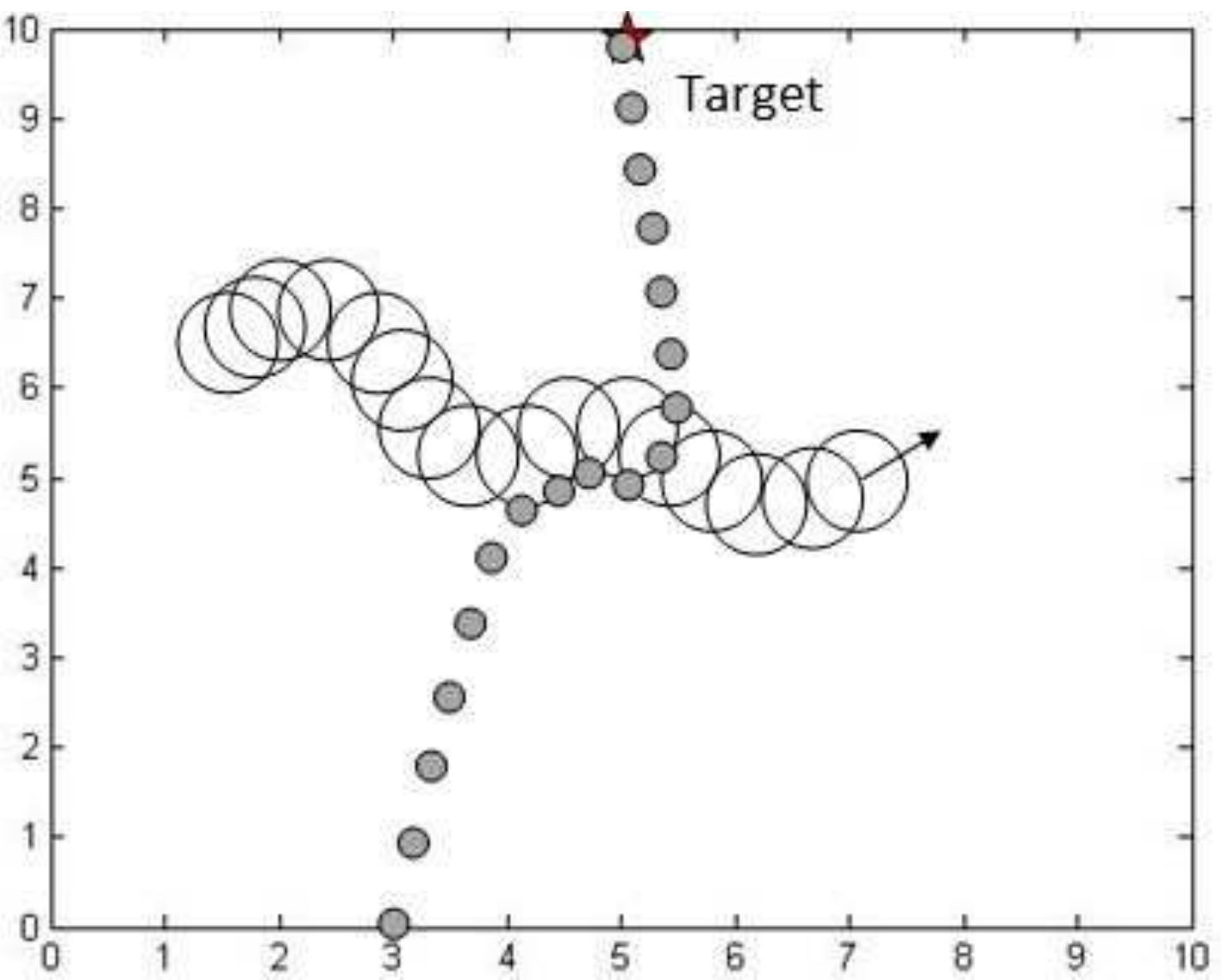}}
		\label{c5.sim81}}
		\subfigure[]{\scalebox{0.4}{\includegraphics{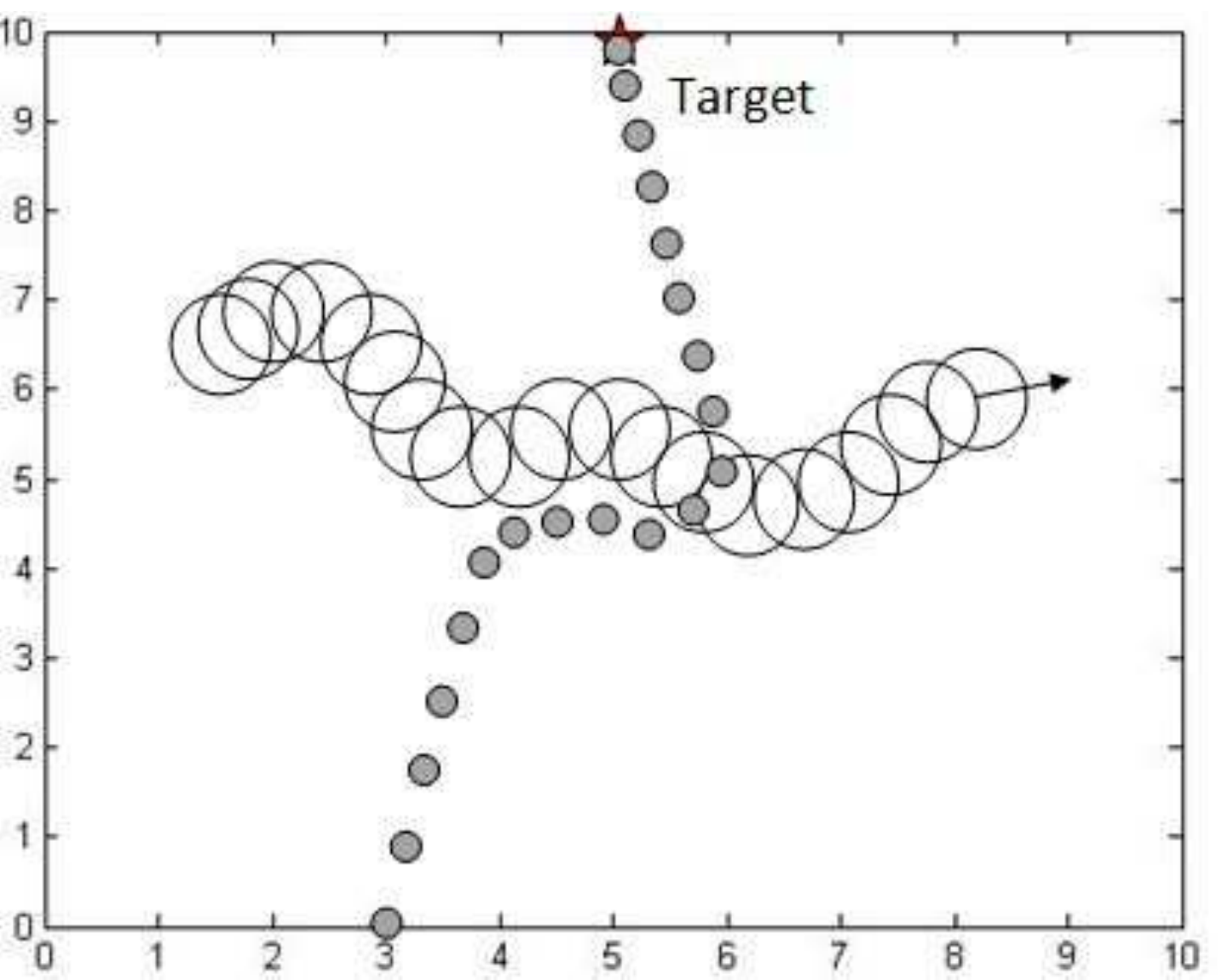}}
		\label{c5.sim82}}
		\subfigure[]{\scalebox{0.4}{\includegraphics{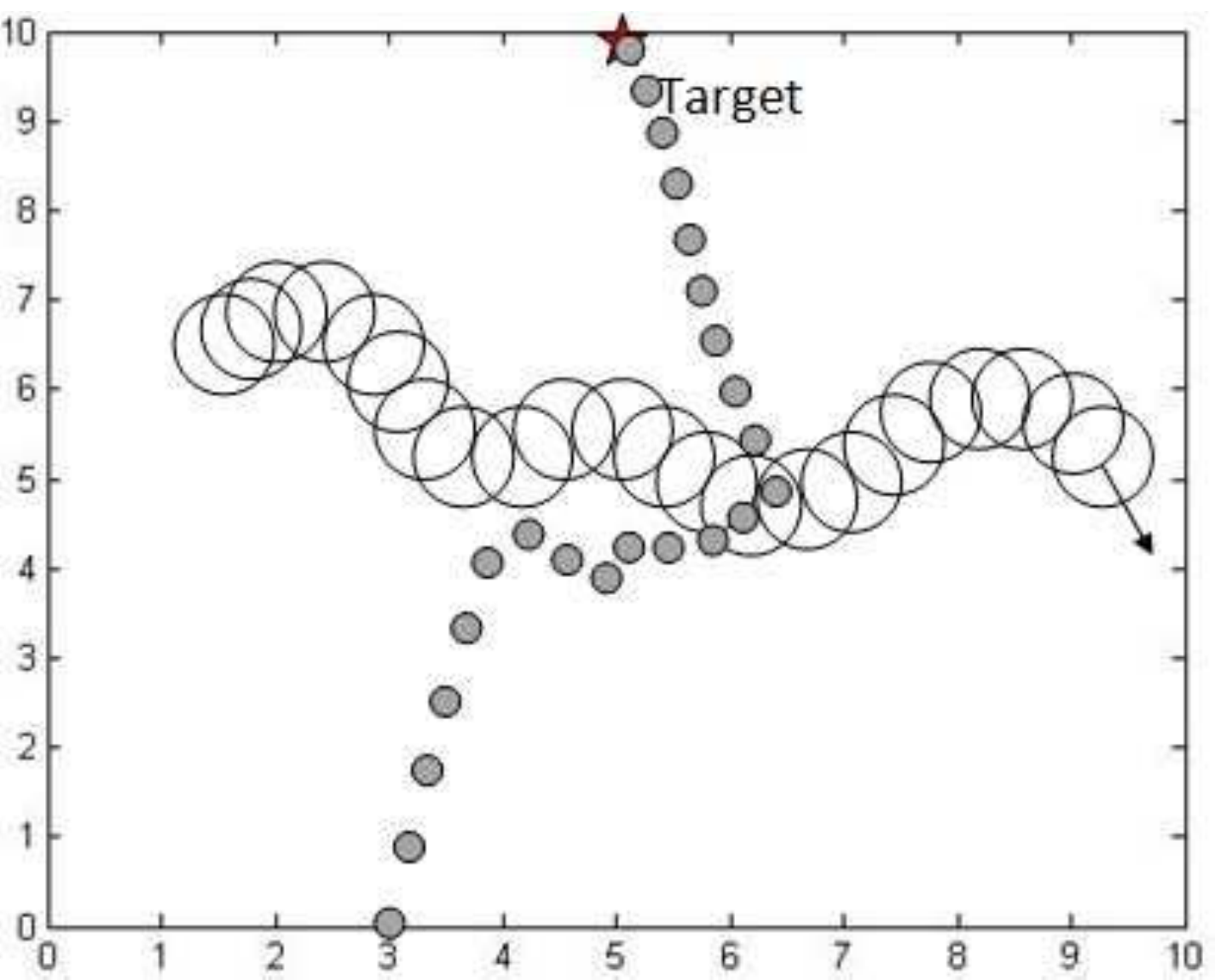}}
		\label{c5.sim83}}
		\caption{Performance comparison: obstacle with non-linear velocity for (a) BINA (b)ENA (C)NAIER}
		\label{c5.sim8}
		\end{figure}

		\begin{table}[!h]
		\caption{Navigation time (in $sec$) for BINA, ENA and NAIER over 20 experiment runs (obstacles moving with non-linear velocities)}
		\centering
		\begin{tabular}{ccccc}
		\hline
				      & Result for BINA & Result for  ENA & Result for NAIER & Best result\\  
		\hline
		Experiment 1 & 18.4 & 15.3  &  17.5 & ENA \\  
		Experiment 2 & 20.1 & 26.4  &  17.3 & NAIER\\  
		Experiment 3 & 19.2 & 22.9  &  20.1 & BINA\\  
		Experiment 4 & 18.8 & 22.6  &  16.5 & NAIER\\  
		Experiment 5 & 25.5 & 27.2  &  29.4  & BINA \\ 
		Experiment 6 & 21.8 & 27.3  &  22.3  & BINA\\ 
		Experiment 7 & 26.8 & 22.5  &  29.2  & ENA\\ 
		Experiment 8 & 25.2 &  29.3 &  21.7  & NAIER \\ 
		Experiment 9 & 25.8&   24.3 &  20.5 & NAIER\\ 
		Experiment 10 & 20.9 &  27.3  & 29.3 & BINA\\
		Experiment 11 & 30.1 & 24.5	 & 28.5 & ENA\\ 
		Experiment 12	& 18.5	& 25.9 & 29.6	& BINA\\                                                                           
		Experiment 13	& 23.4	& 24.7	&18.5	& NAIER\\
		Experiment 14	&27.7	& 29.6	& 21.1	& NAIER\\ 
		Experiment 15	&29.9	& 25.3 & 28.9	& ENA\\ 
		Experiment 16	&21.8	& 32.8 & 28.3	& BINA\\
		Experiment 17	& 28.0 & 21.8	& 29.9	& ENA\\
		Experiment 18	&20.4 & 28.0& 26.7	& BINA\\
		Experiment 19	&19.2	& 23.9& 29.2& BINA\\ 
		Experiment 20	&28.4	& 23.9& 25.7& ENA\\
		\hline
		\label {Nav_non}	
		\end{tabular}	
		\end{table}

\begin{figure}[!h]
		\centering
		\subfigure[]{\scalebox{0.4}{\includegraphics{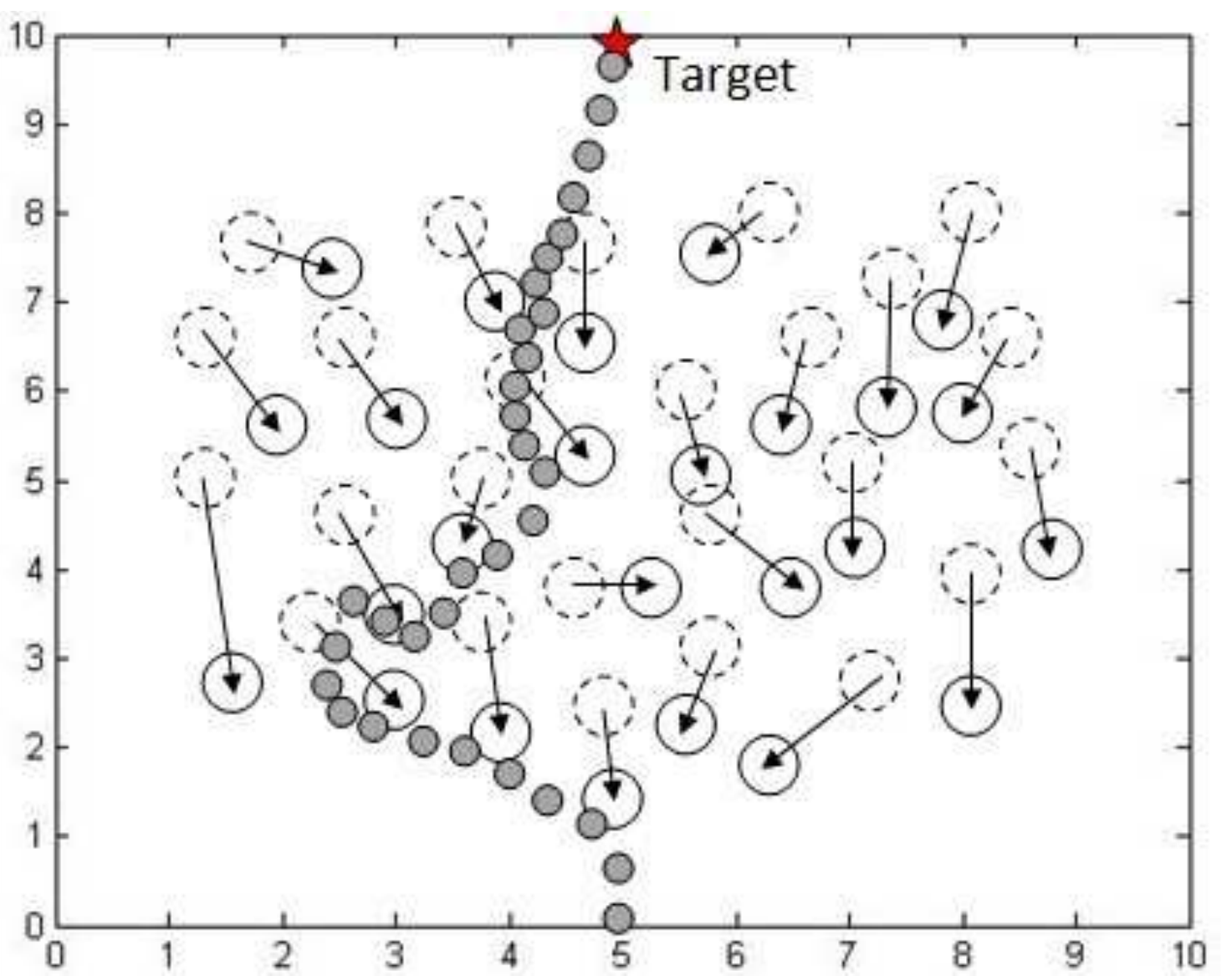}}
		\label{c5.sim91}}
		\subfigure[]{\scalebox{0.4}{\includegraphics{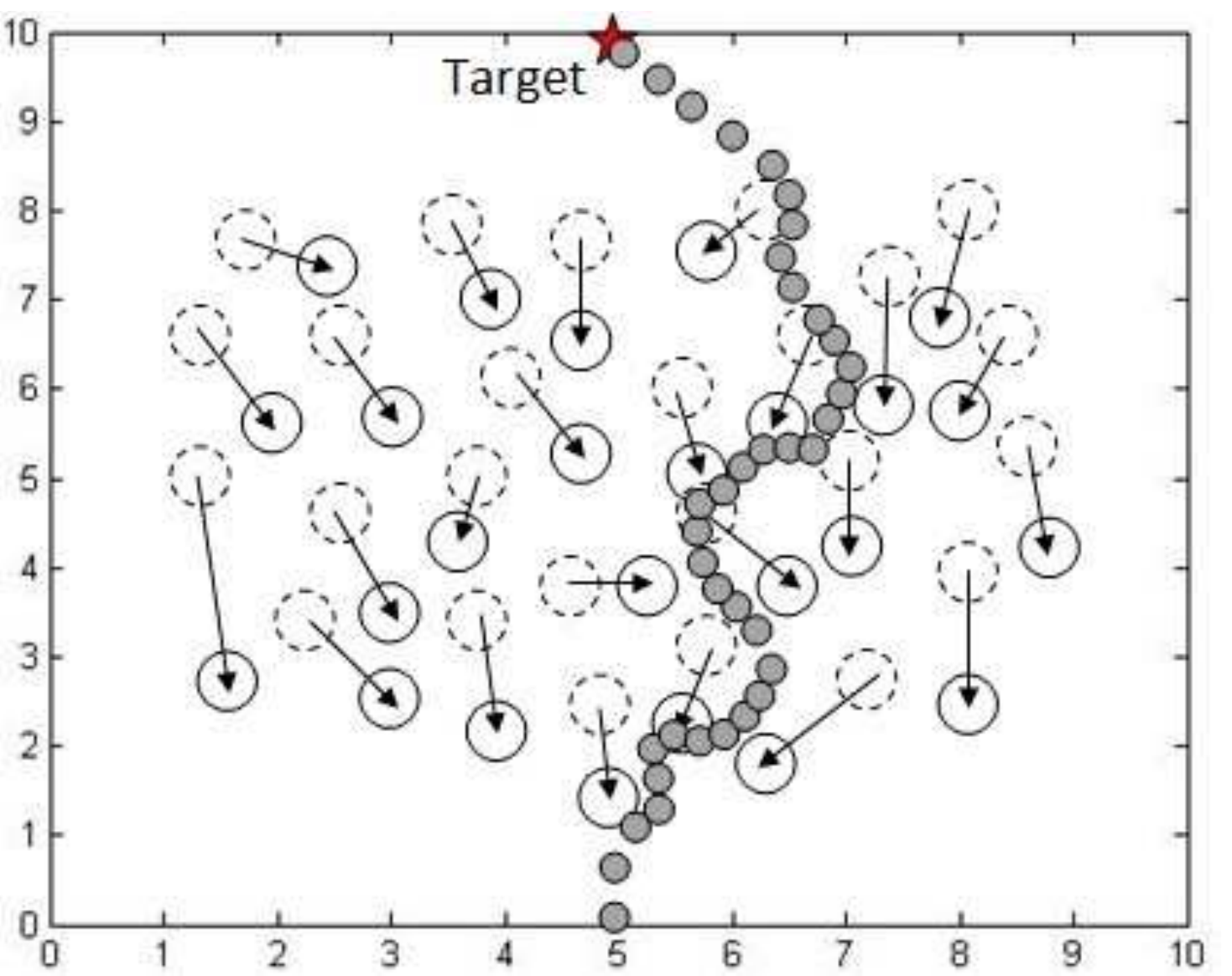}}
		\label{c5.sim92}}
		\subfigure[]{\scalebox{0.4}{\includegraphics{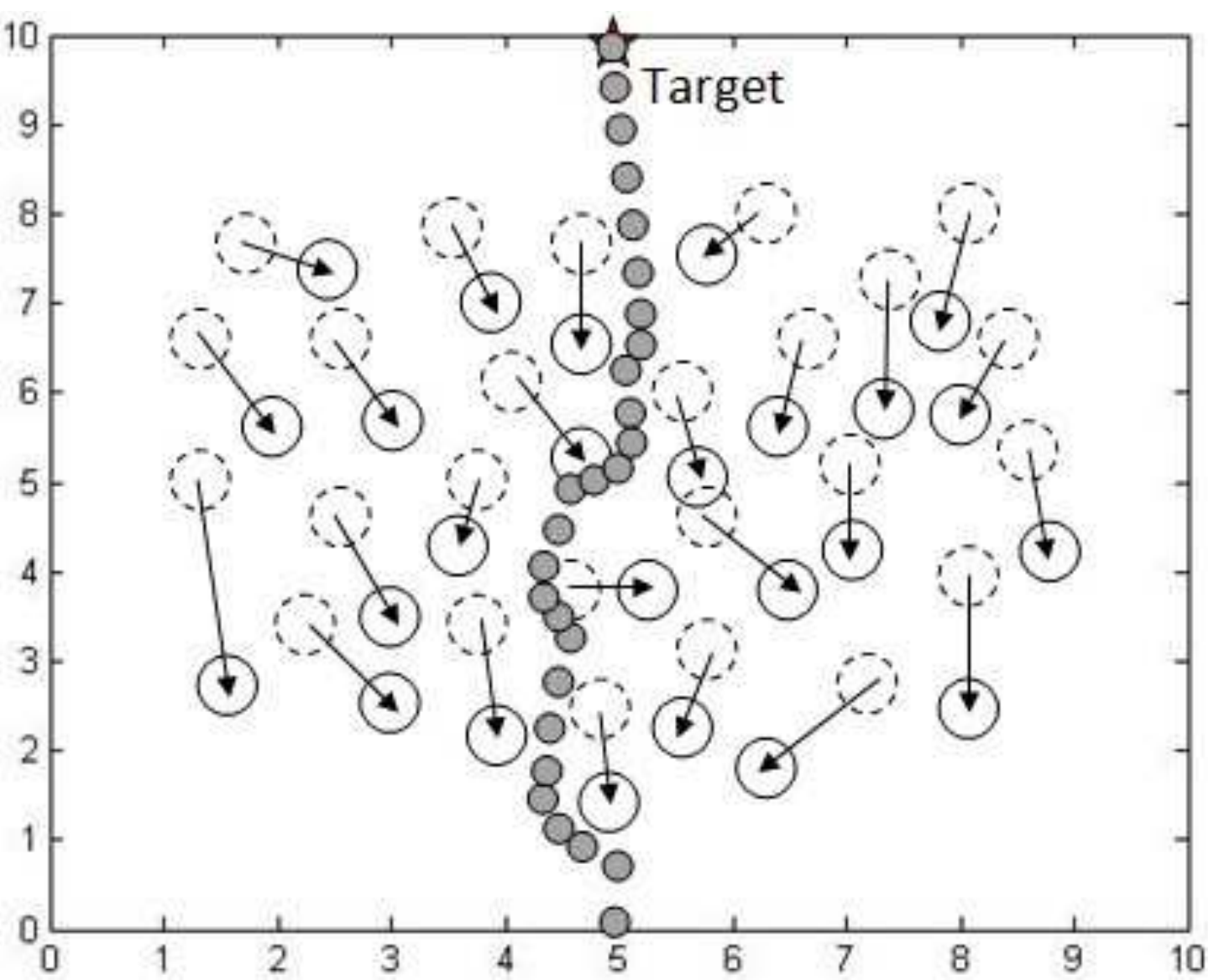}}
		\label{c5.sim93}}
		\caption{Performance comparison: navigation in a extremely cluttered dynamic environments for (a) BINA (b)ENA (C)NAIER}
		\label{c5.sim9}
		\end{figure}

		Finally, we compare the performance of three proposed navigation algorithm in an extremely cluttered environments. Although it is demonstrated that the proposed algorithms are capable of avoiding dynamic obstacles, the performance of the algorithms may varies when the obstacles are extremely cluttered within the same environments, such as passengers in train stations at busy hours. We slightly change the presentation of the result in order to make it clear and easy to understand. In Fig.~\ref{c5.sim9}, the initial position of the obstacle is shown by a dashed circle and moving direction of the obstacle is depicted by an arrow connecting its initial position and final position.
\par
		As we can see from the results in Fig.~\ref{c5.sim9}, NAIER has the best performance over the other two algorithms. The advantage of NAIER is that it is able to find the collision-free angular sectors in the sensing range and steer the moving direction of the robot to the middle of the vacant angular sector. It allows the robot to find a safe path through the cluttered environments. BINA and ENA are also able to guide the robot to the target location while ensuring the safety of the robot in this cluttered environment. However, the employment of enlarged vision cone for BINA often force the robot to a detour when the obstacles are closely positioned, see e.g. the first three obstacles the robot avoids in Fig.~\ref{c5.sim91}. The interpolation technique is put in use for ENA when the obstacles are cluttered, which enlarges the $d_0$-equidistant curve of this group of obstacles for the robot to follow.
\par

		\subsection{Section Summary}

			In this section, we demonstrate that the proposed navigation algorithms are capable of navigating a robot in various scenarios. The simulation results are presented to show the characteristics of the algorithms. BINA has the most efficient navigation time when facing obstacles of regular shapes (circle or regular polygons) in both stationary and dynamic environments. ENA is applicable in a large variety of scenarios and is efficient when avoiding obstacles of irregular shapes. ENA has the average and most stable navigation time in all three algorithms. Finally, the advantage of NAIER is its superior efficiency in extremely cluttered stationary and dynamic  environments, which the other two algorithms take relative longer to deal with.

		\section {Summary}

			In this chapter, we present the comparison between BINA, ENA and NAIER in various aspects including the measurements required by the algorithms which directly influence the implementation cost and the design of the control system, the computation complexity which affects the real time performance of the algorithm and finally the performance comparison between the algorithm in various static and dynamic environments.

\chapter {The Implementations on an Intelligent Wheelchair Control System} \label{C6}

	The fundamental concern for operating a wheelchair is the safety of the user. Conventional wheelchairs can be operate manually, or with the aid of joysticks, levers and other accessories, to transport an impaired person from current location to target location. More advanced intelligent wheelchairs allows the users to control the movements of the wheelchairs by their own efforts, i.e., user-wheelchair interactions. These wheelchair control approaches requires a fair amount of training and practice and their performance can be easily plagued by many external factors. 
\par
	The implementations of the navigation algorithms on electric-power wheelchairs solve these problems and guarantee the safety of the users. In this chapter, the navigation algorithms, Biologically-Inspired Navigation Approach (BINA) and Equidistant Navigation Approach (ENA) which are proposed in Chapter ~\ref{C2} and Chapter ~\ref{C3} respectively,  are implementation on a real intelligent autonomous wheelchair SAM (Semi-Autonomous Machine). The applicability and performance of the intelligent autonomous wheelchair under the guidance of these navigation algorithm are shown by experiments in real life scenarios

	\section {Background and Motivation}

		The population of the impaired persons was growing over the past decades \cite{ABS92,MLJ96}. Numerous people with mobility impairments are experiencing problems such as losing social connections (which leads to social isolation, depression and anxiety), emotional damage (which leads to fear, loss of self-esteem)  \cite{MFT03,LIL01} etc. Studies also show that the utilization of independent mobility aids help the user to ease the problem and significantly improve his/her lifestyle  \cite{SK99,RCS05}. Wheelchairs are one of the most commonly used mobility aids to assist the movement of user \cite{JNR94, MFT03} and a large number of users are benefited from it \cite{RCSE08}. 
\par
		The existing approaches to the safe navigation of wheelchairs can be generally classified into two categories: the user-wheelchair interaction approach and the autonomous navigation approach.
\par
		The user-wheelchair interaction approaches control the movements of the wheelchair by giving "commands" to the wheelchair by different parts of body. e.g. hand gesture \cite{YZJZYL11,TLKDYH10}, head position and movement \cite{ZFHLL10, SRBRIY09,LWHHKY09}, oral motion \cite{STTNKR07} etc. voice vommand \cite{QMT09,SRCLSP02,KKMKK00}, brain signal \cite{RBBE2006,SARC11,BGSTK10}, etc. These approaches give the maximum control freedom to the users, allowing them to control the wheelchairs according to their judgments and efforts. However, this also raises several problem because the information about the environments is acquired by the users' reaction. These perception and reaction can be easily plagued by many factors such as bad perceptional condition (tiredness, stress) or poor operating environment (darkness, noise). Inaccurate information about the environment can dramatically increase the chances of misjudgments, which lead to inappropriate decisions, and so the safety of the users is not guaranteed. These approaches also require the users to stay active and concentrates for any potential dangers during the whole navigation process  which is fatiguesome. Furthermore, a clinical survey shows that forty percent of the users found it difficult to perform steer tasks using these intelligent wheelchairs, and a number of users cannot operate the intelligent wheelchairs due to various reasons \cite{LFE00}. This is an indication that most of user-wheelchair interaction methods cannot be learned easily and require a fair amount of practice and training in order to properly operate the wheelchair.
\par
		The autonomous navigation approaches, on the other hand, take over the control of the wheelchair from the users. The advantages of these approaches are that the wheelchairs can be operated in poor perceptional environments and responses to the changes in the environments are generally faster. Moreover, it does not require any practice and training to operate the wheelchairs. A number of navigation algorithms have been implemented in various intelligent wheelchairs. The Vector Field Histogram (VFH) \cite{BJKY91} and Vector Force Field \cite{BJK91} are implemented in the NavChair Assistive Wheelchair navigation system, the obstacle avoidance strategy based on optimized Bayesian Neural Networks \cite{THT08} and shared control strategy \cite{TH08} are implemented in SAM (semi-Autonomous Machine). A biologically inspired approach \cite{TS09}, which was implemented with a mobile robot and which performance was confirmed by extensive real world experiments, can also be applied to  intelligent wheelchairs. These navigation algorithms have demonstrated great successes in static environments. Furthermore, the velocity obstacle approach (VO) is implemented in a commercial wheelchair model SPRINT in \cite{EPJSPF01,Epjs99}, the VO approach allows the SPRINT wheelchair to cope with environments with moving obstacles.
\par
		These autonomous navigation approaches only deal with static environments which is not relevant to many real world scenarios, where the users of the wheelchairs often find themselves involved in dynamic crowded environments with multiple moving obstacles. The proposed navigation algorithms ensures the safety of the users of the wheelchairs in both static and dynamic environments, the performance of the proposed navigation algorithms BINA and ENA has been confirmed by the simulation results and the experiments with a real non-holonomic mobile robot in Chapter ~\ref{C2} and Chapter ~\ref{C3}. The intelligent wheelchair SPRINT  \cite{EPJSPF01,Epjs99} is a rare example of a wheelchair that can operate in dynamic environments, Both BINA and ENA have their own advantages over SPRINT when they are implemented on SAM. The experiment results of the implementation of BINA and ENA on SAM are presented in this chapter.

	\section {System Model}

		We consider a wheelchair that travels in a plane and has two independently actuated driving wheels mounted on the same axle and castor wheels.
The position of the wheelchair is represented by the absolute Cartesian coordinates $x,y$ of the reference point located at the center of the axle, whereas its orientation is given by the angle $\theta$ between the wheelchair centerline  and the abscissa axis. The driving wheels roll without sliding.
The wheelchair is controlled by the angular velocities $\omega_l$ and $\omega_r$ of the left and right driving wheels, respectively, which are limited by a common and given constant $\Omega$. The relevant mathematical model of kinematics of the wheelchair is as follows:
		\begin{equation}
		\label{ch6:1}
		\begin{array}{l}
		\dot{x} = v \cos \theta,
		\\
		\dot{y} = v \sin \theta,
		\\
		\dot{\theta} = u,
			\end{array}\;
		\begin{array}{l}
		v = \frac{v_l+v_r}{2},
		\\
		 u = \frac{v_r-v_l}{2L},
		\\
		 v_i = R_w \omega_i,
		\end{array}
		\;
		\begin{array}{l}
		 x(0) = x_0
		\\
		 y(0) = y_0
		\\
		 \theta(0) = \theta_0
		\end{array}
		\end{equation}
		where $R_w$ is the radius of the driving wheels, $2L$ is the length of the axle, and
$\omega_i = \omega_i(t) \in [-\Omega,\Omega], i=l,r$. To simplify the matters, we treat $v$ and $u$ as control variables. They uniquely determine the rotational velocities $\omega_r=(v+Lu)/R_w, \omega_l=(v-Lu)/R_w $ and obey the bound:
		\begin{equation}
		\label{control.constr}
		|v|+L|u| \leq V:= R_w \Omega.
		\end{equation}
		This bound implies restrictions on the forward and rotational movements of the wheelchair. In particular, its speed cannot exceed $V$, and for given $v \in (-V,V)$, the turning radius of the wheelchair is bounded from below 
		\begin{equation}
		\label{mini}
		 R = \frac{L|v|}{V-|v|}.
		\end{equation}
		\par	
	\section {Wheelchair System Description}

		The algorithms BINA and ENA are implemented on the wheelchair SAM. Motions of this wheelchair can be controlled by many ways such as  joystick, EEG signals \cite{CDA07}, head movement \cite{THT08} etc. A number of control methods have been proposed and implemented on the SAM wheelchair, see e.g.\cite{NST06,TH08,TN11,NN08}.
\par
\begin{figure}[h]
		\centering
		\subfigure[]{\scalebox{0.28}{\includegraphics{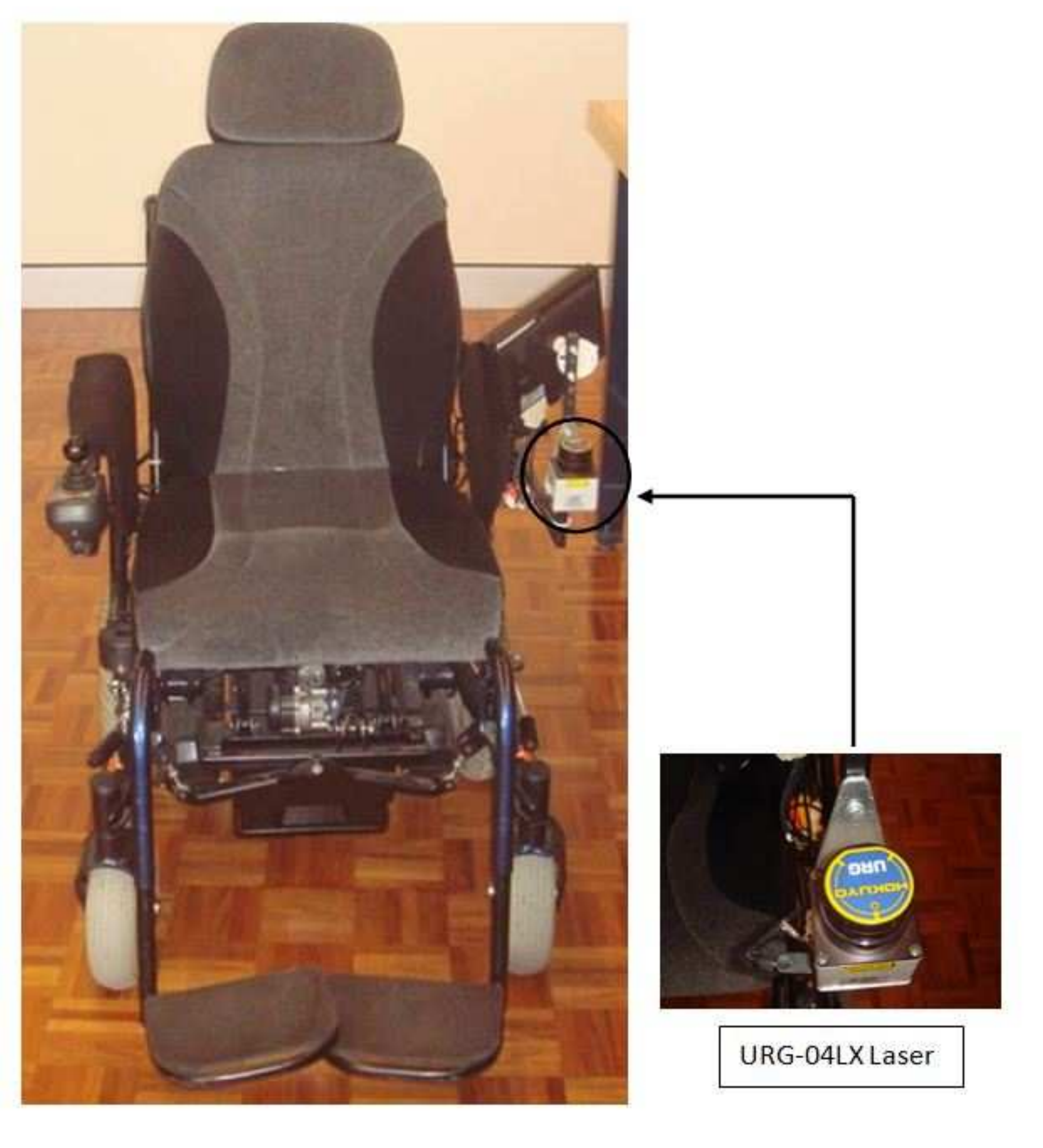}}
		\label{front.fig}}
		\hfill
		\subfigure[]{\scalebox{0.28}{\includegraphics{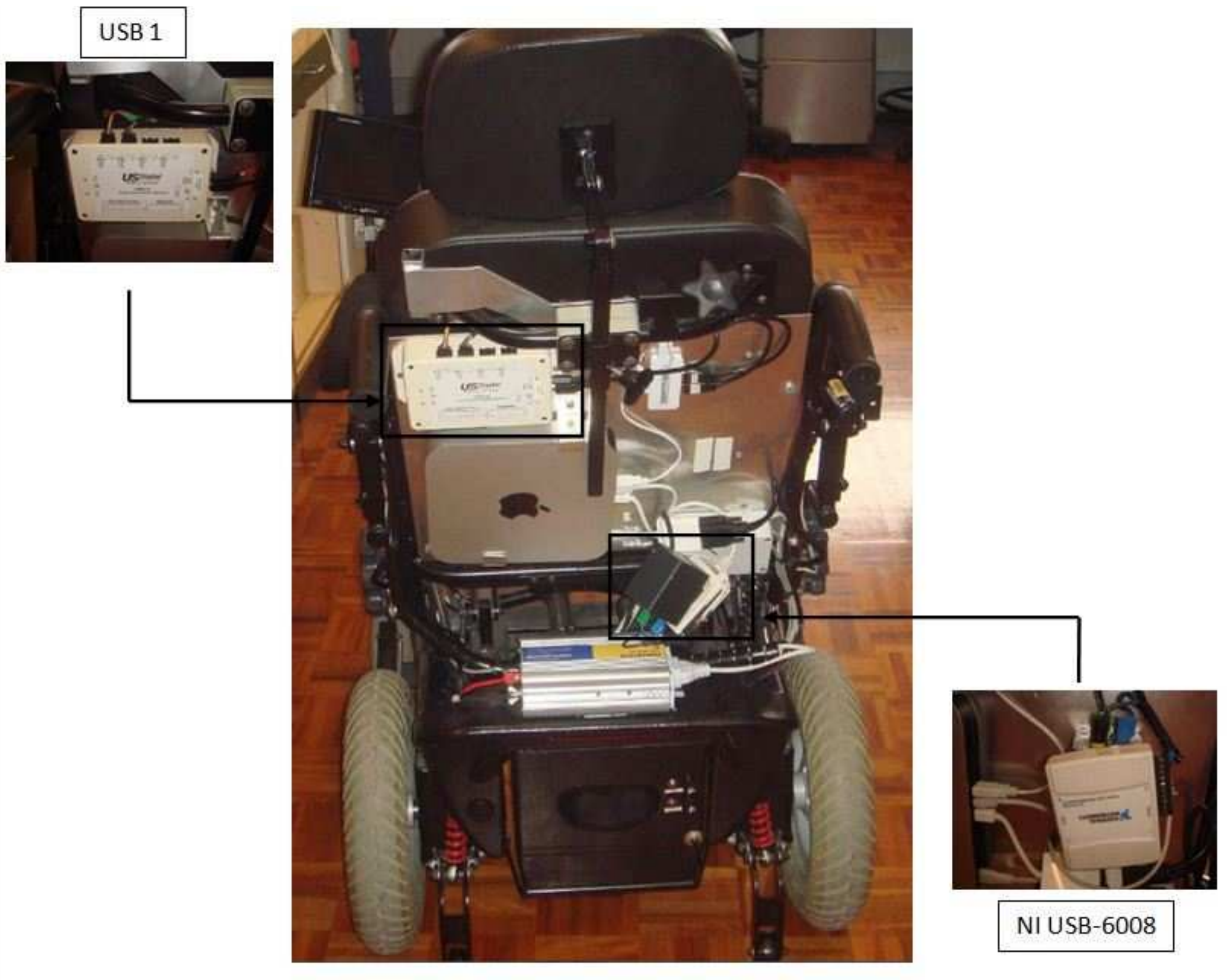}}
		\label{back.fig}}
		\caption{(a)Front view of the wheelchair; (b)Rear view of the wheelchair}
		\end{figure}

		The SAM wheelchair features two rear driving wheels (see Fig.~\ref{front.fig}) and two front caster wheels (see Fig.~\ref{back.fig}). The motor for driving the wheels is powered by a 24V battery. The joystick mounted on the right rack of the wheelchair at the font of the wheelchair which can be used to control the motion of the wheelchair. The LDC monitor is mounted on the left rack, this monitor is used to display information such as the map of the environment, the GPS information or the laser data. There is also a laser range finder on the left rack which is the main interface between the wheelchair to the environment. An on-board computer is available at the back of the wheelchair for those programs running in Linux environment. The DC/AC are responsible to convert the control signals and the data into the correct format.
\par
		The following devices are particular important for our experiments, therefore their functionalities and specifications should be emphsised:
	
		\begin{itemize}

			\item A notebook (Windows XP operating system with dual-core CPU running at 1.66GHz and 2GB RAM). For these experiments, LabWindows are used to program the algorithms. The notebook act as the core of the entire intelligent wheelchair control system, it receives information from various devices and computes the control signals, which is then sent to the driving system.
			\item The URG-04LX laser is mounted on the left rack at the front of the wheelchair. The laser scans the environment and provides necessary information to the wheelchair such as distance to the obstacles (required by both BINA and ENA), the vision cone (required by BINA) etc.  The maximum scan angle for this laser range finder is $240 degree$ and the maximum scan range is $4m$ with accuracy of $\pm 1\%$ of the measurement, the scan frequency is $10 Hz$. Furthermore, the laser is able to scan and provide two-dimensional map of the environment if necessary.
			\item The USB1 adapter (By USDigital). It is an interface between the computer and the encoders which are attached to the driving wheels. The adapter interprets the information from the encoders, convert them into the revolution of both wheels. This information is used to estimate the position and orientation of the wheelchair.
			\item  The National Instrument NI USB-6008 DAQ device. This DAQ devices converts control signals into appropriate voltages and sent them to the motor of the wheelchair. 

		\end{itemize}

\begin{figure}[h]
		\centering
		\includegraphics[width=4.5in]{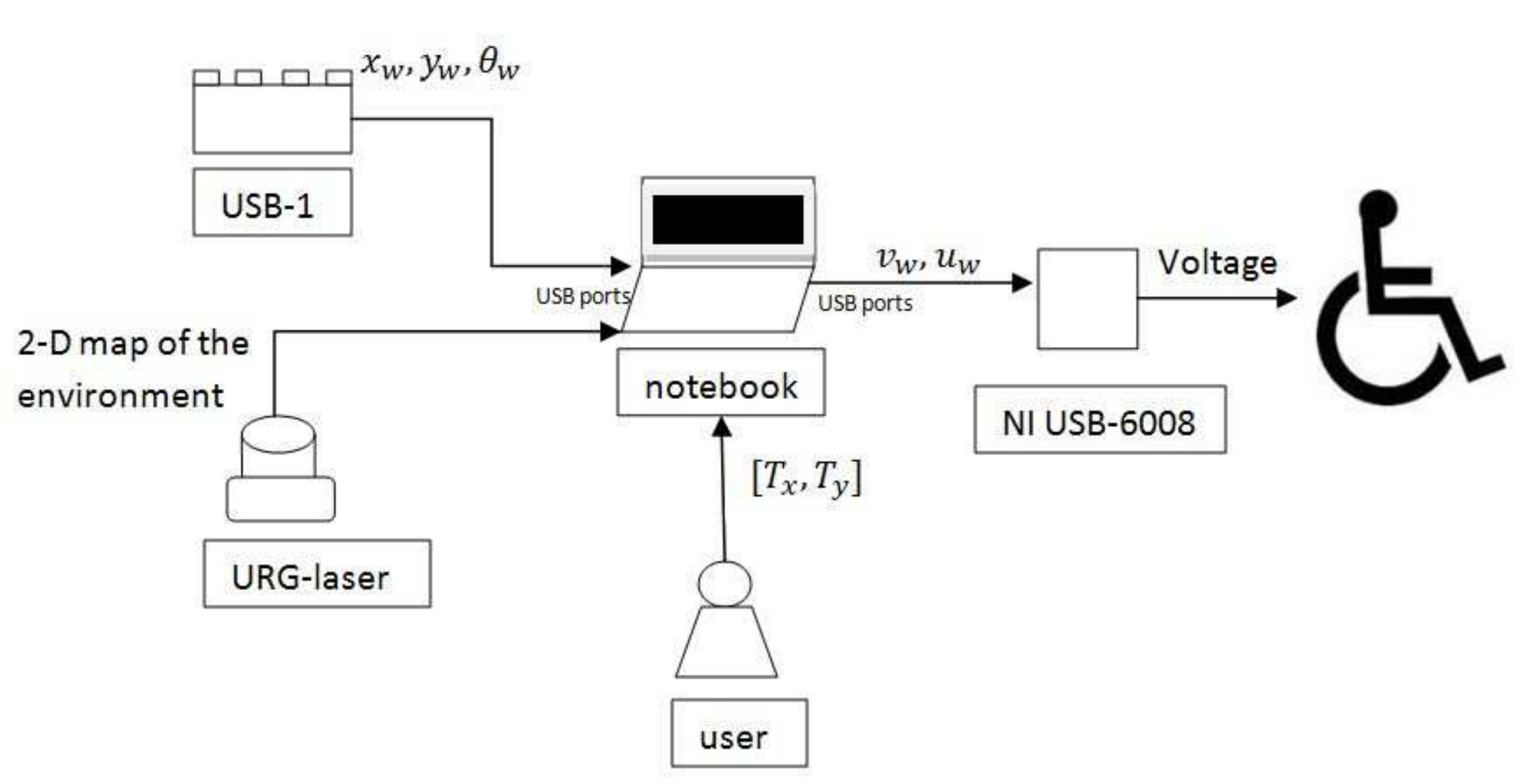}
		\caption{Interconnections between hardwares of wheelchair control system}
		\label{inter}
		\end{figure}

		The interconnections between these hardwares of SAM wheelchair is depicted in Fig.~\ref{inter}. Note that the variables $T_x$ and $T_y$ are the coordinates of the target location. 
		\par

	\section {Implementation and Experiments of Biologically-Inspired Navigation Algorithm (BINA)}

		\subsection{Parameter Measurements} \label{para_meas1}

			In this section, we describe the methodology to acquire the essential parameters required by BINA, which is presented in Section~\ref{PD1}.

			\begin {itemize}

				\item The measurements to the environments are carried out by the URG-04LX laser range finder. The laser returns the measurements  as an array of data. Each of the indices shows the distance to the closest obstacle at a certain angle. If there is no obstacle within the sensing range of the laser, it will return the maximum scan range of the laser ($4m$). Table~\ref{data} shows a fragment of data returned by the laser and the graphic illustration of the data is depicted in Fig.~\ref{c6.graphic}. This data can be displayed on the LCD monitor of the wheelchair.
\par
				\begin{table}[h]
					\caption {Fragment of data from URG-04LX laser range finder}
					
				\begin{tabular}{l |l |l |l |l |l |l |l |l |l |l |l |l |l |l |l | l }				
					\hline
					index & ... & 80 & 81 & 82 & 83 & 84 & 85 & 86 & 87 & 88 & 89 & 90 & 91 &  ...\\
					\hline
			              $dis$(m) & ... & 4.0 & 4.0 & 2.8 & 2.6 & 2.4 &  2.2  & 2.2 & 2.2 & 2.6 & 2.8 & 4.0 & 4.0 &  ...    \\
					\hline 
				\end{tabular}
				\label {data}
				\end{table}

				\begin{figure}[h]
				\centering
				\includegraphics[width=4.5in]{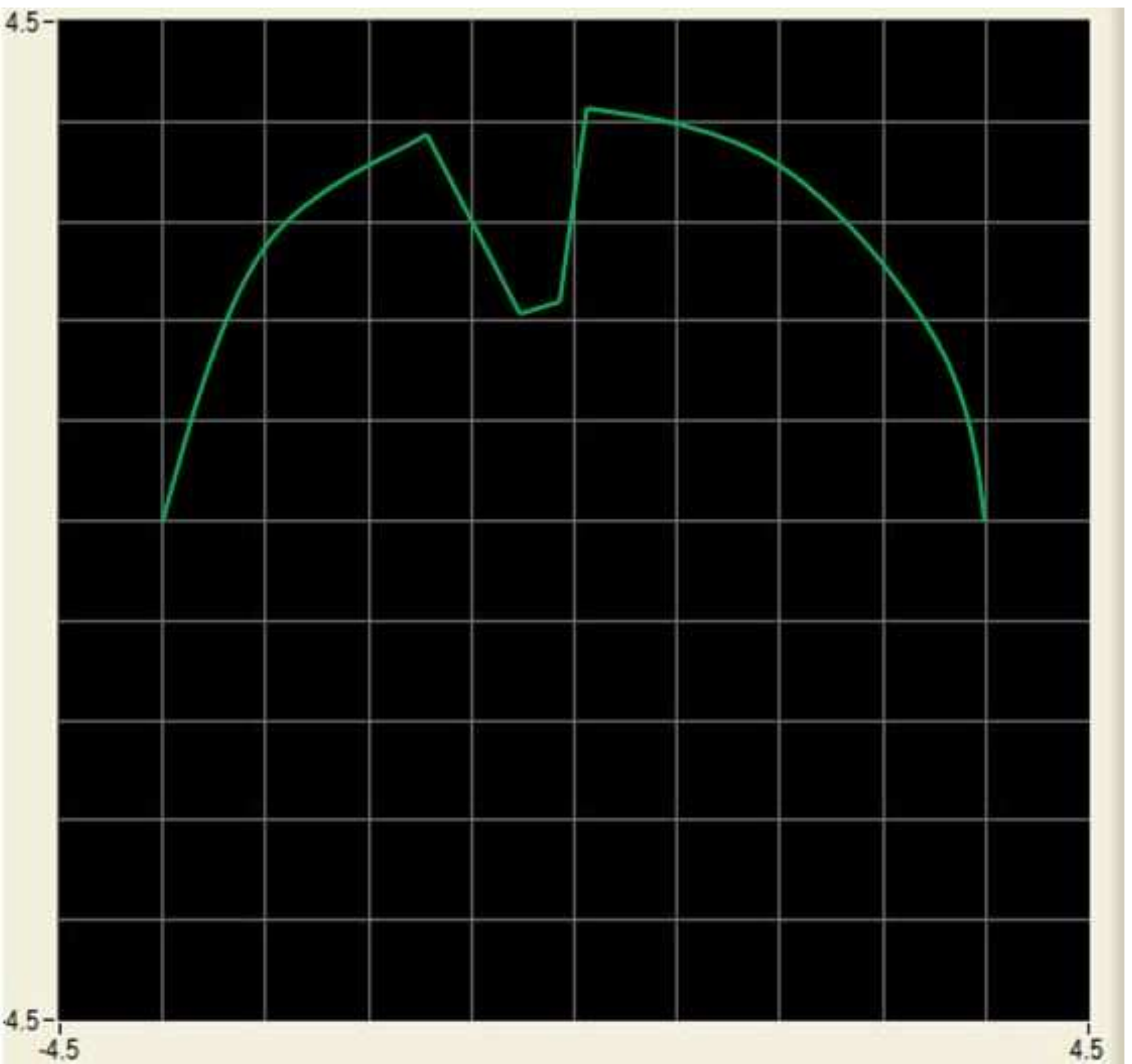}
				\caption{Graphic Illustration of the received data}
				\label{c6.graphic}
				\end{figure}

				The minimum distance $d_i(t)$ and the vision cone ($\alpha_i^{(1)}$ and ($\alpha_i^{2}$)) of the obstacle can be found from the array of data returned from laser range finder. The minimum distance to the obstacle $i$ is simply the minimum values in the array.  The vision cone of the obstacle $i$ can be found by searching for the "start" index and the "end" index for the distance less than the maximum sensing range of the laser, for example, in Table~\ref{data}, the start index is $82$ and the end index is $90$. The appropriate angle of the boundaries of the vision cone are found by converting these two indices into proper angle.
\par
				\item The velocity of the obstacle $i$ can be estimated using the laser data by some existing approaches, see e.g. \cite{JSF00, LG02}. In this case, since we have assumption (\ref{cons1}) that the velocity of the obstacle $v_i(t)$ is less than the maximum speed of the wheelchair, the numerical differentiate method is also applicable if the data are sampled at a reasonably fast rate.

				\item The bearing to the target $H(t)$ can be easily computed when the orientation of the robot is known (by estimation from encoders) and the relative angular position of the target with respect to the wheelchair is known (by simple geometry), see Fig.~\ref{para1}.

			\end{itemize}

			\subsection{Experiments Results with a SAM Wheelchair}
			
				The experiments are conducted in real life scenarios to demonstrate the applicability of BINA on SAM wheelchair. Table~\ref{chair_para} shows the wheelchair and controller parameters.
\par

\begin{figure}[!h]
			\centering
			\subfigure[]{\scalebox{0.30}{\includegraphics{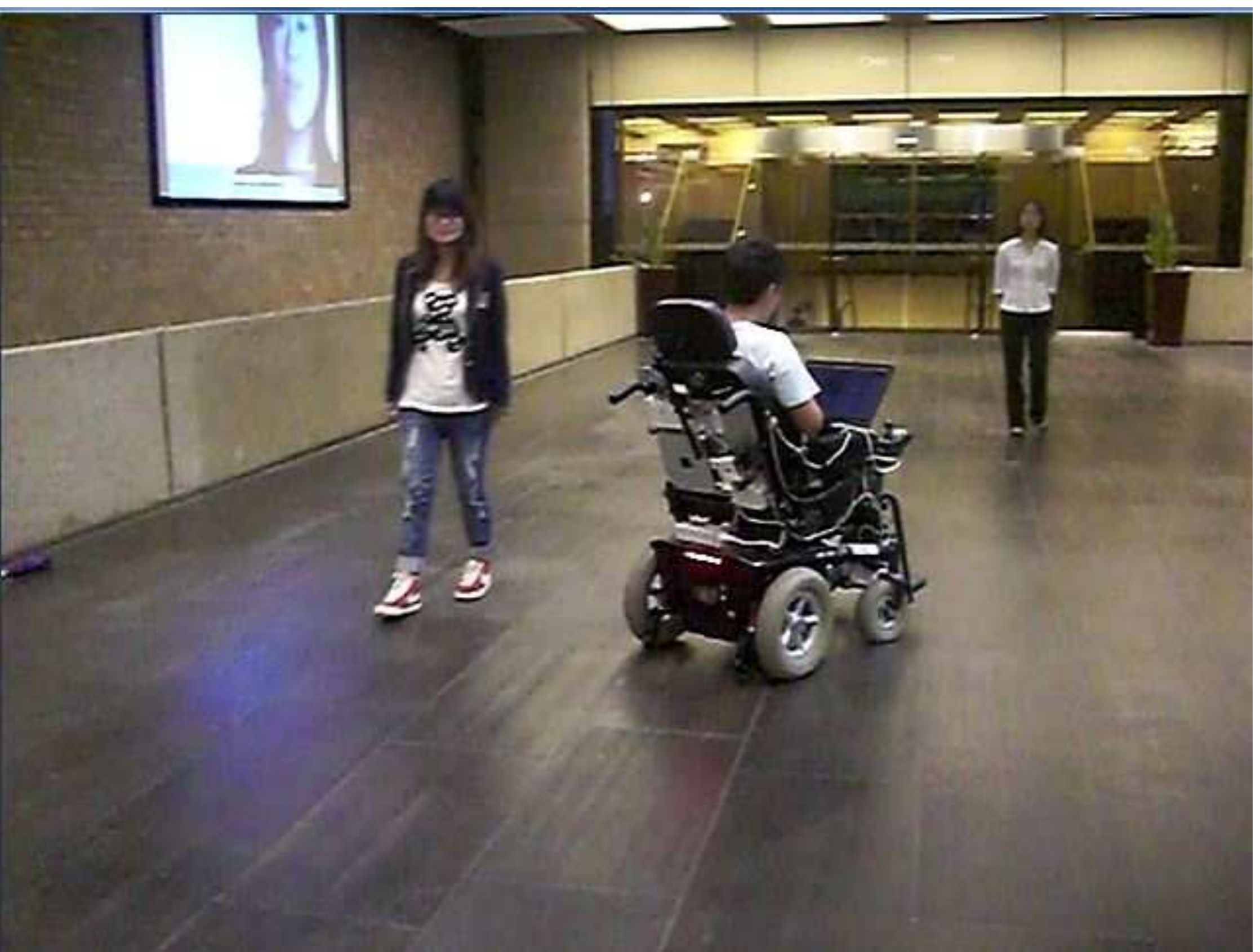}}
			\label{c6.exp11}}
			\subfigure[]{\scalebox{0.30}{\includegraphics{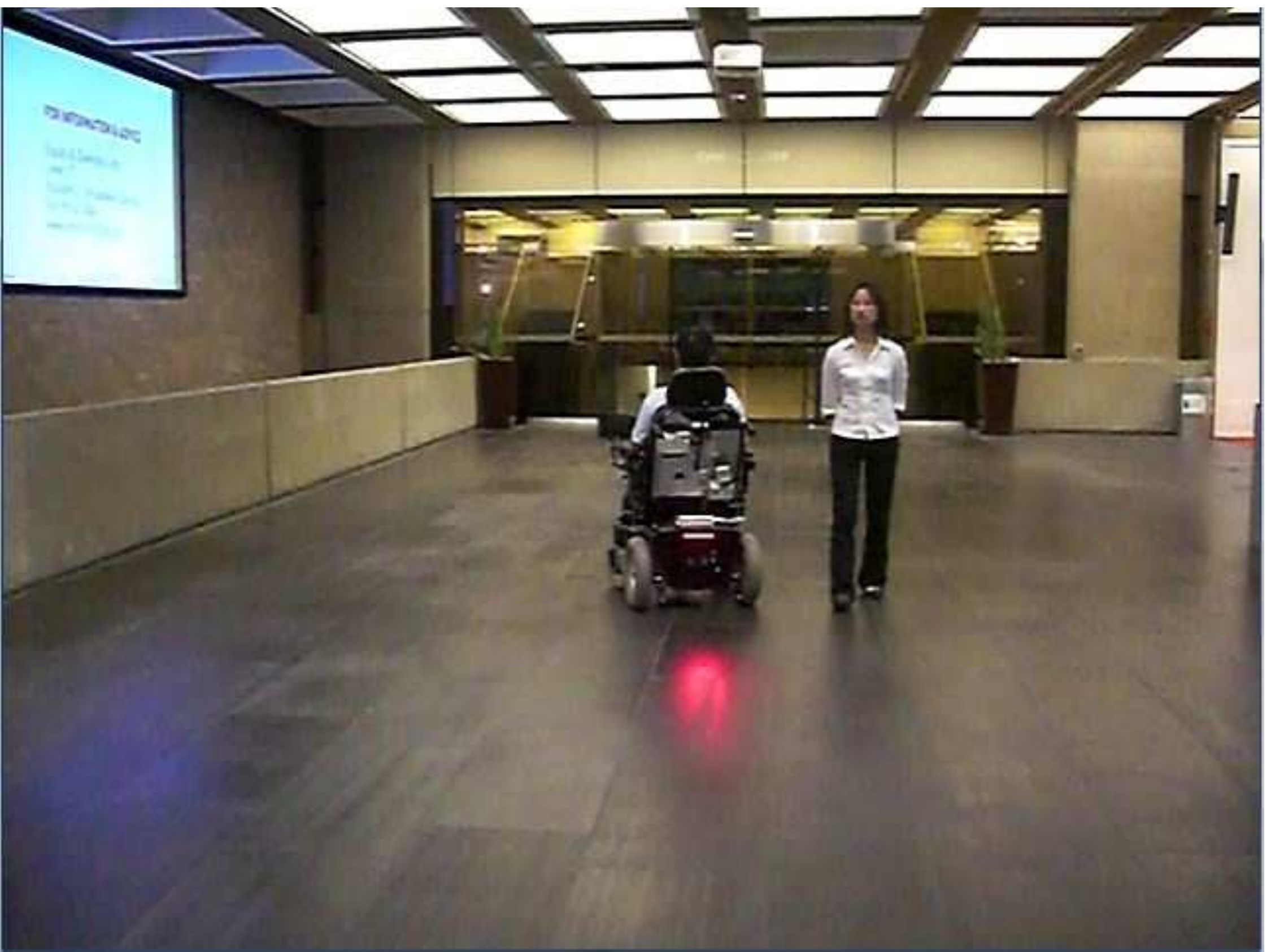}}
			\label{c6.exp12}}
			\subfigure[]{\scalebox{0.45}{\includegraphics{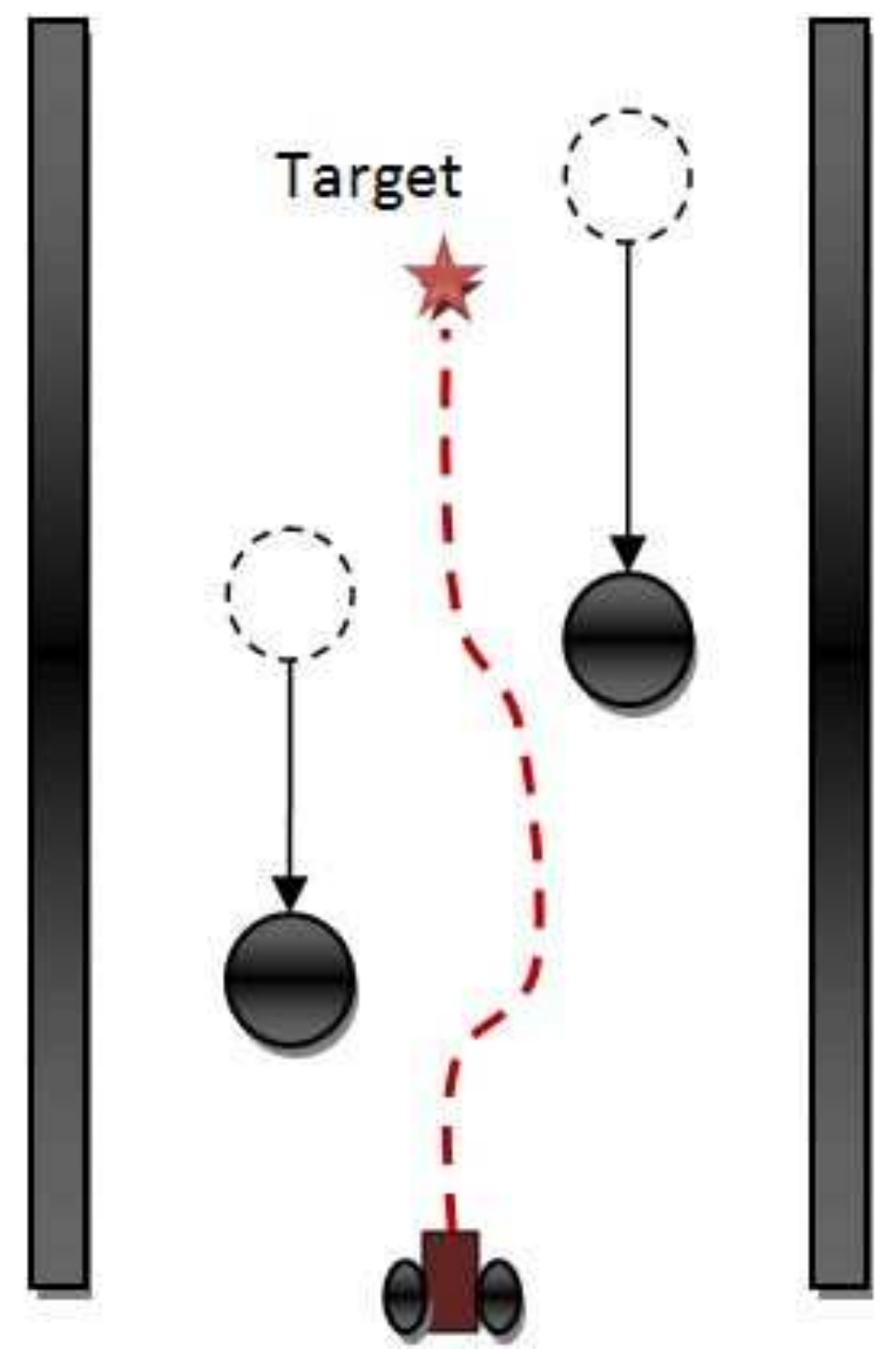}}
			\label{c6.exp13}}
			\caption{Wheelchair navigating in a dynamic environment with two moving obstacles}
			\label{c6.exp1}
			\end{figure}

				\begin{table}[h]
				\caption {Wheelchair and controller parameters}
				\centering
				\begin{tabular}{l | l| l}
					\hline
					Parameter & Value & Comments \\
					\hline
					$T_s$ & $0.1s$ & Sampling interval \\
					\hline
					$2L$ &  $0.55m$ & Distance between two driving wheels (the axle) \\
					\hline
					$D_w$  & $0.35m$ & Diameter of the driving wheels \\
					\hline
					$V_{max}$ & $0.9m/s$ & Maximum speed of the wheelchair \\
					\hline
					$U_{max}$ & $\pi /4 rad/s$ & Maximum angular velocity \\
					\hline
					$\alpha_0$ &  $\pi/3.5 rad$ & Avoiding angle \\ 
					\hline
				\end{tabular}
				\label{chair_para}
				\end{table}
				In the first experiment, we present the simple experimental result of the wheelchair navigating in a dynamic environment with two moving obstacles. Fig.~\ref{c6.exp11} and Fig.~\ref{c6.exp12} show the moments when the wheelchair bypasses each of the obstacles. It can be observed that the space between the wheelchair and the obstacles are reasonable, meaning that the distance is large enough so that potential collision is impossible to happen and small to keep the efficiency of the avoidance maneuver. The complete path is depicted in Fig.~\ref{c6.exp13}. It can be seen that the wheelchair avoids the obstacle very efficiently. This experiments can be easily extended to a complicated scenario with many more obstacles, which represents many real life situation with cluttered obstacles such as train station, busy streets. The wheelchair is able to avoid each of the obstacles with the same fashion as shown in Fig.~\ref{c6.exp1}.

				\par
			\par

\begin{figure}[!h]
			\centering
			\subfigure[]{\scalebox{0.33}{\includegraphics{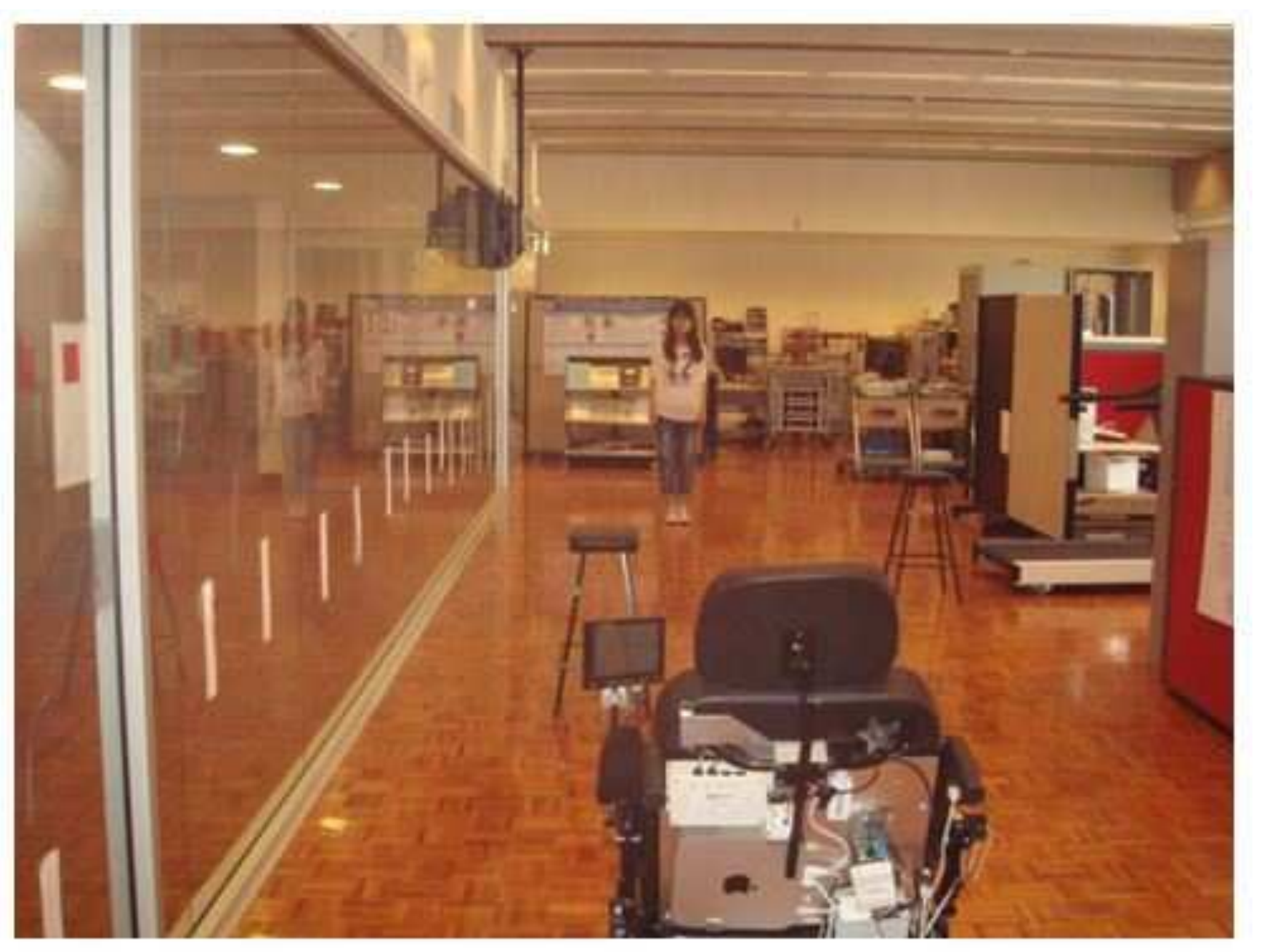}}
			\label{c6.exp21}}
			\subfigure[]{\scalebox{0.33}{\includegraphics{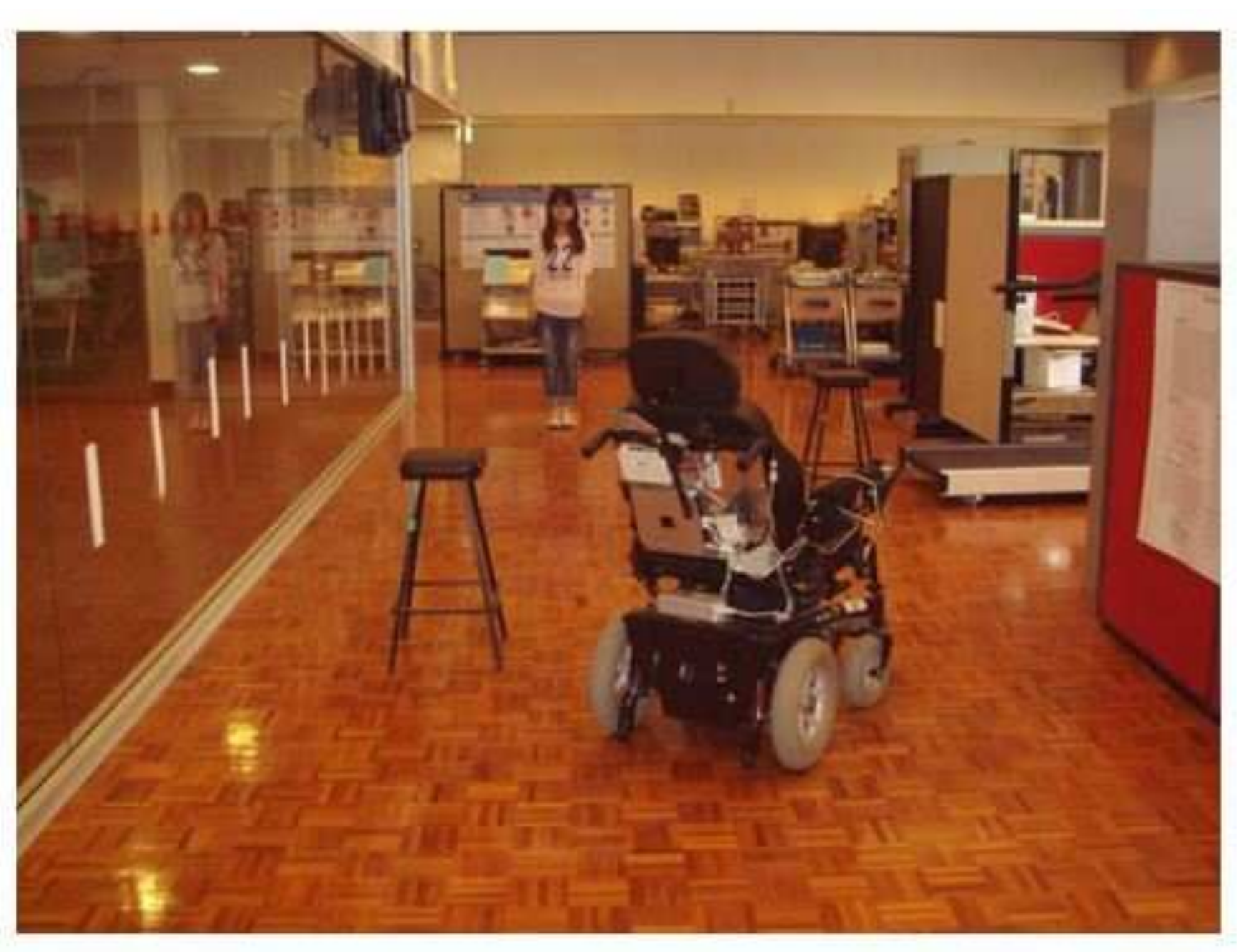}}
			\label{c6.exp22}}
			\subfigure[]{\scalebox{0.33}{\includegraphics{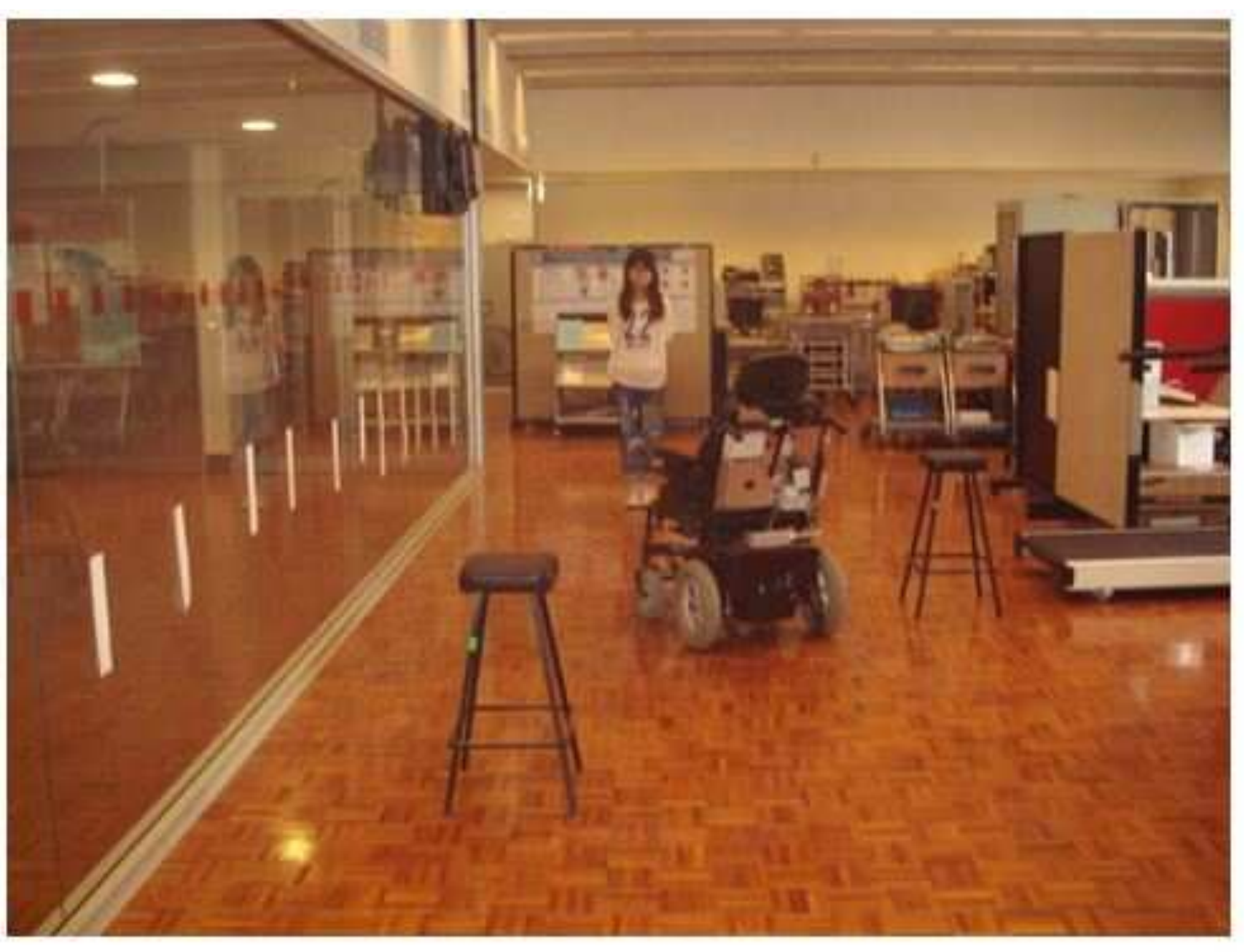}}
			\label{c6.exp23}}
			\subfigure[]{\scalebox{0.33}{\includegraphics{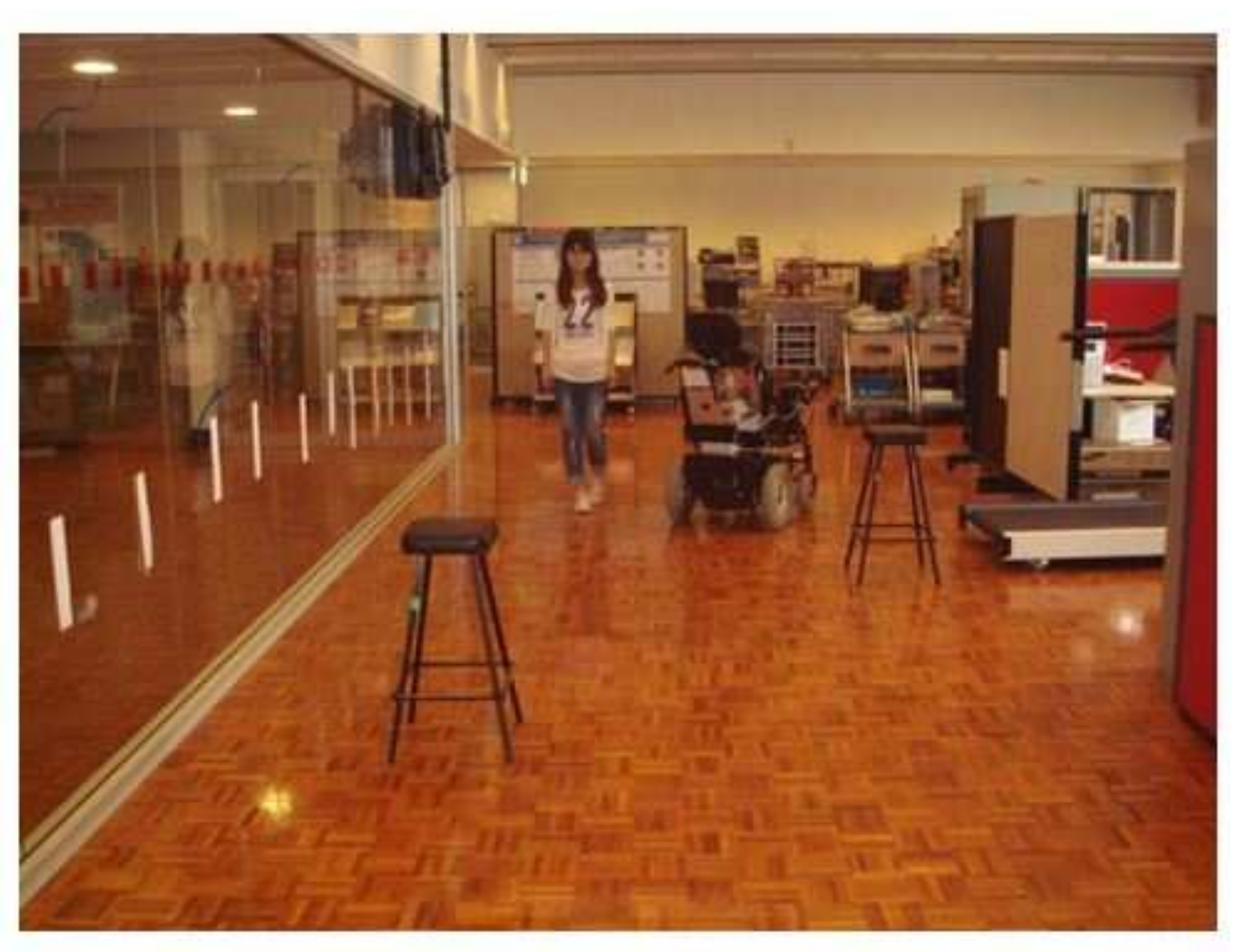}}
			\label{c6.exp24}}
			\subfigure[]{\scalebox{0.33}{\includegraphics{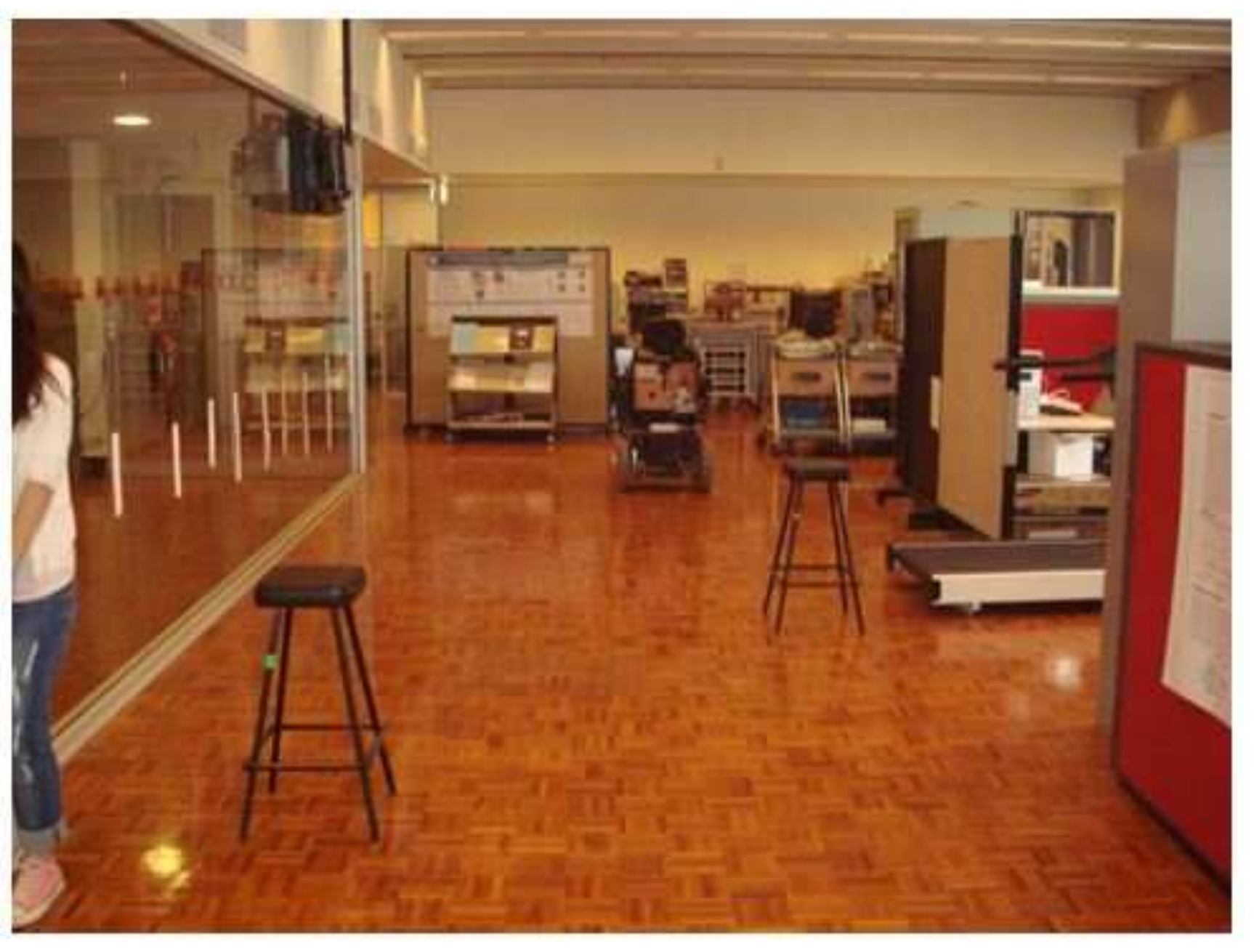}}
			\label{c6.exp25}}
			\subfigure[]{\scalebox{0.45}{\includegraphics{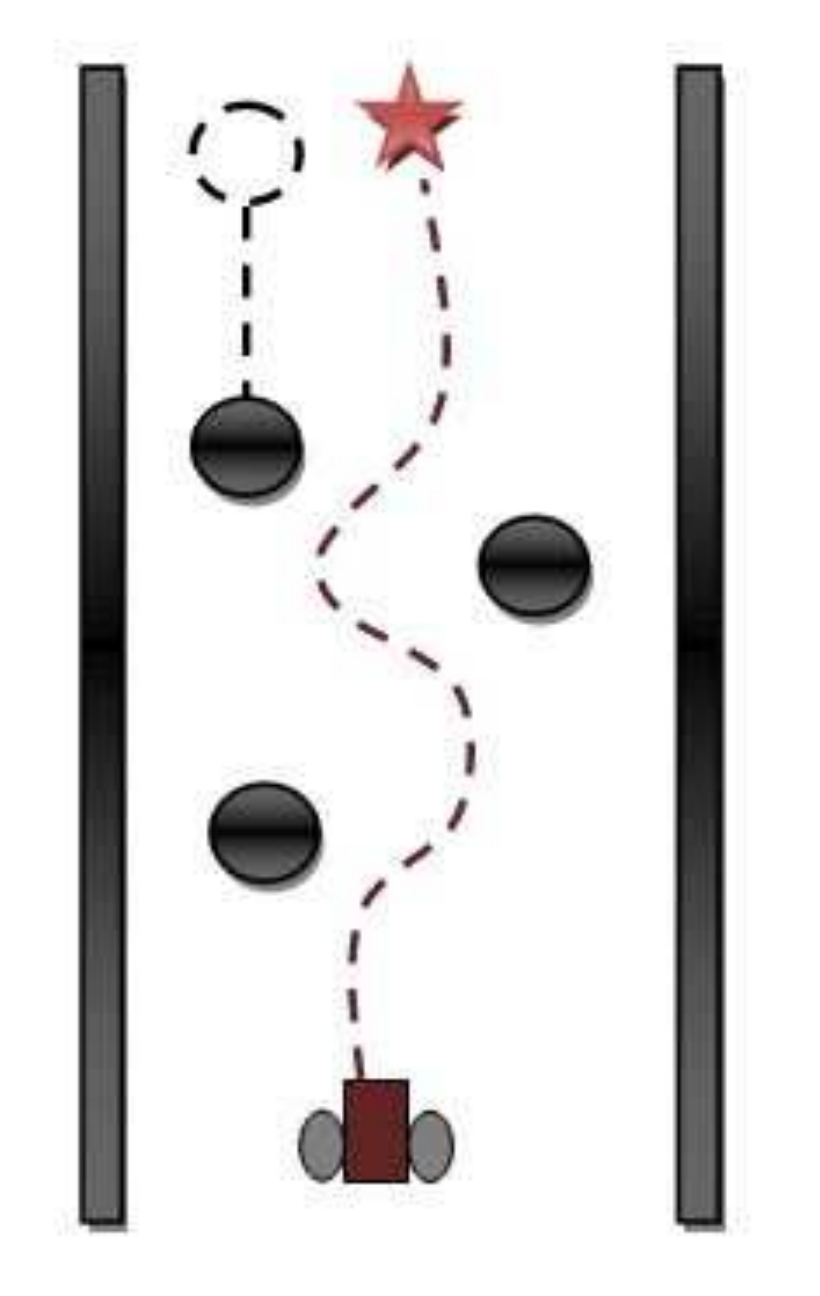}}
			\label{c6.exp26}}

			\caption{Wheelchair avoiding static and moving obstacles}
			\label{c6.exp2}
			\end{figure}

				In many of the real life scenarios, it is likely that there is a mix of stationary and dynamic obstacles in the same environments. Museums and art galleries are examples of such environment with exhibitions (stationary) and many visitors (moving). Fig.~\ref{c6.exp2} shows a similar environment with stationary (two chairs) and dynamic obstacles (the experimenter). The process of avoiding these obstacles are shown in Fig.~\ref{c6.exp22}, Fig.~\ref{c6.exp23} and Fig.~\ref{c6.exp24}. The wheelchair reaches target at Fig.~\ref{c6.exp25}.
\par
			\par

\begin{figure}[!h]
			\centering
			\subfigure[]{\scalebox{0.30}{\includegraphics{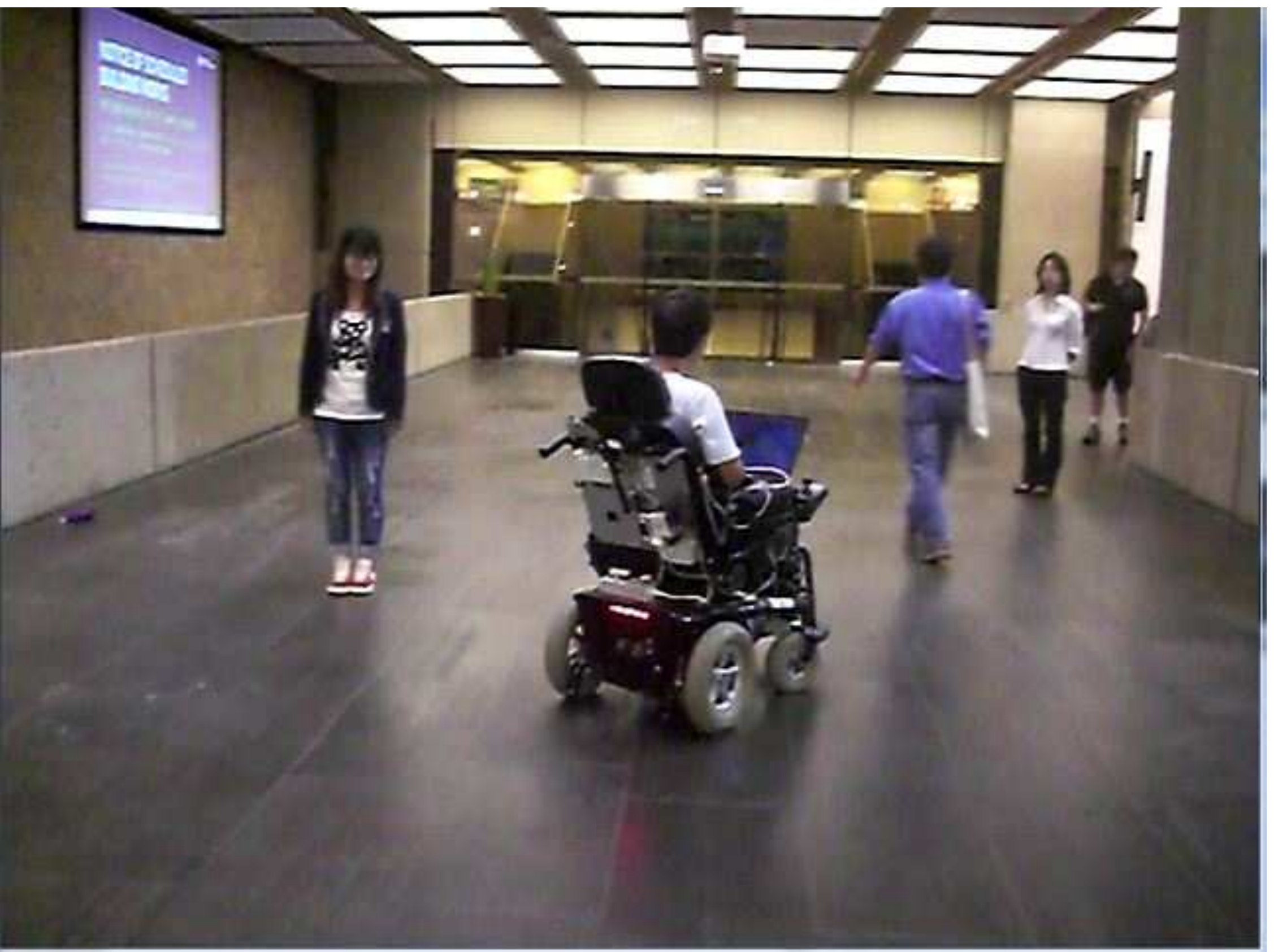}}
			\label{c6.exp31}}
			\subfigure[]{\scalebox{0.30}{\includegraphics{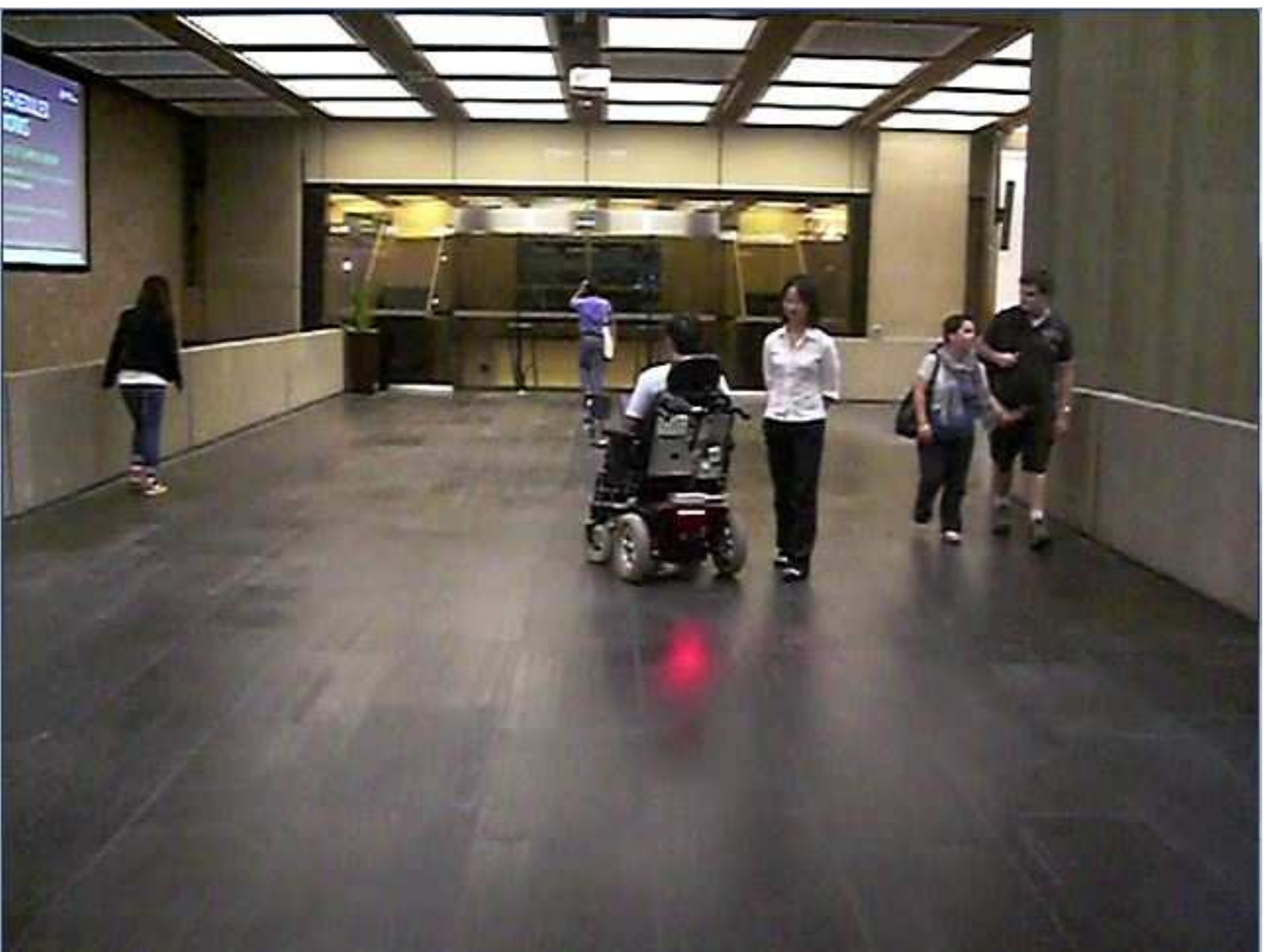}}
			\label{c6.exp32}}
			\subfigure[]{\scalebox{0.30}{\includegraphics{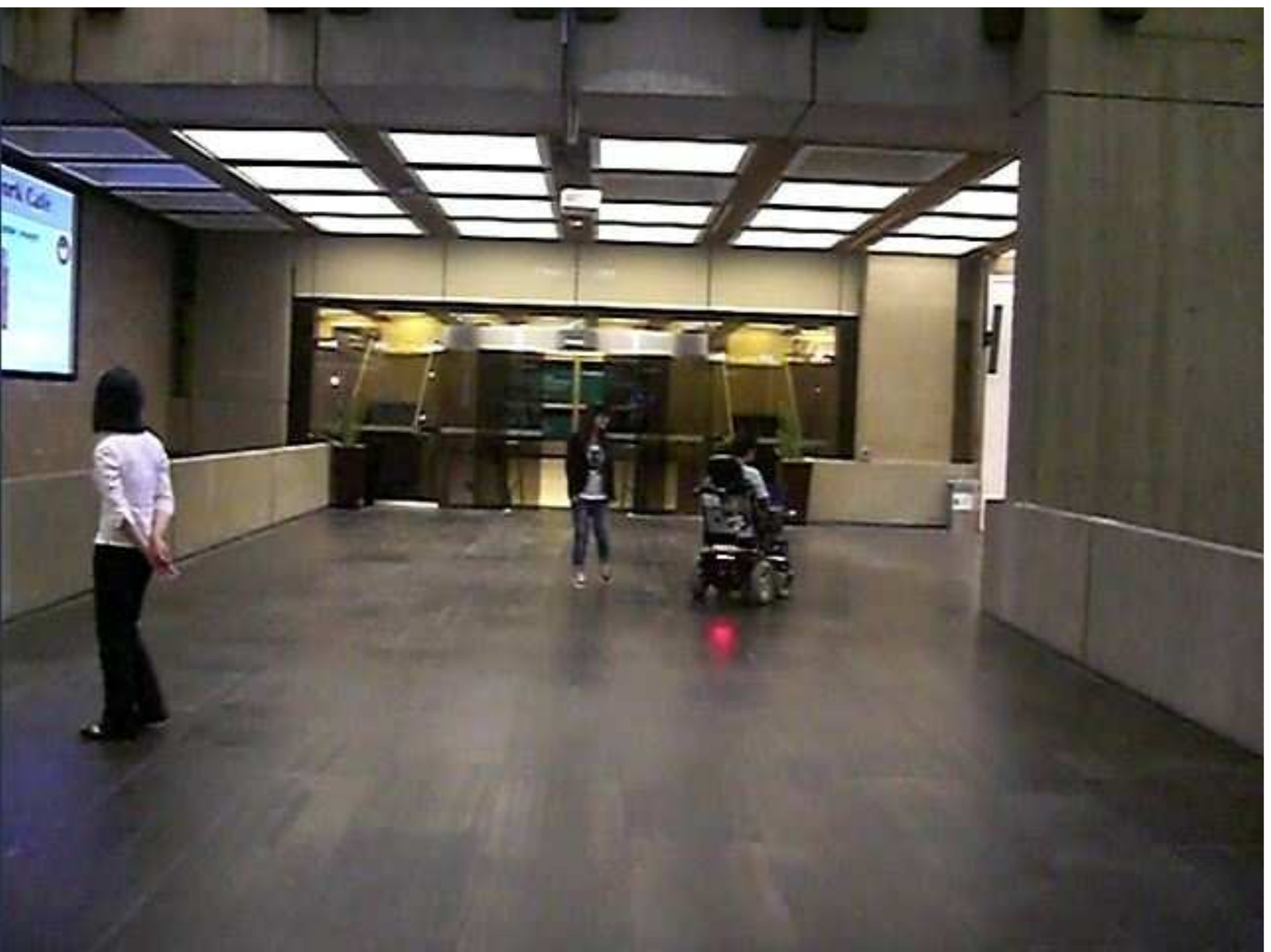}}
			\label{c6.exp33}}
			\subfigure[]{\scalebox{0.30}{\includegraphics{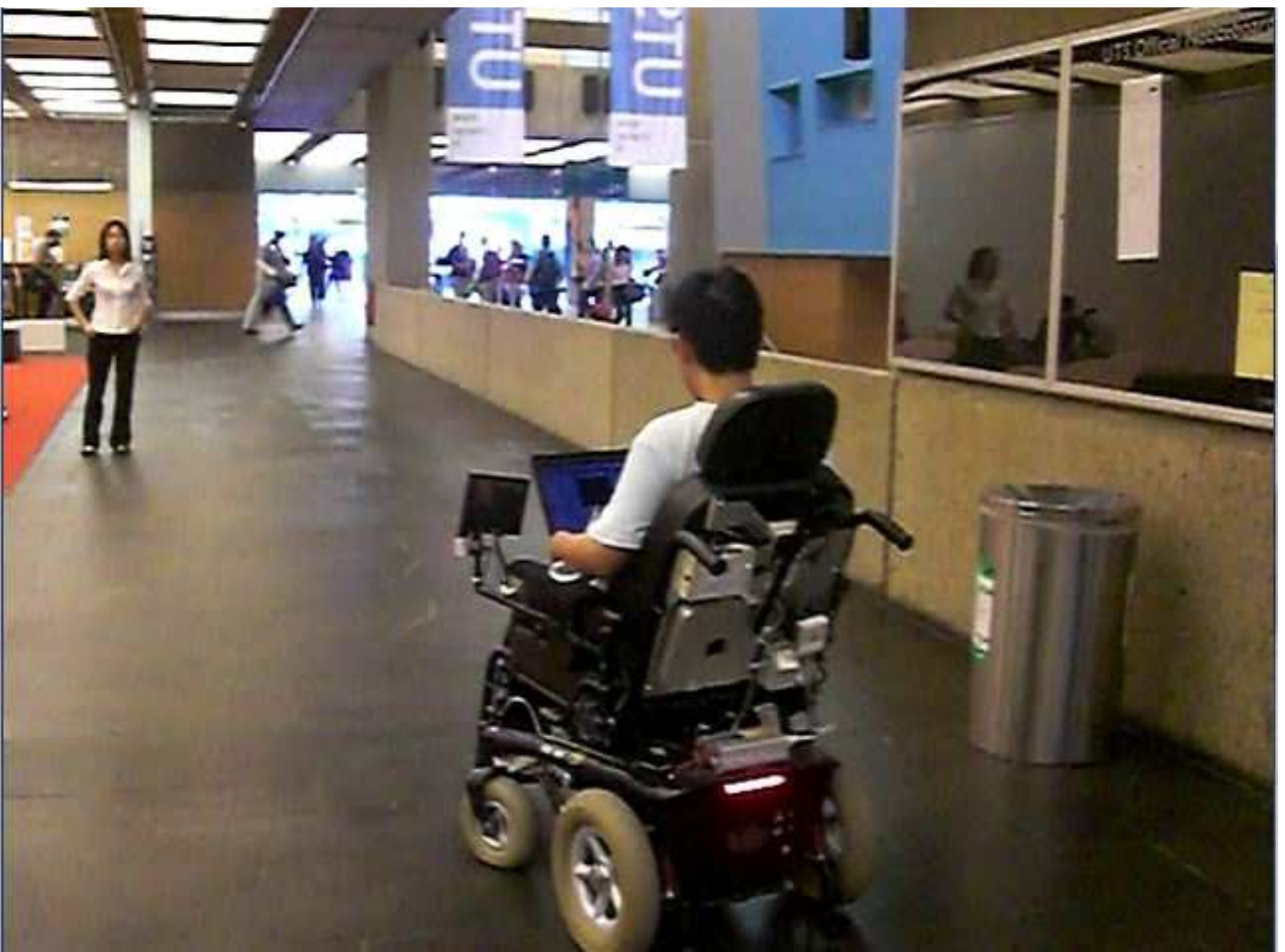}}
			\label{c6.exp34}}
			\subfigure[]{\scalebox{0.30}{\includegraphics{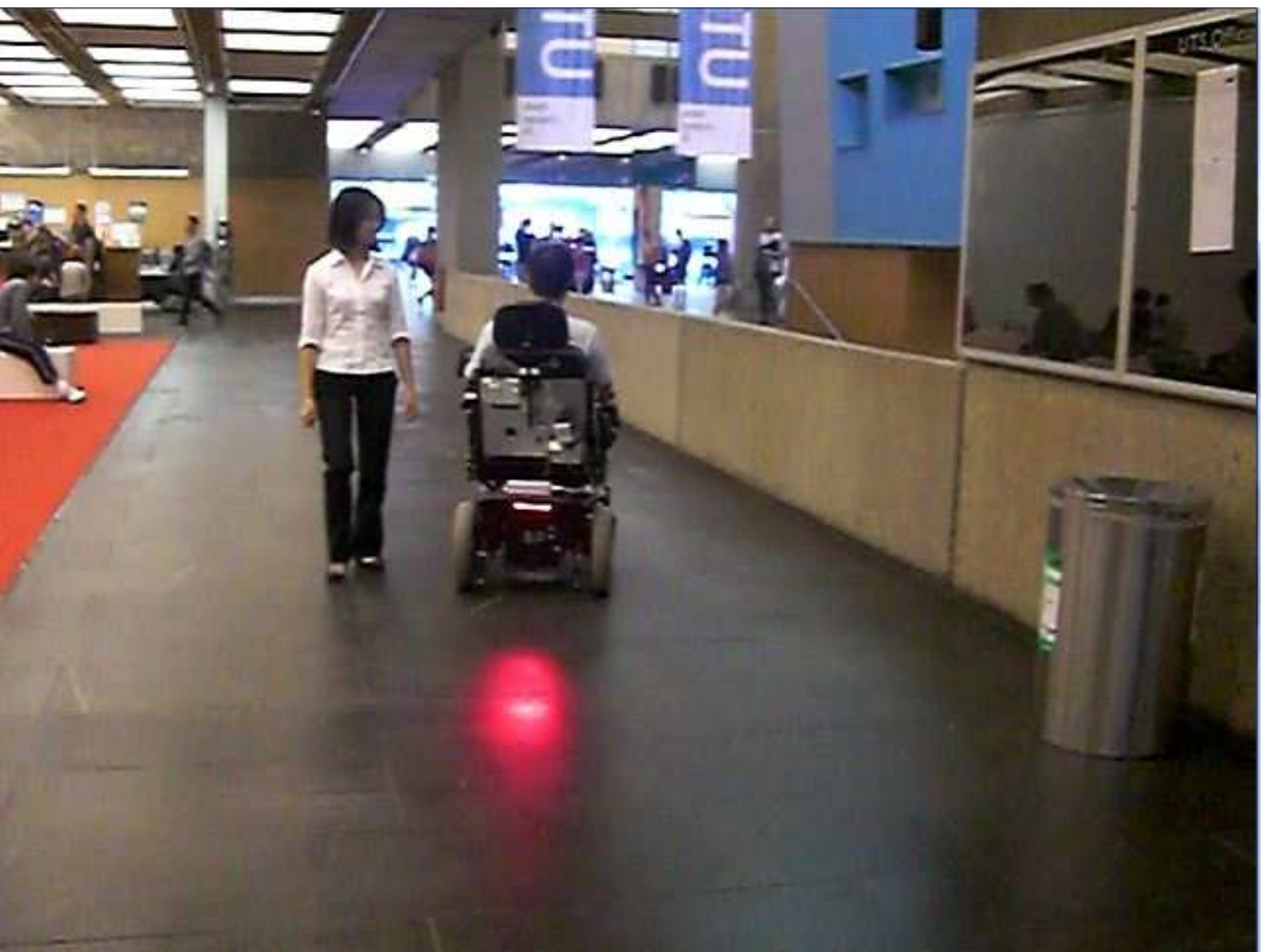}}
			\label{c6.exp35}}
			\subfigure[]{\scalebox{0.31}{\includegraphics{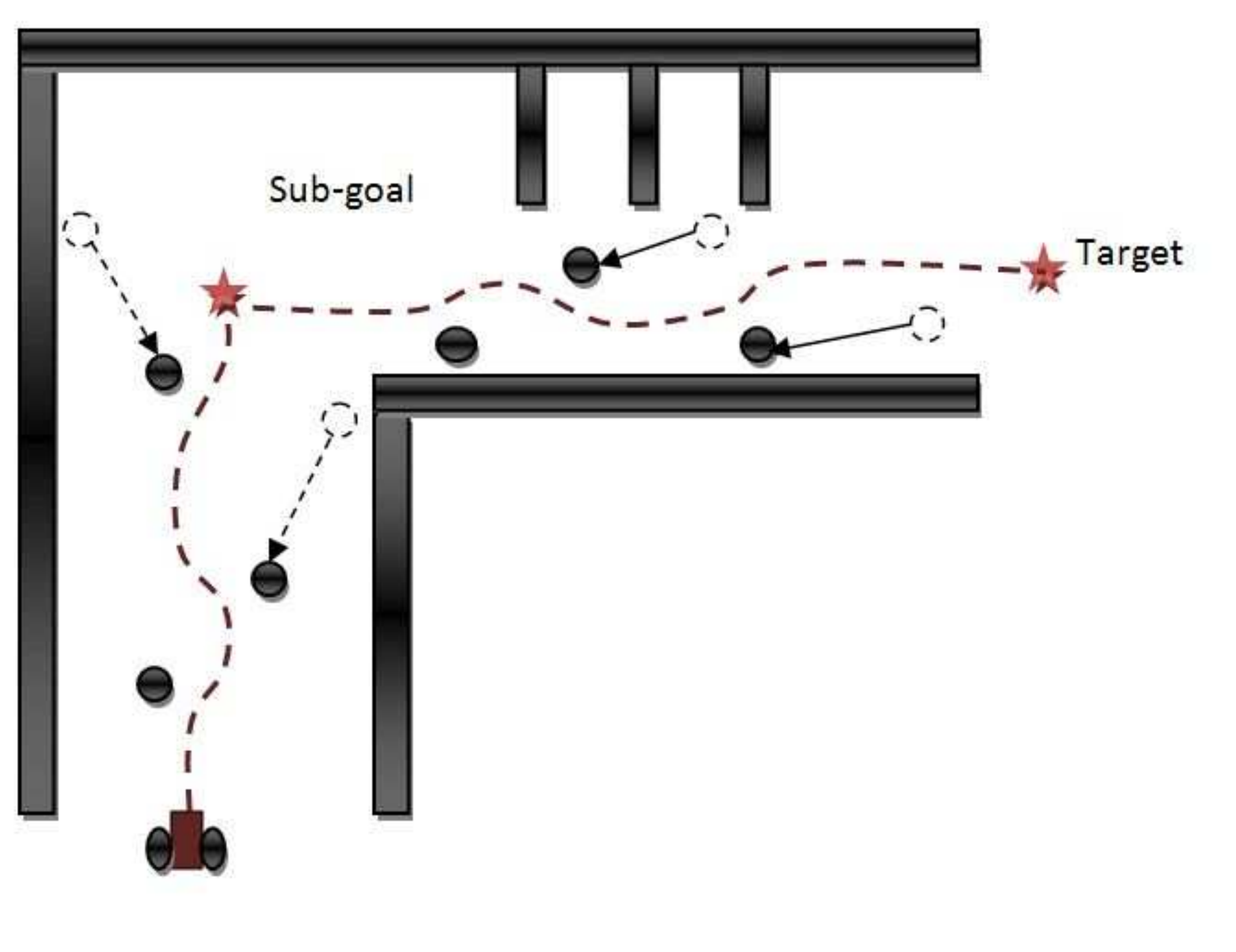}}
			\label{c6.exp37}}
			\caption{Wheelchair navigating in a cluttered corridor with moving pedestrians}
			\label{c6.exp3}
			\end{figure}

			The last experiment in Fig.~\ref{c6.exp3} shows a complex scenario of navigating a wheelchair in a narrow corridor with stationary and dynamic obstacles. Another challenge of this experiment is that the final target can not be seen directly at the starting point. The wheelchair has to reach the target via a sub-target point. The wheelchair avoids a number of en-route obstacles (pedestrians, wall, trash bin and couches) in Fig.~\ref{c6.exp31}, Fig.~\ref{c6.exp32}, Fig.~\ref{c6.exp33}, Fig.~\ref{c6.exp34}, Fig.~\ref{c6.exp35}, Fig.~\ref{c6.exp37}. This experiment environment presents numerous public environments with pedestrians such as parks, streets.

			\par

	\section {Implementation and Experiments of Equidistant Navigation Algorithm (ENA)}

		\subsection {Parameter Measurement}

			The only measurement required by ENA is the minimum distance $d_i(t)$ from the wheelchair to the obstacle $i$. This information is measured by URG-04LX laser range finder in a similar but simpler fashion as introduced in Section~\ref{para_meas1}. Since ENA does not require the direction for which the minimum distance is achieved, the data received by the notebook is an array of data without index, see e.g. Table~\ref{data2}. Therefore, it is more cost-efficient to implement the ENA on the wheelchair because the algorithm can work with other simpler or cheaper detection devices which provide distance information.
\par
				\begin{table}[h]
					\caption {Fragment of data from URG-04LX laser range finder}
					
				\begin{tabular}{l |l |l |l |l |l |l |l |l |l |l |l |l |l |l |l | l }				
					\hline
			              $dis$(m) & ... & 4.0 & 4.0 & 2.8 & 2.6 & 2.4 &  2.2  & 2.2 & 2.2 & 2.6 & 2.8 & 4.0 & 4.0 &  ...    \\
					\hline 
				\end{tabular}
				\label {data2}
				\end{table}
			The bearing to target $H(t)$ can be easily obtained and is described in Section~\ref{para_meas1}.

		\subsection {Experiment Results with SAM Wheelchair}

			  \begin{figure}[!t]
			\centering
			\subfigure[]{\scalebox{0.30}{\includegraphics{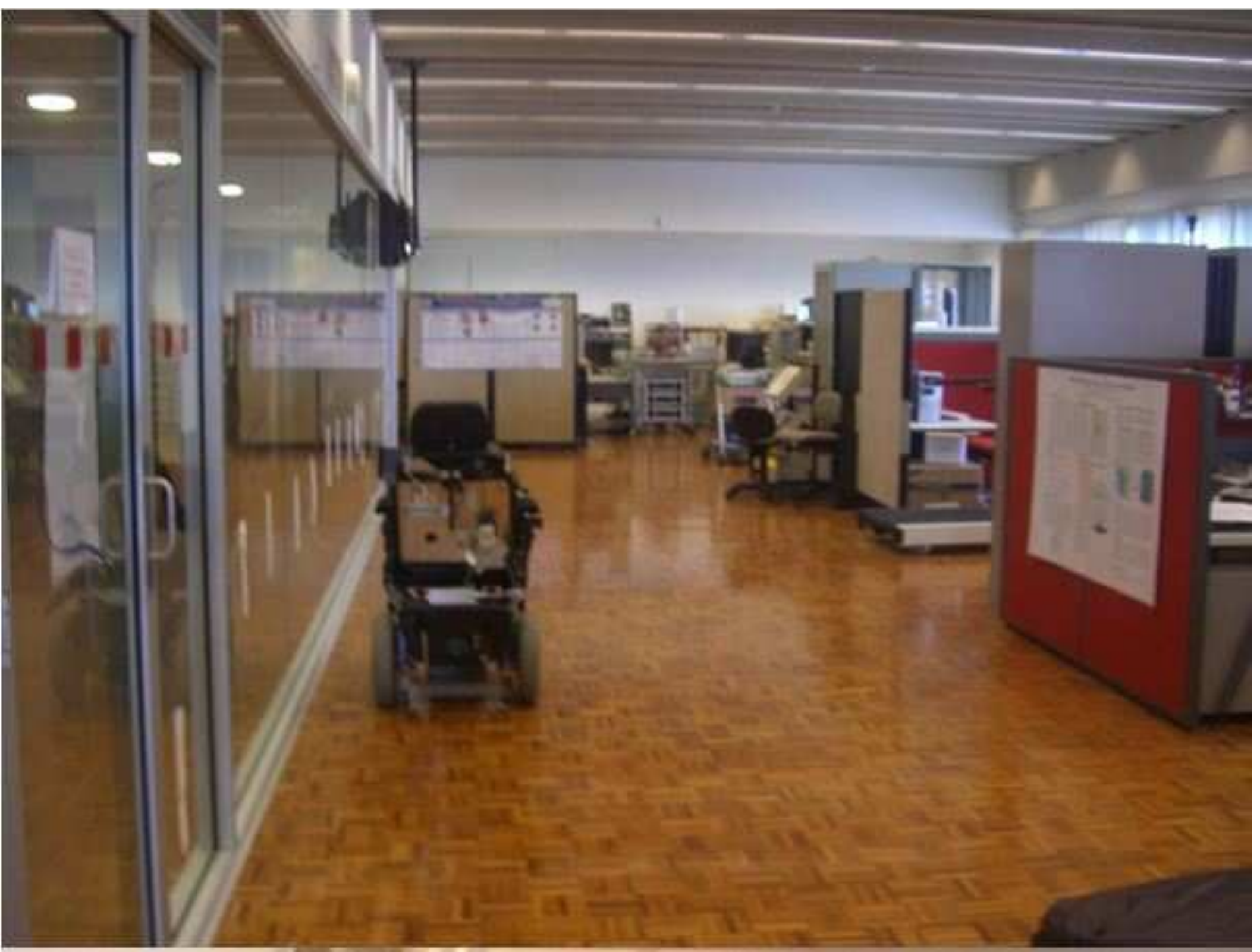}}
			\label{c6.exp41}}
			\subfigure[]{\scalebox{0.30}{\includegraphics{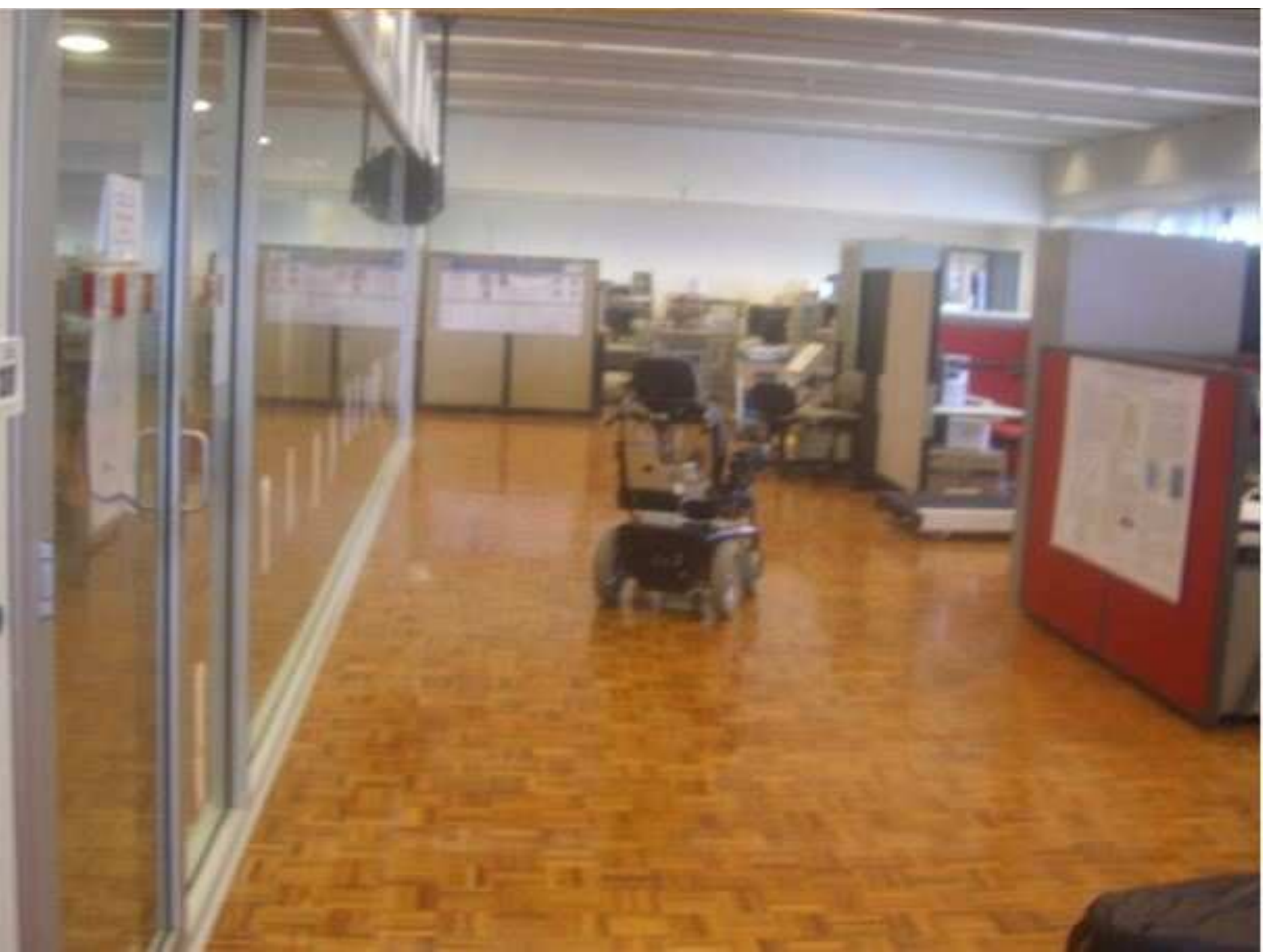}}
			\label{c6.exp42}}
			\subfigure[]{\scalebox{0.30}{\includegraphics{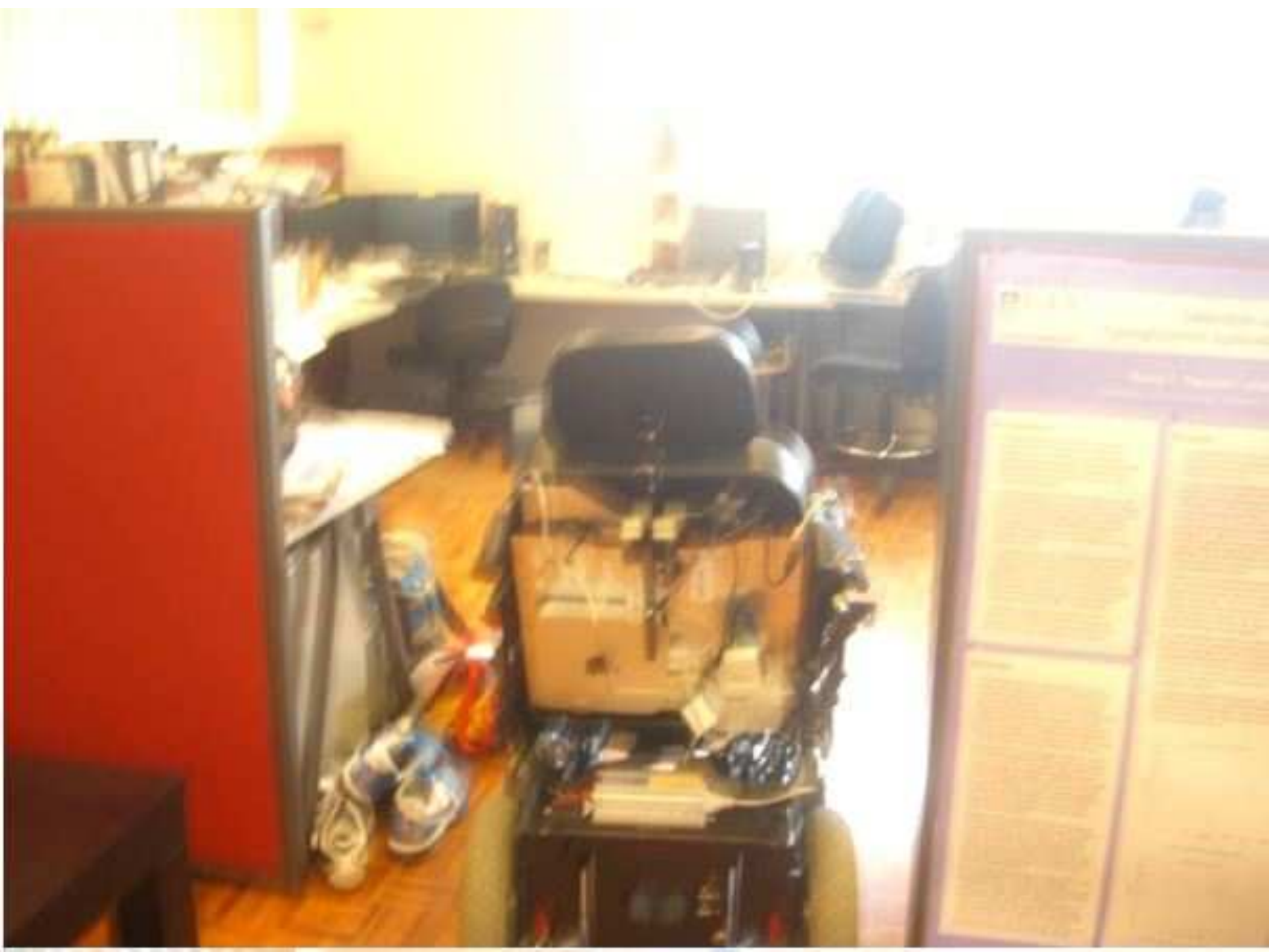}}
			\label{c6.exp43}}
			\subfigure[]{\scalebox{0.30}{\includegraphics{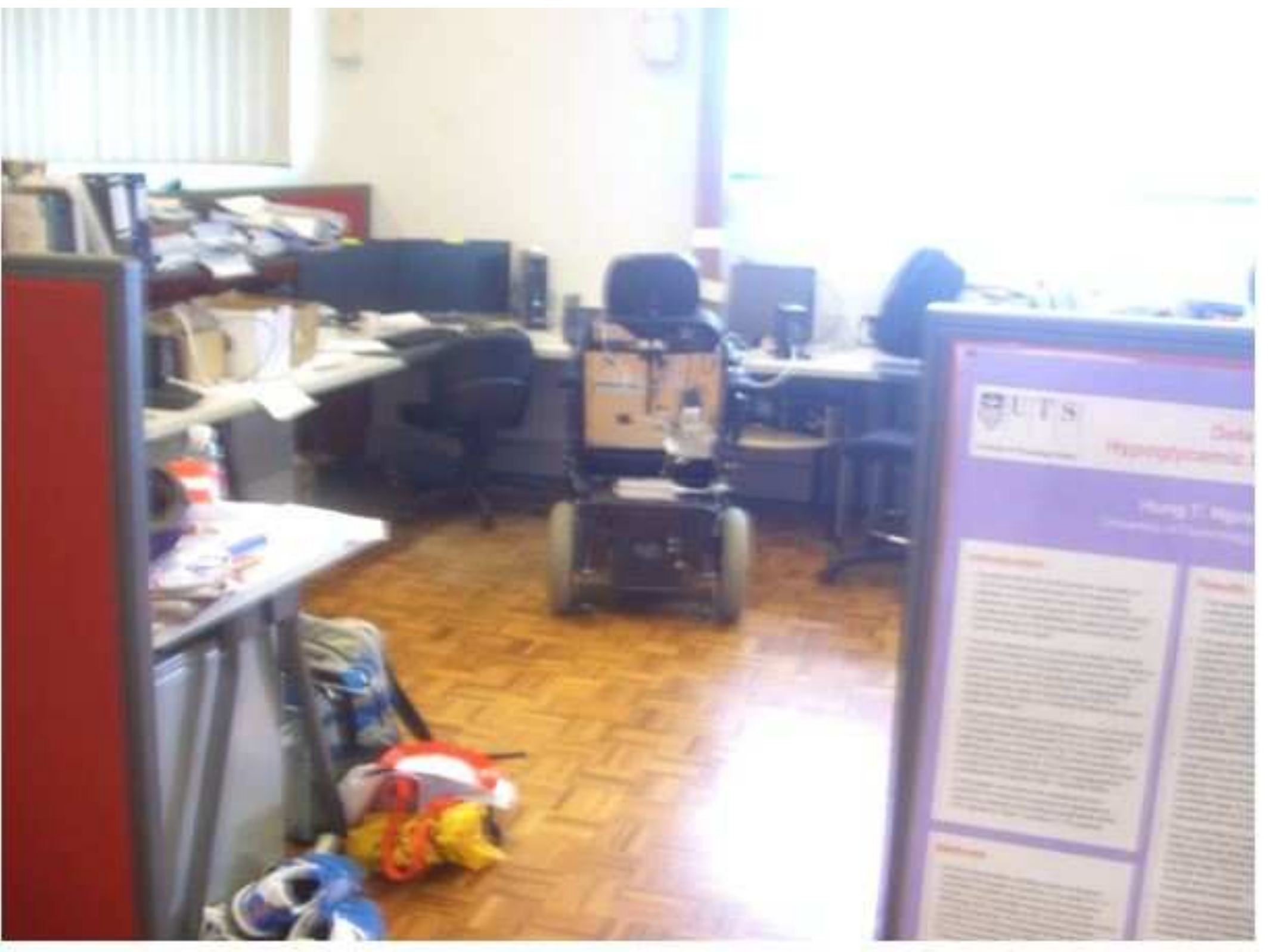}}
			\label{c6.exp44}}
			\subfigure[]{\scalebox{0.30}{\includegraphics{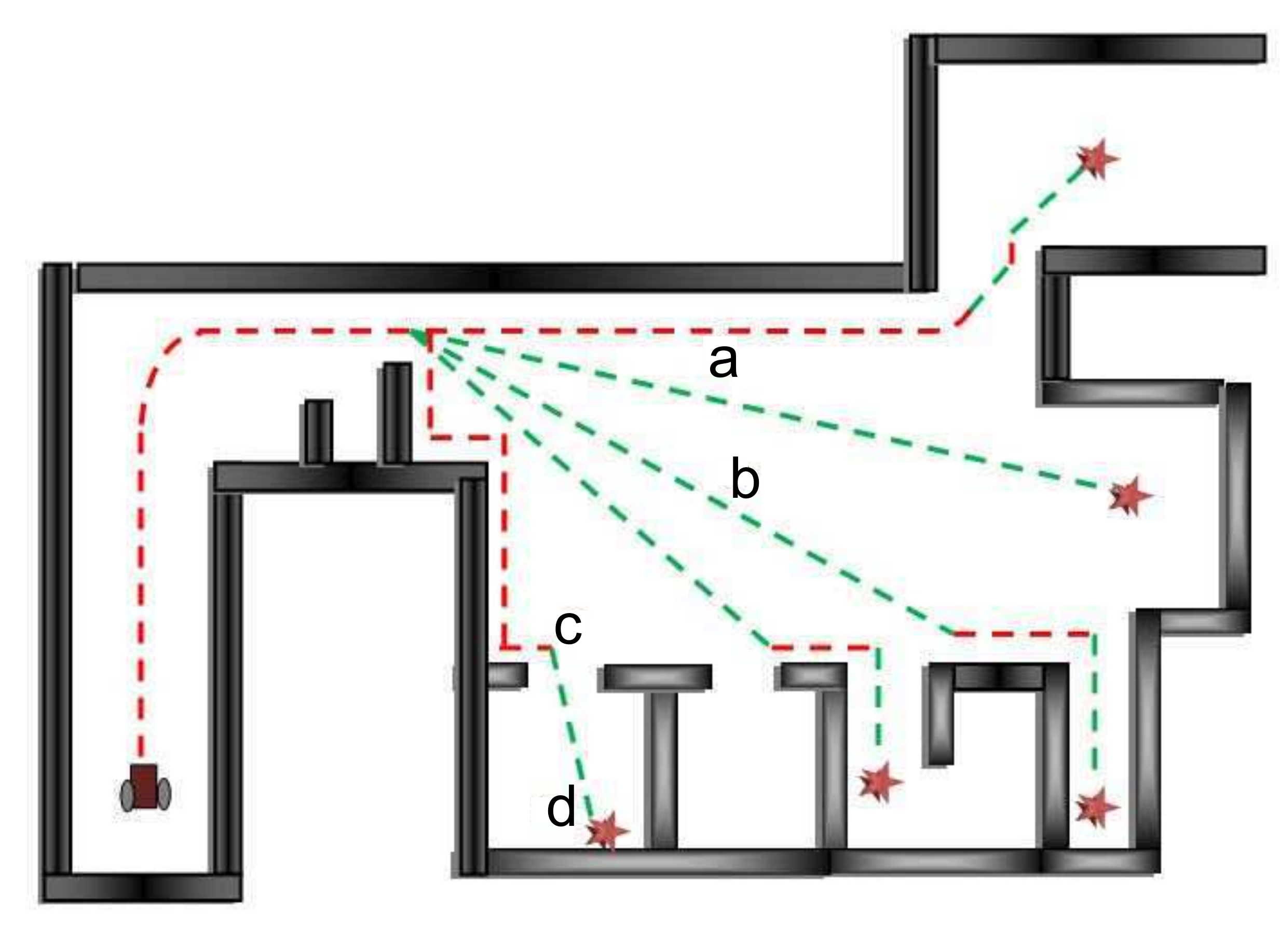}}
			\label{c6.exp45}}
			\caption{Wheelchair performs basic navigation tasks}
			\label{c6.exp4}
			\end{figure}
			\par


			\begin{figure}[!h]
			\centering
			\subfigure[]{\scalebox{0.30}{\includegraphics{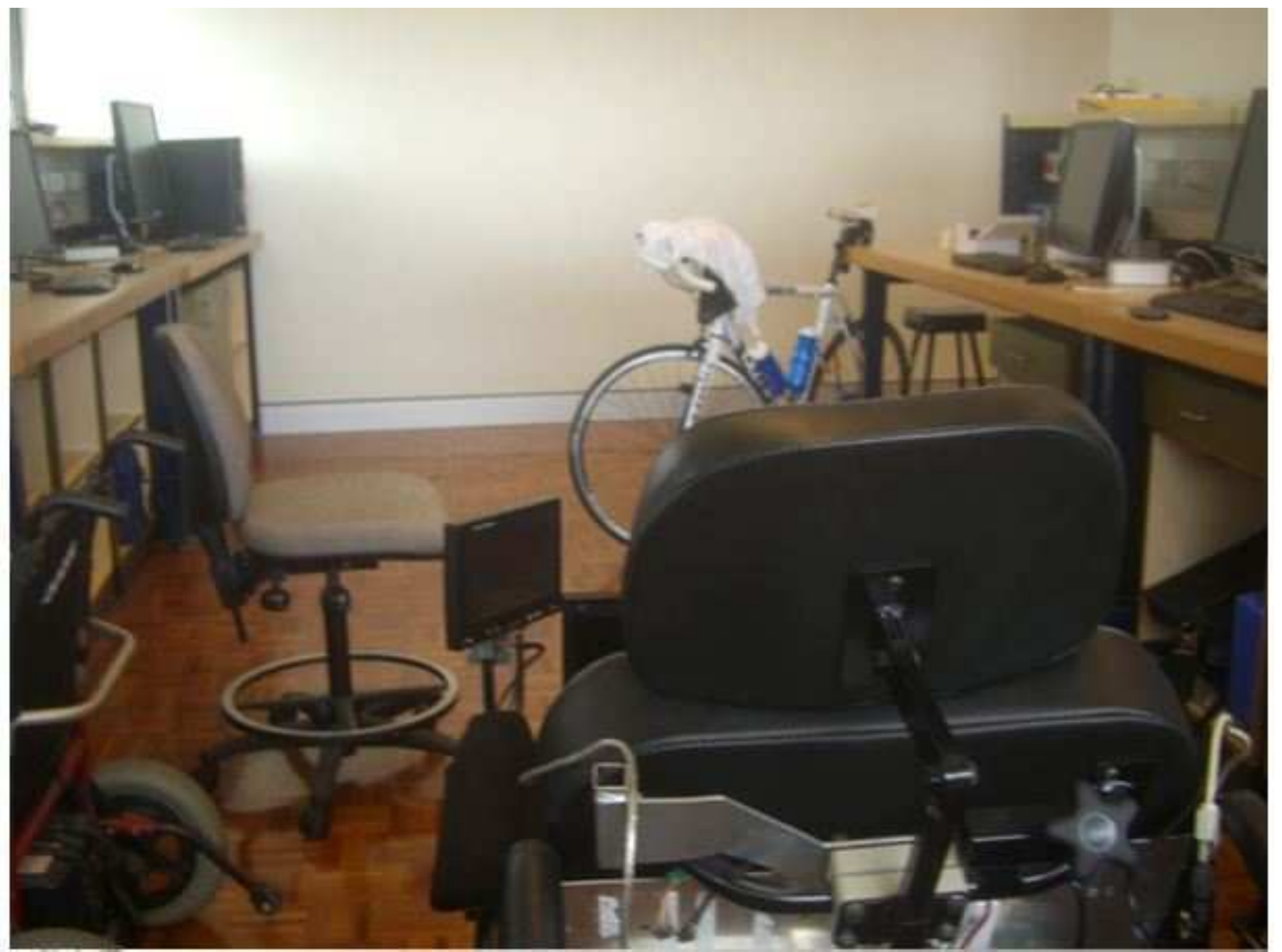}}
			\label{c6.exp51}}
			\subfigure[]{\scalebox{0.30}{\includegraphics{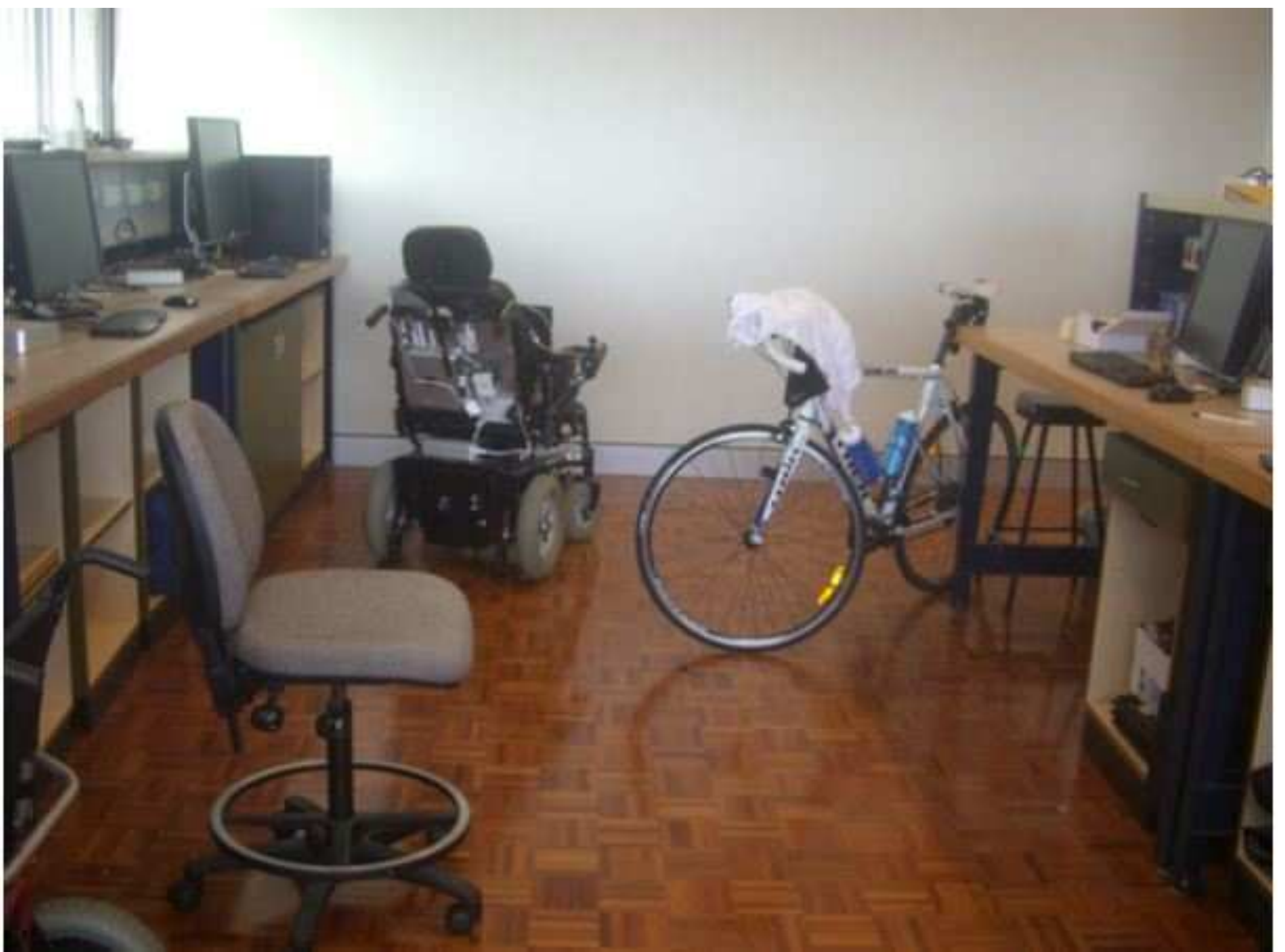}}
			\label{c6.exp52}}
			\subfigure[]{\scalebox{0.30}{\includegraphics{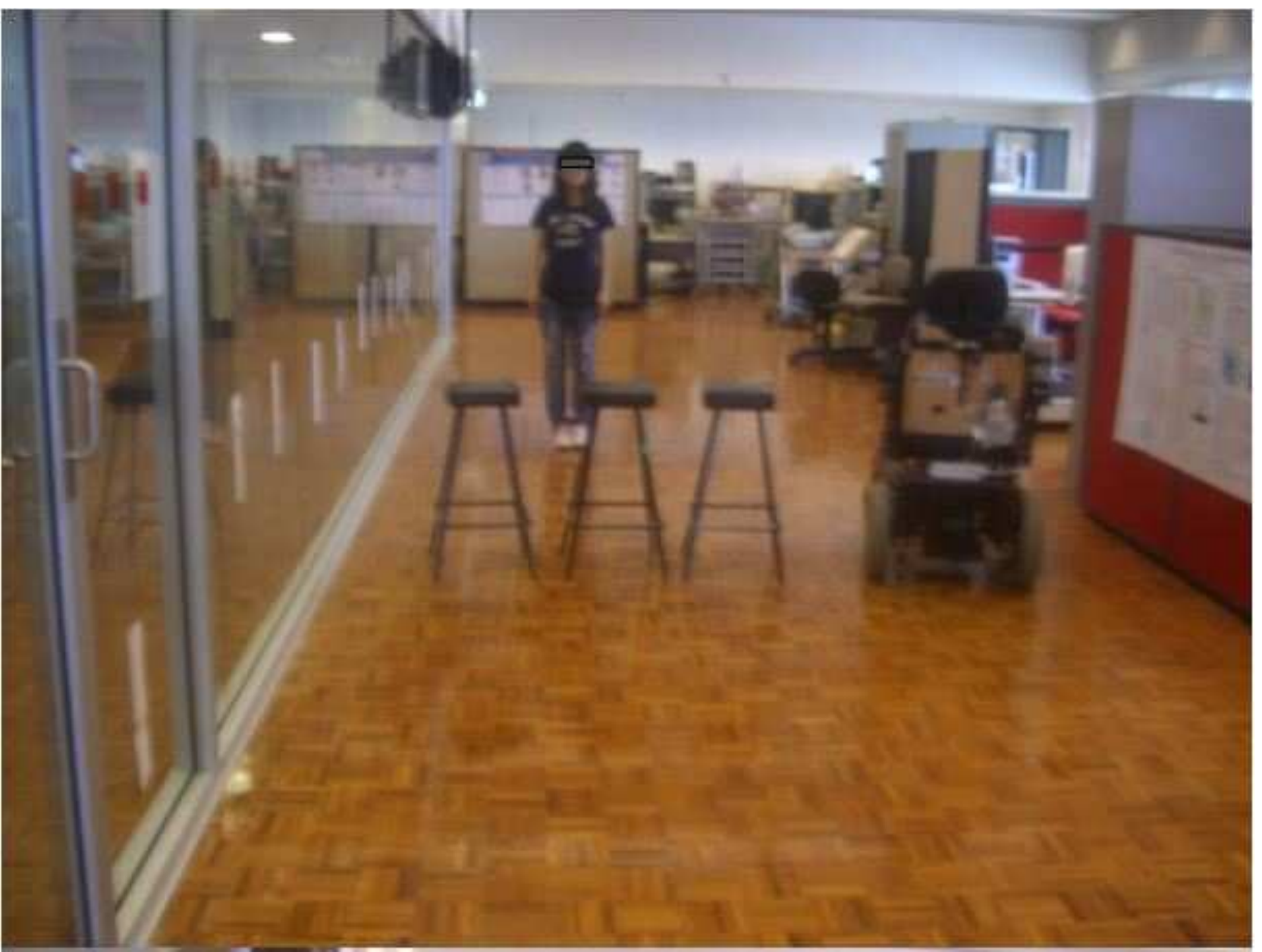}}
			\label{c6.exp53}}
			\subfigure[]{\scalebox{0.30}{\includegraphics{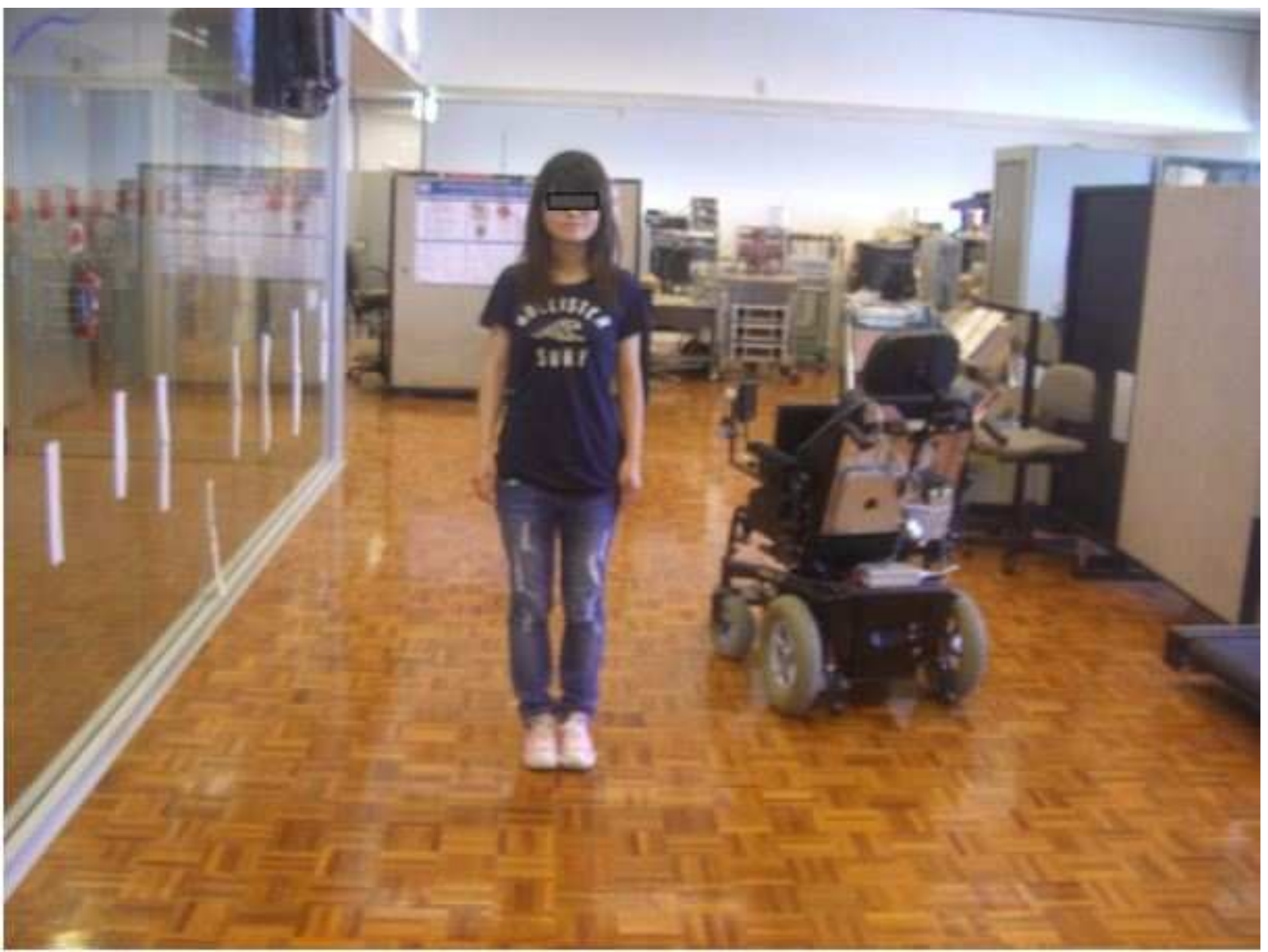}}
			\label{c6.exp54}}
			\subfigure[]{\scalebox{0.30}{\rotatebox{-90}{\includegraphics{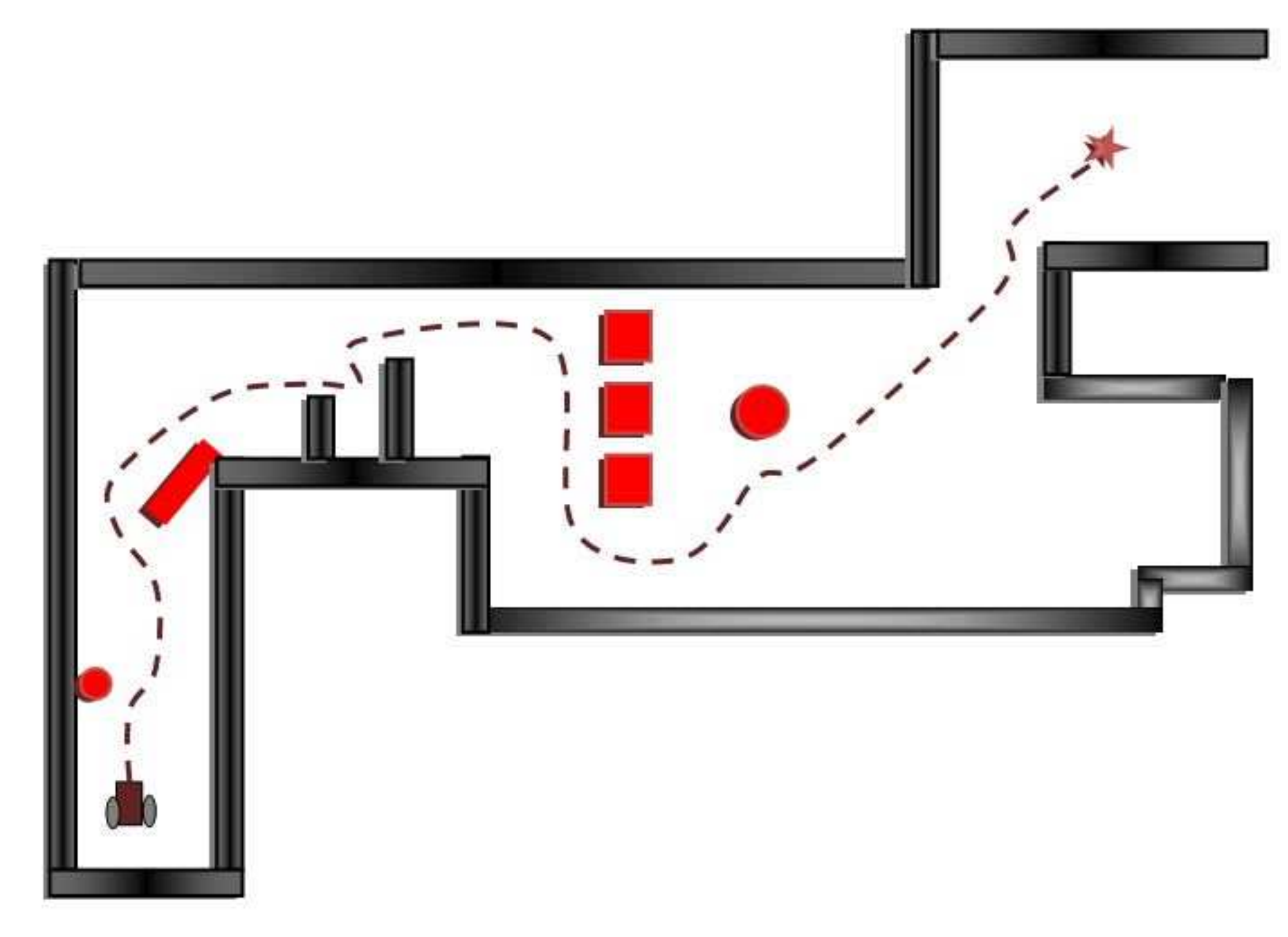}}}
			\label{c6.exp55}}
			\caption{Wheelchair avoids random obstacles}
			\label{c6.exp5}
			\end{figure}
			\par
			    We first show that the wheelchair with ENA is able to perform basic navigation tasks in a static environment.	The wheelchair has to arrive at various target positions without any collision. The most challenging part of the experiment was to maintain a given margin between the wheelchair and the obstacles (mainly the walls). This experiment also presents solutions to the challenging scenarios that can not be solved by many existing approaches, i.e. narrow passage (see location c in Fig.~\ref{c6.exp45}), hidden target location (see all the target locations in Fig.~\ref{c6.exp45}).  The figures capture crucial moment of the experiment and demonstrate maintaining a given margin to the wall (see Fig.~\ref{c6.exp41}), straight motion to the target (Fig.~\ref{c6.exp42}), crossing a narrow passage (similar to a door passage, Fig.~\ref{c6.exp43}), and arrival at target position (Fig.~\ref{c6.exp44}). The corresponding points are indicated by marks in Fig.~\ref{c6.exp45}. Fig.~\ref{c6.exp45} also depicts the complete paths for all experiments. The green dashed lines indicate the path for which the straight motion to the target is activated and the red dashed lines indicates the path for which the obstacle avoidance law is put in use.

			We also added some occasional obstacles (bike, a chain of chairs, personal) in the scene and examined the resultant performance of the proposed navigation algorithm. It can be seen from Fig.~\ref{c6.exp55} that the shapes of the obstacles are arbitrary. The initial position of the wheelchair is shown in Fig.~\ref{c6.exp51}. Fig.~\ref{c6.exp52} and Fig.~\ref{c6.exp53} correspond to the moments when the wheelchair avoids a specific obstacle. The overall path is shown in Fig.~\ref{c6.exp54}. As can be seen, the wheelchair is still able to reach the target while respecting the required safety margin.
\par

			\begin{figure}[!h]
			\centering
			\subfigure[]{\scalebox{0.30}{\includegraphics{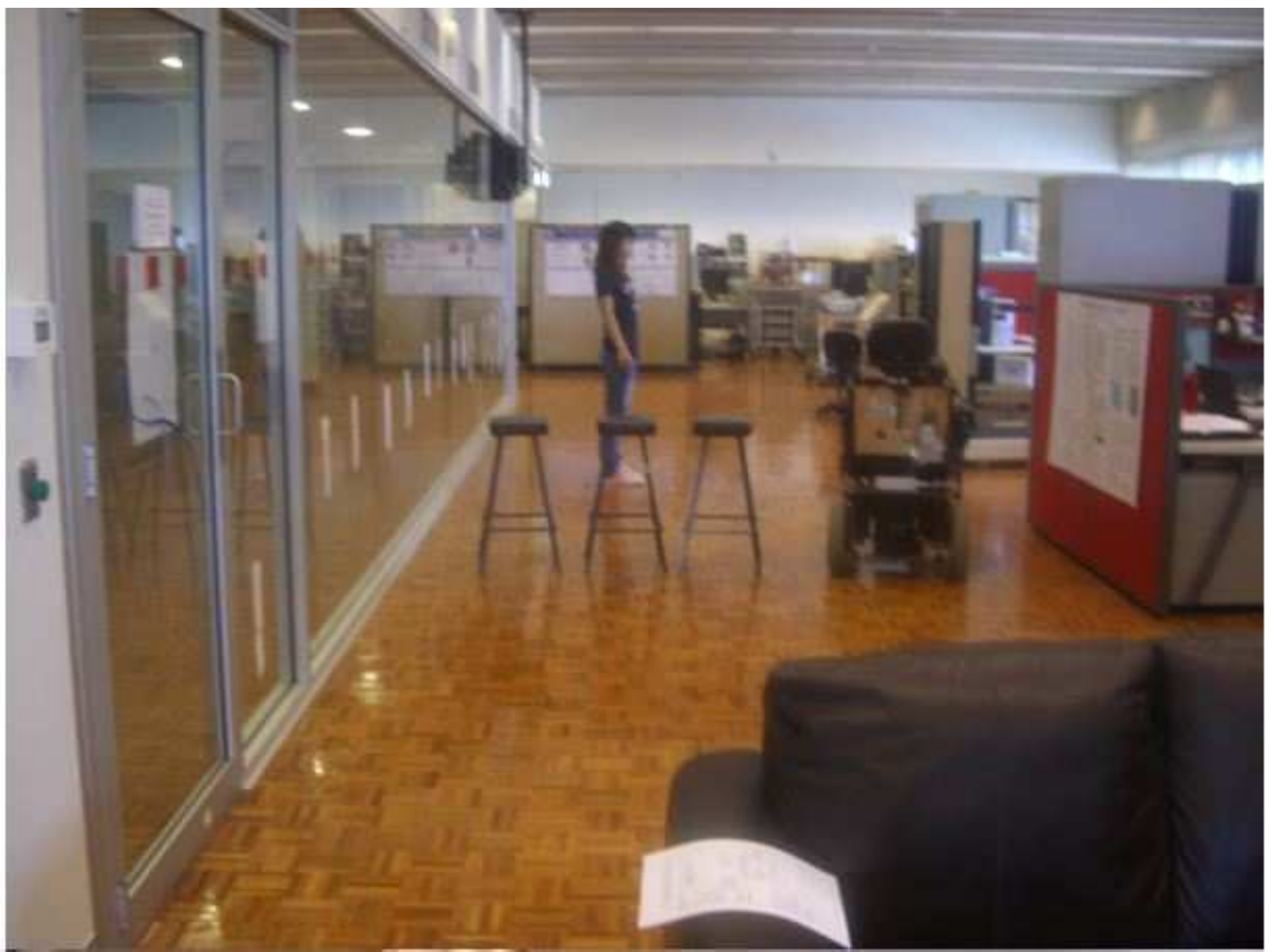}}
			\label{c6.exp61}}
			\subfigure[]{\scalebox{0.30}{\includegraphics{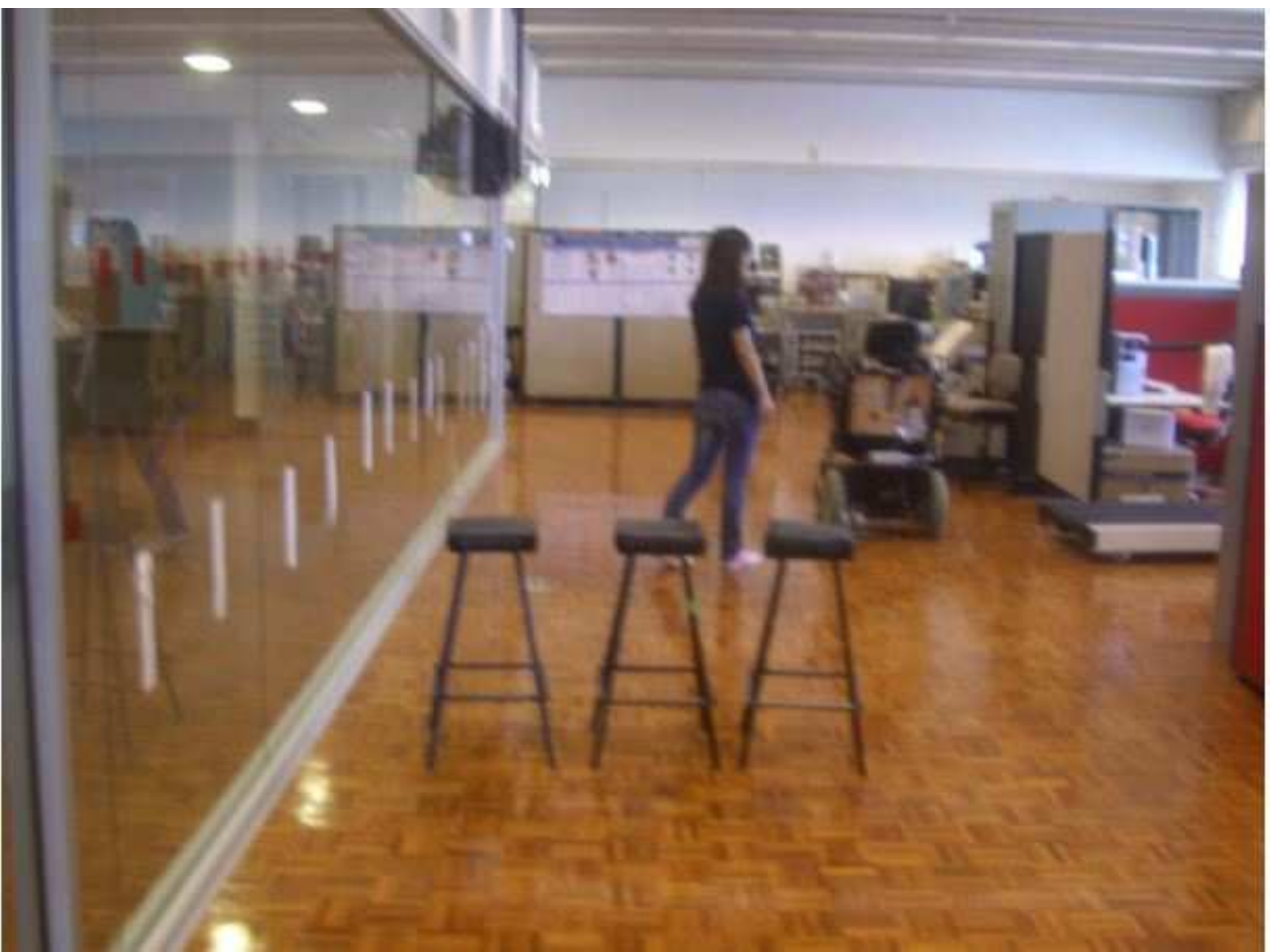}}
			\label{c6.exp62}}
			\subfigure[]{\scalebox{0.30}{\includegraphics{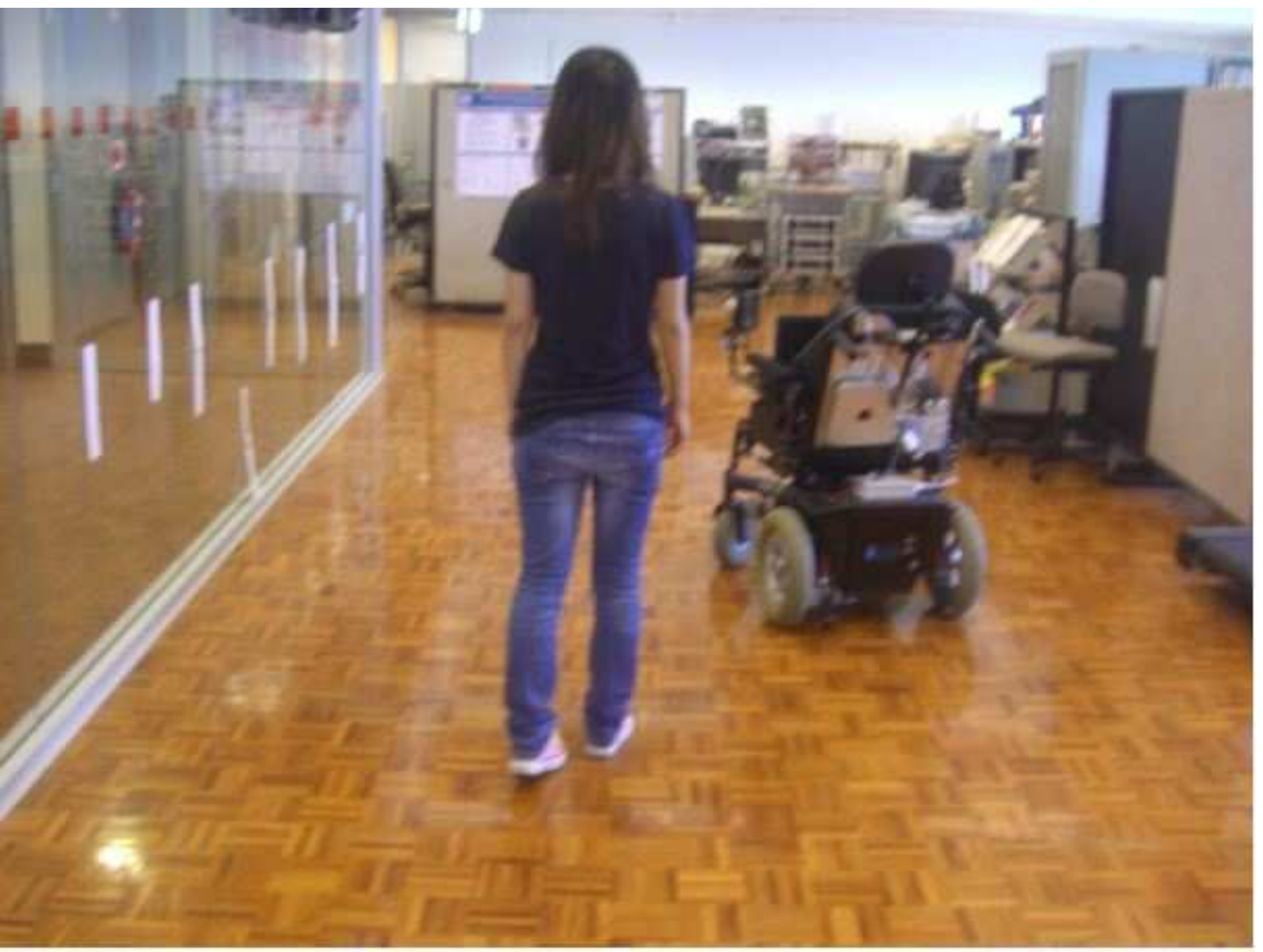}}
			\label{c6.exp63}}
			\subfigure[]{\scalebox{0.30}{\includegraphics{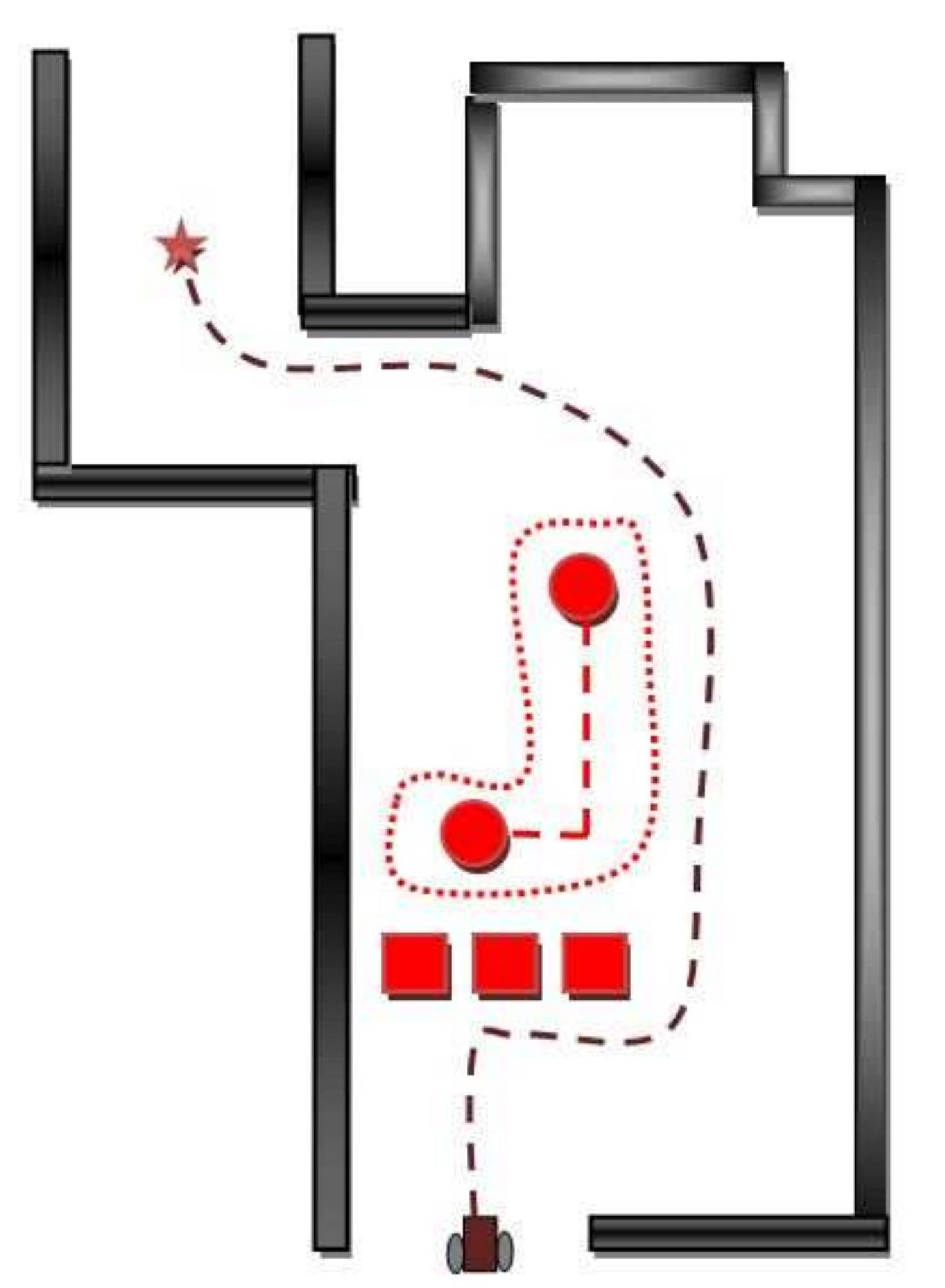}}
			\label{c6.exp64}}
			\caption{Wheelchair navigating among stationary and dynamic obstacles}
			\label{c6.exp6}
			\end{figure}
			\par

			In the next experiment, the wheelchair encounters a moving obstacle (the personal) just after bypassing the chain of chairs. This moving obstacle traverses the path from the wheelchair to the target and moves towards the target. In this scenario, the wheelchair is still able to track the $d_0$-equidistant curve of the moving obstacle, which is shown by red dashed line around the obstacle, see Fig.~\ref{c6.exp62} and Fig.~\ref{c6.exp63}. The complete path of this experiment is shown in Fig.~\ref{c6.exp64}.

\begin{figure}[!h]
			\centering
			\subfigure[]{\scalebox{0.30}{\includegraphics{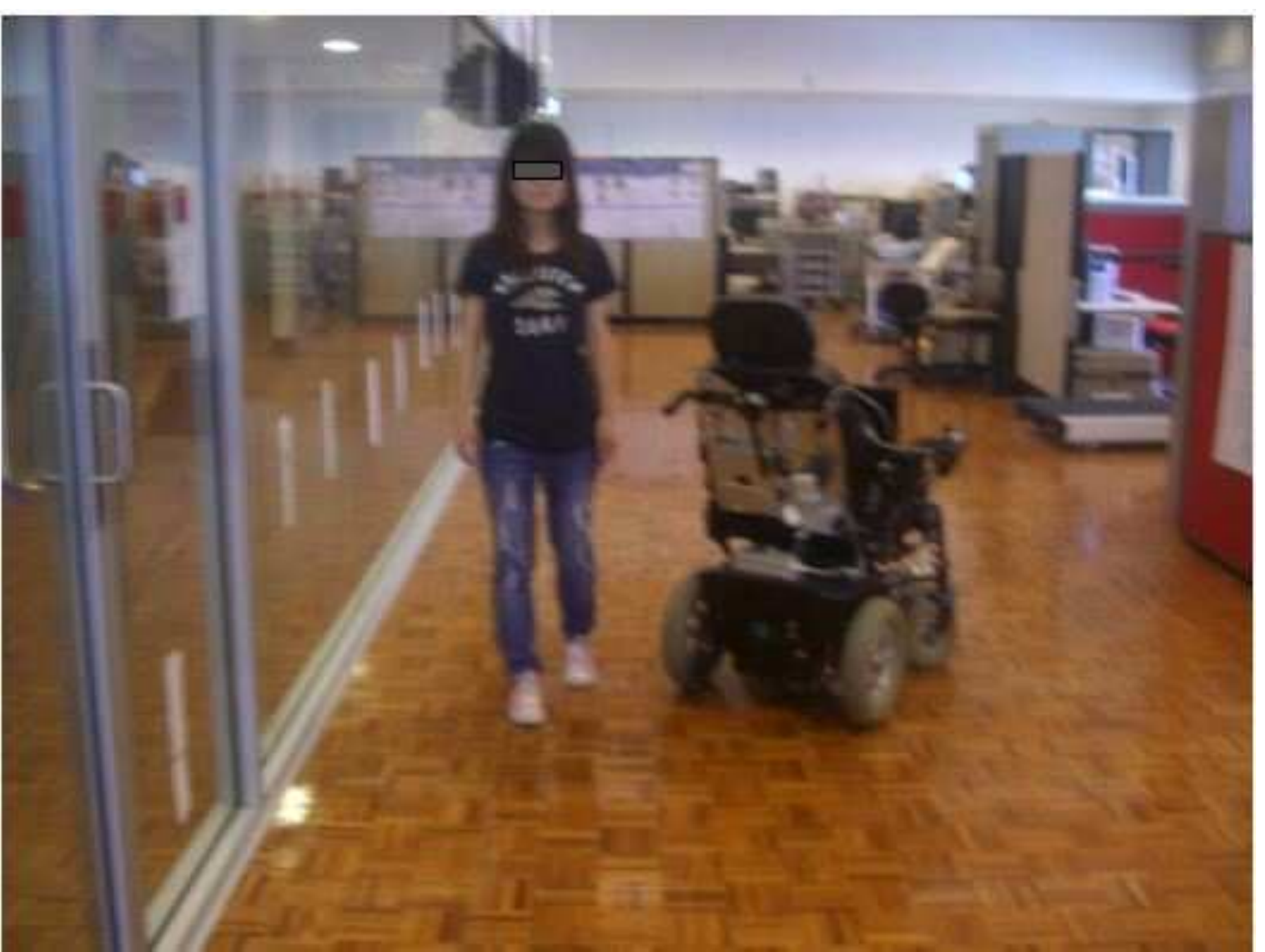}}
			\label{c6.exp71}}
			\subfigure[]{\scalebox{0.30}{\includegraphics{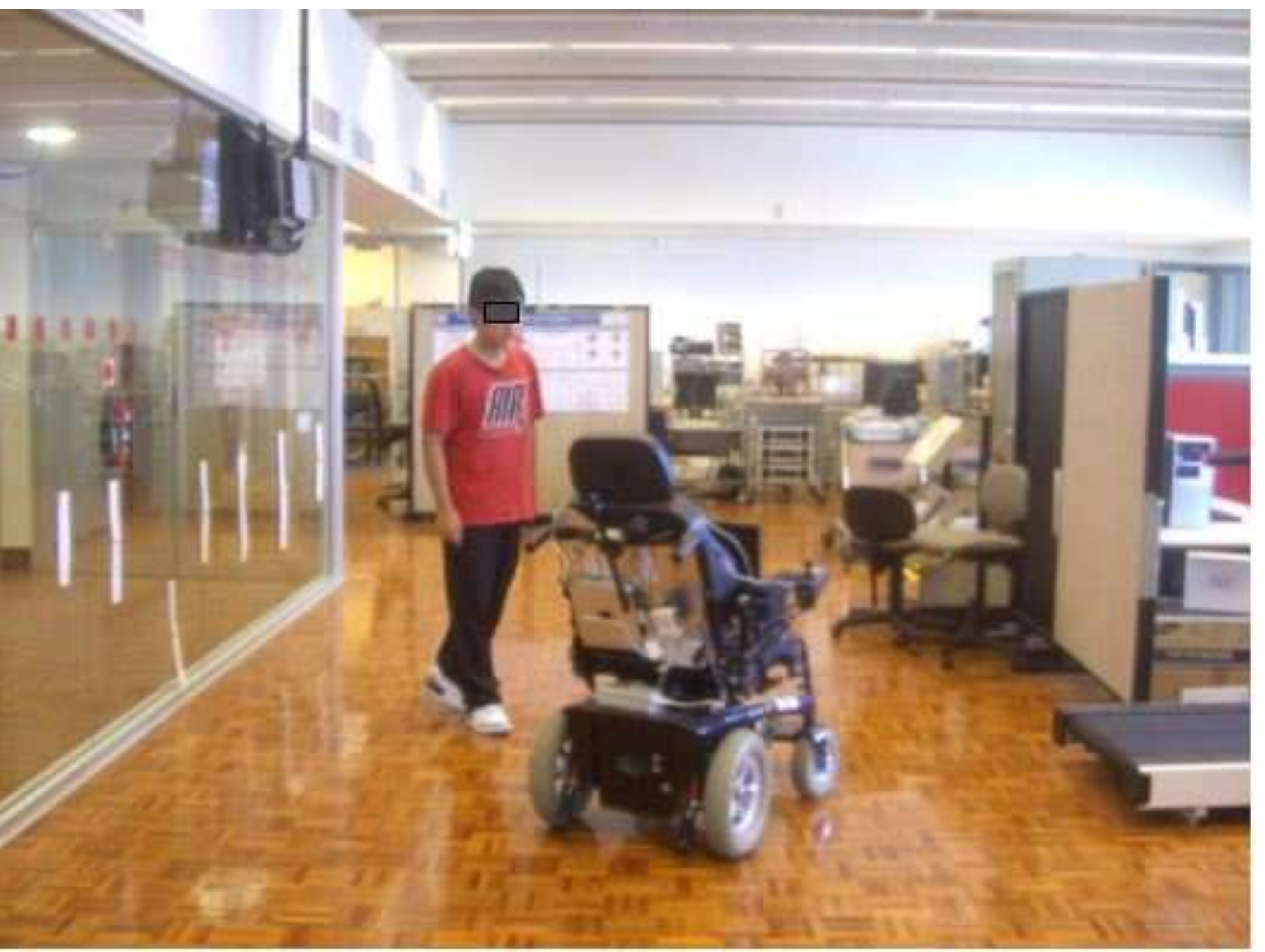}}
			\label{c6.exp72}}
			\subfigure[]{\scalebox{0.30}{\includegraphics{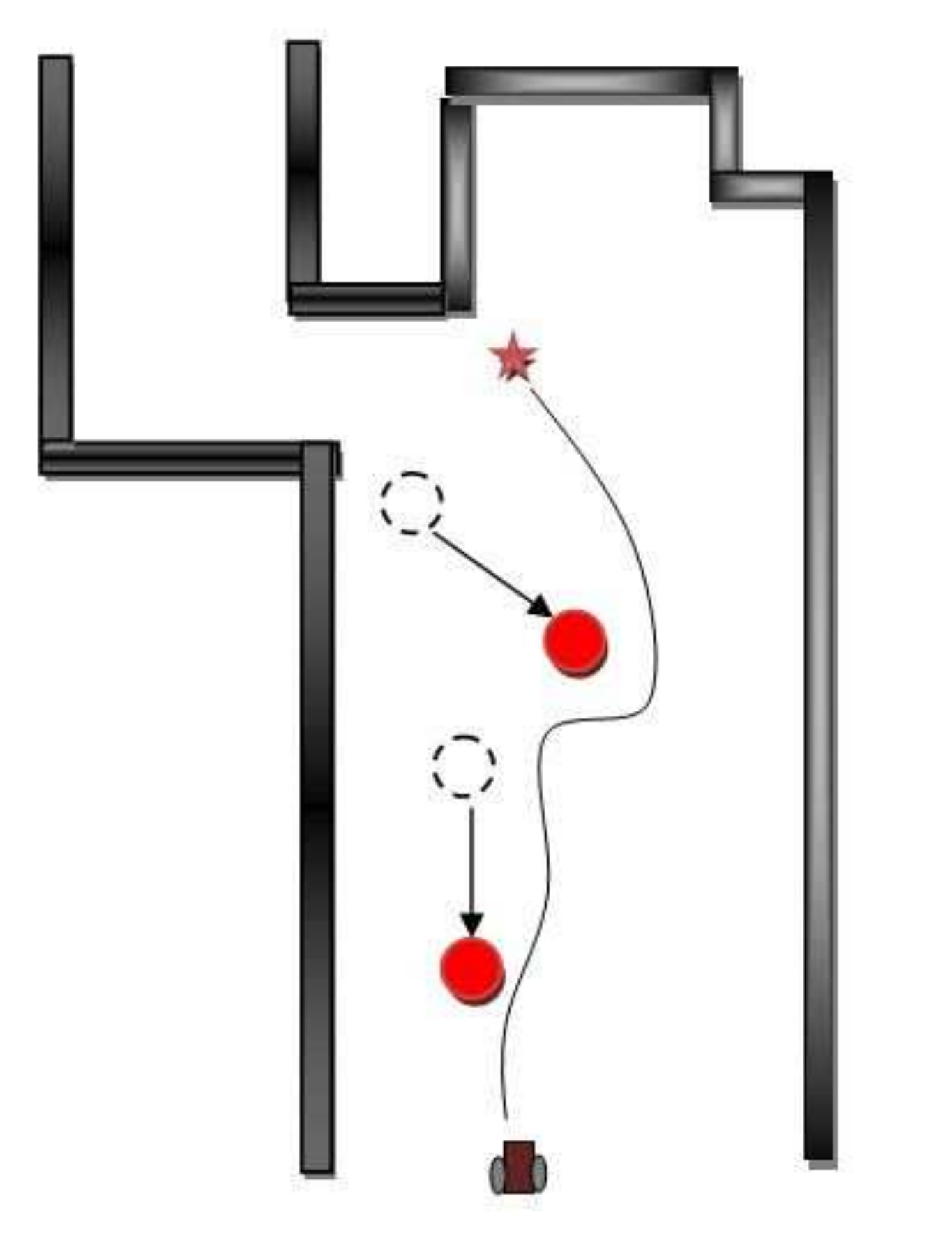}}
			\label{c6.exp73}}
			\subfigure[]{\scalebox{0.30}{\includegraphics{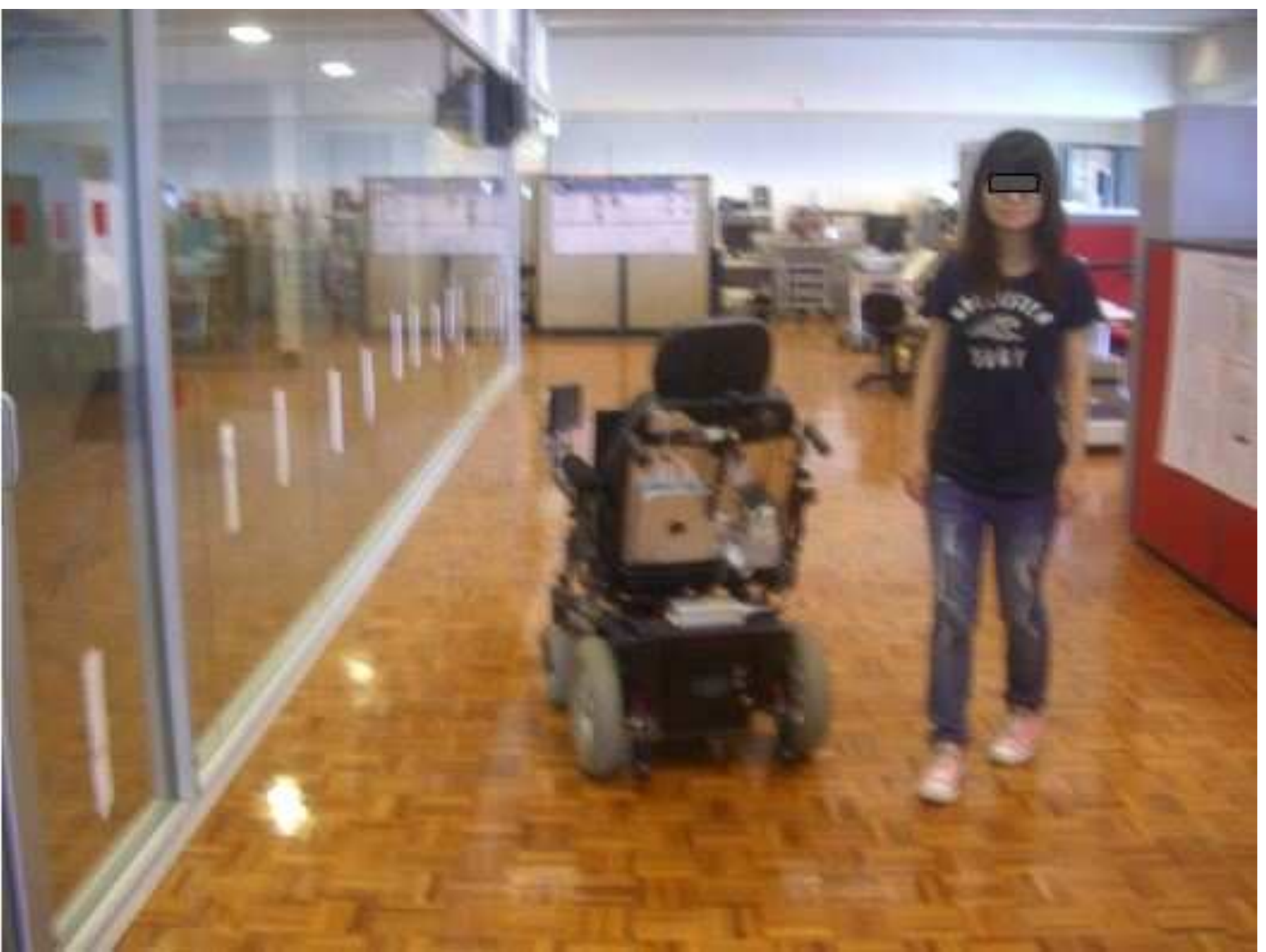}}
			\label{c6.exp74}}
			\subfigure[]{\scalebox{0.30}{\includegraphics{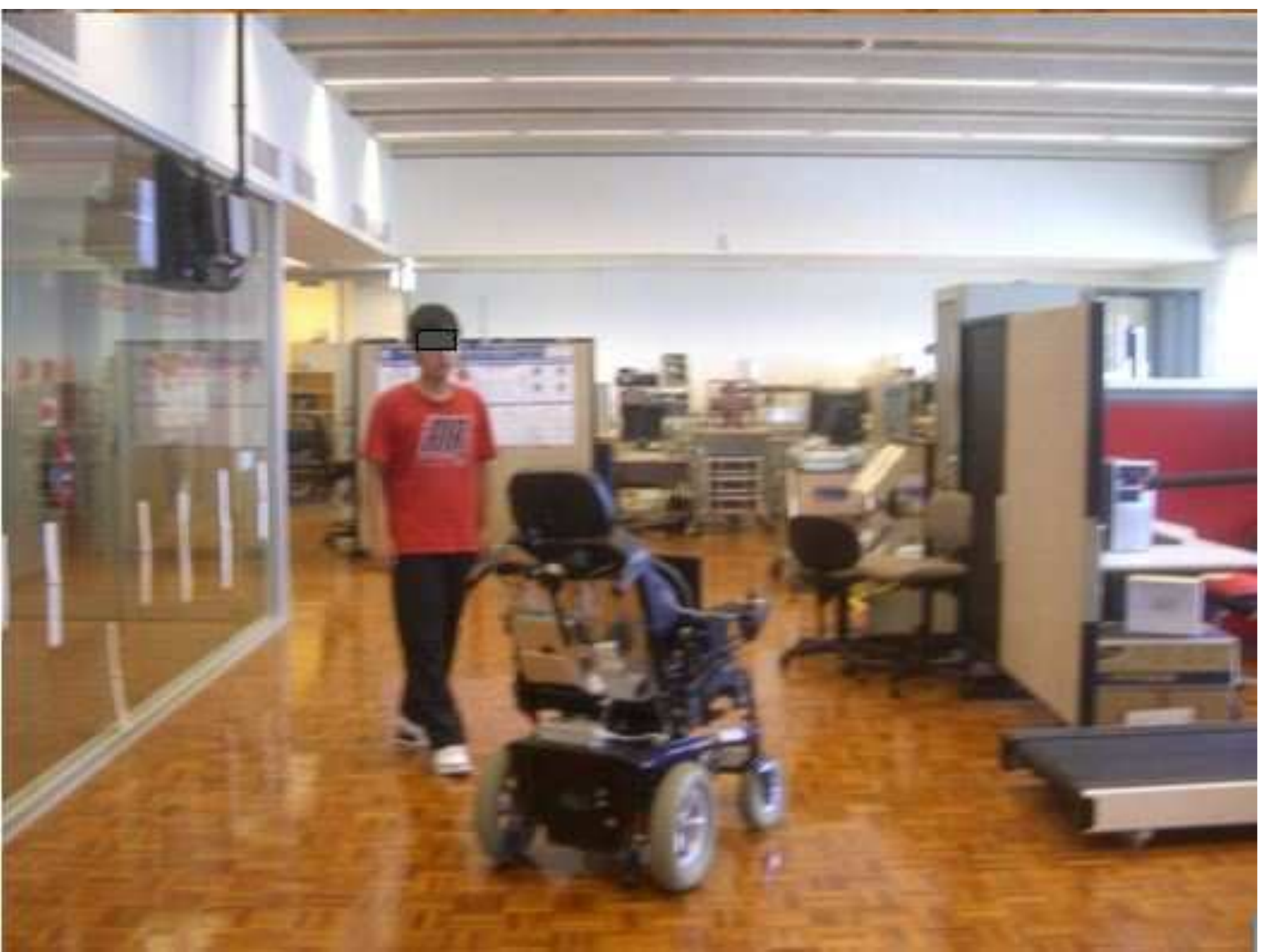}}
			\label{c6.exp75}}
			\subfigure[]{\scalebox{0.30}{\includegraphics{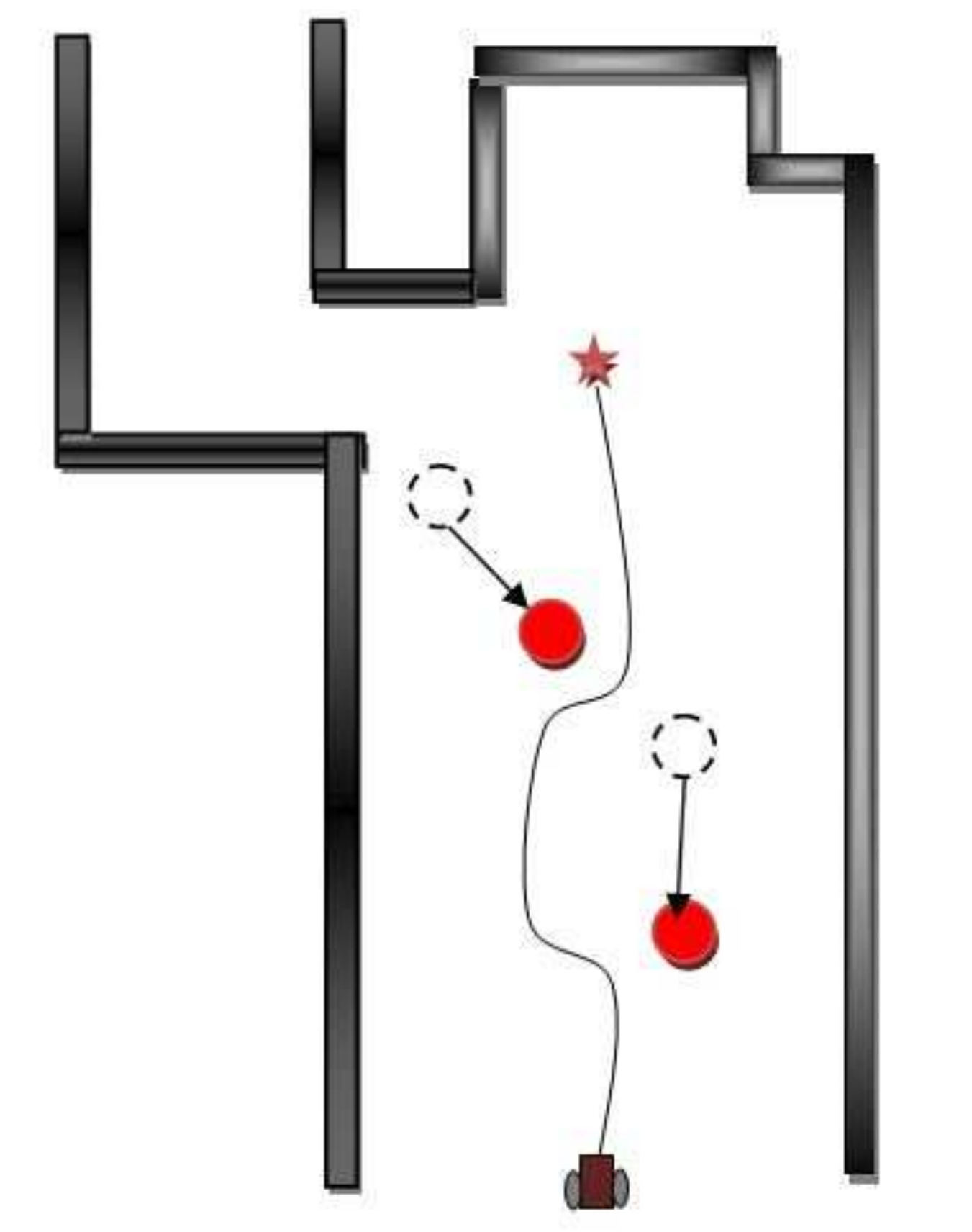}}
			\label{c6.exp76}}
			\caption{Wheelchair avoids multiple moving obstacles}
			\label{c6.exp7}
			\end{figure}

			The capability of the proposed navigation algorithm to avoid multiple moving obstacles is examined in Fig.~\ref{c6.exp7}. In these experiments, the obstacle moves in random directions as shown in Fig.~\ref{c6.exp73} and Fig.~\ref{c6.exp76}. The wheelchair is still able to arrive at the target while avoiding all the obstacles. These experiments are repeated many time with the obstacles moving in various directions, the safety of the wheelchair is always guaranteed under the guidance of the ENA.

			\par

		\section{Summary}

		The implementation of BINA and ENA on the SAM wheelchair are presented in this chapter. Both of the algorithm ensures the safety of the wheelchair in static and dynamic environments with multiple obstacles. The performance of the algorithms has been confirmed by experiments with SAM wheelchair in real life scenarios.
\chapter {The Implementations on an Autonomous Intelligent Hospital Bed} \label{C7}

	The hospital beds had evolved from some crude stretchers to sets of convenient multi-functional medical devices over the centuries. Although we have seen these great improvements on the parts of hospital beds, most of the navigation tasks are still achieved manually. The manual steering of the hospital beds is the earliest and simplest way to control the motion of the hospital beds, but it may not be the most efficient one. 
\par
	 The existing hospital bed systems which are only motorised to adjust their own positions to a certain extend such as Trendlenburg, up, down positions, none of them is motorised for transportation. Designing an intelligent hospital bed system with obstacle avoidance ability is a good solution to the substantial labor required to transport the hospital bed. In this chapter, we present the implementations of Biologically-Inspired Navigation Approach (BINA) and Equidistant Navigation Approach (ENA), which are proposed in Chapter~\ref{C2} and Chapter~\ref{C3}, respectively, on an automated hospital bed FlexBed. The implementations allow Flexbed to avoid en-route obstacles and arrive at target location safely. The performance of the Flexbed with the proposed navigation algorithms is confirmed by real life experiments.

		\section{Background and Motivation}

			The hospital beds had been widely used in hospitals to aid the transpiration the patients who need different types of medical attentions. Over the centuries, the hospital beds had evolved from some crude stretchers to sets of convenient multi-functional medical devices: various features and functions such as side rails, wheels, adjustable sections for different parts of body, etc, see e.g. \cite{FLD92,RK95,TN65,HWV85,WTN96}. These improvements provides better services to patients and conveniences to medical care personnels.
\par
			The most important purpose for utilising hospital beds is to safely transport the patients from current location to the target location. However, despite these great improvements on various parts of hospital beds, the navigation system of the hospital beds has never been improved. The motion of the hospital beds are controlled by manual steering of the hospital personnels. Manual steering is the earliest and simplest method to drive the hospital beds, but it may not be the most efficient and reliable method nowadays. The most convincing reason is that the performance of the transportation tasks by manual steering heavily rely on the personnels' reaction and decision which is subject to various factors such as the personnels' concentration on the navigation task, proficiency on operating hospital beds during transportation, etc. Reports and studies has shown that human-based mistake is a major contributing factor for the potential and unexpected risks of hospital beds transportation within hospitals, especially for critically ill patients \cite{BUG04,PJP07}. Furthermore, the force or torque provided by the hospital personnels may not be sufficient to steer the hospital bed due to its heavy load, especially in the emergency case. The implementation of navigation strategies on different transportation facilities is proven to be successful for improving the safety and efficiency of the facilities, e.g. wheelchairs, transport robots. Therefore, there is no doubt that the performance of the hospital beds can be greatly boosted by helps of a safe and reliable autonomous navigation strategy.
\par
			None of the existing hospital bed system is considered as autonomous intelligent system with obstacle avoidance ability up to date, although some of them are motorised so that they are able to adjust their own position to a certain extend. The implementation of the algorithm BINA and ENA allows the hospital beds to safely navigate in a hospital environments with stationary and dynamic obstacles. The BINA allows the hospital beds to efficiently avoid obstacle and it is easy to compute. The ENA allows the hospital bed to navigate in a variety of unknown environments due to its flexibility and it is cost-efficient. These two algorithm are able to work independently to drive the hospital beds to the target location without collisions. They can also be integrated into other control schemes, for example, when there is no thread of collision, the motion of the hospital beds are controlled by some global navigation algorithms or Ipad controllers. The algorithms take over the control of hospital beds when the sensory system detects some obstacles within certain range.

		\section {System Model}

			We consider a hospital bed modeled as unicycle. The mathematical model of the hospital bed is as follows:

			\begin{equation}
			\label{ch7:1}
			\begin{array}{l}
			\dot{x} = v(t) \cos \theta,
			\\
			\dot{y} = v(t) \sin \theta,
			\\
			\dot{\theta} = u 
			\end{array},~~~ 
			\begin{array}{l}
			x(0) = x_0,
			\\
			y(0) = y_0,	
			\\
			\theta(0) = \theta_0
			\end{array} 
			\end{equation}
			where 
			\begin{equation}
			\label{ch7:max}
			 u\in [-U_{max},U_{max}],~~~v(t)\in [0,V_{max}].
			\end{equation} 

			Here the position of the wheeled mobile robot is represented by the absolute Cartesian coordinates as $(x,y)$ of the reference point, which is the center of mass of the hospital bed. The orientation of hospital bed, measured in counterclockwise direction from the reference axis, is given by $\theta$. $v(t)$ and $u(t)$ are the speed and angular velocity of the hospital bed. $V_{max}$ and $U_{max}$ are the non-holonomic constraints on the speed and angular velocities, which limits the minimum turning radius of the hospital bed by:
			\begin{equation}
			\label{ch7:mini}
			R = v/U_{max}.
			\end{equation}
			\par

		\section {Hospital Bed System Description}

\begin{figure}[h]
			\centering
			\includegraphics[width=5.0in]{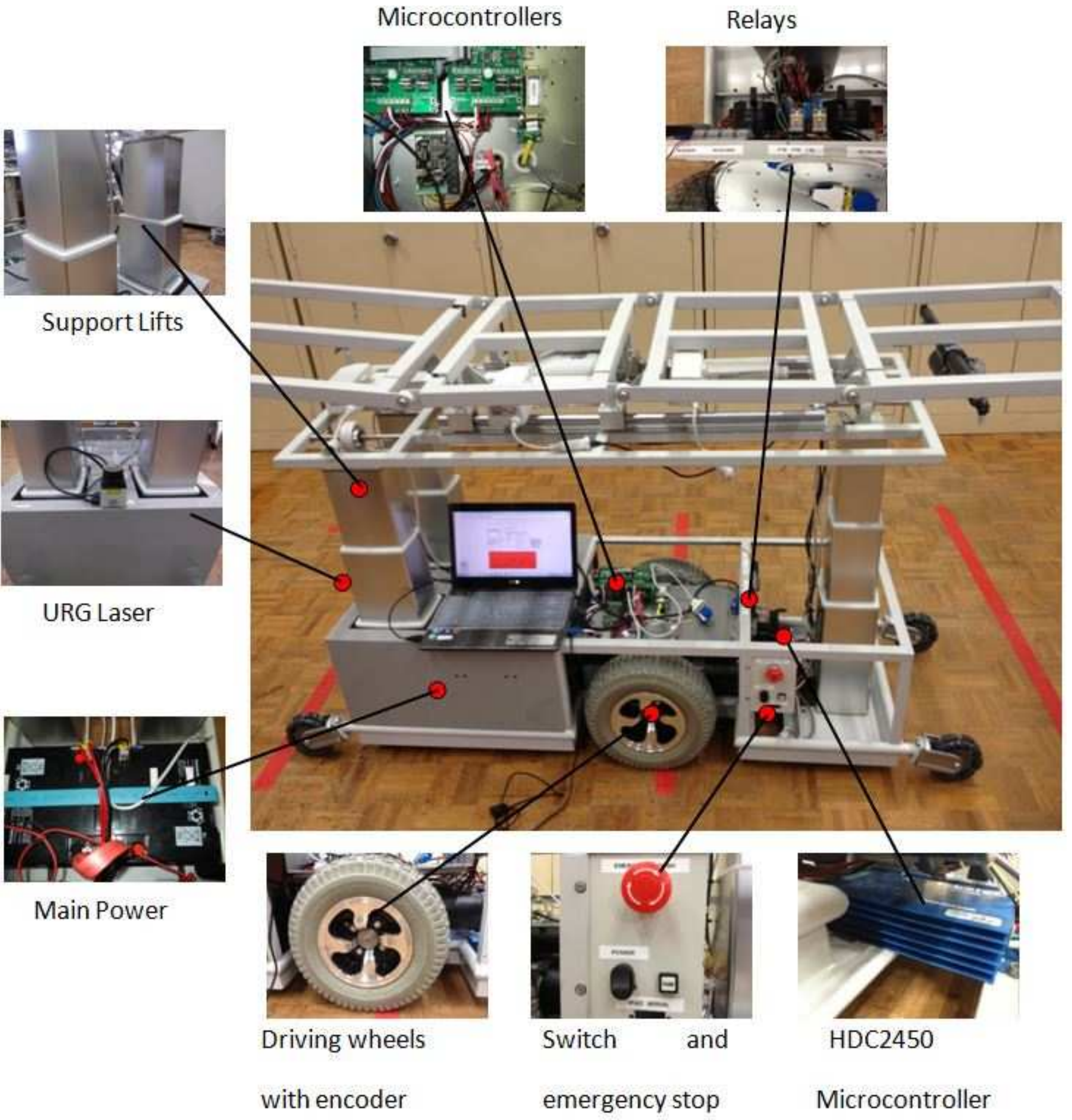}
			\caption{Key components in Flexbed}
			\label{bed}
			\end{figure}

\begin{figure}[h]
			\centering
			\includegraphics[width=4.5in]{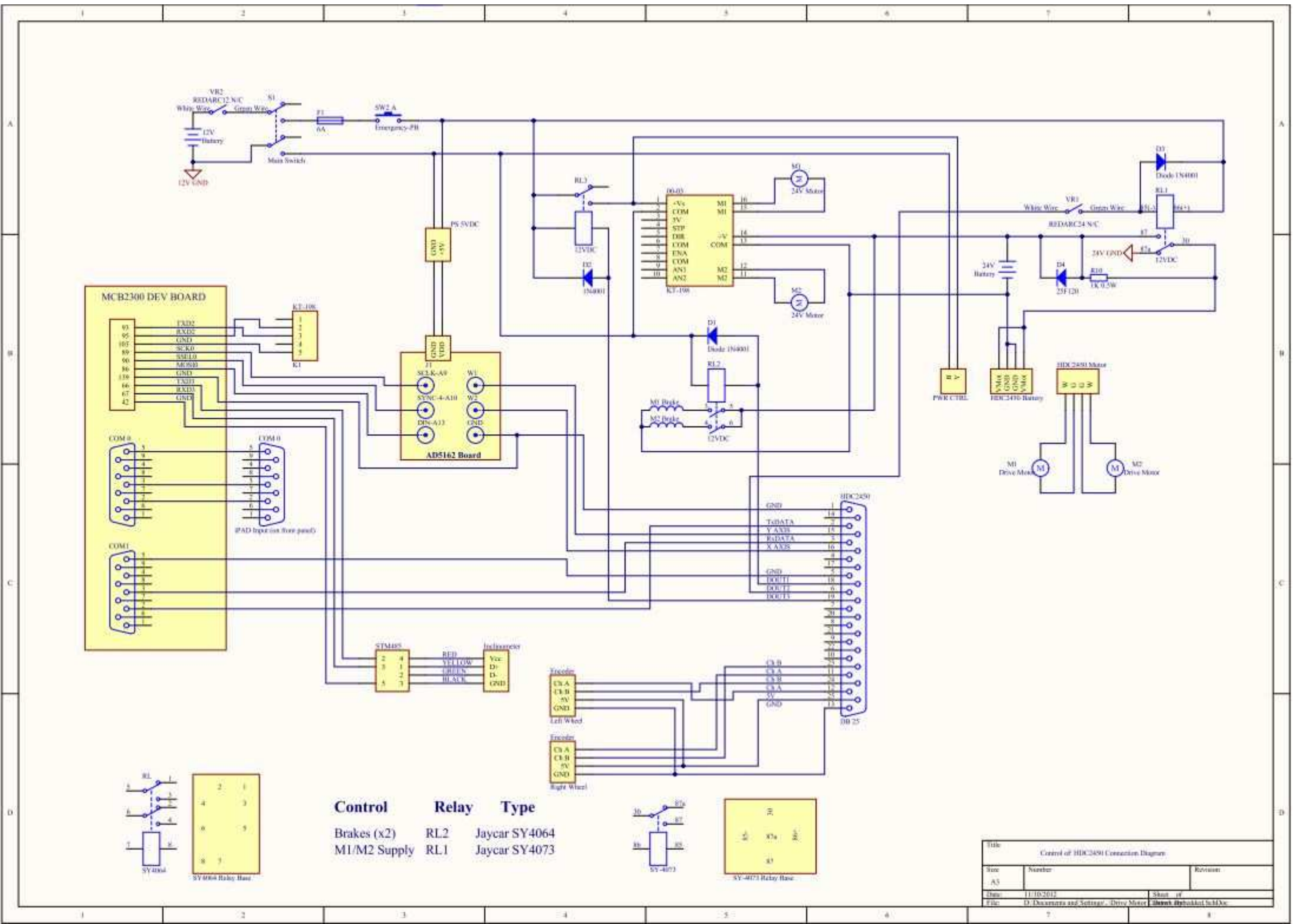}
			\caption{Schematic for the motor control system of Flexbed}
			\label{motor}
			\end{figure}

			The BINA and ENA are implemented on a hospital bed, named Flexbed. Flexbed features two driving wheels located at the sides of the base section, two supportive wheels in the front and the rare sides of the base section. The encoders (model 775) has been attached to the driving wheels to measure the distance traveled by both wheels. There are totally three support lifts upon the base section, these lifts are able to support up tp $150kg$ of weight. They contain DC actuators for extension and contraction movements which alls the personnel to adjust the bed position to Trendelenburg, Anti-Trendelengurg, tilt to right or to left mode from motor controller implemented in ipad. Ipad can also used to control the motion of Flexbed. The available connection between ipad and the Flexbed is RS232 serial cable or WIFI.  The motors for driving Flexbed and the actuators are powered from two $12 volts$ batteries in series, all other electronics are supplied from a separate $12 volts$ battery. The power switch is located at one side of Flexbed together with the emergency stop bottom. Flexbed uses various sensor devices, such as laser range finder and Microsoft Kinect, to interact with the real world. The appearance of Flexbed and the position of these key components are shown in Fig.\ref{bed} and the schematic for the the motor control system of Flexbed is presented in Fig.\ref{motor}.
\par

			Some of these components of Flexbed are particularly critical in the experiments, therefore they are described in more details as follows:

			\begin{itemize}

				\item A notebook (Windows 7 operating system with dual-core CPU running at $2.66 GHz$ and $4GB$ RAM) with the proposed navigation algorithm BINA and ENA implemented in LabView is the core to control the Flexbed system. It acquires all the necessary data from various devices to compute the control signals which is sent to the motor system to control the movement of the Flexbed.

				\item The URG-04LX laser is mounted on the middle at the front of the Flexbed. The laser scans the environment and provides necessary information to the wheelchair such as distance to the obstacles (required by both BINA and ENA), the vision cone (required by BINA) etc.  The maximum scan angle for this laser range finder is $240 degree$ and the maximum scan range is $4m$ with accuracy of $\pm 1\%$ of the measurement, the scan frequency is $10 Hz$. Furthermore, the laser is able to scan and provide two-dimensional map of the environment if necessary.

				\item The encoders that are attached to the driving wheels of the Flexbed are Model 775 Slim Thru-Bore encoders. Each encoder generate two square waves in incremental quadrature mode which provides the distance traveled by each wheel (Channel A leading B for clockwise rotation viewed from mounting face and vice versa). The rising time for these waves are less than $1 ms$. The position $(x,y)$ and the orientation $\theta$ of the Flexbed can be estimated from the distance traveled by both wheels.

				\item The micro-controller RobotQ HDC2450 is responsible for receives information (distances, velocities, accelerations, power etc) from the encoders and send to control signals to the motor. It has four inputs for two incremental encoders to be connects. The control signals are sent to the motor system via serial interface or USB interface. This micro-controller comes with a utility software called Robotrun Plus. This utility allows the user to check the connection on analogue or digital inputs/outputs pins and set the speed of the wheels to perform some simple motions. More importantly, it is able to adjust the power being sent to the wheels which controls the motions of Flexbed. It also allows the users to change the maximum speed and angular velocity of the Flexibed. The build-in PID controller allows the users to optimise the performance of the encoders and the motor.

			\end{itemize}

	\section {Experiments of Flexbed with Biologically-Inspired Navigation Algorithm (BINA)}

			In this section, we present the experiments of navigating Flexbed in various real life scenarios under the guidance of BINA. The data acquisition method has been introduced in~\ref{para_meas1} with the use of same URG-04LX laser range finder.
\par
\begin{figure}[!h]
			\centering
			\subfigure[]{\scalebox{0.3}{\includegraphics{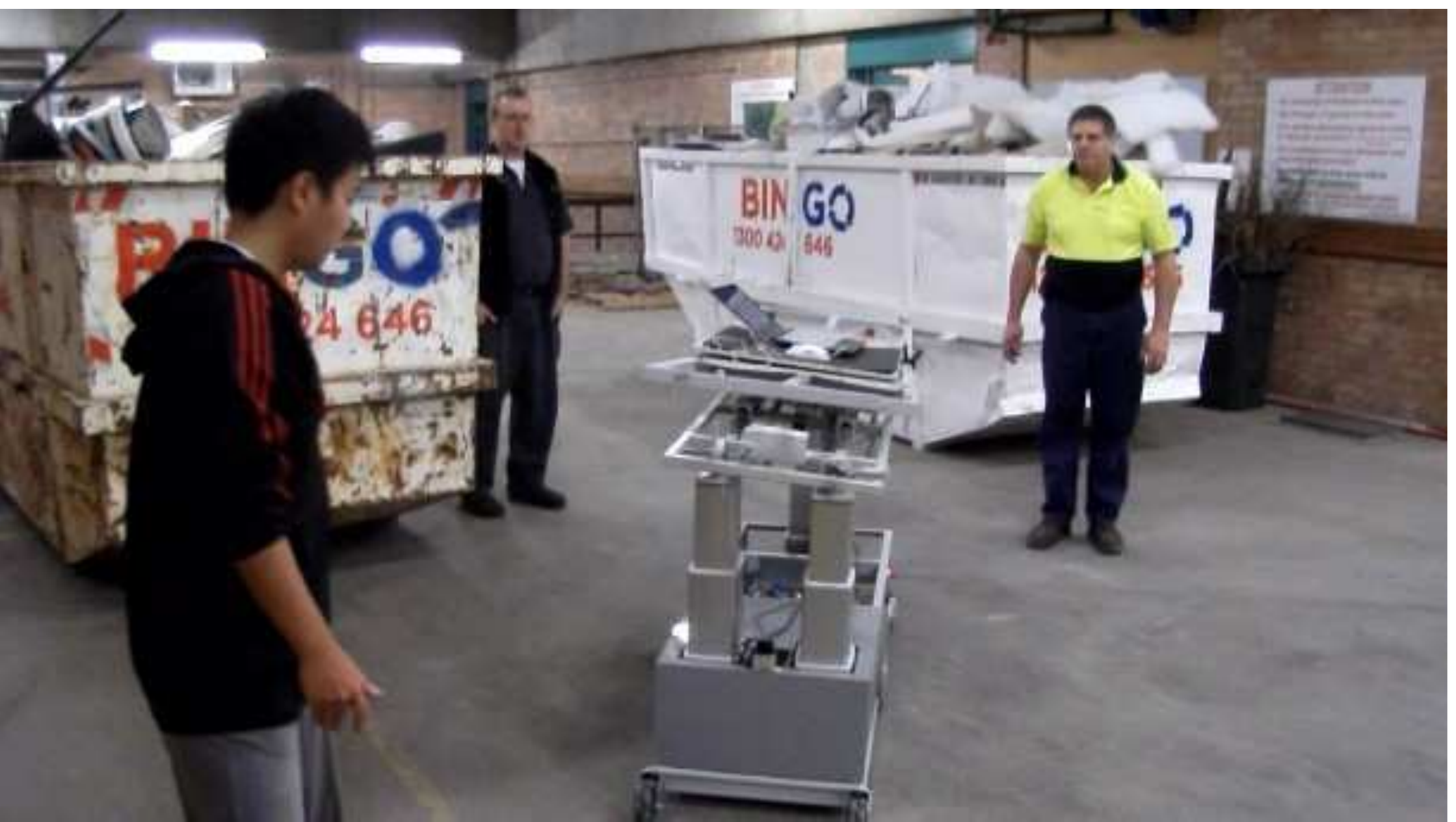}}
			\label{c7.exp11}}
			\subfigure[]{\scalebox{0.3}{\includegraphics{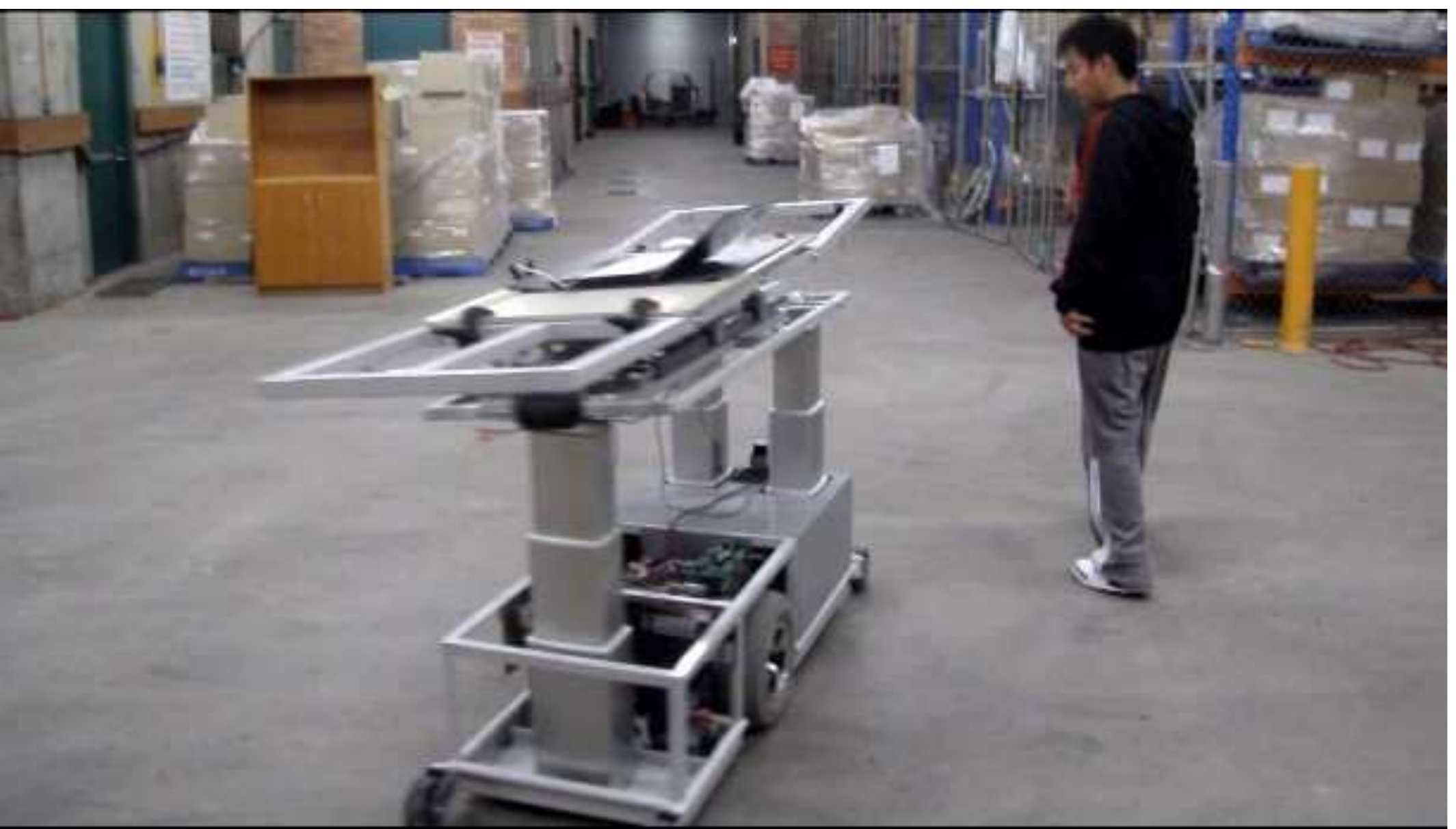}}
			\label{c7.exp12}}
			\subfigure[]{\scalebox{0.3}{\includegraphics{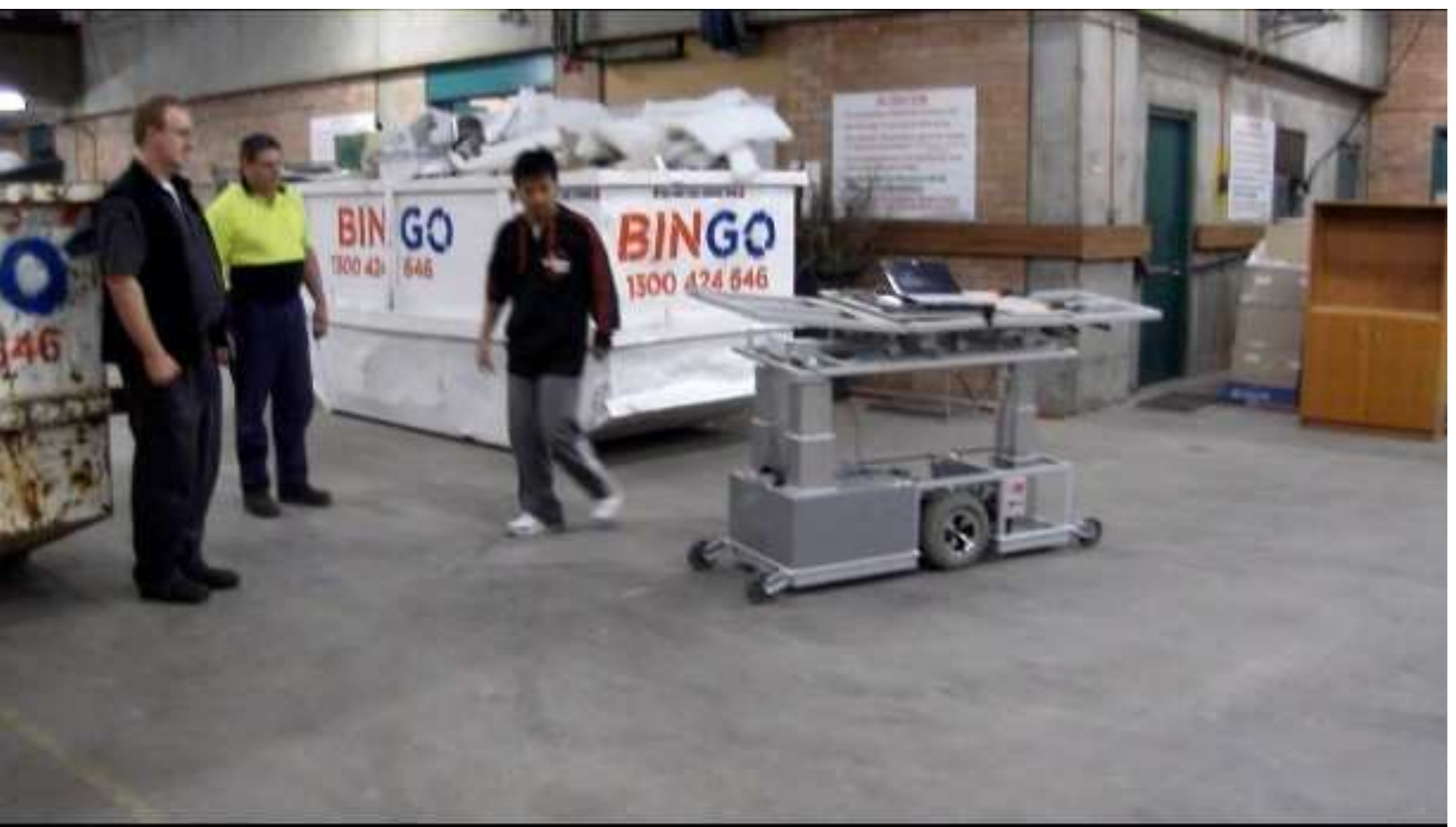}}
			\label{c7.exp13}}
			\subfigure[]{\scalebox{0.4}{\includegraphics{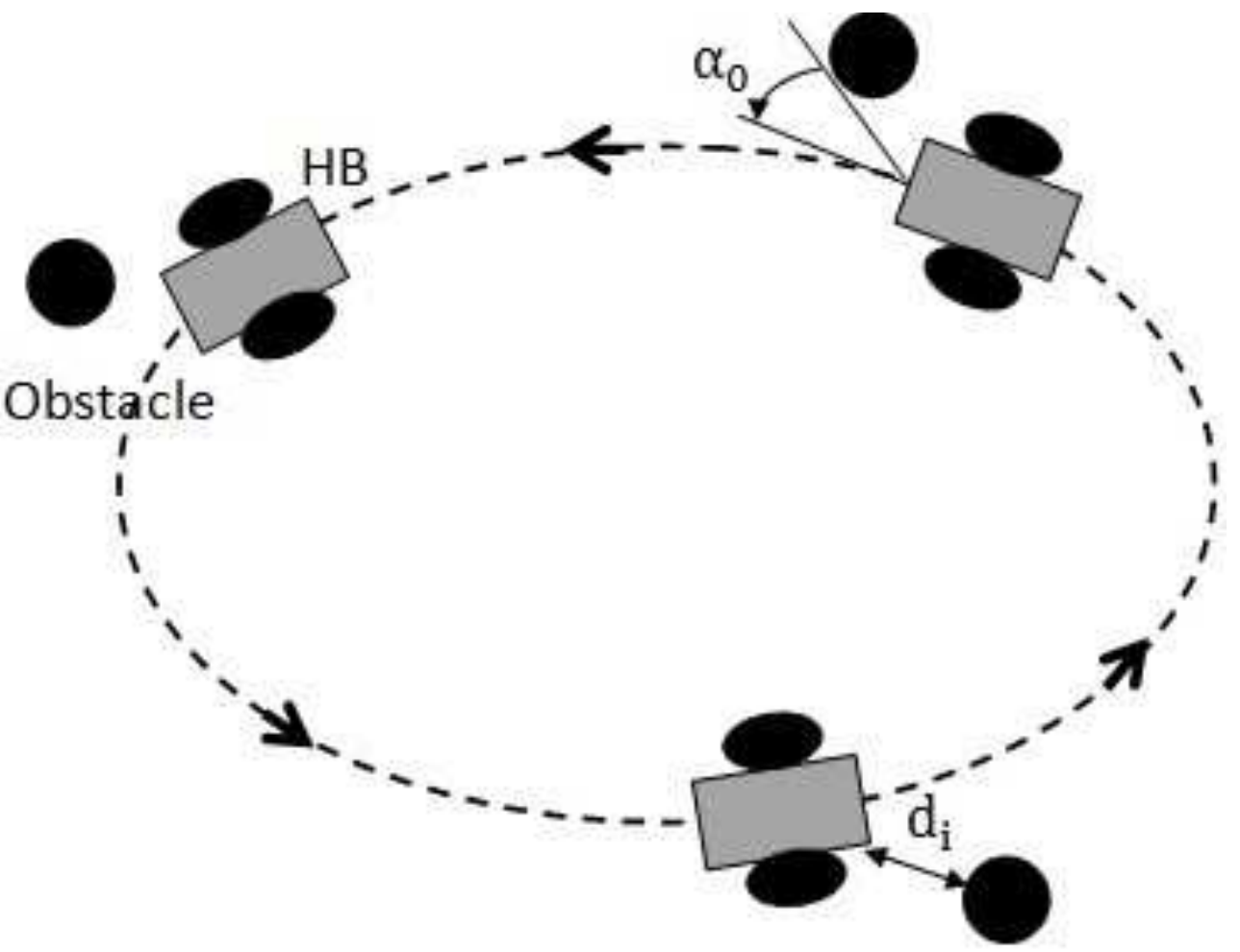}}
			\label{c7.exp14}}
			\subfigure[]{\scalebox{0.50}{\includegraphics{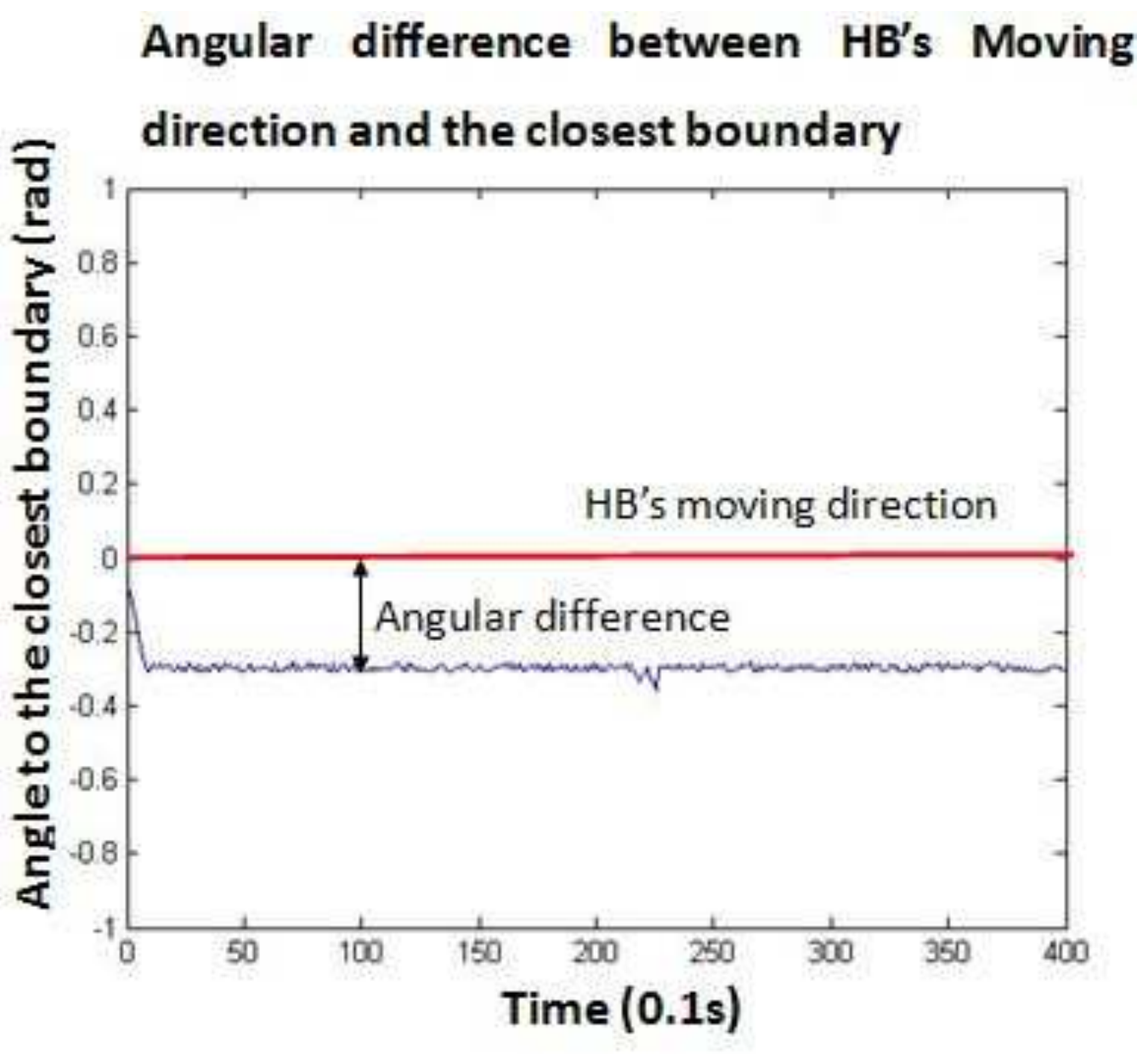}}
			\label{c7.exp15}}
			\subfigure[]{\scalebox{0.50}{\includegraphics{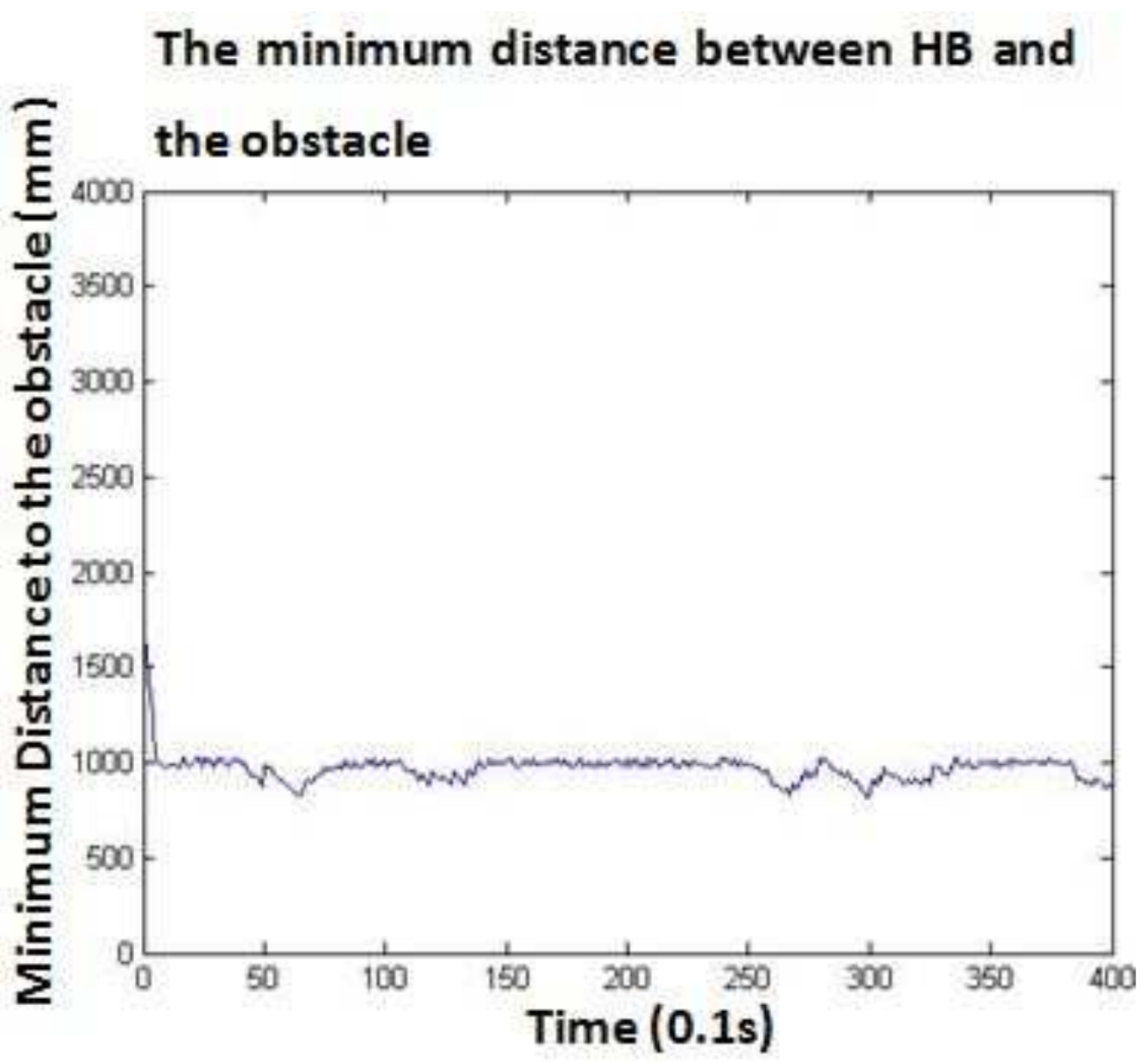}}
			\label{c7.exp16}}
			\caption{safety measurements during obstacle avoidance maneuver}
			\label{c7.exp1}
			\end{figure}

			In the first experiment, we present and discuss two of the important safety measurements when Flexbed is guided by the obstacle avoidance law (\ref{bi_cont}). These two measurements are 1) the angular difference between  the instantaneous moving direction of the Flexbed and the closest boundary of the enlarged vision cone. 2) the minimum distance between the Flexbed and the obstacle. In this experiment, we deliberately position the experimenter within the sensor range of Flexbed and it is treated as an obstacle by Flexbed at all time. The snapshots of this experiment are shown in Fig.~\ref{c7.exp11}, Fig.~\ref{c7.exp12} and Fig.~\ref{c7.exp13}, the complete paths taken by the Flexbed and the obstacle are shown in Fig.~\ref{c7.exp14}. The angular difference between the instantaneous moving direction of the Flexbed and the closest boundary is shown in Fig.\ref{c7.exp15}, it can be observed that there is a constant avoiding angle between them (in this case, the mean value of this angle equals $-0.3287rad$ with standard deviation of 0.0551), which means the Flexbed will never collide into the Flexbed under the guidance of obstacle avoidance law (\ref{cont}). The safety of the Flexbed is further confirmed by the minimum distance between the Flexbed and the obstacle $d_i$ which is always greater than a certain safety distance $d_{safe}$ (in this case, the minimum value of $d_i$ equals $0.816m$) in Fig.\ref{c7.exp16}.
\par

			\par

\begin{figure}[h]
			\centering
			\subfigure[]{\scalebox{0.35}{\includegraphics{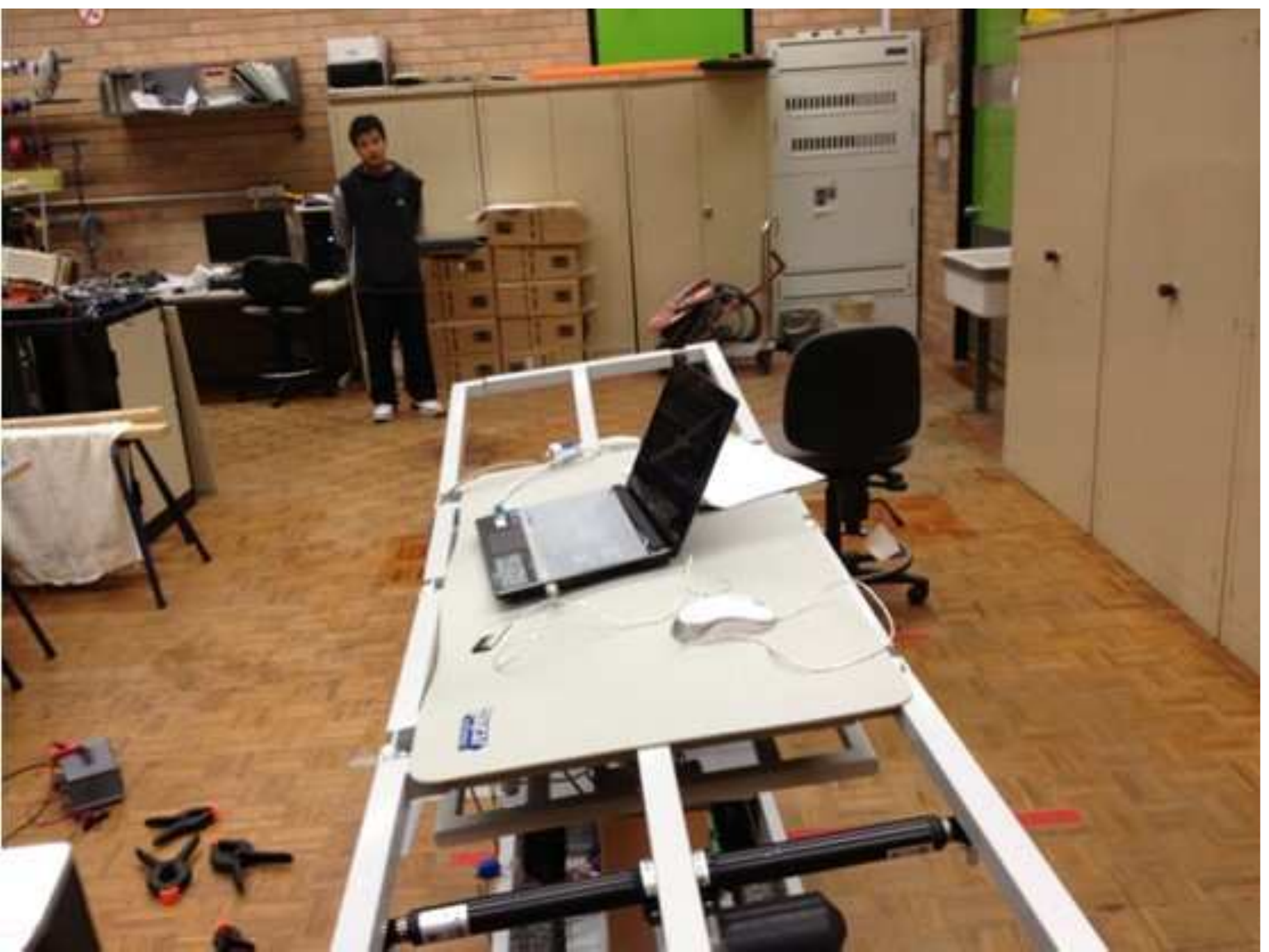}}
			\label{c7.exp21}}
			\subfigure[]{\scalebox{0.35}{\includegraphics{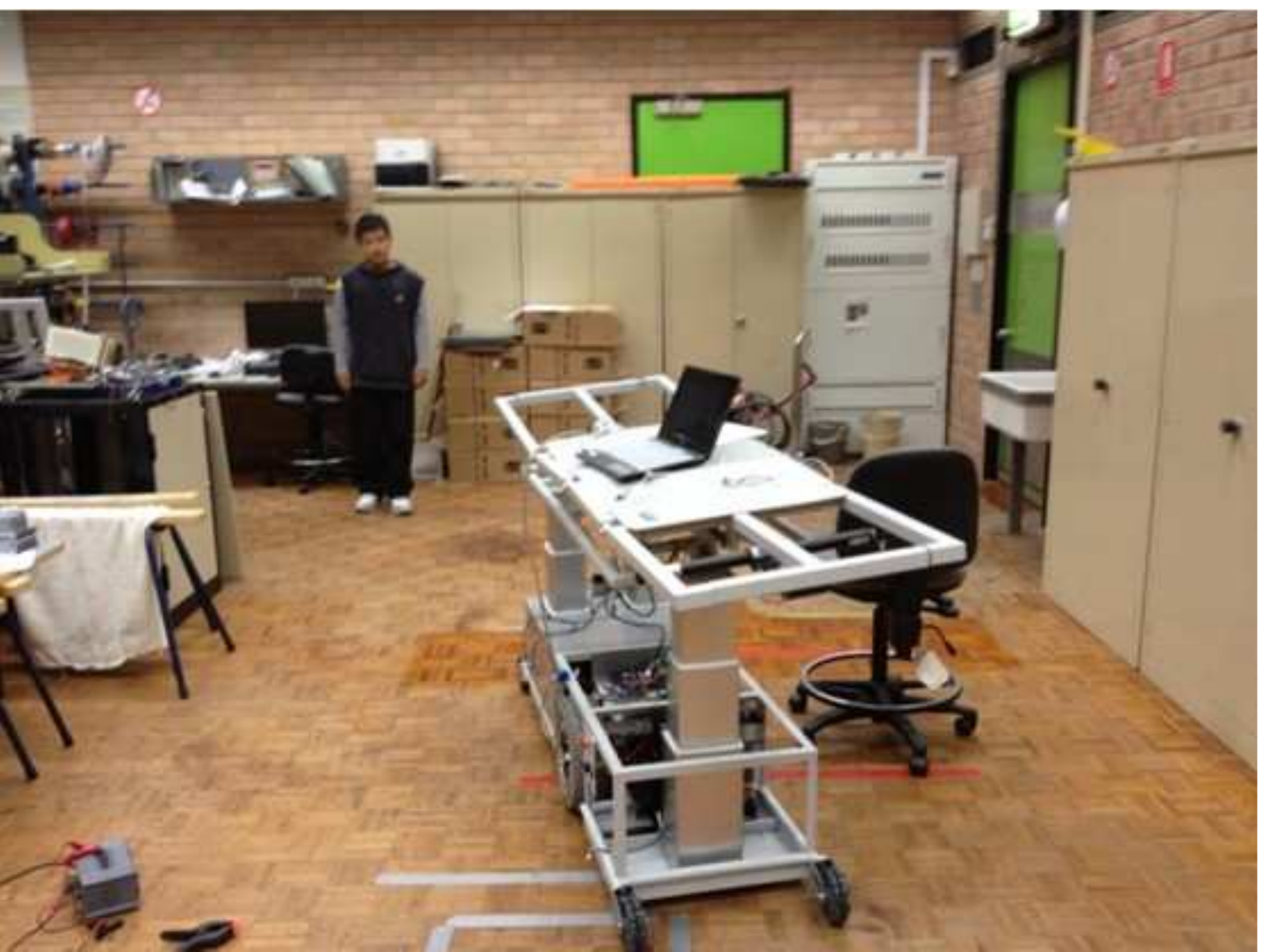}}
			\label{c7.exp22}}
			\subfigure[]{\scalebox{0.35}{\includegraphics{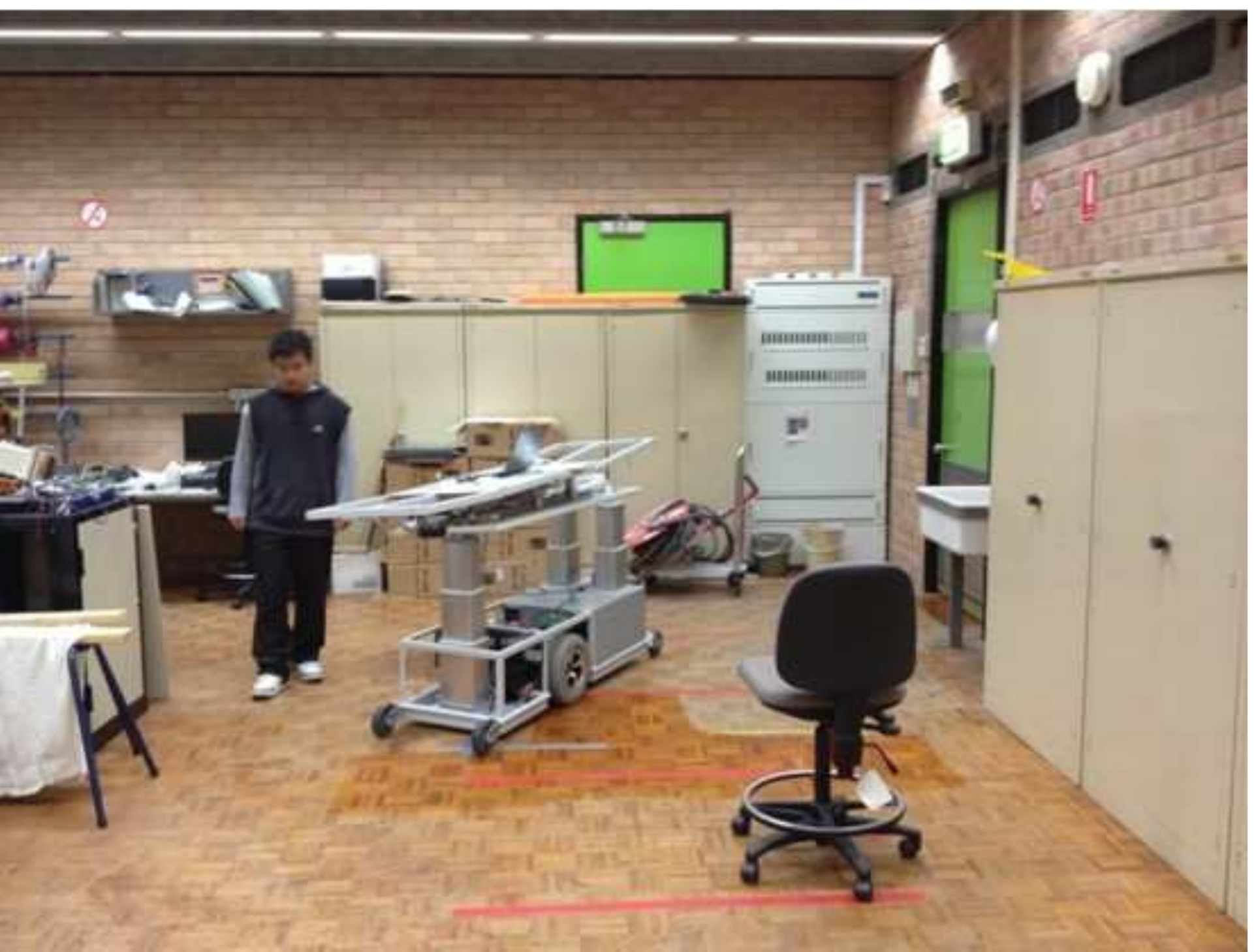}}
			\label{c7.exp23}}
			\subfigure[]{\scalebox{0.40}{\includegraphics{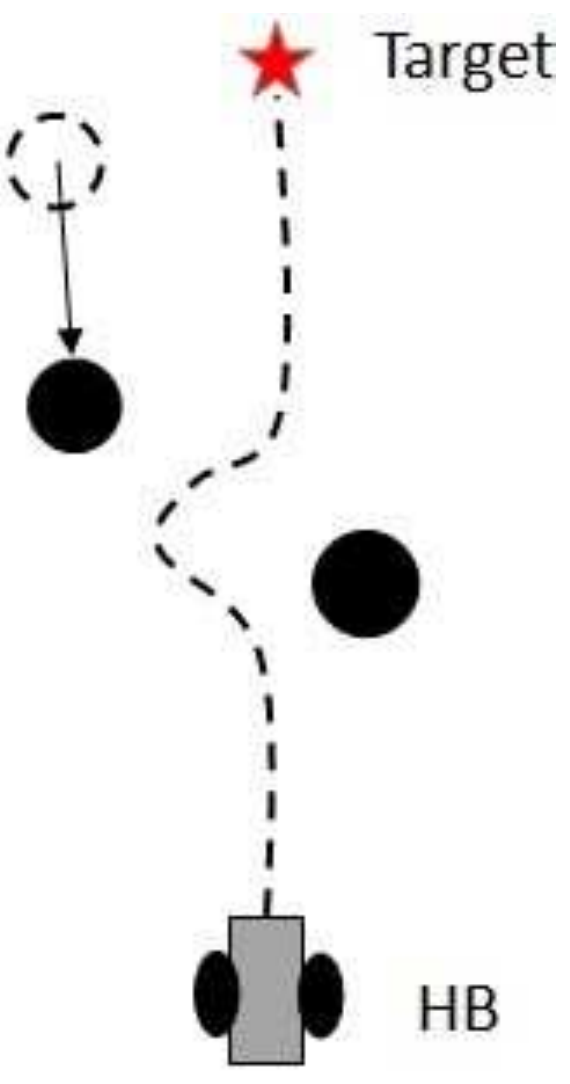}}
			\label{c7.exp24}}
			\caption{Flexbed avoids static and dynamic obstacles}
			\label{c7.exp2}
			\end{figure}

			The ability of BINA to navigate Flexbed in a simple environment with one stationary obstacle and one dynamic obstacle is shown in Fig.~\ref{c7.exp2}. Fig.~\ref{c7.exp22} and Fig.~\ref{c7.exp23} show the moment when Flexbed bypass these two obstacles. Flexbed arrive safely at target location in Fig.~\ref{c7.exp24}. This experiment proves the capability of BINA to guide Flexbed to target location while avoiding en-route obstacles.

			\par

\begin{figure}[!h]
			\begin{minipage}{.5\textwidth}	
			\centering	
			\subfigure[]{\scalebox{0.3}{\includegraphics{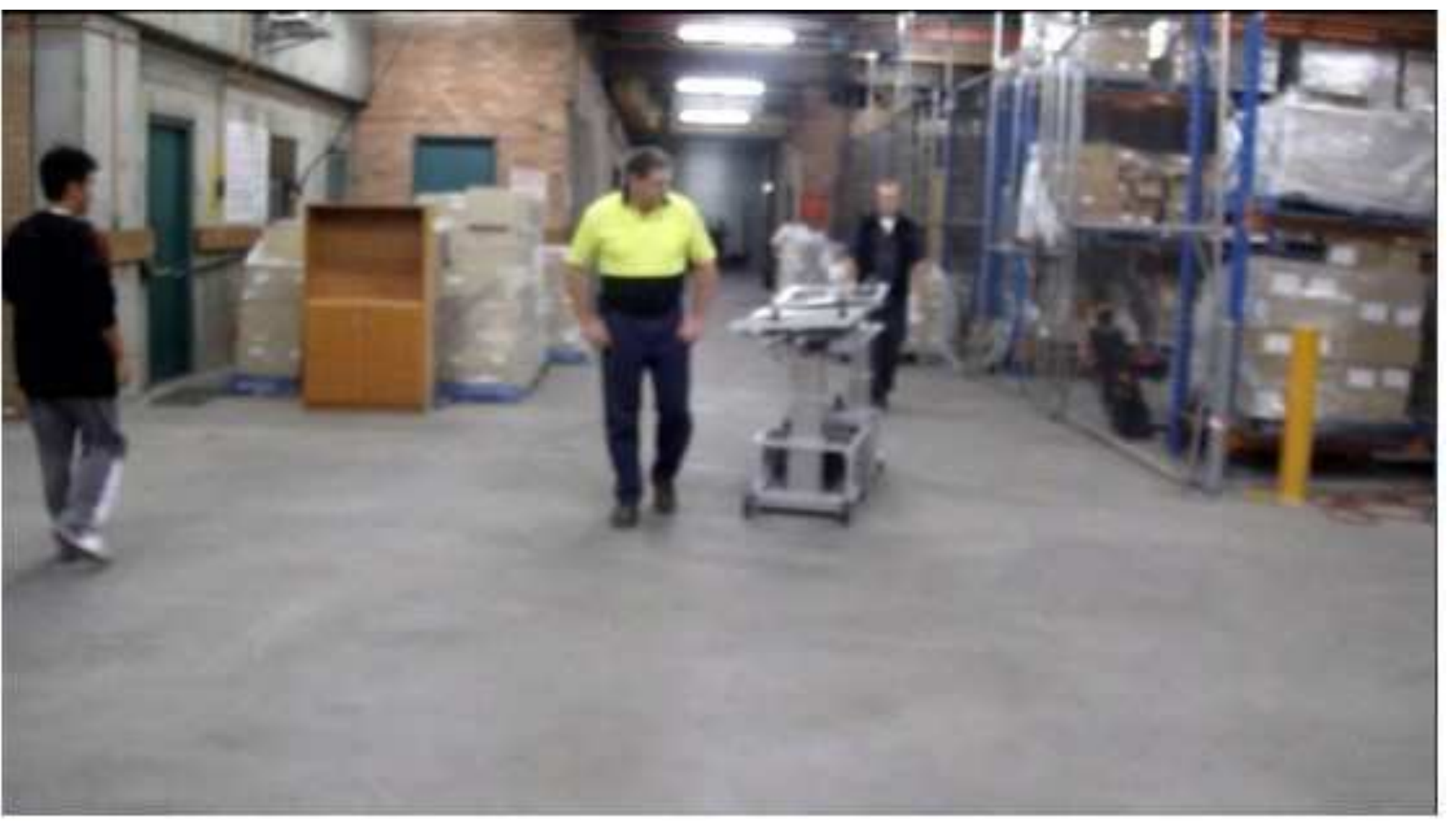}}
			\label{c7.exp31}}
			\subfigure[]{\scalebox{0.3}{\includegraphics{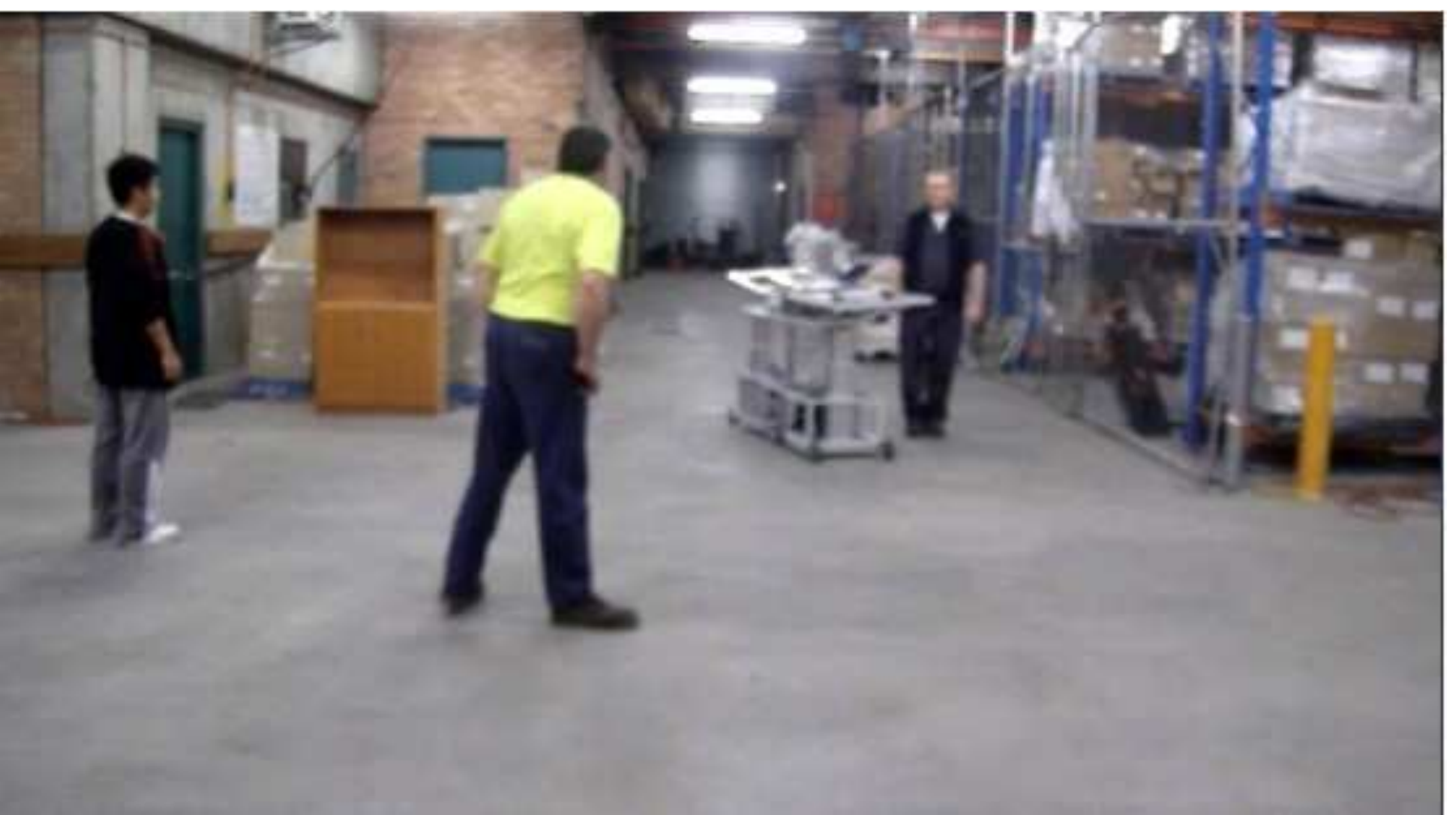}}
			\label{c7.exp32}}
			\subfigure[]{\scalebox{0.3}{\includegraphics{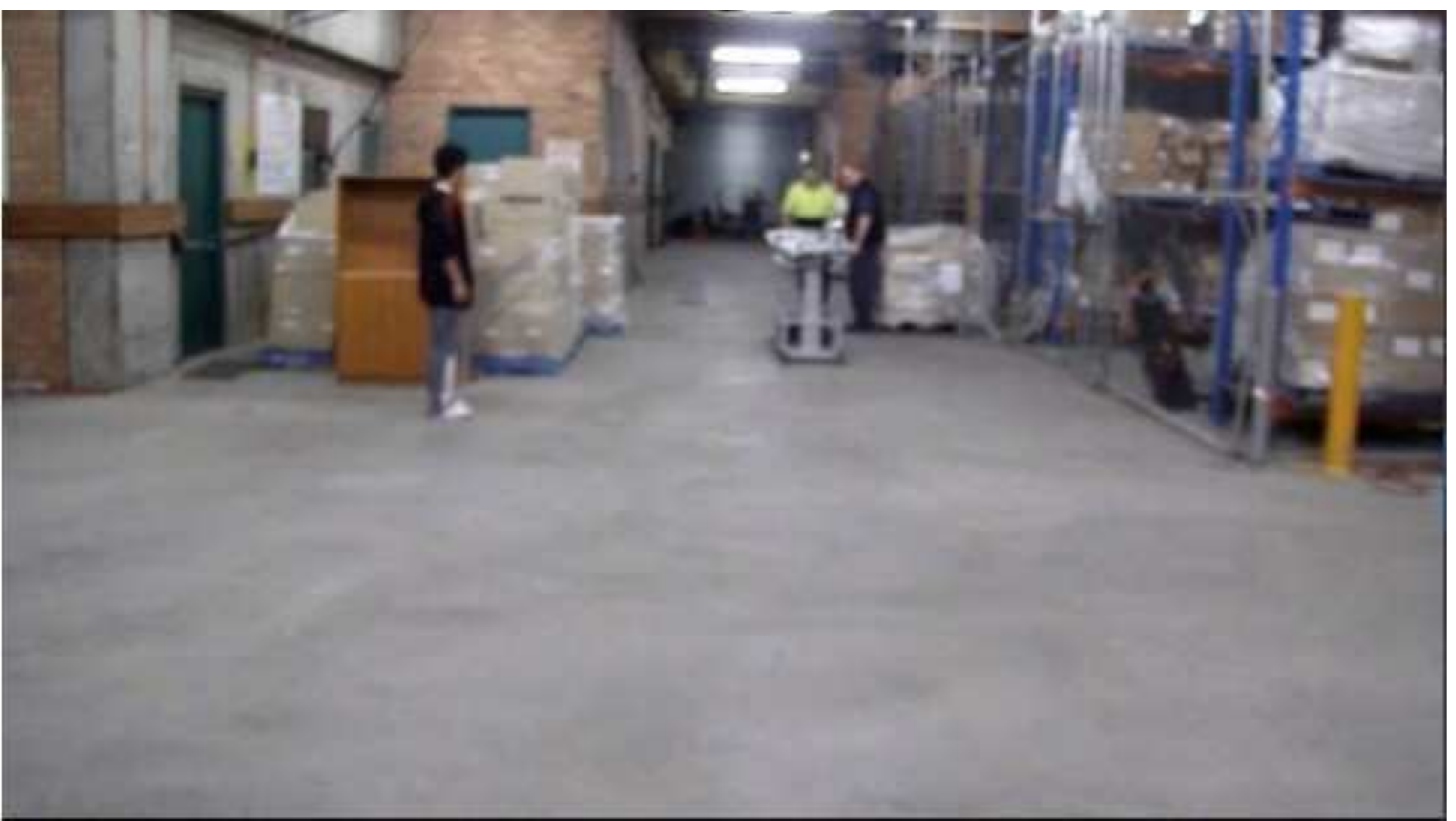}}
			\label{c7.exp33}}
			\subfigure[]{\scalebox{0.3}{\includegraphics{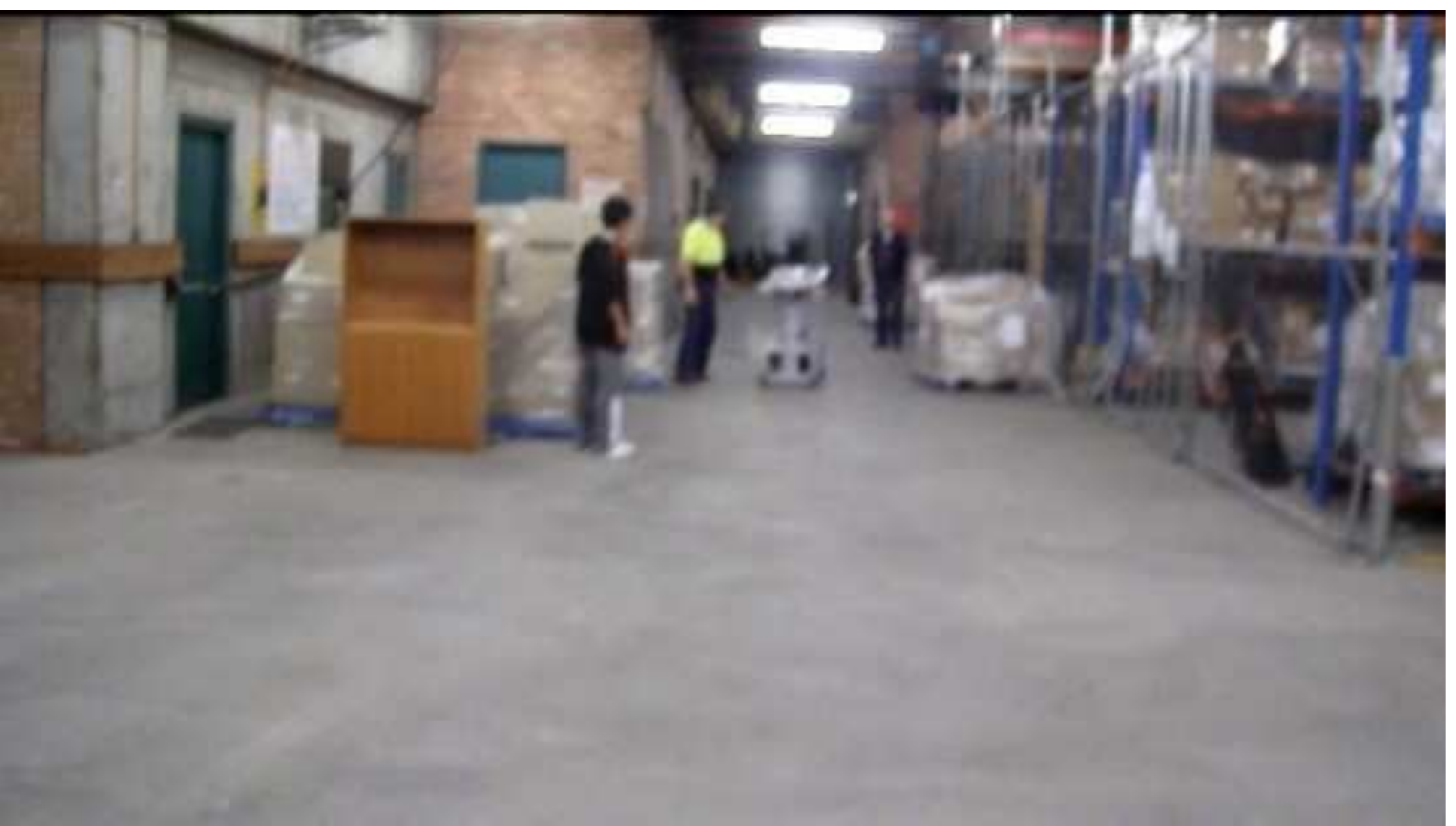}}
			\label{c7.exp34}}
			\end{minipage}
			\begin{minipage}{0.5\textwidth}
			\subfigure[]{\scalebox{0.8}{\includegraphics{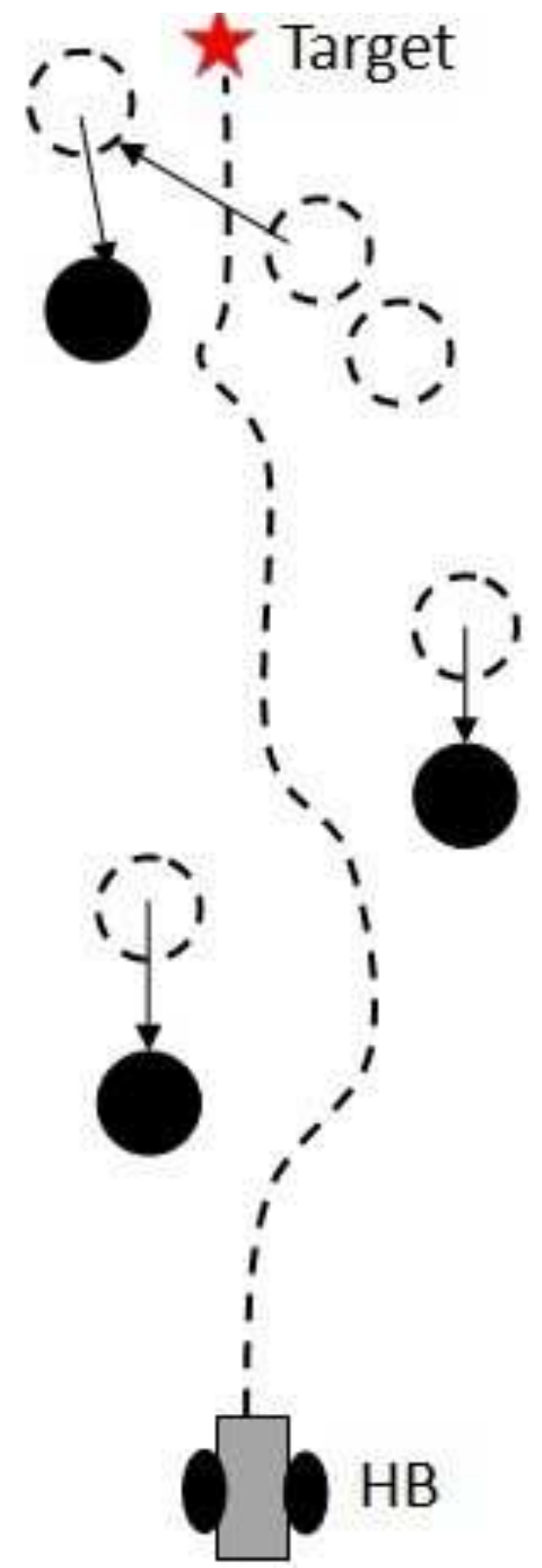}}
			\label{c7.exp35}}
			\end{minipage}
			\caption{Flexbed avoids multiple dynamic obstacles}	
			\label{c7.exp3}
			\end{figure}

			The next experiment shows a more complicated environments with multiple moving obstacles. This is also a more realistic representation of the hospital environments where the obstacles (personals, patients) are moving with various velocities. Fig.~\ref{c7.exp31}, Fig.~\ref{c7.exp32}, Fig.~\ref{c7.exp33} and Fig.~\ref{c7.exp34} show the moments when the Flexbed avoids the obstacles. This experiments shows BINA is capable of navigating Flexbed in dynamic environments with moving obstacles, such as hospital environment.
\par

\begin{figure}[h]
			\centering
			\subfigure[]{\scalebox{0.32}{\includegraphics{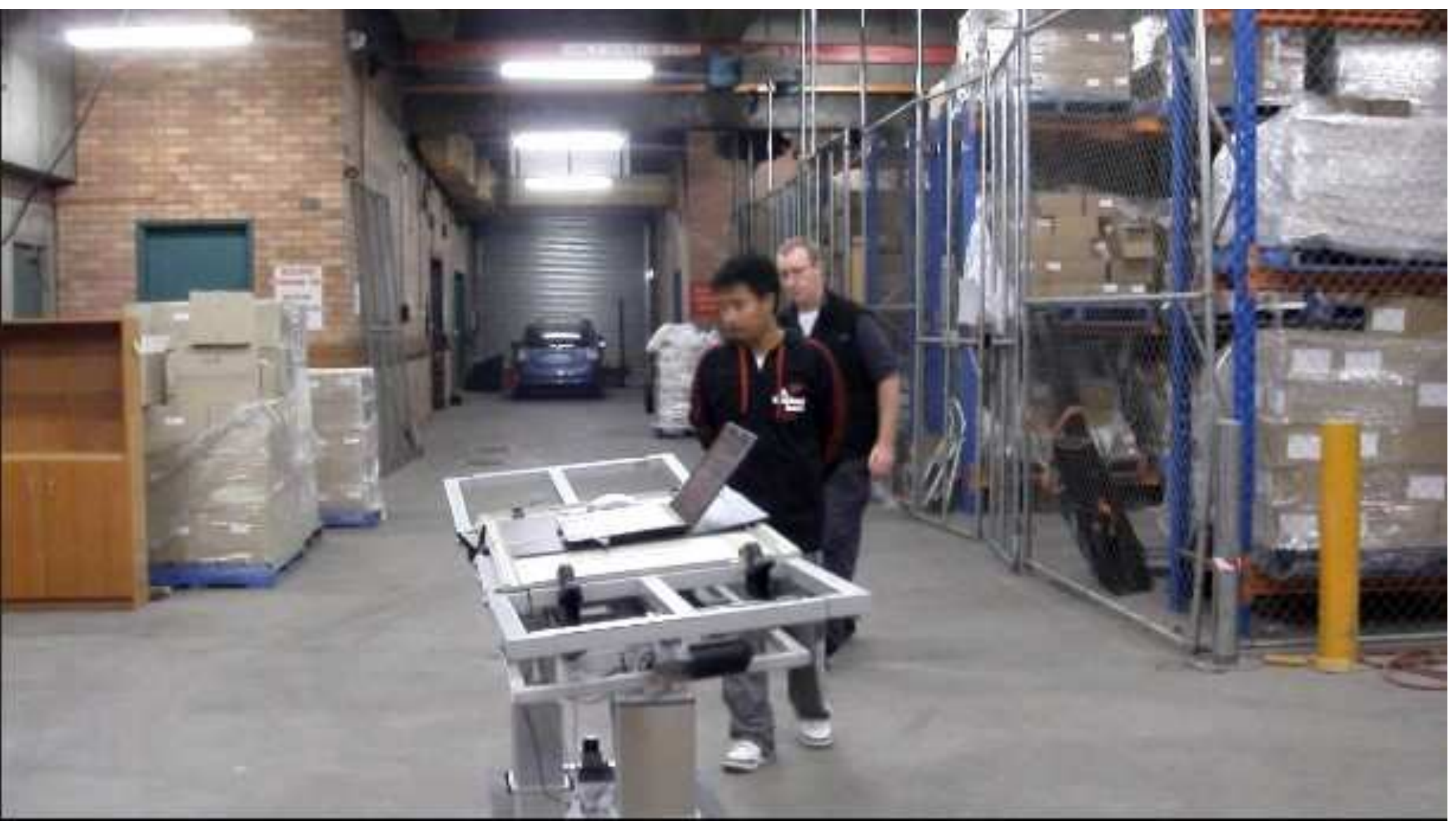}}
			\label{c7.exp41}}
			\subfigure[]{\scalebox{0.32}{\includegraphics{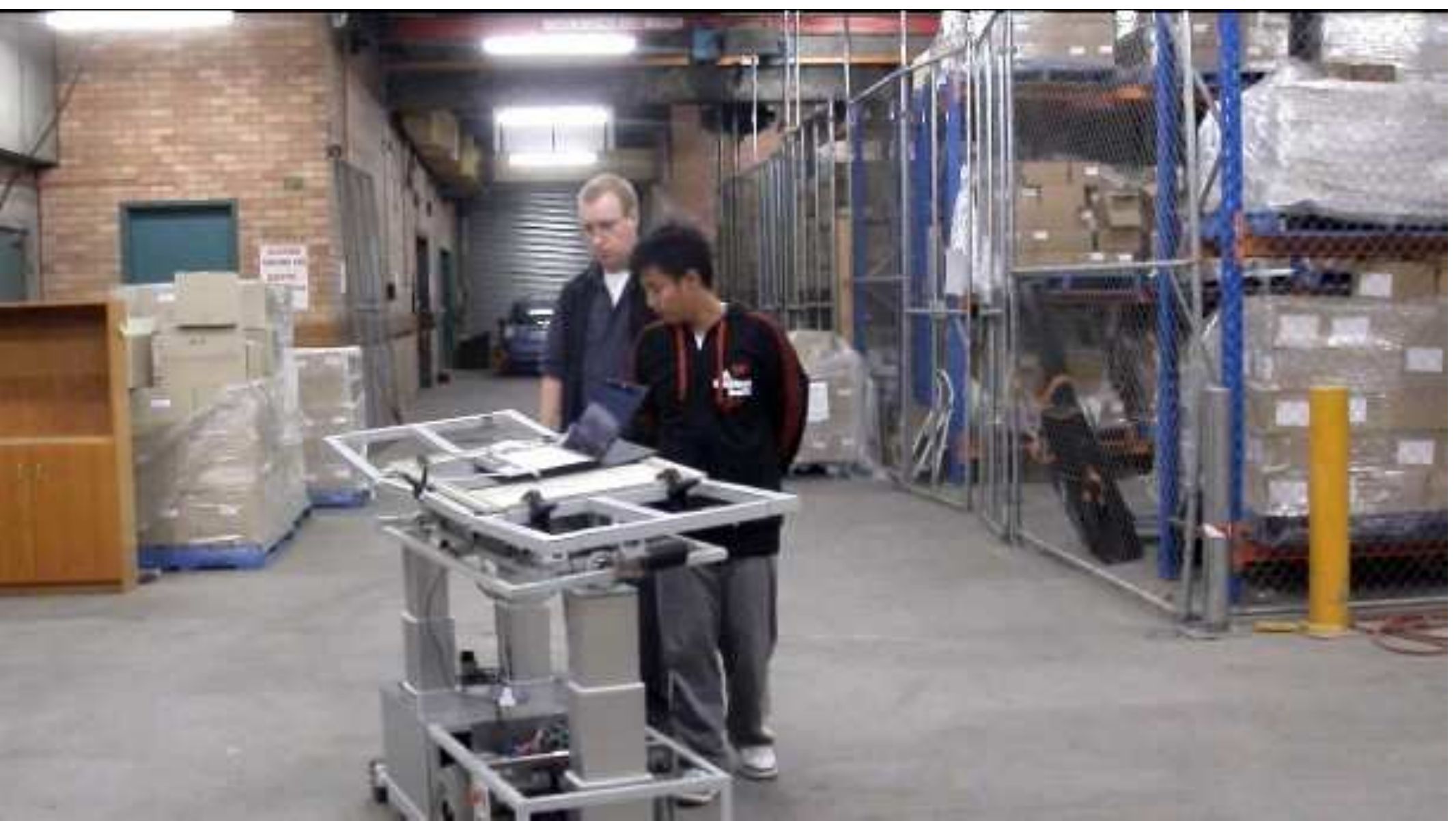}}
			\label{c7.exp42}}
			\subfigure[]{\scalebox{0.5}{\includegraphics{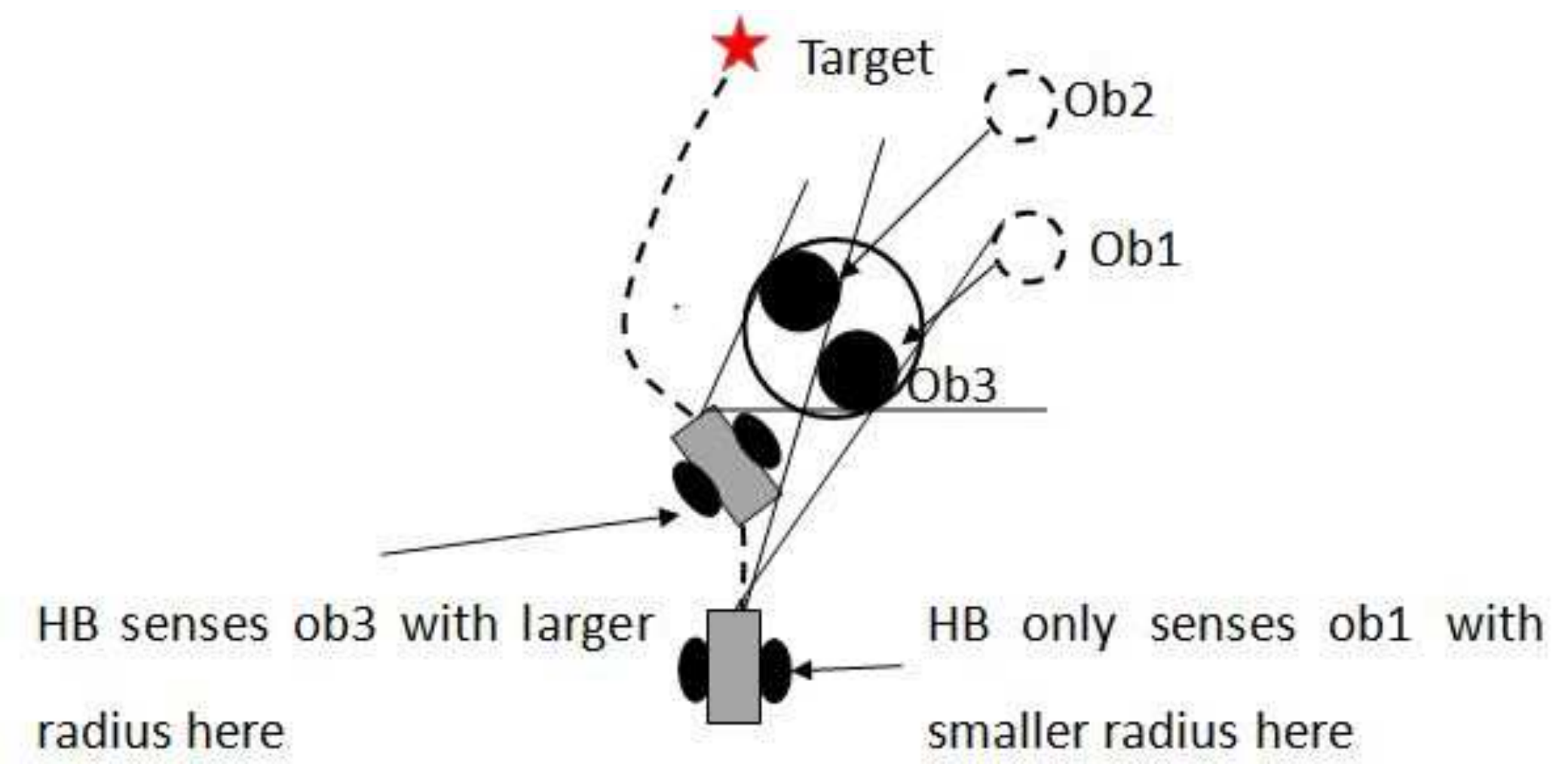}}
			\label{c7.exp43}}

			\caption{Flexbed avoids obstacle with dynamic changing radius}
			\label{c7.exp4}
			\end{figure}

			In the last experiment, we examine the scenario which the radii of the obstacles are dynamically-changing. This is a common scenario in real life where a number of obstacles may join together to form a larger group of obstacles or some obstacles may leave the group. In Fig.~\ref{c7.exp41}, Flexbed senses one obstacle (ob1) with small radius and start to avoid it. At The meantime, the second obstacle (ob2) joins ob1 to form 2 larger obstacle (ob3) in Fig.~\ref{c7.exp42}. It can be seen that BINA is able to avoid obstacles with dynamically changing radii by steering Flexbed towards the closest boundary of vision cone of the new obstacle. This experiment can be easily extended to scenario with more obstacles forming even larger group of obstacles or obstacles leaving the group resulting a group of obstacles with smaller radius.

			\par

	\section {Experiments of Flexbed with Equidistant Navigation Algorithm (ENA)}

		In this section, we present the experimental results of navigating Flexbed in real life scenarios under the guidance of ENA. Since the only information required by ENA to avoid obstacles is the minimum distance between Flexbed and the obstacle, the URG-04LX laser range finder can be replaced by other detection devices such as Microsoft Kinect.
\par
\begin{figure}[h]
			\centering
			\subfigure[]{\scalebox{0.27}{\includegraphics{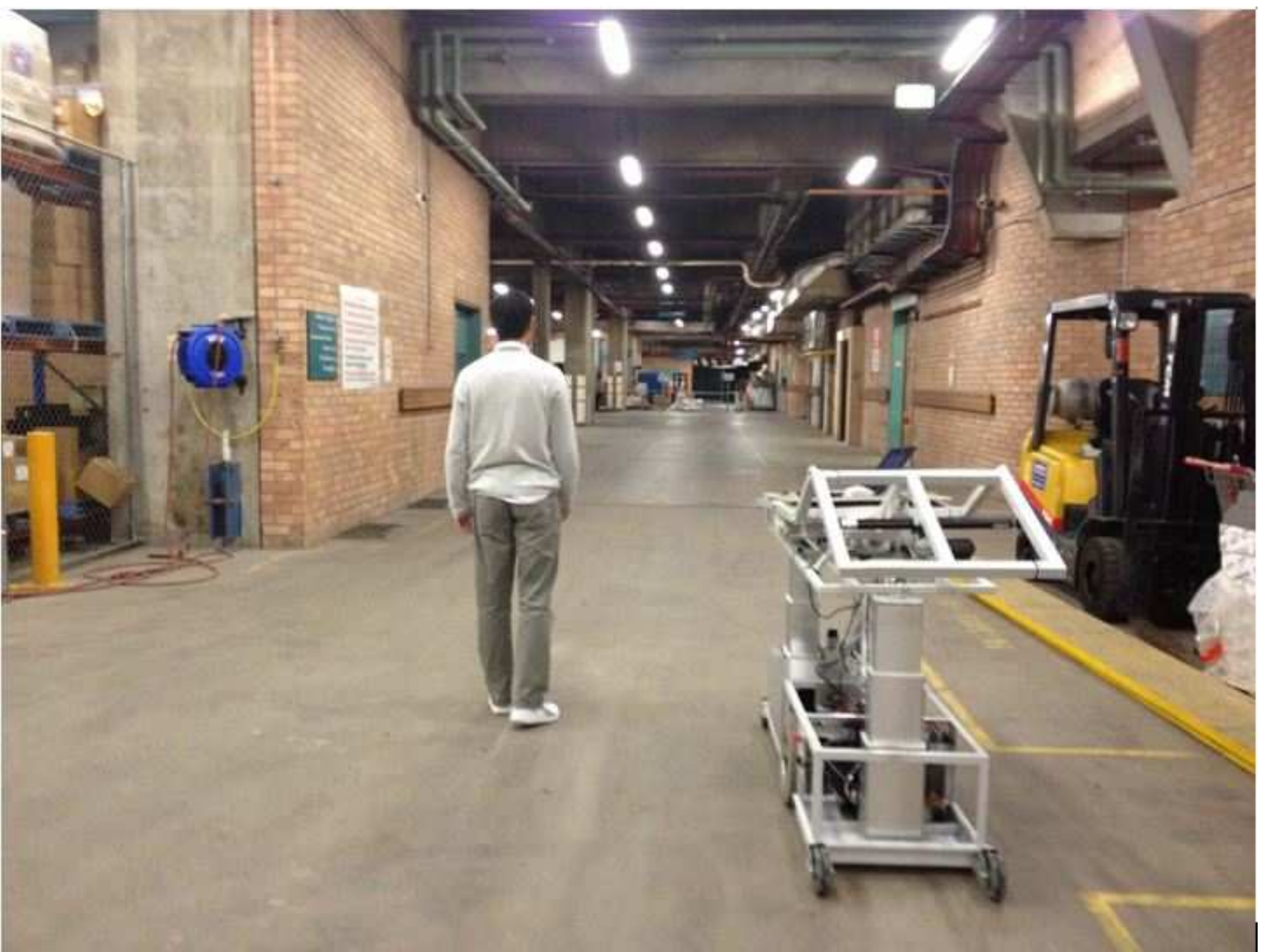}}
			\label{c7.exp51}}
			\subfigure[]{\scalebox{0.27}{\includegraphics{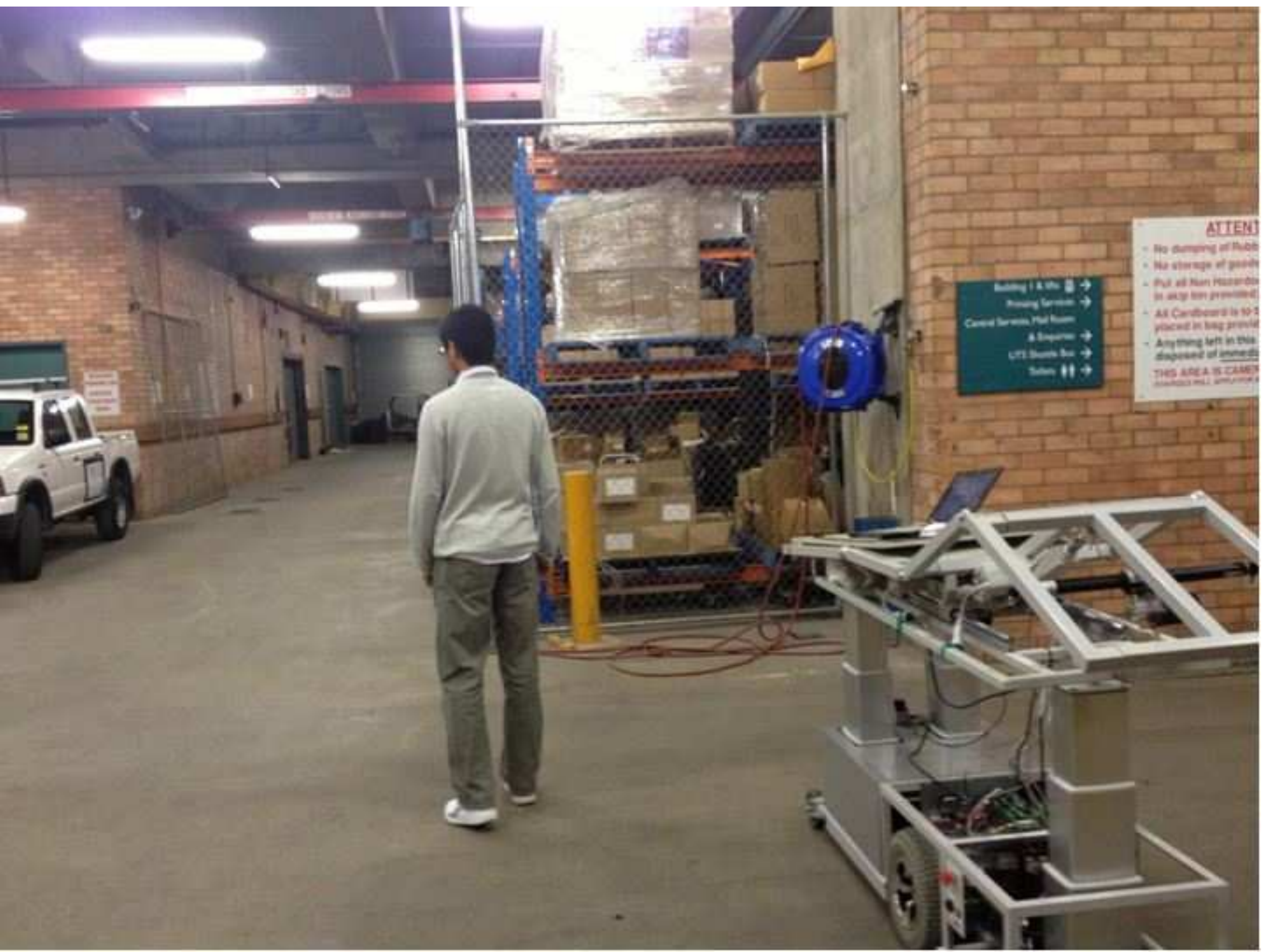}}
			\label{c7.exp52}}
			\subfigure[]{\scalebox{0.27}{\includegraphics{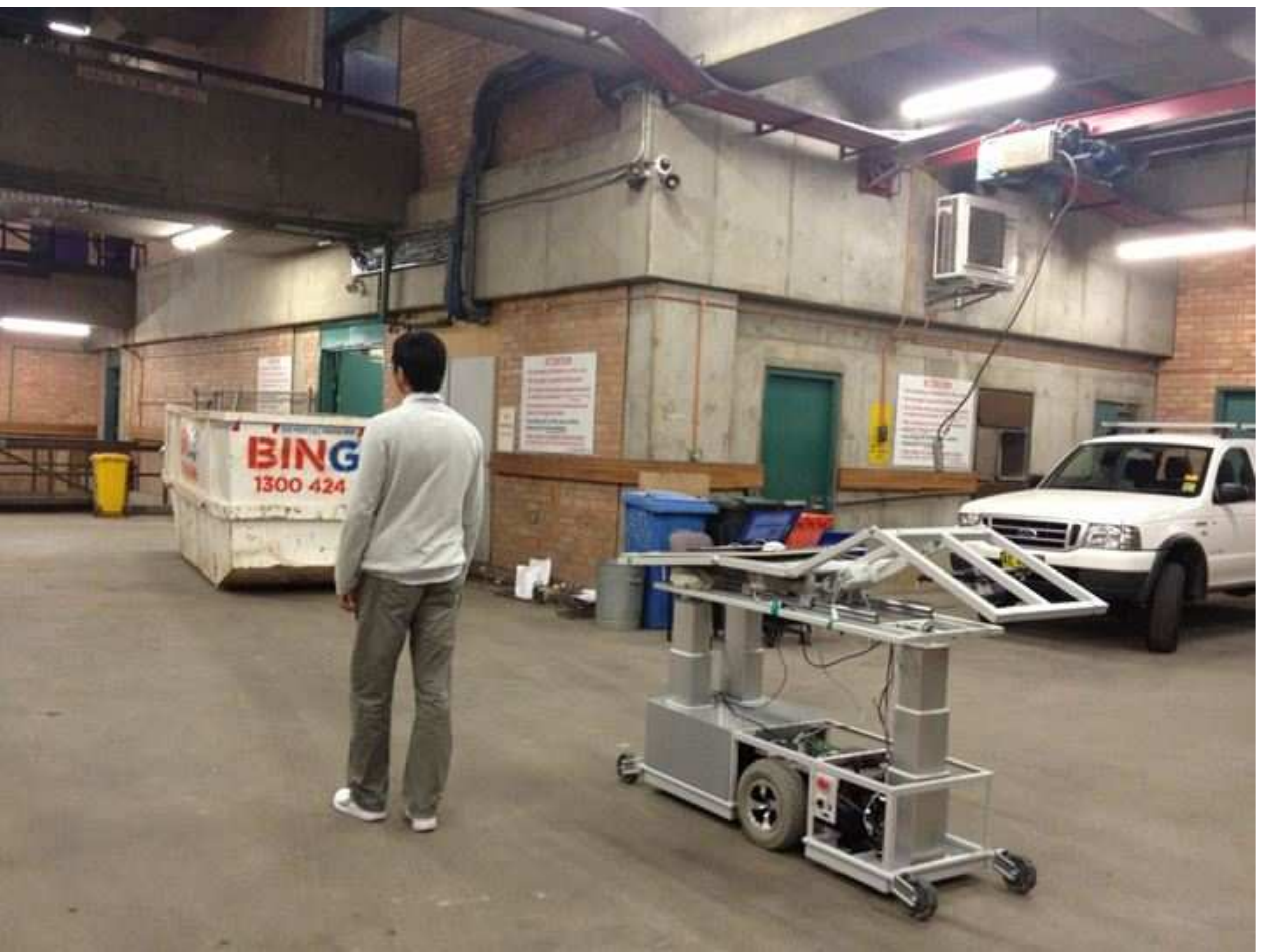}}
			\label{c7.exp53}}
			\subfigure[]{\scalebox{0.40}{\includegraphics{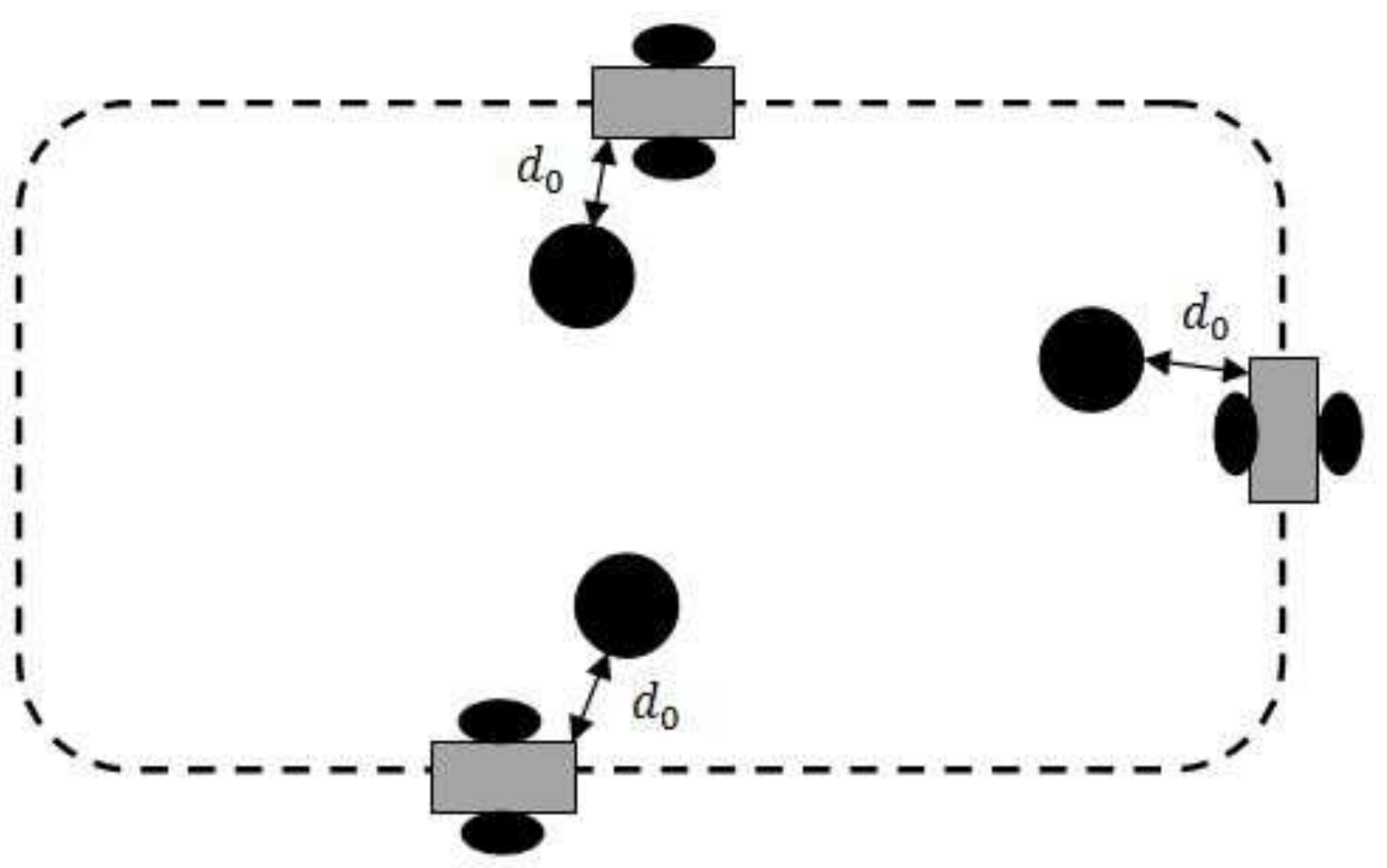}}
			\label{c7.exp54}}
			\caption{Flexbed keeps a constant distance to the obstacle}
			\label{c7.exp5}
			\end{figure}

		In the first experiment, we show that Flexbed is able to keep a constant distance between itself and the obstacle. We deliberately position the experimenter within the switching distance $C$ so that he is considered as an obstacle at all time. The snapshots of the experiment are shown in Fig.~\ref{c7.exp51}, Fig.~\ref{c7.exp52} and Fig.~\ref{c7.exp53}. The complete paths taken by the obstacle and Flexbed is depicted in Fig.~\ref{c7.exp54}.
		
			\par

\begin{figure}[h]
			\centering
			\subfigure[]{\scalebox{0.33}{\includegraphics{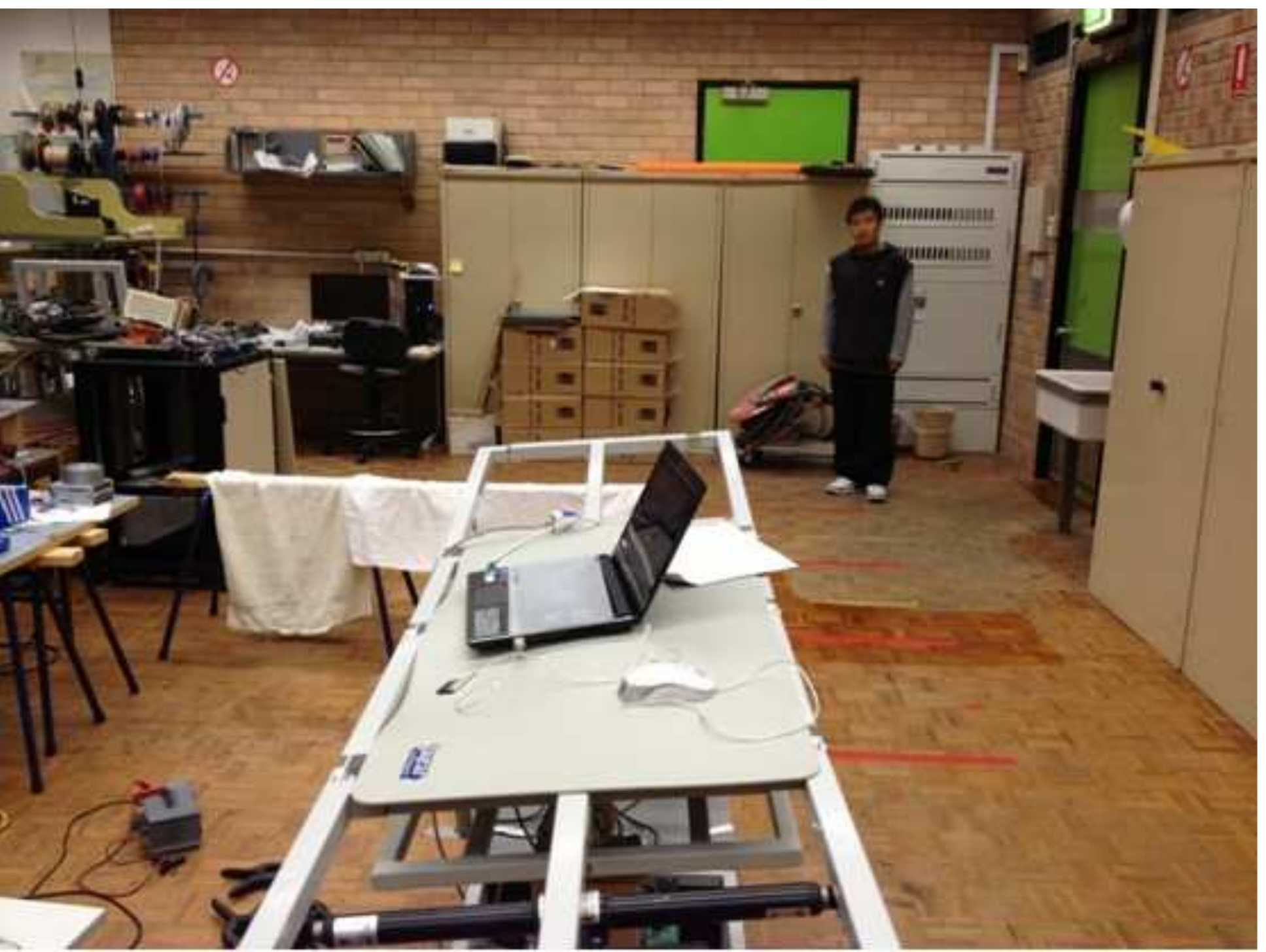}}
			\label{c7.exp61}}
			\subfigure[]{\scalebox{0.33}{\includegraphics{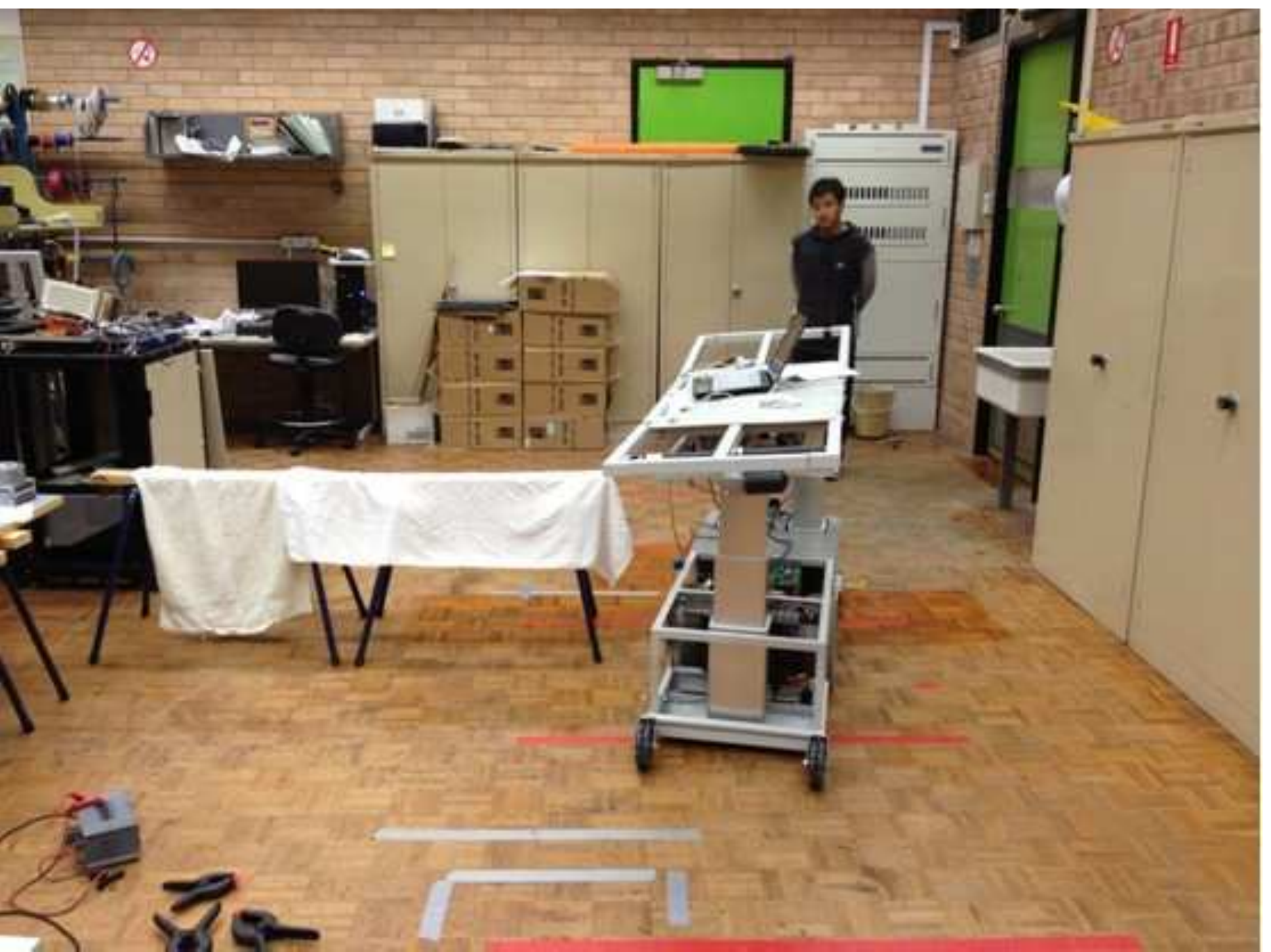}}
			\label{c7.exp62}}
			\subfigure[]{\scalebox{0.33}{\includegraphics{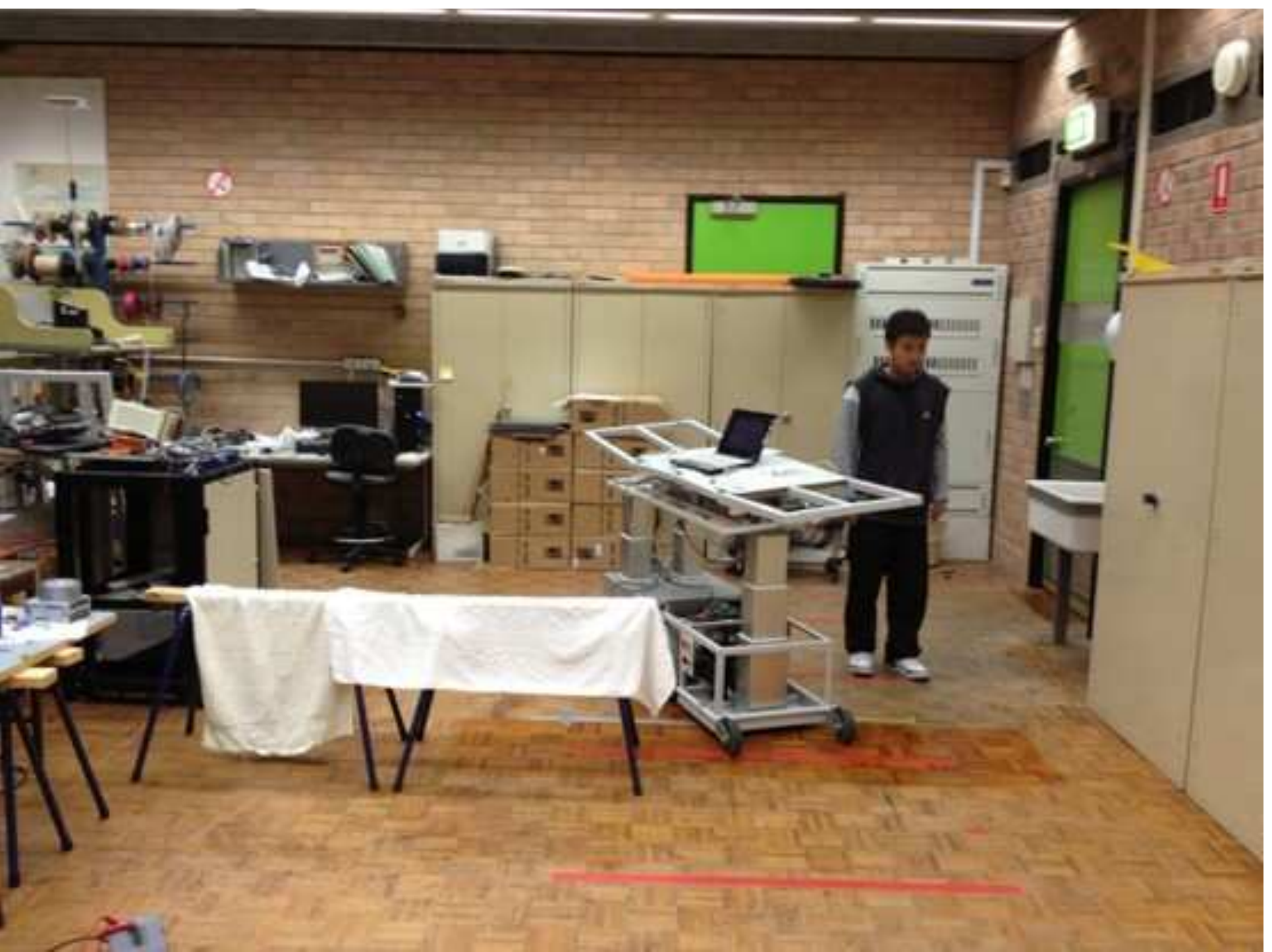}}
			\label{c7.exp63}}
			\subfigure[]{\scalebox{0.50}{\includegraphics{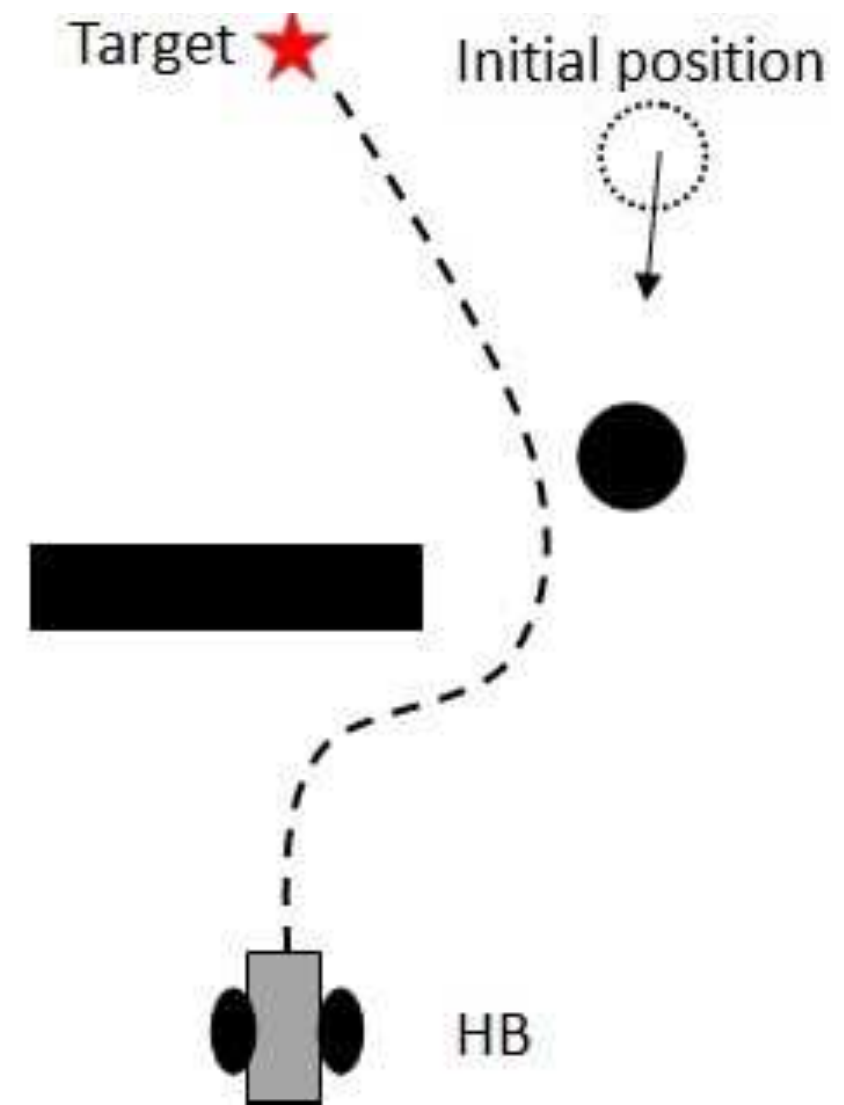}}
			\label{c7.exp64}}
			\caption{Flexbed avoids the stationary obstacle (long bench) and the dynamic obstacle (the experimenter)}
			\label{c7.exp6}
			\end{figure}

		Fig.~\ref{c7.exp6} shows the experiment of navigating Flexbed in an environment with stationary and dynamic obstacles. More importantly, this experiment shows that the ability of ENA to deal with obstacles with various shapes, which is a more realistic representation of the real life scenarios. Fig.~\ref{c7.exp62} and Fig.~\ref{c7.exp63} show the moment when Flexbed bypasses the rectangular obstacle (long bench) and the moving obstacle (the experimenter). The complete path is depicted in Fig.~\ref{c7.exp64}.

			\par

\begin{figure}[!t]
			\begin{minipage}{.5\textwidth}	
			\centering	
			\subfigure[]{\scalebox{0.23}{\includegraphics{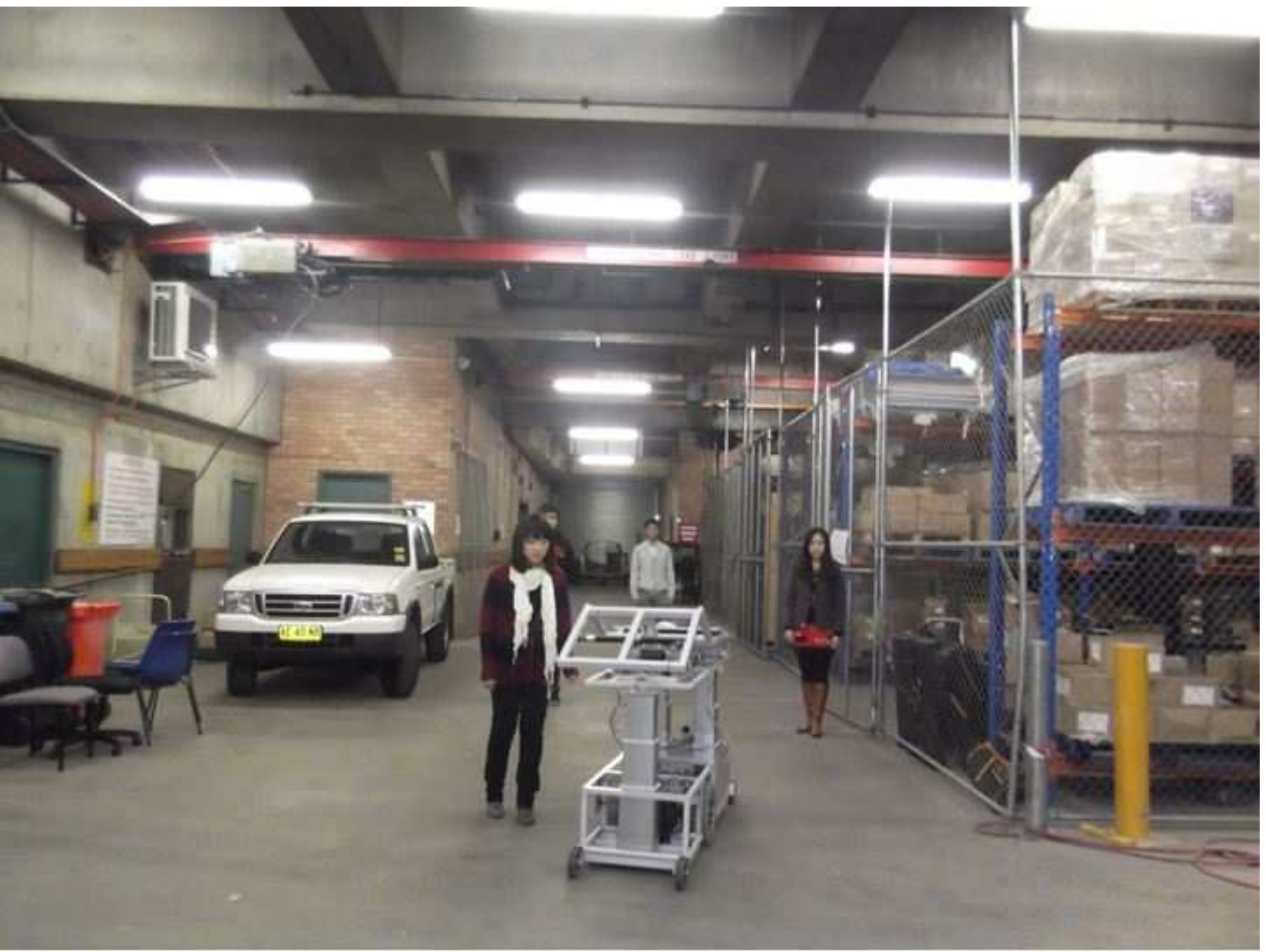}}
			\label{c7.exp71}}
			\subfigure[]{\scalebox{0.23}{\includegraphics{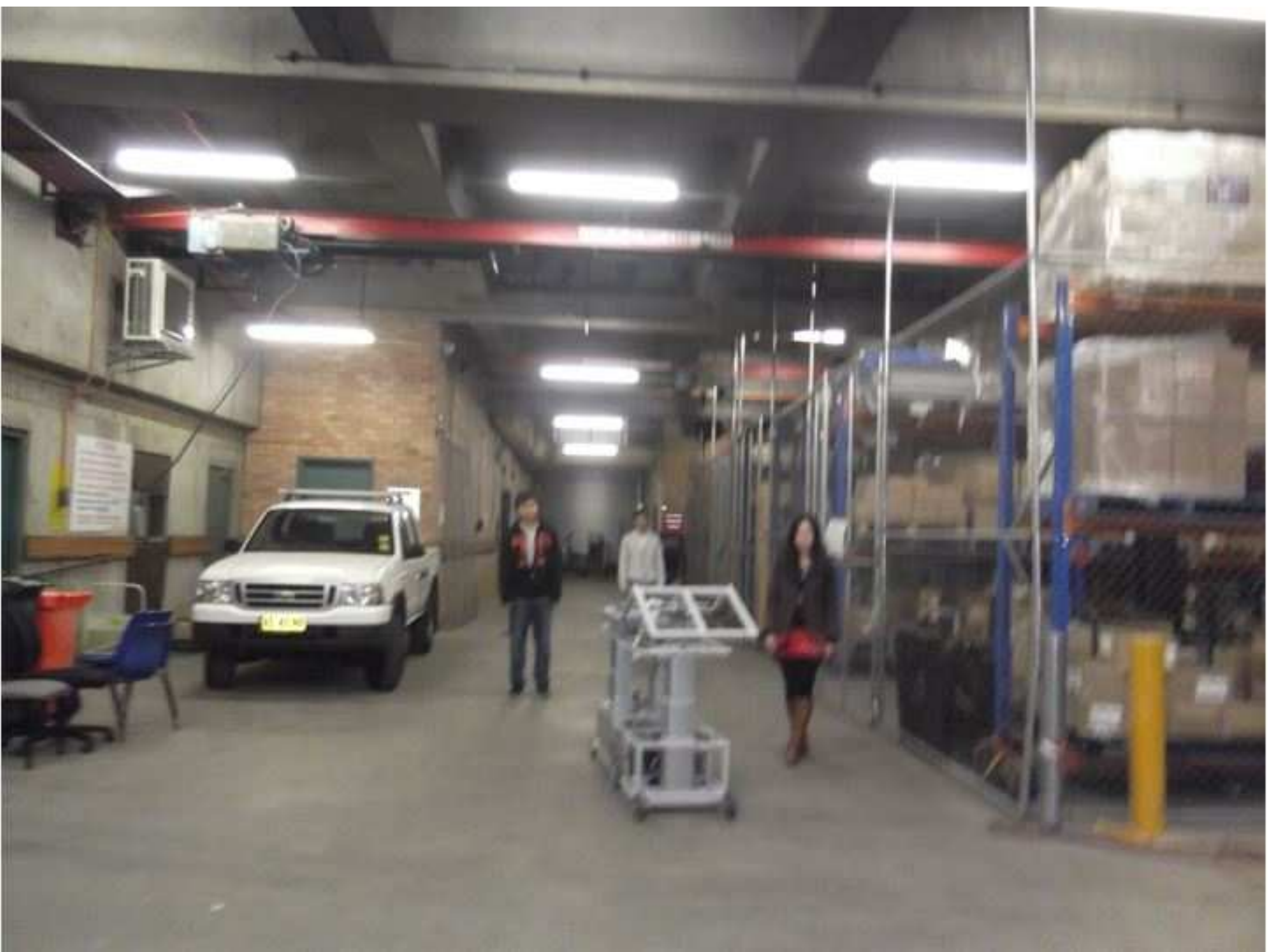}}
			\label{c7.exp72}}
			\subfigure[]{\scalebox{0.23}{\includegraphics{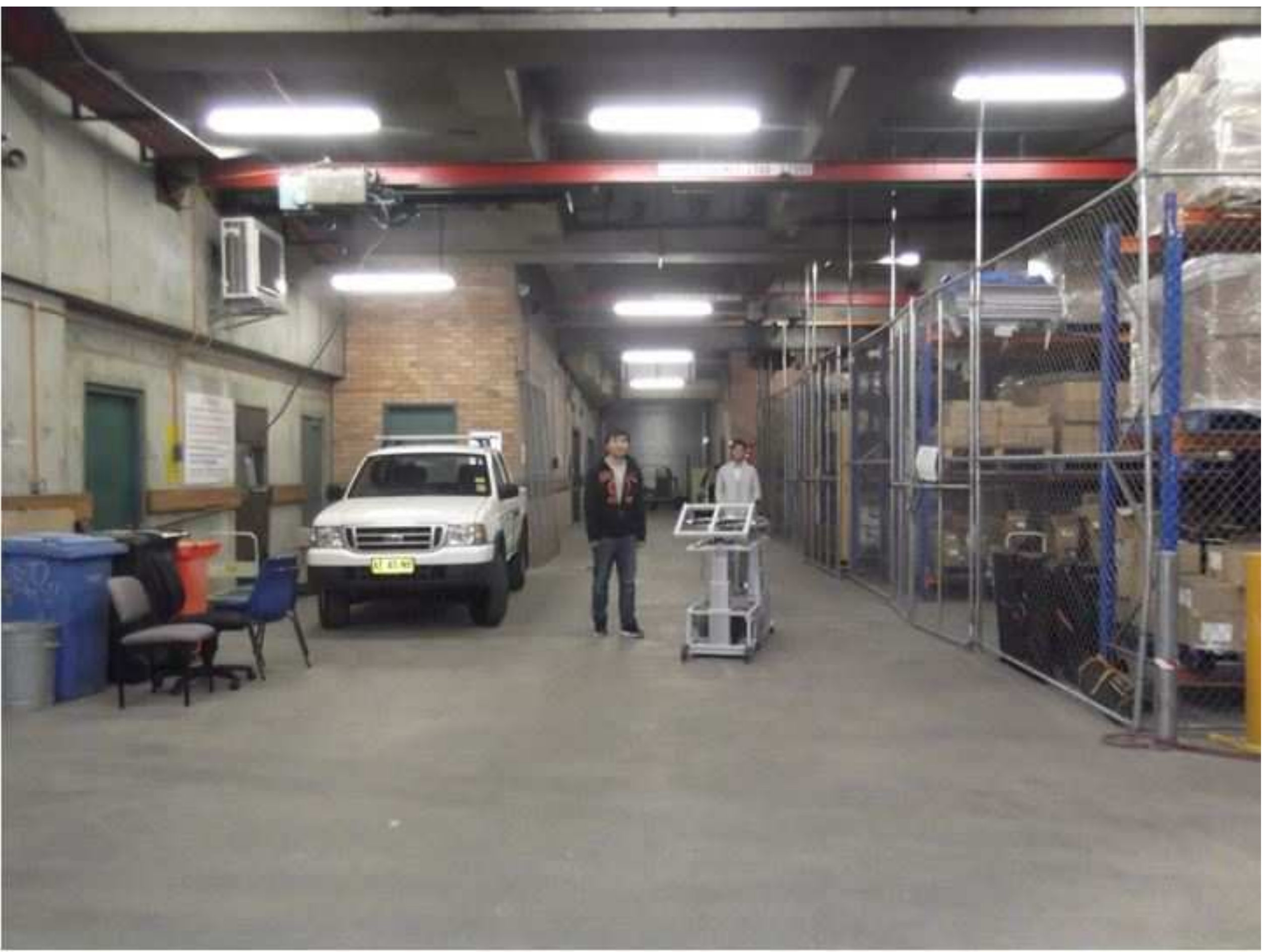}}
			\label{c7.exp73}}
			\subfigure[]{\scalebox{0.23}{\includegraphics{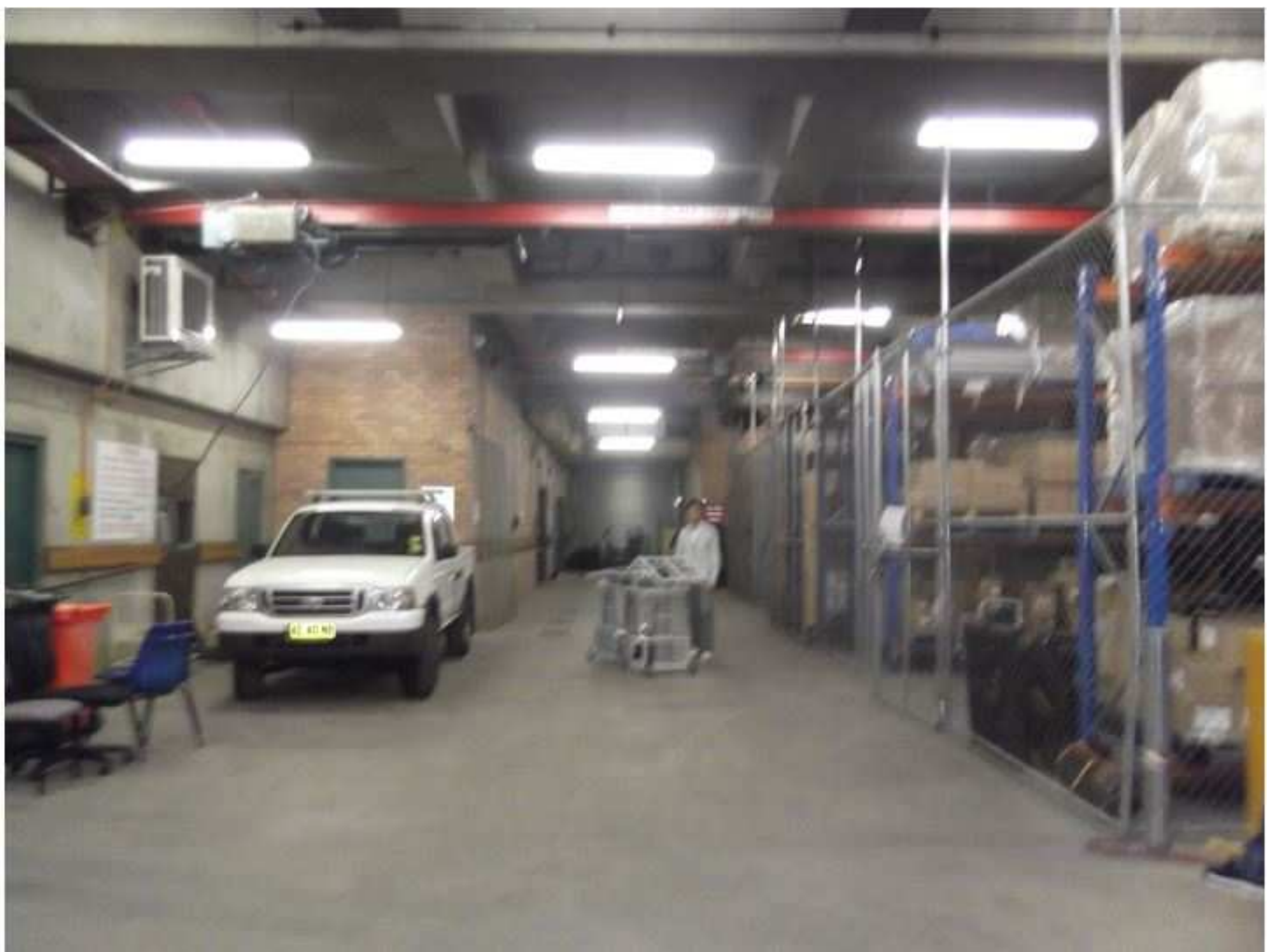}}
			\label{c7.exp74}}
			\end{minipage}
			\begin{minipage}{0.5\textwidth}
			\subfigure[]{\scalebox{0.6}{\includegraphics{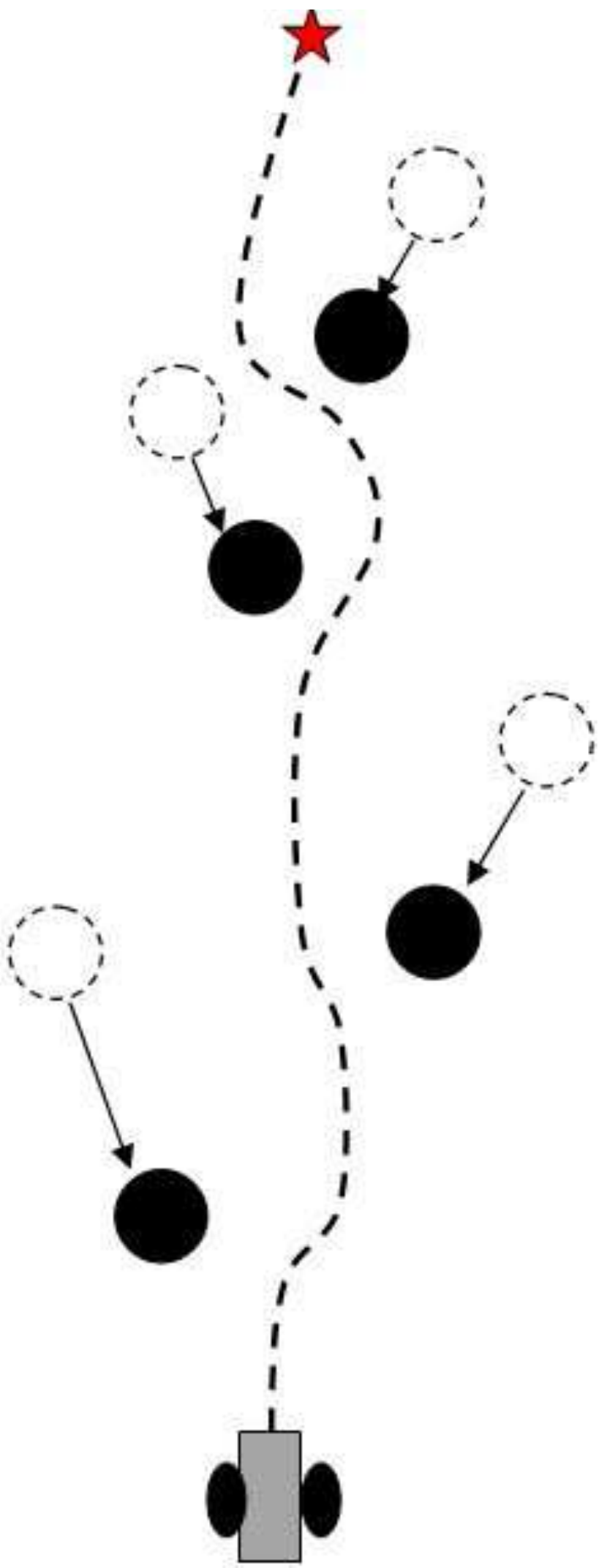}}
			\label{c7.exp75}}
			\end{minipage}
			\caption{Flexbed navigating in dynamic environment with moving obstacles}	
			\label{c7.exp7}
			\end{figure}

		The next experiment shows the ability of the Flexbed to navigate in dynamic environments with multiple moving obstacles. In this experiment, the obstacles are moving with various speed at random directions. Fig.\ref{c7.exp71}, Fig.\ref{c7.exp72} Fig.\ref{c7.exp73} and Fig.~\ref{c7.exp74} show the crucial moments when the Flexbed bypasses each of the obstacles, the Flexbed is able to track the $d_0$-equidistant curve during its avoidance maneuver. The overall path of the Flexbed is depicted in Fig.\ref{c7.exp75}.
\par

			\begin{figure}[!h]
			\begin{minipage}{.5\textwidth}
			\centering
			\subfigure[]{\scalebox{0.24}{\includegraphics{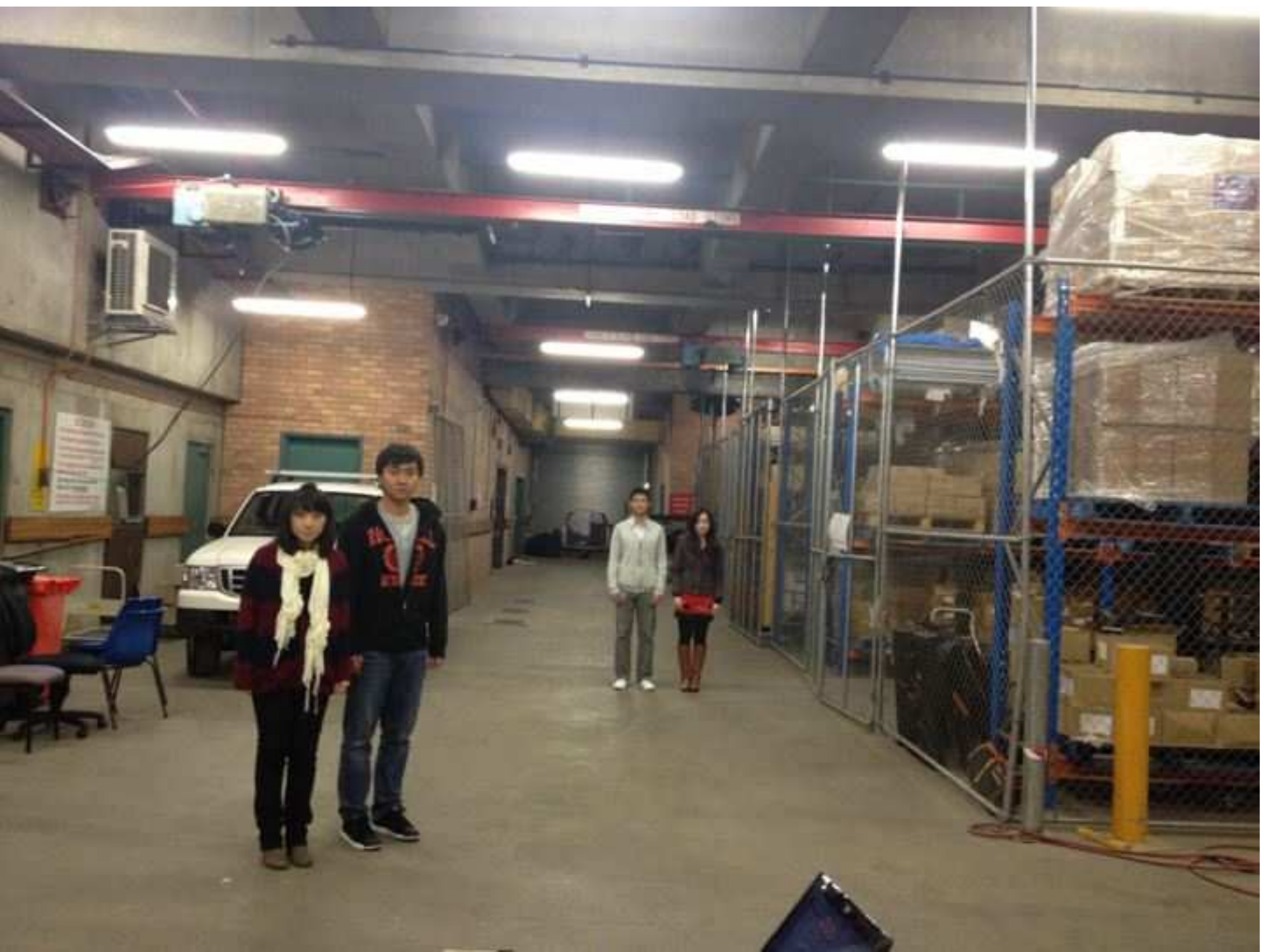}}
			\label{c7.exp81}}
			\subfigure[]{\scalebox{0.24}{\includegraphics{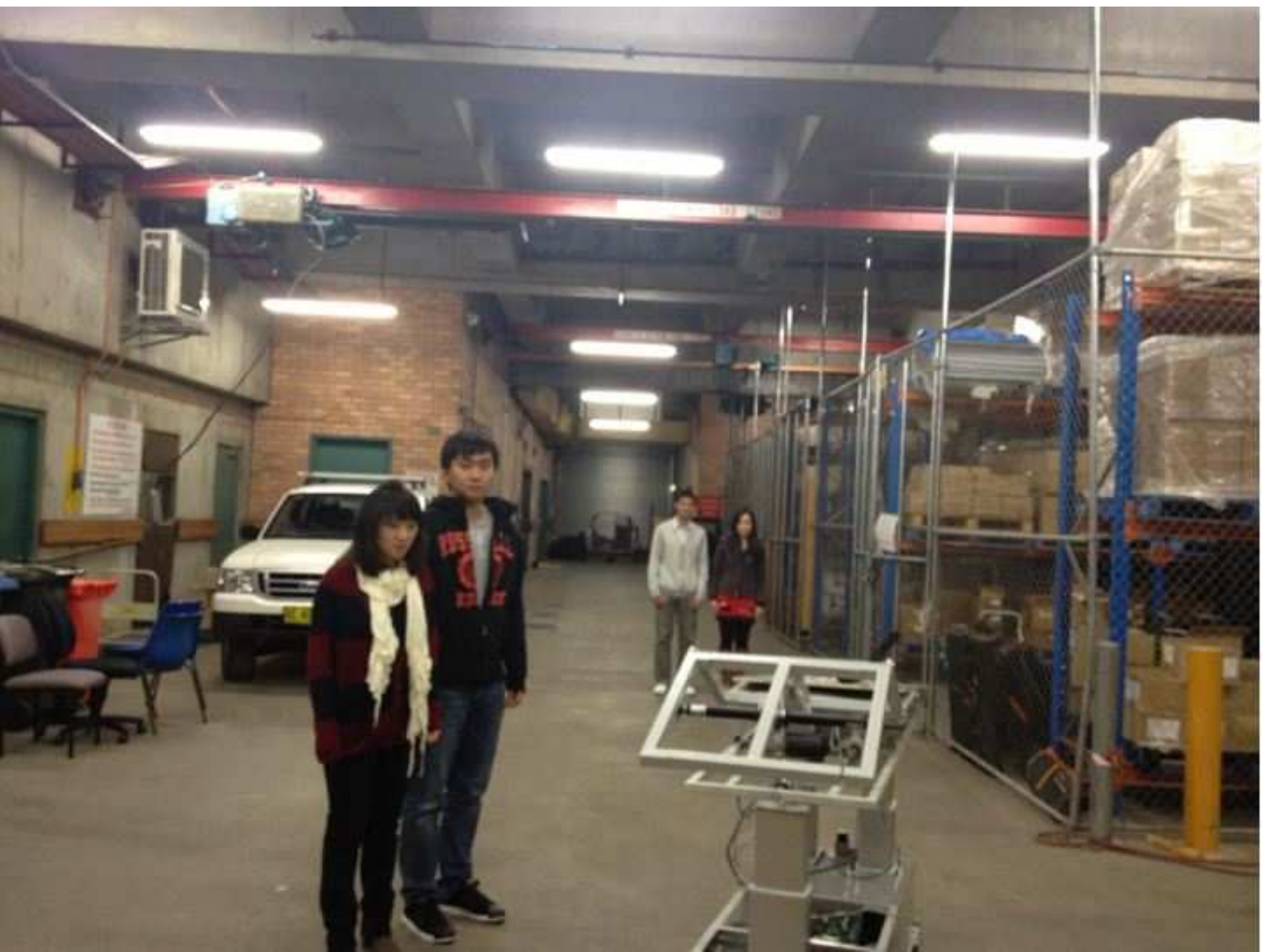}}
			\label{c7.exp82}}
			\subfigure[]{\scalebox{0.24}{\includegraphics{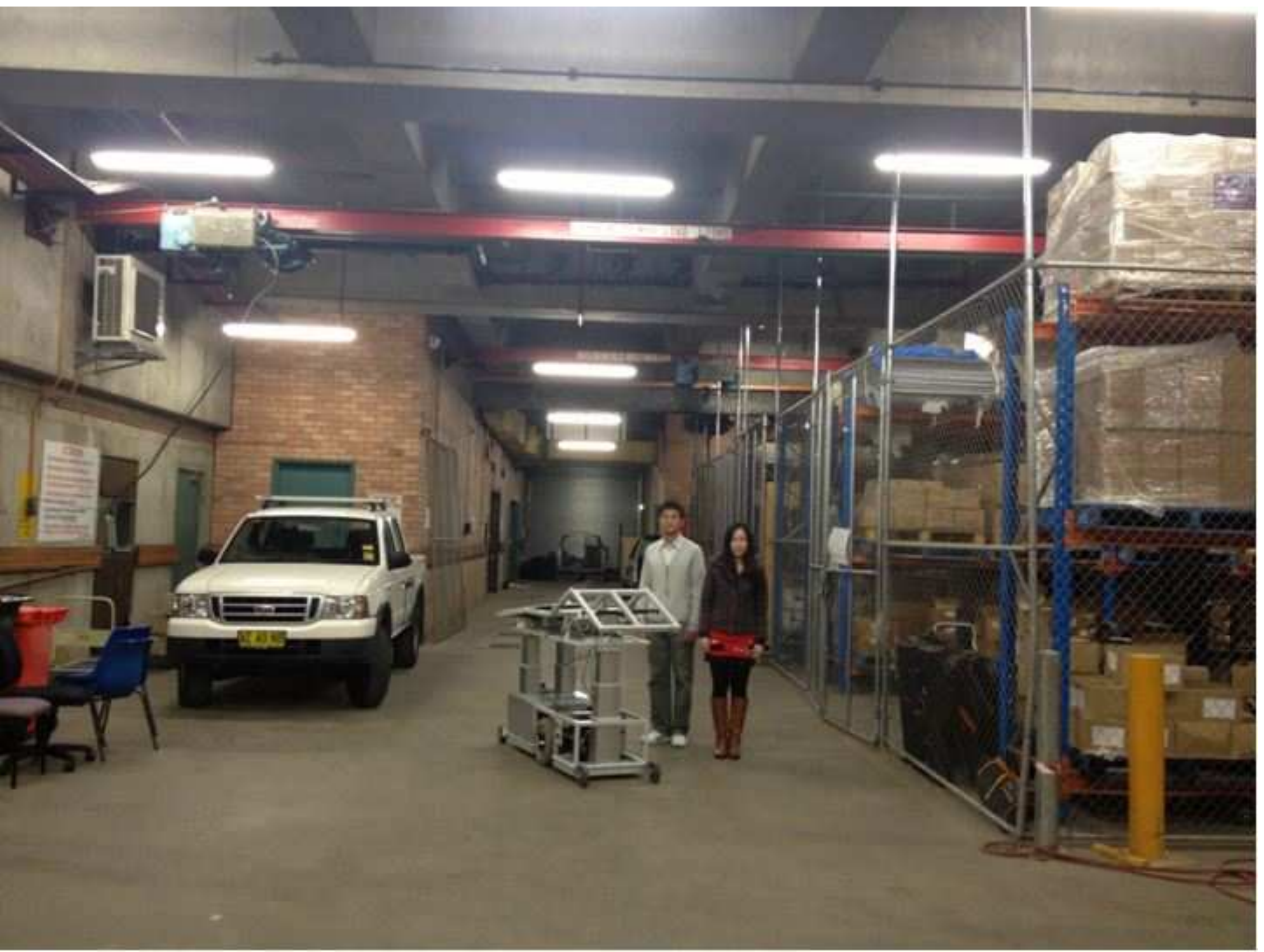}}
			\label{c7.exp83}}
			\end{minipage}
			\begin{minipage}{0.5\textwidth}
			\subfigure[]{\scalebox{0.6}{\includegraphics{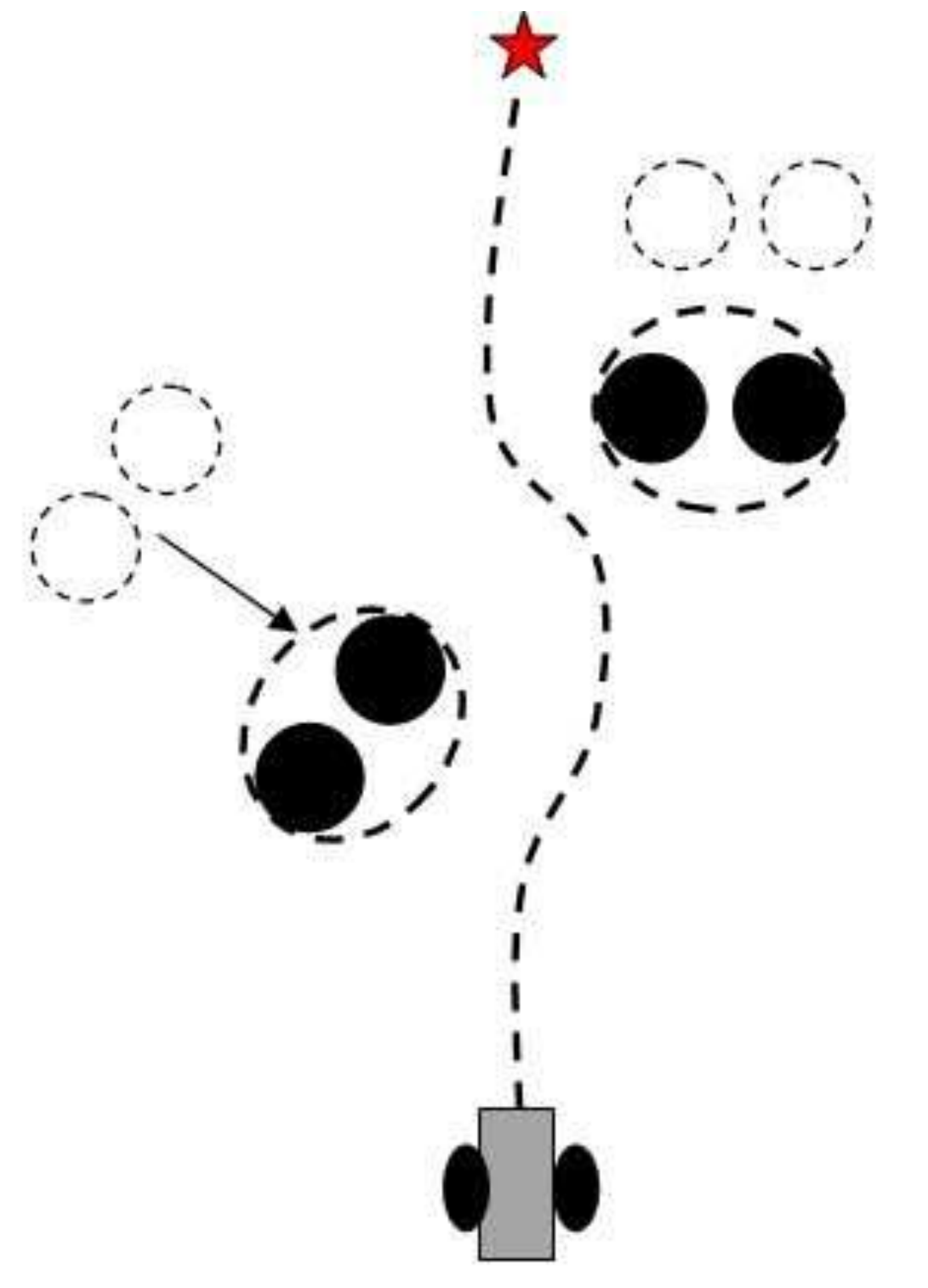}}
			\label{c7.exp84}}
			\end{minipage}
			\caption{Flexbed avoids group of obstacles}
			\label{c7.exp8}
			\end{figure}

			 In Fig.\ref{c7.exp8}, the extension of the experiment shown in Fig.\ref{c7.exp8} is presented. In this scenario,  The Flexbed faces moving obstacles with different shapes, this is a common scenario in many hospital and other work places: two or more people are moving side by side which makes the shapes of the group complicated. In the experiment, the group of two obstacles are interpolated as one larger obstacle depicted in Fig.\ref{c7.exp84}, the Flexbed is still able to track the $d_0$-equidistant curve and avoid the obstacles.
\par

\begin{figure}[h]
			\centering
			\subfigure[]{\scalebox{0.24}{\includegraphics{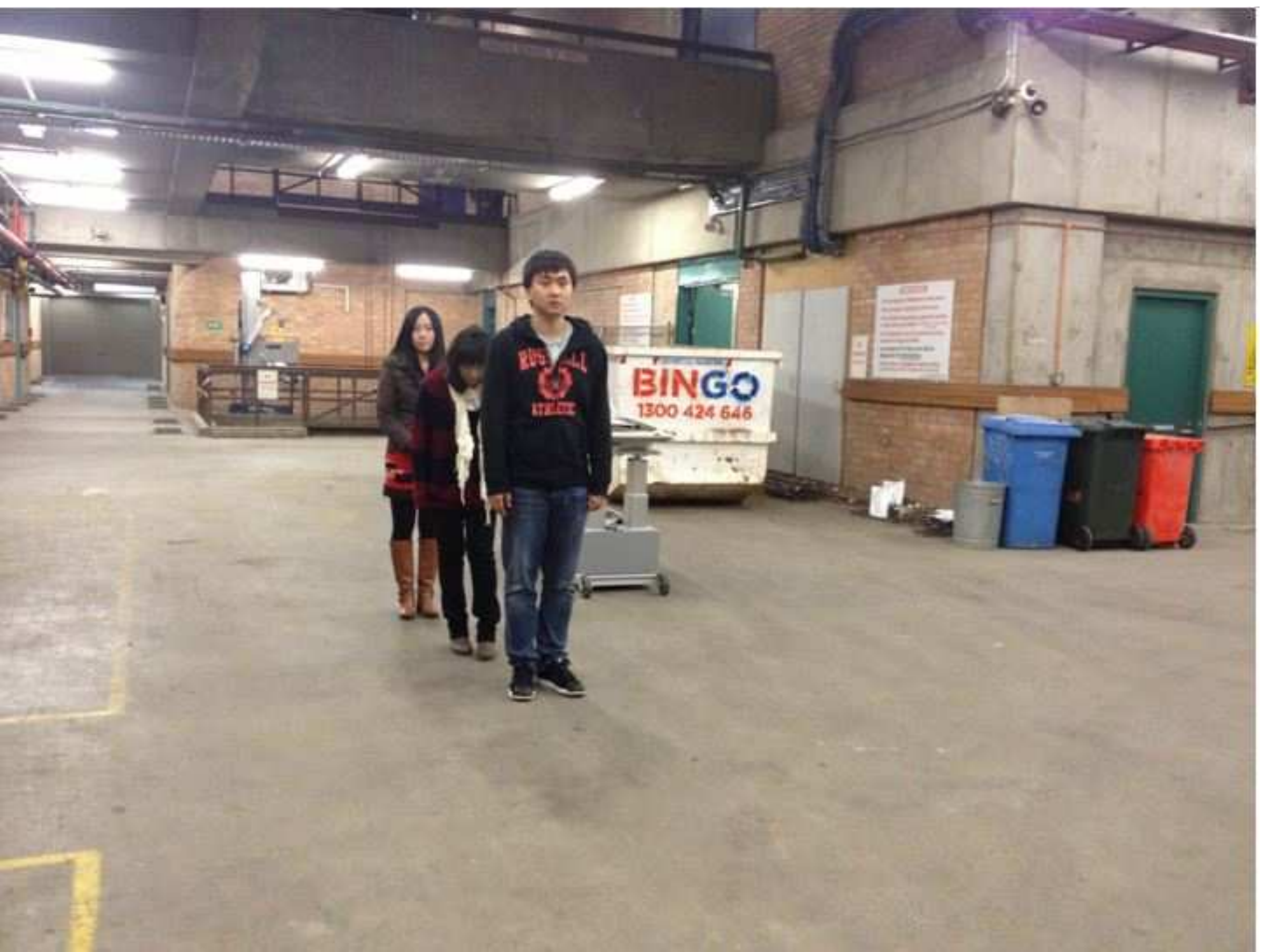}}
			\label{c7.exp91}}
			\subfigure[]{\scalebox{0.24}{\includegraphics{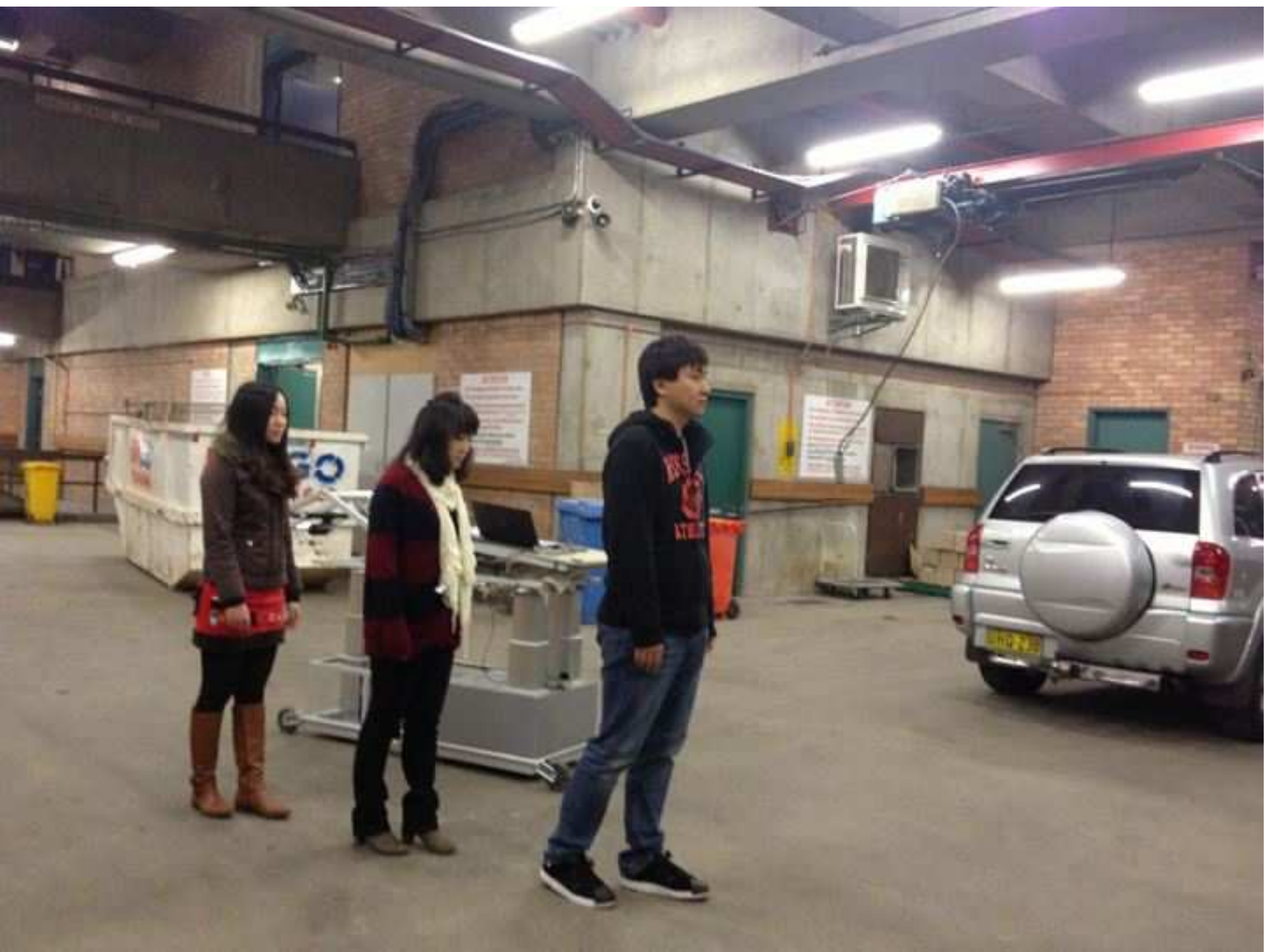}}
			\label{c7.exp92}}
			\subfigure[]{\scalebox{0.24}{\includegraphics{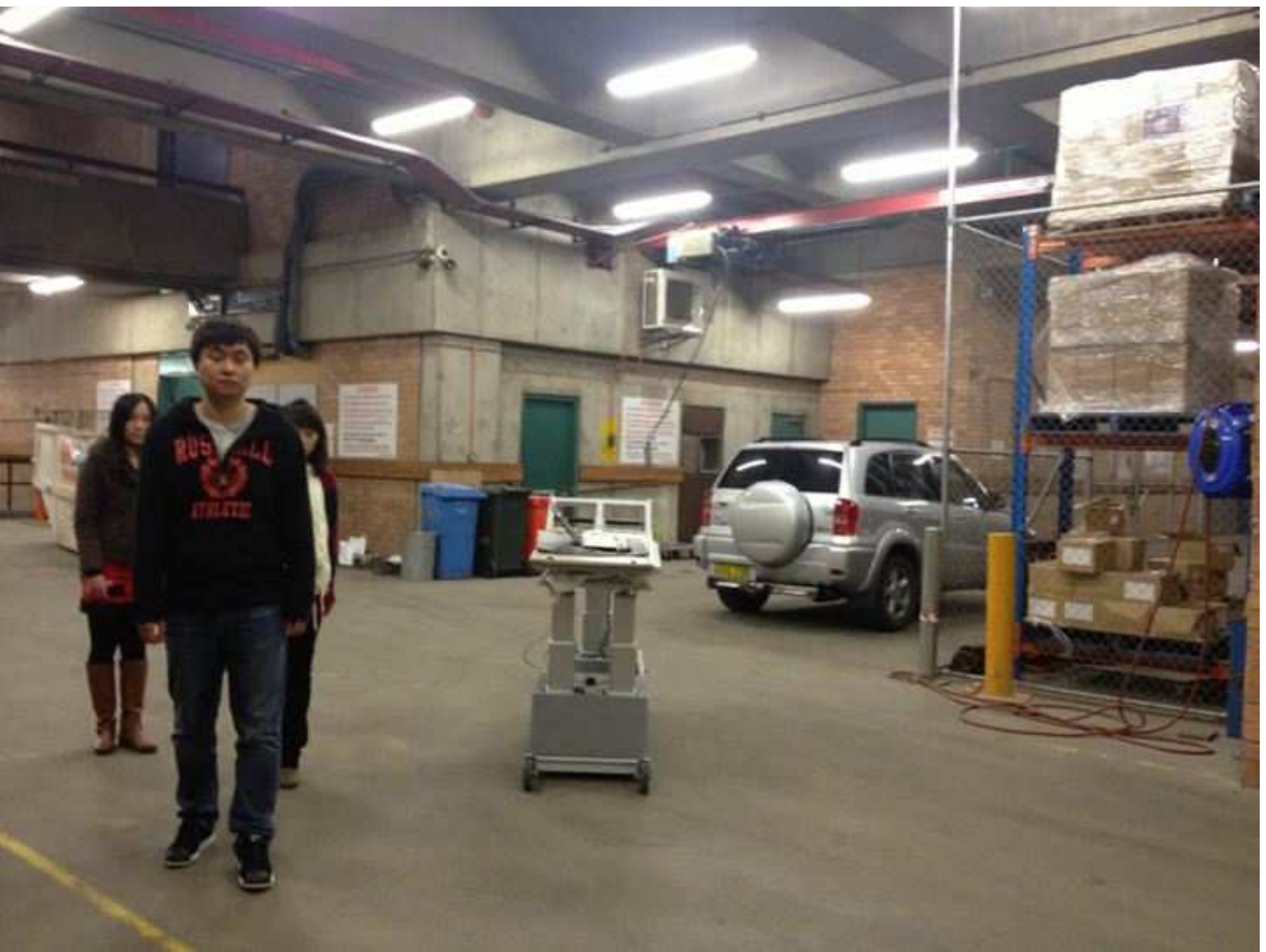}}
			\label{c7.exp93}}
			\subfigure[]{\scalebox{0.35}{\includegraphics{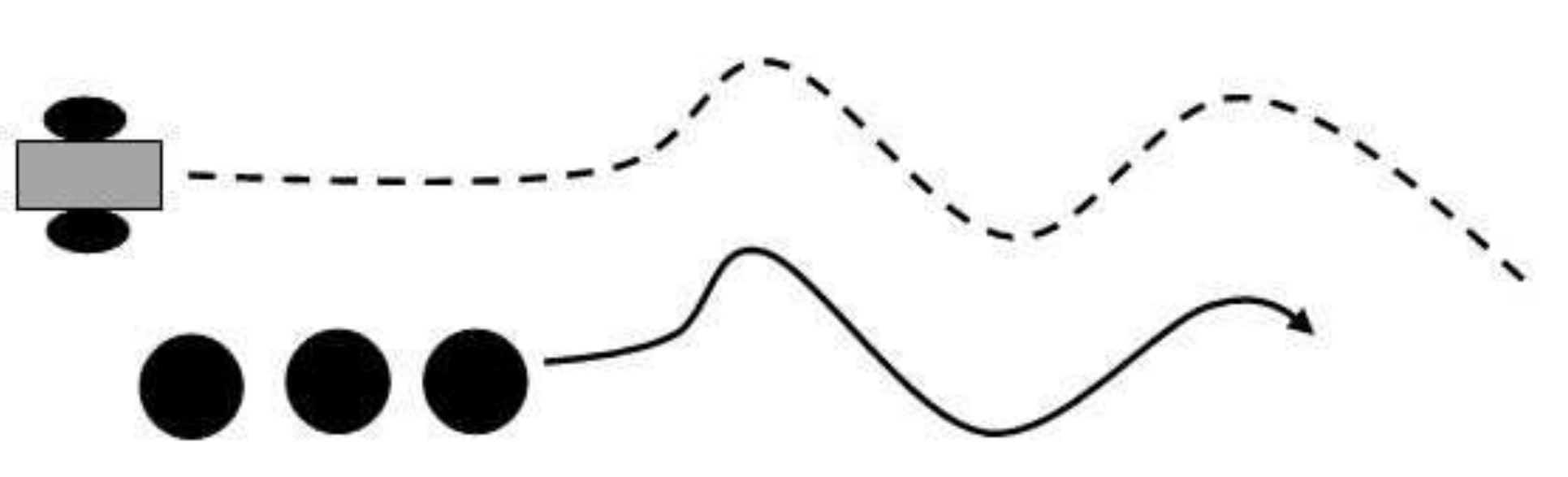}}
			\label{c7.exp94}}
			\caption{Flexbed avoids obstacle with dynamic dyforming shape}
			\label{c7.exp9}
			\end{figure}

			Finally, the Flexbed is able to deal with a dynamic deforming obstacle. In this experiment, the movement of the group of obstacles are lead by the a pilot (the first obstacle), and the pivot is moving in a sinusoidal fashion, resulting a snake-like deforming obstacle. The Flexbed is able to keep a constant distance with the obstacle as shown in Fig.\ref{c7.exp91}, Fig.\ref{c7.exp52} and Fig.\ref{c7.exp53} and  Fig.\ref{c7.exp54}.

	\section {Summary}

		A novel hospital bed system Flexbed and the experimental results of BINA and ENA on Flexbed are presented in this chapter.Unlike the existing hospital bed systems which are not motorised for transportation, Flexbed offers various options to control its motion. The implementations of BINA and ENA allow Flexbed to safely and automatically navigate in a hospital environment with stationary and dynamic obstacles. The performance and applicability of the algorithms are confirmed by experimental results.

\chapter {Decentralised Formation Building Algorithm for Groups of Mobile Robots} \label{C8}

	The decentralised control of groups of multiple mobile robots is another interesting and challenging topic in the field of mobile robotics.  The decentralised control is a robust control scheme since the groups of mobile robots are dynamically decoupled, meaning that the motion of one robot in the group does not directly affect that of the others. Each robot updates its speed and heading based on its closest neighbors rather relying on receiving updates from centralised agent. This control scheme ensures that failure of single robot (the centralised agent) will not cause the failure of the whole group. A decentralised control law guides the robots so that they will eventually move in the same direction with the same speed
\par
	The design of a decentralised control law for formation control of multiple mobile robots is even more difficult in the sense that the robots are only not required to move in the same direction but also in a desire geometric configuration.  Many of the existing formation building approaches consider simplest first or second order linear models for the motion of each robot, which causes the results to be heavily based on tools and methods from linear system theory. Such simple linear models allows the robots to drive with arbitrarily large angular velocity and arbitrarily small turning radius, which is unrealistic in real life application and implementations. Furthermore, many approaches employs the leader-follower scheme which leads to restrictive classes of robot communication graphs. In some other papers the robot communication graph
is assumed to be minimally rigid \cite{T4,T6} or time-invariant and connected \cite{T7} which is also quite restrictive. 
\par
	In this chapter, we present a constructive and easy-to-implement decentralised formation building algorithm for a group of autonomous wheeled mobile robtots,The formation building problem is particularly important for coverage problems for mobile sensor networks \cite{CTM09,CTM11,CTM12,CTM13,GAD08,KSL07,CH01}. The objective is to guide the robots in the group so that they eventually move in the same direction in the desired geometric configuration. This decentralised control law has its advantage of being able to work independently without a leader robot (the centralised agent). The robot uses sine global consensus to achieve and maintain the desired pattern. We consider two problems in this chapter, the first one is formation building with general class of communication graphs which are not assumed to be time-invariant or always connected. The second problem is formation building with anonymous robots which is more challenging because each robot are not aware of his position in the desire configuration, and the robots have to reach a consensus on their positions. The performance of the proposed formation building control laws is confirmed by computer simulations and experiments with real mobile robots.  
\par

	\section{Multi-Robots System Description}

		We consider a group of $n$ autonomous mobile robots traveling in a plane in continuous time, these robots are labeled from $1$ to $n$. The absolute Cartesian coordinates of the vehicle $i$ is represented by $(x_i(t), y_i(t))$. The orientation of the vehicle $i$ is represented by $\theta_i(t)$, which is measured from the x-axis in the counterclockwise direction. The value of $\theta_i (t)$ falls into the interval $\theta_i \in  \in (-\pi, \pi]$ with $\theta_i(t) =0$ corresponds to the direction of positive x-axis. The kinematic equations of the robot motion are given by

		\begin{eqnarray}  \label{UAV}
		\dot{x}_i(t) =& v_i(t)\cos(\theta_i(t))&;  \nonumber \\
		\dot{y}_i(t) =& v_i(t)sin(\theta_i(t))&;  \nonumber \\
		\dot{\theta}_i(t)=&\omega_i(t)&
		\end{eqnarray}

Here, $v_i(t)$ and $\omega_i(t)$ are the speed and angular velocity of the robot $i$, respectively. They are also the control inputs of robot $i$ and their values are limited by the following constraints:

		\begin{eqnarray}  \label{A2}
		-\omega^{max}\leq \omega_i(t)\leq \omega^{max}~~~~~\forall t\geq 0,
		\end{eqnarray}	
		\begin{eqnarray}  \label{A2v}
		V^{m}\leq v_i(t)\leq V^{M}~~~~~\forall t\geq 0
		\end{eqnarray}

		\begin{figure}[!b]
		\centering 
		\includegraphics[width =120mm]{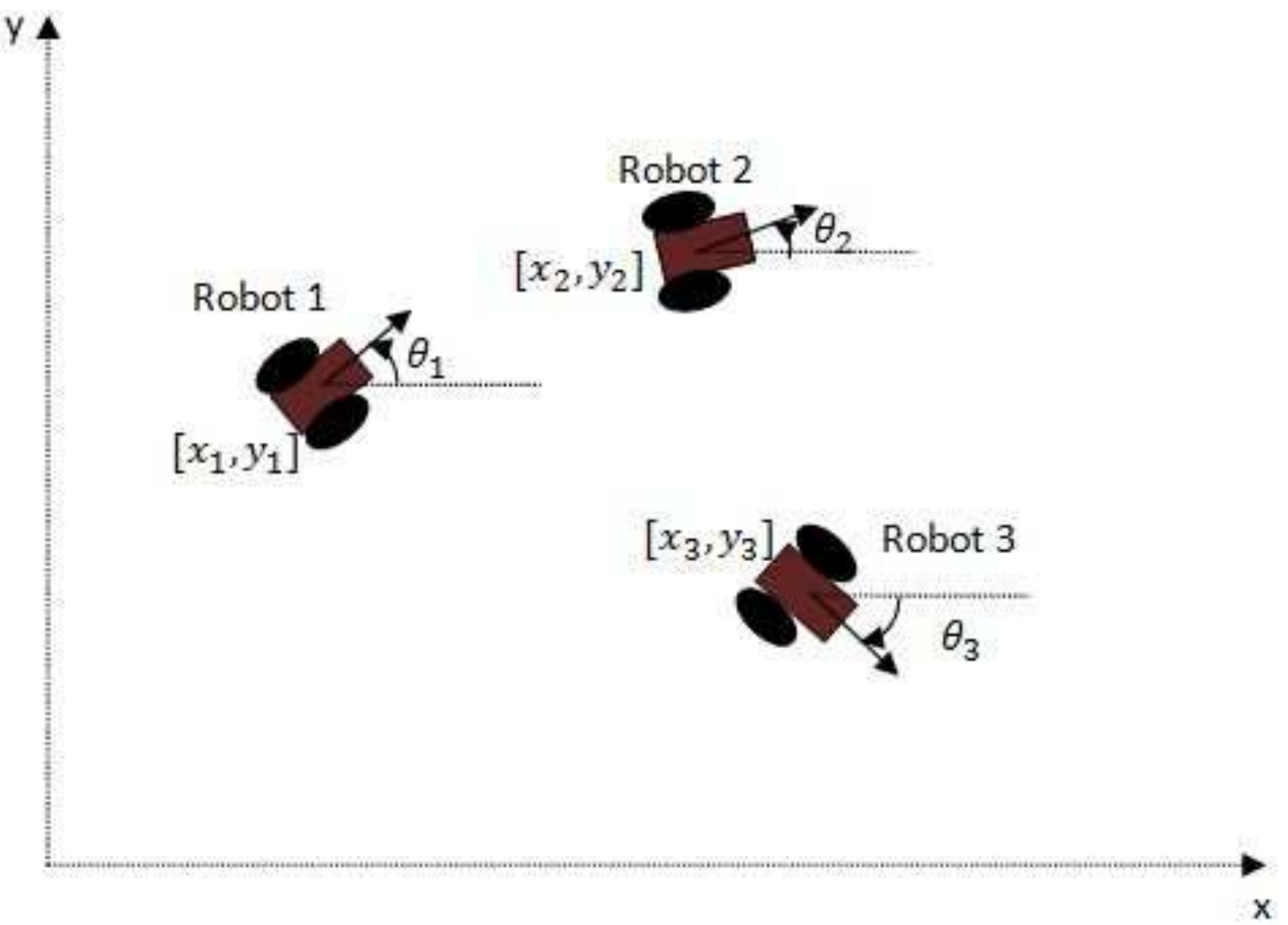}
		\caption{Multi-robots system}
		\label{multi_v}
		\end{figure}

for all robot serial number $i = 1,2, \dots, n$. Here $\omega^{max}>0$ and $0<V^{m}<V^{M}$ are given constants. Notice that since $v_i(t)>0$ for all $t$, the vehicles are assumed not to drive backwards. Fig.~\ref{multi_v} shows the illustration of the multi-robot system.
\par
		Introduce two-dimensional vectors $z_i(t)$ of the vehicles' coordinates and vectors $V_i(t)$ of the robots' velocities by

		\begin{eqnarray}  \label{Vi}
		z_i(t):= \left(
		\begin{array}{l}
		 x_i(t) \\
		 y_i(t)%
		\end{array}
		\right),~~~~ V_i(t):=\left(
		\begin{array}{l}
		 v_i(t)\cos(\theta_i(t)) \\
		 v_i(t)\sin(\theta_i(t))%
		\end{array}
		\right)
		\end{eqnarray}
		for all $i=1,2,\ldots,n$.
\par
		The vehicles in the groups communicate at discrete time instants $k = 0, 1, 2, \dots$.  We use a connected undirected graph $\mathcal{G}(k)$ to determine whether two robots $i$ and $j$ ($i \neq j$) are able to communicate at time $k$. The $n$ vertices of the graph $\mathcal{G}(j)$ corresponds to the $n$ vehicles in the plane. Two vertices are connected by an edge in the graph  $\mathcal{G}(j)$ if and only if the corresponding robots communicate at time $k$. Introduce the following notations.

		\begin{figure}[!h]
		\centering 
		\includegraphics[width =120mm]{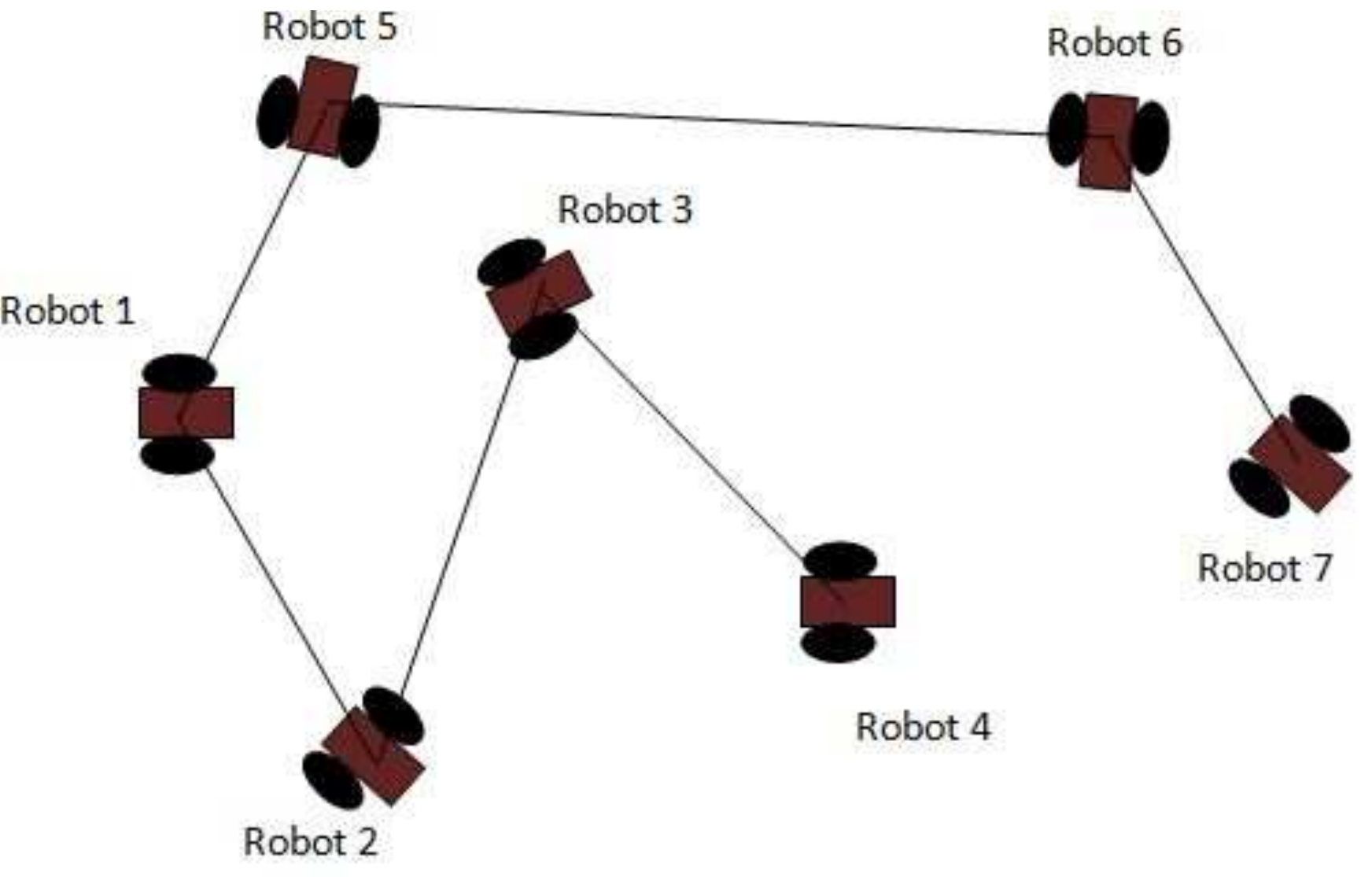}
		\caption{Connected undirected graph $\mathcal{G}(k)$}
		\label{graph}
		\end{figure}
\par
For any vertex $i\in \{1,2,\ldots,n\}$, $\mathcal{N}_i(k)$ denotes the set
of all vertices $j\neq i$ of $\mathcal{G}$ that are connected to $i$ by an
edge. Furthermore, $n_i(k)$ denotes the number of vertices in the set $%
\mathcal{N}_i(k)$.
\par

We will also need the following assumption.

\begin{assumption}
\label{As1} There exists an infinite sequence of contiguous, non-empty,
bounded time-intervals $[k_{j},k_{j+1})$, $j= 0,1,2,\ldots$, starting at $%
k_{0}=0$, such that across each $[k_{j},k_{j+1})$, the union of the
collection $\{\mathcal{G}(k): k\in[k_{j},k_{j+1})\}$ is a connected graph.
\end{assumption}
\par
		We use the consensus variables  $\tilde{\theta}_i(k)$, $\tilde{x}_{i}(k)$, $\tilde{y}_{i}(k)$ and $\tilde{v}_{i}(k)$ to achieve common goals of all the robots.   The
consensus variable $\tilde{\theta}_{i}(k)$ is used to achieve the common
heading of the formation, the consensus variables $\tilde{x}_{i}(k)$ and $%
\tilde{y}_{i}(k)$ are used to achieve the common origin of coordinates of
the formation, and the consensus variable $\tilde{v}_{i}(k)$ is used to
achieve the common speed of the formation.
\par
		All the robots starts with different initial values of $\tilde{\theta}_i(0)$, $\tilde{x}_{i}(0)$, $\tilde{y}_{i}(0)$ and $\tilde{v}_{i}(0)$. At time $k$, each of the robots in the group estimates $\tilde{\theta}_{i}(k)$, $\tilde{x}_{i}(k)$, $\tilde{y}_{i}(k)$ and $\tilde{v}_{i}(k)$ of the consensus formation parameters. This allows the robots to eventually converge to some consensus values which define a common formation and speed for all the robots.
\par

\begin{assumption}
\label{ch8:As2} The initial values of the consensus variables $\tilde{\theta}_{i}$ satisfy $\tilde{\theta}_{i}(0) \in [0, \pi)$ for all $i=1, 2, \ldots, n$.
\end{assumption}

We assume that the information on other vehicles that is available to the
vehicle $i$ at time $k$ is the coordinates $(x_j(k),y_j(k))$ (in practise, in mobile wireless networks, these coordinates can be estimated using robust Kalman filtering approach \cite{PI99,PPNS04,SP95,SAVP98,PPN05}) and the
consensus variables $\tilde{\theta}_{j}(k)$, $\tilde{x}_{j}(k)$, $\tilde{y}%
_{j}(k)$ and $\tilde{v}_{j}(k)$ for all $j\in \mathcal{N}_i(k)$. Also, the
vehicle's own coordinates, orientation and speed are measured at any time $%
t\geq 0$.
		\section {General Formation Building} \label{GFB}

			We propose the following rules for updating the consensus variables $\tilde{%
\theta}_{i}(k)$, $\tilde{x}_{i}(k)$, $\tilde{y}_{i}(k)$ and $\tilde{v}%
_{i}(k) $:
\par
		\begin{align}  \label{theta}
		\tilde{\theta}_i(k+1) &=\frac{\tilde{\theta}_i(k)+\sum_{j\in \mathcal{N}%
		_{i}(k)}\tilde{\theta}_j(k)} {1+|\mathcal{N}_{i}(k)|};  \nonumber \\
		\tilde{x}_i(k+1) &=\frac{x_i(k)+\tilde{x}_i(k)+\sum_{j\in \mathcal{N}%
		_{i}(k)}(x_j(k)+\tilde{x}_j(k))} {1+|\mathcal{N}_{i}(k)|}-x_i(k+1);
		\nonumber \\
		\tilde{y}_i(k+1)&=\frac{y_i(k)+\tilde{y}_i(k)+\sum_{j\in \mathcal{N}%
		_{i}(k)}(y_j(k)+\tilde{y}_j(k))} {1+|\mathcal{N}_{i}(k)|}-y_i(k+1);
		\nonumber \\
		\tilde{v}_i(k+1)&=\frac{\tilde{v}_i(k)+\sum_{j\in \mathcal{N}_{i}(k)}\tilde{v}%
		_j(k)} {1+|\mathcal{N}_{i}(k)|}.
		\end{align}

			In multi-robots systems, the essential informations have to be shared among the robots for them to achieve common goals. In the case of formation building, the use of rules in (\ref{theta}) allow the robots to achieve a consensus on the heading, speed and mass centre of the formation.
\par

\begin{lemma}
\label{T2} Suppose that Assumptions \ref{As1} and \ref{ch8:As2} hold and the
consensus variables are updated according to the decentralized control
algorithm (\ref{theta}). Then there exist constants $\tilde{\theta}_0$, $%
\tilde{X}_0$, $\tilde{Y}_0$ and $\tilde{v}_0$ such that
\begin{eqnarray}  \label{conv}
\lim_{k\rightarrow\infty} \tilde{\theta}_i(k)=\tilde{\theta}_0;  \nonumber \\
\lim_{k\rightarrow\infty} \tilde{v}_i(k)=\tilde{v}_0;  \nonumber \\
\lim_{k\rightarrow\infty} (x_i(k)+\tilde{x}_i(k))=\tilde{X}_0;  \nonumber \\
\lim_{k\rightarrow\infty} (y_i(k)+\tilde{y}_i(k))=\tilde{Y}_0.
\end{eqnarray}
for all $i=1,2,\ldots,n$. Moreover, the convergence in (\ref{conv}) is
exponentially fast.
\end{lemma}

			The statement of Lemma \ref{T2} immediately follows from the main result of \cite{JA03}.
\par

			Since the consensus update rules in (\ref{theta}) is based on the averaging of the consensus variables of neighbouring robots , the consensus variables in (\ref{conv}) should be converged to the same constants $\tilde{\theta}_0$, $%
\tilde{X}_0$, $\tilde{Y}_0$ and $\tilde{v}_0$ for any robot $i$. 
\par

Assume that $X_1,X_2,\ldots,X_n,Y_1,Y_2,\ldots,Y_n$ are given numbers. We define a navigation law that is globally stabilizing with initial conditions $(x_i(0),y_i(0),\theta_i(0))$, $i=1,2,\ldots,n$ and the configuration $\mathcal{C}=\{X_1,X_2,\ldots,X_n,Y_1,Y_2,\ldots,Y_n\}$, if there exist a Cartesian coordinate system and $\tilde{v}_0$ such that the solution of the closed-loop system (\ref{UAV}) with these initial conditions and the proposed navigation law in this Cartesian coordinate system satisfies:
\begin{eqnarray}  \label{conv_co}
\lim_{t\rightarrow\infty} (x_i(t)-x_j(t))=X_i-X_j,  \nonumber \\
\lim_{t\rightarrow\infty} (y_i(t)-y_j(t))=Y_i-Y_j,
\end{eqnarray}
\begin{eqnarray}  \label{conv_v}
\lim_{t\rightarrow\infty} \theta_i(t) = 0,  \nonumber \\
\lim_{t\rightarrow\infty} v_i(t) =\tilde{v}_0,
\end{eqnarray}
for all $1\leq i\neq j\leq n$.
\par

			Equation (\ref{conv_co}) and (\ref{conv_v}) defines the final convergence values for the consensus variables.  (\ref{conv_co}) means that all the robots will move in the same direction (along x-axis of the global Cartesian coordinate system) with the same speed defined by $\tilde{v}_0$. (\ref{conv_co}) means that the robots will move in the desired geometric configuration defined by $\mathcal{C}=\{X_1,X_2,\ldots,X_n,Y_1,Y_2,\ldots,Y_n\}$. Both (\ref{conv_co}) and (\ref{conv_v}) require a global Cartesian coordinate system. They define the final goal for formation building of multi-robot system.
\par

			\begin{figure}[!h]
			\centering
			\subfigure[]{\scalebox{0.60}{\includegraphics{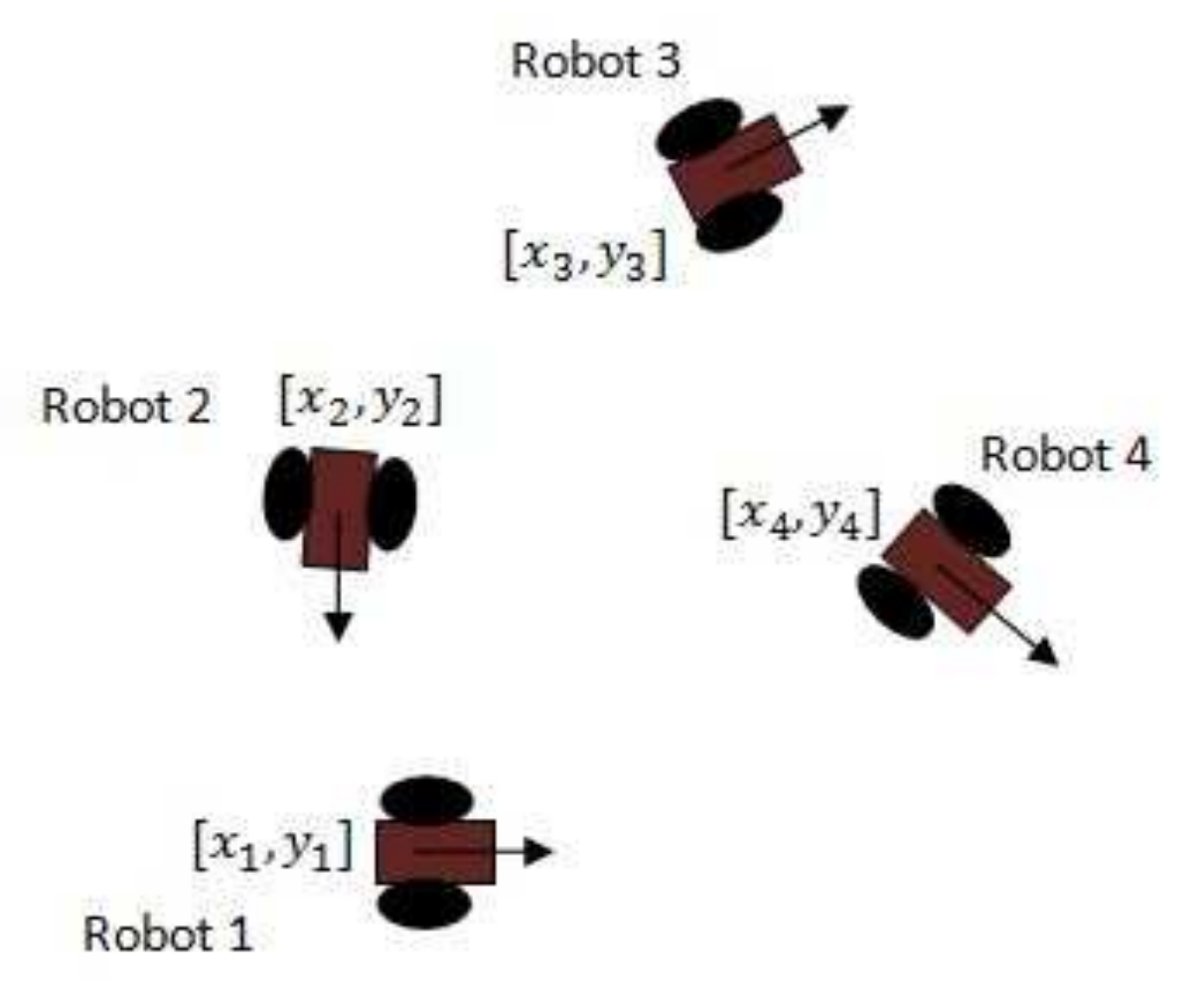}}
			\label{conv_init}}
			\hfill
			\subfigure[]{\scalebox{0.60}{\includegraphics{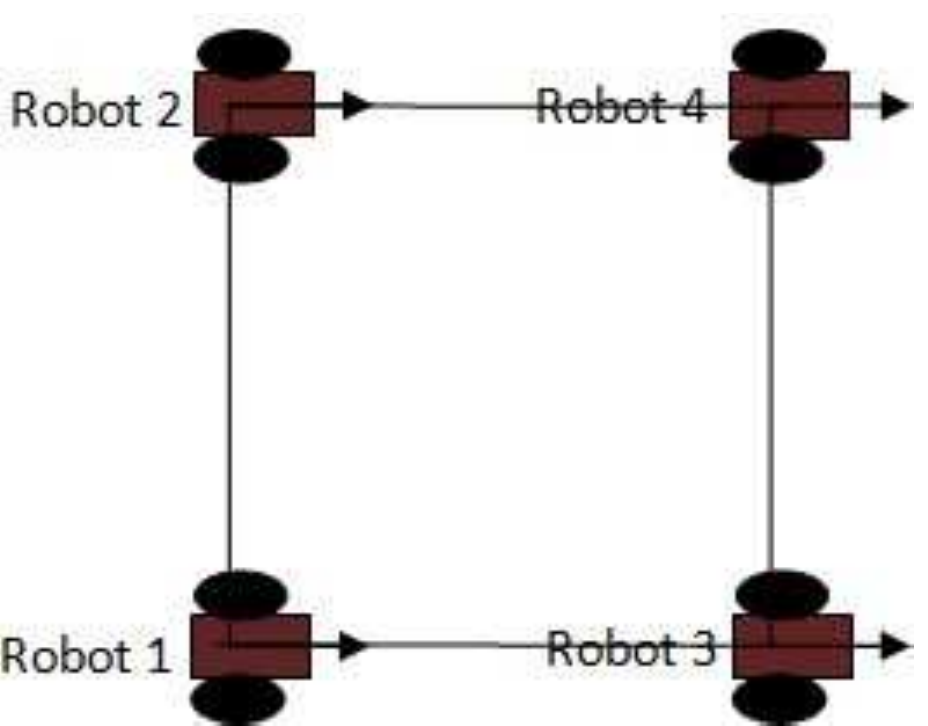}}
			\label{conv_final}}
			\caption{(a) Initial state (robots with arbitrary $\tilde{\theta}_i(0)$, $\tilde{x}_{i}(0)$, $\tilde{y}_{i}(0)$ and $\tilde{v}_{i}(0)$) (b) Final state (robots move in the same speed and direction with a desired square formation)}			
			\label{conv_ex}
			\end{figure}
			\par

			An simple example of formation building scenario for the robots to form a square of side $1$ is shown in Fig.~\ref{conv_ex}. In this case, for the geometric configuration $\mathcal{C}$, we can take $X_1=X_2=0, X_3=X_4=1, Y_1=Y_3=0,Y_2=Y_4=1$.
\par

The objective of this section is to present a globally stabilizing control
law for the multi-vehicle formation under consideration.

In our globally stabilizing law, we will use the consensus variables $\tilde{%
\theta}_{i}(k)$, $\tilde{x}_{i}(k)$, $\tilde{y}_{i}(k)$ and $\tilde{v}%
_{i}(k) $ updated according to (\ref{theta}). Introduce the corresponding
piecewise constant continuous time variables as
\begin{eqnarray}  \label{tilde1}
\tilde{\theta}_{i}(t)&:=\tilde{\theta}_{i}(k)~~~~\forall t\in(k,k+1);
\nonumber \\
\tilde{x}_{i}(t)&:=\tilde{x}_{i}(k)~~~~\forall t\in(k,k+1);  \nonumber \\
\tilde{y}_{i}(t)&:=\tilde{y}_{i}(k)~~~~\forall t\in(k,k+1);  \nonumber \\
\tilde{v}_{i}(t)&:=\tilde{v}_{i}(k)~~~~\forall t\in(k,k+1).
\end{eqnarray}

Let $c>0$ be any constant such that
\begin{eqnarray}  \label{cc}
c>\frac{2V^{M}}{\omega^{max}}.
\end{eqnarray}
\par

Furthermore, let $h_1$ and $h_2$ be non-zero two-dimensional vectors, and
let $\alpha$ be the angle between the vectors $h_1$ and $h_2$ measured from $%
h_1$ in the counter-clockwise direction, $-\pi <\alpha\leq \pi$, i.e. $%
\alpha =0$ if $h_1=h_2$. Now introduce the following function:
\begin{eqnarray}  \label{f}
f(h_1,h_2):= sign(\alpha)
\end{eqnarray}
where $sign(\cdot)$ is defined by 
\begin{eqnarray}  \label{sign}
sign (\alpha) := \left\{
\begin{array}{lll}
-1&  if & \alpha < 0\\
0 &  if &\alpha = 0\\
1 & if &\alpha > 0
\end{array} \right.
\end{eqnarray}
		
For any time $t$ and any robot $i$, we consider a Cartesian coordinate
system with the $x-$axis in the direction $\tilde{\theta}_i(t)$ (according
to the definition (\ref{tilde1}), $\tilde{\theta}_i(t)$ is piecewise
constant). In other words, in this coordinate system $\tilde{\theta}_i(t)=0$%
, $x_i(t),y_i(t)$ are now coordinates of the robot $i$ in this system. Notice that we now formulate
our decentralized control law for each robot in its
own coordinate system. Since according to Lemma \ref{T2}, $\tilde{\theta}_i(k)$ converges to the same value for all $i$, all these robots' coordinate systems converge to the same coordinate system in
which (\ref{conv_co}) holds.

Introduce the functions $h_{i}(t)$ as
\begin{equation}
h(t):=(x_{i}(t)+\tilde{x}_{i}(t))+X_{i}+t\tilde{v}_{i}(t)  \label{hh}
\end{equation}%
for all $i$. Also, for all $i=1,2,\ldots ,n$ introduce two-dimensional vectors
$g_{i}(t)$ as
\begin{eqnarray}
g_{i}^{x}(t) &:&=\left\{
\begin{array}{ll}
h_{i}(t)+c & if~x_{i}(t)\leq h_{i}(t) \\
x_{i}(t)+c & if~x_{i}(t)>h_{i}(t)%
\end{array}%
\right.  \nonumber  \label{gij} \\
g_{i}^{y}(t) &:&=(y_{i}(t)+\tilde{y}_{i}(t))+Y_{i}, \nonumber \\g_{i}(t) &:&=\left(
\begin{array}{l}
g_{i}^{x}(t) \\
g_{i}^{y}(t)%
\end{array}%
\right)
\end{eqnarray}%
and two-dimensional vectors $d_{i}(t)$ as
\[
d_{i}(t):=g_{i}(t)-z_{i}(t)
\]%
where $z_{i}(t)$ is defined by (\ref{Vi}).

\begin{assumption}
\label{AsI}
We assume that the configuration $\mathcal{C} =\{X_1,X_2,\ldots,X_n,Y_1,Y_2,\\\* %
\ldots,Y_n\}$ and the constant $c$ satisfying (\ref{cc}) are known to each
robot $i$.
\end{assumption}

Now we introduce the following decentralized control law:
\begin{eqnarray}  \label{Co3}
v_i(t)=\left\{
\begin{array}{ll}
V^M & if~~~x_i(t)\leq h_i(t) \\
V^m & if~~~x_i(t)> h_i(t)%
\end{array}
\right.  \nonumber \\
\omega_i(t)=\omega^{max}f(V_i(t),d_{i}(t)),
\end{eqnarray}
for all $i=1,2,\ldots,n$. Here $f(\cdot,\cdot), V_i(\cdot),d_{i}(\cdot)$ are
defined by (\ref{Vi}), (\ref{f}).

We will need the following assumption.

\begin{assumption}
\label{Asvv} The initial robots' speeds satisfy
\[
V^{m}< v_i(0)< V^{M}
\]
for all $i=1,2,\ldots,n$.
\end{assumption}

Notice that Assumption \ref{Asvv} is just slightly stronger the requirement (\ref{A2v}) for $t=0$ where 
non-strict inequalities are required.
\par

\begin{theorem}
\label{T1} Consider the autonomous robots described by the equations (\ref%
{UAV}) and the constraints (\ref{A2}), (\ref{A2v}). Let $\mathcal{C}%
=\{X_1,X_2,\ldots,X_n,Y_1,Y_2,\ldots,Y_n \}$ be a given configuration.
Suppose that Assumptions \ref{As1}, \ref{ch8:As2}, \ref{AsI} and \ref{Asvv} hold, and $c$
is a constant satisfying (\ref{cc}). Then, the decentralized control law (%
\ref{theta}), (\ref{Co3}) is globally stabilizing with any initial
conditions and the configuration $\mathcal{C}$.
\end{theorem}

The proof of this theorem is presented in \cite{SAVW13}
\par

The main idea of the navigation law (\ref{Co3}) can be explained as follows. Each vehicle $i$ is
guided towards a fictitious target $T_{i}$ that is always located "ahead" of
the desired vehicle's position relative to its neighbours. The reason why we
guide the vehicle towards a fictitious target but not the desired relative
vehicle's position itself is that if we guided the
vehicle towards the desired relative position, we would have $%
\|d_{i}(t)\|\rightarrow 0$, therefore, $\omega_i(t)\rightarrow\infty$ and
the constraint (\ref{A2}) would be violated. Notice that our method for
guidance towards a fictitious target $T_{i}$ is a pure pursuit type guidance
law~\cite{SAVT10}.

		\section {Formation Building with Anonymous Robots} \label{FBAR}

			In this section, we consider a more challenging problem: formation building with anonymous robots. It is more challenging than general formation building problem in the sense that each robots does not know its position in the configuration $\mathcal{C}=\{X_1,X_2,\ldots,X_n,Y_1,Y_2,%
\ldots,Y_n\}$ in advance, the robots should reach a consensus on their position. In the area of robotics, it is common to use the multi-robot task allocation approach to similar problems. However,  most work on multi-robot task allocation  has been ad hoc and empirical especially in the case of an arbitrarily large number of robots; see e.g. \cite{T8,T9}. In this section, we propose a randomised algorithm to handle this problem. The randomised algorithm for anonymous robots is a modification of the algorithm from \cite {AVSF12}.
\par
			We first state our definition of navigation law with anonymous robots. 
A navigation law is said to be globally stabilizing with
anonymous robots and the configuration $\mathcal{C}=\{X_1,X_2,%
\ldots,X_n,Y_1,Y_2,\ldots,Y_n\}$ if for any initial conditions $%
(x_i(0),y_i(0),\theta_i(0))$, there exists a permutation $r(i)$ of the index
set $\{1,2,\ldots,n\}$ such that for any $i=1,2,\ldots,n$, there exist a
Cartesian coordinate system and $\tilde{v}_0$ such that the solution of the
closed-loop system (\ref{UAV}) with the proposed navigation law in this
Cartesian coordinate system satisfies (\ref{conv_v}) and
\begin{eqnarray}  \label{def2}
\lim_{t\rightarrow\infty} (x_{i}(t)-x_{j}(t))&=X_{r(i)}-X_{r(j)},  \nonumber
\\
\lim_{t\rightarrow\infty} (y_{i}(t)-y_{j}(t))&=Y_{r(i)}-Y_{r(j)},
\end{eqnarray}
for all $1\leq i\neq j\leq n$.
\par

\begin{figure}[!h]
		\centering 
		\includegraphics[width =100mm]{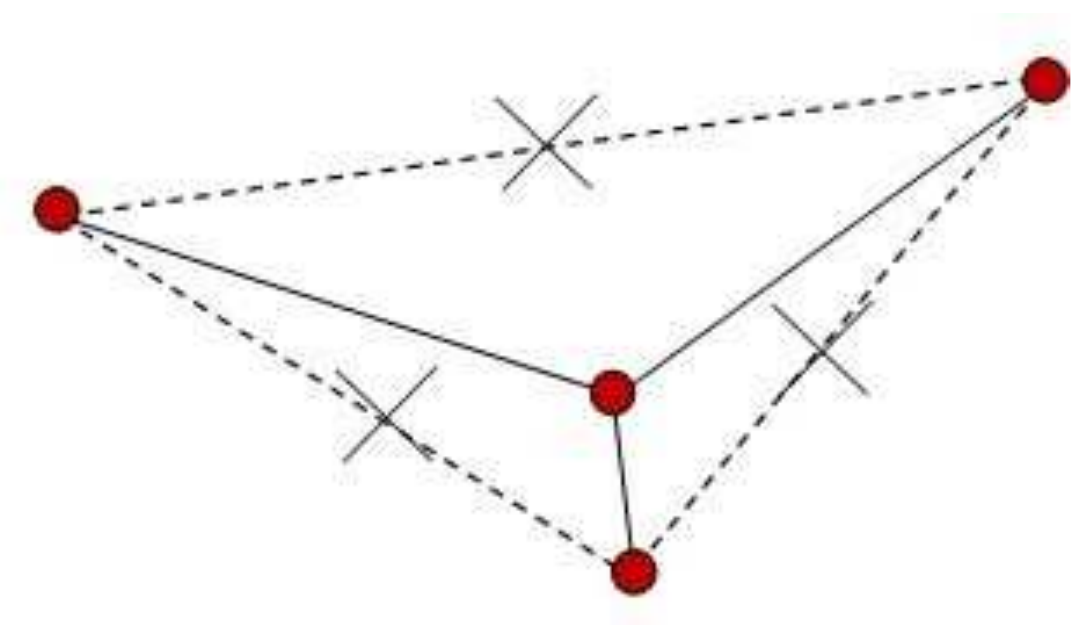}
		\caption{Graph $\mathcal{P}$}
		\label{graph_ex}
		\end{figure}

			Let $R>0$ be the detection range of the robots, i.e, each robot $i$ is able to detect all other robots inside the circle of radius $R$ at the current position of robot $i$.  Furthermore, let $0<\epsilon<\frac{R}{2%
}$ be a given constant. For any configuration $\mathcal{C}%
=\{X_1,X_2,\ldots,X_n,Y_1,Y_2,\ldots,Y_n\}$ introduce a undirected graph $%
\mathcal{P}$ consisting of $n$ vertices. Vertices $i$ and $j$ of the graph $%
\mathcal{P}$ are connected by an edge if and only if $\sqrt{%
(X_i-X_j)^2+(Y_i-Y_j)^2)}\leq R-2\epsilon$, see Fig.~\ref{graph_ex} for example. We will need the following
assumption.

\begin{assumption}
\label{As8} The graph $\mathcal{P}$ is connected.
\end{assumption}

We present a randomised algorithm to build an index permutation function $%
r(i)$. Let $N\geq 1$ be a given integer. Let $r(0,i)\in \{1,2,...,n\}$ be
any initial index values where $i=1,2,...,n$.

As in the navigation law (\ref{theta}), (\ref{Co3}), for any time $t$ and
any robot $i$, we consider a Cartesian coordinate system with the $x-$axis
in the direction $\tilde{\theta}_i(t)$ (according to the definition (\ref%
{tilde1}), $\tilde{\theta}_i(t)$ is piecewise constant). In other words, in
this coordinate system $\tilde{\theta}_i(t)=0$, $x_i(t),y_i(t)$ are now
coordinates of the robot $i$ in this system. Furthermore, we say that a
vertex $j$ of the graph $\mathcal{P}$ is vacant at time $kN$ for robot $i$
if there is no any robot inside the circle of radius $\epsilon$ centred at
the point 
\begin{eqnarray*}
\left(
\begin{array}{l}

(x_i(kN)+\tilde{x}_i(kN))+X_j+ kN\tilde{v}_i(kN) \\
(y_i(kN)+ \tilde{y}_i(kN))+Y_j
\end{array}
\right)
\end{eqnarray*}
Let $\mathcal{S}(kN,i)$ denote the set of vertices of $\mathcal{P}$
consisting of $r(kN,i)$ and those of vertices of $\mathcal{P}$ that are
connected to $r(kN,i)$ and vacant at time $kN$ for robot $i$. Let $|\mathcal{%
S}(kN,i)|$ be the number of elements in $\mathcal{S}(kN,i)$. It is clear
that $1\leq |\mathcal{S}(kN,i)|$ because $r(kN,i)\in \mathcal{S}(kN,i)$%
. Moreover, introduce the Boolean variable $b_i(kN)$ such that $b_i(kN):=1$
if there exists another robot $j\neq i$ that is inside of the circle of
radius $\epsilon$ centred at 
\begin{eqnarray*}
\left(
\begin{array}{l}
(x_i(kN)+\tilde{x}_i(kN))+X_i+kN\tilde{v}_i(kN) \\
(y_i(kN)+\tilde{y}_i(kN))+Y_i
\end{array}
\right)
\end{eqnarray*}
at time $kN$, and $b_i(kN):=0$
otherwise. We propose the following random algorithm:
\begin{eqnarray}  \label{random}
&r((k+1)N,i)=r(kN,i)~~~if \nonumber \\ &\left(b_i(kN)=0~or~\left(b_i(kN)=1~and~|%
\mathcal{S}(kN,i)|=1\right)\right);  \nonumber \\
&r((k+1)N,i)=j \nonumber \\ &~~with~~probability ~~\frac{1}{|\mathcal{S}(kN,i)|}~~~\forall
j\in \mathcal{S}(kN,i)  \nonumber \\
&if \left(b_i(kN)=1~~~and~~~|\mathcal{S}(kN,i)|>1\right).
\end{eqnarray}

Now we are in a position to present the main result of this section.

\begin{theorem}
\label{T3} Consider the autonomous robots described by the equations (\ref%
{UAV}) and the constraints (\ref{A2}), (\ref{A2v}). Let $\mathcal{C}%
=\{X_1,X_2,\ldots,X_n,Y_1,Y_2,\ldots,Y_n \}$ be a given configuration.
Suppose that Assumptions \ref{As1}, \ref{ch8:As2}, \ref{Asvv} and \ref{As8}
hold, and $c$ is a constant satisfying (\ref{cc}). Then, initial conditions $%
(x_i(0),y_i(0),\theta_i(0))$, $i=1,2,\ldots,n$ there exists an integer $%
N_0>0 $ such that for any $N\geq N_0$, the decentralized control law (\ref%
{theta}), (\ref{Co3}), (\ref{random}) with probability $1$ is globally
stabilizing with these initial conditions and the configuration $\mathcal{C}$.
\end{theorem}
\par
	
		The proof of this theorem is presented in \cite{SAVW13}.
\par

		\section {Computer Simulation Results}

			\subsection {General Formation Building}

\begin{figure}[!h]
		\centering 
		\includegraphics[width =150mm]{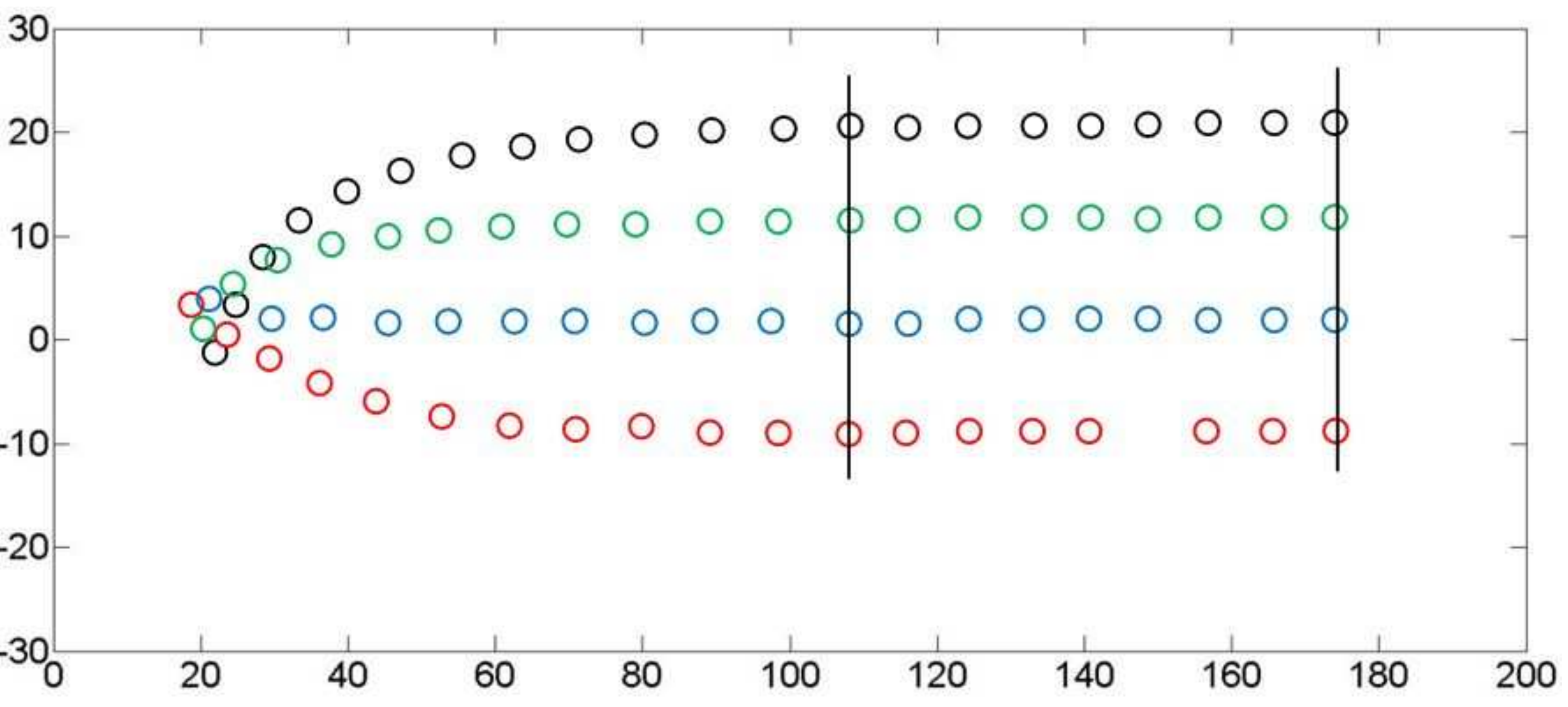}
		\caption{Straight line formation with equal separation between two consecutive robots}
		\label{c8.sim1}
		\end{figure}

			In this section, we present simulation results for the formation building algorithm in  present in Section~\ref{GFB}. 
			In Fig.~\ref{c8.sim1}, the simulation result for straight line formation with four robots are presented. At the beginning of the simulation, the robots are randomly positioned and orientated, they also start with various speeds, i.e. The robots start with different values of $\tilde{\theta}%
_{i}(0)$, $\tilde{x}_{i}(0)$. The motions of the robots are governed by (\ref{UAV}), (\ref{A2})
and (\ref{A2v}) with decentralized law (\ref{theta}), (\ref{Co3}) employed. The robots updates consensus information with the neighbouring robots and successfully form a straight line with equal distance between two consecutive robots. The formation is achieved at which the first straight line is drown, and the robots are able to keep the formations for the rest of time, which demonstrates the consistency of the proposed formation building algorithm over time.
\par

\begin{figure}[!h]
		\centering 
		\includegraphics[width =150mm]{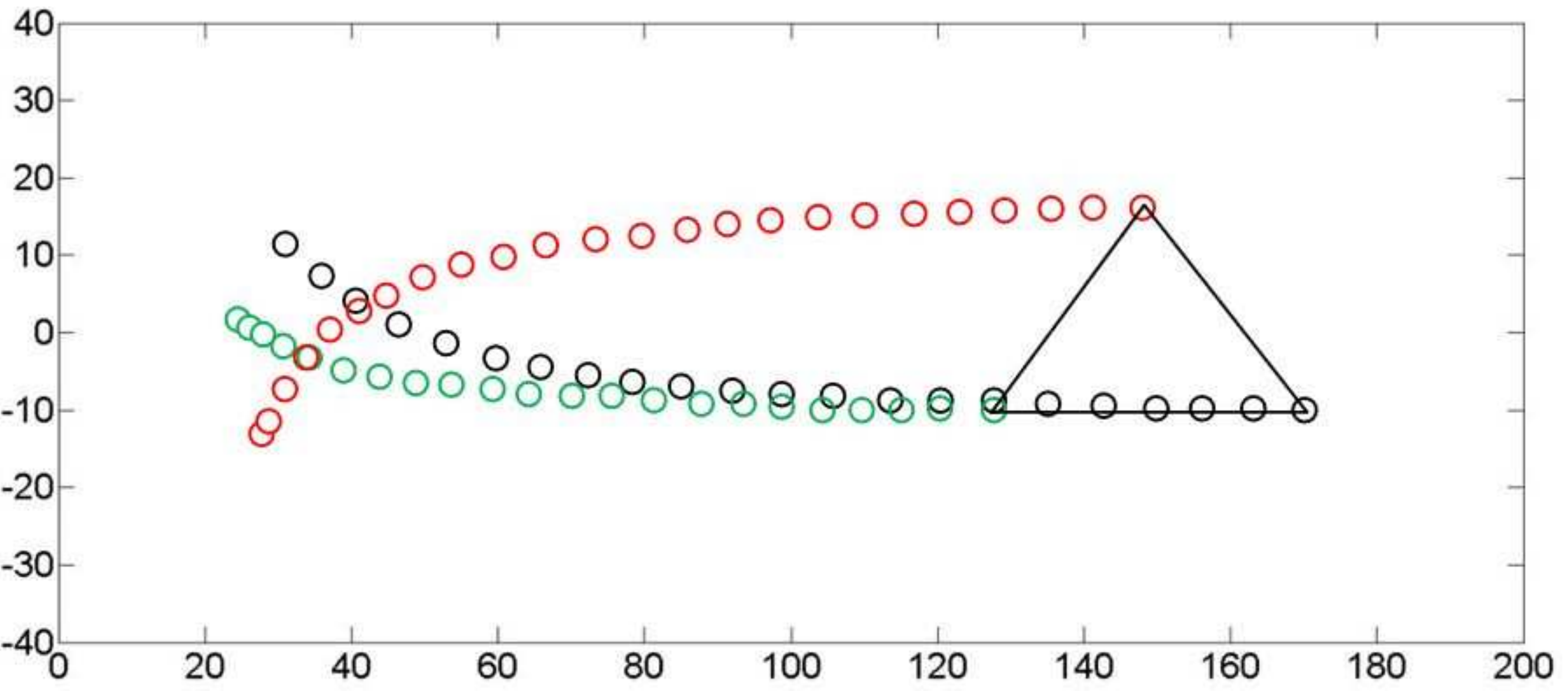}
		\caption{Equilateral Triangle formation}
		\label{c8.sim2}
		\end{figure}

\begin{figure}[!h]
		\centering 
		\includegraphics[width =150mm]{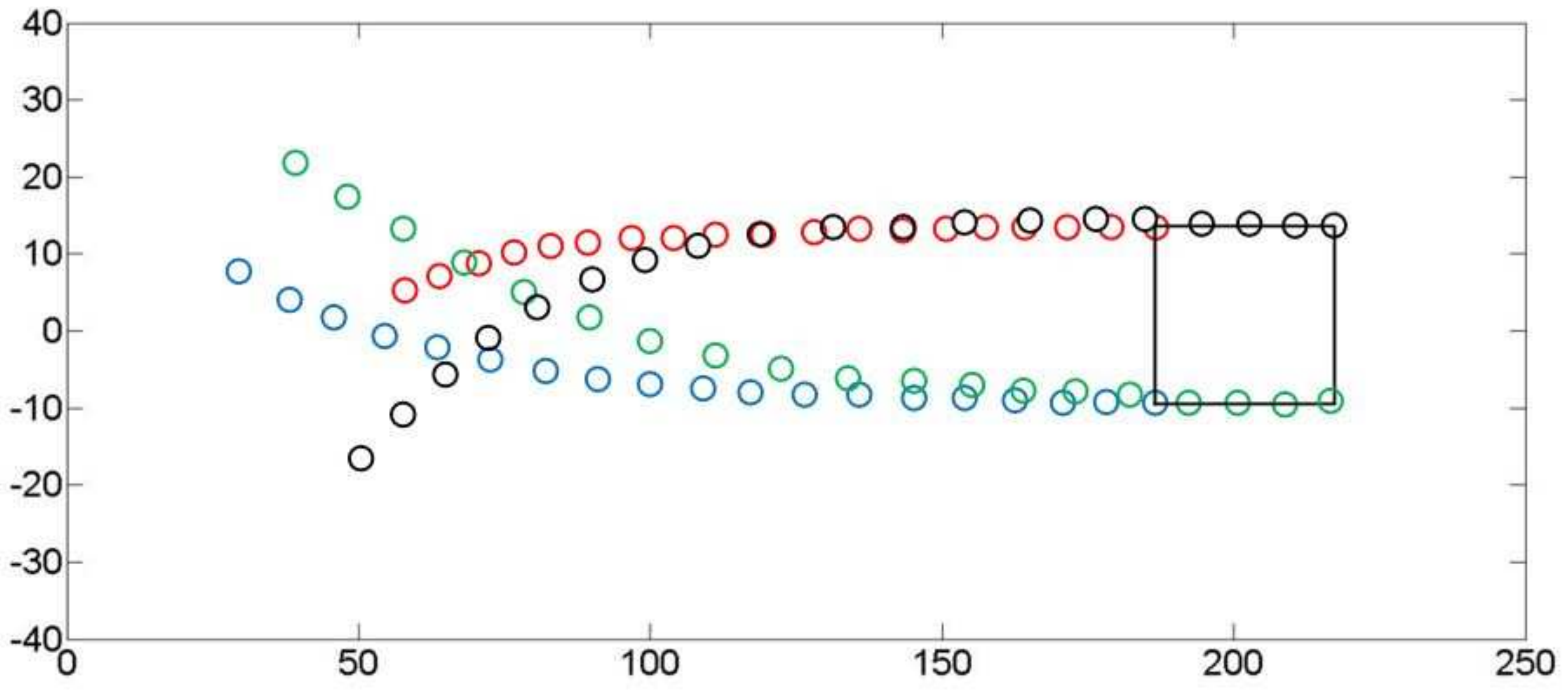}
		\caption{Robots form a square shape}
		\label{c8.sim3}
		\end{figure}

\begin{figure}[!h]
		\centering 
		\includegraphics[width =150mm]{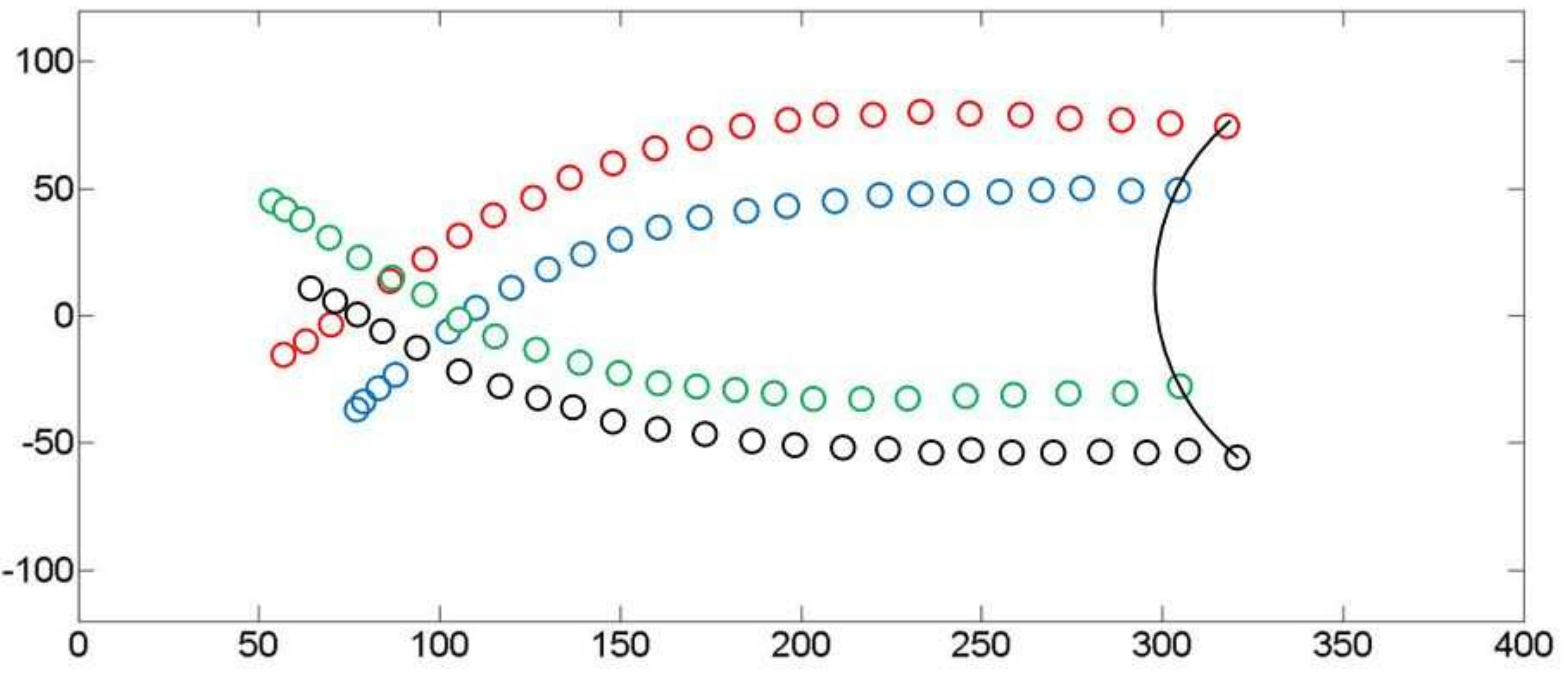}
		\caption{Robots form an arc}
		\label{c8.sim4}
		\end{figure}

\begin{figure}[!h]
		\centering 
		\includegraphics[width =150mm]{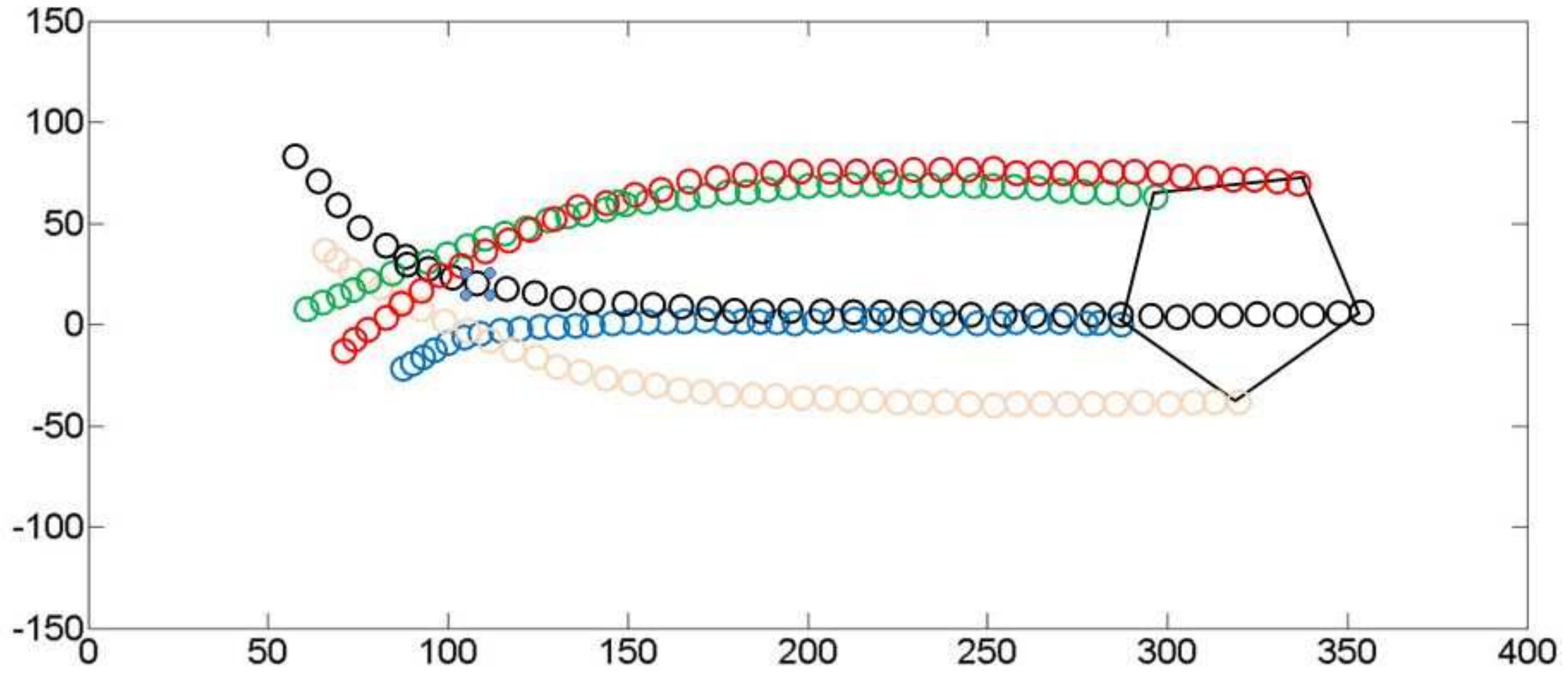}
		\caption{Robots form regular pentagon}
		\label{c8.sim5}
		\end{figure}

		In Fig.~\ref{c8.sim2}, Fig.~\ref{c8.sim3}, Fig.~\ref{c8.sim4} and Fig.~\ref{c8.sim5}, we present more complicated formation simulation results. From these results, we show that with any initial conditions and various configuration $\mathcal{C}$, the robots are always able to achieve global stability.

		The following tables~\ref{xdata} and ~\ref{ydata} show the detailed results over several experiment runs (performed in simulation) for four robots (square formation) .
\par

		\begin{table}[!h]
		\caption{Experiment Results over ten experiment runs (x coordinate)}
		\begin{tabular}{llllll}
		\hline
				      & $X_1 - X_2$ & $X_2 - X_3$ & $X_3 - X_4$ & $X_4 - X_1$  \\
		\hline
		Desired difference (X) & -5.000 & 0 & 5.000  & 0  \\
		\hline
		\hline
		Actual output 1 & -4.9071 &   0.0698  &  5.0868 &   0.0357 \\
		Actual output 2 & -5.0515  &  0.0486  &  4.9784  &  0.0311 \\
		Actual output 3 & -4.9342  &  0.0412  &  4.9064  & -0.0446 \\
		Actual output 4 & -4.9092  &  -0.0806  & 5.0647   & 0.0390 \\
		Actual output 5 & - 4.9634 &   0.0900   & 4.9069  & -0.0123\\
		Actual output 6 & -4.9763   & 0.0531 &   5.0590  & -0.0626  \\
		Actual output 7 & -4.9980  &  -0.0109  & 5.0293  &  0.0419  \\
		Actual output 8 & -5.0509  &  -0.0448 &   5.0359   & 0.0310  \\
		Actual output 9 & -4.9325 &   -0.0762 &  4.9997  &  0.0919  \\
		Actual output 10 & -4.9681   & 0.0171&    4.9448    &0.0503  \\
		\hline
	\label {xdata}	
	\end{tabular}
		\end{table}

		The average error in x-coordinate is $-0.0309m, -0.0107m, -0.0012m,  -0.0201m$. and the standard deviation is $ 0.0521, 0.0609, 0.0652, 0.0465$.
\par

		\begin{table}[!h]
		\caption{Experiment Results over ten experiment runs (y coordinate)}
	\begin{tabular}{llllll}
		\hline
				      & $Y_1 - Y_2$ & $Y_2 - Y_3$ & $Y_3 - Y_4$ & $Y_4 - Y_1$  \\	
		\hline
		Desired difference (Y)  & 0 & -5.000 & 0  & 5.000  \\
		\hline
		\hline
		Actual output 1 & -0.0490 &  -5.0012  &  0.0398  &  5.0782 \\
		Actual output 2 & 0.0919  &  -5.0094  &  -0.0723 &  4.9299 \\
		Actual output 3 & -0.0485&  -5.0681&   -0.0491 &  5.0629  \\
		Actual output 4 & -0.0513  & -5.0859  & -0.0300 &  4.9393  \\
		Actual output 5 & -0.0498 &  -5.0232   & -0.0053  & 4.9703  \\
		Actual output 6 & 0.0662&    -5.0171   & 0.0099  &  5.0834  \\
		Actual output 7 & -0.0428 &  -5.0514    &0.0507  &  4.9761  \\
		Actual output 8 &0.0136 &   -4.9152  &  -0.0892  & 5.0062  \\
		Actual output 9 & 0.0558&    -5.0868 &  -0.0740 &  5.0138 \\
		Actual output 10 & -0.0061 &  -4.9024   & -0.0326  &  4.9324 \\
		\hline
	\label {ydata}	
	\end{tabular}
	\end{table}

	The average error in y-coordinate is $0.0020m, 0.0161m, 0.0252m,  0.0008m$, and the standard deviation is $ 0.0556, 0.0643,0.0483, 0.0595$.
\par
	
			\subsection {Formation Building with Anonymous Robots}

\begin{figure}[!h]
		\centering 
		\includegraphics[width =150mm]{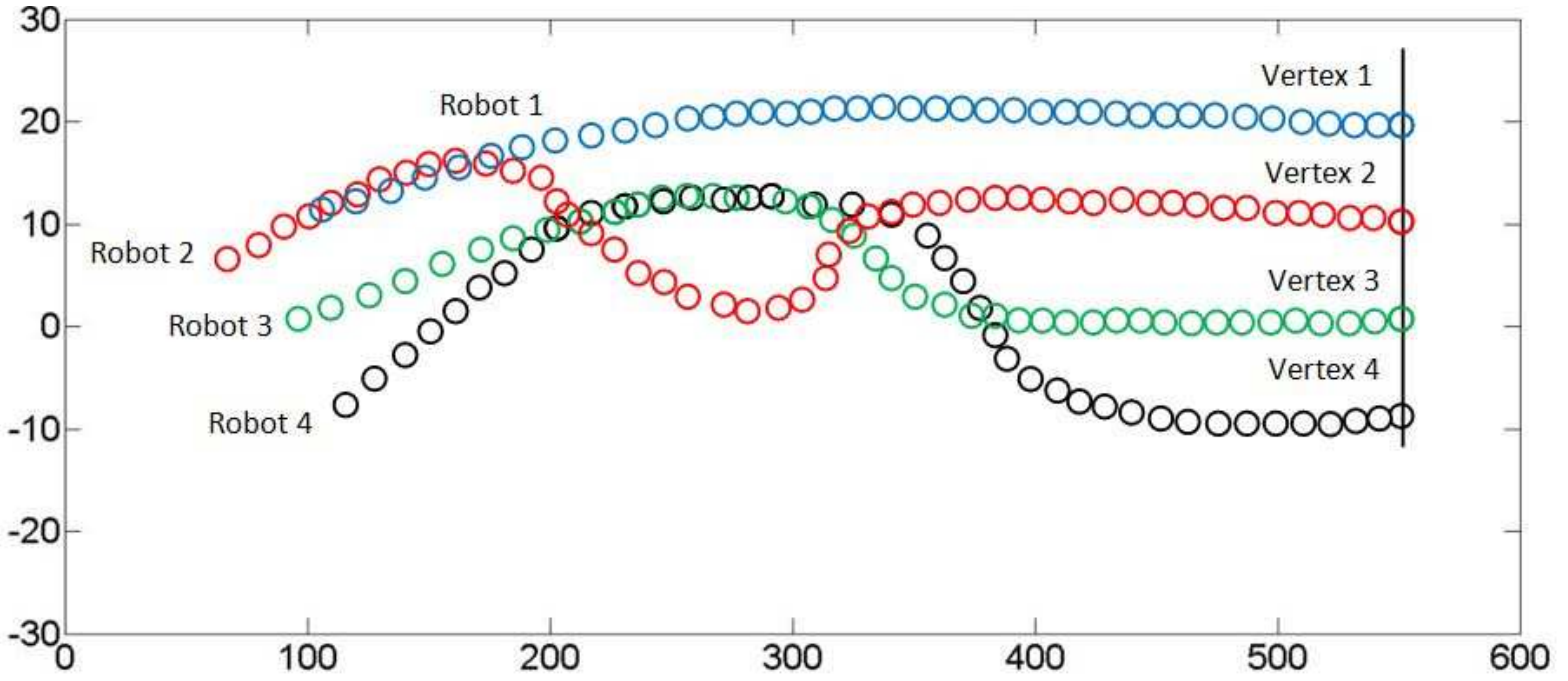}
		\caption{Straight line formation with anonymous robots, permutation applied}
		\label{c8.sim6}
		\end{figure}

				We present simulation results for formation building with anonymous robots. This randomised formation building algorithm is presented in Section~\ref{FBAR}.
				In Fig.~\ref{c8.sim6}, we present a similar simulation as in Fig.~\ref{c8.sim1} with anonymous robots. The major difference is that the robots are not aware of their positions in the configuration $\mathcal{C}$ and the motion and the motions of the robots are guided by the proposed randomised algorithm with index permutation. We can observe from Fig.~\ref{c8.sim6}, two robots may aim for the same position in the configuration $\mathcal{C}$ during the simulation. To resolve this situation,  the index permutation assigns a new vertex to one of the two robots so that they do not "compete" for the same index. This permutation is likely to take place several time in one simulation. Table~\ref{perm_index} shows the indices assigned to all the robots during simulation presented in Fig.~\ref{c8.sim6}, each index corresponds one vertex depicted in Fig.~\ref{c8.sim6}.

\begin{figure}[!h]
		\centering 
		\includegraphics[width =150mm]{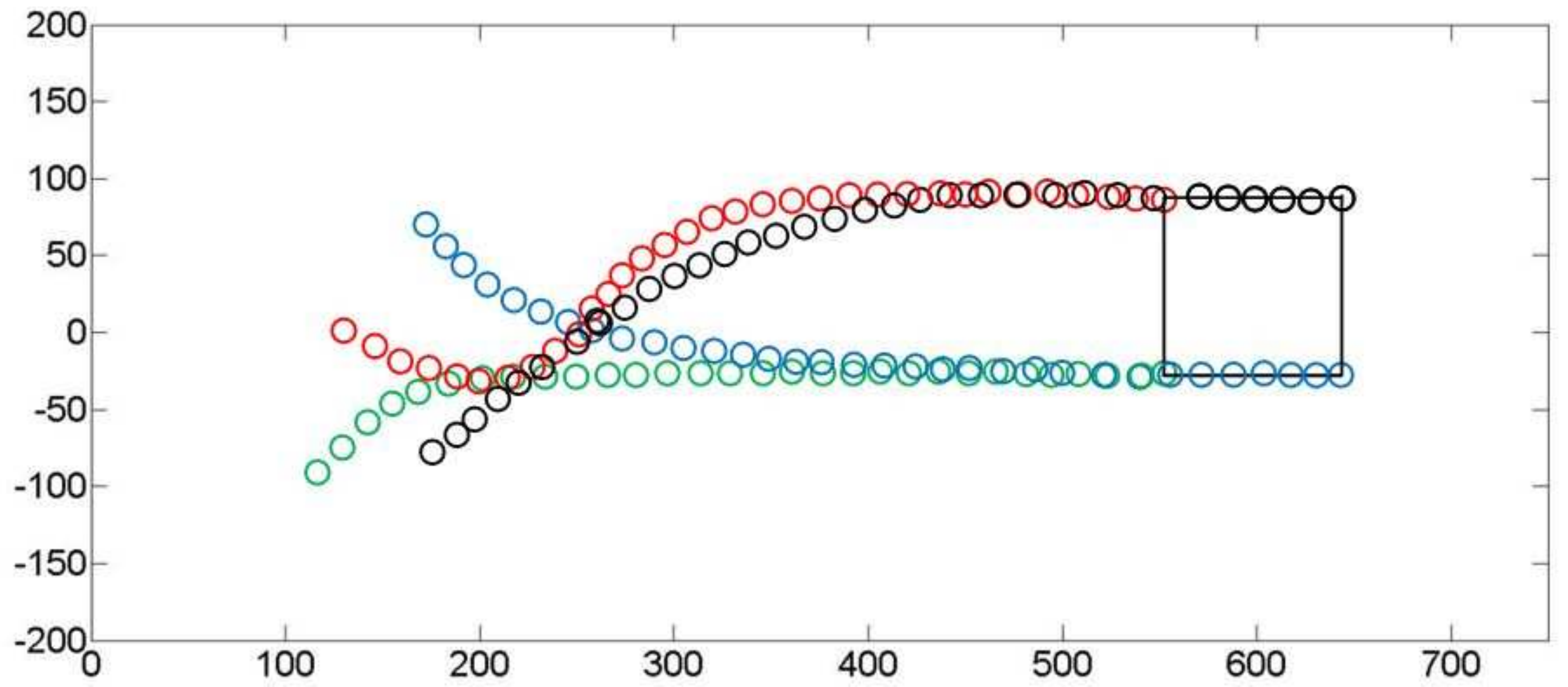}
		\caption{Square formation with anonymous robots, permutation applied}
		\label{c8.sim7}
		\end{figure}

\begin{figure}[!h]
		\centering 
		\includegraphics[width =150mm]{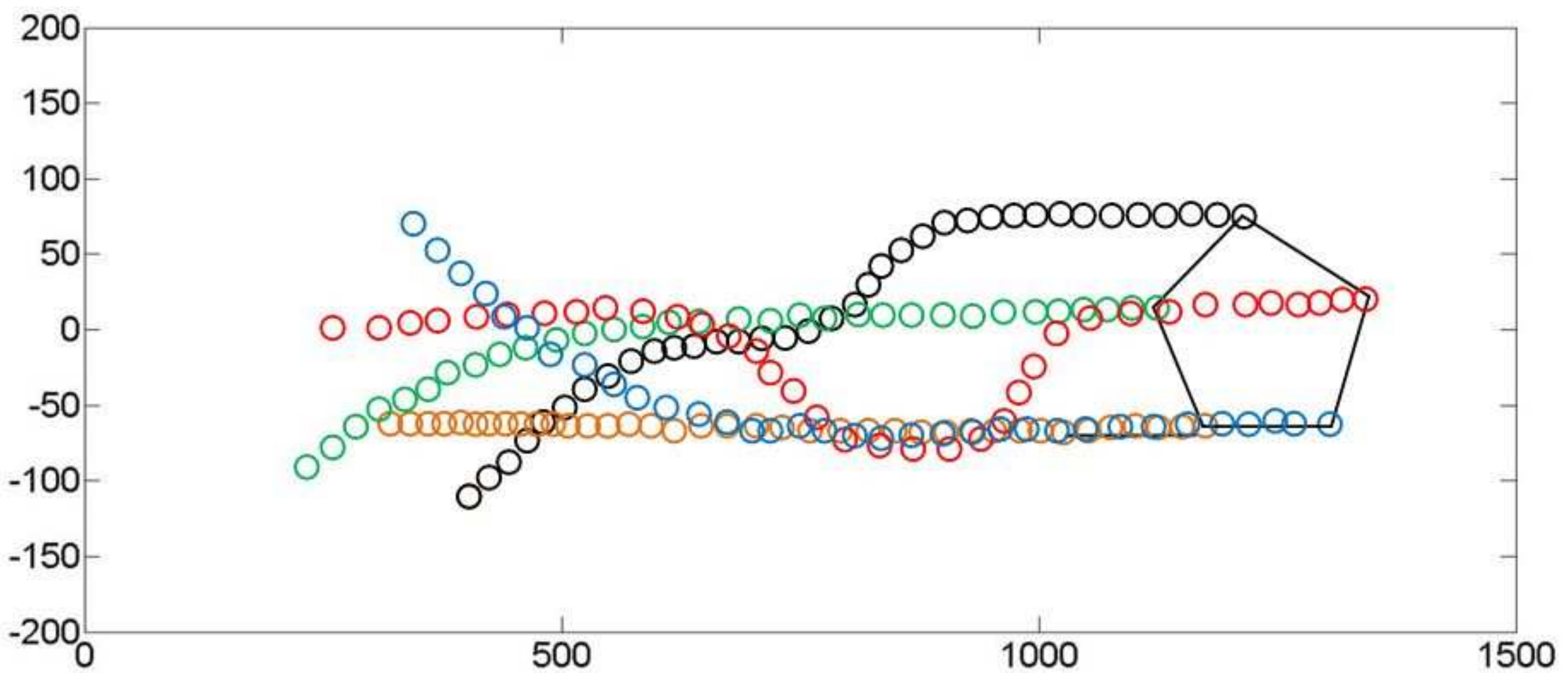}
		\caption{Regular pentagon formation with anonymous robots, permutation applied}
		\label{c8.sim8}
		\end{figure}

\begin{figure}[!h]
		\centering 
		\includegraphics[width =150mm]{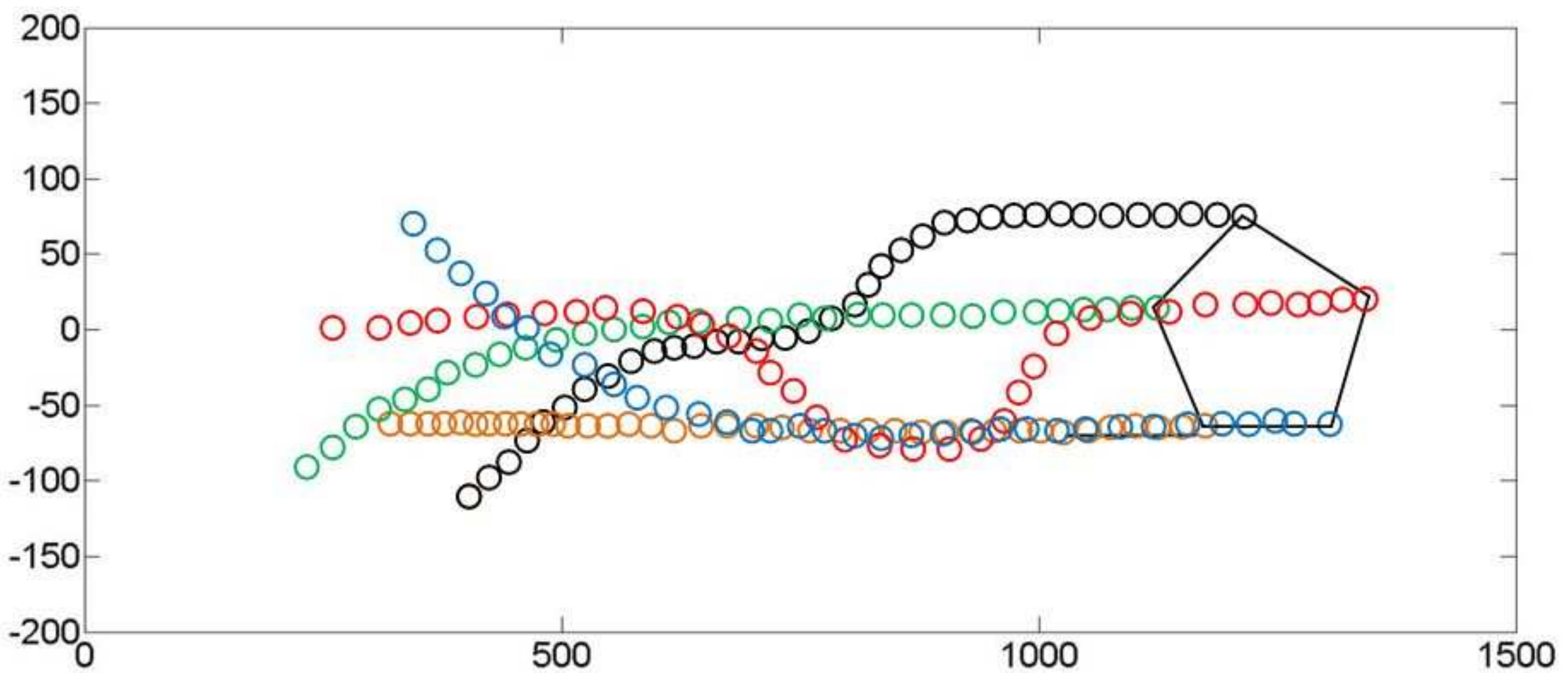}
		\caption{Random shape formation with anonymous robots, permutation applied}
		\label{c8.sim9}
		\end{figure}

				\begin{table}[h]
					\caption {Indices assigned to all the robots}
					
				\begin{tabular}{| l| l| l| l| l|}				
					\hline
					Simulation time & Index of robot 1 & Index of robot 2 & Index of robot 3 & Index of robot 4 \\
					\hline 
					1-12 & 1 & 1 &2 & 2\\
					\hline
					13- 29 & 1 & 3 & 2 & 2 \\
					\hline
					29-53 & 1 & 2 & 3 & 4 \\
					\hline
				\end{tabular}
				\label {perm_index}
				\end{table}  
\par. 
				More complicated formation building with anonymous robots are presented in Fig.~\ref{c8.sim7}, Fig.~\ref{c8.sim8} and Fig.~\ref{c8.sim9}. The randomised algorithm achieve global stability with any initial conditions and various configuration $\mathcal{C}$.
\par


		\section {Experiments with Group of Real Robots}

In this section, we present our experiments with real robots guided by the algorithm proposed in Section \ref{GFB}. In these experiments, the proposed algorithm is implemented on the TurtleBot and Pioneer 3-DX. The encoders are available for both type of the robots, thus all the necessary information for the proposed algorithm (positions, orientations and the velocities) can be easily accessed. We have done experiments with three and four wheeled robots.
\par

\begin{figure}[!h]
\centering
\subfigure[]{\scalebox{0.36}{\includegraphics{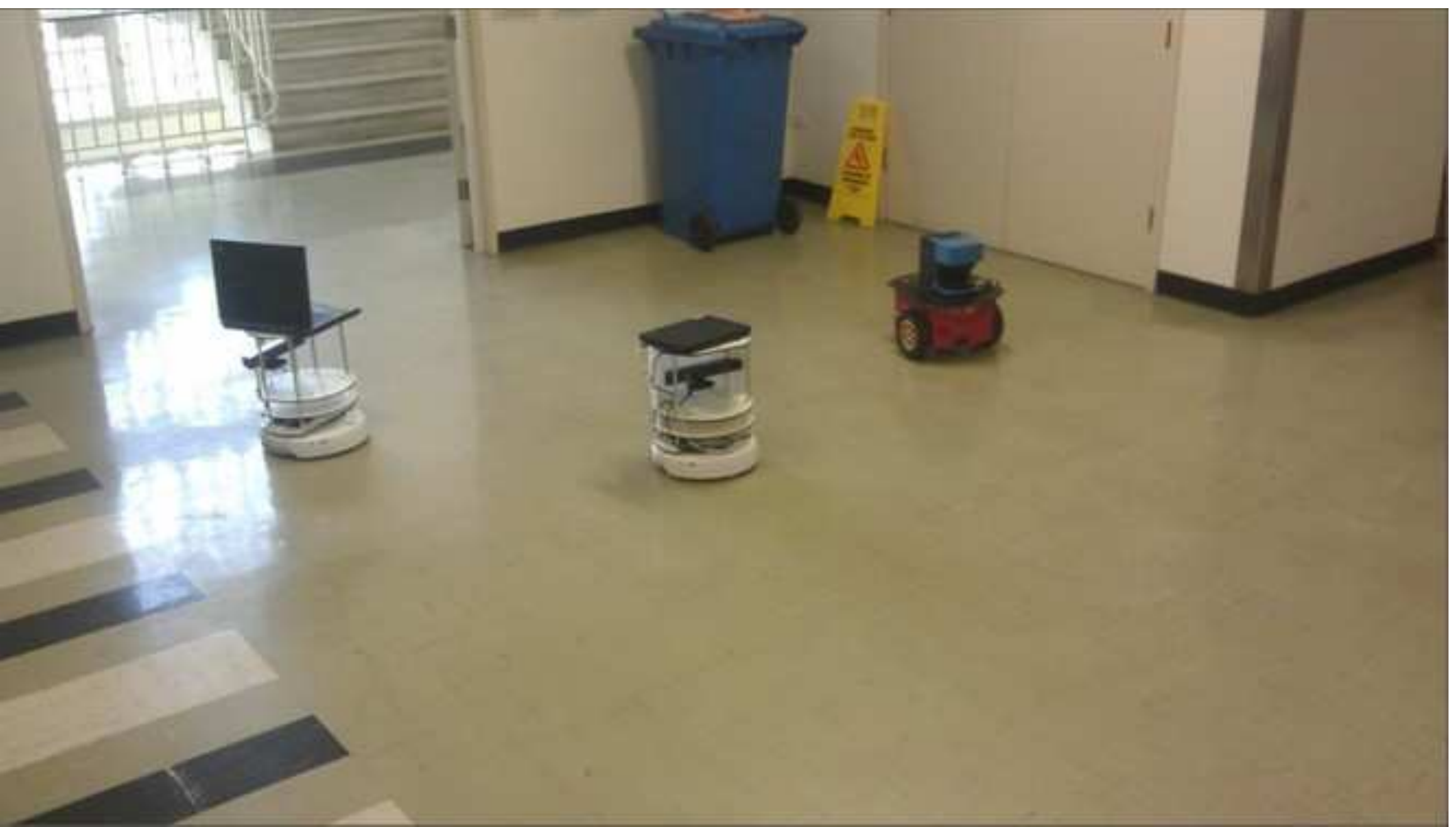}}
\label{c8.exp11}}
\subfigure[]{\scalebox{0.36}{\includegraphics{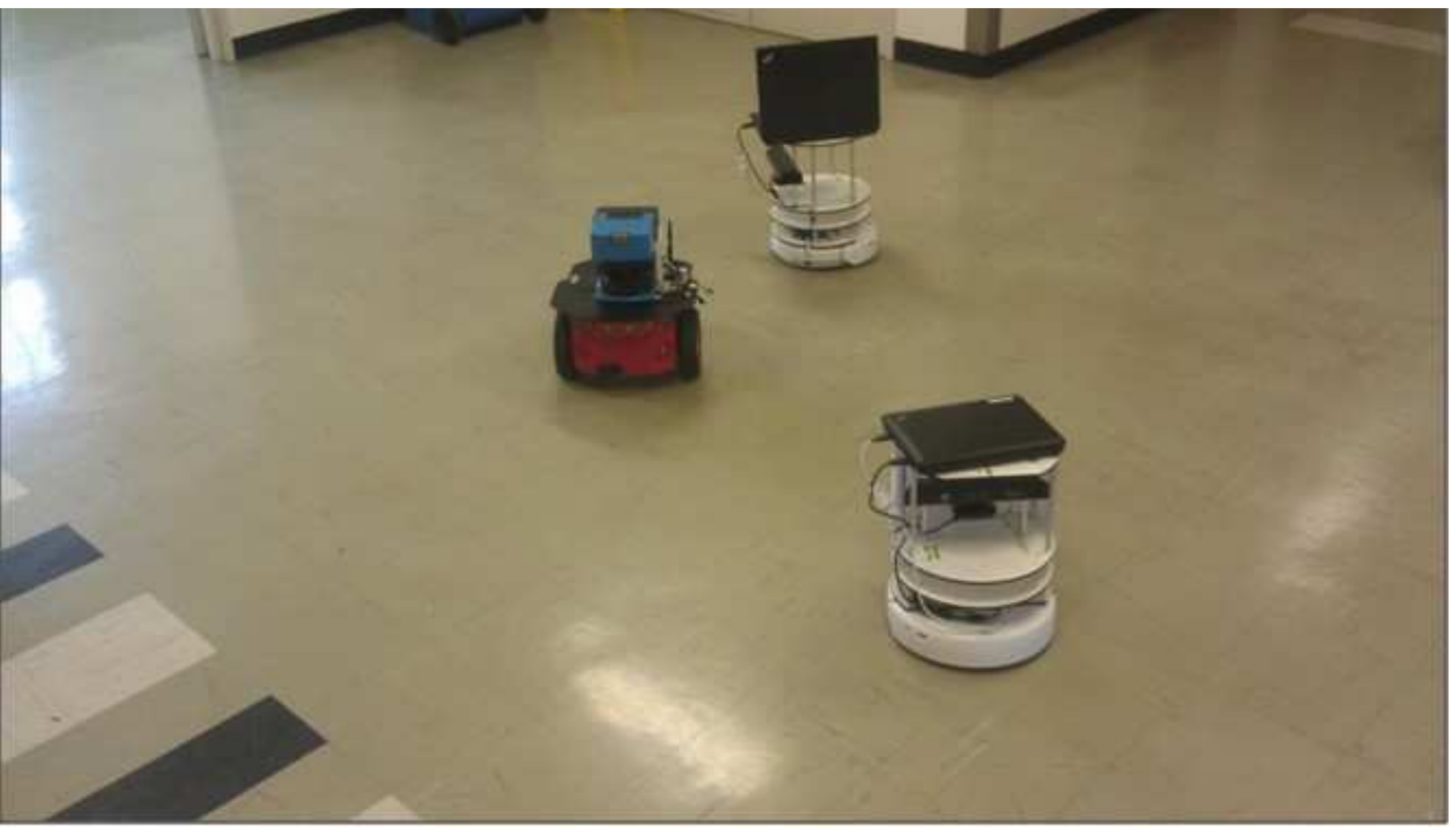}}
\label{c8.exp12}}
\subfigure[]{\scalebox{0.36}{\includegraphics{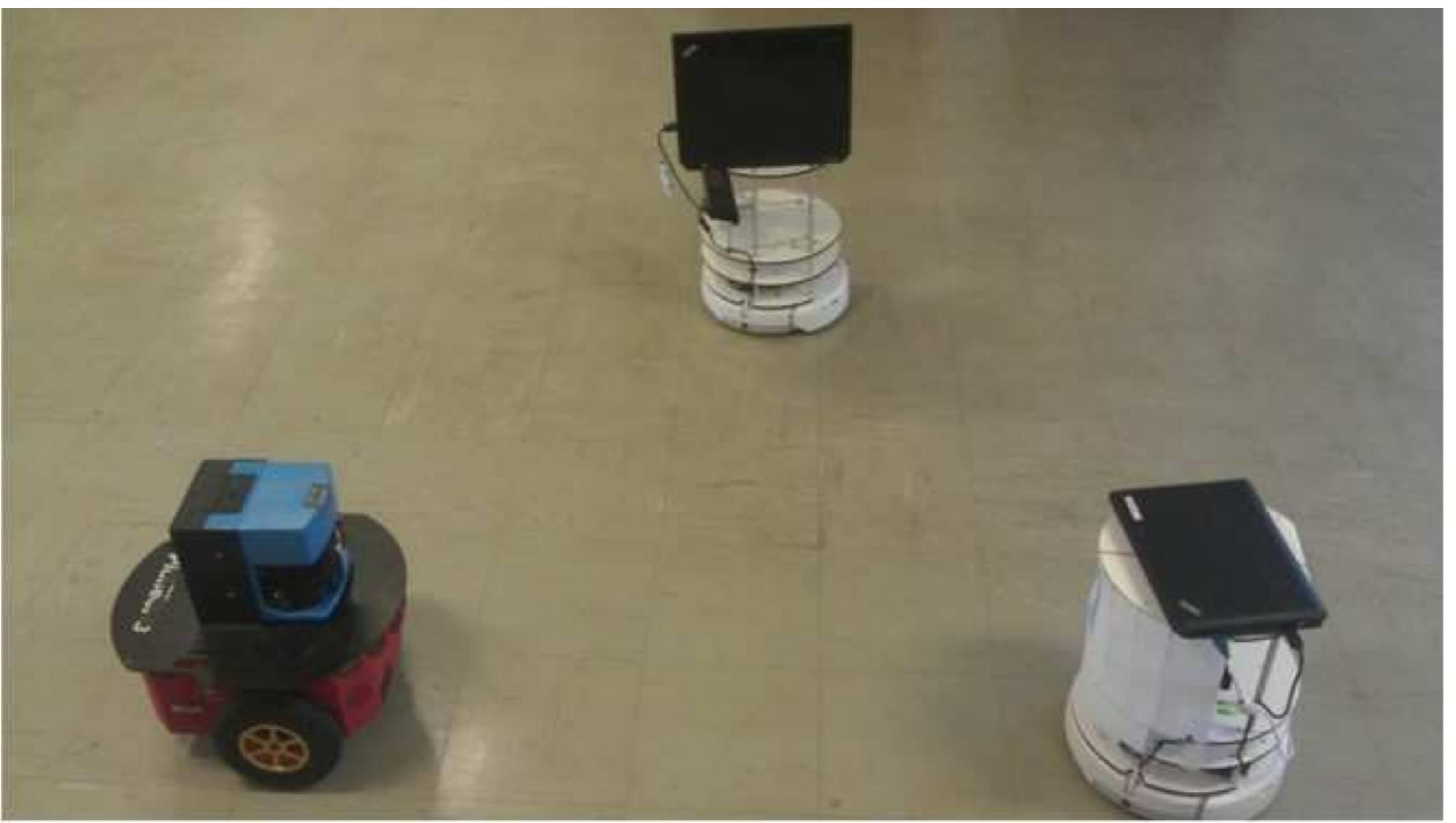}}
\label{c8.exp13}}
\subfigure[]{\scalebox{0.36}{\includegraphics{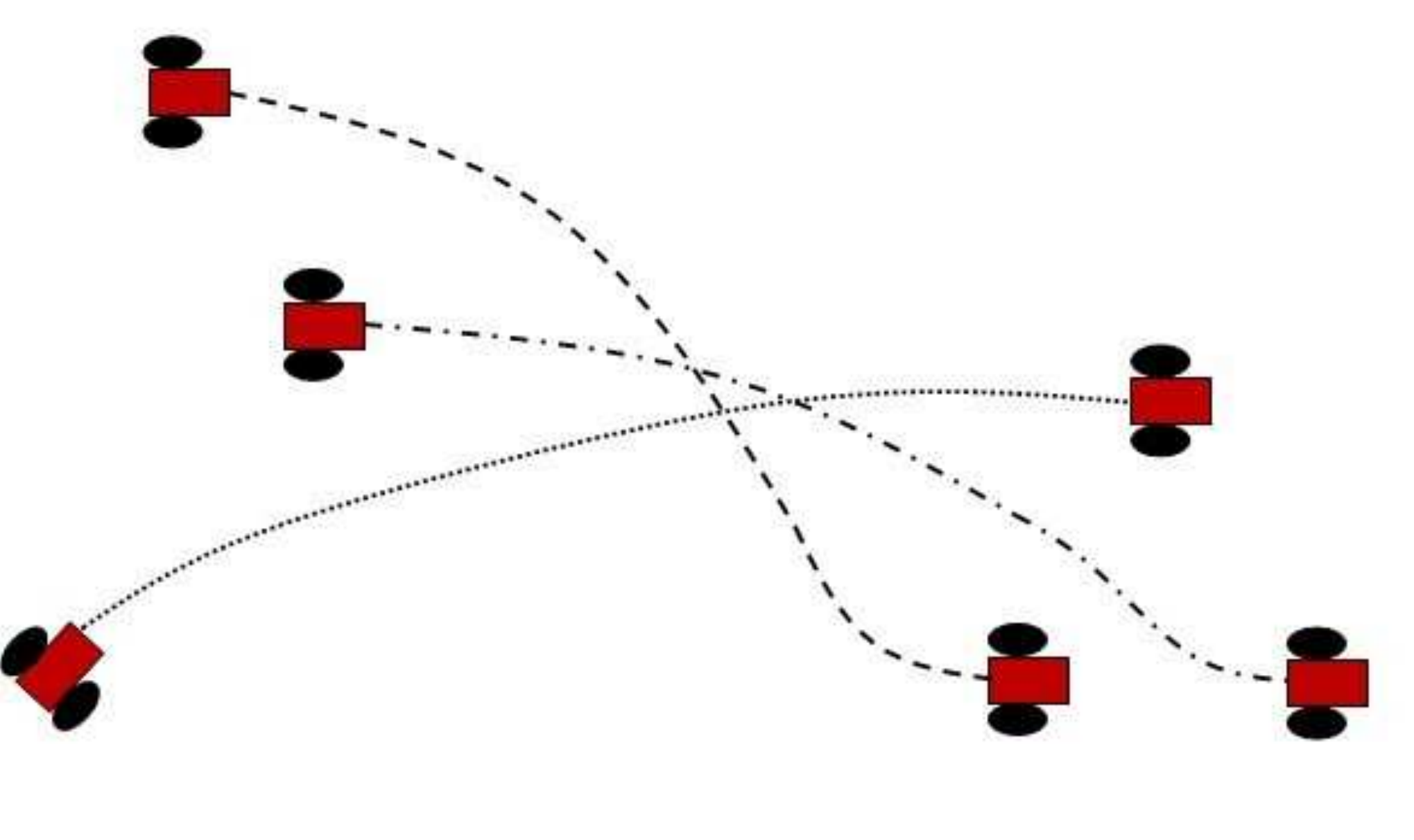}}
\label{c8.exp14}}
\caption{Robots form a regular triangle}
\label{c8.exp1}
\end{figure}

\begin{figure}[!h]
\centering
\subfigure[]{\scalebox{0.36}{\includegraphics{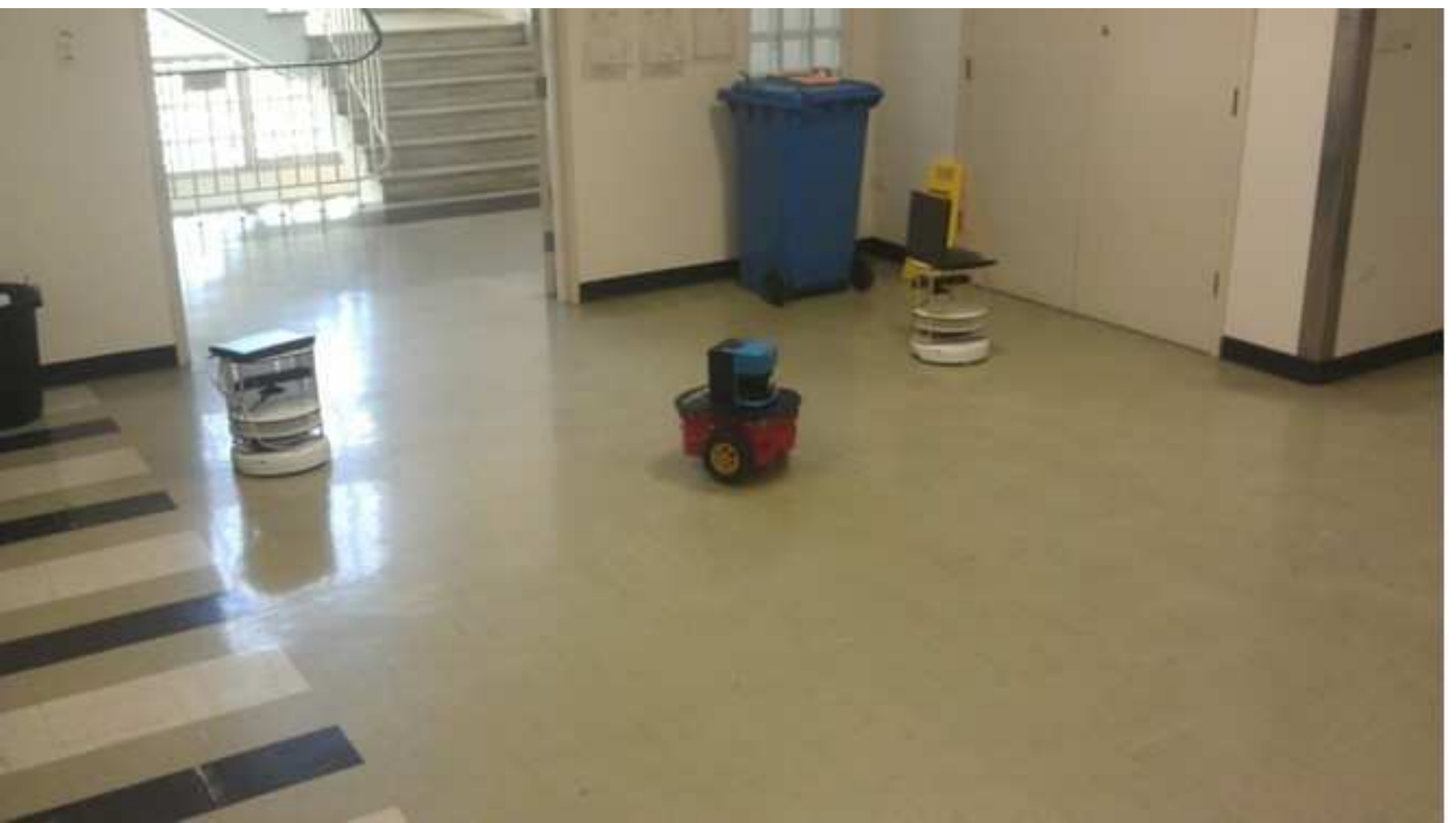}}
\label{c8.exp21}}
\subfigure[]{\scalebox{0.36}{\includegraphics{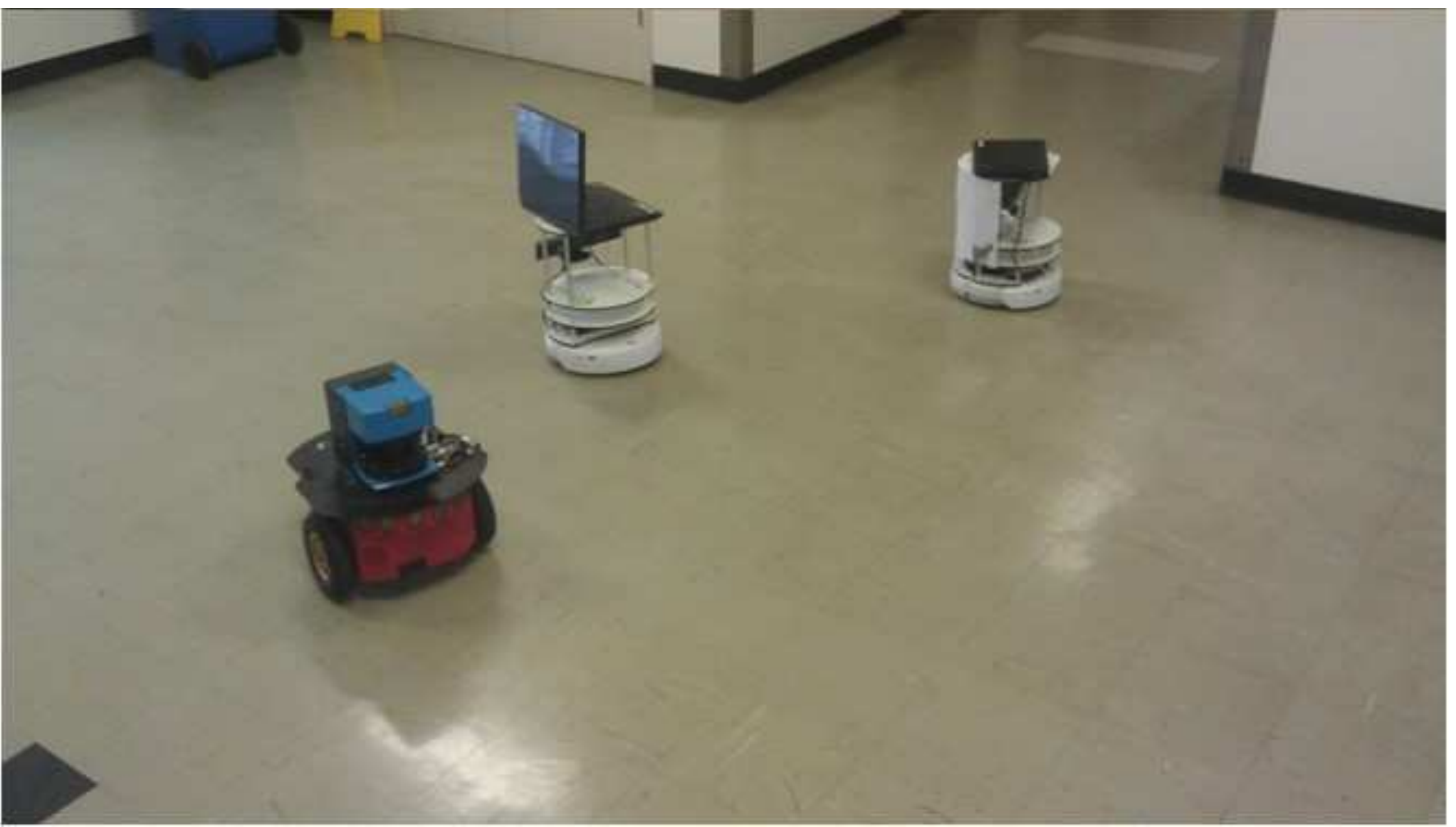}}
\label{c8.exp22}}
\subfigure[]{\scalebox{0.36}{\includegraphics{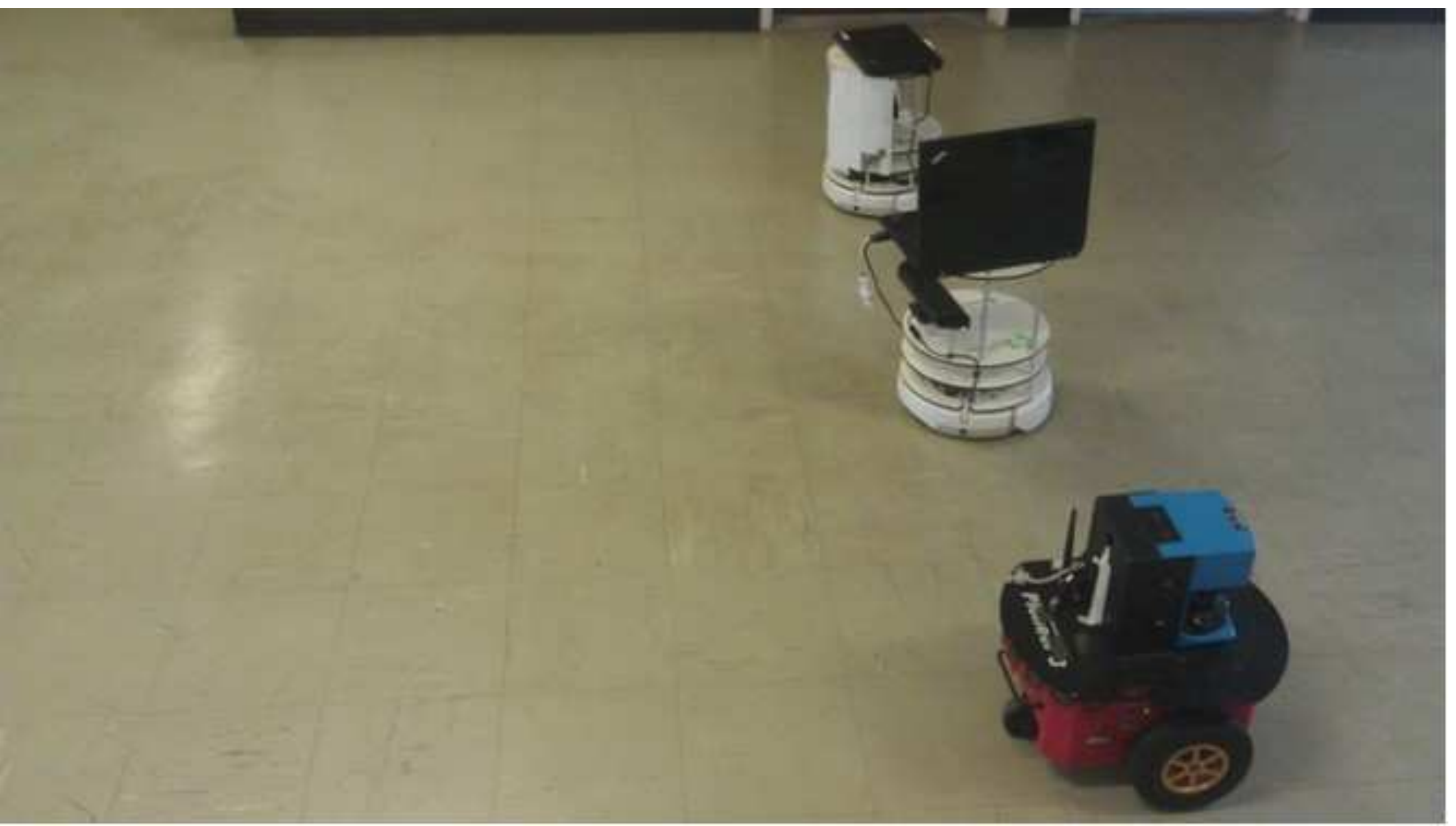}}
\label{c8.exp23}}
\subfigure[]{\scalebox{0.35}{\includegraphics{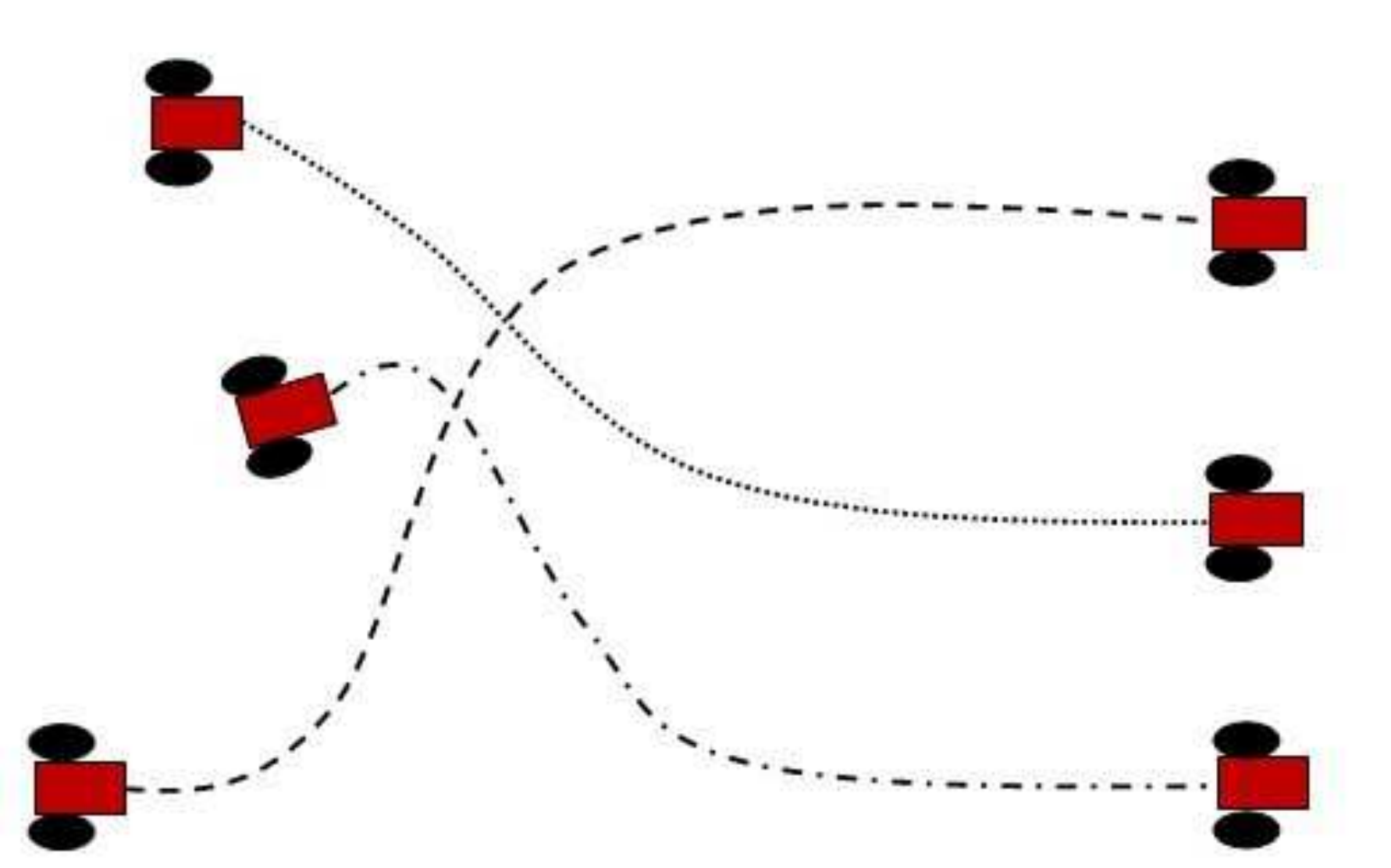}}
\label{c8.exp24}}
\caption{Robots form a straight line}
\label{c8.exp2}
\end{figure}

The experiment snapshots in Fig.~\ref{c8.exp1} and ~\ref{c8.exp2} show the performance of the algorithm with three robots. At the start of each experiment, the robots are placed in random positions with random orientations, see Fig.~\ref{c8.exp11} and ~\ref{c8.exp21}. In Fig.~\ref{c8.exp12} and ~\ref{c8.exp22}, the robots are converging to their final configurations, during this convergence process.
The real time position and orientation information are obtained from encoders which are attached to both wheels of the robots. 
  The final formation configurations for these two experiments, which are regular triangle and straight line, are shown in Fig.~\ref{c8.exp13} and Fig.~\ref{c8.exp23}, respectively. The overall path taken by the robots during these experiments are shown in Fig.~\ref{c8.exp14} and ~\ref{c8.exp24}.

\par

\par

\begin{figure}[!h]
\centering
\subfigure[]{\scalebox{0.36}{\includegraphics{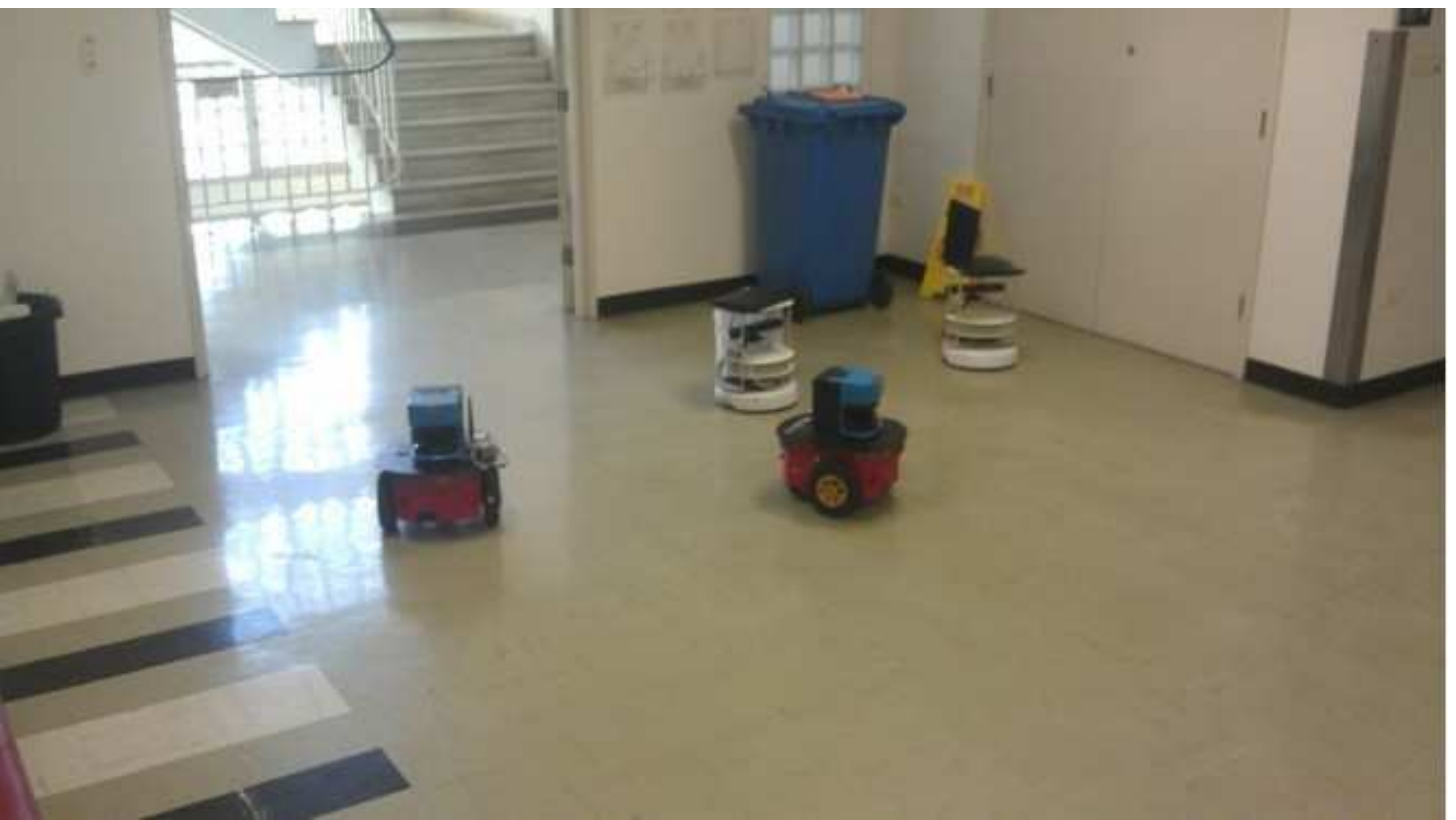}}
\label{c8.exp31}}
\subfigure[]{\scalebox{0.36}{\includegraphics{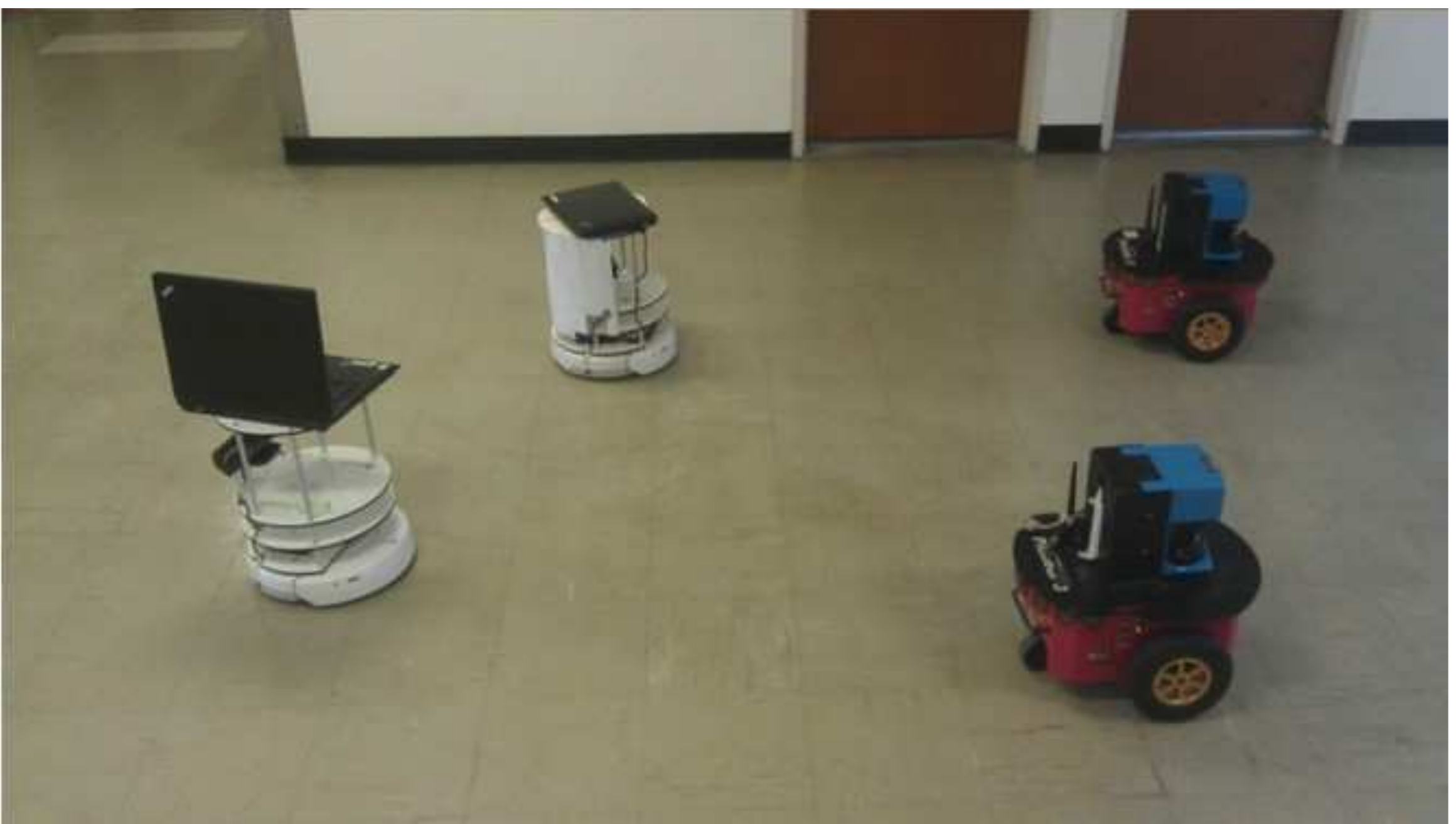}}
\label{c8.exp32}}
\subfigure[]{\scalebox{0.36}{\includegraphics{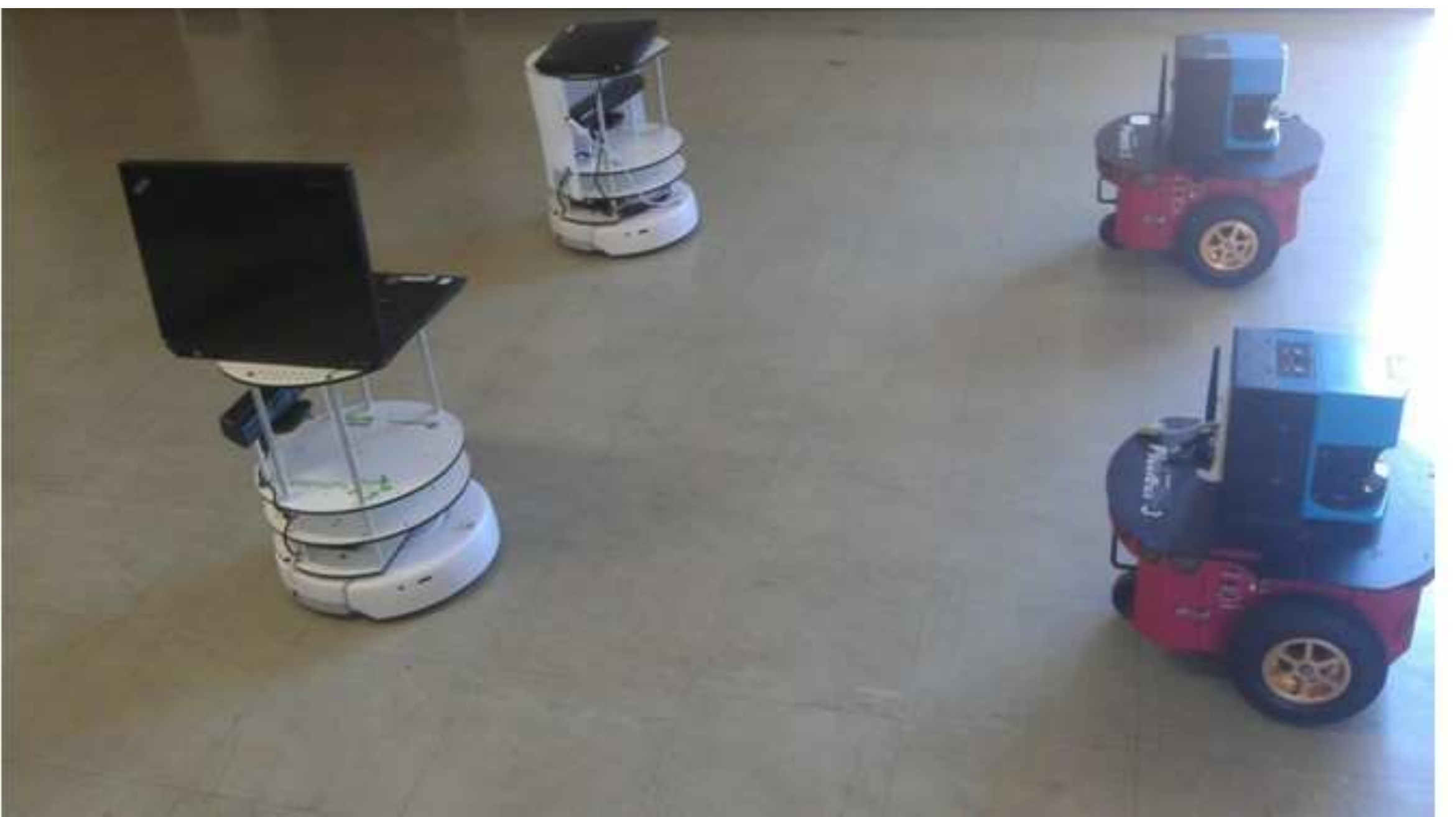}}
\label{c8.exp33}}
\subfigure[]{\scalebox{0.35}{\includegraphics{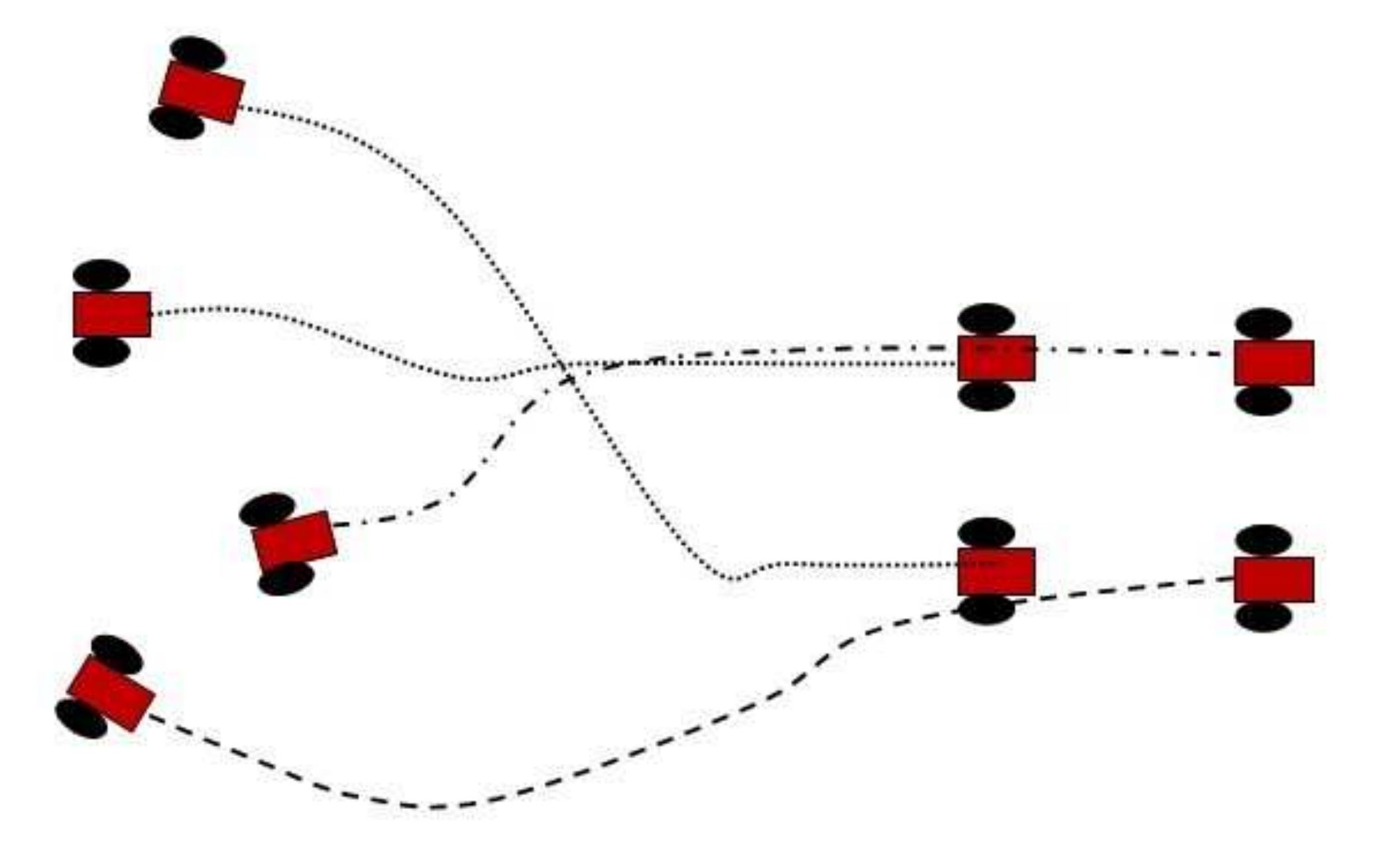}}
\label{c8.exp34}}
\caption{Robots form a rectangle}
\label{c8.exp3}
\end{figure}

\begin{figure}[!h]
\centering
\subfigure[]{\scalebox{0.36}{\includegraphics{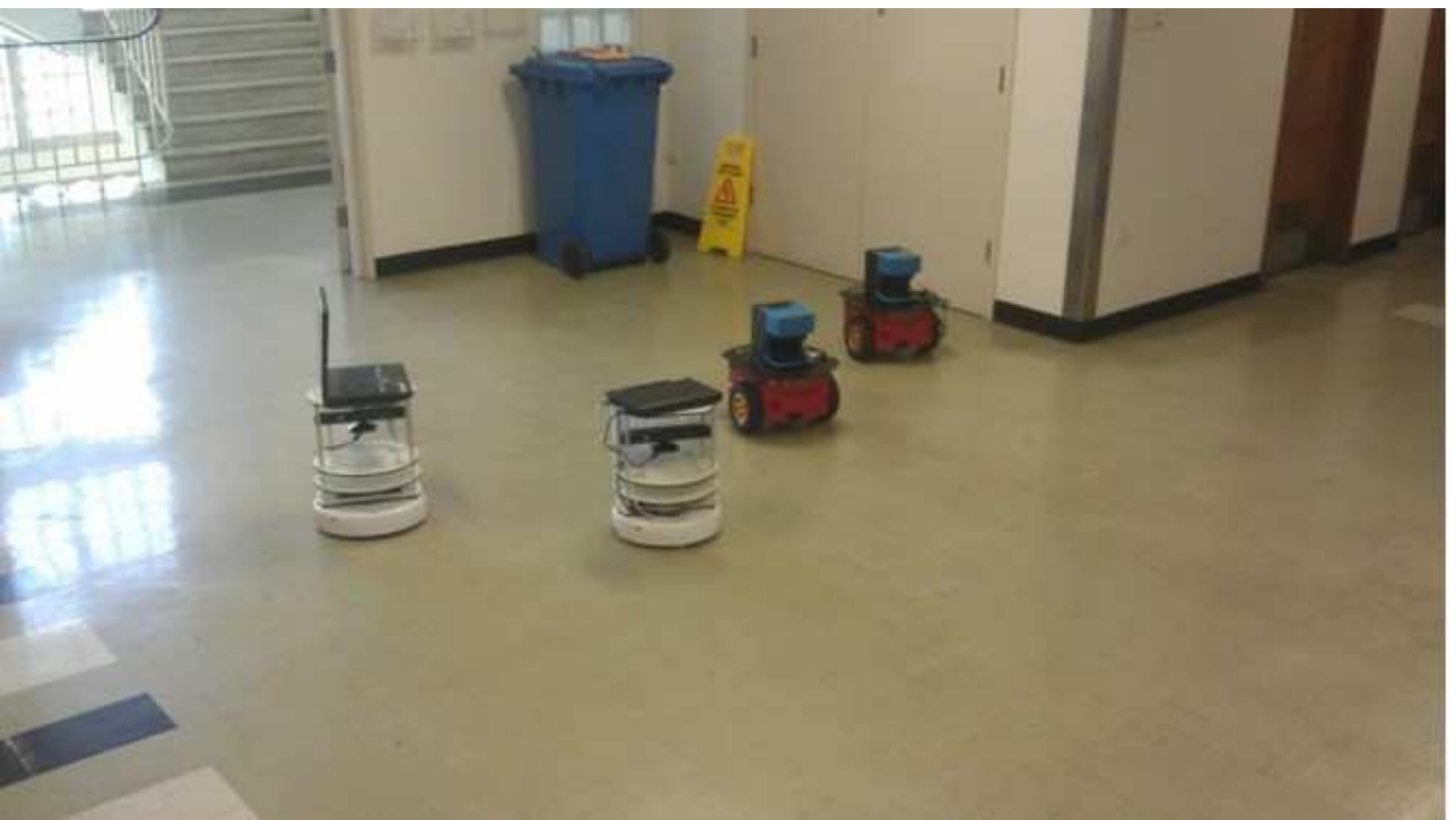}}
\label{c8.exp41}}
\subfigure[]{\scalebox{0.36}{\includegraphics{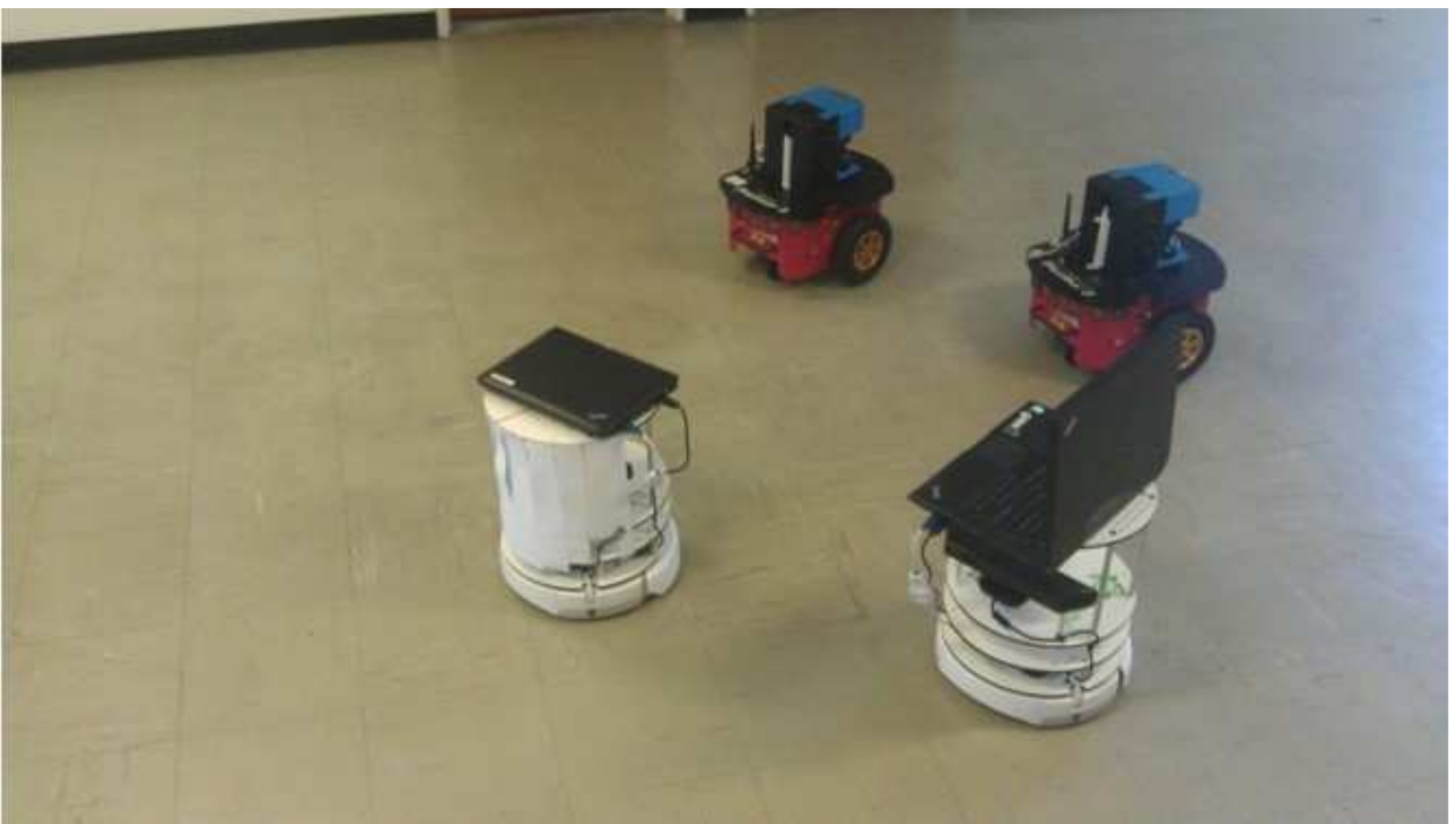}}
\label{c8.exp42}}
\subfigure[]{\scalebox{0.36}{\includegraphics{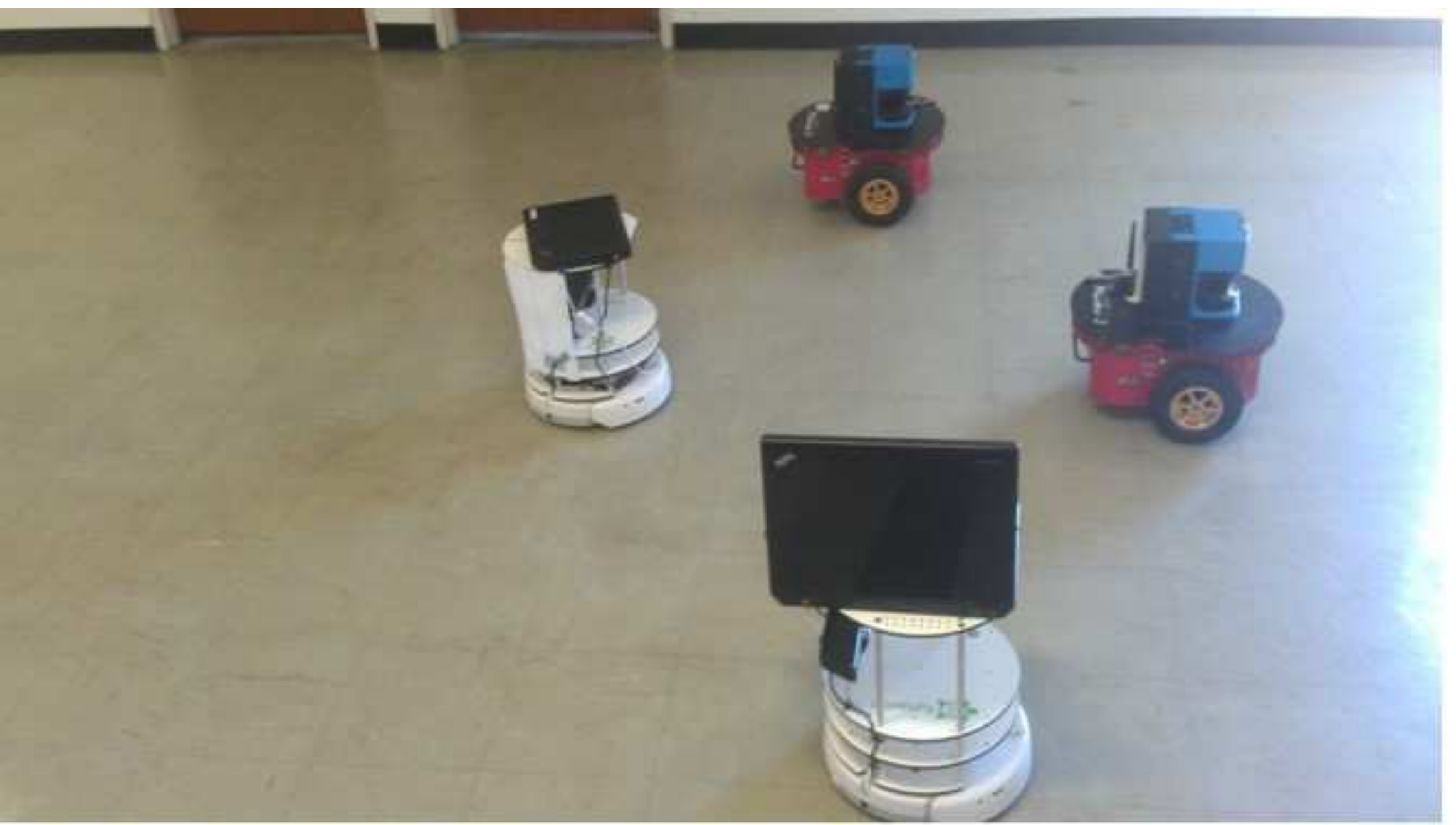}}
\label{c8.exp43}}
\subfigure[]{\scalebox{0.35}{\includegraphics{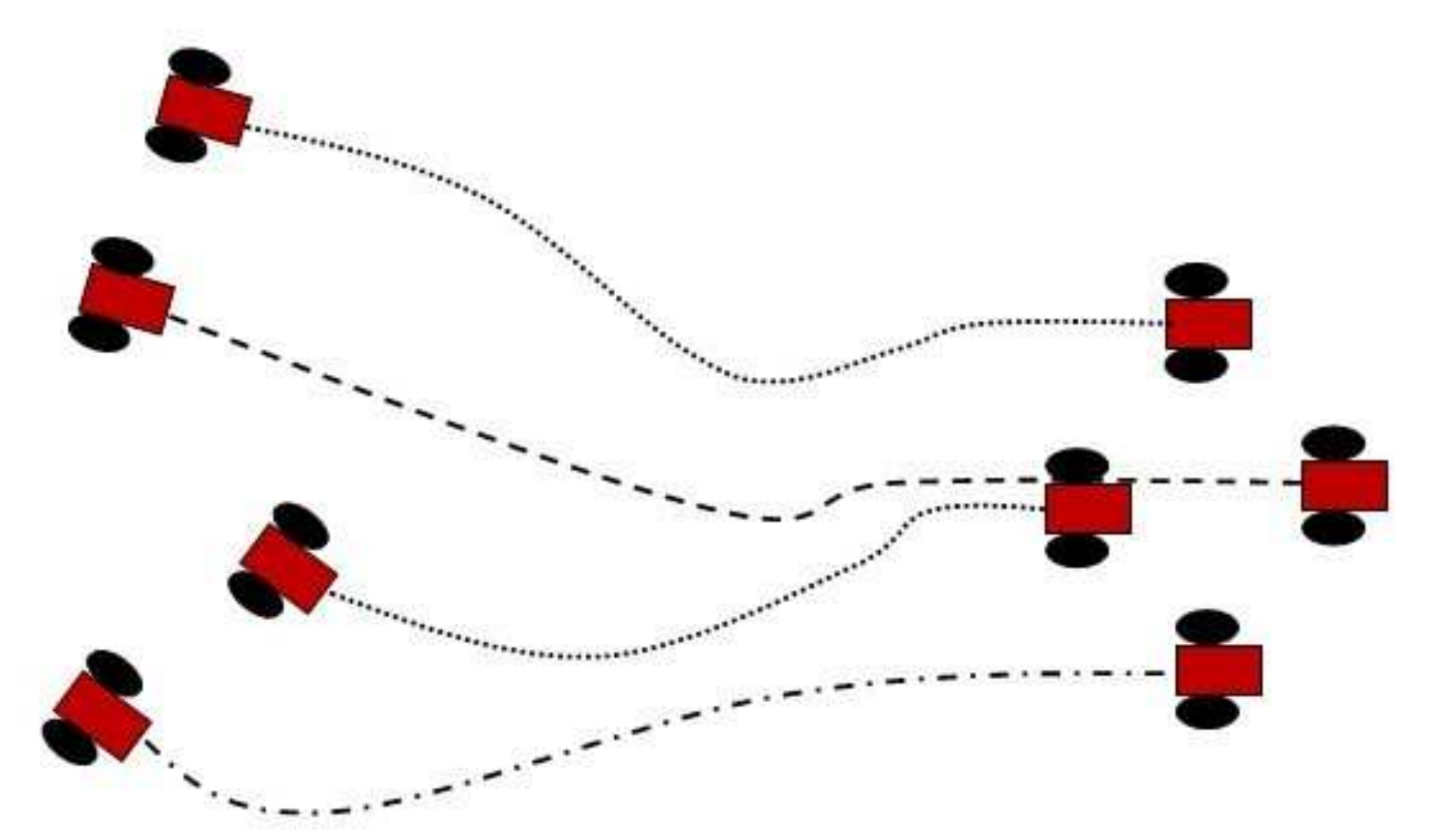}}
\label{c8.exp44}}
\caption{Robots form a kite}
\label{c8.exp4}
\end{figure}

In the next two experiments, one additional robot is added into the group. It is shown in Fig.~\ref{c8.exp3} and Fig.~\ref{c8.exp4} that, the formation of four robots (rectangle in Fig.~\ref{c8.exp3} and kite in Fig.~\ref{c8.exp4}) is achievable  under the guidance of the proposed algorithm. Notice that the figures of this section show  illustrative schematic trajectories of the robots. The convergence time varies in different scenarios, normally,  formations with three mobile robots take approximately 40 sec to converge and formations with four mobile robots take approximately 55 sec to converge.
It can be seen that more complicated formation can be achieved as more robots are added into the group.

\par

\par

		\section{Summary}

		In this chapter, we propose a decentralised formation building algorithm for a group of mobile robots to collectively move with the same speed in a desired geometric formation from any initial positions. Furthermore, we consider the problem of formation with anonymous robots. The robots does not aware of their position in the desired formation in advance and they should reach a consensus on their position. THe multi-robot network considered in this chapter is an example of a network control system as in ~\cite{MA09,MA03,SA06,SA07,MA07,MAA01}. The proposed algorithms are constructive and easy-to-implement. The performance of the proposed algorithms are confirmed by the computer simulations and experiments with a group of real mobile robots.
\chapter {Conclusion} \label{C9}

	The main focus of this report is to design control algorithms for collision free navigation of non-holonomic mobile robots. Unlike most of the exisiting navigation algorithms which focus on robot navigation in static environment, we present three different algorithms for safe navigation of mobile robots in dynamic environments with moving obstacles. We also consider a more realistic and complicated mobile robots of non-holonomic type whose motion is a better representation of a large number of control systems. Each of the proposed navigation algorithm employs different strategy to ensure the safety of the robot.
\par
	The biologically-inspired navigation algorithm (BINA) guides the robot so that there is always a constant avoiding angle between the instantaneous moving direction of the robot and one of boundaries of the enlarged vision cone. The equidistant navigation algorithm drive the robot to a $d_0$-equidistant curve around the obstacle. The navigation algorithm based on integrated environment representation ((NAIER) aims to seek a free path though the crowd of obstacles. These strategy are proven to able to accomplish navigation tasks in various scenarios. The features and merits of the proposed navigation algorithms are confirmed by simulation results and experiment results with real mobile robot Pioneer 3-Dx. 
\par
	These obstacle avoidance strategies determine that one of these proposed navigation algorithms is more efficient in particular scenarios than the others. BINA is the most efficient when the shapes of the obstacles are regular (circle or regular polygons) in both static and dynamic environments. The advantage of ENA is that applicable for a large variety of scenarios, especially when dealing with obstacles of irregular shapes or obstacles with time-varying shapes. Another advantage of ENA is that it is the most stable algorithm in terms of navigation time. NAIER is particularly efficient when the obstacles are extremely cluttered in the environments.
\par
	The successful implementations of BINA and ENA on intelligent wheelchair SAM and autonomous hospital bed Flexbed demonstrate their abilities to accomplish navigation tasks in real life scenarios. The experiment results show that these algorithms is able to solve complicated real life problem such as multiple moving obstacles in a narrow area, obstacles moving with non-linear velocities, obstacles with dynamically changing radii.  They also indicates that these proposed navigation algorithms are applicable in many real control systems.
\par 
	Finally, we investigate the problem of formation building for a group of mobile robots. A constructive and easy-to-implement decentralised controller formation building algorithm is proposed. The proposed controller allows the group of robots to not only move in the same direction but also in the desired geometric formation. This formation building is achieved without a leader robot which means that single point failure does not cause the whole system to fail. This design of formation building strategy enhances the robustness of the whole system. Furthermore, we consider a group of anonymous robots, each robot in this group are not aware of his position in the desired configuration and the robots have to reach a consensus on their positions. A randomised decentralised navigation algorithm is proposed to achieve formation building with anonymous robots. This algorithm achieves the convergence of final formation with probability $1$. We present simulation results and experiments results with a group of robots to demonstrate the performance of the proposed algorithms.

	\section{Future Research Directions}

		The scope of this report can be possibly further extended by the following research directions.

		\begin{itemize}

			\item The problem of navigating mobile robots with complicated models can be considered. In such cases, it is not sufficient to solve the navigation problem by only considering the kinematic models of the mobile robots, the dynamics of the robots should also be studied. We will combine the proposed navigation ideas with advanced methods of modern robust control. see e.g. \cite{PIR00,SAVP95,UVA00}.
			\item It is interesting to investigate the performance of the proposed navigation algorithms against obstacles with even more complicated motions, such as those obstacles whose motions are not restricted by non-holonomic constraints.
			\item The proposed algorithm NAIER had not yet been implemented to any real control system. It is very suitable for systems such as wheelchairs which often find themselves involved in a crowded environments.
			\item The proposed navigation algorithms has been only implemented on ground robots/systems. It is possible to extend their applications to other underwater/aerial (e.g.\cite{MHA12}) systems with reasonable modifications of the algorithms.
			\item The proposed formation building strategies can be combined with the obstacle avoidance strategies so that the robots will not collide into each other during converging process.
		\end{itemize}



\bibliography{bib/References,bib/chaorefs,bib/wheelchair,bib/wheelchair2}

\end{document}